# Toward Global Large Language Models in Medicine

Rui Yang[1,2,3,4], Huitao Li[1,2], Weihao Xuan[5†], Heli Qi[6], Xin Li[7], Kunyu Yu[1,2], Yingjian Chen[8], Rongrong Wang[3], Jacques Behmoaras[1,2], Tianxi Cai[9,10], Bibhas Chakraborty[1,2,11,12,13], Qingyu Chen[14], Lionel Tim-Ee Cheng[15,16,17], Marie-Louise Damwanza[18], Chido Dzinotyiwei[18], Aosong Feng[19], Chuan Hong[13,20], Yusuke Iwasawa[8], Yuhe Ke[1,21,22], Linah Kitala[18], Taehoon Ko[23,24,25], Jisan Lee[26,27], Irene Li[8], Jonathan Chong Kai Liew[1,28], Hongfang Liu[29], Lian Leng Low[30,31,32], Edison Marrese-Taylor[8], Yutaka Matsuo[8], Isheanesu Misi[18], Yilin Ning[1,2], Jasmine Chiat Ling Ong[1,33], Marcus Eng Hock Ong[34,35], Enrico Petretto[1,2], Hossein Rouhizadeh[36], Abiram Sandralegar[36,37], Oren Schreier[36], Iain Beehuat Tan[38,39,40,41], Patrick Tan[38,41,42,43], Daniel Shu Wei Ting[44,45,46,47], Junjue Wang[5], Chunhua Weng[48], Matthew Yu Heng Wong[49], Fang Wu[50], Yunze Xiao[51], Xuhai Xu[48], Qingcheng Zeng[52], Zhuo Zheng[50], Yifan Peng[3,4,53†], Douglas Teodoro[36†], Nan Liu[1,2,13,34,54†]

*[1] Duke-NUS AI + Medical Sciences Initiative, Duke-NUS Medical School, Singapore, Singapore*

*[2] Centre for Biomedical Data Science, Duke-NUS Medical School, Singapore, Singapore*

*[3] Department of Population Health Sciences, Weill Cornell Medicine, New York, NY, USA*

*[4] Systems Engineering, Cornell University, Ithaca, NY, USA*

*[5] Graduate School of Frontier Sciences, The University of Tokyo, Tokyo, Japan*

*[6] Faculty of Science and Engineering, Waseda University, Tokyo, Japan*

*[7] McWilliams School of Biomedical Informatics, University of Texas Health Science Center at Houston, Houston, TX, USA*

*[8] Graduate School of Engineering, The University of Tokyo, Tokyo, Japan*

*[9] Department of Biostatistics, Harvard University, Boston, MA, USA*

*[10] Department of Biomedical Informatics, Harvard Medical School, Boston, MA, USA*

*[11] Health Services Research & Population Health, Duke-NUS Medical School, Singapore, Singapore*

*[12] Department of Statistics and Data Science, National University of Singapore, Singapore, Singapore*

*[13] Department of Biostatistics and Bioinformatics, Duke University, Durham, NC, USA*

*[14] Department of Biomedical Informatics and Data Science, Yale School of Medicine, Yale University, New Haven, CT, USA*
*[15] Radiological Sciences Academic Clinical Programme, SingHealth Duke-NUS, Singapore, Singapore*
*[16] Department of Cardiothoracic and Abdominal Radiology, Singapore General Hospital, Singapore, Singapore*
*[17] SingHealth Community Hospitals, Singapore Health Services, Singapore, Singapore*
*[18] Vambo AI, Johannesburg, South Africa*
*[19] Department of Computer Science, Yale University, New Haven, CT, USA*
*[20] Duke Clinical Research Institute, Durham, NC, USA*
*[21] Department of Anesthesiology, Singapore General Hospital, Singapore, Singapore*
*[22] Data Science and Artificial Intelligence Lab, Singapore General Hospital, Singapore, Singapore*
*[23] Department of Medical Informatics, College of Medicine, The Catholic University of Korea, Seoul, Republic of Korea*
*[24] Department of Medical Sciences, College of Medicine, The Catholic University of Korea, Seoul, Republic of Korea*
*[25] CMC Institute for Basic Medical Science, The Catholic Medical Center of the Catholic University of Korea, Seoul, Republic of Korea*
*[26] Department of Nursing, Wonju Campus, Kangwon National University, Gangwon State, Republic of Korea*
*[27] Clinical Informatics Center, UT Southwestern Medical Center, Dallas, TX, USA*
*[28] Perelman School of Medicine, University of Pennsylvania, Philadelphia, PA, USA*
*[29] Department of Quantitative and Systems Health Sciences, Dell Medical School, The University of Texas at Austin, Austin, USA*
*[30] Family Medicine Academic Clinical Programme, Duke-NUS Medical School, Singapore, Singapore*
*[31] Division of Population Health and Integrated Care, Singapore General Hospital, Singapore, Singapore*
*[32] Centre for Population Health Research and Implementation, Singapore Health Services, Singapore, Singapore*
*[33] Division of Pharmacy, Singapore General Hospital, Singapore, Singapore*

*[34] Pre-hospital & Emergency Research Centre, Health Services Research & Population Health, Duke-NUS Medical School, Singapore, Singapore*
*[35] Department of Emergency Medicine, Singapore General Hospital, Singapore, Singapore*
*[36] Department of Radiology and Medical Informatics, University of Geneva, Geneva, Switzerland*
*[37] Division of Neurosurgery, Geneva University Hospitals, Geneva, Switzerland*
*[38] Cancer and Stem Cell Biology Programme, Duke-NUS Medical School, Singapore, Singapore*
*[39] Division of Medical Oncology, National Cancer Centre Singapore, Singapore, Singapore*
*[40] Office of Deputy Group Chief Medical Informatics Officer (Research), Singapore Health Services, Singapore, Singapore*
*[41] Genome Institute of Singapore, Agency for Science, Technology and Research, Singapore, Singapore*
*[42] SingHealth Duke-NUS Institute of Precision Medicine, Singapore, Singapore*
*[43] Precision Health Research, Singapore, Singapore*
*[44] Ophthalmology and Visual Sciences Academic Clinical Programme, Duke-NUS Medical School, Singapore, Singapore*
*[45] Singapore National Eye Centre, Singapore Eye Research Institute, Singapore, Singapore*
*[46] Artificial Intelligence Office, Singapore Health Services, Singapore, Singapore*
*[47] Byers Eye Institute, Stanford University, Stanford, CA, USA*
*[48] Department of Biomedical Informatics, Columbia University, New York, NY, USA*
*[49] School of Clinical Medicine, University of Cambridge, Cambridge, UK*
*[50] Department of Computer Science, Stanford University, Stanford, CA, USA*
*[51] Language Technologies Institute, Carnegie Mellon University, Pittsburgh, PA, USA*
*[52] Department of Linguistics, Northwestern University, Evanston, IL, USA*
*[53] Institute of Artificial Intelligence for Digital Health, Weill Cornell Medicine, New York, NY, USA*
*[54] NUS Artificial Intelligence Institute, National University of Singapore, Singapore, Singapore*

This work was jointly supervised by Yifan Peng, Douglas Teodoro, and Nan Liu.

***Corresponding Authors:***

*Weihao Xuan, Graduate School of Frontier Sciences, The University of Tokyo, Kiban-to 406, 5-1-5 Kashiwanoha, Kashiwa, Chiba 277-8561, Japan*

*Email: xuan@ms.k.u-tokyo.ac.jp*

*Yifan Peng, Department of Population Health Sciences, Weill Cornell Medicine, 575 Lex ave, New York, 10022, USA*

*Email: yip4002@med.cornell.edu*

*Douglas Teodoro, Department of Radiology and Medical Informatics, University of Geneva, Campus Biotech, G6-N3, Chemin des Mines 9, CH-1202 Geneva, Switzerland*

*Email: douglas.teodoro@unige.ch*

*Nan Liu, Centre for Biomedical Data Science, Duke-NUS Medical School, 8 College Road, Singapore 169857, Singapore*

*Email: liu.nan@duke-nus.edu.sg*

## Abstract

Despite continuous advances in medical technology, the global distribution of health care resources remains uneven. The development of large language models (LLMs) has transformed the landscape of medicine and holds promise for improving health care quality and expanding access to medical information globally. However, existing LLMs are primarily trained on high-resource languages, limiting their applicability in global medical scenarios. To address this gap, we constructed GlobMed, a large multilingual medical dataset containing more than 800,000 entries spanning 20 languages, including seven low-resource languages. Building on this, we established GlobMed-Bench, which systematically assesses 56 state-of-the-art proprietary and open-weight LLMs across 12 of the 20 languages represented in GlobMed, comprising over 20,000 independent experiments and more than 32 million model responses. This large-scale evaluation revealed significant performance disparities across languages, particularly for low-resource languages. Additionally, we introduced GlobMed-LLMs, a suite of multilingual medical LLMs trained across the 12 benchmark languages, with parameters ranging from 1.7B to 8B. GlobMed-LLMs achieved an average performance improvement of nearly 40% relative to baseline LLMs, with a maximum performance improvement of more than threefold in low-resource languages. These resources provide an important foundation for advancing the equitable development and application of LLMs globally, enabling broader language communities to benefit from technological advances.

## Introduction

Despite continuous advances in medical technology, important innovations have not translated into a more equitable global distribution of health care resources, particularly in low-resource regions[1,2]. Currently, nearly half of the global population still lacks access to essential health services[3]. This disparity is largely driven by fragile health systems, shortages of trained health care providers, and inadequate infrastructure, all of which continue to constrain progress in population health[3]. In addition, insufficient access to reliable medical information further limits public health literacy and self-care capability, contributing to delayed diagnoses and preventable health burdens[4].

The rapid development of large language models (LLMs) has begun to reshape the landscape of medicine, with promising applications in clinical consultation, disease diagnosis, health management, and medical education[5–7]. Early evidence suggests that LLMs have the potential to alleviate the workload of clinicians while simultaneously improving the quality and consistency of patient care[8–11]. Beyond frontline clinical settings, LLMs are increasingly used to support scientific research, including manuscript evaluation[12], literature screening[13], quality appraisal[14], and knowledge synthesis[15]. As these models become more capable and widely accessible, they offer a compelling opportunity to strengthen health care delivery and expand access to medical information globally[16,17].

However, the global applicability of LLMs in medicine is limited by several key challenges[18,19]. Most existing LLMs are trained predominantly on data from high-resource languages (i.e., languages with abundant linguistic resources and technical support[20]). For instance, over 92% of GPT-3's pretraining corpus is derived from English sources, while low-resource languages (i.e., languages with scarce linguistic resources and limited technical support[20]) remain severely underrepresented[21]. This imbalance has led to substantial performance disparities across languages, undermining the reliability and generalizability of LLMs in global medical contexts[22,23]. Moreover, current medical benchmarks are limited in both scale and linguistic diversity, making it difficult to systematically evaluate LLM performance across a wide range of real-world use cases[24]. These limitations risk exacerbating global health disparities, particularly affecting communities that rely on low-resource languages, precisely those who stand to benefit most from the equitable development of LLM applications[18,19]. Addressing these challenges is essential to ensure that advances in LLM technology can effectively support health care delivery and access to medical information in diverse global settings.

Therefore, developing well-curated multilingual medical datasets and comprehensive evaluation benchmarks is crucial[24]. Such resources would facilitate the systematic evaluation of LLM performance across languages and uncover gaps in generalizability. Equally important is the inclusion of languages that are currently underrepresented, ensuring that medical LLM innovations benefit a broader range of language communities. Building on these foundational resources, specialized medical LLMs optimized for multilingual contexts, particularly for lower-resource languages, are needed to extend the reach of technological advances to health communities that have historically been excluded.

To advance the global development of medical LLMs, our study makes three core contributions (**Fig. 1**): (1) **GlobMed**. We constructed GlobMed, a large multilingual medical dataset, spanning 20 languages, including seven low-resource languages, that collectively represent over seven billion people (~87% of the global population)[25]. GlobMed contains more than 800,000 entries across three core tasks: natural language inference (NLI), long-form question answering (QA), and multiple-choice question answering (MCQA). Additionally, GlobMed was evaluated by medical experts proficient in the corresponding target languages to ensure linguistic accuracy and clinical validity. (2) **GlobMed-Bench**. We established GlobMed-Bench, a comprehensive evaluation benchmark that assesses 56 state-of-the-art proprietary and open-weight LLMs across 12 of the 20 languages represented in GlobMed: Chinese, English, French, German, Japanese, Korean, Portuguese, Spanish, Swahili, Wolof, Yoruba, and Zulu. GlobMed-Bench comprises over 20,000 independent experiments and more than 32 million model responses. This benchmark provides the most extensive and systematic multilingual medical evaluation of LLMs to date, revealing significant performance disparities across languages, particularly for low-resource languages. (3) **GlobMed-LLMs**. We introduced GlobMed-LLMs, a suite of multilingual medical LLMs ranging from 1.7B to 8B parameters, trained on GlobMed to improve medical performance across languages. Across six multilingual medical benchmarks and the 12 languages evaluated in GlobMed-Bench, GlobMed-LLMs achieved an average performance improvement of nearly 40% relative to baseline LLMs, with a maximum performance improvement of more than threefold in low-resource languages.

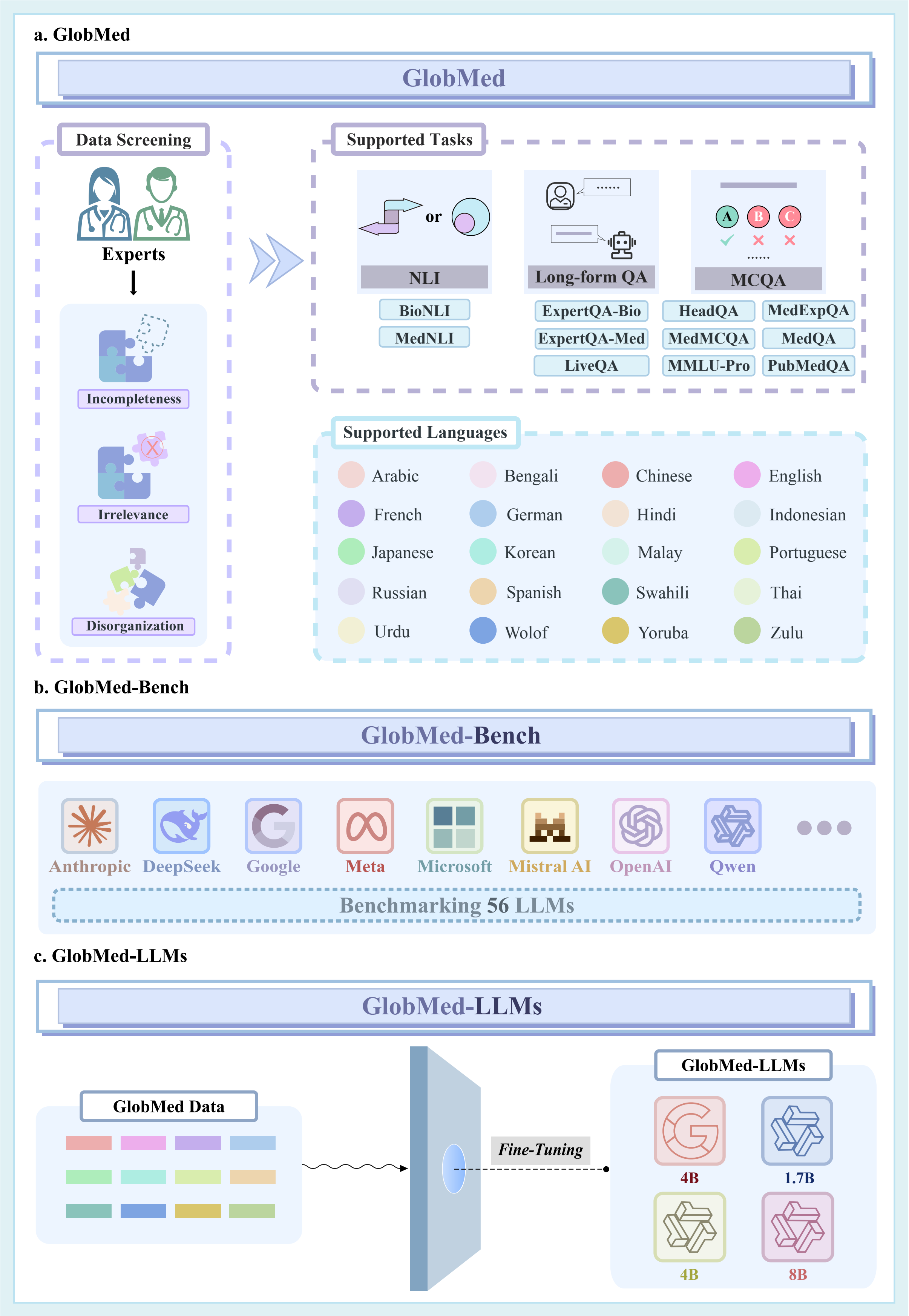

**Fig. 1: Overview of the three main contributions of the study. a, GlobMed:** A large multilingual medical dataset, covering 20 languages across three core tasks: natural language

inference, long-form question answering, and multiple-choice question answering. GlobMed includes 13 high-resource languages (Arabic, Chinese, English, French, German, Hindi, Indonesian, Japanese, Korean, Portuguese, Russian, Spanish, and Thai) and seven low-resource languages (Bengali, Malay, Swahili, Urdu, Wolof, Yoruba, and Zulu). **b, GlobMed-Bench:** A comprehensive multilingual medical benchmark evaluating 56 state-of-the-art proprietary and open-weight LLMs across 12 of the 20 languages represented in GlobMed: Chinese, English, French, German, Japanese, Korean, Portuguese, Spanish, Swahili, Wolof, Yoruba, and Zulu. The benchmark comprises over 20,000 independent experiments and more than 32 million responses, enabling systematic evaluation of performance disparities across languages. **c, GlobMed-LLMs:** A suite of multilingual medical LLMs ranging from 1.7B to 8B parameters, trained on GlobMed to improve medical performance across languages. GlobMed-LLMs were evaluated across six multilingual medical benchmarks and the 12 languages evaluated in GlobMed-Bench.

## Results

### GlobMed

We constructed GlobMed through three steps: data collection and screening, agentic machine translation, and expert evaluation (**Fig. 2**). GlobMed comprises over 800,000 entries across 20 languages, including 13 high-resource languages (Arabic, Chinese, English, French, German, Hindi, Indonesian, Japanese, Korean, Portuguese, Russian, Spanish, and Thai) and seven low-resource languages (Bengali, Malay, Swahili, Urdu, Wolof, Yoruba, and Zulu). These entries cover three core tasks: NLI, long-form QA, and MCQA. Additionally, GlobMed was independently evaluated by a global team of medical experts proficient in the corresponding target languages to ensure both linguistic accuracy and clinical validity.

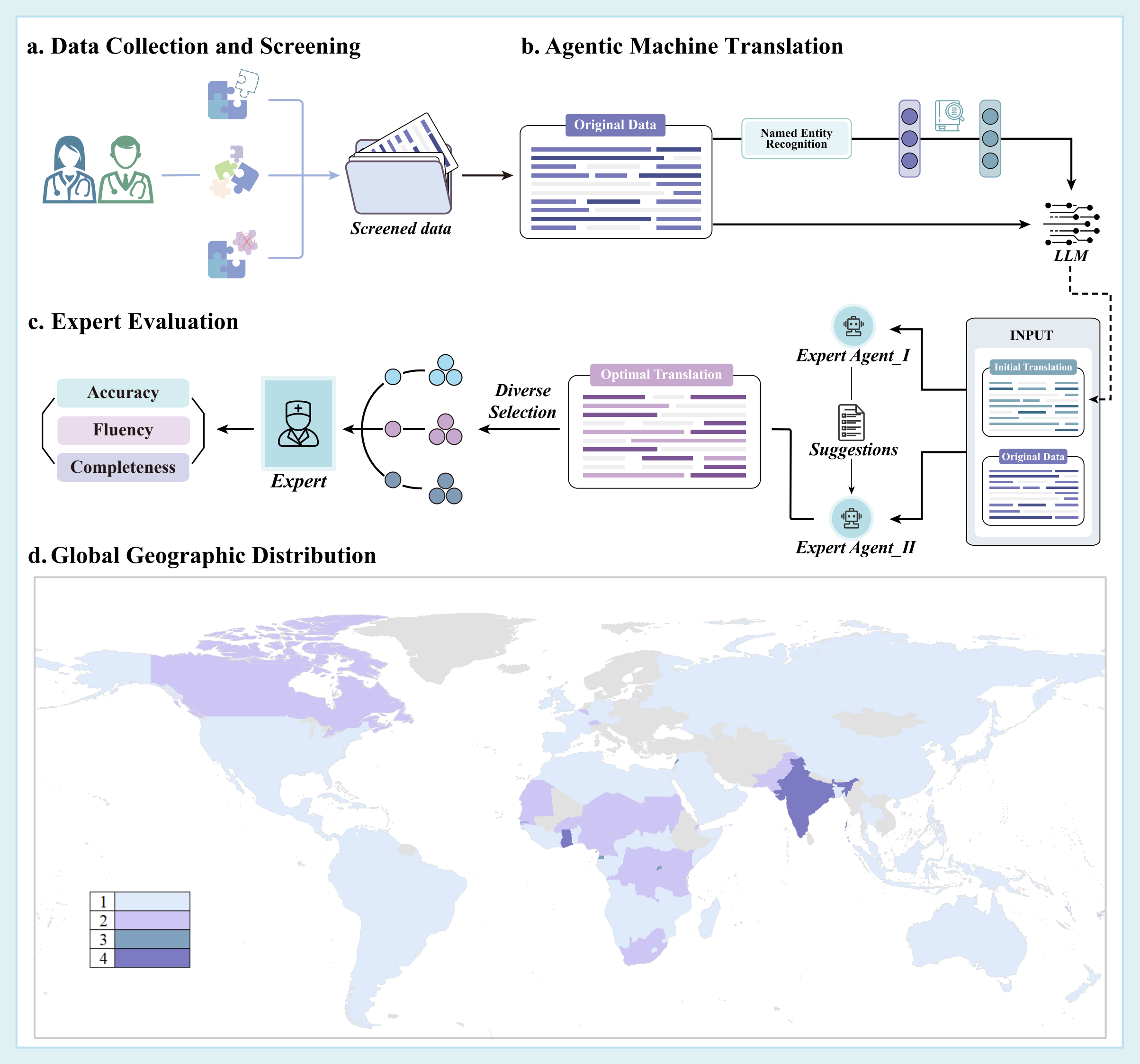

**Fig. 2: Overview of the GlobMed curation workflow. a, Data Collection and Screening:** Data were sourced from 11 medical datasets, including BioNLI and MedNLI for the NLI task, ExpertQA-Bio, ExpertQA-Med, and LiveQA for the long-form QA task, and HeadQA, MedExpQA, MedMCQA, MedQA, MMLU-Pro, and PubMedQA for the MCQA task. All original data (49,147 entries) were manually reviewed, during which 3,222 entries were removed because of incompleteness, irrelevance, and disorganization. The remaining high-quality screened data served as the foundation for subsequent multilingual expansion. **b, Agentic Machine Translation:** This process involved three steps: (1) named entity recognition and medical entity retrieval from a custom-built multilingual medical dictionary, followed by initial translation generation by an LLM; (2) reflection by Expert Agent I to identify semantic or structural issues and provide refinement suggestions; (3) optimized translation generated by Expert Agent II, incorporating the suggested improvements. **c, Expert Evaluation:** To ensure diverse topical coverage, topic modeling was performed on each dataset to select representative samples across multiple thematic clusters. Each selected entry was then independently

evaluated by at least two medical experts proficient in the corresponding target language across three dimensions: accuracy, fluency, and completeness. **d, Global Geographic Distribution:** A world map illustrates the geographic distribution of the 20 languages represented in GlobMed. Countries and regions are shaded to indicate inclusion, with color intensity representing the number of GlobMed-supported languages recognized as official or national languages. GlobMed spans countries and regions across Asia, Europe, Africa, and the Americas.

## GlobMed-Bench

*Overall Performance of Proprietary LLMs and Open-Weight LLMs*

We selected 12 of the 20 languages represented in GlobMed, including Chinese, English, French, German, Japanese, Korean, Portuguese, Spanish, Swahili, Wolof, Yoruba, and Zulu, to construct GlobMed-Bench. Meanwhile, GlobMed-Bench includes six datasets from the NLI and MCQA components of GlobMed: BioNLI and MedNLI for NLI, and HeadQA, MedExpQA, MedQA, and MMLU-Pro for MCQA. Based on this benchmark, we systematically evaluated 12 proprietary LLMs and 44 open-weight LLMs, resulting in more than 20,000 independent experiments and over 32 million model responses (**Fig. 3**).

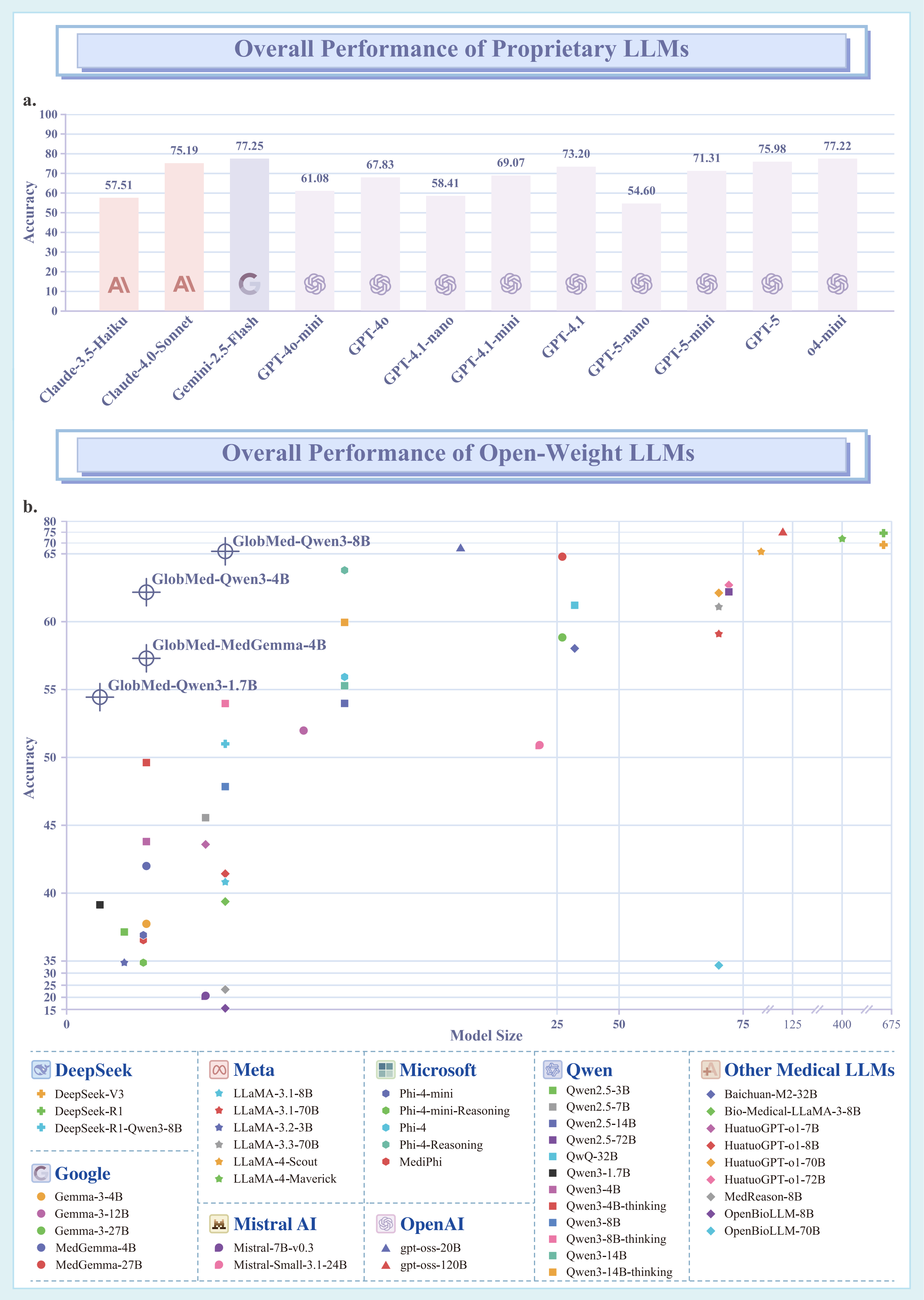

**Fig. 3: Overview of GlobMed-Bench. a, Overall performance of proprietary LLMs.** The bar chart displays the overall performance (y-axis, measured by accuracy) of 12 state-of-the-art proprietary LLMs on GlobMed-Bench. The evaluated LLMs include the Anthropic series (Claude-

3.5-Haiku and Claude-4.0-Sonnet), Google's Gemini-2.5-Flash, and the OpenAI series (GPT-4o-mini, GPT-4o, GPT-4.1-nano, GPT-4.1-mini, GPT-4.1, GPT-5-nano, GPT-5-mini, GPT-5, o4-mini). For the OpenAI GPT-5 series, we set the "reasoning effort" to "minimal." **b, Overall performance of open-weight LLMs:** The scatter plot displays the relationship between model size (x-axis, measured in billions of parameters) and overall performance (y-axis, measured by accuracy) for 44 open-weight LLMs evaluated on GlobMed-Bench. The evaluated LLMs include the DeepSeek, Meta LLaMA, Microsoft Phi, Qwen, Google Gemma, Mistral AI, OpenAI gpt-oss series, and specialized medical LLMs. For Qwen3-1.7B, we set it to "non-thinking" mode. Additionally, we marked the performance of GlobMed-LLMs for comparison.

Proprietary LLMs generally achieved stronger performance, with accuracies ranging from 54.60% to 77.25%. Meanwhile, multiple proprietary LLMs surpassed the 75% accuracy threshold, underscoring their current advantage in multilingual medical tasks. The top-performing LLMs were Gemini-2.5-Flash[26] (77.25%), o4-mini[27] (77.22%), and GPT-5[28] (75.98%), with Claude-4.0-Sonnet[29] (75.19%) also demonstrating strong multilingual medical capability.

Open-weight LLMs ranged from 1.7B to 671B parameters. Overall, performance showed a clear positive correlation with parameter scale. The top-performing LLMs achieved accuracies of approximately 75%, while the lowest-performing LLM achieved an accuracy of approximately 15%. Among all evaluated open-weight LLMs, gpt-oss-120B[30] (74.74%), DeepSeek-R1[31] (74.56%), and LLaMA-4-Maverick[32] (71.94%) achieved the strongest performance and were the only open-weight LLMs to exceed the 70% accuracy threshold. Notably, several medium-sized LLMs, such as gpt-oss-20B[30] (67.37%), delivered unexpectedly strong performance despite their relatively modest parameters, highlighting promising scaling efficiency.

*Cross-Lingual Performance Disparities across 56 LLMs*

To systematically evaluate the cross-lingual performance consistency, we used the original language of each dataset as the reference and compared performance across 56 LLMs for all other languages (**Fig. 4**). Specifically, for datasets originally in English, English served as the reference; for those in Spanish, Spanish served as the reference. When English was the reference, most languages, including high-resource languages, exhibited significant performance gaps relative to English. In contrast, when Spanish served as the reference, disparities among high-resource languages were notably smaller. Under both reference conditions, however, low-resource languages (e.g., Swahili, Wolof, Yoruba, Zulu) consistently showed pronounced performance gaps relative to the reference language across nearly all LLMs, indicating that current LLMs still have markedly insufficient medical knowledge comprehension capability in low-resource languages. To further assess whether cross-lingual performance disparities persist in more open-ended and clinically relevant scenarios, we conducted an additional multilingual HealthBench evaluation[33]. Specifically, we selected 30 samples from HealthBench and evaluated them across six languages, including English, French, Japanese, Korean, Swahili, and Yoruba. The results showed that cross-lingual performance disparities persisted in this setting as well. The detailed evaluation protocol and results are provided in **Supplementary Information S2.5 (HealthBench Evaluation)**.

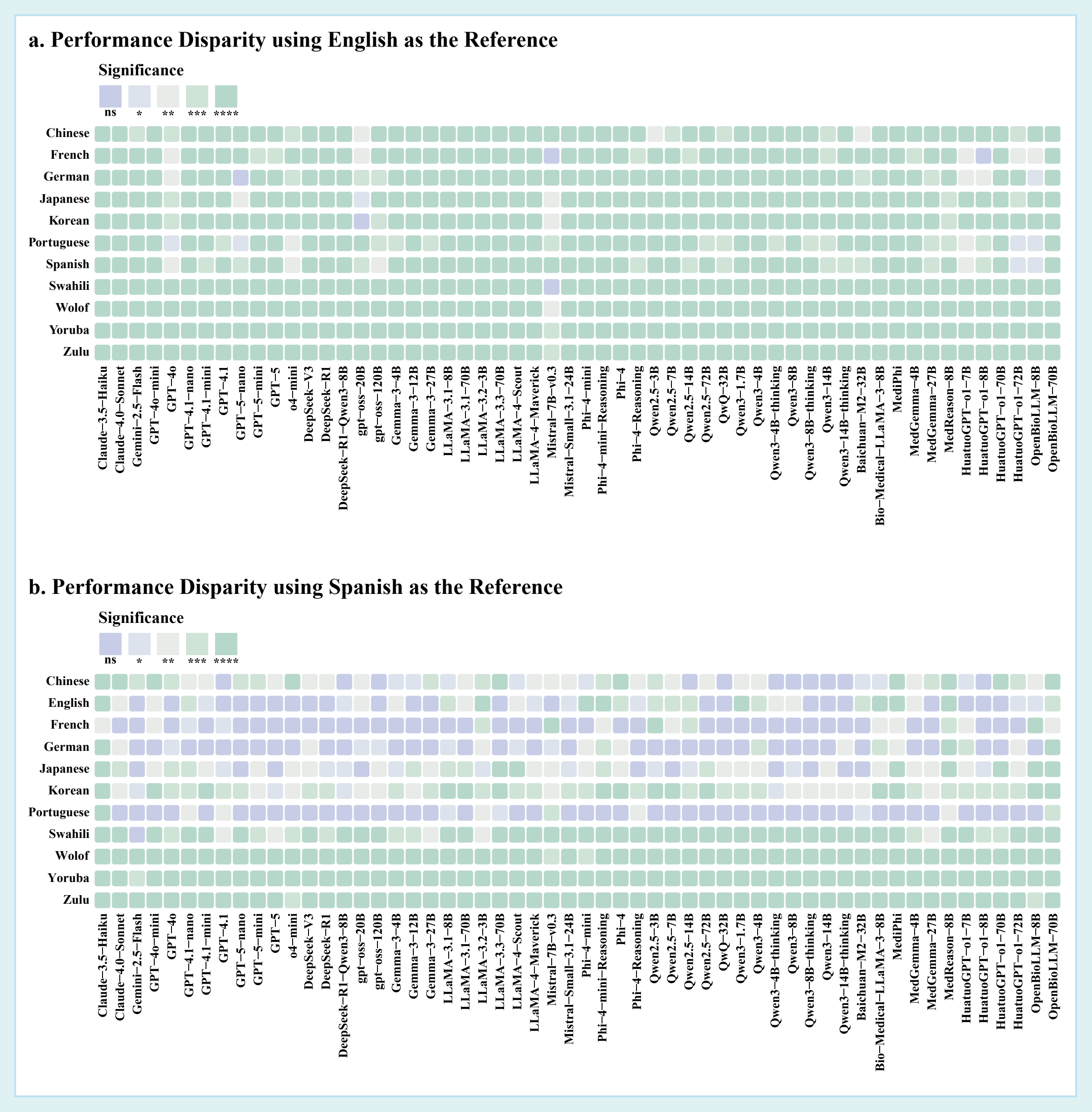

**Fig. 4: Cross-Lingual Performance Disparities across 56 LLMs. a, English as the reference language:** For datasets originally in English, each cell represents the statistical significance of the performance disparity between English and the target language (y-axis) for a given LLM (x-axis). **b, Spanish as the reference language:** For datasets originally in Spanish, each cell represents the statistical significance of the performance disparity between Spanish and the target language (y-axis) for a given LLM (x-axis). Statistical significance is indicated by asterisks (*p<0.05, **p<0.01, ***p<0.001, ****p<0.0001).

*Performance Comparison Between General LLMs and Medical LLMs*

To evaluate the effect of medical-specific training on multilingual performance, we compared general LLMs with their corresponding medical variants on GlobMed-Bench (**Fig. 5**). Overall, medical LLMs did not consistently outperform their general counterparts. Meanwhile, performance varied substantially across LLMs, indicating that the benefits of medical-specific training depend heavily on training strategy.

Among the medical LLMs, the MedGemma series[34] showed the most consistent performance gains, with MedGemma-4B[34] improving accuracy from 37.74% to 42.00% relative to Gemma-3-4B[35], and MedGemma-27B[34] improving accuracy from 58.84% to 64.79% relative to Gemma-3-27B[35]. Radar chart analysis revealed that MedGemma[34] consistently improved performance across all six medical benchmarks and 12 languages. The HuatuoGPT-o1 series[36] achieved modest improvements over their general counterparts at multiple scales (8B, 70B, 72B), but these improvements were generally below 4 percentage points. In contrast, several medical LLMs performed worse than their general counterparts, such as HuatuoGPT-o1-7B[36], Bio-Medical-LLaMA-3-8B[37], MedReason-8B[38], and OpenBioLLM-8B[39]/70B[40], with some models exhibiting accuracy declines exceeding 20%.

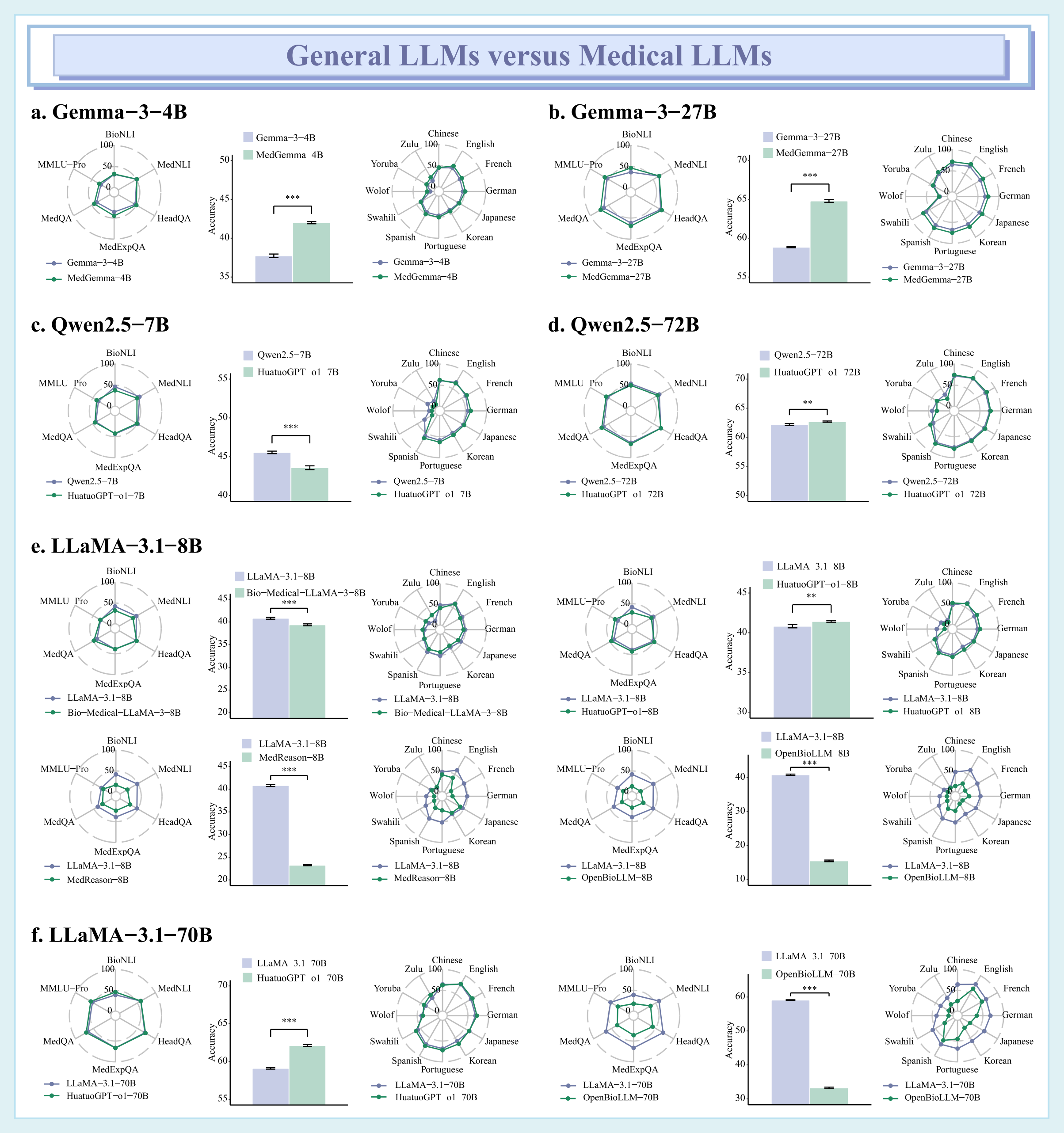

**Fig. 5: Performance comparison between general LLMs and medical LLMs.** The figure contains six panels (a–f), each comparing general LLMs with their medical variants across six medical benchmarks (left radar chart), 12 languages (right radar chart), and overall accuracy (middle bar chart). **a, Gemma-3-4B:** Comparison between Gemma-3-4B and MedGemma-4B. **b, Gemma-3-27B:** Comparison between Gemma-3-27B and MedGemma-27B. **c, Qwen2.5-7B:** Comparison between Qwen2.5-7B and HuatuoGPT-o1-7B. **d, Qwen2.5-72B:** Comparison between Qwen2.5-72B and HuatuoGPT-o1-72B. **e, LLaMA-3.1-8B:** Comparison between LLaMA-3.1-8B and four medical LLMs (Bio-Medical-LLaMA-3-8B, HuatuoGPT-o1-8B, MedReason-8B, OpenBioLLM-8B). **f, LLaMA-3.1-70B:** Comparison between LLaMA-3.1-70B and two medical LLMs (HuatuoGPT-o1-70B, OpenBioLLM-70B). Statistical significance is indicated by asterisks (*p<0.05, **p<0.01, ***p<0.001).

*Performance Comparison Between Non-Reasoning LLMs and Reasoning LLMs*

We compared non-reasoning LLMs with their reasoning-enhanced counterparts on GlobMed-Bench (**Fig. 6**). Overall, reasoning LLMs demonstrated clear advantages in multilingual medical tasks. For instance, DeepSeek-R1 improved accuracy from 69.15% to 74.56%, compared with DeepSeek-V3, and reasoning variants of the Qwen3 series outperformed their non-reasoning counterparts across all scales (4B, 8B, and 14B), with improvements ranging from 4.67 to 6.13 percentage points. Similarly, Phi-4-reasoning improved accuracy from 55.93% to 63.79%. However, the benefits were not universal; some reasoning variants, such as Phi-4-mini-reasoning, underperformed their non-reasoning counterparts, with a decrease of 2.58 percentage points in accuracy.

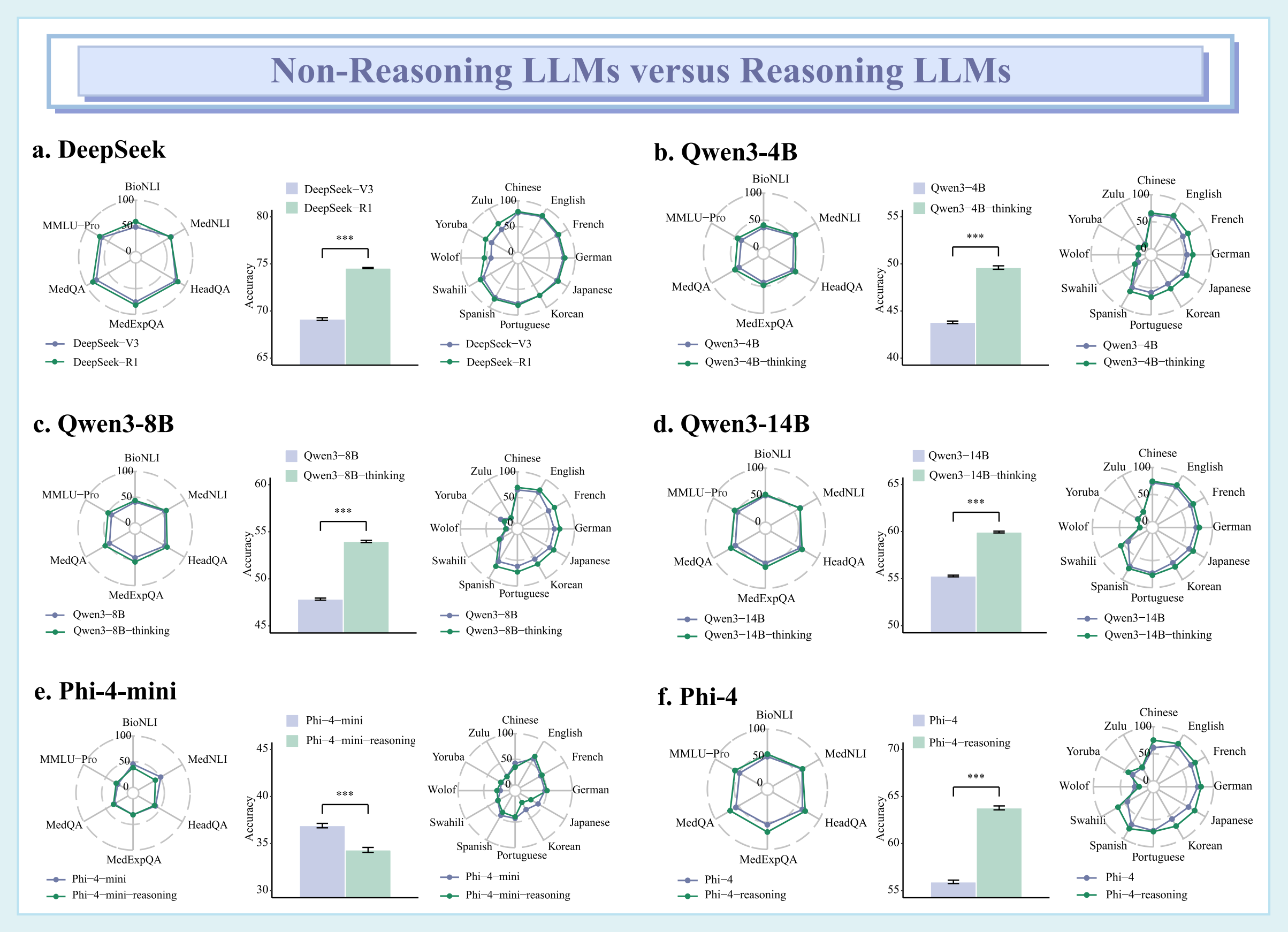

**Fig. 6: Performance comparison between non-reasoning LLMs and reasoning LLMs.** The figure contains six panels (a–f), each comparing non-reasoning LLMs and their reasoning-enhanced counterparts across six medical benchmarks (left radar chart), 12 languages (right radar chart), and overall accuracy (middle bar chart). **a, DeepSeek:** Comparison between DeepSeek-V3 and DeepSeek-R1. **b, Qwen3-4B:** Comparison between Qwen3-4B and Qwen3-4B-

thinking. **c, Qwen3-8B:** Comparison between Qwen3-8B and Qwen3-8B-thinking. **d, Qwen3-14B:** Comparison between Qwen3-14B and Qwen3-14B-thinking. **e, Phi-4-mini:** Comparison between Phi-4-mini and Phi-4-mini-reasoning. **f, Phi-4:** Comparison between Phi-4 and Phi-4-reasoning. Statistical significance is indicated by asterisks (*p<0.05, **p<0.01, ***p<0.001).

### GlobMed-LLMs

We fine-tuned MedGemma-4B[34] and the Qwen3 series[41] (1.7B, 4B, 8B) using GlobMed training data corresponding to the 12 languages included in GlobMed-Bench to enhance multilingual medical capability and mitigate performance disparities in underrepresented languages. The resulting fine-tuned LLMs, collectively referred to as GlobMed-LLMs, demonstrated substantial improvements over their baseline counterparts across multiple benchmarks and these 12 languages (**Fig. 7**).

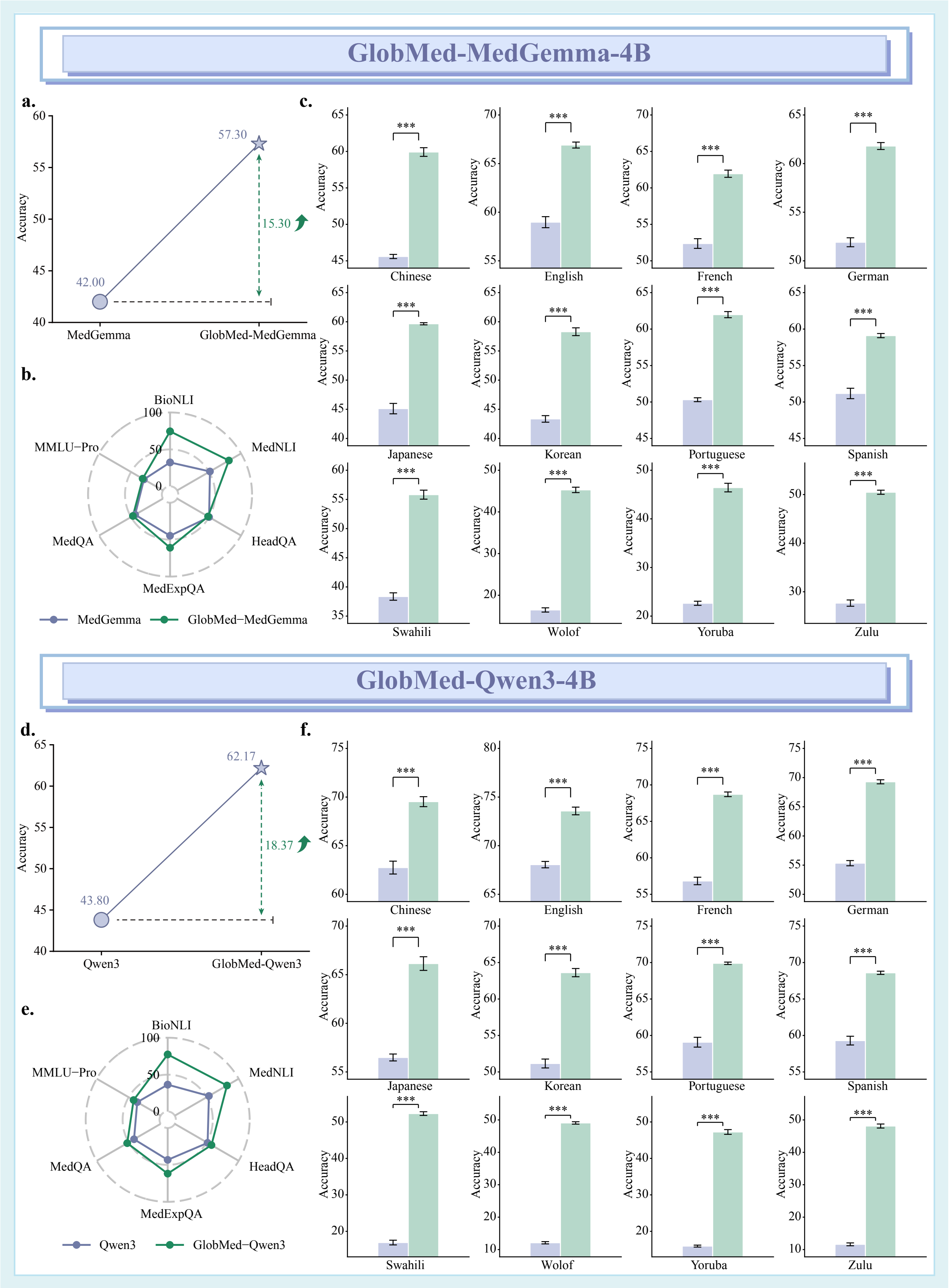

**Fig. 7: Performance comparison of GlobMed-LLMs versus baseline LLMs. a, GlobMed-MedGemma-4B overall performance:** Average accuracy across all benchmarks and languages improved from 42.00% to 57.30% compared with MedGemma-4B. **b, Task-wise performance:** GlobMed-MedGemma-4B outperformed MedGemma-4B on nearly all medical benchmarks, with

a slight decrease on HeadQA. **c, Language-wise performance:** GlobMed-MedGemma-4B achieved higher average accuracy across the 12 benchmark languages compared with MedGemma-4B, with particularly notable improvements in low-resource languages. **d, GlobMed-Qwen3-4B overall performance:** Average accuracy across all benchmarks and languages improved from 43.80% to 62.17% compared with Qwen3-4B. **e, Task-wise performance:** GlobMed-Qwen3-4B consistently outperformed Qwen3-4B across all medical benchmarks. **f, Language-wise performance:** GlobMed-Qwen3-4B achieved higher average accuracy across the 12 benchmark languages compared with Qwen3-4B, with particularly notable improvements in low-resource languages. Statistical significance is indicated by asterisks (*p<0.05, **p<0.01, ***p<0.001).

Specifically, GlobMed-MedGemma-4B increased overall accuracy from 42.00% to 57.30%, outperforming the original MedGemma-4B[34] on nearly all medical benchmarks, with only a slight decrease on HeadQA. Across 12 languages, it maintained consistent improvements in all high-resource languages and showed notable gains in low-resource languages, including Swahili, Wolof, Yoruba, and Zulu.

Similarly, GlobMed-Qwen3-4B improved overall accuracy from 43.80% to 62.17%, outperforming Qwen3-4B[41] across all evaluated benchmarks and demonstrating substantial improvements in every language. Collectively, these results indicate that fine-tuning with GlobMed effectively improved the multilingual medical capabilities of LLMs while narrowing performance gaps across languages.

## Discussion

LLMs are rapidly transforming medical AI, yet their development has unintentionally widened global disparities in access to medical information, particularly in regions where underrepresented languages are spoken. Building medical AI systems that can effectively serve linguistically diverse populations is, therefore, not only a technical challenge but also a critical step toward improving global health equity.

In this study, we address this gap through three key contributions: (1) GlobMed, the largest multilingual medical dataset to date, spanning 20 languages, including seven low-resource languages, and containing more than 800,000 entries; (2) GlobMed-Bench, a large-scale evaluation benchmark assessing the multilingual medical capabilities of 56 proprietary and open-weight LLMs through over 20,000 independent experiments and more than 32 million model responses; and (3) GlobMed-LLMs, a suite of fine-tuned LLMs ranging from 1.7B to 8B parameters that substantially enhance performance and reduce cross-lingual disparities.

Systematic evaluation on GlobMed-Bench revealed distinct performance trends between proprietary and open-weight LLMs. Proprietary LLMs consistently achieved high accuracy within a narrow range, whereas open-weight LLMs exhibited broader variability. These differences likely reflect disparities in training data, computing resources, and optimization strategies. Proprietary LLMs are typically developed by well-resourced technology companies, whereas many open-weight LLMs originate from academic teams with more limited resources.

Cross-lingual analysis highlighted persistent performance disparities across languages. Models generally performed better in high-resource languages, such as English, French, German, Portuguese, and Spanish. In contrast, underrepresented languages, particularly Wolof, Yoruba, and Zulu, remained challenging for all evaluated models. Notably, model performance showed certain associations with their development context. For instance, LLMs developed in China performed particularly well in Chinese, suggesting strong language-specific adaptation.

The cross-lingual performance disparities may be influenced by several interrelated factors. First, low-resource languages are typically underrepresented in large-scale pretraining corpora, and the volume and linguistic coverage of pretraining data are closely associated with multilingual downstream task performance[42]. Second, multilingual tokenizers may induce greater subword fragmentation for certain languages, while the degree to which a tokenizer is adapted to a target language may further affect linguistic representation and downstream task performance[43]. Third, medical corpora are often more limited in low-resource languages, which may further restrict models' exposure to and learning of medical concepts expressed in these languages[22]. Therefore, the cross-lingual disparities observed in this study are unlikely to be driven by a single factor, but may instead reflect the combined effects of data availability, linguistic representation, and domain-specific resource limitations.

Our findings further suggest that domain adaptation strategies and model reasoning capabilities are important determinants of multilingual medical performance. Medical LLMs exhibited substantial variability in performance, with only a subset demonstrating clear advantages over their corresponding general LLMs. This indicates that further training or fine-tuning on medical data alone does not necessarily yield consistent improvements in multilingual medical capability[44]. Instead, effective domain adaptation likely requires high-quality pretraining, optimized training strategies, and careful incorporation of domain knowledge[45]. Meanwhile, LLMs equipped with explicit reasoning mechanisms showed clear gains across multiple multilingual medical benchmarks, though their higher computational demands and longer inference times may limit practical deployment in resource-constrained settings.

In addition, GlobMed-LLMs achieved significant improvements over their baseline LLMs, indicating that targeted multilingual medical training can effectively strengthen cross-lingual medical capability and narrow performance gaps across languages. Despite their relatively modest sizes, GlobMed-LLMs outperformed some much larger LLMs. This efficiency may reduce deployment barriers and operational costs, a critical consideration for under-resourced regions.

In summary, through GlobMed, GlobMed-Bench, and GlobMed-LLMs, this study provides an important foundation for the development of more inclusive medical AI and represents a key step toward enabling medical LLMs to serve broader linguistic communities. Importantly, multilingual capability represents only one dimension of the global applicability of medical AI. Beyond the language dimension, other dimensions, such as regional disease patterns and sociocultural contexts, are equally important. Therefore, the future evaluation and development of medical LLMs should adopt a more comprehensive approach that jointly considers language, regional disease patterns, and sociocultural contexts. This will also be a key focus of the next stage of our work. From this perspective, this study can be viewed as an important foundational step toward advancing global medical AI. At the same time, truly global deployment of medical AI will require sustained collaboration among academia, industry, health care institutions, and governments, together with further validation of applicability and safety, as well as appropriate governance requirements, in real-world clinical settings. Through long-term and coordinated efforts, medical AI can ultimately serve and benefit populations across diverse contexts worldwide[46].

## Limitations

First, although GlobMed currently covers a broad range of languages and populations, its overall scale and linguistic coverage remain limited relative to the global linguistic landscape. Many low-resource languages still lack sufficient high-quality medical data, limiting the applicability of the proposed models across diverse medical scenarios. Second, although we adopted a multi-stage agentic machine translation framework and further validated translation quality through medical expert evaluation, some heterogeneity in translation quality may remain across languages, particularly for low-resource languages with more limited linguistic and medical terminology resources. Therefore, potential biases introduced by translation cannot be completely ruled out. Third, the large-scale systematic evaluation in GlobMed-Bench remains primarily based on structured tasks with clearly defined ground-truth answers, such as NLI and MCQA. Although we further incorporated a multilingual HealthBench evaluation, extending the evaluation to more open-ended and clinically relevant scenarios, this evaluation currently covers only a limited number of questions, languages, and representative models. Consequently, the current evaluation cannot fully capture the complexity of real-world clinical applications, such as multi-turn physician–patient interactions, decision support within clinical workflows, and other more complex open-ended clinical tasks[47]. In addition, dedicated safety evaluation tasks in multilingual settings remain limited. Finally, the current model training framework relies solely on post-training and lacks a multilingual pretraining stage specifically tailored to the medical domain. This may constrain the development of a more comprehensive and balanced representation of multilingual medical knowledge. GlobMed-LLMs were evaluated on structured medical benchmarks, and assessing their performance on open-ended tasks remains an important direction for future work. Addressing these limitations will be important for developing more comprehensive, equitable, and clinically relevant multilingual medical AI systems.

# Methods

## GlobMed: Constructing the Multilingual Medical Dataset

*Data Collection and Screening*

We collected data encompassing three core tasks: NLI, long-form QA, and MCQA. Specifically, the NLI task includes BioNLI[48] and MedNLI[49]; the long-form QA task includes ExpertQA-Bio[50], ExpertQA-Med[50], and LiveQA[51]; and the MCQA task includes HeadQA[52], MedExpQA[53], MedMCQA[54], MedQA[55], MMLU-Pro[56], and PubMedQA[57]. All datasets are publicly accessible, except MedNLI, which can be accessed through the PhysioNet platform[58].

Despite the value of these resources for medical research, our manual review identified three main quality issues within the original data: incompleteness (missing critical information), irrelevance (weak relevance to medicine), and disorganization (inconsistent formatting or structural irregularities). To ensure data reliability, we conducted a multi-stage quality control process on the original datasets, which contained 49,147 entries, resulting in the removal of 3,222 entries that did not meet quality standards. Ambiguous cases were verified by medical experts. This rigorous screening established a high-quality foundation for subsequent multilingual machine translation, benchmark evaluation, and LLM fine-tuning within the GlobMed framework.

Further details on the data curation process are provided in **Supplementary Information S1.1 (Data Curation)**; detailed dataset statistics are provided in **Supplementary Information S1.2 (Data Statistics)**, and representative examples are provided in **Supplementary Information S1.3 (Data Example)**.

*Agentic Machine Translation*

Following data screening, we developed a flexible agentic machine translation framework to expand GlobMed into multiple languages. The framework comprises three stages. In the first stage, we utilized the "Medical NER Model"[59] to extract medical entities from the original data, which were matched to translations from our custom-built multilingual medical dictionary comprising approximately 350,000 translation pairs for resource-available languages (Chinese, English, French, German, Japanese, Korean, Portuguese, Spanish). The extracted entities and dictionary translations, along with the original text, were then provided to an LLM to generate an initial translation. In the second stage, Expert Agent I reviewed the initial translation, identified semantic or structural issues, and provided suggestions for refinement. In the third stage, Expert Agent II incorporated these suggestions to generate the final optimized translation. Detailed prompts for the agentic machine translation framework are provided in **Supplementary Information S1.4 (Agentic Machine Translation Prompt)**.

Leveraging this framework, GlobMed was expanded to 20 languages, including 13 high-resource languages (Arabic, Chinese, English, French, German, Hindi, Indonesian, Japanese, Korean, Portuguese, Russian, Spanish, and Thai) and seven low-resource languages (Bengali, Malay, Swahili, Urdu, Wolof, Yoruba, and Zulu).

For the original translator selection, we evaluated several LLMs and commercial products, including Claude-3.5-Sonnet[60], GPT-4o-mini[61], GPT-4o[61], Google Translate[62], and DeepL Translate[63]. Multiple medical experts independently evaluated translation quality, including terminology precision, semantic consistency, and content fluency. The evaluation results demonstrated that Claude-3.5-Sonnet[60] achieved the best overall performance and was therefore adopted as the primary translator. As more advanced LLMs became available, we subsequently upgraded to Claude-4.0-Sonnet[29], further improving the quality and stability of multilingual translation.

*Expert Evaluation*

To further mitigate potential biases introduced by machine translation and enhance data reliability, we implemented an expert evaluation process. Topic modeling[64] was applied to select representative samples from multiple thematic clusters for each task. For the multilingual data, each selected entry was independently evaluated by at least two medical experts proficient in the corresponding target language. During evaluation, experts assessed translation accuracy, fluency, and completeness using a five-point (1–5) scale. Detailed evaluation criteria for the three dimensions are provided in **Supplementary Information S1.5 (Expert Evaluation Criteria)**.

We selected six representative languages, French, German, Japanese, Korean, Swahili, and Yoruba, for translation quality evaluation. For each language, 300 samples were randomly selected and independently evaluated under blinded conditions. We further used Expert Agreement (EA) to quantify agreement between the two experts. The five-point ratings were grouped into three categories: scores of 1-2, a score of 3, and scores of 4-5; agreement was considered to be achieved when the two experts assigned ratings within the same category. EA was calculated separately for accuracy, fluency, and completeness, while overall EA represents the average agreement rate across the three dimensions. Detailed translation quality evaluation results are provided in **Supplementary Information S1.6 (Translation Quality)**.

## GlobMed-Bench: Evaluating 56 LLMs Across 12 Languages

*LLM Evaluation*

To systematically evaluate current LLMs on multilingual medical benchmarks, we constructed GlobMed-Bench, incorporating proprietary LLMs, open-weight general LLMs, and open-weight medical-specific LLMs, covering both reasoning and non-reasoning variants.

Proprietary LLMs included the Anthropic series[29,65] (Claude-3.5-Haiku, Claude-4.0-Sonnet), Google's Gemini-2.5-Flash[26], and the OpenAI series[27,28,61,66] (GPT-4o-mini, GPT-4o, GPT-4.1-nano, GPT-4.1-mini, GPT-4.1, GPT-5-nano, GPT-5-mini, GPT-5, o4-mini). For the OpenAI GPT-5 series, we set the "reasoning effort" to "minimal". Open-weight general LLMs comprised the DeepSeek series[31,67] (DeepSeek-V3, DeepSeek-R1, DeepSeek-R1-Qwen3-8B), the Gemma series[35] (Gemma-3-4B/12B/27B), the gpt-oss series[30] (gpt-oss-20B/120B), the LLaMA series[32,68–70] (LLaMA-3.1-8B/70B, LLaMA-3.2-3B, LLaMA-3.3-70B, LLaMA-4-Scout, LLaMA-4-Maverick), the Mistral series[71,72] (Mistral-7B-v0.3, Mistral-Small-3.1-24B), the Phi series[73] (Phi-4-mini, Phi-4 and their corresponding reasoning variants), and the Qwen series[41,74] (Qwen2.5-3B/7B/14B/72B, QwQ-32B, Qwen3-1.7B/4B/8B/14B and their corresponding thinking variants). For Qwen3-1.7B, we set it to "non-thinking" mode. Open-weight medical-specific LLMs included the MedGemma series[34] (MedGemma-4B/27B), the HuatuoGPT series[36] (HuatuoGPT-o1-7B/8B/70B/72B), OpenBioLLM-8B[39]/70B[40], Baichuan-M2-32B[75], Bio-Medical-LLaMA-3-8B[37], MediPhi[76], and MedReason-8B[38]. Detailed information on all evaluated LLMs is provided in **Supplementary Information S2.1 (LLM Information)**.

All LLMs were evaluated on the six GlobMed-Bench datasets across the 12 benchmark languages, with five independent runs per evaluation to ensure statistical reliability.

*Prompt Design*

To authentically capture LLM capabilities in multilingual medical scenarios, prompts were delivered in the target language rather than in a mixture of languages or in English, adhering to a strict language-consistency principle. This design is critical, as prompts in non-target languages can introduce comprehension bias and fail to reliably measure true multilingual performance[77]. All prompt templates were carefully designed by medical experts proficient in the corresponding target languages for each of the 12 benchmark languages to guarantee both linguistic naturalness and cultural appropriateness. Detailed evaluation prompts for all 12 languages are provided in **Supplementary Information S2.2 (Evaluation Prompt)**.

*Criteria for Counting Experiments and Model Responses*

We defined one independent experiment as one complete evaluation run of a specific model on a specific dataset in a specific language. GlobMed-Bench includes six medical datasets, 56 LLMs, and 12 languages, with five independent runs conducted for each model–dataset–language combination. Accordingly, GlobMed-Bench comprised 20,160 independent experiments, calculated as 6 datasets × 56 LLMs × 12 languages × 5 independent runs. At the individual test sample level, the six datasets contained a total of 9,670 test samples. Each test sample was evaluated by all 56 LLMs across 12 languages in five independent runs, resulting in 32,491,200 model responses, calculated as 9,670 test samples × 56 LLMs × 12 languages × 5 independent runs.

*Statistical Analysis*

Accuracy was calculated as the percentage of correct predictions, and aggregate accuracy was calculated using an unweighted arithmetic mean. Statistical analyses were conducted across the dataset–language combinations included in the corresponding experiment. Each model–dataset–language combination was evaluated in five independent runs, and the corresponding results from the two groups were paired to calculate p values using two-sided paired-sample t-tests. We further applied multiple-comparison correction separately to cross-lingual performance disparities, comparisons between general and medical LLMs, comparisons between non-reasoning and reasoning LLMs, and comparisons between GlobMed-LLMs and their corresponding baseline LLMs to assess the robustness of the results. Error bars represent the standard deviation of the accuracy values obtained across the five independent runs. Detailed results for all 56 LLMs are provided in **Supplementary Information S2.3 (Detailed Results of 56 Proprietary and Open-Weight LLMs)**, and the cross-lingual performance disparity analysis is provided in **Supplementary Information S2.4 (Language Disparity Analysis)**.

*HealthBench Evaluation*

To further assess whether cross-lingual performance disparities persist in more open-ended and clinically relevant scenarios, we conducted an additional multilingual HealthBench evaluation[33]. We selected 30 samples from HealthBench and extended them to six languages: English, French, Japanese, Korean, Swahili, and Yoruba. We evaluated the high-performing proprietary LLM and open-weight LLM on GlobMed-Bench, Gemini-2.5-Flash[26] and DeepSeek-R1[31], respectively. For the English samples, we followed the original HealthBench evaluation framework and used an LLM-as-a-Judge for scoring. For the other five languages, each model response was independently evaluated under blinded conditions by two medical experts proficient in the corresponding target language according to the HealthBench rubric. Detailed evaluation procedures and results are provided in **Supplementary Information S2.5 (HealthBench Evaluation)**.

**GlobMed-LLMs: Developing Multilingual Medical LLMs**

*Fine-Tuning Multilingual Medical LLMs*

In the fine-tuning stage, we selected MedGemma-4B[34] and the Qwen3 series (1.7B, 4B, 8B)[41] as baseline LLMs. These models represent the top-performing non-reasoning LLMs at comparable parameter scales. Meanwhile, we focused on LLMs with relatively small parameters to improve accessibility in regions with limited AI infrastructure and low-resource medical communities. GlobMed-LLMs were trained using GlobMed data corresponding to the 12 languages included in GlobMed-Bench.

We adopted two complementary supervised fine-tuning (SFT) strategies: direct SFT and distillation-enhanced SFT. For direct SFT, we directly used the training data and adopted a standard generative supervised fine-tuning framework. We organized training samples in an instruction format and optimized the models using an autoregressive language modeling objective, rather than formulating NLI or MCQA as classification tasks with separate heads.

For distillation-enhanced SFT, we used gpt-oss-120B[30] to generate reasoning-enhanced responses containing explicit medical reasoning processes and answers, followed by multi-layer quality control to retain reliable supervision data. For each training sample, gpt-oss-120B[30] independently generated three reasoning-enhanced responses, and we compared the consistency of both the answers and the reasoning pathways. Samples that consistently produced the correct answer with similar core reasoning logic were preferentially retained, whereas those showing substantial reasoning disagreement underwent further quality checking. We subsequently used GPT-5-mini[28] as an independent verifier to assess the logical coherence of the reasoning process, its consistency with the original question, and whether it reasonably supported the answer. In addition, we randomly sampled 10 examples from each dataset used for distillation and asked medical experts to further evaluate the medical accuracy and logical coherence of the generated reasoning. The quality-controlled distilled data were subsequently expanded into 12 target languages using the same agentic machine translation framework as that used for GlobMed construction. During distilled-data construction, we further applied a language randomization strategy, whereby each distilled sample was randomly assigned to one of the 12 target languages to prevent reasoning-enhanced supervision from being disproportionately concentrated in a subset of languages and to promote more balanced cross-lingual reasoning supervision.

For GlobMed-LLMs of different parameter scales, we adopted different combinations of training data. GlobMed-Qwen3-1.7B and GlobMed-Qwen3-8B were trained using direct SFT only, whereas the two 4B models highlighted in the main text, GlobMed-MedGemma-4B and GlobMed-Qwen3-4B, were trained using both direct SFT and distillation-enhanced SFT data. For the two 4B models, the two data sources were first combined into a unified training set and jointly used for full-parameter fine-tuning in a single training stage, rather than being applied sequentially in two separate stages. The two data sources were mixed according to their natural data volumes without imposing an additional manually specified mixing ratio. The direct SFT experiments for GlobMed-Qwen3-1.7B and GlobMed-Qwen3-8B are detailed in **Supplementary Information S3.1 (GlobMed-Qwen3-1.7B/8B through Direct Supervised Fine-Tuning)**, and detailed performance comparisons between all GlobMed-LLMs and their corresponding baseline LLMs are provided in **Supplementary Information S3.2 (Performance Comparison between GlobMed-LLMs and Baseline LLMs)**.

*Training Setup*

All fine-tuning experiments were conducted on a server equipped with 16 NVIDIA H100 GPUs (96GB memory each). We employed full-parameter fine-tuning with an initial learning rate of 1.0e-5, a global batch size of 256, and a single training epoch. Mixed precision training (bfloat16) was employed to improve computational efficiency. The AdamW optimizer[78] was selected with a weight decay coefficient of 0.0. For learning rate scheduling, we applied a cosine annealing strategy[79] with a warmup[80] period during the first 10% of training steps. All experiments were implemented using the Hugging Face Transformers framework[81].

## Data Availability

The GlobMed dataset constructed in this study is publicly available on the Hugging Face platform at https://huggingface.co/collections/ruiyang-medinfo/globmed. It excludes the original MedNLI data and their multilingual translations. Owing to privacy protection requirements and institutional policies governing the use and distribution of MedNLI, access to MedNLI-related data must be requested through the PhysioNet platform (https://physionet.org/content/mednli/1.0.0/).

## Code Availability

All code used for training and evaluation in this study has been made publicly available on GitHub at https://github.com/ruiyang-medinfo/GlobMed. The weights of the GlobMed-LLMs cannot be publicly released, as the training process incorporated MedNLI-related data, which is subject to usage and distribution restrictions.

## Acknowledgements

This work was supported by the Duke-NUS Signature Research Programme funded by the Ministry of Health, Singapore. Any opinions, findings and conclusions or recommendations expressed in this material are those of the author(s) and do not reflect the views of the Ministry of Health. This work was supported by Innosuisse – Swiss Innovation Agency: Innovation project 55441.1 IP-ICT, and European Union's Horizon Europe research and innovation programme under grant agreement No 101287214 (AIM-SAFE). This work was partially supported by the NIH R01LM014344, R01LM014573 and R01LM014604. The findings and conclusions presented in this paper are those of the author(s) and do not necessarily reflect the views of the NIH or the U.S. Department of Health and Human Services. Additionally, we thank Leticia Johnson, Kosuke Kita, and Yohei Okada for supporting parts of the data evaluation; Irene Li for providing partial translation APIs and early-stage HPC resources, which were supported by JST ACT-X (JPMJAX24CU), JSPS KAKENHI (24K20832), Kyushu University Research Institute for Information Technology through the HPCI System Research Project (hp250092), NVIDIA Academic Grant Programme, Google Cloud (Gemma 3 Academic Programme), and Google Research Scholar Award 2025; and Jingchi Liao for supporting parts of the initial data review, statistical analysis, and preliminary data visualization.

## Competing Interests

The authors declare no competing interests.

# Toward Global Large Language Models in Medicine

## Supplementary Information

### Contents

## S1. GlobMed

### S1.1. Data Curation

Our manual review identified three major quality issues within the original data: incompleteness (missing critical information), irrelevance (weak relevance to medicine), and disorganization (inconsistent formatting or structural irregularities), as shown in **Supplementary Fig. 1**. To ensure data reliability, we conducted a multi-stage quality control process on the original datasets, which contained 49,147 entries, resulting in the removal of 3,222 entries that did not meet quality standards. Ambiguous cases were verified by medical experts.

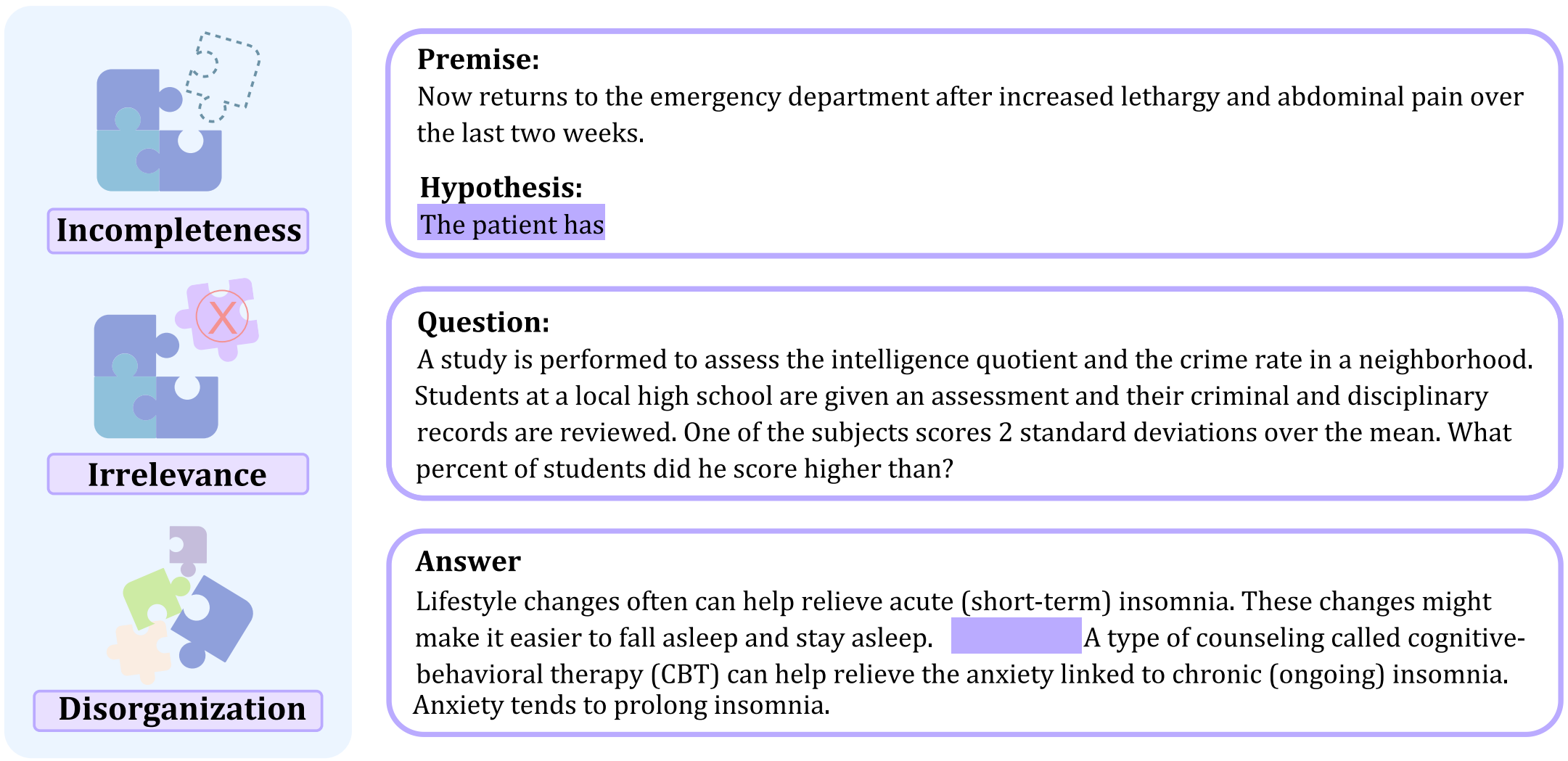

**Supplementary Fig. 1:** Three major quality issues identified during manual review.

### S1.2. Data Statistics

**Supplementary Tab. 1** summarizes the numbers of samples before and after screening in the GlobMed datasets. For datasets with predefined data splits, we generally retained their original splits. Specifically, for BioNLI, we used the original balanced training set and original test set. For MedNLI, HeadQA, MedExpQA, and MedQA, we retained the original training and test sets. For ExpertQA-Bio and ExpertQA-Med, we used all samples from the corresponding ExpertQA domain subsets and assigned all retained samples to the training set. For LiveQA, we used the original training set. For MedMCQA, we used the original validation set as the test set. For MMLU-Pro, we first extracted the health-related subset and then randomly divided it into training and test sets using a 7:3 ratio. For PubMedQA, we used the expert-annotated set of 1,000 samples as the test set.

| Dataset | Train | | Test | | License |
|---|---|---|---|---|---|
| | Original | Screened | Original | Screened | |
| **BioNLI** | 5,544 | 5,540 | 6,308 | 4,450 | CC BY 4.0 |
| **MedNLI** | 11,232 | 11,219 | 1,419 | 1,417 | PhysioNet Credentialed Health Data License 1.5.0 |
| **ExpertQA-Bio** | 96 | 96 | / | / | MIT License |
| **ExpertQA-Med** | 504 | 500 | / | / | MIT License |
| **LiveQA** | 634 | 627 | / | / | Public release for research in medical QA |
| **HeadQA** | 2,657 | 2,164 | 2,742 | 2,218 | MIT License |
| **MedExpQA** | 434 | 434 | 125 | 125 | CC BY 4.0 |
| **MedMCQA** | / | / | 4,183 | 4,081 | MIT License |
| **MedQA** | 10,178 | 10,166 | 1,273 | 1,273 | MIT License |
| **MMLU-Pro** | 573 | 434 | 245 | 187 | MIT License |
| **PubMedQA** | / | / | 1,000 | 994 | MIT License |

**Supplementary Tab. 1:** Statistics on the number of samples in the GlobMed datasets.

**Supplementary Tab. 2–4** report the average token counts across the 20 languages represented in GlobMed for the NLI, long-form QA, and MCQA datasets, respectively. All token counts were calculated using the NLLB-200 tokenizer.

| Average Tokens | BioNLI | | MedNLI | |
|---|---|---|---|---|
| | train | test | train | test |
| **Arabic** | 532 | 560 | / | / |
| **Bengali** | 541 | 577 | / | / |
| **Chinese** | 417 | 443 | 49 | 47 |
| **English** | 406 | 435 | 40 | 39 |
| **French** | 528 | 563 | 56 | 55 |
| **German** | 509 | 545 | 55 | 53 |
| **Hindi** | 499 | 535 | / | / |
| **Indonesian** | 375 | 402 | / | / |
| **Japanese** | 434 | 467 | 49 | 47 |
| **Korean** | 434 | 466 | 49 | 47 |
| **Malay** | 384 | 411 | / | / |
| **Portuguese** | 471 | 499 | 51 | 49 |
| **Russian** | 507 | 537 | / | / |
| **Spanish** | 474 | 504 | 51 | 49 |
| **Swahili** | 470 | 495 | 50 | 49 |
| **Thai** | 560 | 593 | / | / |
| **Urdu** | 513 | 547 | / | / |
| **Wolof** | 477 | 496 | 55 | 53 |
| **Yoruba** | 564 | 584 | 66 | 65 |
| **Zulu** | 510 | 539 | 54 | 53 |

**Supplementary Tab. 2:** Average number of tokens in NLI datasets.

| Average Tokens | ExpertQA-Bio | ExpertQA-Med | LiveQA |
|---|---|---|---|
| **Arabic** | 345 | 324 | 411 |
| **Bengali** | 308 | 294 | 372 |
| **Chinese** | 252 | 241 | 309 |
| **English** | 248 | 240 | 309 |
| **French** | 339 | 324 | 421 |
| **German** | 324 | 317 | 400 |
| **Hindi** | 290 | 277 | 365 |
| **Indonesian** | 229 | 219 | 290 |
| **Japanese** | 247 | 232 | 298 |
| **Korean** | 249 | 238 | 303 |
| **Malay** | 235 | 226 | 299 |
| **Portuguese** | 294 | 281 | 360 |
| **Russian** | 324 | 313 | 408 |
| **Spanish** | 293 | 281 | 360 |
| **Swahili** | 302 | 292 | 368 |
| **Thai** | 354 | 329 | 398 |
| **Urdu** | 305 | 295 | 379 |
| **Wolof** | 341 | 334 | 421 |
| **Yoruba** | 384 | 379 | 471 |
| **Zulu** | 320 | 312 | 392 |

**Supplementary Tab. 3:** Average number of tokens in long-form QA datasets.

| Average Tokens | HeadQA | | MedQA | | MedExpQA | | MedMCQA | | MMLU-Pro | | PubMedQA | |
|---|---|---|---|---|---|---|---|---|---|---|---|---|
| | train | test | train | test | train | test | train | test | train | test | train | test |
| **Arabic** | 122 | 116 | 282 | 289 | 225 | 230 | / | 70 | 202 | 191 | / | 478 |
| **Bengali** | 126 | 119 | 279 | 286 | 232 | 235 | / | 73 | 204 | 194 | / | 450 |
| **Chinese** | 96 | 91 | 223 | 228 | 174 | 176 | / | 56 | 155 | 153 | / | 353 |
| **English** | 98 | 92 | 224 | 230 | 178 | 181 | / | 55 | 158 | 151 | / | 356 |
| **French** | 118 | 112 | 282 | 289 | 215 | 221 | / | 69 | 194 | 188 | / | 468 |
| **German** | 116 | 110 | 271 | 276 | 212 | 217 | / | 66 | 188 | 183 | / | 448 |
| **Hindi** | 117 | 110 | 260 | 266 | 213 | 216 | / | 68 | 189 | 181 | / | 403 |
| **Indonesian** | 90 | 85 | 202 | 207 | 162 | 164 | / | 52 | 145 | 138 | / | 332 |
| **Japanese** | 97 | 91 | 223 | 228 | 173 | 176 | / | 58 | 155 | 152 | / | 354 |
| **Korean** | 95 | 90 | 221 | 227 | 171 | 174 | / | 56 | 153 | 151 | / | 357 |
| **Malay** | 92 | 86 | 208 | 213 | 165 | 167 | / | 53 | 148 | 141 | / | 339 |
| **Portuguese** | 106 | 100 | 247 | 252 | 190 | 194 | / | 62 | 171 | 169 | / | 409 |
| **Russian** | 122 | 115 | 284 | 291 | 227 | 231 | / | 72 | 203 | 191 | / | 464 |
| **Spanish** | 123 | 117 | 249 | 255 | 219 | 224 | / | 63 | 174 | 171 | / | 413 |
| **Swahili** | 109 | 103 | 259 | 266 | 201 | 207 | / | 65 | 184 | 175 | / | 422 |
| **Thai** | 134 | 126 | 291 | 298 | 238 | 246 | / | 81 | 221 | 210 | / | 491 |
| **Urdu** | 119 | 113 | 272 | 278 | 219 | 221 | / | 69 | 195 | 186 | / | 424 |
| **Wolof** | 115 | 109 | 271 | 279 | 205 | 211 | / | 71 | 191 | 181 | / | 459 |
| **Yoruba** | 143 | 135 | 346 | 354 | 265 | 272 | / | 79 | 234 | 219 | / | 507 |
| **Zulu** | 118 | 112 | 280 | 287 | 220 | 226 | / | 72 | 201 | 190 | / | 464 |

**Supplementary Tab. 4:** Average number of tokens in MCQA datasets.

### S1.3. Data Example

*NLI*

**前提:**
七个实验室合作研究两种中等纯度纤溶酶原制剂（64/23，63/6），观察到激活剂（尿激酶或链激酶）的量和纤溶酶原的激活时间影响生成的纤溶酶量。使用酪蛋白和合成多肽（S-2251）作为底物，作者随后证明，在不因纤溶酶自身消化作用而导致活性损失的情况下，很难实现纤溶酶原的完全激活。

**假设:**
通过比较各种 AKT 诱导激活混合物的多肽亚基（在 SDS 电泳）与其纤溶酶活性，可以得出结论：在 AKT 诱导物产生最大量纤溶酶时，一些 AKT 诱导物仍以非活性 AKT 诱导中间体（PLG-i）的形式存在

**答案:** 矛盾

**Supplementary Tab. 5:** NLI data example in Chinese.

**Premise:**
Seven laboratories collaborating in a study of two intermediate purity plasminogen preparations (64/23, 63/6) observed that the amount of activator (urokinase or streptokinase) and the time of activation of plasminogen influenced the amount of plasmin generated. Using casein and a synthetic polypeptide (S-2251) as substrates, the authors subsequently showed that complete activation of plasminogen was difficult to achieve without activity losses due to plasmin autodigestion.

**Hypothesis:**
Comparison of the polypeptide subunits (on SDS electrophoresis) of the various AKT-induced activation mixtures with their plasmin activity allowed the conclusion that at maximum generation of plasmin from AKT-induced, some AKT-induced remains in the form of an inactive AKT-induced intermediate (PLG-i).

**Answer:** Contradiction

**Supplementary Tab. 6:** NLI data example in English.

**Prémisse:**
Sept laboratoires collaborant à une étude sur deux préparations de plasminogène de pureté intermédiaire (64/23, 63/6) ont observé que la quantité d'activateur (urokinase ou streptokinase) et le temps d'activation du plasminogène influençaient la quantité de plasmine générée. En utilisant la caséine et un polypeptide synthétique (S-2251) comme substrats, les auteurs ont par la suite montré qu'une activation complète du plasminogène était difficile à réaliser sans pertes d'activité dues à l'autodigestion de la plasmine.

**Hypothèse:**
La comparaison des sous-unités polypeptidiques (par électrophorèse SDS) des différents mélanges d'activation induits par l'AKT avec leur activité plasmine a permis de conclure que lors de la génération maximale de plasmine à partir du plasminogène induit par l'AKT, une partie du plasminogène induit par l'AKT reste sous forme d'un intermédiaire inactif (PLG-i, plasminogène intermédiaire).

**Réponse:** Contradiction

**Supplementary Tab. 7:** NLI data example in French.

**Prämisse:**
Sieben Labore, die an einer Studie zu zwei Plasminogenpräparaten mittleren Reinheitsgrades (64/23, 63/6) zusammenarbeiteten, beobachteten, dass die Menge der Aktivatorsubstanz (Urokinase oder Streptokinase) und die Aktivierungszeit des Plasminogens die Menge des erzeugten Plasmins beeinflussten. Unter Verwendung von jeweils Casein und einem synthetischen Polypeptid (S-2251) als Substrate zeigten die Autoren anschließend, dass eine vollständige Aktivierung des Plasminogens schwierig zu erzielen war, ohne dass es zu Aktivitätseinbußen durch Plasmin-Auto-Digestion kam.

**Hypothese:**
Der Vergleich der Polypeptid-Untereinheiten (mittels SDS-Elektrophorese) der verschiedenen AKT-induzierten Aktivierungsgemische mit ihrer Plasminaktivität ermöglichte folgende Schlussfolgerung: Bei maximaler Plasminbildung aus AKT-induziertem Plasminogen verbleibt ein Teil des AKT-induzierten Plasminogens in Form eines inaktiven AKT-induzierten Zwischenprodukts (PLG-i, Plasminogen-Intermediat).

**Antwort:** Widerspruch

**Supplementary Tab. 8:** NLI data example in German.

**前提:**
7つの研究室が協力して、2種類の中間純度プラスミノーゲン製剤（64/23、63/6）の研究を行った結果、活性化剤（ウロキナーゼまたはストレプトキナーゼ）の量とプラスミノーゲンの活性化時間が生成されるプラスミンの量に影響を与えることを観察した。カゼインと合成ポリペプチド（S-2251）を基質として用いた後続の研究で、著者らはプラスミンの自己消化による活性の損失なしにプラスミノーゲンの完全な活性化を達成することが困難であることを示した。

**仮説:**
様々なAKT誘導型活性化混合物のポリペプチドサブユニット（SDS電気泳動法による）をそのプラスミン活性と比較した。この比較により、以下の結論が導き出された。AKT誘導型からのプラスミン生成が最大に達した時点でも、一部のAKT誘導型が不活性のAKT誘導型中間体（PLG-i、プラスミノゲン中間体）の形で残っている。

**答え:** 矛盾

**Supplementary Tab. 9:** NLI data example in Japanese.

**전제:**
두 가지 중간 순도의 플라스미노겐 제제(64/23, 63/6)에 대한 연구에 협력한 7개 실험실에서 활성제(우로키나제 또는 스트렙토키나제)의 양과 플라스미노겐의 활성화 시간이 생성된 플라스미의 양에 영향을 미친다는 것을 관찰하였다. 이후 저자들은 카세인과 합성 폴리펩타이드(S-2251)를 기질로 사용하여 연구를 진행하였다. 그 결과, 플라스민의 자가소화(자가분해)로 인한 활성도 손실 없이 플라스미노겐의 완전한 활성화를 달성하기 어렵다는 것을 보여주었다.

**가설:**
다양한 AKT-유도 활성화 혼합물의 폴리펩티드 소단위체(SDS 전기영동법 상에서)를 플라스민 활성과 비교한 결과, AKT-유도로부터 플라스민이 최대로 생성되는 시점에서도 일부 AKT-유도는 비활성 AKT-유도 중간체(PLG-i, 플라스미노겐 중간체) 형태로 남아있다는 결론을 내릴 수 있었다.

**답변:** 모순

**Supplementary Tab. 10:** NLI data example in Korean.

**Premissa:**
Sete laboratórios que colaboraram em um estudo de duas preparações de plasminogênio de pureza intermediária (64/23, 63/6) observaram que a quantidade de ativador (uroquinase ou estreptoquinase) e o tempo de ativação do plasminogênio influenciaram a quantidade de plasmina gerada. Usando caseína e um polipeptídeo sintético (S-2251) como substratos, os autores subsequentemente demonstraram que a ativação completa do plasminogênio era difícil de ser alcançada sem perdas de atividade devido à auto-digestão da plasmina.

**Hipótese:**
A comparação das subunidades polipeptídicas (por eletroforese SDS) das várias misturas de ativação AKT-induzidas com sua atividade de plasmina permitiu concluir que, na geração máxima de plasmina a partir de AKT-induzida, alguma indução por AKT permanece na forma de um intermediário AKT-induzido inativo (PLG-i).

**Resposta:** Contradição

**Supplementary Tab. 11:** NLI data example in Portuguese.

**Premisa:**
Siete laboratorios que colaboraron en un estudio de dos preparaciones de plasminógeno de pureza intermedia (64/23, 63/6) observaron que la cantidad de activador (urocinasa o estreptocinasa) y el tiempo de activación del plasminógeno influían en la cantidad de plasmina generada. Utilizando caseína y un polipéptido sintético (S-2251) como sustratos, los autores posteriormente demostraron que era difícil lograr la activación completa del plasminógeno sin pérdidas de actividad debido a la autodigestión de la plasmina.

**Hipótesis:**
La comparación de las subunidades polipeptídicas (en electroforesis en SDS) de las diversas mezclas de activación AKT-inducida de plasminógeno con su actividad de plasmina permitió concluir que, en el punto de máxima generación de plasmina a partir de AKT-inducida, una parte del AKT-inducido permanece en forma de un intermediario AKT-inducido inactivo (PLG-i).

**Respuesta:** Contradicción

**Supplementary Tab. 12:** NLI data example in Spanish.

**Hoja:**
Maabara saba zilizoshirikiana katika utafiti wa maandalizi mawili ya upeo wa kati wa plasminojeni (64/23, 63/6) zilibaini kwamba kiwango cha kiamsho (urokinasi au streptokinasi) na muda wa kuamsha plasminojeni viliathiri kiwango cha plasmini kilichozalishwa. Kwa kutumia kaseini na polipeptidi ya sintetiki (S-2251) kama substrati, waandishi baadaye walionyesha kwamba uamsho kamili wa plasminojeni ulikuwa mgumu kupatikana bila upotezaji wa utendaji kutokana na kujimeng'enya kwa plasmini.

**Wazo:**
Ulinganisho wa vitengo vidogo vya polipeptidi (kwenye electrophoresis ya SDS) vya mchanganyiko mbalimbali wa AKT-iliyochochewa pamoja na shughuli zao za plasmin uliruhusu hitimisho kwamba wakati wa uzalishaji wa juu wa plasmin kutoka AKT-iliyochochewa, baadhi ya AKT-iliyochochewa inabaki katika muundo wa kati wa AKT-iliyochochewa isiyofanya kazi (PLG-i).

**Jibu:** Kupingana

**Supplementary Tab. 13:** NLI data example in Swahili.

**Digle:**
Juróom ñaari làboratwaar yu bokk ci natt gu ñuy def ci ñaari jumtukaay yu sét yu plasminogen (64/23, 63/6) gis nañu ne limu activateur bi (urokinase walla streptokinase) ak jamono ji ñuy yokk doxiinu plasminogen bi am na solo ci limu plasmin bi ñu meññal. Ñu jëfandikoo casein ak peptide (S-2251) ni substrat yi, bindkat yi mën nañu wone ne yokk doxiinu plasminogen ba mu mat dafa jafe ndax ñàkk doxiin ndax plasmin bi di lekk boppam.

**Njortu:**
Moñ yu polypeptide subunit yi (ci SDS electrophoresis) ci xeeti jaxasé AKT-induced activation ak seen activité plasmin dafa may xalaat ne bu plasmin génnee ba mu eppeeku ci AKT-induced, am na AKT-induced bu des ci xeetu AKT-induced bu taxul dara (PLG-i).

**Tontu:** Dëddu

**Supplementary Tab. 14:** NLI data example in Wolof.

**Ìsọ̀lẹ̀:**
Àwọn ilé-iṣẹ́ tẹ̀sííìgì méje tí wọ́n ṣe ìforúkọsílẹ̀ papọ̀ ní ìwádìí lórí àwọn ìgbékalẹ̀ méjì tí a mọ̀ gẹ́gẹ́ bí intermediate purity plasminogen (64/23, 63/6) ṣe àkíyèsí pé iye activator (urokinase tàbí streptokinase) àti àkókò activation tí plasminogen ni ìpaṣẹ lórí iye plasmin tí a fi lésí. Nípa lílo casein àti synthetic polypeptide (S-2251) gẹ́gẹ́ bí substrates, àwọn onímọ̀ yìí fi hàn lẹ́hìnná pé activation tó yẹ kí ó pé fún plasminogen ṣòro láti ṣe láìsí pípadànù iṣẹ́ nítorí plasmin autodigestion.

**Ìbéèrè:**
Ìfiwéra àwọn abẹṣẹ́ polypeptide (lórí SDS electrophoresis) ti àwọn ìwọ̀n ìdàpọ̀ ìmúnára AKT pẹ̀lú ìṣe plasmin wọn fàyè gba ìparí pé ní ìpele àgbàyégbà ìpilẹ̀ṣẹ̀ plasmin láti AKT-induced, díẹ̀ nínú AKT-induced ṣì wà ní ìrísí ààrín-gbùngbùn AKT-induced tí kò ṣiṣẹ́ (PLG-i).

**Ìdáhùn:** Ìlòdì

**Supplementary Tab. 15:** NLI data example in Yoruba.

**Isitatimende:**
Amalabhorethri ayisikhombisa abebambisene ocwaningweni lwezinhlobo ezimbili ezingangokuphelele zokuhlelwa kweplasminogen (64/23, 63/6) babona ukuthi inani lama-activator (i-urokinase noma i-streptokinase) kanye nesikhathi sokusebenza kweplasminogen kuye kube nomthelela kwinani leplasmin elakhiwayo. Ngokusebenzisa i-casein kanye ne-synthetic polypeptide (S-2251) njengezinto eziyisisekelo, ababhali kamuva babonisa ukuthi ukusebenza okuphelele kweplasminogen bekunzima ukufinyeleleka ngaphandle kokulahlekelwa wumsebenzi ngenxa ye-autodigestion yeplasmin.

**Umqondo:**
Ukuqhathaniswa kwamayunithi amancane epholipeptidi (ku-SDS electrophoresis) inhlanganisela ezahlukene eziqaliswe yi-AKT nezenzo zazo ze-plasmin kuvumele isiphetho sokuthi ekukhiqizweni okukhulu kwe-plasmin esuka kwi-AKT-induced, i-AKT-induced ethile isasele esimweni se-AKT-induced intermediate engasebenzi (PLG-i).

**Impendulo:** Ukuphikisana

**Supplementary Tab. 16:** NLI data example in Zulu.

*Long-form QA*

**问题:**
一位 6 个月大的患儿由母亲带来就诊，母亲报告孩子生长发育迟缓，以及牙列异常、反复呕吐、动作不协调和难以控制的癫痫发作。体格检查显示患儿有脂溢性皮炎、小头畸形和下颌异常。根据患者表现的临床数据，患者呈现的是什么综合征？

**答案:**
根据所呈现的临床数据，患者似乎患有西症候群 (West Syndrome)，也称为婴儿痉挛症 (Infantile Spasms)。西症候群是一种罕见的癫痫性障碍，特征包括生长迟缓、动作不协调、难以控制的发作（婴儿痉挛）、智力障碍和其他症状。脂溢性皮炎、小头症和突出的下巴也可能是该综合征的指示性特征。

**Supplementary Tab. 17:** Long-Form QA data example in Chinese.

**Question:**
A 6-month-old patient is taken to consult by his mother, who reports delayed growth in her son, as well as changes in the dentition, recurrent vomiting, uncoordinated movements, and difficult-to-control seizures. On physical examination, he is noted to have seborrheic skin, microcephaly, and a prominent jaw. Based on the clinical data presented by the patient, what syndrome does the patient present?

**Answer:**
Based on the clinical data presented, the patient appears to have West Syndrome, also known as Infantile Spasms. West Syndrome is a rare epileptic disorder characterized by delayed growth, uncoordinated movements, difficult-to-control seizures (infantile spasms), intellectual disability, and other symptoms. Seborrheic skin, microcephaly, and a prominent jaw may also be indicative of this syndrome.

**Supplementary Tab. 18:** Long-Form QA data example in English.

**Question:**
Un patient de 6 mois est amené en consultation par sa mère, qui signale un retard de croissance, ainsi que des changements dans la dentition, des vomissements récurrents, des mouvements non coordonnés et des convulsions difficiles à contrôler. À l'examen physique, on note une peau séborrhéique, une microcéphalie et une mâchoire proéminente. Sur la base des données cliniques présentées par le patient, quel syndrome présente-t-il ?

**Réponse:**
D'après les données cliniques présentées, le patient semble atteint du syndrome de West, également connu sous le nom de spasmes infantiles (Infantile Spasms). Le syndrome de West est un trouble épileptique rare caractérisé par un retard de croissance, des mouvements non coordonnés, des crises épileptiques difficiles à maîtriser (spasmes infantiles), une déficience intellectuelle et d'autres symptômes. Une peau séborrhéique, une microcéphalie et une mâchoire proéminente peuvent également être indicateurs de ce syndrome.

**Supplementary Tab. 19:** Long-Form QA data example in French.

**Frage:**
Ein 6 Monate alter Patient wird von seiner Mutter zur Untersuchung gebracht, die von einer Wachstumsverzögerung ihres Sohnes berichtet, sowie Veränderungen des Zahnstatus, rezidivierendem Erbrechen, unkoordinierten Bewegungen und schwer kontrollierbaren Anfällen. Bei der körperlichen Untersuchung fallen eine seborrhoische Haut, Mikrozephalie und ein vorspringender Kiefer auf. Basierend auf den klinischen Daten des Patienten, welches Syndrom weist der Patient auf?

**Antwort:**
Basierend auf den vorliegenden klinischen Daten scheint der Patient an West-Syndrom zu leiden, auch bekannt als Infantile Spasmen. Das West-Syndrom ist eine seltene epileptische Erkrankung, die durch verzögertes Wachstum, unkoordinierte Bewegungen, schwer zu kontrollierende Anfälle (Infantile Spasmen), geistige Behinderung und andere Symptome gekennzeichnet und charakterisiert ist. Seborrhoische Dermatitis, Mikrozephalie und ein prominenter Kiefer können ebenfalls auf dieses Syndrom hinweisen.

**Supplementary Tab. 20:** Long-Form QA data example in German.

**質問:**
6 ヶ月の患者が母親に連れられて受診しました。母親は息子の成長遅延、歯列の変化、再発性の嘔吐、協調運動障害、制御困難な発作について報告しています。身体診察では、脂漏性の皮膚、小頭症、そして下顎前突が認められました。これらの臨床所見に基づいて、患者はどの症候群を呈していますか?

**答え:**
提示された臨床データに基づくと、患者はウエスト症候群（別名：点頭てんかん（Infantile Spasms））を有しているように見受けられます。ウエスト症候群は、発育遅延、協調運動障害、制御が困難な発作（点頭発作）、知的障害、およびその他の症状を特徴とする稀なてんかん性障害です。脂漏性皮膚、小頭症、および突出した顎も、この症候群を示唆する可能性があります。

**Supplementary Tab. 21:** Long-Form QA data example in Japanese.

**질문:**
6 개월 된 환자가 어머니와 함께 진료를 받으러 왔습니다. 어머니는 아들의 발육 지연과 함께 치열의 변화, 재발성 구토, 협응되지 않은 움직임, 그리고 조절이 어려운 발작 증상을 보고합니다. 진찰 시 지루성 피부염, 소두증, 그리고 비정상적으로 돌출된 턱이 관찰됩니다. 환자가 나타내는 임상 데이터를 바탕으로, 환자가 나타내는 증후군은 무엇입니까?

**답변:**
제시된 임상 데이터를 바탕으로, 환자는 웨스트 증후군(West Syndrome), 또는 영아연축(Infantile Spasms)으로도 알려진 질환을 가지고 있는 것으로 보입니다. 웨스트 증후군은 성장 지연, 협응 운동 장애, 조절하기 어려운 발작(영아연축), 지적 발달 장애 및 기타 증상을 특징으로 하는 희귀 뇌전증성 장애입니다. 지루성 피부염, 소두증, 그리고 돌출된 턱 또한 이 증후군의 징후일 수 있습니다.

**Supplementary Tab. 22:** Long-Form QA data example in Korean.

**Pergunta:**
Um paciente de 6 meses é levado à consulta pela mãe, que relata atraso no crescimento do filho, bem como alterações na dentição, vômitos recorrentes, movimentos descoordenados e convulsões de difícil controle. No exame físico, observa-se dermatite seborreica, microcefalia e mandíbula protuberante. Com base nos dados clínicos apresentados pelo paciente, qual síndrome o paciente apresenta?

**Resposta:**
Com base nos dados clínicos apresentados, o paciente parece ter Síndrome de West, também conhecida como Espasmos Infantis. A Síndrome de West é um distúrbio epiléptico raro caracterizado por atraso no crescimento, movimentos descoordenados, crises convulsivas de difícil controle (espasmos infantis), deficiência intelectual e outros sintomas. Pele seborreica, microcefalia e mandíbula proeminente podem também ser indicativos desta síndrome.

**Supplementary Tab. 23:** Long-Form QA data example in Portuguese.

**Pregunta:**
Un paciente de 6 meses es llevado a consulta médica por su madre, quien informa retraso en el crecimiento del paciente, así como cambios en la dentición, vómito recurrente, movimientos descoordinados y convulsiones de difícil control. En la exploración física, se observa piel seborreica, microcefalia y mandíbula prognática. Basándose en los datos clínicos presentados por el paciente, ¿qué síndrome presenta el paciente?

**Respuesta:**
Basándose en los datos clínicos presentados, el paciente parece tener Síndrome de West, también conocido como Espasmos Infantiles. El Síndrome de West es un trastorno epiléptico raro caracterizado por retraso en el crecimiento, movimientos descoordinados, convulsiones difíciles de controlar (conocidas como espasmos infantiles), discapacidad intelectual y otros síntomas, entre otros. La piel seborreica, la microcefalia y una mandíbula prominente también pueden ser indicadores de este síndrome.

**Supplementary Tab. 24:** Long-Form QA data example in Spanish.

**Swali:**
Mgonjwa wa miezi 6 anapelekwa kuonana na daktari na mama yake, ambaye anaripoti kuchelewa kukua kwa mwanawe, pamoja na mabadiliko katika meno, kutapika mara kwa mara, mwendo usiokuwa na uratibu, na kifafa kisichoweza kudhibitiwa kwa urahisi. Katika uchunguzi wa kimwili, anaonekana kuwa na ngozi yenye magamba, kichwa kidogo, na taya iliyovimba. Kulingana na data za kitabibu zinazoonyeshwa na mgonjwa, ni mfumo wa dalili gani mgonjwa anaonyesha?

**Jibu:**
Kulingana na data za kitabibu zilizowasilishwa, mgonjwa anaonekana kuwa na Sindrome ya West, pia inajulikana kama Misukosuko ya Watoto Wachanga. Sindrome ya West ni hali adimu ya kifafa inayojulikana kwa ukuaji wa kuchelewa, mienendo isiyoratibiwa, misukosuko isiyodhibitiwa kwa urahisi (misukosuko ya watoto wachanga), ulemavu wa akili na kuzuiwa kwa ukuaji wa akili, na dalili zingine. Ngozi yenye mafuta mengi (seborrhea), utasa wa kichwa (kichwa kidogo kuliko kawaida), na taya inayojitokeza pia zinaweza kuashiria dalili za sindrome hii.

**Supplementary Tab. 25:** Long-Form QA data example in Swahili.

**Laaj:**
Ndey liir bu am juróom benn weer indi na doom am bu góor ci doktoor bi, ndax mu wax ne doom ji dafa yéex ci màgg, ak yëngu-yëngu yu baaxul ci bëñ yi, xëb xëb bu bari, yëngu-yëngu yu amul takku, ak jaxase yu jafe-jafe yu kenn mënul dajale. Bi ñu ko seetee, gis nañu deram bi dafa mel ni lu am diw bu bari, bopp bi dafa tuuti (mikrosefali), ak yaxu gémmiñ gu gëna yaatu. Bu fekkee ni nga seet xaaraay yi liir bi am, ban xeeti feebar la liir bi am?

**Tontu:**
Ci li ñu gis ci baatukaay bi, feebar bi mëna nekk Syndrome West, walla Perlu yu Ndaw. Syndrome West dafa nekk feebar bu ñaaw bu xeeti ci màggate bu yées, yëngatu yu amul topplante, perlu yu metti ñu faj (perlu yu ndaw), ñàkk xam-xam, ak yeneen balaawaan. Dërëm bu teel sew, bopp bu ndaw (microcéphalie), ak sàggay bu génn it mën na nekk ci seeni màndarga.

**Supplementary Tab. 26:** Long-Form QA data example in Wolof.

**Ìbéèrè:**
Ọmọ oṣu mẹfa ni a mu lọ si ayẹwo lati ọwọ iya rẹ, ti o sọ pe idagbasoke ọmọ rẹ ọkunrin ti lọra, pẹlu awọn iyipada ni eto eyín, igbogbo-igbogbo èébì, awọn ìṣísẹ̀ ara ti ko ni ibamu, ati awọn igbẹwọ ti o ṣoro lati ṣakoso. Ni ayẹwo ara, wọn ṣe akiyesi pe o ni awọ ara seborrheic ti ko dara, ori kekere (microcephaly), ati ẹyin-agbọn ti ko ni ipo to yẹ. Ni ipilẹ awọn alaye aisan ti aisan naa fihan, irú àìsàn pataki wo ni o dabi pe alaisan yii ni?

**Ìdáhùn:**
Nípasẹ̀ àwọn ìwádìí ìṣègùn tí a ṣe, ó dàbí pé aláìsàn náà ní West Syndrome (Àìsàn West), tí a tún mọ̀ sí Infantile Spasms (Ìgbàgbẹ́ Ọmọdé). West Syndrome jẹ́ àìsàn ìgbàgbẹ́ tó jẹ́ ọ̀kan lára àwọn àìsàn tó ṣẹlẹ̀ lọ́ọ̀kọ̀ọ̀kan, tí ó ní àwọn àmì bíi ìdàgbàsókè tó lọra, àìní ìdarí ara, ìgbàgbẹ́ tó ṣòro láti dá dúró (infantile spasms), àìlè ronú dáadáa, àti àwọn àmì mìíràn. Awọ̀ ara tó ní ọ̀rá púpọ̀ (seborrheic skin), orí kékeré (microcephaly), àti erẹ̀kẹ́ tó hàn gbangba lè jẹ́ àmì fún àìsàn yìí.

**Supplementary Tab. 27:** Long-Form QA data example in Yoruba.

**Umbuzo:**
Isiguli sezinyanga eziyisithupha sithathwa ukuyobonana nodokotela unina, obika ukukhula kancane kwengane yakhe yomfana, kanye noguquko emazinyweni, ukuhlanza okuphindaphindayo, ukungahambelani kweminyakazo yomzimba, kanye nokuphathwa izifo zokuwa okunzima ukuzilawula. Ekuhlolweni komzimba kwengane, kuyabonakala ukuthi unesifo sesikhumba esibizwa nge-seborrheic, ikhanda elincane, kanye nomhlathi ongajwayelekile. Ngokususela kumininingwane yezempilo evezwe yisiguli, yisiphi isimo esihlangene esivezwa yisiguli?

**Impendulo:**
Ngokususela kwi-data yezokunakekelwa kweziguli ephakanyisiwe, kubonakala ukuthi isiguli sinesifo esibizwa ngokuthi yi-West Syndrome, esaziwa futhi ngokuthi yi-Infantile Spasms (ukudlidliza okuthile kwezingane ezincane). I-West Syndrome iyisifo esiyingcosana sokuwa esichazwa ngokukhula okuhlehlayo, ukunyakaza okungahlelekile, ukudlidliza okungalawuleki (ukudlidliza okukhethekile kwezingane ezincane), ukungakwazi ukucabanga kahle, kanye nezinye izimpawu. Isikhumba esineseborrhoea (esinamafutha amaningi), ikhanda elincane (i-microcephaly), kanye nomhlathi oqavile kungase kube izimpawu zalesi sifo esiyingqayizivele.

**Supplementary Tab. 28:** Long-Form QA data example in Zulu.

*MCQA*

**问题:**
一位 23 岁的孕妇，妊娠 22 周，出现排尿时灼痛。她表示这种症状始于 1 天前，并且尽管增加饮水量和服用蔓越莓提取物，症状仍在恶化。除此之外，她感觉良好，并由医生定期进行产前检查。她的体温为 97.7°F（36.5°C），血压为 122/77 毫米汞柱，脉搏为 80 次/分钟，呼吸频率为 19 次/分钟，室内空气下氧饱和度为 98%。体格检查显示无肋脊角压痛，子宫增大。以下哪项是该患者的最佳治疗方案？

**选项:**
A: 氨苄青霉素
B: 头孢曲松
C: 多西环素
D: 呋喃妥因

**答案:** 呋喃妥因

**Supplementary Tab. 29:** MCQA data example in Chinese.

**Question:**
A 23-year-old pregnant woman at 22 weeks' gestation presents with burning upon urination. She states it started 1 day ago and has been worsening despite drinking more water and taking cranberry extract. She otherwise feels well and is followed by a doctor for her pregnancy. Her temperature is 97.7°F (36.5°C), blood pressure is 122/77 mmHg, pulse is 80/min, respirations are 19/min, and oxygen saturation is 98% on room air. Physical exam is notable for an absence of costovertebral angle tenderness and a gravid uterus. Which of the following is the best treatment for this patient?

**Options:**
A: Ampicillin
B: Ceftriaxone
C: Doxycycline
D: Nitrofurantoin

**Answer:** Nitrofurantoin

**Supplementary Tab. 30:** MCQA data example in English.

**Question:**
Une femme enceinte de 23 ans à 22 semaines de grossesse se présente avec des brûlures urinaires. Elle déclare que cela a commencé il y a 1 jour et s'est aggravé malgré une augmentation de sa consommation d'eau et la prise d'extrait de canneberge. Elle se sent par ailleurs bien et est suivie par un médecin pour sa grossesse. Sa température est de 36,5°C, sa tension artérielle est de 122/77 mmHg, son pouls est de 80/min, sa fréquence respiratoire est de 19/min et sa saturation en oxygène est de 98% à l'air ambiant. L'examen physique est remarquable par l'absence de sensibilité de l'angle costo-vertébral et un utérus gravide palpable. Quel est le meilleur traitement pour cette patiente?

**Options:**
A: Ampicilline
B: Ceftriaxone
C: Doxycycline
D: Nitrofurantoïne

**Réponse:** Nitrofurantoïne

**Supplementary Tab. 31:** MCQA data example in French.

**Frage:**
Eine 23-jährige schwangere Frau in der 22. Schwangerschaftswoche stellt sich mit Brennen beim Urinieren vor. Sie gibt an, dass es vor 1 Tag begann und sich trotz erhöhter Wasseraufnahme und Einnahme von Cranberry-Extrakt verschlimmert hat. Ansonsten fühlt sie sich gut und wird von einem Arzt bezüglich ihrer Schwangerschaft betreut. Ihre Temperatur beträgt 97,7°F (36,5°C), der Blutdruck liegt bei 122/77 mmHg, der Puls bei 80/min, die Atmung bei 19/min und die Sauerstoffsättigung bei 98% unter Raumluft. Die körperliche Untersuchung zeigt keine Druckschmerzhaftigkeit im Costovertebralwinkel und einen graviden Uterus. Welche der folgenden Behandlungen ist für diese Patientin am besten geeignet?

**Optionen:**
A: Ampicillin
B: Ceftriaxon
C: Doxycyclin
D: Nitrofurantoin

**Antwort:** Nitrofurantoin

**Supplementary Tab. 32:** MCQA data example in German.

**質問**:
妊娠 22 週の 23 歳の妊婦が排尿時痛を訴えて来院しました。彼女は 1 日前から症状が始まり、水分摂取量を増やしクランベリーエキスを摂取しているにもかかわらず悪化していると述べています。それ以外は体調は良好で、妊娠については医師によるフォローアップを受けています。彼女の体温は 36.5°C、血圧は 122/77 mmHg、脈拍は 80/分、呼吸数は 19/分、室内気での酸素飽和度は 98%です。身体診察では、肋骨脊椎角の圧痛がないことと、妊娠子宮が認められます。この患者に対する最適な治療は以下のうちどれですか?

**選択肢**:
A: アンピシリン
B: セフトリアキソン
C: ドキシサイクリン
D: ニトロフラントイン

**答え:** ニトロフラントイン

**Supplementary Tab. 33:** MCQA data example in Japanese.

**질문**:
임신 22 주차인 23 세 임산부가 배뇨통을 호소하며 내원했습니다. 이 증상이 1 일 전에 시작되었고 물을 더 많이 마시고 크랜베리 추출물을 복용했음에도 악화되고 있다고 합니다. 그 외에는 전반적으로 건강한 상태이며 임신 관련 의사의 관리를 받고 있습니다. 체온은 97.7°F (36.5°C, 섭씨), 혈압은 122/77 mmHg (수은주 밀리미터), 맥박은 80 회/분, 호흡수는 19 회/분, 그리고 실내 공기에서의 산소포화도는 98%입니다. 신체 검사상 늑골척추각 압통은 없으며 임신자궁이 관찰됩니다. 이 환자에게 가장 적절한 치료는 다음 중 무엇입니까?

**옵션**:
A: 앰피실린
B: 세프트리악손
C: 독시사이클린
D: 니트로푸란토인

**답변:** 니트로푸란토인

**Supplementary Tab. 34:** MCQA data example in Korean.

**Pergunta**:
Uma mulher grávida de 23 anos com 22 semanas de gestação apresenta ardência ao urinar. Ela afirma que começou há 1 dia e tem piorado apesar de beber mais água e tomar extrato de cranberry. De resto, sente-se bem e é acompanhada por um médico durante a gravidez. Sua temperatura é de 36,5°C, a pressão arterial é 122/77 mmHg, o pulso é 80/min, as respirações são 19/min e a saturação de oxigênio é 98% em ar ambiente. O exame físico é notável pela ausência de dor à percussão do ângulo costovertebral e um útero gravídico. Qual das seguintes opções é o melhor tratamento para esta paciente?

**Opções**:
A: Ampicilina
B: Ceftriaxona
C: Doxiciclina
D: Nitrofurantoína

**Resposta:** Nitrofurantoína

**Supplementary Tab. 35:** MCQA data example in Portuguese.

**Pregunta:**
Una mujer embarazada de 23 años con 22 semanas de gestación se presenta con ardor al orinar. Ella afirma que comenzó hace 1 día y ha estado empeorando a pesar de beber más agua y tomar extracto de arándano. Por lo demás, se siente bien y está siendo controlada por un médico durante su embarazo. Su temperatura es de 97.7°F (36.5°C), la presión arterial es de 122/77 mmHg, el pulso es de 80/min, las respiraciones son de 19/min y la saturación de oxígeno es del 98% en aire ambiente. El examen físico es notable por la ausencia de dolor a la palpación en el ángulo costovertebral y un útero gestante. ¿Cuál de los siguientes es el mejor tratamiento para esta paciente?

**Opciones:**
A: Ampicilina
B: Ceftriaxona
C: Doxiciclina
D: Nitrofurantoína

**Respuesta:** Nitrofurantoína

**Supplementary Tab. 36:** MCQA data example in Spanish.

**Swali:**
Mwanamke mjamzito mwenye umri wa miaka 23 akiwa na ujauzito wa wiki 22 anaripoti kuhisi maumivu wakati wa kukojoa. Anasema ilianza siku 1 iliyopita na imeendelea kuwa mbaya licha ya kunywa maji zaidi na kutumia dawa ya cranberry. Kwa vyovyote anaendelea kujisikia vizuri na anafuatiliwa na daktari kwa ujauzito wake. Joto lake la mwili ni 97.7°F (36.5°C), shinikizo la damu ni 122/77 mmHg, mapigo ya moyo ni 80/dakika, upumuaji ni 19/dakika, na ushibaji wa oksijeni ni 98% katika hewa ya kawaida. Uchunguzi wa mwili unaonyesha kutokuwepo kwa maumivu kwenye pembe ya uti wa mgongo na figo (costovertebral angle) na tumbo la ujauzito. Ni matibabu gani kati ya yafuatayo yanayofaa zaidi kwa mgonjwa huyu?

**Chaguzi:**
A: Ampicillin
B: Ceftriaxone
C: Doxycycline
D: Nitrofurantoin

**Jibu:** Nitrofurantoin

**Supplementary Tab. 37:** MCQA data example in Swahili.

**Laaj:**
Jigéen bu am 23 at bu ëmb bu nekk ci 22 ayubés gis na ñu ko ndax lakk bi muy yég saa su santaan. Wax na ni lakk bi tàmbali na bëy 1 bes te di gëna metti doonte mu nàn na ndox bu bari ak di jëfandikoo tànku cranberry. Moo ti dara jakkaaraluko ko te ab doktoor di ko ubbi. Tamperatuur bi 97.7°F (36.5°C) la, tànsiyo bi 122/77 mmHg, puls bi 80/min, noyyi gi 19/min, te oksijeen bi 98% la ci ngelaw mi. Saytu jëmm ji dafa wone ni amul metit ci costovertebral angle bi te biir bi dafa diis. Ban ci fàppkay yi ñu tëral moo gën ci faj gi jaambur bi?

**Tànneefi:**
A: Ampicillin
B: Ceftriaxone
C: Doxycycline
D: Nitrofurantoin

**Tontu:** Nitrofurantoin

**Supplementary Tab. 38:** MCQA data example in Wolof.

**Ìbéèrè:**
Obìnrin tó lóyún tó jẹ́ ọmọ ọdún mẹ́tàlélógún (23) tí oyún rẹ̀ ti tó ọ̀sẹ̀ méjìlélógún (22) wá pẹ̀lú ìrora tí ó ń gbóná nígbà tí ó bá ń tọ̀. Ó sọ pé ó bẹ̀rẹ̀ ní ọjọ́ kan sẹ́yìn tí ó sì ti ń burú sí i láìsí bí òun ti ń mu omi pọ̀ tí òun sì ti ń lo òògùn cranberry extract. Yàtọ̀ sí èyí, ara rẹ̀ dá òun tí dókítà sì ń mojútó oyún rẹ̀. Ìgbóná ara rẹ̀ jẹ́ 97.7°F (36.5°C), ìfunpa rẹ̀ jẹ́ 122/77 mmHg, ìlù ọkàn rẹ̀ jẹ́ 80/dákíkà, ìmísì rẹ̀ jẹ́ 19/dákíkà, àti ìwọ̀n oxygen saturation rẹ̀ jẹ́ 98% ní gbangba. Àyẹ̀wò ara rẹ̀ fi hàn pé kò sí costovertebral angle tenderness àti pé ilé ọmú rẹ̀ ti tòbi nítorí oyún. Nínú àwọn wọ̀nyí, èwo ni ó jẹ́ ìtọ́jú tó dára jùlọ fún aláìsàn yìí?

**Àwọn àṣàyàn:**
A: Ampicillin
B: Ceftriaxone
C: Doxycycline
D: Nitrofurantoin

**Ìdáhùn:** Nitrofurantoin

**Supplementary Tab. 39:** MCQA data example in Yoruba.

**Umbuzo:**
Owesifazane okhulelwe oneminyaka engu-23 ubudala osemasonteni angu-22 ekhulelwe ufika enokusha uma enza umshobingo. Uthi kuqale izolo futhi kuya ngokuba kubi nakuba ephuza amanzi amaningi futhi ethatha okuthi cranberry extract. Ngaphandle kwalokho uzizwa ephilile futhi ulandelwa udokotela ngokukhulelwa kwakhe. Izinga lokushisa komzimba wakhe kungu-97.7°F (36.5°C), umfutho wegazi ungu-122/77 mmHg, inhliziyo ishaya ku-80/min, ukuphefumula kungu-19/min, futhi ukugcwala kwe-oxygen kungamaphesenti angu-98% emoyeni wasekamelweni. Ukuhlolwa komzimba kukhombisa ukungabikho kwezinhlungu zesiphambano semisipha nomgogodla kanye nesisu esikhulelwe. Yikuphi kulokhu okulandelayo okungukwelashwa okungcono kwalesi siguli?

**Izinketho:**
A: I-Ampicillin
B: I-Ceftriaxone
C: I-Doxycycline
D: I-Nitrofurantoin

**Impendulo:** I-Nitrofurantoin

**Supplementary Tab. 40:** MCQA data example in Zulu.

### S1.4. Agentic Machine Translation Prompt

*NLI*

**System Message:**
You are a professional translator specializing in accurate translation of medical content from {source_lang} to {target_lang}.

**Your task is to translate the medical text. Follow these requirements:**

1. Preserve all numerals, decimal points, scientific notation, comparison signs, percentages, and brackets exactly as written (e.g., 2.5, µg/ml, 100%, [3H] leucine). Do NOT localize decimal separators or convert units/currencies.

2. Keep biochemical/clinical entities and abbreviations (e.g., NGF, tyrosine hydroxylase, ELISA, mRNA) exactly as in the source unless the target language has a well-established localized common name (rare). If localized, do NOT add the original in parentheses.

3. Preserve LaTeX/math/code verbatim; do not translate or alter content inside LaTeX/code delimiters.

4. Maintain tone and register appropriate for professional medical assessments. Avoid explanatory additions.

**Task:**
Please translate the following NLI item:

<SOURCE_TEXT>
{source_text}
</SOURCE_TEXT>

**Output:**
Only provide the {target_lang} translation for the above text. Do not include any explanations or text apart from the translation.

Return ONLY a single JSON object with the exact keys: "premise" and "hypothesis". Do not include any explanations or text outside the JSON. Do NOT add extra keys or metadata. All JSON keys must remain in English exactly as shown. Only translate the content inside square brackets.

<TRANSLATION>
{
"premise": "[translation of premise]",
"hypothesis": "[translation of hypothesis]"
}
</TRANSLATION>

**Supplementary Tab. 41:** Agentic machine translation prompt for the NLI task (Initial Translation).

**System Message:**
You are a medical translation expert, specializing in translation from {source_lang} to {target_lang}.

**Task Description:**
Carefully review the source text and its translation from {source_lang} to {target_lang}, and then provide constructive suggestions in English.

**Requirements:**
1. No additions, removals, or explanations of domain content.

2. Preserve anonymization and any specialized placeholders exactly (e.g., masked IDs, [***], {VAR}).

3. Numerals & notation: verify all numbers, decimal points, scientific notation, comparison signs, percentages, brackets, and Greek letters are unchanged; DO NOT localize decimal separators or convert currencies/units.

4. Units & symbols: confirm μg/ml, %, ×, ±, →, ≥/≤, and SI prefixes/symbols are preserved exactly (no written-out forms).

5. Biomedical entities/abbreviations (e.g., NGF, tyrosine hydroxylase, ELISA, mRNA): ensure they remain as in the source unless a well-established {target_lang} equivalent exists; if localized, the original term MUST NOT be added in parentheses; check consistent capitalization.

6. Field integrity: ensure exactly the two fields are present and correctly mapped –"premise": translated premise text –"hypothesis": translated hypothesis text.

7. Tone/register: maintain professional assessment style; remove any explanatory additions or commentary.

**Input:**

<SOURCE_TEXT>
{source_text}
</SOURCE_TEXT>

<INITIAL_TRANSLATION>
{initial_trans}
</INITIAL_TRANSLATION>

**Output:**

<SUGGESTIONS>
[Your suggestions here]
</SUGGESTIONS>

**Supplementary Tab. 42:** Agentic machine translation prompt for the NLI task (Reflection).

**System Message:**
You are a senior medical NLI translation reviser, specializing in translation from {source_lang} to {target_lang}.

**Task Description:**
Carefully review and edit the NLI translation from {source_lang} to {target_lang}, incorporating the expert feedback below.

**Requirements:**
1. DO NOT explain anything; produce the improved translation JSON only.

2. Preserve every single quote from the source; do not add new single or double quotes.

3. Remove unnecessary explanations or original terms from {source_lang} if present in the translation.

**Input:**

<SOURCE_TEXT>
{source_text}
</SOURCE_TEXT>

<INITIAL_TRANSLATION>
{initial_trans}
</INITIAL_TRANSLATION>

<EXPERT_SUGGESTIONS>
{reflection}
</EXPERT_SUGGESTIONS>

**Output:**
Only provide the {target_lang} translation for the above text. Do not include any explanations or text apart from the translation.

Return ONLY a single JSON object with the exact keys: "premise" and "hypothesis". Do not include any explanations or text outside the JSON. Do NOT add extra keys or metadata. All JSON keys must remain in English exactly as shown. Only translate the content inside square brackets.

<IMPROVED_TRANSLATION>
{
"premise": "[improved translation of premise]",
"hypothesis": "[improved translation of hypothesis]"
}
</IMPROVED_TRANSLATION>

**Supplementary Tab. 43:** Agentic machine translation prompt for the NLI task (Improved Translation).

*Long-Form QA*

**System Message:**
You are a professional translator specializing in accurate translation of medical content from {source_lang} to {target_lang}.

**Your task is to translate the medical text. Follow these requirements:**

1. Preserve all numerals, decimal points, scientific notation, comparison signs, percentages, and brackets exactly as written (e.g., 2.5, μg/ml, 100%, [3H] leucine). Do NOT localize decimal separators or convert units/currencies.

2. Keep biochemical/clinical entities and abbreviations (e.g., NGF, tyrosine hydroxylase, ELISA, mRNA) exactly as in the source unless the target language has a well-established localized common name (rare). If localized, do NOT add the original in parentheses.

3. Preserve LaTeX/math/code verbatim; do not translate or alter content inside LaTeX/code delimiters.

4. Preserve the complete meaning and paragraph structure of both the question and the answer. Do not summarize, omit, or add any information. Maintain tone and register appropriate for professional medical assessments. Avoid explanatory additions.

**Task:**
Please translate the following long-form QA item:

<SOURCE_TEXT>
{source_text}
</SOURCE_TEXT>

**Output:**
Only provide the {target_lang} translation for the above text. Do not include any explanations or text apart from the translation.

Return ONLY a single JSON object with the exact keys: "question" and "answer". Do not include any explanations or text outside the JSON. Do NOT add extra keys or metadata. All JSON keys must remain in English exactly as shown. Only translate the content inside square brackets.

<TRANSLATION>
{
"question": "[translation of question]",
"answer": "[translation of answer]"
}
</TRANSLATION>

**Supplementary Tab. 44:** Agentic machine translation prompt for the long-form QA task (Initial Translation).

**System Message:**
You are a medical translation expert, specializing in translation from {source_lang} to {target_lang}.

**Task Description:**
Carefully review the source text and its translation from {source_lang} to {target_lang}, and then provide constructive suggestions in English.

**Requirements:**
1. No additions, removals, summarization, or explanations of domain content.

2. Preserve anonymization and any specialized placeholders exactly (e.g., masked IDs, [***], {VAR}).

3. Numerals & notation: verify all numbers, decimal points, scientific notation, comparison signs, percentages, brackets, and Greek letters are unchanged; DO NOT localize decimal separators or convert currencies/units.

4. Units & symbols: confirm μg/ml, %, ×, ±, →, ≥/≤, and SI prefixes/symbols are preserved exactly (no written-out forms).

5. Biomedical entities/abbreviations (e.g., NGF, tyrosine hydroxylase, ELISA, mRNA): ensure they remain as in the source unless a well-established {target_lang} equivalent exists; if localized, the original term MUST NOT be added in parentheses; check consistent capitalization.

6. Field integrity: ensure exactly the two fields are present and correctly mapped –"question": translated question text –"answer": translated answer text. Ensure that the complete meaning and paragraph structure of both fields are preserved.

7. Tone/register: maintain professional medical style; remove any explanatory additions or commentary.

**Input:**

<SOURCE_TEXT>
{source_text}
</SOURCE_TEXT>

<INITIAL_TRANSLATION>
{initial_trans}
</INITIAL_TRANSLATION>

**Output:**

<SUGGESTIONS>
[Your suggestions here]
</SUGGESTIONS>

**Supplementary Tab. 45:** Agentic machine translation prompt for the long-form QA task (Reflection).

**System Message:**
You are a senior medical long-form QA translation reviser, specializing in translation from {source_lang} to {target_lang}.

**Task Description:**
Carefully review and edit the long-form QA translation from {source_lang} to {target_lang}, incorporating the expert feedback below.

**Requirements:**
1. DO NOT explain anything; produce the improved translation JSON only.

2. Preserve every single quote from the source; do not add new single or double quotes.

3. Preserve the complete meaning and paragraph structure of both the question and the answer. Do not summarize, omit, or add any information. Remove unnecessary explanations or original terms from {source_lang} if present in the translation.

**Input:**

<SOURCE_TEXT>
{source_text}
</SOURCE_TEXT>

<INITIAL_TRANSLATION>
{initial_trans}
</INITIAL_TRANSLATION>

<EXPERT_SUGGESTIONS>
{reflection}
</EXPERT_SUGGESTIONS>

**Output:**
Only provide the {target_lang} translation for the above text. Do not include any explanations or text apart from the translation.

Return ONLY a single JSON object with the exact keys: "question" and "answer". Do not include any explanations or text outside the JSON. Do NOT add extra keys or metadata. All JSON keys must remain in English exactly as shown. Only translate the content inside square brackets.

<IMPROVED_TRANSLATION>
{
"question": "[improved translation of question]",
"answer": "[improved translation of answer]"
}
</IMPROVED_TRANSLATION>

**Supplementary Tab. 46:** Agentic machine translation prompt for the long-form QA task (Improved Translation).

*MCQA*

**System Message:**
You are a professional translator specializing in accurate translation of medical content from {source_lang} to {target_lang}.

**Your task is to translate the medical text. Follow these requirements:**

1. Preserve all numerals, decimal points, scientific notation, comparison signs, percentages, and brackets exactly as written (e.g., 2.5, μg/ml, 100%, [3H] leucine). Do NOT localize decimal separators or convert units/currencies.

2. Keep biochemical/clinical entities and abbreviations (e.g., NGF, tyrosine hydroxylase, ELISA, mRNA) exactly as in the source unless the target language has a well-established localized common name (rare). If localized, do NOT add the original in parentheses.

3. Preserve LaTeX/math/code verbatim; do not translate or alter content inside LaTeX/code delimiters.

4. Keep option labels and the number of options exactly as in the source (e.g., A/B/C/D). Do not merge or split options.

5. Maintain tone and register appropriate for professional medical assessments. Avoid explanatory additions.

**Task:**
Please translate the following MCQA item:

<SOURCE_TEXT>
{source_text}
</SOURCE_TEXT>

**Output:**
Only provide the {target_lang} translation for the above text. Do not include any explanations or text apart from the translation. Different options are separated by newline characters (\n). The number of options in the output must match the input exactly. Do not skip or combine any options.

Return the translation in the following JSON format, with keys: "question" and "options", where the value of "options" is a dictionary with keys option1, option2, option3, etc. All JSON keys must remain in English exactly as shown. Only translate the content inside square brackets:

<TRANSLATION>
{
"question": "[translation of question]",
"options": {
"option1": "[translation of option1]",
"option2": "[translation of option2]",
"option3": "[translation of option3]",
...
}
}
</TRANSLATION>

**Supplementary Tab. 47:** Agentic machine translation prompt for the MCQA task (Initial Translation).

**System Message:**
You are a medical translation expert, specializing in translation from {source_lang} to {target_lang}.

**Task Description:**
Carefully review the source text and its translation from {source_lang} to {target_lang}, and then provide constructive suggestions in English.

**Requirements:**
1. No additions, removals, or explanations of domain content.

2. Preserve anonymization and any specialized placeholders exactly (e.g., masked IDs, [***], {VAR}).

3. Numerals & notation: verify all numbers, decimal points, scientific notation, comparison signs, percentages, brackets, and Greek letters are unchanged; DO NOT localize decimal separators or convert currencies/units.

4. Units & symbols: confirm µg/ml, %, ×, ±, →, ≥/≤, and SI prefixes/symbols are preserved exactly (no written-out forms).

5. Biomedical entities/abbreviations (e.g., NGF, tyrosine hydroxylase, ELISA, mRNA): ensure they remain as in the source unless a well-established {target_lang} equivalent exists; if localized, the original term MUST NOT be added in parentheses; check consistent capitalization.

6. Options integrity: the number of options and their order must match the source exactly; no merge/split; no added symbols or written-out labels; each option remains a single line separated by \n.

7. Tone/register: maintain professional assessment style; remove any explanatory additions or commentary.

Input:

<SOURCE_TEXT>
{source_text}
</SOURCE_TEXT>

<INITIAL_TRANSLATION>
{initial_trans}
</INITIAL_TRANSLATION>

**Output:**

<SUGGESTIONS>
[Your suggestions here]
</SUGGESTIONS>

**Supplementary Tab. 48:** Agentic machine translation prompt for the MCQA task (Reflection).

**System Message:**
You are a senior medical MCQA translation reviser, specializing in translation from {source_lang} to {target_lang}.

**Task Description:**
Carefully review and edit the MCQA translation from {source_lang} to {target_lang}, incorporating the expert feedback below.

**Requirements:**

1. DO NOT explain anything; produce the improved translation JSON only.

2. Preserve every single quote from the source; do not add new single or double quotes.

3. Remove unnecessary explanations or original terms from {source_lang} if present in the translation.

**Input:**

<SOURCE_TEXT>
{source_text}
</SOURCE_TEXT>

<INITIAL_TRANSLATION>
{initial_trans}
</INITIAL_TRANSLATION>

<EXPERT_SUGGESTIONS>
{reflection}
</EXPERT_SUGGESTIONS>

**Output:**
Only provide the improved translation. Do not include any explanations or text apart from the translation. Different options are separated by newline characters (\n). The number of options in the output must match the input exactly. Do not skip or combine any options.

Return the translation in the following JSON format, with keys: "question" and "options", where the value of "options" is a dictionary with keys option1, option2, option3, etc. All JSON keys must remain in English exactly as shown. Only translate the content inside square brackets.

<IMPROVED_TRANSLATION>
{
"question": "[improved translation of question]",
"options": {
"option1": "[improved translation of option1]",
"option2": "[improved translation of option2]",
"option3": "[improved translation of option3]",
...
}
}
</IMPROVED_TRANSLATION>

**Supplementary Tab. 49:** Agentic machine translation prompt for the MCQA task (Improved Translation).

### S1.5. Expert Evaluation Criteria

During evaluation, experts proficient in the corresponding target language assessed translation accuracy, fluency, and completeness using a five-point (1–5) scale. **Supplementary Tab. 50–52** provide detailed evaluation criteria for these three metrics.

*Accuracy Evaluation Criteria*

**Accuracy (1-5)**

Evaluation Criteria:

5 points (Very Accurate)
**Medical terms and concepts are completely and correctly translated, with no errors.**
- All professional terms correspond to the original text, with no mistranslations or incorrect translations.
**The most appropriate and professional medical terms are used.**
- Expressions conform to common medical conventions, with standardized terminology use.

4 points (Accurate)
**Most medical terms and concepts are correctly translated, with only very few minor errors** that do not affect overall understanding.
- Some terms may not be precise, but overall accuracy is maintained.
**Generally appropriate medical terms are used.**
- In a few places, more colloquial words may be used, but they are still understandable to professionals.

3 points (Moderately Accurate)
**Major medical terms and concepts are generally correct, but there are some errors that may cause partial misunderstanding.**
- Some key terms are inaccurately translated, requiring the reader to infer.
**Deviation in the use of medical terms.**
- Occasionally uses less common or outdated terms.

2 points (Not Very Accurate)
**Most medical terms and concepts are mistranslated, severely affecting understanding.**
- Key concepts are mistranslated, possibly leading to misunderstanding of the original meaning.
**Incorrect or inappropriate medical terms are used.**
- Terminology is confused, lacking professionalism.

1 point (Inaccurate)
**Medical terms and concepts are riddled with errors, failing to correctly convey the original information.**
- Most of the content does not match the original text.
**Lacks correct use of medical terms.**
- Terminology is chaotic, possibly using non-medical vocabulary entirely.

**Supplementary Tab. 50:** Evaluation criteria for accuracy.

*Fluency Evaluation Criteria*

**Fluency (1-5)**

Evaluation Criteria:

5 points (Very Fluent)
**Natural and smooth expression, with no reading obstacles.**
- Elegant language, conforming to professional literature style.
**Sentence structure fully adheres to linguistic habits, with no grammatical or lexical errors.**

4 points (Fluent)
**Expression is mostly natural, with occasional minor language flaws that do not affect understanding.**
- Some sentences may seem slightly awkward.
**Sentence structure mostly conforms to habits, with very few grammatical errors.**

3 points (Moderately Fluent)
**Expression is somewhat unnatural, requiring readers to slightly adjust for understanding.**
- Some improper word use or awkward sentence structure.
**Sentence structure is generally correct, but there are some grammatical errors.**

2 points (Not Very Fluent)
**Expression is not fluent, with obvious reading obstacles.**
- Sentences are not smoothly connected, and logic is unclear.
**Sentence structure has many problems, with frequent grammatical errors.**

1 point (Not Fluent)
**Expression is very awkward or disjointed, making understanding difficult.**
- There may be a literal translation with a lack of proper expression.
**Sentence structure is chaotic, with severe grammatical errors, making the text unreadable.**

**Supplementary Tab. 51:** Evaluation criteria for fluency.

*Completeness Evaluation Criteria*

**Completeness (1-5)**

Evaluation Criteria:

5 points (Very Complete)
**Fully retains the original meaning, with no omissions or additions of information.**
- Details, data, and annotations are accurately conveyed.
**The translated content completely matches the original in length and depth.**

4 points (Complete)
**The main meaning of the original is preserved, with only very few minor omissions or unclear details.**
- Some minor information may be omitted.
**The translated content mostly corresponds to the original.**

3 points (Moderately Complete)
**Most of the original meaning is conveyed, but some information is omitted or added.**
- Important details may be overlooked.
**The translated content differs from the original, requiring the reader to infer part of the content.**

2 points (Not Very Complete)
**The main information of the original is not fully conveyed, with significant omissions or unnecessary additions.**
- Information unrelated to the original may be introduced.
**The translated content does not fully correspond to the original, affecting understanding.**

1 point (Incomplete)
**Large amounts of information are missing, or incorrect information is added, failing to reflect the main content of the original.**
- Important paragraphs or sentences are missing.
**The translated content significantly deviates from the original, making it impossible to understand the original meaning.**

**Supplementary Tab. 52:** Evaluation criteria for completeness.

## S1.6. Translation Quality

We selected six representative languages, French, German, Japanese, Korean, Swahili, and Yoruba, for translation quality evaluation. For each language, 300 samples were randomly selected and independently evaluated under blinded conditions. We further used Expert Agreement (EA) to quantify agreement between the two experts. The five-point ratings were grouped into three categories: scores of 1-2, a score of 3, and scores of 4-5; agreement was considered to be achieved when the two experts assigned ratings within the same category. EA was calculated separately for accuracy, fluency, and completeness, while overall EA represents the average agreement rate across the three dimensions.

| Language | Accuracy | Fluency | Completeness | Accuracy EA | Fluency EA | Completeness EA | Overall EA |
|---|---|---|---|---|---|---|---|
| **French** | 4.93 | 4.80 | 4.96 | 100.0% | 97.7% | 99.3% | 99.0% |
| **German** | 4.84 | 4.69 | 4.79 | 96.7% | 93.3% | 97.3% | 95.8% |
| **Japanese** | 4.77 | 4.63 | 4.95 | 100.0% | 100.0% | 100.0% | 100.0% |
| **Korean** | 4.69 | 4.50 | 4.77 | 90.7% | 92.0% | 90.0% | 90.9% |
| **Swahili** | 4.41 | 4.48 | 4.53 | 100.0% | 99.7% | 100.0% | 99.9% |
| **Yoruba** | 3.74 | 3.86 | 4.85 | 53.0% | 60.3% | 96.7% | 70.0% |

**Supplementary Tab. 53:** Medical expert evaluation of translation quality.

# S2. GlobMed-Bench

## S2.1. LLM Information

| LLMs | Developer | Parameters (B) | Domain | Reasoning | License |
|---|---|---|---|---|---|
| **Claude-3.5-Haiku** | Anthropic | Undisclosed | General | No | Proprietary |
| **Claude-4.0-Sonnet** | Anthropic | Undisclosed | General | Yes | Proprietary |
| **Gemini-2.5-Flash** | Google | Undisclosed | General | Yes | Proprietary |
| **GPT-4o-mini** | OpenAI | Undisclosed | General | No | Proprietary |
| **GPT-4o** | OpenAI | Undisclosed | General | No | Proprietary |
| **GPT-4.1-nano** | OpenAI | Undisclosed | General | No | Proprietary |
| **GPT-4.1-mini** | OpenAI | Undisclosed | General | No | Proprietary |
| **GPT-4.1** | OpenAI | Undisclosed | General | No | Proprietary |
| **GPT-5-nano** | OpenAI | Undisclosed | General | Yes | Proprietary |
| **GPT-5-mini** | OpenAI | Undisclosed | General | Yes | Proprietary |
| **GPT-5** | OpenAI | Undisclosed | General | Yes | Proprietary |
| **o4-mini** | OpenAI | Undisclosed | General | Yes | Proprietary |
| **DeepSeek-V3** | DeepSeek | 671B | General | No | MIT |
| **DeepSeek-R1** | DeepSeek | 671B | General | Yes | MIT |
| **DeepSeek-R1-Qwen3-8B** | DeepSeek | 8B | General | Yes | MIT |
| **Gemma-3-4B** | Google | 4B | General | No | Gemma |
| **Gemma-3-12B** | Google | 12B | General | No | Gemma |
| **Gemma-3-27B** | Google | 27B | General | No | Gemma |
| **gpt-oss-20B** | OpenAI | 20B | General | Yes | Apache 2.0 |
| **gpt-oss-120B** | OpenAI | 120B | General | Yes | Apache 2.0 |
| **LLaMA-3.1-8B** | Meta-llama | 8B | General | No | Llama-3.1 |
| **LLaMA-3.1-70B** | Meta-llama | 70B | General | No | Llama-3.1 |
| **LLaMA-3.2-3B** | Meta-llama | 3B | General | No | Llama-3.2 |
| **LLaMA-3.3-70B** | Meta-llama | 70B | General | No | Llama-3.3 |
| **LLaMA-4-Scout** | Meta-llama | 109B | General | No | Llama-4 |
| **LLaMA-4-Maverick** | Meta-llama | 400B | General | No | Llama-4 |
| **Mistral-7B-v0.3** | Mistral AI | 7B | General | No | Apache 2.0 |
| **Mistral-Small-3.1-24B** | Mistral AI | 24B | General | No | Apache 2.0 |
| **Phi-4-mini** | Microsoft | 3.8B | General | No | MIT |
| **Phi-4-mini-Reasoning** | Microsoft | 3.8B | General | Yes | MIT |
| **Phi-4** | Microsoft | 14B | General | No | MIT |
| **Phi-4-Reasoning** | Microsoft | 14B | General | Yes | MIT |
| **Qwen2.5-3B** | Qwen | 3B | General | No | Qwen-Research |
| **Qwen2.5-7B** | Qwen | 7B | General | No | Apache 2.0 |
| **Qwen2.5-14B** | Qwen | 14B | General | No | Apache 2.0 |
| **Qwen2.5-72B** | Qwen | 72B | General | No | Qwen |
| **QwQ-32B** | Qwen | 32B | General | Yes | Apache 2.0 |
| **Qwen3-1.7B** | Qwen | 1.7B | General | No | Apache 2.0 |
| **Qwen3-4B** | Qwen | 4B | General | No | Apache 2.0 |
| **Qwen3-4B-thinking** | Qwen | 4B | General | Yes | Apache 2.0 |
| **Qwen3-8B** | Qwen | 8B | General | No | Apache 2.0 |
| **Qwen3-8B-thinking** | Qwen | 8B | General | Yes | Apache 2.0 |
| **Qwen3-14B** | Qwen | 14B | General | No | Apache 2.0 |
| **Qwen3-14B-thinking** | Qwen | 14B | General | Yes | Apache 2.0 |
| **Baichuan-M2-32B** | Baichuan Inc. | 32B | Medical | Yes | Apache 2.0 |
| **Bio-Medical-LLaMA-3-8B** | ContactDoctor | 8B | Medical | No | Other |
| **MediPhi** | Microsoft | 3.8B | Medical | No | MIT |
| **MedGemma-4B** | Google | 4B | Medical | No | Health-AI-Developer-Foundations |
| **MedGemma-27B** | Google | 27B | Medical | No | Health-AI-Developer-Foundations |
| **MedReason-8B** | UCSC-VLAA | 8B | Medical | Yes | Apache 2.0 |
| **HuatuoGPT-o1-7B** | FreedomAI | 7B | Medical | Yes | Apache 2.0 |
| **HuatuoGPT-o1-8B** | FreedomAI | 8B | Medical | Yes | Apache 2.0 |
| **HuatuoGPT-o1-70B** | FreedomAI | 70B | Medical | Yes | Apache 2.0 |
| **HuatuoGPT-o1-72B** | FreedomAI | 72B | Medical | Yes | Apache 2.0 |
| **OpenBioLLM-8B** | Saama | 8B | Medical | No | Llama-3 |
| **OpenBioLLM-70B** | Saama | 70B | Medical | No | Llama-3 |

**Supplementary Tab. 54:** LLM information.

### S2.2. Evaluation Prompt

*MedNLI*

**任务:**

请通过选择正确的选项来确定给定前提和假设之间的关系。只提供直接答案 "**X**" ，其中 **X** 是正确的字母选择。

前提:
{前提}

假设:
{假设}

选项:
A. 蕴含
B. 中立
C. 矛盾

**答案:**

**Supplementary Tab. 55:** MedNLI evaluation prompt for Chinese.

**TASK:**

Please determine the relationship between the given premise and hypothesis by selecting the correct option. Provide direct answers only with "**X**" where **X** is the correct letter choice.

Premise:
{premise}

Hypothesis:
{hypothesis}

Options:
A. Entailment
B. Neutral
C. Contradiction

**Answer:**

**Supplementary Tab. 56:** MedNLI evaluation prompt for English.

**TÂCHE:**

Veuillez déterminer la relation entre la prémisse et l'hypothèse données en sélectionnant l'option correcte. Fournissez uniquement des réponses directes avec **"X"** où **X** est la lettre correcte.

Prémisse:
{Prémisse}

Hypothèse:
{Hypothèse}

Options:
A. Implication
B. Neutre
C. Contradiction

**Réponse:**

**Supplementary Tab. 57:** MedNLI evaluation prompt for French.

**AUFGABE:**

Bitte bestimmen Sie die Beziehung zwischen der gegebenen Prämisse und Hypothese, indem Sie die richtige Option auswählen. Geben Sie nur direkte Antworten mit **"X"**, wobei **X** die richtige Buchstabenoption ist.

Prämisse:
{Prämisse}

Hypothese:
{Hypothese}

Optionen:
A. Implikation
B. Neutral
C. Widerspruch

**Antwort:**

**Supplementary Tab. 58:** MedNLI evaluation prompt for German.

**タスク:**

与えられた前提と仮説の関係を、正しい選択肢を選んで判断してください。正しい選択肢は「**X**」のみ、直接答えてください。

前提:
{前提}

仮説:
{仮説}

選択肢:
A. 含意
B. 中立
C. 矛盾

**答え:**

**Supplementary Tab. 59:** MedNLI evaluation prompt for Japanese.

**작업:**

주어진 전제와 가설 사이의 관계를 판단하여 올바른 선택지를 고르십시오. 정답이 **"X"** 일 경우 **"X"** 로만 직접적으로 답하십시오.

전제:
{전제}

가설:
{가설}

선택지:
A. 함의
B. 중립
C. 모순

**답변:**

**Supplementary Tab. 60:** MedNLI evaluation prompt for Korean.

**TAREFA:**

Por favor, determine a relação entre a premissa e a hipótese dadas selecionando a opção correta. Forneça apenas respostas diretas com **"X"** onde **X** é a letra correta.

Premissa:
{Premissa}

Hipótese:
{Hipótese}

Opções:
A. Implicação
B. Neutro
C. Contradição

**Resposta:**

**Supplementary Tab. 61:** MedNLI evaluation prompt for Portuguese.

**TAREA:**

Por favor, determine la relación entre la premisa y la hipótesis dadas seleccionando la opción correcta. Proporcione solo respuestas directas con **"X"** donde **X** es la letra correcta.

Premisa:
{Premisa}

Hipótesis:
{Hipótesis}

Opciones:
A. Implicación
B. Neutral
C. Contradicción

**Respuesta:**

**Supplementary Tab. 62:** MedNLI evaluation prompt for Spanish.

**KAZI:**

Tafadhali amua uhusiano kati ya hoja iliyotolewa na wazo kwa kuchagua chaguo sahihi. Toa majibu ya moja kwa moja tu na **"X"** ambapo **X** ni chaguo sahihi cha herufi.

Hoja:
{Hoja}

Wazo:
{Wazo}

Chaguzi:
A. Uthibitisho
B. Katikati
C. Kupingana

**Jibu:**

**Supplementary Tab. 63:** MedNLI evaluation prompt for Swahili.

**LIGGÉEY:**

Jërëjëfël na nga xelal diggante bi nekk ci digle bi ak njortu bi ci tann ci option yi dëgg. Joxeel tontu yu jub rekk ak **"X"** fu **X** di araf bi dëgg.

Digle:
{Digle}

Njortu:
{Njortu}

Options:
A. Jagleel
B. Moytu
C. Dëddu

**Tontu:**

**Supplementary Tab. 64:** MedNLI evaluation prompt for Wolof.

**IṢẸ́:**

Jọwọ pinnu ìbátan láàrin ìsẹ̀lẹ̀ tí a fún ọ àti ìbéèrè nípa yíyan àṣàyàn tó tọ́. Ṣe ìdáhùn tààrà nìkan pẹ̀lú **"X"** níbi tí **X** jẹ́ àṣàyàn lẹ́tà tó tọ́.

Ìsẹ̀lẹ̀:
{Ìsẹ̀lẹ̀}

Ìbéèrè:
{Ìbéèrè}

Àwọn àṣàyàn:
A. Ìbámu
B. Ààrín
C. Ìlòdì

**Ìdáhùn:**

**Supplementary Tab. 65:** MedNLI evaluation prompt for Yoruba.

**UMSEBENZI:**

Sicela unqume ubudlelwano phakathi kwesitatimende esinikeziwe kanye nomqondo ngokukhetha inketho efanele. Nikeza izimpendulo eziqondile kuphela ngo-**"X"** lapho u-**X** eyinketho encwadi efanele.

Isitatimende:
{Isitatimende}

Umqondo:
{Umqondo}

Izinketho:
A. Ukuvumelana
B. Okungathathi hlangothi
C. Ukuphikisana

**Impendulo:**

**Supplementary Tab. 66:** MedNLI evaluation prompt for Zulu.

*BioNLI*

**任务:**

请通过选择正确的选项来确定给定前提和假设之间的关系。只提供直接答案 "**X**" ，其中 **X** 是正确的字母选择。

前提:
{前提}

假设:
{假设}

选项:
A. 蕴含
B. 矛盾

**答案:**

**Supplementary Tab. 67:** BioNLI evaluation prompt for Chinese.

**TASK:**

Please determine the relationship between the given premise and hypothesis by selecting the correct option. Provide direct answers only with "**X**" where **X** is the correct letter choice.

Premise:
{premise}

Hypothesis:
{hypothesis}

Options:
A. Entailment
B. Contradiction

**Answer:**

**Supplementary Tab. 68:** BioNLI evaluation prompt for English.

**TÂCHE:**

Veuillez déterminer la relation entre la prémisse et l'hypothèse données en sélectionnant l'option correcte. Fournissez uniquement des réponses directes avec **"X"** où **X** est la lettre correcte.

Prémisse:
{Prémisse}

Hypothèse:
{Hypothèse}

Options:
A. Implication
B. Contradiction

**Réponse:**

**Supplementary Tab. 69:** BioNLI evaluation prompt for French.

**AUFGABE:**

Bitte bestimmen Sie die Beziehung zwischen der gegebenen Prämisse und Hypothese, indem Sie die richtige Option auswählen. Geben Sie nur direkte Antworten mit **"X"**, wobei **X** die richtige Buchstabenoption ist.

Prämisse:
{Prämisse}

Hypothese:
{Hypothese}

Optionen:
A. Implikation
B. Widerspruch

**Antwort:**

**Supplementary Tab. 70:** BioNLI evaluation prompt for German.

**タスク:**

与えられた前提と仮説の関係を、正しい選択肢を選んで判断してください。正しい選択肢は「**X**」のみ、直接答えてください。

前提:
{前提}

仮説:
{仮説}

選択肢:
A. 含意
B. 矛盾

**答え:**

**Supplementary Tab. 71:** BioNLI evaluation prompt for Japanese.

**작업:**

주어진 전제와 가설 사이의 관계를 판단하여 올바른 선택지를 고르십시오. 정답이 **"X"** 일 경우 **"X"** 로만 직접적으로 답하십시오.

전제:
{전제}

가설:
{가설}

선택지:
A. 함의
B. 모순

**답변:**

**Supplementary Tab. 72:** BioNLI evaluation prompt for Korean.

**TAREFA:**

Por favor, determine a relação entre a premissa e a hipótese dadas selecionando a opção correta. Forneça apenas respostas diretas com **"X"** onde **X** é a letra correta.

Premissa:
{Premissa}

Hipótese:
{Hipótese}

Opções:
A. Implicação
B. Contradição

**Resposta:**

**Supplementary Tab. 73:** BioNLI evaluation prompt for Portuguese.

**TAREA:**

Por favor, determine la relación entre la premisa y la hipótesis dadas seleccionando la opción correcta. Proporcione solo respuestas directas con **"X"** donde **X** es la letra correcta.

Premisa:
{Premisa}

Hipótesis:
{Hipótesis}

Opciones:
A. Implicación
B. Contradicción

**Respuesta:**

**Supplementary Tab. 74:** BioNLI evaluation prompt for Spanish.

**KAZI:**

Tafadhali amua uhusiano kati ya hoja iliyotolewa na wazo kwa kuchagua chaguo sahihi. Toa majibu ya moja kwa moja tu na **"X"** ambapo **X** ni chaguo sahihi cha herufi.

Hoja:
{Hoja}

Wazo:
{Wazo}

Chaguzi:
A. Uthibitisho
B. Kupingana

**Jibu:**

**Supplementary Tab. 75:** BioNLI evaluation prompt for Swahili.

**LIGGÉEY:**

Jëréjëfël na nga xelal diggante bi nekk ci digle bi ak njortu bi ci tann ci option yi dëgg. Joxeel tontu yu jub rekk ak **"X"** fu **X** di araf bi dëgg.

Digle:
{Digle}

Njortu:
{Njortu}

Options:
A. Jagleel
B. Dëddu

**Tontu:**

**Supplementary Tab. 76:** BioNLI evaluation prompt for Wolof.

**IṢẸ́:**

Jọwọ pinnu ìbátan láàrin ìsẹ̀lẹ̀ tí a fún ọ àti ìbéèrè nípa yíyan àṣàyàn tó tọ́. Ṣe ìdáhùn tààrà nìkan pẹ̀lú **"X"** níbi tí **X** jẹ́ àṣàyàn lẹ́tà tó tọ́.

Ìsẹ̀lẹ̀:
{Ìsẹ̀lẹ̀}

Ìbéèrè:
{Ìbéèrè}

Àwọn àṣàyàn:
A. Ìbámu
B. Ìlòdì

**Ìdáhùn:**

**Supplementary Tab. 77:** BioNLI evaluation prompt for Yoruba.

**UMSEBENZI:**

Sicela unqume ubudlelwano phakathi kwesitatimende esinikeziwe kanye nomqondo ngokukhetha inketho efanele. Nikeza izimpendulo eziqondile kuphela ngo-**"X"** lapho u-**X** eyinketho encwadi efanele.

Isitatimende:
{Isitatimende}

Umqondo:
{Umqondo}

Izinketho:
A. Ukuvumelana
B. Ukuphikisana

**Impendulo:**

**Supplementary Tab. 78:** BioNLI evaluation prompt for Zulu.

*MCQA*

**任务:**

根据所给的问题和选项选择正确答案。仅提供直接答案，用 **"X"** 表示正确的字母选项。

问题:
{问题}

选项:
{选项}

**答案:**

**Supplementary Tab. 79:** MCQA evaluation prompt for Chinese.

**TASK:**

Choose the correct answer based on the given question and options. Provide direct answers only with **"X"** where **X** is the correct letter choice.

Question:
{question}

Options:
{options}

**Answer:**

**Supplementary Tab. 80:** MCQA evaluation prompt for English.

**TÂCHE:**

Choisissez la bonne réponse en vous basant sur la question et les options données. Fournissez uniquement la réponse directe avec **"X"**, où **X** est la bonne lettre.

Question:
{Question}

Options:
{Options}

**Réponse:**

**Supplementary Tab. 81:** MCQA evaluation prompt for French.

**AUFGABE:**

Bitte wählen Sie die richtige Antwort basierend auf der gestellten Frage und den Optionen. Geben Sie nur die direkte Antwort mit **"X"**, wobei **X** der richtige Buchstabe ist.

Frage:
{Frage}

Optionen:
{Optionen}

**Antwort:**

**Supplementary Tab. 82:** MCQA evaluation prompt for German.

**タスク:**

与えられた質問と選択肢に基づいて正しい答えを選んでください。正解の「**X**」のみで答えてください。

質問:
{質問}

選択肢:
{選択肢}

**答え:**

**Supplementary Tab. 83:** MCQA evaluation prompt for Japanese.

**작업:**

주어진 질문과 선택지를 바탕으로 정답을 고르십시오. 정답이 **"X"** 일 경우 **"X"** 로만 직접적으로 답하십시오.

질문:
{질문}

옵션:
{옵션}

**답변:**

**Supplementary Tab. 84:** MCQA evaluation prompt for Korean.

**TAREFA:**

Escolha a resposta correta com base na pergunta e nas opções fornecidas. Forneça apenas a resposta direta com **"X"**, onde **X** é a letra correta.

Pergunta:
{Pergunta}

Opções:
{Opções}

**Resposta:**

**Supplementary Tab. 85:** MCQA evaluation prompt for Portuguese.

**TAREA:**

Elige la respuesta correcta según la pregunta y las opciones dadas. Proporciona solo la respuesta directa con **"X"**, donde **X** es la letra correcta.

Pregunta:
{Pregunta}

Opciones:
{Opciones}

**Respuesta:**

**Supplementary Tab. 86:** MCQA evaluation prompt for Spanish.

**KAZI:**

Chagua jibu sahihi kulingana na swali na chaguzi zilizotolewa. Toa majibu ya moja kwa moja tu na **"X"** ambapo **X** ni chaguo sahihi cha herufi.

Swali:
{Swali}

Chaguzi:
{Chaguzi}

**Jibu:**

**Supplementary Tab. 87:** MCQA evaluation prompt for Swahili.

**LIGGÉEY:**

Tànnal tontu bi dëgg ci li ñu laaj ak tànneefi yi. Joxeel tontu yu jub rekk ak **"X"** fu **X** di araf bi dëgg.

Laaj:
{Laaj}

Tànneefi:
{Tànneefi}

**Tontu:**

**Supplementary Tab. 88:** MCQA evaluation prompt for Wolof.

**IṢẸ́:**

Yan ìdáhùn tó tọ́ lórí ìbéèrè àti àwọn àṣàyàn tí a fún ọ. Ṣe ìdáhùn tàárà nìkan pẹ̀lú **"X"**níbi tí **X** jẹ́ àṣàyàn lẹ́tà tó tọ́.

Ìbéèrè:
{Ìbéèrè}

Àwọn àṣàyàn:
{Àwọn àṣàyàn}

**Ìdáhùn:**

**Supplementary Tab. 89:** MCQA evaluation prompt for Yoruba.

**UMSEBENZI:**

Khetha impendulo efanele ngokusekelwe kumbuzo kanye nezinketho ezinikeziwe. Nikeza izimpendulo eziqondile kuphela ngo-**"X"** lapho u-**X** eyinketho encwadi efanele.

Umbuzo:
{Umbuzo}

Izinketho:
{Izinketho}

**Impendulo:**

**Supplementary Tab. 90:** MCQA evaluation prompt for Zulu.

### S2.3. Detailed Results of 56 Proprietary and Open-Weight LLMs

We present detailed evaluation results for 56 proprietary and open-weight LLMs across the six medical datasets in GlobMed-Bench, comprising 36 supplementary tables (**Supplementary Tab. 91–126**). For each dataset, we first report the mean accuracy and standard deviation of each model across five independent runs, followed by the accuracy results from each individual run. To visually illustrate performance disparities across languages, after excluding the original language of each dataset, the three highest-performing target languages for each model are highlighted in dark, medium, and light green, respectively, whereas the three lowest-performing target languages are highlighted in dark, medium, and light red, respectively.

| LLMs | Chinese | English | French | German | Japanese | Korean | Portuguese | Spanish | Swahili | Wolof | Yoruba | Zulu | Overall |
|---|---|---|---|---|---|---|---|---|---|---|---|---|---|
| **Proprietary LLMs** | | | | | | | | | | | | | |
| Claude-3.5-Haiku | 37.75±0.02 | 50.18±0.01 | 39.77±0.01 | 40.47±0.01 | 40.11±0.02 | 44.79±0.01 | 40.49±0.03 | 40.68±0.01 | 37.09±0.02 | 47.12±0.03 | 70.89±0.03 | 25.76±0.03 | 42.92 ± 0.00 |
| Claude-4.0-Sonnet | 68.31±0.30 | 75.46±0.30 | 63.88±0.12 | 59.72±0.41 | 62.98±0.18 | 58.72±0.35 | 68.48±0.15 | 67.95±0.24 | 54.74±0.10 | 45.47±0.53 | 47.38±0.20 | 53.82±0.25 | 60.58 ± 0.10 |
| Gemini-2.5-Flash | 62.44±0.42 | 69.06±0.56 | 58.13±0.24 | 59.40±0.14 | 58.74±0.24 | 61.82±0.19 | 61.29±0.41 | 60.66±0.18 | 53.76±0.36 | 55.54±0.46 | 63.93±0.25 | 54.59±0.16 | 59.95 ± 0.12 |
| GPT-4o-mini | 63.52±0.36 | 63.71±0.22 | 49.23±0.25 | 46.17±0.29 | 53.56±0.25 | 57.05±0.69 | 49.98±0.35 | 50.12±0.28 | 42.23±0.23 | 27.65±0.44 | 40.99±0.30 | 31.44±0.29 | 47.97 ± 0.15 |
| GPT-4o | 38.77±0.30 | 45.35±0.14 | 42.09±0.04 | 40.10±0.37 | 41.20±0.28 | 47.08±0.30 | 42.96±0.22 | 42.76±0.37 | 34.31±0.28 | 15.88±0.38 | 33.78±0.25 | 34.17±0.22 | 38.20 ± 0.09 |
| GPT-4.1-nano | 46.35±0.22 | 55.66±0.25 | 48.35±0.53 | 45.15±0.40 | 53.86±0.87 | 52.72±0.87 | 50.24±0.38 | 49.21±0.30 | 48.22±0.39 | 46.26±1.10 | 43.89±0.48 | 39.74±0.72 | 48.31 ± 0.16 |
| GPT-4.1-mini | 59.10±0.22 | 64.68±0.36 | 55.96±0.33 | 52.49±0.17 | 55.87±0.45 | 54.89±0.20 | 58.95±0.22 | 59.87±0.23 | 46.81±0.34 | 21.39±0.31 | 36.37±0.50 | 37.96±0.40 | 50.36 ± 0.07 |
| GPT-4.1 | 49.67±0.18 | 64.53±0.21 | 51.20±0.14 | 49.07±0.26 | 49.68±0.33 | 52.27±0.33 | 51.45±0.09 | 51.39±0.18 | 40.29±0.32 | 45.09±0.29 | 41.09±0.38 | 34.23±0.15 | 48.33 ± 0.05 |
| GPT-5-nano | 45.90±0.46 | 41.14±0.96 | 34.17±0.56 | 45.48±0.46 | 51.23±0.60 | 42.11±0.65 | 49.31±0.47 | 51.51±0.54 | 58.19±0.48 | 31.55±0.88 | 58.94±0.73 | 55.01±0.87 | 47.04 ± 0.08 |
| GPT-5-mini | 60.99±0.37 | 68.93±0.37 | 60.20±0.22 | 58.04±0.28 | 59.24±0.44 | 59.09±0.27 | 63.31±0.21 | 63.86±0.17 | 53.63±0.18 | 26.32±0.34 | 33.52±0.17 | 52.49±0.13 | 54.97 ± 0.07 |
| GPT-5 | 65.70±0.34 | 73.41±0.21 | 57.73±0.48 | 56.70±0.18 | 63.00±0.25 | 67.47±0.26 | 60.32±0.20 | 62.98±0.19 | 51.64±0.43 | 48.43±0.57 | 45.86±0.41 | 45.43±0.32 | 58.22 ± 0.08 |
| o4-mini | 68.97±0.31 | 73.34±0.20 | 67.13±0.19 | 66.90±0.25 | 68.42±0.22 | 67.08±0.27 | 70.24±0.41 | 71.06±0.23 | 61.42±0.22 | 32.30±0.45 | 49.20±0.31 | 55.24±0.62 | 62.61 ± 0.10 |
| **Open-Weight LLMs** | | | | | | | | | | | | | |
| DeepSeek-V3 | 51.37±0.29 | 63.91±0.26 | 49.80±0.17 | 51.90±0.48 | 49.28±0.37 | 48.66±0.36 | 52.21±0.15 | 50.81±0.18 | 43.59±0.18 | 42.33±0.90 | 37.78±0.52 | 32.85±0.22 | 47.87 ± 0.19 |
| DeepSeek-R1 | 62.59±0.17 | 74.17±0.12 | 60.19±0.08 | 60.48±0.18 | 56.87±0.29 | 55.42±0.30 | 64.32±0.16 | 62.49±0.36 | 51.09±0.35 | 50.52±0.49 | 54.59±0.22 | 49.14±0.40 | 58.49 ± 0.05 |
| DeepSeek-R1-Qwen3-8B | 59.35±0.50 | 64.43±0.22 | 51.34±0.39 | 48.34±0.27 | 52.42±0.21 | 48.93±0.55 | 52.63±0.61 | 54.97±0.58 | 44.67±0.48 | 11.54±0.31 | 18.20±0.31 | 13.92±0.53 | 43.39 ± 0.09 |
| Gemma-3-4B | 35.05±0.29 | 39.87±0.35 | 35.29±0.30 | 34.47±0.11 | 36.01±0.10 | 33.38±0.44 | 34.40±0.29 | 33.80±0.28 | 34.21±0.35 | 2.32±0.26 | 26.58±0.46 | 32.98±0.50 | 31.53 ± 0.17 |
| Gemma-3-12B | 32.39±0.23 | 38.32±0.21 | 34.52±0.08 | 32.29±0.08 | 35.82±0.39 | 37.37±0.30 | 36.00±0.27 | 35.28±0.19 | 39.35±0.38 | 4.83±0.40 | 55.15±0.59 | 35.69±0.33 | 34.75 ± 0.07 |
| Gemma-3-27B | 36.93±0.25 | 49.26±0.29 | 36.72±0.11 | 39.24±0.25 | 39.79±0.29 | 44.00±0.22 | 39.22±0.26 | 39.61±0.17 | 35.47±0.28 | 12.51±0.31 | 31.70±0.46 | 37.02±0.21 | 36.79 ± 0.13 |
| gpt-oss-20B | 68.88±0.18 | 62.77±0.71 | 61.55±0.20 | 60.24±0.40 | 64.22±0.45 | 64.60±0.48 | 62.81±0.09 | 63.91±0.25 | 55.16±0.52 | 24.47±0.49 | 46.64±0.76 | 54.23±0.52 | 57.46 ± 0.06 |
| gpt-oss-120B | 64.07±0.32 | 70.93±0.44 | 59.62±0.24 | 60.13±0.28 | 63.76±0.30 | 64.62±0.37 | 63.10±0.19 | 63.29±0.30 | 67.20±0.37 | 41.49±1.26 | 49.74±0.21 | 58.14±0.21 | 60.51 ± 0.09 |
| LLaMA-3.1-8B | 52.10±0.49 | 38.37±0.29 | 36.39±0.52 | 35.42±0.17 | 68.51±0.55 | 45.40±0.84 | 43.84±0.38 | 40.94±0.41 | 39.99±0.36 | 49.94±1.04 | 38.47±0.29 | 12.67±0.44 | 41.84 ± 0.13 |
| LLaMA-3.1-70B | 35.17±0.23 | 58.56±0.28 | 35.90±0.20 | 35.95±0.49 | 36.19±0.38 | 40.67±0.45 | 37.70±0.12 | 35.69±0.25 | 34.54±0.41 | 58.88±0.39 | 32.91±0.29 | 26.15±0.13 | 39.03 ± 0.13 |
| LLaMA-3.2-3B | 72.46±0.26 | 45.05±0.53 | 36.17±0.18 | 43.47±0.59 | 60.40±0.72 | 53.49±0.39 | 54.48±0.80 | 55.69±0.44 | 31.20±0.55 | 20.09±0.52 | 27.16±0.57 | 20.14±0.90 | 43.32 ± 0.13 |
| LLaMA-3.3-70B | 46.80±0.08 | 58.40±0.32 | 45.35±0.24 | 41.29±0.09 | 42.40±0.57 | 44.20±0.44 | 46.68±0.36 | 46.01±0.23 | 39.00±0.28 | 69.33±0.44 | 48.76±0.27 | 29.38±0.11 | 46.47 ± 0.10 |
| LLaMA-4-Scout | 48.14±0.42 | 56.46±0.41 | 42.52±0.31 | 42.47±0.26 | 43.55±0.28 | 45.14±0.21 | 47.65±0.21 | 44.04±0.29 | 40.84±0.30 | 54.19±0.32 | 62.82±0.15 | 44.04±0.31 | 47.65 ± 0.12 |
| LLaMA-4-Maverick | 50.31±0.28 | 61.59±0.16 | 46.13±0.12 | 45.74±0.11 | 50.17±0.29 | 53.14±0.47 | 48.98±0.32 | 49.04±0.12 | 39.67±0.24 | 63.19±0.36 | 42.39±0.51 | 37.39±0.32 | 48.98 ± 0.12 |
| Mistral-7B-v0.3 | 33.77±0.47 | 2.96±0.24 | 15.37±0.18 | 32.37±0.81 | 26.52±0.49 | 27.15±0.67 | 22.79±0.58 | 19.82±0.64 | 14.17±0.47 | 9.84±0.39 | 10.87±0.66 | 10.78±0.43 | 18.87 ± 0.08 |
| Mistral-Small-3.1-24B | 53.63±0.37 | 65.01±0.36 | 48.41±0.29 | 43.97±0.30 | 47.28±0.15 | 48.82±0.87 | 43.90±0.25 | 48.11±0.55 | 42.66±0.63 | 24.00±0.65 | 21.30±0.38 | 32.49±0.24 | 43.30 ± 0.17 |
| Phi-4-mini | 53.49±0.44 | 54.39±0.27 | 51.75±0.31 | 48.41±0.27 | 54.65±0.33 | 43.90±0.83 | 43.60±0.40 | 49.32±0.32 | 38.82±0.44 | 31.50±1.18 | 31.71±0.74 | 41.10±0.39 | 45.22 ± 0.14 |
| Phi-4-mini-Reasoning | 43.18±0.49 | 59.28±0.50 | 48.11±0.76 | 44.69±0.35 | 31.63±0.47 | 20.37±0.65 | 37.43±0.68 | 37.06±0.47 | 35.98±0.60 | 45.14±0.60 | 35.04±0.37 | 23.85±0.65 | 38.48 ± 0.18 |
| Phi-4 | 52.26±0.74 | 59.40±0.28 | 52.55±0.26 | 51.69±0.37 | 53.78±0.25 | 52.24±0.64 | 54.52±0.41 | 54.96±0.41 | 44.45±0.22 | 34.99±0.54 | 44.64±0.57 | 35.97±0.60 | 49.29 ± 0.20 |
| Phi-4-Reasoning | 63.14±0.30 | 69.16±0.23 | 59.47±0.48 | 55.84±0.25 | 61.87±0.44 | 59.33±0.21 | 57.68±0.36 | 61.65±0.47 | 57.74±0.62 | 15.25±0.35 | 48.78±0.74 | 37.89±0.68 | 53.98 ± 0.11 |
| Qwen2.5-3B | 58.53±0.31 | 63.52±0.61 | 49.35±0.47 | 46.79±0.73 | 58.77±0.60 | 58.81±0.86 | 55.05±0.29 | 54.30±0.24 | 20.77±0.31 | 1.72±0.14 | 12.86±0.76 | 10.49±0.40 | 40.91 ± 0.15 |
| Qwen2.5-7B | 50.22±0.47 | 62.04±0.48 | 53.87±0.67 | 47.88±0.58 | 49.61±0.54 | 44.70±0.75 | 54.62±0.19 | 50.11±0.37 | 45.43±0.97 | 28.77±0.29 | 38.67±0.50 | 29.16±0.82 | 46.26 ± 0.12 |
| Qwen2.5-14B | 52.19±0.35 | 67.40±0.56 | 48.87±0.19 | 45.29±0.34 | 45.68±0.25 | 48.63±0.20 | 51.31±0.36 | 53.24±0.19 | 45.47±0.52 | 59.04±0.33 | 70.53±0.32 | 38.01±0.65 | 52.14 ± 0.13 |
| Qwen2.5-72B | 47.66±0.56 | 65.00±0.41 | 55.39±0.15 | 52.62±0.28 | 50.07±0.51 | 53.59±0.22 | 58.92±0.30 | 60.07±0.21 | 38.96±0.34 | 71.40±0.14 | 55.46±0.83 | 29.40±0.30 | 53.21 ± 0.11 |
| QwQ-32B | 51.84±0.34 | 59.03±0.45 | 49.35±0.10 | 47.15±0.26 | 48.46±0.19 | 47.31±0.40 | 50.77±0.25 | 50.86±0.21 | 47.19±0.69 | 39.39±0.65 | 43.35±0.82 | 45.30±0.89 | 48.33 ± 0.05 |
| Qwen3-1.7B | 51.29±0.36 | 54.77±0.46 | 45.41±0.53 | 55.56±0.44 | 49.77±0.53 | 41.56±0.44 | 45.10±0.52 | 47.07±0.40 | 49.64±0.65 | 37.22±0.44 | 47.39±0.45 | 42.44±1.41 | 47.27 ± 0.18 |
| Qwen3-4B | 52.17±0.26 | 59.04±0.38 | 40.76±0.50 | 38.76±0.44 | 49.69±0.43 | 43.75±0.51 | 50.20±0.32 | 49.51±0.30 | 11.72±0.31 | 3.21±0.19 | 24.30±0.88 | 14.56±0.41 | 36.47 ± 0.21 |
| Qwen3-4B-thinking | 50.26±0.48 | 61.98±0.24 | 50.45±0.28 | 47.11±0.16 | 51.75±0.33 | 49.65±0.31 | 54.19±0.30 | 54.57±0.37 | 26.59±0.43 | 3.43±0.37 | 27.40±0.74 | 11.66±0.25 | 40.75 ± 0.09 |
| Qwen3-8B | 47.02±0.20 | 61.20±0.37 | 41.10±0.30 | 38.88±0.12 | 51.16±0.53 | 45.59±0.50 | 45.29±0.50 | 48.19±0.39 | 35.02±0.53 | 3.79±0.29 | 52.74±0.54 | 17.08±0.11 | 40.59 ± 0.06 |
| Qwen3-8B-thinking | 48.72±0.59 | 62.34±0.23 | 52.52±0.36 | 48.36±0.18 | 56.03±0.40 | 50.50±0.57 | 56.11±0.37 | 56.56±0.35 | 29.98±0.41 | 8.27±0.17 | 29.72±0.87 | 23.39±0.73 | 43.54 ± 0.19 |
| Qwen3-14B | 58.21±0.38 | 68.72±0.32 | 57.62±0.38 | 52.86±0.56 | 58.35±0.41 | 57.16±0.34 | 63.26±0.22 | 61.61±0.28 | 49.15±0.70 | 2.41±0.27 | 32.63±0.70 | 29.04±0.38 | 49.25 ± 0.11 |
| Qwen3-14B-thinking | 57.79±0.32 | 69.96±0.21 | 59.83±0.44 | 56.71±0.15 | 61.92±0.36 | 60.43±0.53 | 65.14±0.56 | 62.68±0.41 | 55.64±0.47 | 2.60±0.19 | 27.19±0.84 | 33.88±0.91 | 51.15 ± 0.17 |
| Baichuan-M2-32B | 62.71±0.33 | 57.50±0.43 | 50.36±0.45 | 59.86±0.37 | 58.83±0.41 | 59.71±0.47 | 54.70±0.51 | 55.67±0.44 | 36.48±0.69 | 30.79±0.20 | 41.07±0.74 | 43.31±1.02 | 50.92 ± 0.21 |
| Bio-Medical-LLaMA-3-8B | 35.03±0.44 | 40.84±0.43 | 25.97±0.09 | 31.76±0.26 | 34.83±0.49 | 25.73±0.15 | 32.19±0.30 | 29.83±0.31 | 26.07±0.06 | 36.28±0.79 | 30.89±0.20 | 35.01±0.32 | 32.04 ± 0.15 |
| MediPhi | 64.56±0.81 | 72.46±0.37 | 68.53±0.16 | 73.89±0.22 | 60.16±0.42 | 41.85±0.53 | 72.57±0.27 | 74.08±0.45 | 25.73±0.87 | 44.63±0.91 | 21.35±0.59 | 42.70±0.58 | 55.21 ± 0.11 |
| MedGemma-4B | 30.72±0.30 | 36.53±0.19 | 39.85±0.30 | 35.28±0.54 | 32.59±0.22 | 36.40±0.33 | 35.67±0.20 | 36.33±0.25 | 32.33±0.23 | 16.32±0.80 | 28.52±0.55 | 28.07±0.19 | 32.38 ± 0.11 |
| MedGemma-27B | 46.39±0.52 | 54.81±0.56 | 45.30±0.17 | 51.22±0.42 | 48.42±0.37 | 52.05±0.39 | 51.33±0.39 | 50.76±0.26 | 48.75±0.29 | 24.67±0.69 | 41.94±0.69 | 46.10±0.75 | 46.81 ± 0.10 |
| MedReason-8B | 27.60±0.20 | 6.41±0.22 | 4.32±0.21 | 4.15±0.21 | 25.53±0.50 | 30.91±0.22 | 24.10±0.12 | 1.69±0.18 | 8.43±0.31 | 0.49±0.10 | 37.09±0.43 | 23.86±0.11 | 16.22 ± 0.07 |
| HuatuoGPT-o1-7B | 49.02±0.18 | 57.66±0.80 | 52.03±0.39 | 51.43±0.40 | 54.46±0.56 | 49.95±0.20 | 52.49±0.28 | 55.11±0.35 | 16.97±0.56 | 2.74±0.16 | 6.52±0.21 | 4.01±0.32 | 37.70 ± 0.08 |
| HuatuoGPT-o1-8B | 33.68±0.31 | 34.41±0.13 | 36.26±0.37 | 31.37±0.48 | 44.02±0.39 | 35.35±0.45 | 37.86±0.53 | 37.12±0.31 | 27.49±0.39 | 1.78±0.18 | 21.64±0.68 | 7.56±0.37 | 29.04 ± 0.09 |
| HuatuoGPT-o1-70B | 44.62±0.29 | 58.75±0.13 | 47.65±0.22 | 46.30±0.27 | 43.58±0.48 | 46.88±0.20 | 48.04±0.27 | 49.35±0.08 | 37.81±0.20 | 55.35±0.63 | 39.05±0.21 | 33.87±0.33 | 45.94 ± 0.13 |
| HuatuoGPT-o1-72B | 52.81±0.40 | 64.92±0.17 | 59.08±0.35 | 57.54±0.24 | 53.93±0.26 | 55.59±0.50 | 61.61±0.33 | 63.33±0.30 | 44.26±0.51 | 14.88±0.52 | 40.32±0.60 | 25.41±0.13 | 49.47 ± 0.11 |
| OpenBioLLM-8B | 9.19±0.27 | 11.68±0.57 | 13.83±0.56 | 17.50±0.29 | 8.92±0.52 | 8.73±0.55 | 17.15±0.39 | 19.13±0.64 | 10.21±0.41 | 14.51±0.40 | 11.25±0.72 | 12.28±0.55 | 12.86 ± 0.07 |
| OpenBioLLM-70B | 12.77±0.41 | 32.56±0.13 | 26.86±0.21 | 15.55±0.22 | 10.26±0.47 | 10.92±0.56 | 22.36±0.43 | 27.53±0.20 | 11.47±0.24 | 11.47±0.55 | 16.16±0.40 | 15.93±0.50 | 17.82 ± 0.07 |

**Supplementary Tab. 91:** Performance evaluation of 56 LLMs on BioNLI.

| LLMs | Chinese | English | French | German | Japanese | Korean | Portuguese | Spanish | Swahili | Wolof | Yoruba | Zulu |
|---|---|---|---|---|---|---|---|---|---|---|---|---|
| **Proprietary LLMs** | | | | | | | | | | | | |
| **Claude-3.5-Haiku** | 37.75 | 50.18 | 39.78 | 40.47 | 40.09 | 44.79 | 40.49 | 40.67 | 37.10 | 47.15 | 70.90 | 25.78 |
| **Claude-4.0-Sonnet** | 68.45 | 75.17 | 63.91 | 59.96 | 63.12 | 58.54 | 68.29 | 67.62 | 54.74 | 45.03 | 47.24 | 53.42 |
| **Gemini-2.5-Flash** | 61.98 | 69.71 | 57.87 | 59.57 | 58.58 | 61.64 | 61.46 | 60.67 | 53.60 | 55.39 | 63.53 | 54.58 |
| **GPT-4o-mini** | 63.98 | 63.93 | 49.53 | 46.13 | 53.21 | 57.87 | 50.47 | 50.45 | 42.25 | 27.84 | 41.17 | 31.24 |
| **GPT-4o** | 38.58 | 45.39 | 42.04 | 40.31 | 41.21 | 46.81 | 42.85 | 42.56 | 34.29 | 15.62 | 34.00 | 34.13 |
| **GPT-4.1-nano** | 46.04 | 55.39 | 47.98 | 45.46 | 53.06 | 51.24 | 49.89 | 49.51 | 48.49 | 46.09 | 43.33 | 39.87 |
| **GPT-4.1-mini** | 59.01 | 64.38 | 56.09 | 52.22 | 55.69 | 54.94 | 59.03 | 59.78 | 46.92 | 21.71 | 36.74 | 38.38 |
| **GPT-4.1** | 49.89 | 64.67 | 51.24 | 49.35 | 49.73 | 52.36 | 51.55 | 51.66 | 39.93 | 44.99 | 40.85 | 34.36 |
| **GPT-5-nano** | 45.73 | 42.18 | 34.54 | 45.73 | 50.58 | 41.03 | 48.85 | 52.16 | 57.71 | 30.58 | 58.29 | 55.66 |
| **GPT-5-mini** | 60.56 | 68.97 | 60.11 | 57.80 | 58.79 | 59.33 | 63.39 | 63.89 | 53.84 | 25.84 | 33.37 | 52.34 |
| **GPT-5** | 66.27 | 73.37 | 57.64 | 56.65 | 62.70 | 67.82 | 60.29 | 62.76 | 51.69 | 47.84 | 46.00 | 45.30 |
| **o4-mini** | 69.30 | 73.35 | 66.94 | 67.15 | 68.52 | 66.67 | 70.09 | 71.17 | 61.46 | 32.04 | 48.72 | 54.45 |
| **Open-Weight LLMs** | | | | | | | | | | | | |
| **DeepSeek-V3** | 51.30 | 64.00 | 49.98 | 51.96 | 48.92 | 49.10 | 52.25 | 50.74 | 43.48 | 41.91 | 37.15 | 32.47 |
| **DeepSeek-R1** | 62.58 | 74.34 | 60.27 | 60.67 | 56.94 | 55.37 | 64.36 | 62.11 | 50.92 | 50.18 | 54.52 | 49.51 |
| **DeepSeek-R1-Qwen3-8B** | 59.66 | 64.38 | 51.37 | 48.36 | 52.63 | 49.26 | 51.62 | 55.51 | 44.47 | 12.00 | 18.56 | 14.07 |
| **Gemma-3-4B** | 34.72 | 40.22 | 35.15 | 34.36 | 35.84 | 32.72 | 34.45 | 33.78 | 33.89 | 2.22 | 25.87 | 32.22 |
| **Gemma-3-12B** | 32.74 | 38.02 | 34.45 | 32.20 | 35.48 | 37.66 | 36.27 | 35.21 | 38.83 | 4.99 | 55.15 | 36.04 |
| **Gemma-3-27B** | 37.19 | 49.75 | 36.61 | 39.28 | 39.89 | 44.27 | 39.62 | 39.73 | 35.51 | 12.11 | 32.34 | 37.35 |
| **gpt-oss-20B** | 68.99 | 63.69 | 61.28 | 60.16 | 64.22 | 63.91 | 62.70 | 64.34 | 54.90 | 24.11 | 45.96 | 54.49 |
| **gpt-oss-120B** | 64.25 | 70.61 | 59.91 | 59.89 | 63.69 | 64.40 | 62.94 | 63.03 | 67.73 | 42.36 | 49.75 | 57.78 |
| **LLaMA-3.1-8B** | 52.61 | 38.02 | 36.27 | 35.15 | 68.07 | 46.00 | 43.39 | 40.70 | 40.27 | 50.29 | 38.67 | 12.45 |
| **LLaMA-3.1-70B** | 34.90 | 58.36 | 35.80 | 35.53 | 36.11 | 40.92 | 37.62 | 35.91 | 34.83 | 59.15 | 32.99 | 26.36 |
| **LLaMA-3.2-3B** | 72.29 | 45.78 | 36.43 | 43.87 | 59.73 | 53.37 | 55.57 | 56.16 | 30.61 | 20.65 | 26.83 | 19.82 |
| **LLaMA-3.3-70B** | 46.85 | 58.47 | 45.55 | 41.37 | 42.65 | 44.47 | 46.45 | 45.87 | 38.90 | 69.12 | 48.90 | 29.39 |
| **LLaMA-4-Scout** | 48.07 | 56.90 | 42.83 | 42.13 | 44.02 | 45.26 | 47.62 | 43.98 | 40.90 | 54.31 | 63.01 | 44.45 |
| **LLaMA-4-Maverick** | 50.49 | 61.71 | 46.22 | 45.80 | 50.25 | 53.03 | 49.15 | 49.03 | 39.87 | 63.39 | 43.12 | 37.87 |
| **Mistral-7B-v0.3** | 34.11 | 2.81 | 15.60 | 31.98 | 26.22 | 27.75 | 22.94 | 19.21 | 13.89 | 10.11 | 11.19 | 10.25 |
| **Mistral-Small-3.1-24B** | 53.35 | 64.94 | 48.00 | 44.02 | 47.21 | 48.40 | 44.04 | 48.16 | 42.20 | 23.71 | 21.84 | 32.47 |
| **Phi-4-mini** | 53.08 | 54.81 | 52.07 | 48.27 | 54.79 | 43.37 | 43.80 | 49.21 | 39.21 | 31.26 | 32.61 | 40.65 |
| **Phi-4-mini-Reasoning** | 43.19 | 58.99 | 48.97 | 44.49 | 31.28 | 20.31 | 36.67 | 37.75 | 36.38 | 44.27 | 34.99 | 24.43 |
| **Phi-4** | 51.21 | 59.48 | 52.43 | 51.26 | 53.57 | 52.38 | 55.03 | 54.40 | 44.34 | 34.09 | 44.13 | 35.93 |
| **Phi-4-Reasoning** | 63.46 | 69.51 | 58.81 | 56.18 | 61.17 | 59.01 | 57.87 | 61.42 | 58.02 | 15.75 | 49.21 | 36.90 |
| **Qwen2.5-3B** | 58.88 | 63.44 | 48.70 | 47.35 | 59.26 | 59.55 | 55.12 | 54.40 | 20.79 | 1.82 | 13.80 | 10.34 |
| **Qwen2.5-7B** | 50.29 | 62.20 | 53.96 | 47.01 | 50.43 | 44.72 | 54.61 | 50.20 | 45.80 | 28.63 | 38.09 | 27.89 |
| **Qwen2.5-14B** | 52.45 | 68.11 | 48.79 | 45.48 | 45.37 | 48.45 | 51.19 | 53.39 | 45.64 | 59.08 | 70.72 | 38.16 |
| **Qwen2.5-72B** | 47.17 | 65.46 | 55.60 | 52.20 | 49.51 | 53.66 | 59.10 | 60.29 | 39.33 | 71.35 | 56.16 | 29.26 |
| **QwQ-32B** | 51.82 | 59.17 | 49.39 | 46.88 | 48.52 | 46.61 | 50.40 | 51.12 | 46.61 | 39.75 | 44.70 | 45.21 |
| **Qwen3-1.7B** | 51.24 | 54.67 | 46.07 | 55.30 | 50.02 | 41.46 | 45.24 | 46.47 | 50.40 | 37.51 | 47.10 | 42.74 |
| **Qwen3-4B** | 51.91 | 58.92 | 40.47 | 38.20 | 49.55 | 43.06 | 50.29 | 49.24 | 11.35 | 3.17 | 23.15 | 15.19 |
| **Qwen3-4B-thinking** | 50.67 | 61.87 | 50.07 | 47.35 | 52.31 | 50.09 | 53.80 | 55.19 | 26.54 | 3.66 | 27.30 | 11.48 |
| **Qwen3-8B** | 47.17 | 61.26 | 41.21 | 39.01 | 51.12 | 45.78 | 45.73 | 48.74 | 34.83 | 3.55 | 52.52 | 17.12 |
| **Qwen3-8B-thinking** | 48.90 | 62.58 | 52.34 | 48.45 | 56.18 | 50.34 | 55.98 | 57.03 | 29.44 | 8.07 | 28.92 | 23.53 |
| **Qwen3-14B** | 58.38 | 69.06 | 57.39 | 53.44 | 58.13 | 57.46 | 63.44 | 61.30 | 50.20 | 2.07 | 33.24 | 29.01 |
| **Qwen3-14B-thinking** | 57.73 | 70.18 | 59.82 | 56.47 | 62.13 | 60.76 | 64.49 | 63.06 | 55.60 | 2.63 | 26.52 | 32.65 |
| **Baichuan-M2-32B** | 62.29 | 57.10 | 49.82 | 60.36 | 58.45 | 59.19 | 54.56 | 55.39 | 35.69 | 30.85 | 42.34 | 41.73 |
| **Bio-Medical-LLaMA-3-8B** | 35.53 | 40.70 | 26.07 | 31.66 | 34.40 | 25.48 | 32.00 | 29.89 | 26.11 | 36.02 | 31.17 | 34.83 |
| **MediPhi** | 65.08 | 71.82 | 68.79 | 73.91 | 60.43 | 41.24 | 72.18 | 74.20 | 26.11 | 43.66 | 21.62 | 42.58 |
| **MedGemma-4B** | 30.79 | 36.36 | 39.55 | 34.38 | 32.88 | 36.31 | 35.89 | 36.63 | 32.36 | 14.92 | 28.11 | 28.20 |
| **MedGemma-27B** | 45.64 | 53.93 | 45.35 | 51.53 | 47.98 | 51.91 | 51.30 | 50.45 | 48.94 | 24.94 | 42.63 | 45.75 |
| **MedReason-8B** | 27.82 | 6.72 | 4.22 | 4.02 | 25.08 | 31.08 | 24.22 | 1.87 | 8.38 | 0.34 | 37.75 | 23.84 |
| **HuatuoGPT-o1-7B** | 48.79 | 57.33 | 52.18 | 51.42 | 54.43 | 49.84 | 52.94 | 54.74 | 17.46 | 2.72 | 6.52 | 4.52 |
| **HuatuoGPT-o1-8B** | 33.24 | 34.52 | 36.72 | 31.82 | 44.13 | 35.12 | 37.24 | 37.64 | 28.04 | 1.91 | 22.29 | 7.24 |
| **HuatuoGPT-o1-70B** | 44.90 | 58.88 | 47.53 | 46.20 | 43.42 | 46.94 | 48.02 | 49.39 | 37.55 | 54.99 | 38.94 | 33.75 |
| **HuatuoGPT-o1-72B** | 52.67 | 65.15 | 59.46 | 57.33 | 53.96 | 55.12 | 61.24 | 63.46 | 43.82 | 14.67 | 41.06 | 25.28 |
| **OpenBioLLM-8B** | 9.28 | 11.19 | 13.10 | 17.35 | 9.10 | 9.39 | 17.42 | 19.66 | 9.71 | 14.70 | 11.10 | 12.29 |
| **OpenBioLLM-70B** | 12.81 | 32.47 | 26.74 | 15.57 | 9.98 | 11.30 | 22.65 | 27.71 | 11.75 | 11.48 | 15.69 | 16.04 |

**Supplementary Tab. 92:** Zero-Shot performance evaluation of 56 LLMs on BioNLI (Run 1).

| LLMs | Chinese | English | French | German | Japanese | Korean | Portuguese | Spanish | Swahili | Wolof | Yoruba | Zulu |
|---|---|---|---|---|---|---|---|---|---|---|---|---|
| **Proprietary LLMs** | | | | | | | | | | | | |
| **Claude-3.5-Haiku** | 37.75 | 50.18 | 39.78 | 40.45 | 40.09 | 44.79 | 40.54 | 40.70 | 37.08 | 47.15 | 70.88 | 25.73 |
| **Claude-4.0-Sonnet** | 68.61 | 75.73 | 63.98 | 59.44 | 62.97 | 58.25 | 68.65 | 68.04 | 54.63 | 45.35 | 47.42 | 53.80 |
| **Gemini-2.5-Flash** | 63.08 | 69.28 | 58.27 | 59.26 | 58.99 | 61.82 | 61.93 | 60.85 | 53.64 | 55.98 | 64.13 | 54.40 |
| **GPT-4o-mini** | 63.19 | 63.37 | 48.90 | 46.38 | 53.62 | 57.60 | 49.51 | 49.71 | 42.20 | 27.82 | 41.15 | 31.10 |
| **GPT-4o** | 38.58 | 45.37 | 42.09 | 40.22 | 41.28 | 47.42 | 42.85 | 42.29 | 34.04 | 15.82 | 33.64 | 34.27 |
| **GPT-4.1-nano** | 46.63 | 55.46 | 47.60 | 44.45 | 55.17 | 52.92 | 50.07 | 49.53 | 47.89 | 47.91 | 43.82 | 39.93 |
| **GPT-4.1-mini** | 59.19 | 64.67 | 55.42 | 52.52 | 56.09 | 54.85 | 58.79 | 59.75 | 47.06 | 21.44 | 36.94 | 38.18 |
| **GPT-4.1** | 49.60 | 64.20 | 51.37 | 48.90 | 49.89 | 52.13 | 51.44 | 51.21 | 40.09 | 45.06 | 41.28 | 33.98 |
| **GPT-5-nano** | 46.27 | 40.25 | 34.72 | 45.91 | 50.88 | 42.79 | 48.97 | 50.90 | 58.88 | 31.35 | 59.78 | 54.31 |
| **GPT-5-mini** | 61.51 | 69.21 | 59.93 | 57.73 | 59.46 | 59.19 | 63.21 | 63.73 | 53.66 | 26.65 | 33.69 | 52.40 |
| **GPT-5** | 65.51 | 73.48 | 58.09 | 56.67 | 62.83 | 67.37 | 60.54 | 62.85 | 52.29 | 49.12 | 46.27 | 44.97 |
| **o4-mini** | 69.10 | 73.10 | 67.19 | 66.67 | 68.31 | 66.99 | 69.98 | 71.30 | 61.60 | 32.40 | 49.46 | 56.16 |
| **Open-Weight LLMs** | | | | | | | | | | | | |
| **DeepSeek-V3** | 51.71 | 63.82 | 49.71 | 52.65 | 49.89 | 48.45 | 52.25 | 50.72 | 43.60 | 43.75 | 38.18 | 32.97 |
| **DeepSeek-R1** | 62.36 | 74.04 | 60.11 | 60.22 | 56.72 | 55.57 | 64.56 | 62.11 | 50.72 | 51.24 | 54.27 | 49.33 |
| **DeepSeek-R1-Qwen3-8B** | 59.06 | 64.65 | 50.67 | 48.11 | 52.38 | 49.37 | 52.54 | 54.02 | 44.16 | 11.28 | 18.45 | 14.25 |
| **Gemma-3-4B** | 34.97 | 39.96 | 35.33 | 34.38 | 36.09 | 33.42 | 34.04 | 33.39 | 34.02 | 1.98 | 26.63 | 33.30 |
| **Gemma-3-12B** | 32.18 | 38.25 | 34.49 | 32.22 | 35.64 | 37.64 | 35.73 | 35.57 | 39.82 | 4.36 | 56.04 | 35.60 |
| **Gemma-3-27B** | 37.03 | 49.03 | 36.81 | 39.66 | 39.91 | 44.16 | 39.01 | 39.69 | 35.33 | 12.54 | 31.24 | 36.79 |
| **gpt-oss-20B** | 68.65 | 61.98 | 61.84 | 60.67 | 64.22 | 65.17 | 62.94 | 63.84 | 54.81 | 24.52 | 47.51 | 54.25 |
| **gpt-oss-120B** | 64.09 | 70.34 | 59.64 | 60.11 | 64.04 | 64.65 | 63.15 | 63.55 | 67.24 | 41.84 | 49.51 | 58.27 |
| **LLaMA-3.1-8B** | 52.36 | 38.74 | 36.83 | 35.55 | 68.43 | 46.40 | 44.13 | 41.48 | 40.31 | 49.08 | 38.13 | 12.34 |
| **LLaMA-3.1-70B** | 35.48 | 58.90 | 35.96 | 35.69 | 36.49 | 40.90 | 37.75 | 36.00 | 34.97 | 59.19 | 33.06 | 26.16 |
| **LLaMA-3.2-3B** | 72.47 | 44.49 | 36.09 | 42.85 | 60.45 | 53.78 | 53.62 | 55.78 | 31.55 | 20.11 | 26.97 | 19.06 |
| **LLaMA-3.3-70B** | 46.85 | 58.02 | 45.35 | 41.35 | 42.00 | 43.46 | 47.26 | 45.98 | 38.74 | 69.69 | 48.85 | 29.33 |
| **LLaMA-4-Scout** | 48.43 | 56.72 | 42.36 | 42.72 | 43.35 | 45.08 | 47.55 | 43.87 | 40.92 | 54.09 | 62.79 | 44.11 |
| **LLaMA-4-Maverick** | 50.11 | 61.57 | 45.96 | 45.84 | 50.47 | 52.40 | 48.72 | 48.94 | 39.26 | 63.57 | 42.02 | 37.24 |
| **Mistral-7B-v0.3** | 34.20 | 3.24 | 15.15 | 33.26 | 25.80 | 26.74 | 22.94 | 19.28 | 14.76 | 9.37 | 11.39 | 11.26 |
| **Mistral-Small-3.1-24B** | 54.07 | 64.58 | 48.58 | 43.46 | 47.08 | 48.79 | 44.18 | 48.04 | 41.80 | 23.91 | 21.37 | 32.76 |
| **Phi-4-mini** | 53.55 | 54.13 | 51.80 | 48.58 | 54.70 | 43.93 | 43.33 | 49.84 | 38.92 | 31.01 | 31.26 | 41.10 |
| **Phi-4-mini-Reasoning** | 42.54 | 60.11 | 47.78 | 44.47 | 31.33 | 19.89 | 36.79 | 36.99 | 35.26 | 44.83 | 35.10 | 23.26 |
| **Phi-4** | 53.24 | 58.92 | 52.81 | 51.80 | 53.48 | 52.72 | 54.65 | 55.17 | 44.79 | 34.90 | 45.42 | 36.79 |
| **Phi-4-Reasoning** | 62.67 | 69.01 | 59.51 | 55.51 | 62.20 | 59.39 | 57.37 | 61.06 | 56.85 | 15.35 | 49.30 | 38.79 |
| **Qwen2.5-3B** | 58.61 | 63.62 | 49.26 | 45.93 | 58.79 | 58.99 | 54.65 | 54.36 | 21.24 | 1.55 | 13.46 | 11.15 |
| **Qwen2.5-7B** | 50.92 | 62.40 | 53.15 | 47.69 | 49.42 | 43.93 | 54.58 | 49.98 | 45.69 | 28.97 | 38.22 | 29.37 |
| **Qwen2.5-14B** | 52.22 | 67.48 | 49.15 | 45.62 | 45.64 | 48.47 | 50.83 | 53.17 | 45.06 | 58.97 | 70.65 | 37.57 |
| **Qwen2.5-72B** | 48.29 | 64.45 | 55.30 | 52.61 | 50.70 | 53.75 | 58.92 | 60.18 | 39.21 | 71.48 | 54.85 | 29.53 |
| **QwQ-32B** | 51.60 | 58.92 | 49.44 | 47.19 | 48.43 | 47.46 | 51.03 | 50.99 | 47.51 | 39.12 | 43.37 | 45.44 |
| **Qwen3-1.7B** | 51.51 | 54.52 | 44.76 | 55.01 | 50.22 | 42.00 | 44.25 | 47.30 | 49.84 | 36.81 | 48.13 | 40.25 |
| **Qwen3-4B** | 52.13 | 59.17 | 40.18 | 39.08 | 49.44 | 44.09 | 50.20 | 49.15 | 11.60 | 3.35 | 23.64 | 14.72 |
| **Qwen3-4B-thinking** | 50.65 | 61.66 | 50.79 | 46.99 | 51.48 | 49.51 | 54.45 | 54.54 | 27.26 | 3.96 | 26.27 | 12.09 |
| **Qwen3-8B** | 46.74 | 60.65 | 41.55 | 38.99 | 51.66 | 44.97 | 44.94 | 48.22 | 35.21 | 3.91 | 53.21 | 17.03 |
| **Qwen3-8B-thinking** | 48.40 | 62.31 | 53.10 | 48.61 | 56.31 | 51.10 | 56.70 | 56.63 | 30.29 | 8.25 | 29.73 | 24.56 |
| **Qwen3-14B** | 58.40 | 68.34 | 57.73 | 52.76 | 59.06 | 57.10 | 63.12 | 62.02 | 49.46 | 2.79 | 31.80 | 29.01 |
| **Qwen3-14B-thinking** | 57.48 | 69.82 | 60.11 | 56.88 | 62.18 | 60.67 | 65.24 | 62.25 | 56.22 | 2.40 | 27.17 | 33.44 |
| **Baichuan-M2-32B** | 62.63 | 57.48 | 50.13 | 59.60 | 58.97 | 60.40 | 54.67 | 56.00 | 36.90 | 30.72 | 41.10 | 44.27 |
| **Bio-Medical-LLaMA-3-8B** | 35.28 | 40.20 | 26.02 | 31.37 | 34.67 | 25.75 | 32.54 | 29.48 | 25.96 | 35.55 | 30.67 | 34.56 |
| **MediPhi** | 64.99 | 72.52 | 68.40 | 73.51 | 60.27 | 41.37 | 72.81 | 74.72 | 24.58 | 43.98 | 21.06 | 43.28 |
| **MedGemma-4B** | 30.65 | 36.31 | 39.84 | 35.78 | 32.40 | 36.90 | 35.75 | 36.18 | 32.20 | 16.70 | 28.11 | 27.75 |
| **MedGemma-27B** | 47.10 | 55.28 | 45.10 | 51.55 | 48.81 | 52.45 | 51.82 | 50.72 | 48.45 | 23.89 | 41.75 | 46.72 |
| **MedReason-8B** | 27.33 | 6.20 | 4.40 | 4.29 | 25.75 | 31.06 | 23.98 | 1.62 | 8.70 | 0.61 | 36.61 | 23.87 |
| **HuatuoGPT-o1-7B** | 49.10 | 58.16 | 51.75 | 51.91 | 53.51 | 49.80 | 52.40 | 54.92 | 17.30 | 2.56 | 6.52 | 3.80 |
| **HuatuoGPT-o1-8B** | 33.55 | 34.54 | 36.13 | 31.10 | 44.31 | 34.90 | 38.52 | 37.10 | 27.17 | 1.55 | 20.72 | 8.09 |
| **HuatuoGPT-o1-70B** | 44.18 | 58.76 | 47.39 | 46.27 | 43.87 | 46.85 | 48.34 | 49.26 | 38.02 | 55.33 | 38.90 | 33.78 |
| **HuatuoGPT-o1-72B** | 52.56 | 64.88 | 59.08 | 57.84 | 54.29 | 55.42 | 61.78 | 63.28 | 44.27 | 14.11 | 39.73 | 25.55 |
| **OpenBioLLM-8B** | 8.76 | 12.02 | 14.04 | 17.35 | 9.69 | 7.91 | 17.48 | 19.60 | 10.58 | 14.99 | 10.72 | 11.69 |
| **OpenBioLLM-70B** | 12.52 | 32.61 | 26.92 | 15.91 | 10.25 | 11.44 | 21.89 | 27.71 | 11.42 | 10.63 | 16.00 | 16.52 |

**Supplementary Tab. 93:** Zero-Shot performance evaluation of 56 LLMs on BioNLI (Run 2).

| LLMs | Chinese | English | French | German | Japanese | Korean | Portuguese | Spanish | Swahili | Wolof | Yoruba | Zulu |
|---|---|---|---|---|---|---|---|---|---|---|---|---|
| **Proprietary LLMs** | | | | | | | | | | | | |
| **Claude-3.5-Haiku** | 37.73 | 50.20 | 39.78 | 40.47 | 40.11 | 44.81 | 40.49 | 40.67 | 37.06 | 47.12 | 70.85 | 25.75 |
| **Claude-4.0-Sonnet** | 67.82 | 75.82 | 63.82 | 59.33 | 62.67 | 58.76 | 68.61 | 68.25 | 54.85 | 46.09 | 47.39 | 53.80 |
| **Gemini-2.5-Flash** | 62.40 | 68.20 | 58.40 | 59.26 | 59.03 | 61.62 | 61.03 | 60.43 | 54.20 | 54.88 | 63.98 | 54.47 |
| **GPT-4o-mini** | 63.33 | 63.60 | 49.42 | 45.69 | 53.46 | 56.34 | 50.04 | 50.04 | 42.07 | 26.88 | 40.72 | 31.64 |
| **GPT-4o** | 38.72 | 45.39 | 42.09 | 40.49 | 41.21 | 46.97 | 43.33 | 43.01 | 34.52 | 16.25 | 33.84 | 34.16 |
| **GPT-4.1-nano** | 46.31 | 55.60 | 48.63 | 45.35 | 54.29 | 53.26 | 50.22 | 49.17 | 48.47 | 46.27 | 43.60 | 39.37 |
| **GPT-4.1-mini** | 58.83 | 64.52 | 56.16 | 52.70 | 55.21 | 55.17 | 58.92 | 59.84 | 46.22 | 21.33 | 36.13 | 38.16 |
| **GPT-4.1** | 49.42 | 64.63 | 51.12 | 49.35 | 49.10 | 52.70 | 51.53 | 51.37 | 40.74 | 44.88 | 41.42 | 34.31 |
| **GPT-5-nano** | 45.44 | 40.34 | 34.45 | 44.74 | 50.97 | 42.13 | 49.10 | 51.91 | 58.49 | 32.54 | 58.58 | 56.20 |
| **GPT-5-mini** | 60.88 | 69.33 | 60.45 | 58.40 | 59.08 | 59.15 | 63.08 | 63.87 | 53.73 | 26.20 | 33.66 | 52.58 |
| **GPT-5** | 65.39 | 73.08 | 57.03 | 57.01 | 63.03 | 67.55 | 60.00 | 62.97 | 51.26 | 48.49 | 45.98 | 45.46 |
| **o4-mini** | 69.12 | 73.44 | 67.28 | 66.94 | 68.52 | 67.39 | 70.76 | 70.74 | 61.51 | 32.11 | 49.08 | 55.33 |
| **Open-Weight LLMs** | | | | | | | | | | | | |
| **DeepSeek-V3** | 51.53 | 64.29 | 49.93 | 51.71 | 49.12 | 48.18 | 52.16 | 50.63 | 43.69 | 42.02 | 38.25 | 32.99 |
| **DeepSeek-R1** | 62.58 | 74.18 | 60.13 | 60.56 | 57.12 | 55.62 | 64.13 | 62.88 | 51.01 | 50.81 | 54.85 | 48.70 |
| **DeepSeek-R1-Qwen3-8B** | 59.51 | 64.09 | 51.62 | 48.61 | 52.09 | 49.33 | 53.21 | 55.37 | 45.46 | 11.26 | 17.78 | 13.17 |
| **Gemma-3-4B** | 35.51 | 40.11 | 35.37 | 34.63 | 36.04 | 33.39 | 34.76 | 33.89 | 34.65 | 2.54 | 26.47 | 33.48 |
| **Gemma-3-12B** | 32.47 | 38.31 | 34.45 | 32.34 | 35.84 | 36.99 | 35.78 | 35.15 | 39.51 | 4.52 | 55.08 | 35.17 |
| **Gemma-3-27B** | 36.83 | 49.12 | 36.72 | 39.15 | 40.04 | 43.84 | 39.28 | 39.39 | 35.82 | 12.83 | 31.73 | 37.06 |
| **gpt-oss-20B** | 68.99 | 62.58 | 61.55 | 59.62 | 63.64 | 64.52 | 62.74 | 63.78 | 55.53 | 24.67 | 47.42 | 54.76 |
| **gpt-oss-120B** | 63.53 | 71.17 | 59.37 | 59.89 | 64.04 | 65.15 | 63.24 | 62.99 | 67.03 | 42.40 | 49.91 | 58.16 |
| **LLaMA-3.1-8B** | 51.75 | 38.16 | 37.01 | 35.39 | 67.91 | 45.17 | 43.48 | 41.26 | 40.02 | 48.61 | 38.18 | 12.61 |
| **LLaMA-3.1-70B** | 35.15 | 58.72 | 35.96 | 35.71 | 35.55 | 40.38 | 37.55 | 35.55 | 34.70 | 58.49 | 33.28 | 26.11 |
| **LLaMA-3.2-3B** | 72.81 | 45.39 | 36.29 | 43.87 | 60.63 | 53.06 | 55.06 | 54.97 | 31.78 | 20.49 | 26.49 | 19.75 |
| **LLaMA-3.3-70B** | 46.74 | 58.38 | 45.10 | 41.30 | 42.72 | 44.61 | 46.40 | 46.11 | 38.76 | 69.26 | 48.27 | 29.57 |
| **LLaMA-4-Scout** | 47.44 | 56.54 | 42.07 | 42.25 | 43.53 | 45.33 | 48.00 | 43.82 | 41.28 | 53.69 | 62.67 | 44.00 |
| **LLaMA-4-Maverick** | 50.70 | 61.39 | 46.09 | 45.55 | 49.84 | 53.66 | 48.76 | 49.10 | 39.78 | 62.63 | 42.63 | 37.53 |
| **Mistral-7B-v0.3** | 33.39 | 2.85 | 15.44 | 31.48 | 26.90 | 27.55 | 22.79 | 20.54 | 13.93 | 9.80 | 11.39 | 10.99 |
| **Mistral-Small-3.1-24B** | 53.96 | 65.24 | 48.49 | 44.09 | 47.46 | 47.89 | 43.51 | 48.76 | 42.97 | 24.43 | 20.79 | 32.43 |
| **Phi-4-mini** | 53.98 | 54.25 | 51.46 | 48.00 | 54.07 | 43.66 | 44.13 | 49.24 | 39.19 | 31.48 | 31.53 | 41.55 |
| **Phi-4-mini-Reasoning** | 42.97 | 58.81 | 47.08 | 45.30 | 31.64 | 21.48 | 38.27 | 37.19 | 36.61 | 45.84 | 35.62 | 23.28 |
| **Phi-4** | 52.54 | 59.53 | 52.18 | 52.02 | 54.09 | 52.94 | 53.98 | 55.48 | 44.22 | 35.24 | 44.83 | 35.57 |
| **Phi-4-Reasoning** | 63.15 | 68.97 | 59.91 | 55.93 | 62.22 | 59.53 | 57.57 | 61.80 | 58.18 | 14.94 | 49.42 | 38.02 |
| **Qwen2.5-3B** | 58.04 | 64.31 | 49.57 | 47.64 | 58.29 | 59.35 | 54.88 | 54.27 | 20.40 | 1.84 | 12.02 | 10.11 |
| **Qwen2.5-7B** | 50.09 | 61.53 | 54.20 | 48.16 | 49.62 | 44.74 | 54.94 | 50.70 | 45.51 | 28.67 | 39.06 | 29.84 |
| **Qwen2.5-14B** | 52.11 | 67.48 | 48.65 | 44.74 | 45.71 | 48.56 | 51.19 | 52.94 | 44.90 | 58.56 | 70.45 | 38.88 |
| **Qwen2.5-72B** | 47.48 | 65.15 | 55.44 | 52.83 | 50.43 | 53.44 | 58.85 | 59.75 | 39.01 | 71.53 | 55.89 | 29.15 |
| **QwQ-32B** | 51.66 | 58.31 | 49.24 | 47.42 | 48.20 | 47.44 | 50.90 | 50.88 | 48.07 | 38.43 | 42.90 | 46.72 |
| **Qwen3-1.7B** | 51.78 | 54.18 | 44.99 | 56.02 | 49.71 | 40.94 | 45.55 | 47.42 | 49.44 | 37.35 | 47.03 | 42.45 |
| **Qwen3-4B** | 52.45 | 59.42 | 41.51 | 39.28 | 49.19 | 44.36 | 50.67 | 49.60 | 12.16 | 3.21 | 24.90 | 14.49 |
| **Qwen3-4B-thinking** | 50.43 | 62.09 | 50.49 | 46.97 | 51.71 | 49.26 | 54.52 | 54.22 | 26.22 | 3.12 | 27.93 | 11.48 |
| **Qwen3-8B** | 47.01 | 61.21 | 41.03 | 38.83 | 50.31 | 46.25 | 45.82 | 47.69 | 35.71 | 4.13 | 52.38 | 17.08 |
| **Qwen3-8B-thinking** | 49.39 | 61.98 | 52.63 | 48.27 | 55.71 | 50.38 | 55.69 | 56.11 | 30.36 | 8.54 | 31.06 | 23.33 |
| **Qwen3-14B** | 57.69 | 69.03 | 57.19 | 53.39 | 58.11 | 57.53 | 63.17 | 61.44 | 48.45 | 2.27 | 32.11 | 29.39 |
| **Qwen3-14B-thinking** | 58.27 | 70.13 | 60.02 | 56.76 | 61.55 | 60.36 | 64.94 | 62.29 | 55.51 | 2.40 | 27.39 | 35.08 |
| **Baichuan-M2-32B** | 63.06 | 57.51 | 50.76 | 60.11 | 58.97 | 59.82 | 55.08 | 55.35 | 37.44 | 30.79 | 40.79 | 42.94 |
| **Bio-Medical-LLaMA-3-8B** | 35.15 | 41.21 | 25.87 | 31.82 | 35.37 | 25.75 | 32.09 | 30.31 | 26.09 | 37.62 | 30.72 | 35.12 |
| **MediPhi** | 65.10 | 72.56 | 68.40 | 74.07 | 60.18 | 42.43 | 72.72 | 74.13 | 25.69 | 44.47 | 20.94 | 42.74 |
| **MedGemma-4B** | 31.03 | 36.76 | 39.62 | 35.39 | 32.56 | 36.36 | 35.78 | 36.45 | 32.20 | 16.81 | 28.52 | 28.22 |
| **MedGemma-27B** | 46.54 | 55.33 | 45.26 | 50.58 | 48.49 | 51.62 | 50.85 | 50.61 | 48.43 | 25.71 | 40.88 | 46.76 |
| **MedReason-8B** | 27.48 | 6.52 | 4.47 | 4.09 | 25.10 | 30.54 | 24.07 | 1.62 | 8.16 | 0.47 | 36.94 | 23.69 |
| **HuatuoGPT-o1-7B** | 48.90 | 56.94 | 52.56 | 51.44 | 54.70 | 49.82 | 52.18 | 55.33 | 16.07 | 2.99 | 6.29 | 3.93 |
| **HuatuoGPT-o1-8B** | 33.66 | 34.36 | 36.00 | 31.53 | 43.84 | 35.19 | 38.13 | 36.94 | 27.60 | 2.00 | 21.15 | 7.26 |
| **HuatuoGPT-o1-70B** | 44.81 | 58.54 | 47.87 | 46.07 | 43.89 | 46.63 | 48.22 | 49.26 | 38.00 | 55.60 | 39.42 | 34.13 |
| **HuatuoGPT-o1-72B** | 53.17 | 65.01 | 59.35 | 57.73 | 53.91 | 55.39 | 61.35 | 63.48 | 44.70 | 15.08 | 39.98 | 25.51 |
| **OpenBioLLM-8B** | 9.10 | 11.64 | 13.60 | 17.24 | 8.49 | 8.81 | 17.33 | 18.47 | 10.63 | 14.00 | 12.27 | 11.78 |
| **OpenBioLLM-70B** | 13.46 | 32.67 | 26.56 | 15.39 | 10.94 | 11.17 | 22.36 | 27.28 | 11.15 | 11.46 | 16.72 | 15.75 |

**Supplementary Tab. 94:** Zero-Shot performance evaluation of 56 LLMs on BioNLI (Run 3).

| LLMs | Chinese | English | French | German | Japanese | Korean | Portuguese | Spanish | Swahili | Wolof | Yoruba | Zulu |
|---|---|---|---|---|---|---|---|---|---|---|---|---|
| **Proprietary LLMs** | | | | | | | | | | | | |
| **Claude-3.5-Haiku** | 37.78 | 50.18 | 39.75 | 40.47 | 40.11 | 44.79 | 40.47 | 40.67 | 37.10 | 47.08 | 70.88 | 25.80 |
| **Claude-4.0-Sonnet** | 68.27 | 75.39 | 63.98 | 60.31 | 63.03 | 58.90 | 68.40 | 68.02 | 54.65 | 45.96 | 47.69 | 54.04 |
| **Gemini-2.5-Flash** | 62.54 | 68.90 | 58.22 | 59.46 | 58.58 | 62.00 | 61.08 | 60.52 | 54.04 | 55.48 | 63.87 | 54.74 |
| **GPT-4o-mini** | 63.26 | 63.82 | 49.17 | 46.29 | 53.89 | 56.40 | 49.82 | 50.29 | 42.61 | 27.80 | 41.30 | 31.80 |
| **GPT-4o** | 39.30 | 45.10 | 42.16 | 39.55 | 41.53 | 47.39 | 42.97 | 42.72 | 34.65 | 15.42 | 34.02 | 34.45 |
| **GPT-4.1-nano** | 46.31 | 55.89 | 48.74 | 45.24 | 53.28 | 53.44 | 50.13 | 48.90 | 48.54 | 46.22 | 44.13 | 38.81 |
| **GPT-4.1-mini** | 59.42 | 64.54 | 55.89 | 52.54 | 56.40 | 54.61 | 59.28 | 59.71 | 46.99 | 20.90 | 35.69 | 37.60 |
| **GPT-4.1** | 49.80 | 64.47 | 51.01 | 48.81 | 49.91 | 52.34 | 51.37 | 51.26 | 40.22 | 45.60 | 40.54 | 34.22 |
| **GPT-5-nano** | 46.49 | 40.79 | 33.55 | 45.33 | 51.66 | 42.27 | 49.69 | 51.53 | 58.00 | 32.38 | 59.69 | 54.27 |
| **GPT-5-mini** | 60.83 | 68.40 | 60.40 | 58.04 | 59.89 | 59.15 | 63.24 | 64.13 | 53.37 | 26.29 | 33.30 | 52.67 |
| **GPT-5** | 65.60 | 73.48 | 57.64 | 56.56 | 63.08 | 67.10 | 60.38 | 63.15 | 51.24 | 48.83 | 45.89 | 45.82 |
| **o4-mini** | 68.85 | 73.60 | 67.33 | 67.12 | 68.09 | 67.15 | 69.80 | 70.92 | 61.03 | 31.91 | 49.46 | 55.01 |
| **Open-Weight LLMs** | | | | | | | | | | | | |
| **DeepSeek-V3** | 50.94 | 63.82 | 49.82 | 51.82 | 49.19 | 48.90 | 52.40 | 51.10 | 43.82 | 42.58 | 38.02 | 32.94 |
| **DeepSeek-R1** | 62.85 | 74.09 | 60.29 | 60.38 | 56.45 | 55.64 | 64.22 | 62.63 | 51.66 | 50.07 | 54.70 | 49.44 |
| **DeepSeek-R1-Qwen3-8B** | 59.89 | 64.61 | 51.51 | 48.02 | 52.45 | 48.47 | 52.83 | 54.99 | 44.58 | 11.69 | 18.09 | 13.60 |
| **Gemma-3-4B** | 35.03 | 39.35 | 35.71 | 34.49 | 36.07 | 33.96 | 34.58 | 33.75 | 33.98 | 2.27 | 27.10 | 32.76 |
| **Gemma-3-12B** | 32.38 | 38.58 | 34.63 | 32.38 | 35.66 | 37.39 | 36.29 | 35.10 | 39.15 | 5.37 | 54.38 | 35.82 |
| **Gemma-3-27B** | 36.56 | 49.28 | 36.83 | 39.01 | 39.80 | 43.96 | 38.99 | 39.78 | 35.60 | 12.79 | 31.89 | 37.01 |
| **gpt-oss-20B** | 69.06 | 62.29 | 61.48 | 60.27 | 64.11 | 64.90 | 62.85 | 63.75 | 55.89 | 23.91 | 46.07 | 54.27 |
| **gpt-oss-120B** | 64.13 | 71.37 | 59.39 | 60.16 | 63.71 | 64.72 | 62.85 | 63.66 | 67.30 | 41.51 | 49.55 | 58.29 |
| **LLaMA-3.1-8B** | 51.44 | 38.56 | 35.78 | 35.57 | 69.15 | 45.21 | 44.18 | 40.61 | 39.42 | 50.99 | 38.67 | 12.49 |
| **LLaMA-3.1-70B** | 35.33 | 58.20 | 35.62 | 36.76 | 36.36 | 40.02 | 37.73 | 35.46 | 34.04 | 58.43 | 32.61 | 26.09 |
| **LLaMA-3.2-3B** | 72.13 | 44.70 | 36.00 | 43.96 | 59.75 | 53.26 | 54.04 | 55.84 | 31.44 | 19.82 | 27.71 | 21.39 |
| **LLaMA-3.3-70B** | 46.70 | 58.90 | 45.62 | 41.15 | 42.99 | 44.22 | 46.79 | 45.75 | 39.33 | 69.84 | 48.92 | 29.35 |
| **LLaMA-4-Scout** | 48.25 | 55.87 | 42.72 | 42.61 | 43.30 | 44.81 | 47.44 | 44.00 | 40.54 | 54.38 | 62.70 | 43.57 |
| **LLaMA-4-Maverick** | 50.00 | 61.78 | 46.11 | 45.73 | 50.38 | 53.35 | 48.81 | 49.21 | 39.71 | 63.08 | 41.84 | 37.03 |
| **Mistral-7B-v0.3** | 33.98 | 2.70 | 15.44 | 31.91 | 26.83 | 26.16 | 23.42 | 20.45 | 14.58 | 10.34 | 10.45 | 10.40 |
| **Mistral-Small-3.1-24B** | 53.24 | 65.48 | 48.72 | 44.22 | 47.39 | 50.22 | 43.93 | 48.34 | 43.17 | 24.81 | 21.24 | 32.13 |
| **Phi-4-mini** | 52.99 | 54.27 | 51.39 | 48.52 | 54.83 | 45.30 | 43.10 | 49.33 | 38.63 | 30.29 | 30.83 | 40.79 |
| **Phi-4-mini-Reasoning** | 43.87 | 59.33 | 48.00 | 44.52 | 31.48 | 20.25 | 37.80 | 36.88 | 35.42 | 45.44 | 34.83 | 23.64 |
| **Phi-4** | 52.04 | 59.44 | 52.58 | 51.33 | 53.87 | 51.42 | 54.67 | 55.01 | 44.54 | 35.28 | 44.79 | 35.26 |
| **Phi-4-Reasoning** | 63.33 | 69.01 | 59.93 | 55.82 | 61.71 | 59.48 | 57.37 | 61.64 | 57.35 | 15.30 | 47.82 | 37.71 |
| **Qwen2.5-3B** | 58.47 | 62.61 | 49.98 | 46.81 | 59.46 | 58.81 | 55.19 | 53.91 | 20.81 | 1.60 | 12.27 | 10.54 |
| **Qwen2.5-7B** | 49.62 | 62.54 | 54.76 | 48.56 | 48.94 | 45.89 | 54.56 | 50.00 | 43.80 | 28.43 | 39.21 | 29.84 |
| **Qwen2.5-14B** | 51.64 | 66.54 | 48.97 | 45.37 | 45.62 | 48.92 | 51.64 | 53.39 | 45.57 | 59.12 | 70.81 | 37.21 |
| **Qwen2.5-72B** | 47.15 | 64.72 | 55.21 | 52.90 | 49.62 | 53.28 | 59.28 | 59.98 | 38.49 | 71.46 | 54.31 | 29.87 |
| **QwQ-32B** | 52.43 | 59.39 | 49.26 | 46.88 | 48.72 | 47.55 | 50.63 | 50.56 | 46.38 | 39.53 | 43.26 | 44.49 |
| **Qwen3-1.7B** | 50.90 | 55.19 | 45.69 | 55.51 | 48.88 | 41.46 | 45.44 | 46.88 | 48.65 | 37.73 | 47.17 | 42.58 |
| **Qwen3-4B** | 52.43 | 58.43 | 40.94 | 38.79 | 50.00 | 43.51 | 49.89 | 49.75 | 11.91 | 2.92 | 24.58 | 14.13 |
| **Qwen3-4B-thinking** | 49.96 | 62.29 | 50.29 | 47.19 | 51.60 | 49.75 | 54.18 | 54.52 | 26.70 | 3.10 | 28.18 | 11.57 |
| **Qwen3-8B** | 46.94 | 61.21 | 40.74 | 38.83 | 51.53 | 45.73 | 44.65 | 48.34 | 34.27 | 3.42 | 53.42 | 16.94 |
| **Qwen3-8B-thinking** | 47.87 | 62.38 | 52.20 | 48.16 | 56.43 | 51.01 | 56.09 | 56.36 | 29.64 | 8.22 | 29.91 | 22.70 |
| **Qwen3-14B** | 58.63 | 68.72 | 58.20 | 52.11 | 58.38 | 56.72 | 63.03 | 61.57 | 48.65 | 2.43 | 32.56 | 29.35 |
| **Qwen3-14B-thinking** | 57.93 | 69.69 | 59.08 | 56.67 | 61.51 | 59.53 | 66.02 | 62.67 | 54.97 | 2.81 | 26.40 | 34.25 |
| **Baichuan-M2-32B** | 62.56 | 57.19 | 50.20 | 59.73 | 58.38 | 59.80 | 53.93 | 55.33 | 36.18 | 31.06 | 40.67 | 43.60 |
| **Bio-Medical-LLaMA-3-8B** | 34.76 | 40.83 | 26.02 | 32.07 | 35.33 | 25.91 | 31.87 | 29.69 | 26.09 | 35.93 | 30.94 | 35.19 |
| **MediPhi** | 63.19 | 72.76 | 68.52 | 73.98 | 59.44 | 41.93 | 72.40 | 73.80 | 26.92 | 45.96 | 20.88 | 41.78 |
| **MedGemma-4B** | 30.25 | 36.56 | 39.96 | 35.26 | 32.38 | 36.43 | 35.42 | 36.40 | 32.72 | 16.38 | 29.44 | 28.16 |
| **MedGemma-27B** | 46.31 | 54.74 | 45.57 | 51.42 | 48.72 | 51.78 | 51.60 | 51.10 | 49.03 | 24.43 | 41.98 | 44.97 |
| **MedReason-8B** | 27.78 | 6.20 | 4.00 | 4.43 | 25.46 | 30.99 | 24.00 | 1.89 | 8.81 | 0.49 | 37.21 | 24.00 |
| **HuatuoGPT-o1-7B** | 49.06 | 58.81 | 52.13 | 50.81 | 54.83 | 50.02 | 52.49 | 54.94 | 16.79 | 2.67 | 6.85 | 4.07 |
| **HuatuoGPT-o1-8B** | 34.00 | 34.22 | 35.89 | 31.73 | 44.40 | 36.07 | 38.00 | 36.85 | 27.06 | 1.75 | 21.87 | 7.80 |
| **HuatuoGPT-o1-70B** | 44.72 | 58.85 | 47.89 | 46.76 | 43.91 | 46.83 | 47.64 | 49.39 | 37.73 | 56.25 | 38.97 | 34.27 |
| **HuatuoGPT-o1-72B** | 52.36 | 64.72 | 58.58 | 57.30 | 53.57 | 55.57 | 61.60 | 63.60 | 43.69 | 15.01 | 40.88 | 25.46 |
| **OpenBioLLM-8B** | 9.33 | 11.08 | 13.80 | 17.62 | 8.90 | 9.01 | 16.97 | 18.38 | 10.22 | 14.20 | 11.66 | 12.76 |
| **OpenBioLLM-70B** | 12.67 | 32.38 | 27.01 | 15.37 | 10.43 | 10.54 | 22.00 | 27.35 | 11.66 | 11.60 | 16.40 | 15.19 |

**Supplementary Tab. 95:** Zero-Shot performance evaluation of 56 LLMs on BioNLI (Run 4).

| LLMs | Chinese | English | French | German | Japanese | Korean | Portuguese | Spanish | Swahili | Wolof | Yoruba | Zulu |
|---|---|---|---|---|---|---|---|---|---|---|---|---|
| **Proprietary LLMs** | | | | | | | | | | | | |
| **Claude-3.5-Haiku** | 37.75 | 50.18 | 39.78 | 40.47 | 40.13 | 44.79 | 40.47 | 40.67 | 37.10 | 47.12 | 70.92 | 25.73 |
| **Claude-4.0-Sonnet** | 68.38 | 75.21 | 63.71 | 59.57 | 63.10 | 59.17 | 68.47 | 67.80 | 54.81 | 44.94 | 47.17 | 54.02 |
| **Gemini-2.5-Flash** | 62.20 | 69.19 | 57.89 | 59.44 | 58.54 | 62.00 | 60.94 | 60.81 | 53.30 | 55.98 | 64.16 | 54.76 |
| **GPT-4o-mini** | 63.84 | 63.82 | 49.15 | 46.38 | 53.60 | 57.03 | 50.07 | 50.09 | 42.04 | 27.93 | 40.63 | 31.42 |
| **GPT-4o** | 38.65 | 45.48 | 42.07 | 39.91 | 40.76 | 46.81 | 42.79 | 43.24 | 34.04 | 16.27 | 33.42 | 33.84 |
| **GPT-4.1-nano** | 46.47 | 55.96 | 48.81 | 45.24 | 53.51 | 52.76 | 50.88 | 48.94 | 47.71 | 44.81 | 44.58 | 40.74 |
| **GPT-4.1-mini** | 59.03 | 65.30 | 56.25 | 52.47 | 55.98 | 54.88 | 58.72 | 60.27 | 46.85 | 21.55 | 36.36 | 37.48 |
| **GPT-4.1** | 49.62 | 64.70 | 51.24 | 48.94 | 49.78 | 51.80 | 51.35 | 51.44 | 40.49 | 44.94 | 41.35 | 34.29 |
| **GPT-5-nano** | 45.55 | 42.16 | 33.57 | 45.69 | 52.04 | 42.31 | 49.93 | 51.06 | 57.89 | 30.88 | 58.36 | 54.63 |
| **GPT-5-mini** | 61.19 | 68.76 | 60.11 | 58.22 | 58.97 | 58.63 | 63.62 | 63.69 | 53.57 | 26.63 | 33.57 | 52.47 |
| **GPT-5** | 65.71 | 73.64 | 58.25 | 56.61 | 63.35 | 67.51 | 60.40 | 63.19 | 51.71 | 47.87 | 45.17 | 45.62 |
| **o4-mini** | 68.49 | 73.19 | 66.92 | 66.61 | 68.65 | 67.21 | 70.56 | 71.17 | 61.48 | 33.03 | 49.30 | 55.26 |
| **Open-Weight LLMs** | | | | | | | | | | | | |
| **DeepSeek-V3** | 51.37 | 63.60 | 49.55 | 51.35 | 49.30 | 48.65 | 51.98 | 50.88 | 43.37 | 41.37 | 37.30 | 32.90 |
| **DeepSeek-R1** | 62.58 | 74.22 | 60.16 | 60.58 | 57.12 | 54.92 | 64.31 | 62.74 | 51.15 | 50.31 | 54.61 | 48.72 |
| **DeepSeek-R1-Qwen3-8B** | 58.63 | 64.40 | 51.55 | 48.58 | 52.54 | 48.20 | 52.94 | 54.94 | 44.70 | 11.46 | 18.11 | 14.49 |
| **Gemma-3-4B** | 35.03 | 39.69 | 34.88 | 34.47 | 36.02 | 33.42 | 34.18 | 34.18 | 34.52 | 2.61 | 26.83 | 33.15 |
| **Gemma-3-12B** | 32.20 | 38.43 | 34.56 | 32.31 | 36.47 | 37.15 | 35.93 | 35.39 | 39.46 | 4.92 | 55.10 | 35.82 |
| **Gemma-3-27B** | 37.06 | 49.12 | 36.61 | 39.10 | 39.30 | 43.75 | 39.21 | 39.46 | 35.08 | 12.29 | 31.28 | 36.90 |
| **gpt-oss-20B** | 68.72 | 63.30 | 61.60 | 60.47 | 64.90 | 64.49 | 62.83 | 63.82 | 54.67 | 25.15 | 46.25 | 53.37 |
| **gpt-oss-120B** | 64.34 | 71.17 | 59.78 | 60.58 | 63.33 | 64.18 | 63.30 | 63.24 | 66.72 | 39.33 | 49.98 | 58.20 |
| **LLaMA-3.1-8B** | 52.36 | 38.38 | 36.07 | 35.46 | 68.99 | 44.22 | 44.02 | 40.63 | 39.93 | 50.72 | 38.70 | 13.44 |
| **LLaMA-3.1-70B** | 35.01 | 58.61 | 36.16 | 36.04 | 36.43 | 41.12 | 37.87 | 35.53 | 34.18 | 59.15 | 32.63 | 26.02 |
| **LLaMA-3.2-3B** | 72.58 | 44.88 | 36.04 | 42.79 | 61.46 | 54.00 | 54.13 | 55.69 | 30.61 | 19.37 | 27.80 | 20.67 |
| **LLaMA-3.3-70B** | 46.88 | 58.25 | 45.12 | 41.26 | 41.62 | 44.25 | 46.49 | 46.34 | 39.28 | 68.74 | 48.85 | 29.28 |
| **LLaMA-4-Scout** | 48.49 | 56.25 | 42.61 | 42.63 | 43.55 | 45.24 | 47.62 | 44.54 | 40.56 | 54.49 | 62.94 | 44.07 |
| **LLaMA-4-Maverick** | 50.25 | 61.48 | 46.25 | 45.78 | 49.89 | 53.24 | 49.46 | 48.92 | 39.71 | 63.26 | 42.36 | 37.28 |
| **Mistral-7B-v0.3** | 33.15 | 3.19 | 15.24 | 33.21 | 26.83 | 27.53 | 21.84 | 19.62 | 13.69 | 9.60 | 9.91 | 11.01 |
| **Mistral-Small-3.1-24B** | 53.51 | 64.79 | 48.25 | 44.07 | 47.26 | 48.81 | 43.84 | 47.26 | 43.17 | 23.12 | 21.24 | 32.65 |
| **Phi-4-mini** | 53.84 | 54.47 | 52.02 | 48.67 | 54.85 | 43.24 | 43.64 | 48.97 | 38.16 | 33.44 | 32.34 | 41.42 |
| **Phi-4-mini-Reasoning** | 43.33 | 59.17 | 48.74 | 44.67 | 32.43 | 19.93 | 37.60 | 36.47 | 36.25 | 45.30 | 34.65 | 24.65 |
| **Phi-4** | 52.29 | 59.62 | 52.76 | 52.02 | 53.87 | 51.75 | 54.25 | 54.72 | 44.38 | 35.42 | 44.04 | 36.31 |
| **Phi-4-Reasoning** | 63.10 | 69.28 | 59.17 | 55.75 | 62.07 | 59.26 | 58.22 | 62.34 | 58.29 | 14.90 | 48.13 | 38.02 |
| **Qwen2.5-3B** | 58.65 | 63.60 | 49.26 | 46.20 | 58.07 | 57.37 | 55.39 | 54.56 | 20.63 | 1.80 | 12.74 | 10.29 |
| **Qwen2.5-7B** | 50.16 | 61.55 | 53.28 | 47.96 | 49.66 | 44.22 | 54.43 | 49.69 | 46.36 | 29.15 | 38.79 | 28.85 |
| **Qwen2.5-14B** | 52.54 | 67.39 | 48.81 | 45.26 | 46.07 | 48.74 | 51.69 | 53.33 | 46.20 | 59.48 | 70.00 | 38.25 |
| **Qwen2.5-72B** | 48.22 | 65.21 | 55.39 | 52.54 | 50.09 | 53.82 | 58.47 | 60.13 | 38.76 | 71.17 | 56.07 | 29.21 |
| **QwQ-32B** | 51.71 | 59.37 | 49.44 | 47.37 | 48.45 | 47.51 | 50.88 | 50.76 | 47.37 | 40.13 | 42.54 | 44.63 |
| **Qwen3-1.7B** | 51.03 | 55.28 | 45.55 | 55.98 | 50.02 | 41.96 | 45.03 | 47.30 | 49.87 | 36.72 | 47.51 | 44.18 |
| **Qwen3-4B** | 51.93 | 59.24 | 40.70 | 38.47 | 50.25 | 43.71 | 49.93 | 49.80 | 11.60 | 3.42 | 25.24 | 14.29 |
| **Qwen3-4B-thinking** | 49.57 | 62.00 | 50.61 | 47.03 | 51.64 | 49.64 | 54.02 | 54.38 | 26.22 | 3.30 | 27.33 | 11.66 |
| **Qwen3-8B** | 47.24 | 61.69 | 40.97 | 38.72 | 51.17 | 45.21 | 45.33 | 47.98 | 35.10 | 3.93 | 52.18 | 17.24 |
| **Qwen3-8B-thinking** | 49.03 | 62.47 | 52.34 | 48.29 | 55.51 | 49.69 | 56.07 | 56.65 | 30.16 | 8.25 | 28.99 | 22.85 |
| **Qwen3-14B** | 57.96 | 68.47 | 57.60 | 52.58 | 58.09 | 56.97 | 63.53 | 61.73 | 49.01 | 2.47 | 33.42 | 28.43 |
| **Qwen3-14B-thinking** | 57.55 | 69.98 | 60.13 | 56.76 | 62.22 | 60.83 | 64.99 | 63.12 | 55.91 | 2.76 | 28.49 | 33.98 |
| **Baichuan-M2-32B** | 63.03 | 58.20 | 50.90 | 59.48 | 59.37 | 59.35 | 55.24 | 56.27 | 36.18 | 30.52 | 40.47 | 44.02 |
| **Bio-Medical-LLaMA-3-8B** | 34.43 | 41.26 | 25.87 | 31.87 | 34.38 | 25.75 | 32.47 | 29.80 | 26.09 | 36.29 | 30.97 | 35.37 |
| **MediPhi** | 64.45 | 72.63 | 68.52 | 73.96 | 60.47 | 42.27 | 72.74 | 73.53 | 25.33 | 45.06 | 22.27 | 43.10 |
| **MedGemma-4B** | 30.88 | 36.65 | 40.29 | 35.60 | 32.74 | 35.98 | 35.51 | 35.98 | 32.18 | 16.81 | 28.40 | 28.04 |
| **MedGemma-27B** | 46.36 | 54.79 | 45.24 | 51.03 | 48.11 | 52.47 | 51.10 | 50.94 | 48.92 | 24.36 | 42.47 | 46.29 |
| **MedReason-8B** | 27.60 | 6.43 | 4.49 | 3.91 | 26.27 | 30.90 | 24.22 | 1.46 | 8.11 | 0.56 | 36.92 | 23.91 |
| **HuatuoGPT-o1-7B** | 49.24 | 57.08 | 51.55 | 51.55 | 54.83 | 50.27 | 52.43 | 55.60 | 17.21 | 2.74 | 6.43 | 3.71 |
| **HuatuoGPT-o1-8B** | 33.93 | 34.43 | 36.58 | 30.67 | 43.44 | 35.46 | 37.39 | 37.06 | 27.57 | 1.71 | 22.18 | 7.39 |
| **HuatuoGPT-o1-70B** | 44.49 | 58.74 | 47.55 | 46.22 | 42.81 | 47.17 | 47.96 | 49.44 | 37.75 | 54.58 | 39.03 | 33.44 |
| **HuatuoGPT-o1-72B** | 53.30 | 64.83 | 58.94 | 57.48 | 53.91 | 56.43 | 62.07 | 62.83 | 44.83 | 15.51 | 39.96 | 25.26 |
| **OpenBioLLM-8B** | 9.46 | 12.45 | 14.61 | 17.96 | 8.40 | 8.54 | 16.54 | 19.53 | 9.89 | 14.67 | 10.49 | 12.90 |
| **OpenBioLLM-70B** | 12.40 | 32.65 | 27.06 | 15.53 | 9.69 | 10.13 | 22.92 | 27.60 | 11.39 | 12.16 | 16.00 | 16.13 |

**Supplementary Tab. 96:** Zero-Shot performance evaluation of 56 LLMs on BioNLI (Run 5).

| LLMs | Chinese | English | French | German | Japanese | Korean | Portuguese | Spanish | Swahili | Wolof | Yoruba | Zulu | Overall |
|---|---|---|---|---|---|---|---|---|---|---|---|---|---|
| **Proprietary LLMs** | | | | | | | | | | | | | |
| Claude-3.5-Haiku | 69.28±0.48 | 78.14±0.74 | 67.47±0.25 | 66.41±0.48 | 67.89±0.48 | 69.54±0.50 | 66.73±0.41 | 68.78±0.24 | 51.38±0.69 | 9.98±0.60 | 34.05±0.99 | 32.65±1.10 | 56.86 ± 0.22 |
| Claude-4.0-Sonnet | 81.40±0.57 | 85.66±0.43 | 79.32±0.42 | 75.75±0.34 | 78.53±0.42 | 80.10±0.36 | 76.13±0.46 | 78.30±0.26 | 68.35±0.18 | 34.52±0.57 | 57.84±0.45 | 56.88±0.34 | 71.07 ± 0.06 |
| Gemini-2.5-Flash | 71.83±0.84 | 83.48±0.91 | 70.71±0.45 | 69.60±0.52 | 67.78±0.26 | 69.80±0.58 | 69.26±0.41 | 70.12±0.36 | 67.14±0.44 | 46.91±0.76 | 59.20±0.51 | 62.22±0.63 | 67.34 ± 0.09 |
| GPT-4o-mini | 79.77±1.03 | 84.84±0.54 | 81.31±0.94 | 79.19±0.80 | 78.11±0.74 | 77.05±0.87 | 80.30±1.23 | 79.65±0.42 | 62.03±0.61 | 29.12±1.81 | 34.66±0.50 | 50.77±0.81 | 68.07 ± 0.41 |
| GPT-4o | 78.18±0.63 | 81.75±1.12 | 75.44±0.96 | 75.47±0.26 | 76.12±1.13 | 76.87±0.38 | 78.62±0.74 | 75.15±0.24 | 71.29±0.34 | 27.27±1.42 | 46.88±0.84 | 61.18±0.78 | 68.68 ± 0.19 |
| GPT-4.1-nano | 71.52±0.92 | 83.50±0.42 | 75.13±0.96 | 75.09±0.46 | 70.97±1.15 | 69.63±0.44 | 72.42±0.80 | 75.81±0.88 | 56.56±1.18 | 24.53±1.16 | 33.28±0.44 | 47.57±1.04 | 63.00 ± 0.17 |
| GPT-4.1-mini | 82.74±0.46 | 87.72±0.43 | 85.39±0.42 | 82.86±0.42 | 79.27±0.54 | 80.93±0.39 | 82.65±0.26 | 82.75±0.29 | 63.47±0.61 | 19.90±1.20 | 47.30±0.80 | 59.69±0.53 | 71.22 ± 0.16 |
| GPT-4.1 | 83.12±0.39 | 87.75±0.23 | 85.33±0.33 | 84.88±0.96 | 82.24±0.35 | 82.41±0.23 | 85.12±0.26 | 86.43±0.44 | 71.60±0.37 | 38.81±0.74 | 50.02±0.98 | 63.42±0.39 | 75.10 ± 0.25 |
| GPT-5-nano | 73.25±0.75 | 77.88±1.02 | 70.02±0.54 | 78.97±0.89 | 72.28±1.02 | 69.65±1.18 | 78.62±0.24 | 80.93±0.99 | 54.82±1.30 | 16.75±0.54 | 32.38±0.80 | 41.40±0.65 | 62.25 ± 0.17 |
| GPT-5-mini | 83.77±0.42 | 88.56±0.17 | 87.71±0.35 | 86.48±0.57 | 83.36±0.34 | 82.86±0.70 | 86.79±0.63 | 87.23±0.36 | 72.55±0.65 | 32.19±0.86 | 46.96±0.45 | 63.09±0.65 | 75.13 ± 0.19 |
| GPT-5 | 81.00±0.54 | 86.35±0.11 | 85.89±0.18 | 85.76±0.35 | 82.12±0.62 | 83.36±0.75 | 84.74±0.31 | 85.63±0.24 | 80.39±0.50 | 43.22±0.81 | 54.05±0.70 | 65.65±0.41 | 76.51 ± 0.06 |
| o4-mini | 84.92±0.54 | 88.76±0.45 | 87.69±0.39 | 87.45±0.35 | 85.05±0.23 | 84.40±0.57 | 87.26±0.25 | 87.79±0.42 | 79.10±0.68 | 21.67±1.34 | 51.86±1.05 | 67.75±0.28 | 76.14 ± 0.15 |
| **Open-Weight LLMs** | | | | | | | | | | | | | |
| DeepSeek-V3 | 78.42±0.52 | 84.35±0.39 | 77.99±0.47 | 79.15±0.72 | 75.03±0.32 | 75.05±0.17 | 76.50±0.56 | 78.53±0.35 | 68.10±0.89 | 34.21±0.98 | 45.91±0.38 | 51.76±1.25 | 68.75 ± 0.17 |
| DeepSeek-R1 | 77.93±0.34 | 83.37±0.34 | 74.91±0.56 | 76.75±0.44 | 76.10±0.45 | 73.03±0.76 | 74.65±0.29 | 77.52±0.50 | 59.46±0.43 | 34.96±0.36 | 45.97±0.46 | 54.67±0.76 | 67.44 ± 0.12 |
| DeepSeek-R1-Qwen3-8B | 77.73±0.77 | 80.58±0.48 | 75.88±1.38 | 76.00±0.74 | 75.51±0.49 | 71.28±0.94 | 76.36±0.78 | 76.50±0.73 | 31.53±1.16 | 12.21±0.17 | 12.41±0.71 | 15.53±1.54 | 56.79 ± 0.10 |
| Gemma-3-4B | 64.84±0.99 | 73.72±0.70 | 61.31±0.59 | 58.89±0.79 | 58.08±0.70 | 63.25±0.95 | 62.67±0.64 | 62.81±0.85 | 45.93±1.03 | 7.56±0.67 | 29.63±0.34 | 21.35±0.80 | 50.84 ± 0.15 |
| Gemma-3-12B | 63.54±0.80 | 82.12±0.63 | 66.92±0.78 | 67.02±0.72 | 67.97±0.59 | 69.49±0.52 | 64.67±0.31 | 66.41±0.44 | 65.55±0.84 | 7.31±0.80 | 34.95±1.60 | 49.30±0.97 | 58.77 ± 0.28 |
| Gemma-3-27B | 77.50±0.91 | 82.75±0.61 | 70.16±0.57 | 73.69±0.23 | 73.55±0.77 | 78.00±0.50 | 72.50±0.46 | 73.32±0.28 | 71.93±1.09 | 24.54±1.28 | 38.73±1.27 | 55.62±0.39 | 66.03 ± 0.27 |
| gpt-oss-20B | 81.82±0.52 | 76.82±1.09 | 83.56±0.61 | 84.19±0.89 | 80.89±0.88 | 79.97±0.39 | 83.44±0.68 | 82.98±0.61 | 53.11±0.73 | 15.86±1.26 | 34.36±0.87 | 57.05±1.63 | 67.84 ± 0.24 |
| gpt-oss-120B | 82.23±0.68 | 87.21±0.26 | 84.71±0.60 | 86.07±0.32 | 82.29±0.18 | 82.71±0.18 | 85.29±0.29 | 85.70±0.40 | 72.53±1.18 | 25.42±0.76 | 48.85±0.95 | 65.45±0.39 | 74.04 ± 0.10 |
| LLaMA-3.1-8B | 50.57±0.54 | 66.56±0.67 | 61.71±0.66 | 60.61±0.74 | 52.70±1.37 | 57.93±1.55 | 62.37±0.43 | 63.23±0.81 | 38.53±1.15 | 23.84±0.43 | 22.57±1.18 | 18.14±0.78 | 48.23 ± 0.15 |
| LLaMA-3.1-70B | 66.27±0.62 | 77.45±0.74 | 67.85±0.43 | 67.61±0.50 | 65.86±0.60 | 67.23±0.77 | 68.10±0.52 | 67.88±0.38 | 61.96±0.72 | 32.90±1.11 | 34.73±1.39 | 36.19±0.95 | 59.50 ± 0.20 |
| LLaMA-3.2-3B | 30.59±0.98 | 63.46±1.11 | 49.88±1.15 | 40.78±0.30 | 36.44±0.74 | 31.01±0.40 | 50.42±1.49 | 53.90±0.69 | 33.29±1.26 | 20.80±0.76 | 22.85±0.93 | 24.62±0.51 | 38.17 ± 0.15 |
| LLaMA-3.3-70B | 61.49±0.61 | 82.16±0.35 | 67.65±0.44 | 67.31±0.45 | 66.58±0.37 | 69.79±0.52 | 67.23±0.27 | 67.78±0.54 | 64.08±0.56 | 33.47±0.76 | 37.08±0.86 | 39.49±0.64 | 60.34 ± 0.24 |
| LLaMA-4-Scout | 76.71±0.36 | 81.45±0.79 | 71.25±0.28 | 72.28±0.24 | 68.40±0.40 | 70.58±0.69 | 72.71±0.64 | 73.14±0.59 | 68.75±0.43 | 34.59±0.46 | 42.49±0.57 | 56.80±0.50 | 65.76 ± 0.20 |
| LLaMA-4-Maverick | 74.03±0.31 | 85.01±0.50 | 73.75±0.65 | 70.56±0.48 | 71.07±0.37 | 78.16±0.58 | 72.79±0.56 | 75.44±0.18 | 65.80±0.34 | 36.39±0.58 | 48.48±0.42 | 62.78±0.54 | 67.85 ± 0.06 |
| Mistral-7B-v0.3 | 37.22±0.50 | 13.69±0.71 | 19.82±0.79 | 41.54±1.38 | 28.31±0.91 | 24.83±0.35 | 34.68±1.22 | 41.27±0.74 | 16.09±0.52 | 12.76±0.56 | 11.32±0.18 | 11.76±0.71 | 24.44 ± 0.19 |
| Mistral-Small-3.1-24B | 75.69±0.84 | 83.56±0.75 | 73.73±0.90 | 68.58±0.76 | 63.95±0.79 | 64.12±0.71 | 73.38±0.66 | 74.17±0.86 | 30.15±0.87 | 12.15±1.33 | 11.46±0.86 | 19.92±0.51 | 54.24 ± 0.19 |
| Phi-4-mini | 64.33±0.67 | 77.51±0.47 | 68.52±0.78 | 66.94±0.78 | 61.85±0.69 | 45.12±1.09 | 65.94±0.86 | 68.17±0.94 | 32.56±1.40 | 18.47±0.57 | 21.54±0.61 | 23.67±1.08 | 51.22 ± 0.11 |
| Phi-4-mini-Reasoning | 49.33±0.43 | 75.65±0.74 | 58.34±0.79 | 60.86±1.35 | 21.36±0.92 | 17.60±1.33 | 40.69±1.13 | 39.79±1.06 | 30.43±1.14 | 24.38±1.09 | 25.00±0.95 | 26.59±0.62 | 39.17 ± 0.48 |
| Phi-4 | 75.92±0.62 | 83.88±0.24 | 77.91±0.93 | 79.39±0.72 | 77.80±0.78 | 74.16±0.67 | 76.64±0.71 | 78.47±0.67 | 47.84±0.92 | 23.90±1.01 | 31.68±1.18 | 30.09±0.53 | 63.14 ± 0.09 |
| Phi-4-Reasoning | 79.82±0.87 | 85.53±0.33 | 81.30±0.38 | 82.06±0.70 | 79.59±0.74 | 78.39±0.40 | 76.99±0.70 | 82.38±0.55 | 57.47±0.70 | 11.31±0.58 | 32.99±1.26 | 23.34±0.97 | 64.26 ± 0.27 |
| Qwen2.5-3B | 75.10±0.44 | 79.17±0.43 | 69.74±0.48 | 65.89±0.71 | 59.68±0.43 | 61.72±0.95 | 69.79±0.69 | 72.15±1.17 | 12.83±0.66 | 14.55±0.56 | 12.21±0.88 | 13.85±0.75 | 50.56 ± 0.20 |
| Qwen2.5-7B | 79.41±0.47 | 82.99±0.44 | 79.06±0.74 | 77.25±0.45 | 67.76±0.56 | 71.57±0.83 | 78.36±0.41 | 77.23±0.65 | 27.11±1.23 | 9.27±0.56 | 13.83±0.80 | 10.55±0.95 | 56.20 ± 0.38 |
| Qwen2.5-14B | 80.06±0.55 | 82.03±0.42 | 80.88±0.84 | 79.15±0.87 | 75.21±0.62 | 76.81±0.46 | 81.14±0.30 | 81.00±0.82 | 33.38±1.60 | 13.06±0.40 | 27.34±0.92 | 14.27±0.73 | 60.36 ± 0.28 |
| Qwen2.5-72B | 80.45±0.48 | 85.31±0.55 | 83.80±0.41 | 83.60±0.29 | 77.57±0.42 | 78.61±0.70 | 82.78±0.69 | 82.73±0.44 | 49.33±0.60 | 34.06±0.27 | 33.65±0.44 | 32.43±0.78 | 67.03 ± 0.15 |
| QwQ-32B | 77.71±0.31 | 83.81±0.27 | 76.68±0.58 | 77.61±0.61 | 75.47±0.37 | 78.12±0.63 | 77.08±0.52 | 77.22±0.46 | 34.14±1.00 | 28.13±1.93 | 31.02±0.94 | 31.81±0.64 | 62.40 ± 0.34 |
| Qwen3-1.7B | 68.44±0.96 | 75.29±0.66 | 66.58±0.43 | 62.15±0.82 | 54.20±0.75 | 49.33±0.60 | 65.14±0.63 | 63.79±0.91 | 27.38±1.61 | 23.59±0.96 | 30.61±1.47 | 32.43±1.20 | 51.58 ± 0.32 |
| Qwen3-4B | 76.78±0.51 | 82.33±0.69 | 68.57±0.78 | 57.73±1.32 | 72.90±1.18 | 71.87±0.50 | 75.40±0.68 | 73.37±0.71 | 17.78±1.22 | 18.53±1.25 | 11.01±0.71 | 14.99±0.92 | 53.44 ± 0.11 |
| Qwen3-4B-thinking | 78.77±0.65 | 84.53±0.29 | 78.05±0.77 | 76.06±0.42 | 78.52±0.62 | 76.79±0.63 | 76.67±0.57 | 76.84±0.59 | 23.82±0.43 | 11.90±0.58 | 11.97±0.78 | 11.49±0.82 | 57.12 ± 0.25 |
| Qwen3-8B | 75.98±1.16 | 77.49±0.50 | 72.01±0.49 | 69.49±0.81 | 77.15±0.63 | 72.55±0.31 | 70.97±0.36 | 70.90±1.01 | 27.30±0.78 | 13.06±1.01 | 24.80±0.88 | 20.24±0.48 | 55.99 ± 0.18 |
| Qwen3-8B-thinking | 77.85±0.41 | 82.29±0.48 | 80.20±0.62 | 79.73±0.51 | 78.29±0.40 | 75.65±0.76 | 80.20±0.86 | 80.42±0.71 | 31.52±1.52 | 8.33±0.59 | 19.17±1.67 | 15.24±0.98 | 59.07 ± 0.31 |
| Qwen3-14B | 80.72±0.42 | 82.72±0.56 | 81.33±0.50 | 77.88±0.78 | 79.06±0.51 | 78.19±0.49 | 81.57±0.74 | 82.53±0.30 | 44.42±0.88 | 21.27±0.96 | 22.81±0.97 | 32.14±1.52 | 63.72 ± 0.23 |
| Qwen3-14B-thinking | 80.49±0.71 | 84.69±0.41 | 80.59±0.53 | 79.25±0.61 | 79.52±0.95 | 78.62±0.58 | 81.86±0.75 | 80.99±0.75 | 53.27±1.04 | 8.55±0.42 | 17.37±0.58 | 28.89±0.63 | 62.84 ± 0.13 |
| Baichuan-M2-32B | 76.61±0.24 | 80.88±0.29 | 78.55±0.38 | 77.59±0.80 | 74.61±0.69 | 73.32±0.44 | 78.19±0.36 | 78.05±0.32 | 30.67±1.50 | 19.55±0.31 | 30.53±1.52 | 28.69±2.05 | 60.60 ± 0.27 |
| Bio-Medical-LLaMA-3-8B | 40.38±0.55 | 57.53±0.97 | 44.53±0.99 | 39.49±0.45 | 38.57±0.31 | 33.69±0.77 | 38.51±0.64 | 40.30±0.71 | 32.67±0.69 | 33.51±0.83 | 33.07±1.11 | 33.17±0.93 | 38.78 ± 0.22 |
| MediPhi | 29.33±1.28 | 74.04±0.94 | 61.41±0.89 | 56.37±0.62 | 36.73±0.24 | 35.47±0.63 | 56.75±0.60 | 58.73±0.30 | 18.19±0.61 | 27.75±0.55 | 11.98±0.87 | 21.20±1.93 | 40.66 ± 0.16 |
| MedGemma-4B | 58.70±0.53 | 72.20±0.68 | 60.00±0.84 | 61.92±0.79 | 59.16±0.46 | 58.16±0.72 | 59.66±0.63 | 63.66±0.78 | 44.26±1.44 | 15.23±0.58 | 31.80±0.95 | 31.39±1.30 | 51.34 ± 0.21 |
| MedGemma-27B | 77.19±0.34 | 81.58±0.74 | 74.30±0.51 | 74.86±0.36 | 75.03±0.53 | 77.84±0.71 | 76.17±0.54 | 77.26±0.51 | 67.96±0.44 | 26.27±0.53 | 36.51±1.20 | 54.88±0.87 | 66.66 ± 0.19 |
| MedReason-8B | 41.61±0.92 | 11.25±0.48 | 8.75±0.32 | 9.56±0.50 | 40.85±1.40 | 45.33±0.76 | 23.77±0.29 | 3.66±0.29 | 8.44±0.99 | 11.32±0.51 | 24.25±0.71 | 24.53±0.64 | 21.11 ± 0.27 |
| HuatuoGPT-o1-7B | 73.71±1.13 | 79.32±1.05 | 71.21±0.33 | 72.52±0.53 | 57.18±0.56 | 65.87±0.85 | 72.67±0.39 | 73.97±0.56 | 5.66±0.61 | 5.58±0.50 | 7.96±0.88 | 4.98±0.37 | 49.22 ± 0.12 |
| HuatuoGPT-o1-8B | 53.14±1.03 | 60.78±0.82 | 58.17±0.48 | 59.45±0.66 | 45.80±1.43 | 58.17±0.88 | 59.05±1.09 | 59.03±0.67 | 32.63±1.56 | 5.91±0.53 | 11.04±1.24 | 9.95±0.85 | 42.76 ± 0.20 |
| HuatuoGPT-o1-70B | 67.28±0.34 | 79.24±0.75 | 66.79±0.36 | 66.27±0.16 | 67.58±0.65 | 66.96±1.00 | 66.76±0.40 | 67.44±0.80 | 62.03±0.86 | 31.38±0.88 | 34.00±1.21 | 39.58±0.83 | 59.61 ± 0.14 |
| HuatuoGPT-o1-72B | 77.98±0.63 | 84.57±0.34 | 80.87±0.41 | 78.70±0.60 | 74.19±0.46 | 77.09±0.55 | 79.44±0.53 | 80.14±0.31 | 47.41±0.31 | 19.27±0.93 | 28.76±1.04 | 19.68±0.65 | 62.34 ± 0.15 |
| OpenBioLLM-8B | 5.59±0.93 | 20.93±1.12 | 11.95±0.76 | 17.06±0.74 | 5.12±0.56 | 6.31±1.13 | 20.14±1.27 | 21.20±0.63 | 12.29±0.64 | 9.67±0.79 | 10.22±0.52 | 15.62±0.76 | 13.01 ± 0.16 |
| OpenBioLLM-70B | 28.23±0.80 | 72.63±0.61 | 58.03±0.82 | 30.44±0.83 | 29.19±1.36 | 31.16±1.31 | 52.11±1.59 | 62.29±0.57 | 23.02±1.23 | 12.01±1.10 | 14.55±0.52 | 19.19±0.73 | 36.07 ± 0.49 |

**Supplementary Tab. 97:** Performance evaluation of 56 LLMs on MedNLI.

| LLMs | Chinese | English | French | German | Japanese | Korean | Portuguese | Spanish | Swahili | Wolof | Yoruba | Zulu |
|---|---|---|---|---|---|---|---|---|---|---|---|---|
| **Proprietary LLMs** | | | | | | | | | | | | |
| **Claude-3.5-Haiku** | 69.58 | 78.33 | 67.68 | 66.69 | 68.24 | 68.67 | 67.11 | 68.88 | 52.01 | 9.74 | 35.22 | 33.31 |
| **Claude-4.0-Sonnet** | 80.45 | 86.24 | 79.39 | 75.51 | 78.69 | 80.03 | 75.44 | 78.76 | 68.45 | 35.07 | 58.01 | 56.81 |
| **Gemini-2.5-Flash** | 71.77 | 83.27 | 70.15 | 69.37 | 68.17 | 69.44 | 69.72 | 70.15 | 67.82 | 47.21 | 59.28 | 62.24 |
| **GPT-4o-mini** | 80.31 | 85.11 | 82.07 | 78.76 | 77.42 | 77.56 | 81.79 | 79.96 | 62.81 | 31.05 | 35.14 | 51.31 |
| **GPT-4o** | 79.04 | 80.80 | 76.36 | 75.72 | 74.81 | 77.06 | 78.12 | 75.30 | 70.92 | 26.53 | 46.37 | 60.76 |
| **GPT-4.1-nano** | 70.85 | 82.99 | 75.94 | 74.66 | 71.28 | 69.02 | 72.12 | 74.31 | 56.60 | 24.84 | 33.73 | 46.65 |
| **GPT-4.1-mini** | 82.43 | 87.44 | 85.11 | 83.27 | 78.83 | 80.66 | 82.43 | 83.06 | 63.09 | 19.76 | 46.44 | 60.27 |
| **GPT-4.1** | 82.71 | 87.58 | 85.11 | 83.84 | 81.93 | 82.07 | 85.18 | 85.67 | 71.28 | 37.69 | 50.53 | 62.81 |
| **GPT-5-nano** | 73.25 | 79.18 | 70.29 | 78.19 | 72.12 | 69.51 | 78.97 | 81.51 | 52.51 | 16.94 | 31.55 | 41.07 |
| **GPT-5-mini** | 84.26 | 88.64 | 88.07 | 87.01 | 83.42 | 82.71 | 86.66 | 87.30 | 71.63 | 31.40 | 46.65 | 63.16 |
| **GPT-5** | 80.24 | 86.31 | 85.74 | 85.39 | 81.37 | 84.47 | 84.90 | 85.74 | 79.60 | 44.32 | 54.27 | 65.28 |
| **o4-mini** | 84.40 | 89.20 | 87.58 | 88.07 | 84.90 | 83.84 | 87.09 | 87.86 | 78.55 | 23.08 | 53.00 | 67.89 |
| **Open-Weight LLMs** | | | | | | | | | | | | |
| **DeepSeek-V3** | 78.69 | 83.77 | 78.33 | 78.76 | 75.09 | 74.81 | 77.35 | 78.55 | 68.53 | 33.38 | 46.08 | 51.02 |
| **DeepSeek-R1** | 78.41 | 82.92 | 75.23 | 77.28 | 76.29 | 72.76 | 74.38 | 77.49 | 59.56 | 35.29 | 45.66 | 55.26 |
| **DeepSeek-R1-Qwen3-8B** | 76.50 | 80.31 | 75.16 | 75.58 | 75.37 | 71.91 | 76.22 | 76.50 | 30.63 | 12.00 | 13.20 | 16.65 |
| **Gemma-3-4B** | 63.87 | 74.38 | 61.12 | 59.84 | 58.50 | 62.60 | 63.30 | 63.37 | 44.46 | 7.69 | 29.22 | 21.03 |
| **Gemma-3-12B** | 63.02 | 81.23 | 67.40 | 66.27 | 68.38 | 69.80 | 64.57 | 67.04 | 64.71 | 8.19 | 35.57 | 49.68 |
| **Gemma-3-27B** | 76.50 | 82.64 | 70.43 | 73.47 | 74.24 | 77.28 | 72.69 | 73.18 | 71.77 | 24.28 | 37.40 | 55.96 |
| **gpt-oss-20B** | 81.58 | 77.77 | 82.64 | 83.27 | 81.02 | 79.53 | 83.35 | 83.63 | 53.28 | 13.97 | 33.17 | 57.80 |
| **gpt-oss-120B** | 81.72 | 86.87 | 85.60 | 85.96 | 82.29 | 82.64 | 85.04 | 85.67 | 72.62 | 24.49 | 47.71 | 66.06 |
| **LLaMA-3.1-8B** | 50.81 | 65.63 | 61.47 | 60.62 | 53.49 | 55.82 | 61.75 | 62.67 | 39.03 | 23.71 | 21.81 | 18.98 |
| **LLaMA-3.1-70B** | 65.91 | 77.35 | 67.75 | 67.04 | 66.41 | 66.41 | 67.75 | 67.47 | 61.68 | 33.45 | 34.51 | 35.85 |
| **LLaMA-3.2-3B** | 31.05 | 62.17 | 49.40 | 41.14 | 35.57 | 31.40 | 51.66 | 54.55 | 35.00 | 20.89 | 21.95 | 24.14 |
| **LLaMA-3.3-70B** | 62.24 | 82.64 | 68.31 | 67.04 | 65.98 | 70.57 | 66.90 | 68.24 | 64.50 | 34.16 | 36.56 | 39.66 |
| **LLaMA-4-Scout** | 76.71 | 80.80 | 71.28 | 72.19 | 68.24 | 70.78 | 71.98 | 73.75 | 68.31 | 33.94 | 42.70 | 56.18 |
| **LLaMA-4-Maverick** | 74.31 | 84.76 | 74.45 | 70.15 | 71.00 | 77.91 | 72.76 | 75.44 | 65.63 | 36.63 | 48.84 | 62.67 |
| **Mistral-7B-v0.3** | 37.12 | 14.61 | 19.12 | 43.19 | 27.95 | 24.56 | 36.63 | 41.92 | 16.09 | 12.42 | 11.08 | 12.49 |
| **Mistral-Small-3.1-24B** | 76.22 | 84.40 | 73.82 | 69.72 | 63.80 | 65.21 | 72.41 | 74.17 | 30.63 | 12.14 | 12.63 | 19.48 |
| **Phi-4-mini** | 64.57 | 77.42 | 67.96 | 67.04 | 61.68 | 46.65 | 65.28 | 69.02 | 34.09 | 18.91 | 21.17 | 22.37 |
| **Phi-4-mini-Reasoning** | 48.84 | 75.51 | 58.72 | 61.12 | 22.72 | 18.98 | 41.99 | 40.86 | 32.39 | 24.56 | 26.39 | 27.17 |
| **Phi-4** | 75.86 | 83.98 | 78.33 | 79.75 | 78.55 | 74.66 | 77.13 | 78.26 | 46.51 | 23.01 | 31.26 | 29.43 |
| **Phi-4-Reasoning** | 79.32 | 85.32 | 81.65 | 82.50 | 80.66 | 78.33 | 77.13 | 82.99 | 56.81 | 10.87 | 34.93 | 24.06 |
| **Qwen2.5-3B** | 75.79 | 79.46 | 69.80 | 66.27 | 59.21 | 60.41 | 69.58 | 73.18 | 13.34 | 13.97 | 13.76 | 14.04 |
| **Qwen2.5-7B** | 79.60 | 83.13 | 79.68 | 77.77 | 68.03 | 71.14 | 77.70 | 77.06 | 28.86 | 9.32 | 14.54 | 10.94 |
| **Qwen2.5-14B** | 80.59 | 82.36 | 81.58 | 79.82 | 75.79 | 77.49 | 80.88 | 80.38 | 32.96 | 13.48 | 26.89 | 13.62 |
| **Qwen2.5-72B** | 80.95 | 84.47 | 83.70 | 83.84 | 77.70 | 79.11 | 82.78 | 82.43 | 49.61 | 33.73 | 34.30 | 33.45 |
| **QwQ-32B** | 77.35 | 84.05 | 76.15 | 76.85 | 75.65 | 77.91 | 77.13 | 76.71 | 34.02 | 25.41 | 30.28 | 30.91 |
| **Qwen3-1.7B** | 68.03 | 75.37 | 66.69 | 61.54 | 54.20 | 49.26 | 64.22 | 63.30 | 27.45 | 22.23 | 31.90 | 33.17 |
| **Qwen3-4B** | 76.71 | 82.99 | 67.68 | 58.65 | 72.34 | 71.63 | 75.44 | 72.34 | 18.35 | 19.97 | 10.80 | 14.68 |
| **Qwen3-4B-thinking** | 78.48 | 84.90 | 77.70 | 75.51 | 77.49 | 76.64 | 75.65 | 75.79 | 23.57 | 12.14 | 12.91 | 10.94 |
| **Qwen3-8B** | 76.64 | 77.42 | 72.12 | 68.45 | 77.63 | 72.90 | 71.21 | 70.71 | 27.52 | 14.11 | 24.35 | 20.89 |
| **Qwen3-8B-thinking** | 77.98 | 82.15 | 80.31 | 80.24 | 78.33 | 74.74 | 81.23 | 81.44 | 31.90 | 8.82 | 17.64 | 15.31 |
| **Qwen3-14B** | 81.16 | 82.92 | 81.30 | 78.76 | 79.46 | 78.05 | 80.80 | 82.43 | 45.10 | 20.68 | 22.65 | 31.26 |
| **Qwen3-14B-thinking** | 81.37 | 84.33 | 80.31 | 79.04 | 79.46 | 79.39 | 81.72 | 80.52 | 52.51 | 8.19 | 16.44 | 28.93 |
| **Baichuan-M2-32B** | 76.57 | 80.88 | 78.41 | 76.57 | 73.82 | 73.39 | 77.70 | 77.70 | 30.20 | 19.34 | 31.26 | 28.86 |
| **Bio-Medical-LLaMA-3-8B** | 40.79 | 57.94 | 45.80 | 39.31 | 39.03 | 34.79 | 38.18 | 39.94 | 32.32 | 32.39 | 33.52 | 34.23 |
| **MediPhi** | 30.77 | 73.25 | 62.39 | 56.88 | 36.34 | 35.64 | 55.96 | 58.65 | 17.71 | 28.23 | 11.64 | 22.02 |
| **MedGemma-4B** | 59.35 | 72.05 | 59.00 | 62.31 | 59.07 | 57.73 | 59.77 | 64.29 | 45.73 | 15.60 | 30.28 | 30.77 |
| **MedGemma-27B** | 77.21 | 81.51 | 74.74 | 74.66 | 74.88 | 78.76 | 75.51 | 76.85 | 67.54 | 26.46 | 37.40 | 55.33 |
| **MedReason-8B** | 41.92 | 11.93 | 9.10 | 9.46 | 41.99 | 45.02 | 23.64 | 3.53 | 8.33 | 11.15 | 24.91 | 25.19 |
| **HuatuoGPT-o1-7B** | 74.03 | 78.05 | 71.63 | 72.69 | 57.23 | 65.35 | 72.05 | 74.59 | 5.72 | 4.80 | 8.68 | 5.58 |
| **HuatuoGPT-o1-8B** | 54.06 | 61.40 | 57.73 | 59.14 | 43.75 | 57.02 | 59.14 | 60.06 | 31.62 | 5.50 | 9.10 | 10.80 |
| **HuatuoGPT-o1-70B** | 67.33 | 80.24 | 66.62 | 66.13 | 66.69 | 67.75 | 66.27 | 67.96 | 62.31 | 31.40 | 34.44 | 39.31 |
| **HuatuoGPT-o1-72B** | 77.84 | 84.69 | 81.44 | 79.32 | 74.45 | 76.78 | 79.89 | 79.89 | 47.85 | 19.76 | 27.73 | 19.97 |
| **OpenBioLLM-8B** | 4.30 | 20.18 | 11.43 | 17.50 | 5.36 | 7.69 | 21.38 | 20.89 | 12.00 | 9.88 | 9.74 | 15.10 |
| **OpenBioLLM-70B** | 27.52 | 73.32 | 57.52 | 31.33 | 29.36 | 32.67 | 54.41 | 63.02 | 25.19 | 13.34 | 15.03 | 20.47 |

**Supplementary Tab. 98:** Zero-Shot performance evaluation of 56 LLMs on MedNLI (Run 1).

| LLMs | Chinese | English | French | German | Japanese | Korean | Portuguese | Spanish | Swahili | Wolof | Yoruba | Zulu |
|---|---|---|---|---|---|---|---|---|---|---|---|---|
| **Proprietary LLMs** | | | | | | | | | | | | |
| **Claude-3.5-Haiku** | 68.45 | 77.84 | 67.68 | 65.63 | 68.03 | 69.87 | 66.62 | 69.02 | 50.32 | 10.30 | 33.24 | 33.66 |
| **Claude-4.0-Sonnet** | 81.37 | 85.67 | 79.82 | 76.01 | 77.91 | 79.96 | 76.01 | 78.19 | 68.24 | 34.23 | 57.23 | 57.23 |
| **Gemini-2.5-Flash** | 70.57 | 84.54 | 70.36 | 70.08 | 67.82 | 69.02 | 68.67 | 70.01 | 67.11 | 46.22 | 59.99 | 62.17 |
| **GPT-4o-mini** | 79.39 | 85.25 | 80.17 | 79.11 | 77.21 | 76.78 | 81.16 | 79.89 | 62.24 | 27.52 | 35.07 | 51.24 |
| **GPT-4o** | 77.91 | 83.27 | 74.45 | 75.58 | 77.91 | 76.92 | 78.69 | 74.81 | 71.28 | 27.03 | 45.73 | 61.54 |
| **GPT-4.1-nano** | 72.48 | 83.13 | 74.45 | 74.95 | 70.64 | 69.80 | 72.62 | 75.79 | 58.15 | 23.99 | 33.52 | 49.19 |
| **GPT-4.1-mini** | 83.27 | 87.37 | 85.89 | 83.13 | 79.32 | 81.16 | 83.06 | 82.99 | 63.66 | 18.49 | 47.28 | 59.63 |
| **GPT-4.1** | 83.63 | 87.44 | 84.97 | 85.60 | 82.36 | 82.71 | 85.11 | 86.52 | 71.84 | 39.66 | 48.41 | 63.66 |
| **GPT-5-nano** | 72.69 | 78.62 | 70.01 | 78.26 | 72.05 | 68.10 | 78.48 | 81.86 | 55.33 | 17.57 | 32.18 | 41.07 |
| **GPT-5-mini** | 83.77 | 88.43 | 87.16 | 86.73 | 83.63 | 83.91 | 87.01 | 87.37 | 72.97 | 31.97 | 46.37 | 62.60 |
| **GPT-5** | 81.72 | 86.38 | 85.89 | 86.24 | 82.15 | 83.27 | 84.40 | 85.82 | 80.31 | 43.68 | 53.14 | 66.27 |
| **o4-mini** | 85.67 | 88.14 | 87.30 | 87.30 | 85.04 | 84.47 | 87.51 | 88.50 | 79.82 | 20.18 | 51.02 | 67.47 |
| **Open-Weight LLMs** | | | | | | | | | | | | |
| **DeepSeek-V3** | 78.97 | 84.40 | 77.21 | 79.89 | 74.52 | 74.95 | 76.50 | 78.97 | 68.38 | 34.51 | 45.59 | 53.35 |
| **DeepSeek-R1** | 78.12 | 83.20 | 74.17 | 76.64 | 76.50 | 74.17 | 74.31 | 77.28 | 58.79 | 35.07 | 45.80 | 54.34 |
| **DeepSeek-R1-Qwen3-8B** | 77.91 | 81.23 | 75.51 | 75.37 | 75.09 | 71.84 | 77.42 | 75.79 | 32.11 | 12.42 | 13.06 | 15.46 |
| **Gemma-3-4B** | 66.48 | 73.25 | 61.33 | 58.86 | 56.88 | 62.46 | 62.81 | 62.39 | 45.45 | 7.69 | 29.85 | 20.18 |
| **Gemma-3-12B** | 63.80 | 82.00 | 67.82 | 66.27 | 68.17 | 69.94 | 64.71 | 66.69 | 65.28 | 6.70 | 36.13 | 48.27 |
| **Gemma-3-27B** | 76.57 | 82.78 | 70.29 | 73.47 | 73.96 | 77.98 | 72.83 | 73.25 | 71.49 | 22.94 | 38.39 | 55.47 |
| **gpt-oss-20B** | 81.44 | 75.16 | 83.56 | 84.47 | 80.38 | 80.45 | 82.85 | 82.00 | 53.56 | 15.81 | 34.23 | 59.14 |
| **gpt-oss-120B** | 81.72 | 87.09 | 84.83 | 85.82 | 82.50 | 82.71 | 85.04 | 86.03 | 72.12 | 26.53 | 49.75 | 65.00 |
| **LLaMA-3.1-8B** | 49.89 | 67.18 | 62.17 | 59.70 | 54.69 | 57.45 | 62.60 | 64.64 | 38.39 | 23.64 | 21.88 | 18.42 |
| **LLaMA-3.1-70B** | 65.35 | 77.28 | 67.75 | 68.10 | 66.13 | 66.69 | 68.88 | 68.10 | 62.60 | 32.18 | 34.72 | 37.47 |
| **LLaMA-3.2-3B** | 29.57 | 64.57 | 51.09 | 40.93 | 37.33 | 30.56 | 50.67 | 54.69 | 31.69 | 20.04 | 23.57 | 24.84 |
| **LLaMA-3.3-70B** | 60.90 | 81.79 | 67.75 | 68.10 | 66.83 | 69.65 | 66.97 | 67.68 | 64.57 | 33.73 | 37.90 | 39.66 |
| **LLaMA-4-Scout** | 77.28 | 80.95 | 70.85 | 72.19 | 68.60 | 70.57 | 73.32 | 72.34 | 68.74 | 35.22 | 41.85 | 57.16 |
| **LLaMA-4-Maverick** | 73.54 | 85.53 | 72.97 | 70.15 | 70.78 | 78.33 | 73.61 | 75.65 | 65.63 | 36.27 | 48.13 | 62.53 |
| **Mistral-7B-v0.3** | 37.97 | 13.13 | 20.32 | 39.59 | 28.65 | 24.35 | 34.79 | 41.14 | 16.65 | 12.07 | 11.22 | 12.42 |
| **Mistral-Small-3.1-24B** | 74.59 | 83.13 | 72.97 | 68.38 | 62.88 | 63.94 | 73.25 | 74.45 | 30.42 | 13.97 | 11.36 | 20.68 |
| **Phi-4-mini** | 65.35 | 77.13 | 69.37 | 67.96 | 61.12 | 44.88 | 65.77 | 67.89 | 31.47 | 18.84 | 22.23 | 22.79 |
| **Phi-4-mini-Reasoning** | 49.12 | 74.59 | 58.15 | 58.65 | 21.10 | 16.51 | 39.73 | 40.72 | 29.57 | 23.50 | 24.98 | 27.17 |
| **Phi-4** | 75.94 | 83.70 | 77.98 | 80.03 | 78.62 | 73.68 | 76.22 | 78.69 | 49.05 | 24.49 | 30.20 | 30.42 |
| **Phi-4-Reasoning** | 78.97 | 85.32 | 80.73 | 81.44 | 79.46 | 78.69 | 76.57 | 82.85 | 57.16 | 10.52 | 32.18 | 22.09 |
| **Qwen2.5-3B** | 75.16 | 78.48 | 68.95 | 66.27 | 60.06 | 62.31 | 69.58 | 72.55 | 11.86 | 14.40 | 11.86 | 12.63 |
| **Qwen2.5-7B** | 80.10 | 83.49 | 78.48 | 77.63 | 67.82 | 72.12 | 78.69 | 77.98 | 27.73 | 9.46 | 14.26 | 11.86 |
| **Qwen2.5-14B** | 79.75 | 81.79 | 81.16 | 80.31 | 75.02 | 76.78 | 81.44 | 79.89 | 31.62 | 12.91 | 26.39 | 14.33 |
| **Qwen2.5-72B** | 80.88 | 85.96 | 83.35 | 83.63 | 77.06 | 78.05 | 83.49 | 82.78 | 48.84 | 33.87 | 33.45 | 31.47 |
| **QwQ-32B** | 77.77 | 83.49 | 76.78 | 77.13 | 74.95 | 77.13 | 76.36 | 76.92 | 35.85 | 28.58 | 31.26 | 31.47 |
| **Qwen3-1.7B** | 67.47 | 76.29 | 66.06 | 63.37 | 53.00 | 49.82 | 65.21 | 63.44 | 27.38 | 22.94 | 30.70 | 32.53 |
| **Qwen3-4B** | 77.21 | 82.07 | 69.58 | 55.96 | 73.39 | 71.35 | 74.95 | 73.82 | 17.64 | 16.73 | 10.94 | 16.30 |
| **Qwen3-4B-thinking** | 79.04 | 84.40 | 76.99 | 76.08 | 78.90 | 77.13 | 76.78 | 77.13 | 23.92 | 11.64 | 11.29 | 11.93 |
| **Qwen3-8B** | 76.57 | 76.85 | 72.27 | 70.29 | 77.56 | 72.69 | 71.00 | 71.56 | 28.51 | 13.41 | 23.57 | 19.83 |
| **Qwen3-8B-thinking** | 77.13 | 82.99 | 79.89 | 80.03 | 78.05 | 75.51 | 79.32 | 80.52 | 32.67 | 8.05 | 19.62 | 15.53 |
| **Qwen3-14B** | 80.52 | 82.36 | 82.07 | 76.92 | 78.41 | 78.55 | 82.78 | 82.99 | 44.04 | 21.52 | 21.24 | 31.83 |
| **Qwen3-14B-thinking** | 80.59 | 84.83 | 80.03 | 78.90 | 78.33 | 78.12 | 82.64 | 81.79 | 54.90 | 8.82 | 17.64 | 28.23 |
| **Baichuan-M2-32B** | 76.78 | 81.30 | 78.33 | 77.56 | 74.24 | 74.03 | 78.33 | 78.19 | 29.08 | 19.48 | 30.84 | 25.76 |
| **Bio-Medical-LLaMA-3-8B** | 41.07 | 58.15 | 44.32 | 39.45 | 38.25 | 34.16 | 37.76 | 39.31 | 32.60 | 33.52 | 31.83 | 31.83 |
| **MediPhi** | 28.44 | 75.65 | 61.33 | 57.09 | 36.98 | 34.79 | 56.46 | 58.72 | 17.71 | 28.44 | 10.94 | 19.90 |
| **MedGemma-4B** | 59.00 | 72.27 | 59.56 | 62.88 | 59.14 | 58.50 | 58.86 | 62.67 | 42.70 | 15.74 | 31.90 | 31.55 |
| **MedGemma-27B** | 76.64 | 81.37 | 73.96 | 75.30 | 75.58 | 78.12 | 76.85 | 77.98 | 68.53 | 26.25 | 35.22 | 55.96 |
| **MedReason-8B** | 42.98 | 10.87 | 8.47 | 9.53 | 39.73 | 44.39 | 24.14 | 3.32 | 8.54 | 11.79 | 24.06 | 23.85 |
| **HuatuoGPT-o1-7B** | 72.76 | 79.75 | 71.14 | 73.32 | 58.08 | 66.48 | 72.97 | 73.39 | 5.15 | 5.43 | 8.82 | 4.66 |
| **HuatuoGPT-o1-8B** | 51.80 | 60.90 | 58.22 | 58.72 | 46.15 | 58.86 | 59.77 | 58.43 | 34.58 | 6.28 | 10.87 | 8.75 |
| **HuatuoGPT-o1-70B** | 67.11 | 79.75 | 67.04 | 66.41 | 68.53 | 68.10 | 67.18 | 66.97 | 62.24 | 30.77 | 33.66 | 39.10 |
| **HuatuoGPT-o1-72B** | 78.55 | 84.33 | 80.45 | 78.62 | 73.75 | 78.05 | 79.18 | 80.59 | 47.21 | 19.97 | 30.20 | 19.27 |
| **OpenBioLLM-8B** | 6.07 | 20.75 | 13.27 | 16.44 | 5.29 | 6.56 | 19.90 | 22.30 | 12.07 | 8.33 | 9.74 | 16.51 |
| **OpenBioLLM-70B** | 28.16 | 73.18 | 58.79 | 31.19 | 28.23 | 30.28 | 50.53 | 62.10 | 22.23 | 12.28 | 15.17 | 18.70 |

**Supplementary Tab. 99:** Zero-Shot performance evaluation of 56 LLMs on MedNLI (Run 2).

| LLMs | Chinese | English | French | German | Japanese | Korean | Portuguese | Spanish | Swahili | Wolof | Yoruba | Zulu |
|---|---|---|---|---|---|---|---|---|---|---|---|---|
| **Proprietary LLMs** | | | | | | | | | | | | |
| **Claude-3.5-Haiku** | 69.44 | 78.76 | 67.33 | 66.27 | 67.25 | 69.72 | 67.11 | 68.38 | 51.45 | 9.39 | 34.58 | 31.19 |
| **Claude-4.0-Sonnet** | 81.51 | 85.18 | 79.18 | 75.44 | 78.76 | 79.89 | 76.64 | 78.12 | 68.53 | 34.86 | 58.43 | 57.23 |
| **Gemini-2.5-Flash** | 72.48 | 82.71 | 70.78 | 70.01 | 67.47 | 70.08 | 69.16 | 69.58 | 66.97 | 47.14 | 58.65 | 63.02 |
| **GPT-4o-mini** | 78.12 | 83.91 | 82.29 | 80.38 | 78.83 | 76.01 | 80.10 | 79.04 | 61.26 | 28.86 | 34.16 | 51.24 |
| **GPT-4o** | 77.98 | 82.22 | 74.74 | 75.09 | 75.79 | 76.64 | 79.82 | 75.16 | 71.21 | 25.41 | 47.64 | 60.27 |
| **GPT-4.1-nano** | 72.55 | 83.70 | 73.89 | 74.88 | 72.34 | 69.44 | 72.34 | 76.15 | 55.05 | 25.41 | 33.10 | 46.93 |
| **GPT-4.1-mini** | 82.50 | 88.07 | 85.32 | 83.06 | 78.62 | 80.73 | 82.64 | 82.57 | 63.51 | 19.12 | 48.41 | 58.86 |
| **GPT-4.1** | 82.99 | 88.00 | 85.25 | 85.18 | 81.86 | 82.43 | 84.90 | 86.52 | 71.28 | 38.60 | 49.82 | 63.66 |
| **GPT-5-nano** | 74.03 | 76.78 | 69.09 | 80.24 | 70.92 | 70.43 | 78.62 | 79.32 | 55.40 | 16.65 | 31.76 | 42.48 |
| **GPT-5-mini** | 84.12 | 88.36 | 87.72 | 86.87 | 83.56 | 83.13 | 86.52 | 86.66 | 73.32 | 33.03 | 47.49 | 63.94 |
| **GPT-5** | 81.09 | 86.31 | 85.74 | 85.74 | 82.99 | 83.06 | 85.18 | 85.82 | 80.52 | 42.55 | 53.85 | 65.28 |
| **o4-mini** | 84.97 | 89.20 | 88.14 | 87.37 | 85.25 | 84.54 | 87.51 | 87.51 | 78.26 | 23.08 | 52.58 | 67.47 |
| **Open-Weight LLMs** | | | | | | | | | | | | |
| **DeepSeek-V3** | 77.98 | 84.69 | 77.91 | 78.19 | 75.30 | 75.09 | 76.36 | 78.19 | 68.03 | 34.79 | 45.94 | 50.25 |
| **DeepSeek-R1** | 77.70 | 83.70 | 74.74 | 76.85 | 75.86 | 73.11 | 74.88 | 78.19 | 59.99 | 35.29 | 46.65 | 54.06 |
| **DeepSeek-R1-Qwen3-8B** | 77.56 | 79.96 | 77.91 | 75.44 | 76.36 | 71.07 | 75.79 | 76.08 | 30.20 | 12.35 | 11.57 | 16.87 |
| **Gemma-3-4B** | 64.93 | 73.89 | 60.90 | 57.66 | 58.65 | 63.51 | 62.67 | 61.54 | 46.51 | 7.13 | 29.85 | 22.30 |
| **Gemma-3-12B** | 64.78 | 82.99 | 65.77 | 67.25 | 68.38 | 69.16 | 64.78 | 66.20 | 66.13 | 7.83 | 35.85 | 50.39 |
| **Gemma-3-27B** | 77.84 | 82.22 | 69.16 | 74.03 | 73.47 | 77.91 | 72.97 | 73.11 | 71.98 | 23.99 | 40.30 | 55.19 |
| **gpt-oss-20B** | 82.57 | 77.84 | 83.77 | 83.35 | 80.52 | 80.17 | 82.92 | 82.92 | 52.36 | 16.02 | 35.57 | 54.69 |
| **gpt-oss-120B** | 81.86 | 87.51 | 84.83 | 85.82 | 82.36 | 82.50 | 85.53 | 85.25 | 74.38 | 25.05 | 49.12 | 65.35 |
| **LLaMA-3.1-8B** | 51.31 | 66.41 | 61.33 | 60.55 | 51.31 | 60.06 | 62.60 | 62.95 | 40.23 | 23.29 | 21.45 | 18.28 |
| **LLaMA-3.1-70B** | 66.69 | 77.42 | 67.47 | 67.18 | 65.00 | 68.38 | 67.89 | 67.89 | 62.31 | 33.24 | 32.60 | 35.85 |
| **LLaMA-3.2-3B** | 30.56 | 63.80 | 48.13 | 40.86 | 35.78 | 31.05 | 51.45 | 53.21 | 33.45 | 22.02 | 23.29 | 25.41 |
| **LLaMA-3.3-70B** | 60.83 | 82.36 | 67.47 | 67.04 | 66.69 | 69.16 | 67.47 | 68.17 | 63.37 | 32.46 | 36.56 | 38.39 |
| **LLaMA-4-Scout** | 76.64 | 82.00 | 71.28 | 72.69 | 67.82 | 71.42 | 73.39 | 72.83 | 69.37 | 34.44 | 43.26 | 57.23 |
| **LLaMA-4-Maverick** | 74.10 | 84.83 | 73.89 | 70.43 | 71.63 | 78.90 | 72.12 | 75.30 | 66.20 | 35.85 | 48.76 | 63.09 |
| **Mistral-7B-v0.3** | 37.26 | 14.18 | 19.34 | 41.21 | 27.03 | 24.98 | 33.31 | 40.72 | 16.51 | 13.48 | 11.50 | 11.79 |
| **Mistral-Small-3.1-24B** | 76.78 | 84.05 | 73.75 | 68.88 | 64.36 | 64.15 | 73.75 | 72.76 | 29.43 | 12.00 | 11.36 | 19.48 |
| **Phi-4-mini** | 64.22 | 77.06 | 67.82 | 66.62 | 62.95 | 44.11 | 65.35 | 67.40 | 34.09 | 17.78 | 21.81 | 23.85 |
| **Phi-4-mini-Reasoning** | 49.75 | 75.94 | 59.49 | 60.76 | 20.68 | 15.88 | 41.14 | 38.39 | 29.78 | 26.04 | 25.34 | 26.18 |
| **Phi-4** | 76.01 | 84.05 | 77.28 | 79.32 | 76.78 | 74.95 | 76.29 | 77.91 | 47.64 | 24.35 | 31.19 | 30.77 |
| **Phi-4-Reasoning** | 79.39 | 85.25 | 81.16 | 82.85 | 79.96 | 78.90 | 77.91 | 82.36 | 57.73 | 11.86 | 32.75 | 22.51 |
| **Qwen2.5-3B** | 75.09 | 79.04 | 70.15 | 64.93 | 59.70 | 62.88 | 70.29 | 71.21 | 12.70 | 14.82 | 11.71 | 14.33 |
| **Qwen2.5-7B** | 79.39 | 83.27 | 78.62 | 77.21 | 68.53 | 72.55 | 78.48 | 76.99 | 25.62 | 9.10 | 12.49 | 10.66 |
| **Qwen2.5-14B** | 79.25 | 82.57 | 79.75 | 78.76 | 74.31 | 76.43 | 81.44 | 81.79 | 33.10 | 12.99 | 27.59 | 13.76 |
| **Qwen2.5-72B** | 79.82 | 85.39 | 83.77 | 83.13 | 77.21 | 78.19 | 82.50 | 82.71 | 48.62 | 34.09 | 33.59 | 32.96 |
| **QwQ-32B** | 77.70 | 84.12 | 76.08 | 78.33 | 75.94 | 78.48 | 76.78 | 77.77 | 34.02 | 29.64 | 29.99 | 32.25 |
| **Qwen3-1.7B** | 68.17 | 75.09 | 66.97 | 61.47 | 54.41 | 48.34 | 65.28 | 65.42 | 26.61 | 24.28 | 28.86 | 30.42 |
| **Qwen3-4B** | 75.94 | 81.79 | 68.88 | 59.21 | 72.12 | 71.56 | 76.15 | 73.89 | 17.71 | 19.27 | 12.07 | 14.61 |
| **Qwen3-4B-thinking** | 77.77 | 84.69 | 77.98 | 76.22 | 78.41 | 75.79 | 76.92 | 77.06 | 23.92 | 11.15 | 11.29 | 12.42 |
| **Qwen3-8B** | 77.13 | 78.19 | 71.42 | 68.81 | 77.56 | 72.12 | 70.64 | 69.23 | 26.46 | 12.21 | 25.34 | 20.61 |
| **Qwen3-8B-thinking** | 78.12 | 82.36 | 80.52 | 79.68 | 78.97 | 75.58 | 80.03 | 79.46 | 33.10 | 7.83 | 21.03 | 15.88 |
| **Qwen3-14B** | 80.24 | 83.27 | 81.51 | 77.28 | 78.90 | 78.12 | 81.44 | 82.64 | 43.61 | 20.04 | 23.36 | 30.35 |
| **Qwen3-14B-thinking** | 80.59 | 84.62 | 80.52 | 79.39 | 78.97 | 78.69 | 82.00 | 80.10 | 53.35 | 8.68 | 18.00 | 29.92 |
| **Baichuan-M2-32B** | 76.22 | 80.73 | 78.12 | 78.48 | 75.30 | 73.18 | 78.69 | 77.77 | 32.89 | 19.20 | 30.98 | 28.86 |
| **Bio-Medical-LLaMA-3-8B** | 39.73 | 55.96 | 44.81 | 38.88 | 38.32 | 33.24 | 39.10 | 40.58 | 33.80 | 34.72 | 32.25 | 33.52 |
| **MediPhi** | 30.70 | 73.68 | 62.17 | 55.68 | 36.84 | 34.93 | 57.52 | 58.29 | 19.20 | 27.24 | 12.70 | 19.12 |
| **MedGemma-4B** | 58.01 | 71.14 | 60.97 | 62.10 | 58.93 | 59.21 | 59.21 | 64.29 | 45.73 | 14.26 | 32.89 | 30.35 |
| **MedGemma-27B** | 77.35 | 82.00 | 74.88 | 74.45 | 75.44 | 77.56 | 75.86 | 76.71 | 67.61 | 26.18 | 35.78 | 55.05 |
| **MedReason-8B** | 41.35 | 11.57 | 8.75 | 10.30 | 42.70 | 46.01 | 23.85 | 3.88 | 7.20 | 11.50 | 24.98 | 25.19 |
| **HuatuoGPT-o1-7B** | 72.34 | 80.52 | 71.28 | 72.48 | 56.74 | 66.97 | 72.69 | 74.10 | 6.35 | 5.79 | 7.13 | 4.73 |
| **HuatuoGPT-o1-8B** | 53.92 | 61.04 | 58.79 | 59.14 | 47.49 | 57.80 | 59.49 | 58.65 | 33.17 | 5.22 | 11.86 | 9.39 |
| **HuatuoGPT-o1-70B** | 67.40 | 78.48 | 66.27 | 66.41 | 67.61 | 65.70 | 66.41 | 68.17 | 61.12 | 31.33 | 35.57 | 38.60 |
| **HuatuoGPT-o1-72B** | 78.69 | 84.83 | 81.16 | 78.05 | 74.81 | 76.85 | 79.11 | 80.17 | 47.07 | 18.00 | 29.29 | 19.97 |
| **OpenBioLLM-8B** | 6.56 | 19.69 | 11.50 | 17.57 | 5.01 | 6.99 | 18.77 | 21.10 | 11.50 | 10.16 | 10.66 | 15.81 |
| **OpenBioLLM-70B** | 29.15 | 71.91 | 58.01 | 29.99 | 28.65 | 32.53 | 51.24 | 62.74 | 22.72 | 12.28 | 14.04 | 18.77 |

**Supplementary Tab. 100:** Zero-Shot performance evaluation of 56 LLMs on MedNLI (Run 3).

| LLMs | Chinese | English | French | German | Japanese | Korean | Portuguese | Spanish | Swahili | Wolof | Yoruba | Zulu |
|---|---|---|---|---|---|---|---|---|---|---|---|---|
| **Proprietary LLMs** | | | | | | | | | | | | |
| **Claude-3.5-Haiku** | 69.58 | 76.99 | 67.11 | 66.69 | 67.54 | 69.58 | 66.69 | 68.81 | 51.16 | 10.87 | 32.82 | 31.76 |
| **Claude-4.0-Sonnet** | 81.72 | 85.89 | 78.69 | 75.58 | 78.33 | 80.73 | 76.15 | 78.19 | 68.45 | 34.79 | 57.59 | 56.67 |
| **Gemini-2.5-Flash** | 72.69 | 82.57 | 71.14 | 69.72 | 67.68 | 69.94 | 69.16 | 70.36 | 66.62 | 46.08 | 59.21 | 62.39 |
| **GPT-4o-mini** | 80.52 | 85.04 | 80.52 | 78.26 | 78.55 | 76.64 | 78.62 | 79.96 | 62.24 | 27.24 | 34.09 | 49.40 |
| **GPT-4o** | 77.42 | 80.52 | 76.57 | 75.30 | 76.22 | 77.35 | 77.91 | 75.02 | 71.84 | 28.72 | 47.71 | 62.31 |
| **GPT-4.1-nano** | 71.07 | 83.98 | 75.23 | 75.09 | 71.35 | 70.22 | 73.61 | 76.29 | 57.09 | 22.79 | 32.60 | 47.07 |
| **GPT-4.1-mini** | 83.20 | 88.29 | 84.90 | 82.57 | 79.89 | 80.59 | 82.43 | 82.36 | 64.36 | 21.52 | 47.71 | 59.99 |
| **GPT-4.1** | 82.85 | 87.86 | 85.60 | 83.91 | 82.71 | 82.36 | 85.53 | 86.73 | 72.12 | 39.24 | 50.46 | 63.23 |
| **GPT-5-nano** | 72.34 | 77.06 | 70.36 | 78.62 | 73.75 | 69.09 | 78.69 | 80.73 | 55.68 | 16.16 | 33.38 | 41.50 |
| **GPT-5-mini** | 83.35 | 88.78 | 87.93 | 86.10 | 83.42 | 82.07 | 86.03 | 87.16 | 72.48 | 31.40 | 47.21 | 62.31 |
| **GPT-5** | 80.80 | 86.52 | 85.89 | 85.46 | 81.72 | 83.56 | 84.69 | 85.25 | 80.95 | 43.19 | 53.92 | 65.63 |
| **o4-mini** | 85.18 | 88.71 | 88.07 | 87.23 | 85.32 | 85.25 | 86.94 | 87.51 | 79.18 | 21.24 | 50.53 | 67.82 |
| **Open-Weight LLMs** | | | | | | | | | | | | |
| **DeepSeek-V3** | 78.69 | 84.19 | 78.19 | 79.82 | 75.30 | 75.23 | 75.79 | 78.76 | 68.95 | 33.03 | 46.44 | 52.65 |
| **DeepSeek-R1** | 77.56 | 83.70 | 75.65 | 76.08 | 76.43 | 72.05 | 74.95 | 76.85 | 59.56 | 34.65 | 46.22 | 53.99 |
| **DeepSeek-R1-Qwen3-8B** | 78.48 | 80.73 | 76.50 | 76.78 | 75.30 | 69.72 | 75.51 | 77.70 | 33.10 | 12.14 | 12.28 | 12.99 |
| **Gemma-3-4B** | 64.50 | 72.76 | 60.90 | 59.14 | 58.08 | 62.88 | 62.95 | 63.09 | 47.14 | 8.54 | 29.92 | 21.67 |
| **Gemma-3-12B** | 62.74 | 82.07 | 66.69 | 67.89 | 66.97 | 69.80 | 65.07 | 66.13 | 66.69 | 7.55 | 35.00 | 48.27 |
| **Gemma-3-27B** | 78.48 | 83.77 | 70.36 | 73.75 | 73.82 | 78.12 | 71.91 | 73.25 | 70.71 | 25.19 | 37.76 | 55.40 |
| **gpt-oss-20B** | 82.15 | 76.78 | 83.49 | 85.39 | 80.17 | 79.60 | 84.54 | 83.27 | 52.36 | 16.02 | 34.65 | 56.67 |
| **gpt-oss-120B** | 82.57 | 87.44 | 84.12 | 86.59 | 82.00 | 82.71 | 85.18 | 86.17 | 72.41 | 25.34 | 47.99 | 65.35 |
| **LLaMA-3.1-8B** | 50.25 | 67.25 | 62.60 | 61.75 | 52.01 | 57.80 | 62.10 | 62.74 | 37.61 | 24.28 | 23.85 | 16.87 |
| **LLaMA-3.1-70B** | 66.55 | 78.62 | 67.68 | 67.61 | 65.49 | 67.11 | 68.38 | 67.54 | 62.39 | 31.40 | 36.34 | 35.00 |
| **LLaMA-3.2-3B** | 29.78 | 64.36 | 50.53 | 40.58 | 36.77 | 30.63 | 50.39 | 53.28 | 32.53 | 20.75 | 23.71 | 24.35 |
| **LLaMA-3.3-70B** | 61.82 | 81.86 | 67.61 | 67.18 | 66.90 | 69.94 | 67.47 | 67.89 | 64.36 | 32.89 | 38.11 | 40.08 |
| **LLaMA-4-Scout** | 76.64 | 82.57 | 71.21 | 72.27 | 68.88 | 70.64 | 72.19 | 73.68 | 68.95 | 34.72 | 42.63 | 56.32 |
| **LLaMA-4-Maverick** | 74.24 | 85.53 | 73.18 | 70.78 | 71.21 | 77.35 | 72.97 | 75.58 | 65.42 | 35.92 | 48.76 | 63.51 |
| **Mistral-7B-v0.3** | 36.56 | 13.62 | 20.96 | 42.48 | 28.44 | 25.12 | 34.30 | 40.44 | 15.38 | 13.13 | 11.50 | 11.01 |
| **Mistral-Small-3.1-24B** | 75.30 | 82.50 | 72.97 | 67.75 | 65.00 | 64.08 | 74.17 | 75.09 | 29.08 | 12.42 | 10.23 | 19.76 |
| **Phi-4-mini** | 63.59 | 78.19 | 68.10 | 67.25 | 61.54 | 45.80 | 67.40 | 67.25 | 31.47 | 17.93 | 20.68 | 24.35 |
| **Phi-4-mini-Reasoning** | 49.12 | 76.64 | 57.80 | 61.54 | 20.47 | 18.21 | 41.28 | 39.10 | 30.42 | 23.29 | 23.92 | 25.76 |
| **Phi-4** | 75.02 | 83.56 | 79.18 | 78.19 | 77.42 | 73.32 | 77.63 | 77.98 | 47.85 | 24.98 | 32.67 | 30.06 |
| **Phi-4-Reasoning** | 80.31 | 85.89 | 81.65 | 82.29 | 78.83 | 78.05 | 77.28 | 81.65 | 58.57 | 11.57 | 31.69 | 23.85 |
| **Qwen2.5-3B** | 74.81 | 79.32 | 70.08 | 66.62 | 60.13 | 61.26 | 70.64 | 73.18 | 12.70 | 14.18 | 11.71 | 14.54 |
| **Qwen2.5-7B** | 78.97 | 82.57 | 78.48 | 76.78 | 67.11 | 70.43 | 78.26 | 76.36 | 26.75 | 8.47 | 13.83 | 9.81 |
| **Qwen2.5-14B** | 80.31 | 81.86 | 81.65 | 78.62 | 75.16 | 76.36 | 80.80 | 81.58 | 33.24 | 12.49 | 27.03 | 14.18 |
| **Qwen2.5-72B** | 80.17 | 85.53 | 84.47 | 83.56 | 77.98 | 79.60 | 83.35 | 82.29 | 49.47 | 34.16 | 33.10 | 32.11 |
| **QwQ-32B** | 78.19 | 83.63 | 77.49 | 77.77 | 75.51 | 78.33 | 77.49 | 77.06 | 33.45 | 30.06 | 32.39 | 31.90 |
| **Qwen3-1.7B** | 70.01 | 74.45 | 66.20 | 61.75 | 55.05 | 49.47 | 65.00 | 63.44 | 29.92 | 24.42 | 32.18 | 33.52 |
| **Qwen3-4B** | 77.13 | 83.13 | 67.89 | 57.94 | 74.74 | 72.41 | 75.94 | 72.90 | 15.95 | 17.93 | 10.09 | 15.46 |
| **Qwen3-4B-thinking** | 79.18 | 84.12 | 78.83 | 75.86 | 78.76 | 77.42 | 76.99 | 77.21 | 23.29 | 11.86 | 11.64 | 11.79 |
| **Qwen3-8B** | 74.38 | 77.28 | 72.62 | 69.87 | 76.22 | 72.69 | 70.57 | 71.77 | 27.03 | 11.79 | 25.83 | 19.83 |
| **Qwen3-8B-thinking** | 78.05 | 82.29 | 80.95 | 79.82 | 78.12 | 75.58 | 79.46 | 80.17 | 29.50 | 7.83 | 17.22 | 13.55 |
| **Qwen3-14B** | 81.16 | 83.13 | 80.80 | 77.98 | 78.83 | 78.76 | 81.58 | 82.22 | 43.75 | 21.52 | 23.01 | 34.23 |
| **Qwen3-14B-thinking** | 80.52 | 85.32 | 80.66 | 78.69 | 80.03 | 77.98 | 82.29 | 80.80 | 52.22 | 8.05 | 17.50 | 28.72 |
| **Baichuan-M2-32B** | 76.71 | 80.95 | 78.83 | 78.26 | 74.31 | 73.18 | 78.05 | 78.12 | 31.40 | 19.97 | 27.88 | 28.44 |
| **Bio-Medical-LLaMA-3-8B** | 40.30 | 57.23 | 43.05 | 40.08 | 38.67 | 33.38 | 39.24 | 40.44 | 31.97 | 33.59 | 33.10 | 32.67 |
| **MediPhi** | 28.44 | 73.61 | 60.20 | 56.39 | 36.77 | 36.34 | 56.74 | 59.07 | 18.14 | 27.52 | 11.57 | 20.89 |
| **MedGemma-4B** | 58.79 | 72.55 | 60.76 | 61.47 | 59.92 | 57.37 | 60.06 | 64.08 | 44.11 | 15.31 | 31.83 | 33.59 |
| **MedGemma-27B** | 77.56 | 82.50 | 74.24 | 74.74 | 75.02 | 77.91 | 76.57 | 77.42 | 68.31 | 25.48 | 38.11 | 53.71 |
| **MedReason-8B** | 40.51 | 11.01 | 8.40 | 8.89 | 40.16 | 45.02 | 23.36 | 4.02 | 9.95 | 10.52 | 23.29 | 24.42 |
| **HuatuoGPT-o1-7B** | 75.09 | 79.89 | 71.28 | 71.98 | 57.16 | 65.70 | 72.62 | 73.39 | 6.14 | 6.14 | 6.92 | 5.08 |
| **HuatuoGPT-o1-8B** | 52.29 | 61.19 | 58.43 | 60.34 | 46.51 | 57.94 | 57.16 | 59.35 | 30.56 | 6.49 | 12.35 | 10.37 |
| **HuatuoGPT-o1-70B** | 67.75 | 79.11 | 67.18 | 66.34 | 67.47 | 66.27 | 67.04 | 66.27 | 63.23 | 30.56 | 32.25 | 40.44 |
| **HuatuoGPT-o1-72B** | 77.21 | 84.12 | 80.59 | 79.32 | 73.75 | 76.71 | 78.90 | 79.82 | 47.35 | 20.04 | 27.81 | 20.40 |
| **OpenBioLLM-8B** | 6.07 | 22.51 | 11.93 | 17.71 | 4.23 | 5.22 | 21.52 | 20.75 | 12.91 | 9.67 | 10.09 | 16.09 |
| **OpenBioLLM-70B** | 27.38 | 72.19 | 58.86 | 29.36 | 31.47 | 30.13 | 51.31 | 61.75 | 22.51 | 10.30 | 14.40 | 18.91 |

**Supplementary Tab. 101:** Zero-Shot performance evaluation of 56 LLMs on MedNLI (Run 4).

| LLMs | Chinese | English | French | German | Japanese | Korean | Portuguese | Spanish | Swahili | Wolof | Yoruba | Zulu |
|---|---|---|---|---|---|---|---|---|---|---|---|---|
| **Proprietary LLMs** | | | | | | | | | | | | |
| **Claude-3.5-Haiku** | 69.37 | 78.76 | 67.54 | 66.76 | 68.38 | 69.87 | 66.13 | 68.81 | 51.94 | 9.60 | 34.37 | 33.31 |
| **Claude-4.0-Sonnet** | 81.93 | 85.32 | 79.53 | 76.22 | 78.97 | 79.89 | 76.43 | 78.26 | 68.10 | 33.66 | 57.94 | 56.46 |
| **Gemini-2.5-Flash** | 71.63 | 84.33 | 71.14 | 68.81 | 67.75 | 70.50 | 69.58 | 70.50 | 67.18 | 47.92 | 58.86 | 61.26 |
| **GPT-4o-mini** | 80.52 | 84.90 | 81.51 | 79.46 | 78.55 | 78.26 | 79.82 | 79.39 | 61.61 | 30.91 | 34.86 | 50.67 |
| **GPT-4o** | 78.55 | 81.93 | 75.09 | 75.65 | 75.86 | 76.36 | 78.55 | 75.44 | 71.21 | 28.65 | 46.93 | 61.04 |
| **GPT-4.1-nano** | 70.64 | 83.70 | 76.15 | 75.86 | 69.23 | 69.65 | 71.42 | 76.50 | 55.89 | 25.62 | 33.45 | 47.99 |
| **GPT-4.1-mini** | 82.29 | 87.44 | 85.74 | 82.29 | 79.68 | 81.51 | 82.71 | 82.78 | 62.74 | 20.61 | 46.65 | 59.70 |
| **GPT-4.1** | 83.42 | 87.86 | 85.74 | 85.89 | 82.36 | 82.50 | 84.90 | 86.73 | 71.49 | 38.88 | 50.88 | 63.73 |
| **GPT-5-nano** | 73.96 | 77.77 | 70.36 | 79.53 | 72.55 | 71.14 | 78.33 | 81.23 | 55.19 | 16.44 | 33.03 | 40.86 |
| **GPT-5-mini** | 83.35 | 88.57 | 87.65 | 85.67 | 82.78 | 82.50 | 87.72 | 87.65 | 72.34 | 33.17 | 47.07 | 63.44 |
| **GPT-5** | 81.16 | 86.24 | 86.17 | 85.96 | 82.36 | 82.43 | 84.54 | 85.53 | 80.59 | 42.34 | 55.05 | 65.77 |
| **o4-mini** | 84.40 | 88.57 | 87.37 | 87.30 | 84.76 | 83.91 | 87.23 | 87.58 | 79.68 | 20.75 | 52.15 | 68.10 |
| **Open-Weight LLMs** | | | | | | | | | | | | |
| **DeepSeek-V3** | 77.77 | 84.69 | 78.33 | 79.11 | 74.95 | 75.16 | 76.50 | 78.19 | 66.62 | 35.36 | 45.52 | 51.52 |
| **DeepSeek-R1** | 77.84 | 83.35 | 74.74 | 76.92 | 75.44 | 73.04 | 74.74 | 77.77 | 59.42 | 34.51 | 45.52 | 55.68 |
| **DeepSeek-R1-Qwen3-8B** | 78.19 | 80.66 | 74.31 | 76.85 | 75.44 | 71.84 | 76.85 | 76.43 | 31.62 | 12.14 | 11.93 | 15.67 |
| **Gemma-3-4B** | 64.43 | 74.31 | 62.31 | 58.93 | 58.29 | 64.78 | 61.61 | 63.66 | 46.08 | 6.77 | 29.29 | 21.59 |
| **Gemma-3-12B** | 63.37 | 82.29 | 66.90 | 67.40 | 67.96 | 68.74 | 64.22 | 65.98 | 64.93 | 6.28 | 32.18 | 49.89 |
| **Gemma-3-27B** | 78.12 | 82.36 | 70.57 | 73.75 | 72.27 | 78.69 | 72.12 | 73.82 | 73.68 | 26.32 | 39.80 | 56.10 |
| **gpt-oss-20B** | 81.37 | 76.57 | 84.33 | 84.47 | 82.36 | 80.10 | 83.56 | 83.06 | 53.99 | 17.50 | 34.16 | 56.95 |
| **gpt-oss-120B** | 83.27 | 87.16 | 84.19 | 86.17 | 82.29 | 82.99 | 85.67 | 85.39 | 71.14 | 25.69 | 49.68 | 65.49 |
| **LLaMA-3.1-8B** | 50.60 | 66.34 | 60.97 | 60.41 | 52.01 | 58.50 | 62.81 | 63.16 | 37.40 | 24.28 | 23.85 | 18.14 |
| **LLaMA-3.1-70B** | 66.83 | 76.57 | 68.60 | 68.10 | 66.27 | 67.54 | 67.61 | 68.38 | 60.83 | 34.23 | 35.50 | 36.77 |
| **LLaMA-3.2-3B** | 31.97 | 62.39 | 50.25 | 40.37 | 36.77 | 31.40 | 47.92 | 53.78 | 33.80 | 20.32 | 21.74 | 24.35 |
| **LLaMA-3.3-70B** | 61.68 | 82.15 | 67.11 | 67.18 | 66.48 | 69.65 | 67.33 | 66.90 | 63.59 | 34.09 | 36.27 | 39.66 |
| **LLaMA-4-Scout** | 76.29 | 80.95 | 71.63 | 72.05 | 68.45 | 69.51 | 72.69 | 73.11 | 68.38 | 34.65 | 41.99 | 57.09 |
| **LLaMA-4-Maverick** | 73.96 | 84.40 | 74.24 | 71.28 | 70.71 | 78.33 | 72.48 | 75.23 | 66.13 | 37.26 | 47.92 | 62.10 |
| **Mistral-7B-v0.3** | 37.19 | 12.91 | 19.34 | 41.21 | 29.50 | 25.12 | 34.37 | 42.13 | 15.81 | 12.70 | 11.29 | 11.08 |
| **Mistral-Small-3.1-24B** | 75.58 | 83.70 | 75.16 | 68.17 | 63.73 | 63.23 | 73.32 | 74.38 | 31.19 | 10.23 | 11.71 | 20.18 |
| **Phi-4-mini** | 63.94 | 77.77 | 69.37 | 65.84 | 61.96 | 44.18 | 65.91 | 69.30 | 31.69 | 18.91 | 21.81 | 24.98 |
| **Phi-4-mini-Reasoning** | 49.82 | 75.58 | 57.52 | 62.24 | 21.81 | 18.42 | 39.31 | 39.87 | 29.99 | 24.49 | 24.35 | 26.68 |
| **Phi-4** | 76.78 | 84.12 | 76.78 | 79.68 | 77.63 | 74.17 | 75.94 | 79.53 | 48.13 | 22.65 | 33.10 | 29.78 |
| **Phi-4-Reasoning** | 81.09 | 85.89 | 81.30 | 81.23 | 79.04 | 77.98 | 76.08 | 82.07 | 57.09 | 11.71 | 33.38 | 24.21 |
| **Qwen2.5-3B** | 74.66 | 79.53 | 69.72 | 65.35 | 59.28 | 61.75 | 68.88 | 70.64 | 13.55 | 15.38 | 12.00 | 13.69 |
| **Qwen2.5-7B** | 78.97 | 82.50 | 80.03 | 76.85 | 67.33 | 71.63 | 78.69 | 77.77 | 26.61 | 10.02 | 14.04 | 9.46 |
| **Qwen2.5-14B** | 80.38 | 81.58 | 80.24 | 78.26 | 75.79 | 76.99 | 81.16 | 81.37 | 35.99 | 13.41 | 28.79 | 15.46 |
| **Qwen2.5-72B** | 80.45 | 85.18 | 83.70 | 83.84 | 77.91 | 78.12 | 81.79 | 83.42 | 50.11 | 34.44 | 33.80 | 32.18 |
| **QwQ-32B** | 77.56 | 83.77 | 76.92 | 77.98 | 75.30 | 78.76 | 77.63 | 77.63 | 33.38 | 26.96 | 31.19 | 32.53 |
| **Qwen3-1.7B** | 68.53 | 75.23 | 66.97 | 62.60 | 54.34 | 49.75 | 65.98 | 63.37 | 25.55 | 24.06 | 29.43 | 32.53 |
| **Qwen3-4B** | 76.92 | 81.65 | 68.81 | 56.88 | 71.91 | 72.41 | 74.52 | 73.89 | 19.27 | 18.77 | 11.15 | 13.90 |
| **Qwen3-4B-thinking** | 79.39 | 84.54 | 78.76 | 76.64 | 79.04 | 76.99 | 76.99 | 76.99 | 24.42 | 12.70 | 12.70 | 10.37 |
| **Qwen3-8B** | 75.16 | 77.70 | 71.63 | 70.01 | 76.78 | 72.34 | 71.42 | 71.21 | 26.96 | 13.76 | 24.91 | 20.04 |
| **Qwen3-8B-thinking** | 77.98 | 81.65 | 79.32 | 78.90 | 77.98 | 76.85 | 80.95 | 80.52 | 30.42 | 9.10 | 20.32 | 15.95 |
| **Qwen3-14B** | 80.52 | 81.93 | 80.95 | 78.48 | 79.68 | 77.49 | 81.23 | 82.36 | 45.59 | 22.58 | 23.78 | 33.03 |
| **Qwen3-14B-thinking** | 79.39 | 84.33 | 81.44 | 80.24 | 80.80 | 78.90 | 80.66 | 81.72 | 53.35 | 9.03 | 17.29 | 28.65 |
| **Baichuan-M2-32B** | 76.78 | 80.52 | 79.04 | 77.06 | 75.37 | 72.83 | 78.19 | 78.48 | 29.78 | 19.76 | 31.69 | 31.55 |
| **Bio-Medical-LLaMA-3-8B** | 40.01 | 58.36 | 44.67 | 39.73 | 38.60 | 32.89 | 38.25 | 41.21 | 32.67 | 33.31 | 34.65 | 33.59 |
| **MediPhi** | 28.30 | 74.03 | 60.97 | 55.82 | 36.70 | 35.64 | 57.09 | 58.93 | 18.21 | 27.31 | 13.06 | 24.06 |
| **MedGemma-4B** | 58.36 | 72.97 | 59.70 | 60.83 | 58.72 | 58.01 | 60.41 | 62.95 | 43.05 | 15.24 | 32.11 | 30.70 |
| **MedGemma-27B** | 77.21 | 80.52 | 73.68 | 75.16 | 74.24 | 76.85 | 76.08 | 77.35 | 67.82 | 26.96 | 36.06 | 54.34 |
| **MedReason-8B** | 41.28 | 10.87 | 9.03 | 9.60 | 39.66 | 46.22 | 23.85 | 3.53 | 8.19 | 11.64 | 23.99 | 23.99 |
| **HuatuoGPT-o1-7B** | 74.31 | 78.41 | 70.71 | 72.12 | 56.67 | 64.86 | 73.04 | 74.38 | 4.94 | 5.72 | 8.26 | 4.87 |
| **HuatuoGPT-o1-8B** | 53.63 | 59.35 | 57.66 | 59.92 | 45.10 | 59.21 | 59.70 | 58.65 | 33.24 | 6.07 | 11.01 | 10.44 |
| **HuatuoGPT-o1-70B** | 66.83 | 78.62 | 66.83 | 66.06 | 67.61 | 66.97 | 66.90 | 67.82 | 61.26 | 32.82 | 34.09 | 40.44 |
| **HuatuoGPT-o1-72B** | 77.63 | 84.90 | 80.73 | 78.19 | 74.17 | 77.06 | 80.10 | 80.24 | 47.57 | 18.56 | 28.79 | 18.77 |
| **OpenBioLLM-8B** | 4.94 | 21.52 | 11.64 | 16.09 | 5.72 | 5.08 | 19.12 | 20.96 | 12.99 | 10.30 | 10.87 | 14.61 |
| **OpenBioLLM-70B** | 28.93 | 72.55 | 56.95 | 30.35 | 28.23 | 30.20 | 53.07 | 61.82 | 22.44 | 11.86 | 14.11 | 19.12 |

**Supplementary Tab. 102:** Zero-Shot performance evaluation of 56 LLMs on MedNLI (Run 5).

| LLMs | Chinese | English | French | German | Japanese | Korean | Portuguese | Spanish | Swahili | Wolof | Yoruba | Zulu | Overall |
|---|---|---|---|---|---|---|---|---|---|---|---|---|---|
| **Proprietary LLMs** | | | | | | | | | | | | | |
| Claude-3.5-Haiku | 76.96±0.05 | 83.14±0.00 | 82.79±0.04 | 81.82±0.02 | 79.36±0.04 | 74.99±0.06 | 83.34±0.05 | 83.04±0.04 | 59.26±0.04 | 19.45±0.46 | 42.53±0.13 | 42.95±0.08 | 67.47 ± 0.04 |
| Claude-4.0-Sonnet | 87.78±0.29 | 89.56±0.14 | 89.97±0.14 | 89.32±0.10 | 88.42±0.21 | 87.83±0.34 | 90.32±0.27 | 90.72±0.30 | 80.47±0.31 | 55.56±0.56 | 70.72±0.43 | 76.12±0.49 | 83.07 ± 0.09 |
| Gemini-2.5-Flash | 88.83±0.16 | 88.11±0.48 | 90.35±0.28 | 89.60±0.21 | 88.84±0.16 | 88.96±0.23 | 90.27±0.20 | 90.61±0.33 | 88.23±0.34 | 78.26±0.54 | 80.43±0.55 | 83.53±0.32 | 87.17 ± 0.14 |
| GPT-4o-mini | 76.93±0.11 | 83.54±0.38 | 82.68±0.47 | 81.97±0.28 | 78.10±0.37 | 72.34±0.29 | 82.37±0.35 | 82.18±0.25 | 69.38±0.39 | 35.69±0.52 | 48.67±0.47 | 58.10±0.44 | 71.00 ± 0.06 |
| GPT-4o | 86.96±0.32 | 87.87±0.16 | 88.95±0.23 | 88.29±0.28 | 87.69±0.20 | 85.07±0.14 | 88.81±0.21 | 89.18±0.09 | 83.97±0.25 | 37.49±0.51 | 61.44±0.36 | 73.68±0.28 | 79.95 ± 0.18 |
| GPT-4.1-nano | 76.59±0.44 | 84.59±0.19 | 81.86±0.54 | 80.96±0.74 | 74.56±0.89 | 72.63±0.81 | 82.20±0.27 | 81.20±0.47 | 59.13±0.54 | 33.51±0.70 | 44.72±0.87 | 53.54±0.34 | 68.79 ± 0.21 |
| GPT-4.1-mini | 86.13±0.31 | 88.59±0.15 | 87.75±0.23 | 87.61±0.27 | 86.53±0.26 | 83.05±0.35 | 88.89±0.25 | 88.41±0.22 | 77.55±0.23 | 27.20±0.85 | 61.77±0.36 | 72.59±0.25 | 78.01 ± 0.09 |
| GPT-4.1 | 88.57±0.39 | 90.05±0.17 | 90.19±0.18 | 89.59±0.37 | 89.51±0.33 | 86.14±0.47 | 90.52±0.10 | 89.28±0.11 | 86.67±0.42 | 58.72±0.57 | 67.49±0.28 | 78.58±0.36 | 83.78 ± 0.04 |
| GPT-5-nano | 67.63±0.69 | 80.66±0.40 | 78.25±0.35 | 77.47±0.61 | 74.14±0.57 | 66.74±0.53 | 75.31±0.28 | 77.05±0.66 | 57.85±1.09 | 17.02±0.47 | 41.53±0.46 | 50.74±0.56 | 63.70 ± 0.08 |
| GPT-5-mini | 85.53±0.14 | 88.36±0.43 | 87.90±0.24 | 87.37±0.40 | 86.53±0.41 | 82.71±0.18 | 88.22±0.37 | 88.34±0.22 | 77.87±0.41 | 33.48±0.40 | 59.40±0.42 | 73.02±0.57 | 78.23 ± 0.12 |
| GPT-5 | 88.30±0.11 | 90.44±0.29 | 89.99±0.23 | 89.61±0.19 | 89.11±0.15 | 86.01±0.37 | 90.41±0.32 | 90.86±0.16 | 86.71±0.20 | 57.47±0.09 | 71.58±0.31 | 79.55±0.45 | 84.17 ± 0.05 |
| o4-mini | 88.88±0.17 | 90.28±0.16 | 90.22±0.09 | 89.97±0.32 | 89.48±0.13 | 87.08±0.42 | 90.77±0.17 | 90.50±0.21 | 86.74±0.38 | 47.23±0.87 | 76.07±0.20 | 82.47±0.27 | 84.14 ± 0.13 |
| **Open-Weight LLMs** | | | | | | | | | | | | | |
| DeepSeek-V3 | 87.64±0.13 | 89.67±0.13 | 88.59±0.32 | 88.74±0.21 | 87.44±0.44 | 84.10±0.47 | 88.48±0.22 | 89.02±0.09 | 77.07±0.61 | 51.05±0.44 | 58.45±0.88 | 64.69±0.36 | 79.58 ± 0.11 |
| DeepSeek-R1 | 88.28±0.03 | 89.64±0.17 | 89.63±0.14 | 89.25±0.18 | 87.85±0.41 | 86.69±0.17 | 90.10±0.21 | 89.85±0.33 | 84.13±0.60 | 65.64±0.41 | 69.79±0.41 | 77.36±0.48 | 84.02 ± 0.09 |
| DeepSeek-R1-Qwen3-8B | 79.34±0.58 | 82.46±0.35 | 77.59±0.47 | 77.49±0.62 | 75.66±0.44 | 73.91±0.41 | 78.48±0.68 | 78.63±0.15 | 36.65±1.00 | 11.08±0.49 | 10.83±0.63 | 13.17±0.67 | 57.94 ± 0.24 |
| Gemma-3-4B | 52.83±0.43 | 64.58±0.47 | 59.58±0.36 | 58.23±0.55 | 53.69±0.35 | 48.51±0.43 | 59.74±0.76 | 59.42±0.50 | 42.89±0.78 | 12.90±0.53 | 25.55±0.69 | 20.01±0.45 | 46.49 ± 0.08 |
| Gemma-3-12B | 70.80±0.43 | 78.31±0.29 | 75.85±0.52 | 75.65±0.44 | 71.93±0.68 | 69.44±0.49 | 75.22±0.58 | 75.47±0.28 | 62.87±0.57 | 11.15±0.46 | 43.19±0.87 | 54.35±0.76 | 63.69 ± 0.11 |
| Gemma-3-27B | 76.73±0.57 | 82.14±0.38 | 80.95±0.52 | 80.88±0.43 | 77.86±0.51 | 76.83±0.65 | 80.71±0.40 | 81.34±0.42 | 71.33±0.63 | 23.93±0.40 | 52.41±0.61 | 62.27±0.19 | 70.62 ± 0.08 |
| gpt-oss-20B | 82.64±0.46 | 78.47±0.75 | 84.91±0.19 | 84.63±0.41 | 82.95±0.61 | 81.27±0.27 | 85.03±0.34 | 84.74±0.23 | 68.96±0.52 | 33.72±0.68 | 61.98±0.66 | 69.13±0.62 | 74.87 ± 0.03 |
| gpt-oss-120B | 86.91±0.25 | 88.00±0.12 | 88.48±0.21 | 88.13±0.20 | 86.81±0.13 | 85.46±0.37 | 88.13±0.28 | 88.26±0.33 | 77.86±0.41 | 52.62±0.84 | 68.26±0.44 | 76.32±0.65 | 81.27 ± 0.10 |
| LLaMA-3.1-8B | 56.79±0.93 | 74.27±0.94 | 63.01±0.65 | 62.34±0.50 | 50.87±0.46 | 42.49±0.63 | 62.70±0.85 | 62.16±1.07 | 37.13±0.65 | 23.72±0.59 | 19.75±0.84 | 14.36±0.68 | 47.47 ± 0.19 |
| LLaMA-3.1-70B | 78.36±0.44 | 85.67±0.35 | 84.26±0.62 | 83.43±0.63 | 77.57±1.15 | 67.09±0.72 | 83.76±0.36 | 84.28±0.39 | 70.40±0.56 | 41.28±0.68 | 45.09±0.50 | 49.39±0.51 | 70.88 ± 0.29 |
| LLaMA-3.2-3B | 49.20±0.79 | 64.07±0.52 | 53.22±1.29 | 49.62±0.68 | 41.37±0.25 | 31.63±0.97 | 50.68±0.68 | 43.97±1.11 | 33.28±0.60 | 14.19±0.69 | 18.89±0.78 | 17.28±0.52 | 38.95 ± 0.21 |
| LLaMA-3.3-70B | 61.85±1.13 | 86.28±0.33 | 85.89±0.28 | 84.91±0.27 | 65.94±0.74 | 76.32±0.35 | 85.67±0.42 | 85.93±0.24 | 72.72±0.36 | 41.81±0.51 | 47.93±0.52 | 51.22±0.34 | 70.54 ± 0.08 |
| LLaMA-4-Scout | 84.12±0.26 | 87.10±0.27 | 86.31±0.23 | 86.83±0.16 | 84.23±0.22 | 82.29±0.27 | 86.43±0.35 | 87.00±0.32 | 77.98±0.31 | 50.25±0.78 | 50.64±0.15 | 69.34±0.27 | 77.71 ± 0.08 |
| LLaMA-4-Maverick | 87.73±0.24 | 89.55±0.15 | 90.02±0.21 | 89.57±0.14 | 87.02±0.32 | 86.00±0.35 | 89.98±0.17 | 90.05±0.16 | 84.15±0.30 | 54.47±0.27 | 69.27±0.58 | 77.79±0.32 | 82.97 ± 0.13 |
| Mistral-7B-v0.3 | 28.39±0.73 | 36.61±0.44 | 21.04±0.80 | 40.89±1.09 | 24.09±0.59 | 27.67±0.66 | 30.07±0.56 | 37.35±0.91 | 18.13±0.24 | 14.69±0.66 | 10.40±0.50 | 9.78±0.79 | 24.93 ± 0.10 |
| Mistral-Small-3.1-24B | 76.22±0.81 | 83.46±0.24 | 80.23±0.47 | 80.87±0.55 | 74.67±0.75 | 68.63±1.04 | 81.02±0.55 | 81.13±0.68 | 38.85±0.36 | 11.82±1.19 | 13.73±0.47 | 20.57±0.44 | 59.27 ± 0.37 |
| Phi-4-mini | 42.01±0.90 | 71.17±0.59 | 52.71±0.88 | 55.27±1.11 | 39.40±0.81 | 31.97±0.78 | 50.59±1.09 | 50.39±0.79 | 30.29±0.29 | 14.39±0.76 | 19.51±0.48 | 17.02±0.91 | 39.56 ± 0.13 |
| Phi-4-mini-Reasoning | 36.42±0.55 | 74.05±0.48 | 52.82±1.02 | 60.22±0.66 | 26.83±1.00 | 14.56±0.55 | 49.83±0.52 | 44.06±0.47 | 28.71±0.83 | 20.84±0.93 | 18.50±0.55 | 19.80±0.89 | 37.22 ± 0.18 |
| Phi-4 | 70.17±0.66 | 84.60±0.55 | 81.31±0.15 | 81.12±0.29 | 72.99±0.38 | 66.18±0.82 | 80.93±0.59 | 80.43±0.24 | 52.61±0.45 | 23.43±0.53 | 38.22±0.80 | 34.67±0.70 | 63.89 ± 0.17 |
| Phi-4-Reasoning | 79.89±0.71 | 85.50±0.39 | 84.47±0.13 | 84.67±0.22 | 83.64±0.34 | 79.93±0.58 | 78.34±0.36 | 85.09±0.64 | 71.75±0.76 | 17.47±0.55 | 48.91±0.69 | 37.77±0.61 | 69.79 ± 0.16 |
| Qwen2.5-3B | 61.05±0.53 | 66.95±0.35 | 58.48±0.58 | 55.81±0.68 | 50.25±0.47 | 44.89±0.48 | 56.09±0.72 | 55.37±0.27 | 16.09±0.67 | 9.17±0.24 | 13.37±0.55 | 11.75±0.81 | 41.61 ± 0.06 |
| Qwen2.5-7B | 71.29±0.31 | 74.45±0.57 | 68.09±0.20 | 65.88±0.83 | 63.10±0.50 | 57.78±0.58 | 66.76±0.12 | 66.51±0.55 | 32.25±0.59 | 9.13±0.44 | 24.63±0.58 | 18.55±0.66 | 51.54 ± 0.13 |
| Qwen2.5-14B | 77.17±0.72 | 80.30±0.33 | 76.86±0.37 | 75.79±0.56 | 72.86±0.31 | 68.43±0.35 | 75.99±0.49 | 75.73±0.54 | 37.04±0.25 | 26.76±0.76 | 35.73±0.46 | 35.38±0.67 | 61.50 ± 0.11 |
| Qwen2.5-72B | 83.21±0.33 | 85.62±0.33 | 84.48±0.19 | 84.17±0.09 | 81.56±0.26 | 78.96±0.38 | 85.10±0.33 | 84.97±0.37 | 54.45±0.67 | 43.90±0.23 | 41.74±0.43 | 43.98±0.32 | 71.01 ± 0.07 |
| QwQ-32B | 85.03±0.33 | 86.24±0.36 | 86.57±0.45 | 85.38±0.20 | 82.29±0.26 | 79.60±0.73 | 86.68±0.30 | 86.00±0.12 | 62.21±0.46 | 23.75±0.67 | 29.49±1.19 | 29.87±0.77 | 68.59 ± 0.18 |
| Qwen3-1.7B | 57.75±0.84 | 65.09±0.52 | 55.49±0.62 | 52.98±0.72 | 45.46±0.65 | 42.18±0.44 | 52.99±0.51 | 52.82±0.50 | 23.75±0.59 | 22.52±0.38 | 26.30±0.85 | 23.17±0.88 | 43.38 ± 0.26 |
| Qwen3-4B | 72.91±0.56 | 76.47±0.37 | 70.40±0.15 | 70.43±0.35 | 66.57±0.97 | 61.10±1.07 | 70.00±0.54 | 69.67±0.39 | 19.33±0.91 | 14.16±0.41 | 16.25±0.53 | 9.66±0.83 | 51.41 ± 0.17 |
| Qwen3-4B-thinking | 76.64±0.28 | 80.40±0.43 | 78.85±0.46 | 78.31±0.65 | 75.53±0.34 | 72.08±0.50 | 77.34±0.70 | 78.18±0.48 | 29.54±0.79 | 17.14±1.02 | 13.21±0.65 | 8.99±0.98 | 57.18 ± 0.19 |
| Qwen3-8B | 76.95±0.56 | 80.75±0.42 | 75.54±0.11 | 76.37±0.49 | 72.29±0.25 | 67.65±0.70 | 75.36±0.67 | 75.22±0.37 | 31.28±0.81 | 10.53±0.50 | 22.73±0.61 | 11.42±0.82 | 56.34 ± 0.18 |
| Qwen3-8B-thinking | 81.15±0.36 | 83.88±0.22 | 82.17±0.40 | 82.52±0.48 | 79.96±0.42 | 78.42±0.65 | 81.90±0.16 | 82.18±0.30 | 36.87±0.69 | 14.73±0.69 | 17.74±0.43 | 12.06±0.74 | 61.13 ± 0.11 |
| Qwen3-14B | 80.88±0.51 | 83.93±0.27 | 80.96±0.51 | 80.53±0.32 | 76.98±0.53 | 73.88±0.71 | 80.59±0.43 | 80.92±0.71 | 46.86±0.89 | 15.01±0.25 | 24.17±1.02 | 27.17±0.95 | 62.66 ± 0.21 |
| Qwen3-14B-thinking | 83.59±0.31 | 86.03±0.34 | 84.71±0.29 | 84.81±0.33 | 83.42±0.30 | 80.83±0.49 | 84.66±0.40 | 85.02±0.37 | 64.36±0.73 | 17.63±0.63 | 24.89±0.79 | 27.53±0.62 | 67.29 ± 0.12 |
| Baichuan-M2-32B | 84.76±0.28 | 86.10±0.28 | 83.98±0.30 | 82.91±0.52 | 81.52±0.44 | 78.42±0.66 | 83.88±0.67 | 83.72±0.21 | 34.33±0.87 | 25.98±0.47 | 33.35±0.87 | 29.88±0.61 | 65.74 ± 0.25 |
| Bio-Medical-LLaMA-3-8B | 50.07±0.36 | 69.06±0.45 | 59.37±0.26 | 58.85±0.34 | 49.40±0.41 | 44.72±0.42 | 57.13±0.13 | 54.77±0.47 | 38.93±0.49 | 35.79±0.46 | 34.01±0.47 | 33.82±0.55 | 48.83 ± 0.08 |
| MediPhi | 38.13±0.66 | 68.04±0.32 | 62.01±0.55 | 61.21±0.33 | 31.25±1.05 | 32.76±0.35 | 57.91±0.32 | 53.80±0.52 | 16.53±0.63 | 14.77±0.78 | 15.83±0.90 | 17.52±0.70 | 39.15 ± 0.20 |
| MedGemma-4B | 55.00±0.66 | 69.39±0.50 | 63.51±0.68 | 61.79±0.91 | 55.02±0.42 | 50.91±1.44 | 62.94±0.47 | 62.57±0.61 | 45.48±0.32 | 20.66±0.57 | 21.98±0.68 | 30.95±0.75 | 50.02 ± 0.37 |
| MedGemma-27B | 82.20±0.33 | 86.11±0.38 | 84.85±0.47 | 84.64±0.37 | 81.91±0.20 | 80.72±0.73 | 85.48±0.28 | 85.96±0.23 | 77.75±0.32 | 18.04±0.24 | 50.94±0.34 | 65.88±0.77 | 73.71 ± 0.12 |
| MedReason-8B | 47.48±0.29 | 63.47±0.87 | 18.40±0.90 | 14.44±0.35 | 48.26±0.44 | 41.14±0.79 | 26.60±0.31 | 34.88±0.93 | 12.92±0.15 | 8.11±0.28 | 13.19±0.53 | 11.66±0.29 | 28.38 ± 0.12 |
| HuatuoGPT-o1-7B | 73.80±0.39 | 69.96±0.56 | 72.11±0.52 | 71.80±0.67 | 65.12±0.63 | 62.83±0.55 | 72.32±0.26 | 70.88±0.17 | 7.32±0.42 | 7.04±0.30 | 7.19±0.40 | 6.26±0.42 | 48.89 ± 0.16 |
| HuatuoGPT-o1-8B | 64.48±0.74 | 73.33±0.34 | 70.71±0.36 | 70.50±0.82 | 62.24±0.27 | 54.89±0.25 | 69.30±0.54 | 69.15±0.47 | 50.34±0.27 | 9.62±0.49 | 9.04±0.56 | 8.46±0.51 | 51.01 ± 0.14 |
| HuatuoGPT-o1-70B | 72.87±0.53 | 85.19±0.23 | 84.22±0.23 | 84.27±0.36 | 72.69±0.32 | 77.43±0.46 | 84.77±0.46 | 84.83±0.26 | 74.90±0.38 | 43.44±0.95 | 50.15±0.59 | 56.66±0.82 | 72.62 ± 0.22 |
| HuatuoGPT-o1-72B | 84.34±0.08 | 85.54±0.18 | 86.20±0.22 | 84.46±0.48 | 83.03±0.39 | 78.55±0.61 | 86.84±0.21 | 86.85±0.33 | 62.18±0.44 | 44.44±0.61 | 44.91±1.39 | 29.51±0.71 | 71.41 ± 0.19 |
| OpenBioLLM-8B | 22.73±0.63 | 35.28±0.59 | 24.55±0.71 | 32.01±0.36 | 8.99±0.34 | 12.06±0.54 | 34.79±0.97 | 35.48±1.40 | 11.49±0.82 | 6.95±0.21 | 10.10±0.76 | 11.10±0.75 | 20.46 ± 0.19 |
| OpenBioLLM-70B | 31.62±0.98 | 73.19±0.85 | 74.86±0.39 | 52.14±0.98 | 32.53±0.58 | 31.09±0.66 | 58.81±0.64 | 70.78±0.41 | 31.04±0.91 | 10.13±0.36 | 11.79±0.56 | 23.73±0.75 | 41.81 ± 0.26 |

**Supplementary Tab. 103:** Performance evaluation of 56 LLMs on HeadQA.

| LLMs | Chinese | English | French | German | Japanese | Korean | Portuguese | Spanish | Swahili | Wolof | Yoruba | Zulu |
|---|---|---|---|---|---|---|---|---|---|---|---|---|
| **Proprietary LLMs** | | | | | | | | | | | | |
| **Claude-3.5-Haiku** | 76.92 | 83.14 | 82.78 | 81.83 | 79.40 | 75.02 | 83.32 | 83.00 | 59.29 | 19.43 | 42.56 | 42.88 |
| **Claude-4.0-Sonnet** | 88.28 | 89.50 | 90.08 | 89.18 | 88.77 | 87.56 | 90.17 | 90.22 | 79.94 | 55.09 | 70.74 | 75.92 |
| **Gemini-2.5-Flash** | 89.04 | 88.59 | 90.22 | 89.72 | 88.73 | 89.18 | 90.49 | 90.98 | 87.96 | 78.54 | 79.71 | 83.81 |
| **GPT-4o-mini** | 76.87 | 83.72 | 82.37 | 82.19 | 78.18 | 71.96 | 82.64 | 82.19 | 69.39 | 35.66 | 48.87 | 57.75 |
| **GPT-4o** | 87.06 | 88.10 | 89.09 | 88.68 | 87.92 | 85.12 | 88.82 | 89.22 | 84.22 | 38.10 | 61.63 | 73.62 |
| **GPT-4.1-nano** | 77.05 | 84.72 | 81.51 | 81.51 | 75.07 | 72.50 | 81.79 | 81.51 | 59.83 | 32.64 | 45.31 | 53.20 |
| **GPT-4.1-mini** | 86.52 | 88.68 | 88.05 | 87.92 | 86.43 | 83.41 | 88.73 | 88.32 | 77.37 | 28.00 | 61.59 | 72.86 |
| **GPT-4.1** | 89.09 | 90.22 | 90.26 | 89.50 | 89.22 | 86.20 | 90.53 | 89.18 | 86.61 | 59.38 | 67.13 | 78.22 |
| **GPT-5-nano** | 67.49 | 80.84 | 77.64 | 76.47 | 73.72 | 67.40 | 75.56 | 78.09 | 56.67 | 17.27 | 40.98 | 50.63 |
| **GPT-5-mini** | 85.48 | 88.77 | 87.69 | 87.02 | 86.20 | 82.64 | 88.10 | 88.59 | 77.68 | 33.72 | 59.33 | 73.94 |
| **GPT-5** | 88.23 | 90.85 | 90.31 | 89.36 | 88.95 | 86.20 | 90.62 | 90.89 | 86.61 | 57.53 | 71.37 | 79.89 |
| **o4-mini** | 88.91 | 90.17 | 90.13 | 90.35 | 89.27 | 87.65 | 91.03 | 90.44 | 87.33 | 47.88 | 76.28 | 82.60 |
| **Open-Weight LLMs** | | | | | | | | | | | | |
| **DeepSeek-V3** | 87.60 | 89.63 | 88.37 | 88.59 | 86.97 | 83.32 | 88.32 | 89.09 | 77.10 | 51.35 | 58.57 | 64.70 |
| **DeepSeek-R1** | 88.28 | 89.72 | 89.59 | 88.95 | 87.51 | 86.93 | 89.77 | 89.90 | 84.17 | 65.42 | 70.51 | 77.23 |
| **DeepSeek-R1-Qwen3-8B** | 79.94 | 82.82 | 78.36 | 76.56 | 76.19 | 74.21 | 78.13 | 78.49 | 38.10 | 11.27 | 10.69 | 12.76 |
| **Gemma-3-4B** | 52.52 | 64.20 | 59.74 | 57.75 | 54.28 | 48.11 | 60.14 | 59.51 | 41.84 | 12.98 | 25.88 | 20.20 |
| **Gemma-3-12B** | 70.78 | 78.09 | 76.24 | 75.97 | 71.24 | 69.75 | 75.07 | 75.70 | 62.08 | 11.18 | 44.14 | 55.14 |
| **Gemma-3-27B** | 77.64 | 82.28 | 80.52 | 80.30 | 78.18 | 75.70 | 80.93 | 81.24 | 72.09 | 24.44 | 52.34 | 62.13 |
| **gpt-oss-20B** | 82.28 | 78.40 | 84.99 | 84.36 | 83.72 | 81.33 | 85.17 | 84.45 | 69.21 | 32.91 | 62.76 | 68.94 |
| **gpt-oss-120B** | 86.52 | 88.01 | 88.41 | 88.41 | 87.02 | 85.17 | 88.19 | 88.37 | 78.00 | 53.47 | 68.35 | 75.38 |
| **LLaMA-3.1-8B** | 57.26 | 73.22 | 61.95 | 62.13 | 50.50 | 42.88 | 62.62 | 62.71 | 37.06 | 24.17 | 18.98 | 13.80 |
| **LLaMA-3.1-70B** | 79.08 | 86.11 | 83.32 | 83.14 | 76.51 | 67.94 | 84.27 | 83.68 | 69.66 | 40.62 | 45.45 | 49.05 |
| **LLaMA-3.2-3B** | 47.88 | 64.88 | 51.35 | 49.95 | 41.16 | 30.88 | 51.44 | 42.70 | 32.46 | 14.56 | 19.07 | 17.49 |
| **LLaMA-3.3-70B** | 60.64 | 86.74 | 86.16 | 84.81 | 65.78 | 76.74 | 85.66 | 85.89 | 73.08 | 41.34 | 48.60 | 51.08 |
| **LLaMA-4-Scout** | 83.72 | 87.51 | 86.20 | 86.88 | 83.90 | 82.37 | 86.25 | 86.93 | 78.40 | 49.19 | 50.41 | 69.70 |
| **LLaMA-4-Maverick** | 87.87 | 89.63 | 90.31 | 89.54 | 87.29 | 85.53 | 89.77 | 90.04 | 84.27 | 54.73 | 69.88 | 77.86 |
| **Mistral-7B-v0.3** | 28.67 | 36.47 | 22.14 | 40.13 | 23.35 | 27.23 | 30.61 | 37.24 | 18.21 | 14.83 | 10.10 | 10.19 |
| **Mistral-Small-3.1-24B** | 75.02 | 83.05 | 79.44 | 81.47 | 73.72 | 68.03 | 81.24 | 81.15 | 38.77 | 10.82 | 13.84 | 20.15 |
| **Phi-4-mini** | 42.70 | 70.20 | 52.12 | 53.38 | 39.00 | 32.69 | 51.71 | 50.05 | 30.61 | 13.80 | 20.15 | 16.32 |
| **Phi-4-mini-Reasoning** | 35.80 | 74.44 | 54.01 | 59.24 | 25.70 | 14.20 | 49.95 | 43.78 | 29.17 | 19.75 | 18.03 | 20.87 |
| **Phi-4** | 70.74 | 84.22 | 81.38 | 81.02 | 72.50 | 66.23 | 81.65 | 80.57 | 52.52 | 24.21 | 38.82 | 33.86 |
| **Phi-4-Reasoning** | 78.81 | 86.02 | 84.45 | 84.76 | 83.41 | 80.84 | 78.04 | 84.04 | 72.09 | 16.86 | 49.73 | 38.05 |
| **Qwen2.5-3B** | 61.00 | 66.95 | 57.98 | 54.87 | 50.63 | 45.18 | 55.86 | 55.28 | 17.09 | 8.75 | 13.39 | 11.68 |
| **Qwen2.5-7B** | 71.33 | 74.39 | 67.94 | 66.73 | 63.44 | 58.03 | 66.73 | 66.46 | 32.69 | 9.42 | 24.35 | 18.94 |
| **Qwen2.5-14B** | 78.36 | 79.98 | 76.60 | 75.88 | 73.08 | 67.85 | 76.65 | 75.11 | 36.79 | 26.56 | 35.17 | 34.90 |
| **Qwen2.5-72B** | 82.69 | 85.75 | 84.17 | 84.13 | 81.79 | 78.58 | 85.21 | 85.26 | 54.96 | 44.18 | 41.79 | 44.41 |
| **QwQ-32B** | 85.44 | 86.02 | 86.70 | 85.30 | 82.24 | 80.57 | 86.88 | 86.02 | 61.86 | 24.03 | 30.66 | 30.03 |
| **Qwen3-1.7B** | 57.39 | 65.15 | 54.73 | 52.21 | 44.72 | 42.06 | 53.11 | 52.98 | 23.62 | 22.45 | 26.38 | 22.09 |
| **Qwen3-4B** | 72.72 | 76.56 | 70.24 | 70.65 | 67.45 | 62.53 | 69.48 | 69.16 | 18.94 | 14.52 | 16.14 | 10.37 |
| **Qwen3-4B-thinking** | 77.05 | 80.66 | 78.67 | 79.35 | 75.56 | 71.69 | 77.01 | 78.18 | 28.36 | 18.03 | 12.53 | 8.93 |
| **Qwen3-8B** | 76.92 | 80.39 | 75.52 | 77.19 | 72.27 | 67.40 | 75.11 | 74.80 | 32.42 | 11.00 | 22.99 | 11.18 |
| **Qwen3-8B-thinking** | 80.88 | 83.99 | 82.24 | 81.92 | 79.94 | 79.08 | 81.74 | 82.06 | 36.02 | 14.61 | 18.12 | 11.27 |
| **Qwen3-14B** | 81.47 | 83.59 | 80.61 | 80.79 | 76.47 | 73.94 | 80.25 | 79.71 | 45.31 | 14.70 | 22.99 | 28.36 |
| **Qwen3-14B-thinking** | 83.90 | 86.61 | 84.58 | 84.67 | 83.09 | 80.79 | 84.45 | 85.48 | 64.83 | 18.03 | 24.89 | 27.37 |
| **Baichuan-M2-32B** | 84.63 | 86.20 | 83.63 | 83.36 | 81.38 | 77.73 | 84.08 | 83.95 | 34.85 | 26.65 | 32.28 | 30.88 |
| **Bio-Medical-LLaMA-3-8B** | 50.50 | 69.03 | 59.69 | 59.33 | 49.37 | 44.50 | 56.94 | 55.41 | 38.14 | 35.57 | 33.54 | 34.40 |
| **MediPhi** | 38.28 | 68.17 | 61.14 | 61.18 | 32.78 | 32.82 | 57.75 | 54.19 | 16.23 | 14.43 | 14.65 | 17.45 |
| **MedGemma-4B** | 55.23 | 69.57 | 63.71 | 61.00 | 54.69 | 50.36 | 63.17 | 62.67 | 45.18 | 20.20 | 21.46 | 29.94 |
| **MedGemma-27B** | 82.64 | 85.66 | 84.85 | 84.17 | 81.79 | 80.84 | 85.08 | 86.02 | 78.00 | 18.21 | 50.50 | 66.41 |
| **MedReason-8B** | 47.29 | 62.31 | 18.98 | 14.02 | 48.78 | 41.16 | 26.51 | 35.48 | 13.03 | 7.62 | 12.44 | 11.54 |
| **HuatuoGPT-o1-7B** | 73.17 | 70.33 | 72.59 | 71.78 | 65.64 | 62.17 | 72.54 | 70.83 | 7.71 | 7.08 | 7.53 | 6.40 |
| **HuatuoGPT-o1-8B** | 63.48 | 73.72 | 71.06 | 69.52 | 62.08 | 55.05 | 69.30 | 69.12 | 50.63 | 9.65 | 9.11 | 8.34 |
| **HuatuoGPT-o1-70B** | 72.36 | 85.44 | 84.13 | 84.45 | 72.77 | 77.46 | 85.48 | 84.58 | 74.93 | 42.92 | 50.23 | 56.94 |
| **HuatuoGPT-o1-72B** | 84.27 | 85.48 | 85.93 | 84.76 | 82.60 | 78.76 | 87.11 | 86.52 | 61.54 | 44.77 | 45.22 | 30.52 |
| **OpenBioLLM-8B** | 22.36 | 35.84 | 25.43 | 31.88 | 8.84 | 11.36 | 34.36 | 35.12 | 12.62 | 6.58 | 8.79 | 10.78 |
| **OpenBioLLM-70B** | 30.12 | 74.21 | 74.57 | 52.48 | 32.37 | 32.15 | 57.75 | 70.51 | 31.92 | 10.14 | 11.68 | 23.08 |

**Supplementary Tab. 104:** Zero-Shot performance evaluation of 56 LLMs on HeadQA (Run 1).

| LLMs | Chinese | English | French | German | Japanese | Korean | Portuguese | Spanish | Swahili | Wolof | Yoruba | Zulu |
|---|---|---|---|---|---|---|---|---|---|---|---|---|
| **Proprietary LLMs** | | | | | | | | | | | | |
| **Claude-3.5-Haiku** | 76.96 | 83.14 | 82.73 | 81.83 | 79.35 | 74.98 | 83.36 | 83.09 | 59.29 | 19.57 | 42.56 | 42.88 |
| **Claude-4.0-Sonnet** | 87.51 | 89.63 | 89.95 | 89.40 | 88.28 | 87.96 | 90.22 | 90.89 | 80.75 | 55.86 | 70.96 | 75.43 |
| **Gemini-2.5-Flash** | 88.91 | 88.55 | 90.85 | 89.81 | 88.86 | 89.18 | 90.31 | 90.94 | 88.05 | 78.04 | 81.24 | 83.59 |
| **GPT-4o-mini** | 76.83 | 83.36 | 83.32 | 82.06 | 77.73 | 72.59 | 82.28 | 81.97 | 69.97 | 35.84 | 48.20 | 57.57 |
| **GPT-4o** | 86.47 | 87.65 | 89.00 | 88.01 | 87.42 | 84.90 | 88.55 | 89.04 | 83.81 | 36.79 | 61.14 | 73.94 |
| **GPT-4.1-nano** | 75.88 | 84.58 | 82.19 | 81.51 | 75.74 | 73.76 | 82.55 | 80.39 | 59.42 | 34.58 | 45.09 | 53.56 |
| **GPT-4.1-mini** | 86.34 | 88.46 | 87.60 | 87.60 | 86.52 | 83.05 | 88.77 | 88.19 | 77.32 | 26.96 | 61.81 | 72.86 |
| **GPT-4.1** | 88.86 | 90.22 | 89.95 | 89.95 | 89.31 | 85.44 | 90.53 | 89.22 | 87.33 | 58.16 | 67.67 | 79.13 |
| **GPT-5-nano** | 68.30 | 80.16 | 78.36 | 77.64 | 74.21 | 67.13 | 75.34 | 77.23 | 57.26 | 16.77 | 41.75 | 50.95 |
| **GPT-5-mini** | 85.30 | 88.05 | 87.74 | 87.02 | 86.29 | 82.73 | 87.87 | 88.32 | 77.46 | 33.81 | 59.38 | 72.68 |
| **GPT-5** | 88.19 | 90.53 | 89.86 | 89.81 | 89.22 | 85.35 | 90.71 | 90.80 | 86.56 | 57.57 | 71.78 | 78.81 |
| **o4-mini** | 89.00 | 90.13 | 90.17 | 89.59 | 89.45 | 87.02 | 90.80 | 90.49 | 86.38 | 45.76 | 76.15 | 82.73 |
| **Open-Weight LLMs** | | | | | | | | | | | | |
| **DeepSeek-V3** | 87.74 | 89.54 | 88.23 | 88.95 | 87.69 | 84.27 | 88.32 | 89.04 | 76.06 | 51.17 | 58.57 | 64.16 |
| **DeepSeek-R1** | 88.28 | 89.68 | 89.54 | 89.36 | 88.32 | 86.79 | 90.04 | 89.72 | 84.90 | 65.60 | 69.75 | 77.55 |
| **DeepSeek-R1-Qwen3-8B** | 78.72 | 82.82 | 77.55 | 77.91 | 75.83 | 73.40 | 78.40 | 78.72 | 36.29 | 11.27 | 11.05 | 13.44 |
| **Gemma-3-4B** | 53.43 | 65.19 | 59.20 | 58.30 | 53.56 | 48.65 | 60.05 | 59.24 | 43.73 | 13.57 | 25.34 | 19.30 |
| **Gemma-3-12B** | 71.37 | 78.27 | 76.33 | 75.34 | 73.04 | 68.58 | 75.25 | 75.34 | 62.62 | 10.82 | 41.97 | 53.20 |
| **Gemma-3-27B** | 76.87 | 82.24 | 81.51 | 80.97 | 77.41 | 77.19 | 80.52 | 80.79 | 70.38 | 24.12 | 53.34 | 62.22 |
| **gpt-oss-20B** | 82.64 | 77.32 | 84.67 | 85.12 | 82.87 | 80.84 | 85.53 | 84.85 | 69.66 | 34.67 | 62.26 | 68.62 |
| **gpt-oss-120B** | 87.11 | 87.87 | 88.59 | 87.87 | 86.83 | 85.57 | 88.55 | 88.28 | 77.14 | 52.30 | 67.94 | 77.05 |
| **LLaMA-3.1-8B** | 57.53 | 75.61 | 63.66 | 61.63 | 50.81 | 41.75 | 64.16 | 63.17 | 36.88 | 22.77 | 19.21 | 15.42 |
| **LLaMA-3.1-70B** | 77.95 | 85.80 | 84.49 | 83.27 | 77.91 | 67.40 | 83.63 | 84.72 | 70.74 | 40.71 | 44.68 | 49.10 |
| **LLaMA-3.2-3B** | 49.41 | 64.20 | 54.33 | 49.41 | 41.34 | 31.42 | 50.41 | 45.40 | 33.59 | 13.57 | 18.39 | 17.76 |
| **LLaMA-3.3-70B** | 63.44 | 86.20 | 85.84 | 85.30 | 66.14 | 76.28 | 85.08 | 85.71 | 72.81 | 41.79 | 48.29 | 51.17 |
| **LLaMA-4-Scout** | 84.31 | 86.93 | 86.43 | 86.97 | 84.36 | 82.46 | 85.98 | 86.52 | 78.04 | 50.18 | 50.63 | 69.21 |
| **LLaMA-4-Maverick** | 87.60 | 89.31 | 89.95 | 89.77 | 86.74 | 86.29 | 90.17 | 89.81 | 84.13 | 54.01 | 69.07 | 77.37 |
| **Mistral-7B-v0.3** | 28.36 | 37.06 | 21.01 | 39.81 | 24.84 | 28.27 | 29.53 | 37.78 | 17.94 | 14.56 | 10.78 | 10.41 |
| **Mistral-Small-3.1-24B** | 76.56 | 83.59 | 80.30 | 81.47 | 75.11 | 70.38 | 81.79 | 81.56 | 38.46 | 13.53 | 14.16 | 21.19 |
| **Phi-4-mini** | 41.84 | 71.15 | 53.65 | 55.55 | 38.19 | 31.20 | 49.82 | 51.53 | 30.57 | 15.15 | 19.84 | 16.95 |
| **Phi-4-mini-Reasoning** | 36.11 | 73.67 | 52.25 | 60.10 | 27.37 | 13.89 | 50.45 | 43.64 | 27.82 | 21.55 | 18.67 | 19.79 |
| **Phi-4** | 70.11 | 85.57 | 81.47 | 81.11 | 73.22 | 66.73 | 81.24 | 80.34 | 52.80 | 23.17 | 38.55 | 34.13 |
| **Phi-4-Reasoning** | 79.89 | 84.94 | 84.49 | 84.31 | 83.50 | 80.03 | 78.13 | 84.99 | 72.23 | 17.18 | 49.01 | 38.64 |
| **Qwen2.5-3B** | 61.41 | 67.04 | 58.16 | 56.54 | 50.81 | 44.50 | 55.77 | 55.59 | 15.46 | 9.33 | 13.53 | 11.36 |
| **Qwen2.5-7B** | 70.92 | 73.76 | 68.35 | 65.96 | 63.12 | 57.62 | 66.77 | 66.10 | 32.46 | 9.20 | 24.84 | 19.03 |
| **Qwen2.5-14B** | 76.51 | 80.66 | 76.60 | 76.69 | 72.90 | 68.49 | 76.10 | 75.65 | 37.38 | 27.28 | 36.20 | 35.08 |
| **Qwen2.5-72B** | 83.59 | 85.57 | 84.49 | 84.22 | 81.70 | 79.31 | 84.63 | 85.44 | 55.23 | 43.78 | 41.12 | 43.91 |
| **QwQ-32B** | 84.81 | 86.02 | 87.11 | 85.53 | 82.10 | 79.35 | 87.11 | 85.93 | 62.62 | 23.26 | 27.50 | 30.43 |
| **Qwen3-1.7B** | 57.57 | 64.74 | 55.91 | 53.74 | 46.39 | 42.92 | 53.34 | 53.61 | 22.77 | 23.13 | 26.96 | 23.99 |
| **Qwen3-4B** | 73.31 | 76.83 | 70.65 | 70.47 | 66.46 | 61.77 | 70.65 | 69.57 | 19.88 | 13.53 | 15.55 | 10.14 |
| **Qwen3-4B-thinking** | 76.51 | 79.80 | 79.35 | 77.95 | 75.02 | 72.72 | 76.69 | 78.72 | 30.03 | 16.95 | 13.84 | 7.66 |
| **Qwen3-8B** | 77.41 | 81.02 | 75.70 | 76.38 | 72.05 | 68.39 | 75.02 | 74.84 | 31.83 | 10.37 | 23.62 | 12.04 |
| **Qwen3-8B-thinking** | 81.61 | 83.68 | 81.47 | 83.23 | 80.61 | 78.09 | 82.01 | 82.42 | 36.93 | 14.29 | 17.94 | 12.62 |
| **Qwen3-14B** | 80.34 | 84.13 | 80.84 | 80.16 | 77.50 | 74.93 | 80.93 | 81.56 | 47.29 | 15.24 | 24.71 | 26.28 |
| **Qwen3-14B-thinking** | 83.54 | 86.02 | 84.85 | 84.90 | 83.72 | 81.24 | 84.81 | 84.49 | 64.16 | 17.63 | 25.25 | 27.64 |
| **Baichuan-M2-32B** | 84.72 | 86.11 | 83.81 | 82.60 | 81.15 | 78.40 | 83.23 | 83.45 | 32.87 | 25.97 | 33.41 | 29.49 |
| **Bio-Medical-LLaMA-3-8B** | 50.32 | 69.52 | 59.02 | 58.57 | 48.87 | 44.54 | 57.12 | 54.42 | 38.91 | 35.84 | 33.59 | 33.54 |
| **MediPhi** | 38.10 | 67.94 | 62.35 | 61.09 | 30.21 | 32.33 | 58.25 | 54.28 | 16.01 | 14.20 | 15.73 | 16.68 |
| **MedGemma-4B** | 54.55 | 68.53 | 63.35 | 61.27 | 54.46 | 49.28 | 62.85 | 62.35 | 45.49 | 20.20 | 21.06 | 31.51 |
| **MedGemma-27B** | 82.24 | 86.38 | 84.22 | 84.63 | 81.74 | 79.80 | 85.57 | 86.20 | 77.50 | 17.72 | 51.08 | 65.01 |
| **MedReason-8B** | 47.20 | 62.89 | 17.49 | 14.83 | 47.66 | 39.81 | 26.47 | 35.89 | 13.03 | 8.12 | 13.66 | 12.13 |
| **HuatuoGPT-o1-7B** | 74.17 | 69.30 | 71.24 | 71.55 | 65.33 | 62.35 | 72.36 | 70.78 | 6.67 | 6.90 | 6.76 | 5.95 |
| **HuatuoGPT-o1-8B** | 64.29 | 73.35 | 71.10 | 70.06 | 62.13 | 54.91 | 69.39 | 69.12 | 49.91 | 9.02 | 8.48 | 8.84 |
| **HuatuoGPT-o1-70B** | 72.86 | 85.39 | 84.36 | 84.17 | 72.95 | 77.55 | 84.40 | 84.54 | 74.75 | 43.10 | 49.50 | 56.54 |
| **HuatuoGPT-o1-72B** | 84.31 | 85.30 | 86.07 | 83.95 | 83.09 | 78.58 | 86.65 | 86.47 | 62.31 | 43.60 | 43.19 | 29.35 |
| **OpenBioLLM-8B** | 22.45 | 35.84 | 23.53 | 31.70 | 8.66 | 12.04 | 34.13 | 37.11 | 11.36 | 6.99 | 10.28 | 10.10 |
| **OpenBioLLM-70B** | 31.24 | 72.23 | 75.29 | 51.04 | 32.78 | 30.52 | 59.11 | 71.10 | 30.75 | 9.60 | 11.72 | 23.62 |

**Supplementary Tab. 105:** Zero-Shot performance evaluation of 56 LLMs on HeadQA (Run 2).

| LLMs | Chinese | English | French | German | Japanese | Korean | Portuguese | Spanish | Swahili | Wolof | Yoruba | Zulu |
|---|---|---|---|---|---|---|---|---|---|---|---|---|
| **Proprietary LLMs** | | | | | | | | | | | | |
| **Claude-3.5-Haiku** | 77.01 | 83.14 | 82.82 | 81.79 | 79.35 | 75.02 | 83.41 | 83.05 | 59.29 | 18.67 | 42.34 | 43.01 |
| **Claude-4.0-Sonnet** | 87.74 | 89.40 | 90.13 | 89.27 | 88.37 | 88.37 | 90.80 | 90.71 | 80.52 | 55.59 | 71.15 | 76.47 |
| **Gemini-2.5-Flash** | 88.68 | 87.92 | 90.26 | 89.36 | 89.09 | 88.91 | 90.08 | 90.22 | 87.96 | 78.63 | 80.52 | 83.23 |
| **GPT-4o-mini** | 77.05 | 82.96 | 82.87 | 82.15 | 78.18 | 72.59 | 81.88 | 82.60 | 68.89 | 36.47 | 48.24 | 58.66 |
| **GPT-4o** | 87.24 | 87.92 | 89.22 | 88.05 | 87.56 | 85.08 | 89.09 | 89.27 | 83.77 | 37.24 | 61.41 | 73.85 |
| **GPT-4.1-nano** | 76.69 | 84.76 | 82.64 | 79.89 | 74.12 | 72.27 | 82.19 | 81.24 | 58.52 | 33.63 | 43.37 | 54.10 |
| **GPT-4.1-mini** | 86.11 | 88.73 | 87.87 | 87.47 | 86.34 | 82.82 | 88.68 | 88.41 | 77.46 | 28.18 | 61.95 | 72.45 |
| **GPT-4.1** | 88.28 | 89.90 | 90.35 | 89.63 | 89.77 | 86.25 | 90.62 | 89.18 | 86.20 | 58.48 | 67.45 | 78.36 |
| **GPT-5-nano** | 67.13 | 81.20 | 78.45 | 77.68 | 73.81 | 66.14 | 74.84 | 76.69 | 59.29 | 17.72 | 41.39 | 49.82 |
| **GPT-5-mini** | 85.62 | 88.86 | 87.74 | 87.92 | 87.15 | 82.96 | 88.01 | 88.01 | 77.77 | 33.77 | 59.02 | 72.95 |
| **GPT-5** | 88.23 | 90.49 | 89.77 | 89.59 | 89.09 | 86.11 | 90.17 | 90.76 | 86.93 | 57.44 | 71.96 | 79.53 |
| **o4-mini** | 89.09 | 90.53 | 90.35 | 90.04 | 89.63 | 86.52 | 90.67 | 90.80 | 86.74 | 47.39 | 76.10 | 82.10 |
| **Open-Weight LLMs** | | | | | | | | | | | | |
| **DeepSeek-V3** | 87.74 | 89.59 | 88.91 | 88.95 | 87.65 | 84.22 | 88.50 | 88.91 | 77.41 | 50.41 | 57.53 | 64.92 |
| **DeepSeek-R1** | 88.32 | 89.50 | 89.68 | 89.40 | 87.78 | 86.56 | 90.13 | 90.40 | 83.68 | 66.23 | 69.61 | 77.95 |
| **DeepSeek-R1-Qwen3-8B** | 79.67 | 82.42 | 77.59 | 77.91 | 75.25 | 74.35 | 78.54 | 78.81 | 37.15 | 11.18 | 10.05 | 13.93 |
| **Gemma-3-4B** | 52.84 | 64.11 | 60.10 | 57.66 | 53.65 | 48.20 | 59.42 | 60.23 | 43.55 | 13.12 | 24.62 | 20.38 |
| **Gemma-3-12B** | 71.01 | 78.54 | 75.07 | 75.79 | 72.05 | 69.66 | 74.48 | 75.43 | 63.07 | 11.90 | 42.74 | 54.46 |
| **Gemma-3-27B** | 76.51 | 82.33 | 81.51 | 80.70 | 78.09 | 76.92 | 81.29 | 81.70 | 71.46 | 23.94 | 51.62 | 62.53 |
| **gpt-oss-20B** | 83.00 | 78.85 | 84.76 | 84.08 | 82.15 | 81.24 | 84.72 | 84.85 | 68.98 | 33.27 | 62.22 | 69.97 |
| **gpt-oss-120B** | 87.11 | 88.19 | 88.77 | 88.05 | 86.70 | 85.03 | 88.10 | 88.19 | 77.95 | 51.40 | 67.76 | 76.01 |
| **LLaMA-3.1-8B** | 56.00 | 74.71 | 62.89 | 62.40 | 51.40 | 43.33 | 62.49 | 60.50 | 37.11 | 24.26 | 19.25 | 13.75 |
| **LLaMA-3.1-70B** | 78.36 | 85.75 | 84.72 | 84.40 | 78.04 | 67.31 | 83.77 | 84.31 | 70.87 | 42.29 | 45.76 | 49.10 |
| **LLaMA-3.2-3B** | 49.14 | 63.48 | 54.55 | 49.19 | 41.79 | 30.61 | 50.59 | 44.54 | 33.45 | 13.80 | 18.94 | 17.09 |
| **LLaMA-3.3-70B** | 60.96 | 86.47 | 86.20 | 84.85 | 66.77 | 76.33 | 85.48 | 85.71 | 72.32 | 41.93 | 47.84 | 50.90 |
| **LLaMA-4-Scout** | 84.36 | 87.11 | 86.65 | 86.79 | 84.13 | 81.88 | 86.79 | 87.06 | 77.77 | 51.40 | 50.68 | 69.43 |
| **LLaMA-4-Maverick** | 88.05 | 89.68 | 89.72 | 89.45 | 87.38 | 86.07 | 90.13 | 90.17 | 84.58 | 54.55 | 69.84 | 78.18 |
| **Mistral-7B-v0.3** | 28.00 | 35.93 | 21.46 | 42.61 | 24.21 | 28.00 | 29.76 | 37.65 | 18.44 | 13.80 | 10.91 | 9.24 |
| **Mistral-Small-3.1-24B** | 76.69 | 83.63 | 80.30 | 80.48 | 74.17 | 68.71 | 81.02 | 81.79 | 38.59 | 12.40 | 13.80 | 20.78 |
| **Phi-4-mini** | 40.98 | 71.73 | 53.38 | 56.04 | 40.13 | 32.55 | 50.23 | 50.59 | 30.16 | 15.19 | 18.94 | 16.46 |
| **Phi-4-mini-Reasoning** | 36.74 | 74.66 | 51.44 | 60.14 | 28.09 | 14.56 | 49.01 | 44.18 | 29.80 | 21.82 | 17.94 | 20.02 |
| **Phi-4** | 69.75 | 84.54 | 81.15 | 81.29 | 72.86 | 66.86 | 80.61 | 80.25 | 53.29 | 23.40 | 38.64 | 35.57 |
| **Phi-4-Reasoning** | 80.30 | 85.66 | 84.45 | 84.90 | 83.86 | 79.44 | 78.94 | 85.53 | 72.54 | 18.21 | 49.23 | 37.65 |
| **Qwen2.5-3B** | 60.69 | 66.37 | 58.61 | 55.50 | 50.23 | 44.50 | 55.91 | 55.37 | 15.55 | 9.24 | 13.66 | 12.80 |
| **Qwen2.5-7B** | 71.78 | 74.62 | 68.21 | 66.46 | 62.31 | 58.34 | 66.91 | 67.04 | 31.20 | 9.65 | 24.30 | 18.30 |
| **Qwen2.5-14B** | 76.87 | 80.07 | 77.37 | 75.43 | 72.32 | 68.76 | 75.83 | 75.56 | 36.83 | 27.64 | 36.20 | 35.98 |
| **Qwen2.5-72B** | 83.32 | 85.26 | 84.54 | 84.31 | 81.61 | 78.54 | 85.53 | 84.72 | 53.83 | 44.09 | 42.34 | 43.91 |
| **QwQ-32B** | 84.94 | 86.11 | 86.74 | 85.66 | 82.73 | 79.67 | 86.52 | 85.84 | 62.08 | 24.80 | 29.53 | 30.57 |
| **Qwen3-1.7B** | 56.76 | 64.70 | 55.00 | 53.02 | 45.81 | 42.06 | 53.43 | 52.66 | 24.12 | 22.09 | 26.96 | 22.81 |
| **Qwen3-4B** | 72.45 | 76.78 | 70.42 | 69.84 | 66.41 | 60.78 | 69.66 | 69.88 | 20.56 | 14.47 | 16.01 | 9.24 |
| **Qwen3-4B-thinking** | 76.51 | 80.66 | 78.76 | 77.64 | 75.47 | 72.05 | 78.00 | 77.82 | 29.13 | 16.32 | 13.17 | 9.15 |
| **Qwen3-8B** | 77.50 | 80.34 | 75.38 | 75.92 | 72.05 | 68.39 | 74.57 | 75.47 | 30.88 | 10.28 | 22.72 | 12.08 |
| **Qwen3-8B-thinking** | 81.24 | 83.72 | 82.46 | 82.33 | 79.44 | 77.64 | 81.74 | 82.10 | 37.47 | 15.73 | 17.99 | 12.98 |
| **Qwen3-14B** | 80.61 | 84.13 | 81.70 | 80.25 | 76.78 | 74.08 | 80.16 | 81.15 | 47.43 | 15.28 | 24.26 | 26.42 |
| **Qwen3-14B-thinking** | 83.09 | 85.89 | 84.63 | 84.99 | 83.27 | 80.12 | 84.31 | 85.12 | 65.37 | 17.31 | 24.17 | 26.78 |
| **Baichuan-M2-32B** | 85.26 | 85.66 | 83.86 | 83.50 | 81.38 | 79.22 | 84.94 | 83.72 | 34.94 | 25.52 | 34.36 | 30.07 |
| **Bio-Medical-LLaMA-3-8B** | 49.95 | 69.03 | 59.38 | 58.75 | 50.00 | 44.23 | 57.12 | 55.14 | 39.04 | 36.47 | 33.99 | 33.68 |
| **MediPhi** | 37.33 | 67.54 | 62.53 | 61.23 | 31.83 | 33.09 | 58.25 | 53.47 | 16.41 | 14.02 | 15.92 | 18.62 |
| **MedGemma-4B** | 55.68 | 69.66 | 64.07 | 63.03 | 55.32 | 51.98 | 62.22 | 63.07 | 45.94 | 21.19 | 22.32 | 31.83 |
| **MedGemma-27B** | 82.37 | 85.75 | 85.44 | 85.21 | 81.88 | 81.02 | 85.84 | 85.98 | 77.77 | 18.17 | 50.72 | 65.15 |
| **MedReason-8B** | 47.88 | 64.34 | 19.52 | 14.29 | 48.20 | 41.88 | 26.51 | 33.45 | 12.80 | 8.34 | 13.21 | 11.59 |
| **HuatuoGPT-o1-7B** | 74.03 | 70.65 | 72.36 | 71.46 | 65.06 | 62.98 | 72.05 | 71.06 | 7.17 | 6.63 | 7.35 | 6.90 |
| **HuatuoGPT-o1-8B** | 65.46 | 73.49 | 70.29 | 71.46 | 62.62 | 54.51 | 69.93 | 68.44 | 50.50 | 9.24 | 8.79 | 8.84 |
| **HuatuoGPT-o1-70B** | 73.26 | 85.08 | 84.40 | 83.72 | 72.59 | 78.13 | 84.94 | 84.90 | 75.34 | 42.65 | 50.41 | 56.40 |
| **HuatuoGPT-o1-72B** | 84.27 | 85.80 | 86.52 | 84.49 | 83.00 | 77.50 | 87.02 | 87.06 | 62.44 | 45.22 | 44.00 | 29.62 |
| **OpenBioLLM-8B** | 23.13 | 34.63 | 24.89 | 31.70 | 8.75 | 12.17 | 34.76 | 36.61 | 11.95 | 7.12 | 10.32 | 12.08 |
| **OpenBioLLM-70B** | 32.55 | 72.95 | 75.07 | 51.44 | 31.79 | 30.84 | 58.84 | 70.20 | 30.84 | 10.55 | 10.96 | 25.02 |

**Supplementary Tab. 106:** Zero-Shot performance evaluation of 56 LLMs on HeadQA (Run 3).

| LLMs | Chinese | English | French | German | Japanese | Korean | Portuguese | Spanish | Swahili | Wolof | Yoruba | Zulu |
|---|---|---|---|---|---|---|---|---|---|---|---|---|
| | | | | | | Proprietary LLMs | | | | | | |
| Claude-3.5-Haiku | 76.92 | 83.14 | 82.82 | 81.83 | 79.40 | 74.89 | 83.32 | 83.00 | 59.24 | 19.79 | 42.70 | 42.92 |
| Claude-4.0-Sonnet | 87.65 | 89.77 | 89.77 | 89.40 | 88.46 | 87.69 | 90.22 | 90.98 | 80.52 | 54.96 | 70.74 | 76.10 |
| Gemini-2.5-Flash | 88.68 | 87.42 | 90.26 | 89.72 | 88.68 | 88.91 | 90.44 | 90.44 | 88.50 | 78.67 | 80.30 | 83.86 |
| GPT-4o-mini | 76.87 | 83.81 | 82.73 | 81.92 | 78.63 | 72.09 | 82.28 | 82.15 | 69.25 | 35.35 | 49.32 | 58.30 |
| GPT-4o | 86.83 | 87.83 | 88.64 | 88.23 | 87.83 | 84.99 | 88.68 | 89.13 | 83.77 | 37.51 | 61.09 | 73.22 |
| GPT-4.1-nano | 76.78 | 84.27 | 81.38 | 80.48 | 74.48 | 73.04 | 82.19 | 81.29 | 58.66 | 33.27 | 45.49 | 53.34 |
| GPT-4.1-mini | 85.75 | 88.41 | 87.47 | 87.83 | 86.97 | 83.36 | 89.00 | 88.37 | 77.82 | 26.60 | 61.27 | 72.45 |
| GPT-4.1 | 88.23 | 90.04 | 90.04 | 89.86 | 89.31 | 86.74 | 90.58 | 89.40 | 86.47 | 58.30 | 67.85 | 78.72 |
| GPT-5-nano | 66.86 | 80.66 | 78.45 | 78.13 | 73.85 | 66.73 | 75.34 | 76.87 | 58.70 | 16.68 | 41.34 | 51.04 |
| GPT-5-mini | 85.66 | 87.92 | 88.14 | 87.24 | 86.74 | 82.78 | 88.82 | 88.28 | 78.54 | 32.96 | 60.10 | 73.08 |
| GPT-5 | 88.37 | 90.26 | 89.86 | 89.50 | 89.31 | 86.25 | 90.58 | 91.12 | 86.52 | 57.35 | 71.19 | 79.58 |
| o4-mini | 88.73 | 90.26 | 90.17 | 89.68 | 89.54 | 86.93 | 90.58 | 90.53 | 86.83 | 47.25 | 76.10 | 82.28 |
| | | | | | | Open-Weight LLMs | | | | | | |
| DeepSeek-V3 | 87.42 | 89.86 | 88.50 | 88.50 | 86.97 | 84.13 | 88.41 | 88.95 | 77.68 | 51.49 | 59.78 | 65.10 |
| DeepSeek-R1 | 88.28 | 89.86 | 89.86 | 89.22 | 88.23 | 86.65 | 90.26 | 89.59 | 83.41 | 65.78 | 69.52 | 76.65 |
| DeepSeek-R1-Qwen3-8B | 79.67 | 82.10 | 77.10 | 77.91 | 75.88 | 74.03 | 79.58 | 78.45 | 36.16 | 10.23 | 11.77 | 13.48 |
| Gemma-3-4B | 53.02 | 64.47 | 59.56 | 59.02 | 53.34 | 49.19 | 58.57 | 58.93 | 42.88 | 12.13 | 26.47 | 20.33 |
| Gemma-3-12B | 70.20 | 78.00 | 76.01 | 76.10 | 71.60 | 69.66 | 76.10 | 75.79 | 62.98 | 10.73 | 43.82 | 54.10 |
| Gemma-3-27B | 76.19 | 82.37 | 80.75 | 81.47 | 77.23 | 77.01 | 80.52 | 81.15 | 71.19 | 23.81 | 52.34 | 62.08 |
| gpt-oss-20B | 83.18 | 78.45 | 85.03 | 84.76 | 82.64 | 81.38 | 85.03 | 84.54 | 68.35 | 33.77 | 61.59 | 69.57 |
| gpt-oss-120B | 87.02 | 87.96 | 88.41 | 88.14 | 86.79 | 85.98 | 87.78 | 87.78 | 78.18 | 53.34 | 68.35 | 76.47 |
| LLaMA-3.1-8B | 57.57 | 73.58 | 63.21 | 62.62 | 50.36 | 42.43 | 61.99 | 61.72 | 36.43 | 23.76 | 20.47 | 14.52 |
| LLaMA-3.1-70B | 78.31 | 85.17 | 83.95 | 82.73 | 76.33 | 66.05 | 83.27 | 84.49 | 69.93 | 41.21 | 44.95 | 49.41 |
| LLaMA-3.2-3B | 49.82 | 63.89 | 52.89 | 48.92 | 41.21 | 32.37 | 51.22 | 43.01 | 32.91 | 13.80 | 20.06 | 16.46 |
| LLaMA-3.3-70B | 62.40 | 86.11 | 85.62 | 84.58 | 64.79 | 76.47 | 85.98 | 86.20 | 72.36 | 42.61 | 47.57 | 51.17 |
| LLaMA-4-Scout | 84.17 | 86.79 | 86.16 | 86.93 | 84.31 | 82.55 | 86.34 | 87.42 | 77.59 | 50.27 | 50.81 | 69.39 |
| LLaMA-4-Maverick | 87.42 | 89.50 | 90.08 | 89.45 | 87.02 | 86.34 | 89.95 | 90.22 | 83.95 | 54.55 | 68.53 | 77.95 |
| Mistral-7B-v0.3 | 29.44 | 36.70 | 20.11 | 41.03 | 24.39 | 26.74 | 29.71 | 38.23 | 17.85 | 14.61 | 9.69 | 8.66 |
| Mistral-Small-3.1-24B | 77.05 | 83.50 | 80.70 | 80.52 | 75.61 | 68.30 | 80.34 | 81.11 | 39.18 | 10.64 | 13.93 | 20.15 |
| Phi-4-mini | 41.39 | 71.28 | 51.53 | 55.28 | 40.08 | 31.06 | 51.76 | 50.41 | 30.16 | 14.29 | 19.30 | 16.77 |
| Phi-4-mini-Reasoning | 36.25 | 73.90 | 53.56 | 60.91 | 25.92 | 14.88 | 49.82 | 44.82 | 27.95 | 19.97 | 18.58 | 18.39 |
| Phi-4 | 70.92 | 84.31 | 81.42 | 80.70 | 72.86 | 66.28 | 81.02 | 80.21 | 52.21 | 22.77 | 36.83 | 34.67 |
| Phi-4-Reasoning | 80.70 | 85.39 | 84.31 | 84.67 | 83.32 | 79.94 | 78.40 | 85.26 | 70.78 | 17.85 | 47.88 | 37.02 |
| Qwen2.5-3B | 60.41 | 67.27 | 59.42 | 56.40 | 49.86 | 44.68 | 57.35 | 55.64 | 16.41 | 9.24 | 12.44 | 10.69 |
| Qwen2.5-7B | 71.19 | 74.17 | 68.08 | 65.64 | 63.62 | 56.85 | 66.59 | 65.87 | 32.46 | 8.61 | 25.56 | 17.49 |
| Qwen2.5-14B | 77.28 | 80.12 | 77.14 | 75.70 | 73.08 | 68.44 | 76.06 | 75.74 | 37.02 | 25.65 | 35.62 | 34.72 |
| Qwen2.5-72B | 83.14 | 85.39 | 84.67 | 84.13 | 81.61 | 79.31 | 84.99 | 84.85 | 53.70 | 43.87 | 41.70 | 43.55 |
| QwQ-32B | 85.30 | 86.16 | 86.43 | 85.21 | 82.28 | 78.58 | 86.38 | 86.11 | 62.76 | 23.26 | 29.85 | 29.71 |
| Qwen3-1.7B | 59.02 | 65.96 | 55.59 | 53.65 | 45.13 | 41.75 | 52.93 | 52.30 | 24.17 | 22.41 | 24.89 | 24.17 |
| Qwen3-4B | 72.41 | 76.01 | 70.38 | 70.74 | 65.10 | 60.69 | 69.70 | 69.52 | 18.21 | 13.98 | 16.86 | 10.14 |
| Qwen3-4B-thinking | 76.33 | 80.79 | 79.26 | 78.22 | 75.97 | 71.51 | 78.18 | 78.58 | 30.30 | 18.35 | 13.89 | 8.79 |
| Qwen3-8B | 76.83 | 80.66 | 75.56 | 76.19 | 72.54 | 66.86 | 76.10 | 75.47 | 30.48 | 9.92 | 22.14 | 10.10 |
| Qwen3-8B-thinking | 81.33 | 83.81 | 82.42 | 82.46 | 79.98 | 79.13 | 81.92 | 82.55 | 36.34 | 15.06 | 17.04 | 11.45 |
| Qwen3-14B | 81.38 | 84.13 | 81.20 | 80.57 | 76.56 | 73.17 | 80.48 | 80.93 | 47.34 | 14.97 | 25.52 | 28.00 |
| Qwen3-14B-thinking | 83.68 | 85.89 | 85.12 | 84.31 | 83.77 | 80.66 | 84.45 | 85.17 | 63.62 | 18.39 | 24.12 | 28.49 |
| Baichuan-M2-32B | 84.58 | 86.43 | 84.27 | 82.87 | 82.28 | 78.94 | 83.59 | 83.59 | 34.81 | 25.56 | 33.99 | 29.53 |
| Bio-Medical-LLaMA-3-8B | 50.05 | 68.35 | 59.51 | 59.06 | 49.28 | 45.13 | 57.30 | 54.37 | 39.45 | 35.84 | 34.67 | 33.14 |
| MediPhi | 39.13 | 68.39 | 61.86 | 61.72 | 30.48 | 32.46 | 57.71 | 54.01 | 17.63 | 15.42 | 17.18 | 17.58 |
| MedGemma-4B | 55.46 | 69.79 | 64.02 | 61.18 | 55.37 | 52.80 | 63.48 | 61.63 | 45.18 | 20.33 | 22.45 | 30.88 |
| MedGemma-27B | 81.83 | 86.29 | 85.12 | 84.58 | 81.92 | 81.70 | 85.57 | 85.57 | 77.37 | 18.26 | 51.40 | 66.05 |
| MedReason-8B | 47.34 | 63.57 | 18.53 | 14.29 | 48.60 | 41.52 | 27.14 | 34.85 | 12.71 | 8.21 | 13.71 | 11.68 |
| HuatuoGPT-o1-7B | 73.72 | 69.52 | 72.27 | 71.28 | 65.51 | 63.44 | 72.05 | 71.06 | 7.62 | 7.44 | 6.76 | 6.22 |
| HuatuoGPT-o1-8B | 64.29 | 73.26 | 70.47 | 71.24 | 62.40 | 54.82 | 69.43 | 69.39 | 50.32 | 10.05 | 9.96 | 8.66 |
| HuatuoGPT-o1-70B | 72.36 | 84.90 | 84.36 | 84.67 | 72.18 | 77.01 | 84.36 | 85.17 | 74.35 | 43.51 | 49.64 | 55.59 |
| HuatuoGPT-o1-72B | 84.45 | 85.57 | 86.20 | 84.04 | 83.63 | 78.85 | 86.65 | 87.20 | 62.67 | 44.27 | 45.31 | 29.53 |
| OpenBioLLM-8B | 23.62 | 35.39 | 24.26 | 32.51 | 9.33 | 12.85 | 36.47 | 33.59 | 10.96 | 7.03 | 10.32 | 11.54 |
| OpenBioLLM-70B | 31.83 | 72.63 | 74.35 | 52.16 | 33.36 | 30.66 | 58.88 | 70.96 | 29.76 | 10.01 | 12.44 | 23.53 |

**Supplementary Tab. 107:** Zero-Shot performance evaluation of 56 LLMs on HeadQA (Run 4).

| LLMs | Chinese | English | French | German | Japanese | Korean | Portuguese | Spanish | Swahili | Wolof | Yoruba | Zulu |
|---|---|---|---|---|---|---|---|---|---|---|---|---|
| **Proprietary LLMs** | | | | | | | | | | | | |
| **Claude-3.5-Haiku** | 77.01 | 83.14 | 82.82 | 81.83 | 79.31 | 75.02 | 83.27 | 83.05 | 59.20 | 19.79 | 42.47 | 43.06 |
| **Claude-4.0-Sonnet** | 87.74 | 89.50 | 89.90 | 89.36 | 88.23 | 87.56 | 90.17 | 90.80 | 80.61 | 56.31 | 70.02 | 76.69 |
| **Gemini-2.5-Flash** | 88.86 | 88.05 | 90.17 | 89.40 | 88.86 | 88.64 | 90.04 | 90.49 | 88.68 | 77.41 | 80.39 | 83.18 |
| **GPT-4o-mini** | 77.05 | 83.86 | 82.10 | 81.51 | 77.77 | 72.45 | 82.78 | 82.01 | 69.39 | 35.12 | 48.74 | 58.21 |
| **GPT-4o** | 87.20 | 87.83 | 88.82 | 88.46 | 87.74 | 85.26 | 88.91 | 89.22 | 84.27 | 37.83 | 61.95 | 73.76 |
| **GPT-4.1-nano** | 76.56 | 84.63 | 81.56 | 81.42 | 73.40 | 71.60 | 82.28 | 81.56 | 59.20 | 33.41 | 44.36 | 53.52 |
| **GPT-4.1-mini** | 85.93 | 88.68 | 87.74 | 87.24 | 86.38 | 82.60 | 89.27 | 88.77 | 77.77 | 26.28 | 62.22 | 72.32 |
| **GPT-4.1** | 88.37 | 89.86 | 90.35 | 89.00 | 89.95 | 86.07 | 90.35 | 89.40 | 86.74 | 59.29 | 67.36 | 78.49 |
| **GPT-5-nano** | 68.39 | 80.43 | 78.36 | 77.41 | 75.11 | 66.32 | 75.47 | 76.38 | 57.35 | 16.64 | 42.20 | 51.26 |
| **GPT-5-mini** | 85.57 | 88.19 | 88.19 | 87.65 | 86.25 | 82.46 | 88.32 | 88.50 | 77.91 | 33.14 | 59.15 | 72.45 |
| **GPT-5** | 88.46 | 90.08 | 90.13 | 89.77 | 89.00 | 86.16 | 89.99 | 90.71 | 86.93 | 57.44 | 71.60 | 79.94 |
| **o4-mini** | 88.68 | 90.31 | 90.26 | 90.17 | 89.50 | 87.29 | 90.76 | 90.22 | 86.43 | 47.88 | 75.74 | 82.64 |
| **Open-Weight LLMs** | | | | | | | | | | | | |
| **DeepSeek-V3** | 87.69 | 89.72 | 88.95 | 88.73 | 87.92 | 84.58 | 88.86 | 89.13 | 77.10 | 50.81 | 57.80 | 64.56 |
| **DeepSeek-R1** | 88.23 | 89.45 | 89.50 | 89.31 | 87.42 | 86.52 | 90.31 | 89.63 | 84.49 | 65.15 | 69.57 | 77.41 |
| **DeepSeek-R1-Qwen3-8B** | 78.72 | 82.15 | 77.37 | 77.14 | 75.16 | 73.58 | 77.77 | 78.67 | 35.53 | 11.45 | 10.60 | 12.22 |
| **Gemma-3-4B** | 52.34 | 64.92 | 59.29 | 58.43 | 53.61 | 48.42 | 60.50 | 59.20 | 42.47 | 12.71 | 25.43 | 19.84 |
| **Gemma-3-12B** | 70.65 | 78.67 | 75.61 | 75.07 | 71.73 | 69.57 | 75.20 | 75.11 | 63.62 | 11.14 | 43.28 | 54.87 |
| **Gemma-3-27B** | 76.42 | 81.47 | 80.48 | 80.97 | 78.40 | 77.32 | 80.30 | 81.83 | 71.55 | 23.35 | 52.39 | 62.40 |
| **gpt-oss-20B** | 82.10 | 79.35 | 85.12 | 84.85 | 83.36 | 81.56 | 84.72 | 84.99 | 68.58 | 33.99 | 61.05 | 68.53 |
| **gpt-oss-120B** | 86.79 | 87.96 | 88.23 | 88.19 | 86.70 | 85.53 | 88.01 | 88.68 | 78.04 | 52.61 | 68.89 | 76.69 |
| **LLaMA-3.1-8B** | 55.59 | 74.21 | 63.35 | 62.94 | 51.26 | 42.06 | 62.26 | 62.71 | 38.19 | 23.62 | 20.83 | 14.29 |
| **LLaMA-3.1-70B** | 78.09 | 85.53 | 84.81 | 83.63 | 79.08 | 66.77 | 83.86 | 84.22 | 70.78 | 41.57 | 44.59 | 50.27 |
| **LLaMA-3.2-3B** | 49.77 | 63.89 | 52.98 | 50.63 | 41.34 | 32.87 | 49.73 | 44.18 | 33.99 | 15.24 | 17.99 | 17.58 |
| **LLaMA-3.3-70B** | 61.81 | 85.89 | 85.62 | 85.03 | 66.23 | 75.79 | 86.16 | 86.16 | 73.04 | 41.39 | 47.34 | 51.80 |
| **LLaMA-4-Scout** | 84.04 | 87.15 | 86.11 | 86.56 | 84.45 | 82.19 | 86.79 | 87.06 | 78.09 | 50.23 | 50.68 | 68.98 |
| **LLaMA-4-Maverick** | 87.69 | 89.63 | 90.04 | 89.63 | 86.65 | 85.75 | 89.90 | 89.99 | 83.81 | 54.51 | 69.03 | 77.59 |
| **Mistral-7B-v0.3** | 27.50 | 36.88 | 20.47 | 40.89 | 23.67 | 28.13 | 30.75 | 35.84 | 18.21 | 15.64 | 10.50 | 10.41 |
| **Mistral-Small-3.1-24B** | 75.79 | 83.54 | 80.43 | 80.39 | 74.75 | 67.72 | 80.70 | 80.03 | 39.27 | 11.72 | 12.94 | 20.60 |
| **Phi-4-mini** | 43.15 | 71.51 | 52.89 | 56.09 | 39.59 | 32.33 | 49.41 | 49.37 | 29.94 | 13.53 | 19.34 | 18.58 |
| **Phi-4-mini-Reasoning** | 37.20 | 73.58 | 52.84 | 60.73 | 27.05 | 15.28 | 49.91 | 43.87 | 28.81 | 21.10 | 19.30 | 19.93 |
| **Phi-4** | 69.34 | 84.36 | 81.15 | 81.47 | 73.49 | 64.79 | 80.12 | 80.79 | 52.21 | 23.58 | 38.28 | 35.12 |
| **Phi-4-Reasoning** | 79.76 | 85.48 | 84.67 | 84.72 | 84.13 | 79.40 | 78.18 | 85.62 | 71.10 | 17.27 | 48.69 | 37.47 |
| **Qwen2.5-3B** | 61.72 | 67.13 | 58.21 | 55.73 | 49.73 | 45.58 | 55.55 | 54.96 | 15.96 | 9.29 | 13.84 | 12.22 |
| **Qwen2.5-7B** | 71.24 | 75.29 | 67.85 | 64.61 | 63.03 | 58.07 | 66.82 | 67.09 | 32.42 | 8.75 | 24.12 | 18.98 |
| **Qwen2.5-14B** | 76.83 | 80.66 | 76.60 | 75.25 | 72.90 | 68.62 | 75.29 | 76.60 | 37.20 | 26.69 | 35.48 | 36.20 |
| **Qwen2.5-72B** | 83.32 | 86.11 | 84.54 | 84.08 | 81.11 | 79.08 | 85.12 | 84.58 | 54.51 | 43.60 | 41.75 | 44.14 |
| **QwQ-32B** | 84.67 | 86.88 | 85.89 | 85.21 | 82.10 | 79.85 | 86.52 | 86.11 | 61.72 | 23.40 | 29.89 | 28.63 |
| **Qwen3-1.7B** | 58.03 | 64.88 | 56.22 | 52.30 | 45.27 | 42.11 | 52.16 | 52.57 | 24.08 | 22.54 | 26.33 | 22.77 |
| **Qwen3-4B** | 73.67 | 76.15 | 70.33 | 70.47 | 67.45 | 59.74 | 70.51 | 70.20 | 19.07 | 14.29 | 16.68 | 8.39 |
| **Qwen3-4B-thinking** | 76.78 | 80.07 | 78.22 | 78.40 | 75.65 | 72.41 | 76.83 | 77.59 | 29.89 | 16.05 | 12.62 | 10.41 |
| **Qwen3-8B** | 76.10 | 81.33 | 75.56 | 76.15 | 72.54 | 67.22 | 76.01 | 75.52 | 30.79 | 11.09 | 22.18 | 11.68 |
| **Qwen3-8B-thinking** | 80.70 | 84.22 | 82.28 | 82.64 | 79.85 | 78.18 | 82.10 | 81.79 | 37.60 | 13.98 | 17.63 | 11.99 |
| **Qwen3-14B** | 80.61 | 83.68 | 80.43 | 80.88 | 77.59 | 73.26 | 81.15 | 81.24 | 46.93 | 14.88 | 23.35 | 26.78 |
| **Qwen3-14B-thinking** | 83.72 | 85.75 | 84.36 | 85.17 | 83.27 | 81.33 | 85.30 | 84.85 | 63.80 | 16.77 | 26.01 | 27.37 |
| **Baichuan-M2-32B** | 84.63 | 86.11 | 84.31 | 82.24 | 81.42 | 77.82 | 83.54 | 83.90 | 34.17 | 26.19 | 32.69 | 29.44 |
| **Bio-Medical-LLaMA-3-8B** | 49.55 | 69.39 | 59.24 | 58.52 | 49.46 | 45.18 | 57.17 | 54.51 | 39.13 | 35.21 | 34.27 | 34.36 |
| **MediPhi** | 37.83 | 68.17 | 62.17 | 60.82 | 30.97 | 33.09 | 57.57 | 53.07 | 16.37 | 15.78 | 15.69 | 17.27 |
| **MedGemma-4B** | 54.10 | 69.39 | 62.40 | 62.49 | 55.28 | 50.14 | 62.98 | 63.12 | 45.63 | 21.37 | 22.59 | 30.57 |
| **MedGemma-27B** | 81.92 | 86.47 | 84.63 | 84.63 | 82.24 | 80.25 | 85.35 | 86.02 | 78.13 | 17.85 | 50.99 | 66.77 |
| **MedReason-8B** | 47.70 | 64.25 | 17.49 | 14.79 | 48.06 | 41.34 | 26.38 | 34.72 | 13.03 | 8.25 | 12.94 | 11.36 |
| **HuatuoGPT-o1-7B** | 73.90 | 70.02 | 72.09 | 72.95 | 64.07 | 63.21 | 72.59 | 70.69 | 7.44 | 7.17 | 7.53 | 5.82 |
| **HuatuoGPT-o1-8B** | 64.88 | 72.81 | 70.65 | 70.24 | 61.95 | 55.18 | 68.44 | 69.70 | 50.36 | 10.14 | 8.84 | 7.62 |
| **HuatuoGPT-o1-70B** | 73.53 | 85.12 | 83.86 | 84.36 | 72.95 | 77.01 | 84.67 | 84.94 | 75.11 | 45.04 | 50.95 | 57.84 |
| **HuatuoGPT-o1-72B** | 84.40 | 85.57 | 86.29 | 85.08 | 82.82 | 79.04 | 86.79 | 87.02 | 61.95 | 44.32 | 46.84 | 28.54 |
| **OpenBioLLM-8B** | 22.09 | 34.72 | 24.62 | 32.24 | 9.38 | 11.86 | 34.22 | 34.99 | 10.55 | 7.03 | 10.78 | 11.00 |
| **OpenBioLLM-70B** | 32.37 | 73.94 | 75.02 | 53.56 | 32.37 | 31.29 | 59.47 | 71.15 | 31.92 | 10.37 | 12.17 | 23.40 |

**Supplementary Tab. 108:** Zero-Shot performance evaluation of 56 LLMs on HeadQA (Run 5).

| LLMs | Chinese | English | French | German | Japanese | Korean | Portuguese | Spanish | Swahili | Wolof | Yoruba | Zulu | Overall |
|---|---|---|---|---|---|---|---|---|---|---|---|---|---|
| **Proprietary LLMs** | | | | | | | | | | | | | |
| Claude-3.5-Haiku | 72.64±0.36 | 78.40±0.00 | 80.00±0.00 | 75.36±0.36 | 74.40±0.00 | 72.00±0.00 | 80.80±0.00 | 80.00±0.00 | 49.76±0.36 | 33.76±0.67 | 43.20±0.00 | 35.20±0.00 | 64.63±0.08 |
| Claude-4.0-Sonnet | 83.04±1.04 | 87.04±1.19 | 88.16±1.04 | 86.88±0.72 | 83.68±1.21 | 87.04±0.67 | 88.16±1.04 | 88.32±0.91 | 80.00±0.98 | 56.32±2.30 | 68.96±0.36 | 70.24±1.04 | 80.65±0.45 |
| Gemini-2.5-Flash | 85.92±0.91 | 88.00±0.80 | 88.96±1.04 | 88.64±1.04 | 87.68±1.21 | 86.08±1.34 | 88.48±0.72 | 88.48±1.66 | 88.96±1.31 | 77.76±0.67 | 83.04±1.91 | 82.24±0.67 | 86.19±0.18 |
| GPT-4o-mini | 68.16±1.73 | 75.52±1.34 | 78.88±2.30 | 73.28±1.66 | 70.24±1.19 | 62.88±1.75 | 76.32±2.92 | 74.72±1.45 | 61.76±1.73 | 27.84±1.73 | 38.08±2.97 | 45.28±2.92 | 62.75±0.28 |
| GPT-4o | 83.20±0.98 | 85.44±1.04 | 87.04±1.54 | 87.52±0.91 | 82.88±1.07 | 82.40±1.60 | 85.76±0.88 | 85.44±0.36 | 82.08±2.09 | 36.16±3.46 | 52.48±1.45 | 71.84±2.49 | 76.85±0.63 |
| GPT-4.1-nano | 66.08±3.69 | 78.08±2.30 | 73.44±1.99 | 72.96±3.27 | 61.92±2.01 | 57.12±4.22 | 74.56±1.31 | 71.68±1.56 | 52.00±1.39 | 29.44±1.19 | 35.20±2.19 | 42.72±2.30 | 59.60±0.89 |
| GPT-4.1-mini | 80.32±2.09 | 86.08±0.91 | 85.92±0.72 | 86.40±1.70 | 82.08±1.21 | 80.48±1.45 | 85.12±1.21 | 84.64±0.67 | 75.84±2.07 | 25.44±3.46 | 56.64±2.49 | 68.48±2.30 | 74.79±0.74 |
| GPT-4.1 | 84.32±1.56 | 84.00±1.50 | 87.36±1.73 | 84.80±0.98 | 86.72±1.34 | 81.12±1.21 | 88.16±0.67 | 85.28±0.72 | 84.96±1.19 | 52.64±1.19 | 64.80±2.33 | 73.92±1.34 | 79.84±0.57 |
| GPT-5-nano | 51.84±3.89 | 64.00±1.88 | 60.96±1.82 | 65.92±3.82 | 68.00±1.50 | 52.64±2.74 | 64.48±0.91 | 63.68±4.14 | 57.12±2.30 | 17.76±3.46 | 39.20±2.47 | 40.00±0.98 | 53.80±0.81 |
| GPT-5-mini | 80.00±1.26 | 84.00±1.13 | 85.60±1.70 | 85.76±1.91 | 82.88±0.91 | 80.32±3.08 | 85.44±1.31 | 84.48±0.91 | 80.00±2.04 | 29.92±3.47 | 61.60±2.99 | 73.28±2.30 | 76.11±0.78 |
| GPT-5 | 84.96±1.43 | 87.20±0.98 | 87.52±0.91 | 87.68±1.84 | 85.92±1.21 | 86.40±1.50 | 86.72±2.16 | 86.24±1.73 | 86.72±2.63 | 56.96±2.07 | 73.28±0.91 | 81.92±1.45 | 82.63±0.26 |
| o4-mini | 84.16±1.04 | 88.48±1.21 | 90.08±1.34 | 88.64±0.67 | 87.04±0.67 | 85.44±1.43 | 88.80±0.80 | 88.64±0.67 | 87.84±0.36 | 44.96±4.54 | 79.04±1.73 | 82.72±2.30 | 82.99±0.53 |
| **Open-Weight LLMs** | | | | | | | | | | | | | |
| DeepSeek-V3 | 82.56±0.88 | 84.48±1.07 | 86.08±2.23 | 82.40±1.13 | 84.96±0.67 | 80.96±1.19 | 84.48±2.57 | 87.36±1.82 | 76.96±1.82 | 47.36±2.49 | 54.56±1.82 | 59.36±1.43 | 75.96±0.64 |
| DeepSeek-R1 | 84.96±1.31 | 87.68±0.91 | 88.32±1.66 | 87.84±2.29 | 88.00±1.60 | 78.72±3.38 | 87.52±0.91 | 88.64±1.82 | 84.16±1.54 | 64.80±3.35 | 71.52±0.44 | 74.88±2.09 | 82.25±0.54 |
| DeepSeek-R1-Qwen3-8B | 70.24±1.73 | 73.76±3.46 | 66.56±3.81 | 64.00±2.71 | 63.84±1.91 | 63.68±4.14 | 66.56±1.04 | 66.24±3.51 | 28.32±2.30 | 11.68±1.56 | 12.96±4.85 | 13.12±4.48 | 50.08±0.72 |
| Gemma-3-4B | 42.88±1.84 | 52.96±3.37 | 47.68±2.01 | 45.76±2.85 | 42.08±1.84 | 33.60±4.83 | 44.96±2.49 | 44.96±3.68 | 35.52±2.23 | 10.72±2.57 | 15.36±3.85 | 11.84±1.54 | 35.69±0.98 |
| Gemma-3-12B | 61.92±2.75 | 68.00±2.71 | 64.80±2.71 | 65.44±1.91 | 55.36±1.43 | 54.72±4.26 | 66.24±1.73 | 65.76±3.46 | 53.60±1.60 | 10.08±3.03 | 27.04±2.07 | 41.12±2.37 | 52.84±0.65 |
| Gemma-3-27B | 67.84±1.54 | 74.72±3.33 | 71.36±2.85 | 74.24±2.43 | 66.88±2.81 | 64.16±1.54 | 72.32±1.07 | 72.96±2.15 | 62.40±3.49 | 16.16±1.54 | 41.76±3.81 | 56.64±1.54 | 61.79±0.21 |
| gpt-oss-20B | 81.28±0.91 | 76.96±2.96 | 82.08±2.63 | 84.32±2.57 | 80.80±2.47 | 76.16±2.62 | 82.40±1.60 | 81.12±1.34 | 68.64±2.96 | 28.48±2.30 | 58.56±4.71 | 66.88±2.23 | 72.31±0.64 |
| gpt-oss-120B | 86.40±2.88 | 88.16±1.54 | 88.64±1.04 | 90.24±0.67 | 87.04±0.67 | 84.96±1.64 | 88.00±0.98 | 87.68±0.91 | 78.08±1.56 | 55.20±2.71 | 74.56±0.67 | 73.92±2.30 | 81.91±0.56 |
| LLaMA-3.1-8B | 44.96±3.73 | 68.64±3.55 | 52.32±3.86 | 50.72±1.45 | 41.60±4.35 | 30.24±5.04 | 55.04±4.09 | 47.84±2.29 | 30.56±3.68 | 21.92±3.65 | 13.12±2.09 | 10.56±2.49 | 38.96±1.12 |
| LLaMA-3.1-70B | 73.60±1.50 | 80.64±2.62 | 78.88±1.45 | 78.08±2.75 | 72.48±0.72 | 66.56±1.91 | 76.48±1.07 | 77.28±1.66 | 67.84±1.99 | 43.68±2.01 | 38.56±3.17 | 43.68±1.93 | 66.48±0.70 |
| LLaMA-3.2-3B | 39.84±3.12 | 58.88±1.66 | 45.60±3.62 | 39.52±4.26 | 37.44±0.67 | 22.88±2.37 | 43.84±4.43 | 27.84±4.71 | 25.76±3.81 | 12.64±2.36 | 16.32±4.69 | 12.64±3.01 | 31.93±1.72 |
| LLaMA-3.3-70B | 66.08±2.50 | 82.40±1.96 | 81.44±1.04 | 80.48±1.75 | 58.72±2.16 | 66.88±2.16 | 78.56±0.67 | 81.44±1.64 | 67.52±1.21 | 43.04±1.54 | 42.72±1.66 | 48.80±1.88 | 66.51±0.67 |
| LLaMA-4-Scout | 75.20±3.20 | 83.20±1.26 | 79.20±2.71 | 81.28±0.91 | 72.16±1.54 | 72.80±1.50 | 78.56±1.04 | 77.12±1.93 | 68.00±0.98 | 44.32±2.81 | 38.24±0.67 | 58.40±1.88 | 69.04±0.36 |
| LLaMA-4-Maverick | 83.36±1.54 | 86.24±0.36 | 88.16±0.88 | 86.88±1.07 | 83.36±0.88 | 82.88±0.91 | 88.16±1.99 | 87.68±1.45 | 80.48±1.21 | 54.72±2.01 | 67.68±2.92 | 72.48±2.37 | 80.17±0.11 |
| Mistral-7B-v0.3 | 28.48±2.63 | 32.64±1.43 | 15.52±2.09 | 39.04±3.60 | 19.36±5.32 | 19.52±3.28 | 19.20±3.39 | 30.40±2.71 | 16.96±4.32 | 18.56±2.29 | 11.20±2.19 | 10.72±1.56 | 21.80±0.52 |
| Mistral-Small-3.1-24B | 68.96±1.82 | 78.24±1.73 | 71.36±1.91 | 76.16±2.68 | 66.88±3.18 | 54.24±3.46 | 72.96±2.96 | 72.48±2.69 | 30.72±4.98 | 10.88±3.51 | 12.48±1.84 | 12.00±2.71 | 52.28±0.92 |
| Phi-4-mini | 34.56±2.74 | 58.08±1.21 | 36.80±2.33 | 40.80±1.13 | 35.84±2.62 | 28.00±2.47 | 35.68±3.08 | 34.08±3.08 | 22.88±3.03 | 14.56±6.02 | 16.64±1.04 | 16.64±3.41 | 31.21±0.73 |
| Phi-4-mini-Reasoning | 28.96±3.55 | 61.12±1.66 | 46.24±4.13 | 49.60±1.26 | 24.80±6.25 | 15.68±2.09 | 35.20±1.60 | 39.04±3.32 | 24.00±1.88 | 18.40±4.90 | 16.32±2.16 | 16.48±3.51 | 31.32±1.57 |
| Phi-4 | 59.04±3.51 | 79.20±2.65 | 69.60±2.99 | 74.56±2.07 | 62.40±2.71 | 52.00±3.88 | 71.36±3.37 | 69.44±2.29 | 41.44±2.22 | 19.04±4.13 | 30.56±5.53 | 26.24±1.73 | 54.57±1.20 |
| Phi-4-Reasoning | 79.52±2.37 | 85.60±0.80 | 85.44±0.36 | 84.16±2.07 | 83.52±2.23 | 75.52±1.75 | 74.08±3.60 | 84.00±1.70 | 72.00±2.33 | 17.76±2.36 | 44.16±3.27 | 30.72±3.42 | 68.04±0.98 |
| Qwen2.5-3B | 50.40±2.47 | 49.28±4.02 | 46.88±1.21 | 43.68±2.75 | 44.00±2.83 | 36.16±2.62 | 43.84±3.46 | 44.80±1.60 | 14.88±2.69 | 13.12±0.91 | 14.88±2.69 | 12.48±2.09 | 34.53±0.86 |
| Qwen2.5-7B | 57.76±2.43 | 64.80±1.13 | 59.52±1.45 | 52.32±5.26 | 52.96±2.07 | 48.16±1.99 | 55.20±1.13 | 54.08±2.30 | 31.20±2.83 | 14.24±1.04 | 20.16±2.36 | 14.56±1.31 | 43.75±0.66 |
| Qwen2.5-14B | 67.52±2.69 | 73.28±1.07 | 60.00±1.26 | 69.28±1.75 | 62.08±3.60 | 55.36±3.73 | 68.00±2.53 | 67.36±1.31 | 31.84±1.91 | 20.32±1.75 | 28.16±0.88 | 31.84±3.22 | 52.92±0.47 |
| Qwen2.5-72B | 76.48±1.84 | 78.88±1.21 | 79.68±1.75 | 77.60±1.26 | 75.52±1.84 | 70.08±1.66 | 80.48±1.21 | 79.20±1.13 | 50.24±1.82 | 42.24±1.04 | 28.96±2.22 | 35.68±1.84 | 64.59±0.62 |
| QwQ-32B | 81.92±1.07 | 83.68±1.45 | 84.48±1.84 | 82.56±0.88 | 79.04±1.73 | 68.96±1.04 | 84.16±1.31 | 83.68±2.09 | 64.00±3.20 | 17.92±4.62 | 15.84±3.12 | 19.84±2.62 | 63.84±0.76 |
| Qwen3-1.7B | 49.28±3.60 | 51.36±1.43 | 44.00±2.94 | 41.28±2.30 | 35.52±2.50 | 33.44±1.73 | 41.44±2.07 | 41.92±3.94 | 15.84±1.31 | 22.40±2.83 | 25.76±2.96 | 21.60±2.88 | 35.32±0.50 |
| Qwen3-4B | 61.92±1.93 | 66.88±1.21 | 56.16±4.21 | 61.28±2.50 | 53.12±1.66 | 47.36±2.29 | 56.00±3.44 | 56.96±2.15 | 17.28±2.01 | 15.52±2.01 | 16.16±3.81 | 10.08±1.56 | 43.23±0.87 |
| Qwen3-4B-thinking | 66.08±2.69 | 72.00±1.60 | 66.24±2.07 | 66.08±2.01 | 62.40±1.88 | 56.00±2.83 | 66.56±2.49 | 63.52±3.60 | 21.12±4.62 | 18.08±0.91 | 13.60±2.88 | 8.80±2.33 | 48.37±0.91 |
| Qwen3-8B | 63.68±2.37 | 70.24±2.22 | 58.88±4.85 | 63.36±3.89 | 58.56±1.19 | 53.44±3.41 | 61.44±2.74 | 60.32±3.18 | 16.96±3.37 | 11.52±2.57 | 22.24±1.54 | 9.12±1.45 | 45.81±1.01 |
| Qwen3-8B-thinking | 73.28±1.21 | 76.64±1.04 | 73.28±1.34 | 74.72±1.07 | 70.88±3.03 | 67.36±1.43 | 74.24±1.99 | 74.72±2.97 | 26.08±4.10 | 14.72±2.16 | 13.28±3.60 | 10.08±3.23 | 54.11±0.78 |
| Qwen3-14B | 73.12±1.34 | 74.08±1.45 | 73.28±2.30 | 72.80±1.96 | 67.84±2.15 | 58.88±2.30 | 75.52±1.21 | 73.44±3.32 | 38.24±3.01 | 13.12±2.44 | 13.92±2.30 | 15.84±3.46 | 54.17±0.43 |
| Qwen3-14B-thinking | 78.72±0.72 | 79.68±1.45 | 79.52±1.21 | 81.76±1.19 | 80.64±1.31 | 73.60±2.40 | 80.16±0.88 | 77.92±1.21 | 57.60±1.50 | 18.08±3.42 | 11.68±3.33 | 13.60±3.39 | 61.08±0.74 |
| Baichuan-M2-32B | 79.36±1.64 | 83.04±0.88 | 76.80±2.40 | 79.52±1.66 | 77.12±2.01 | 67.52±2.44 | 80.00±1.60 | 74.56±2.29 | 26.56±3.94 | 20.64±3.12 | 23.04±1.91 | 23.84±5.35 | 59.33±1.06 |
| Bio-Medical-LLaMA-3-8B | 43.84±1.73 | 62.88±1.34 | 46.40±2.33 | 48.16±1.31 | 41.76±1.73 | 31.52±2.16 | 40.00±1.26 | 42.88±1.34 | 30.88±1.45 | 24.80±2.94 | 27.36±2.74 | 24.48±1.75 | 38.75±1.06 |
| MediPhi | 28.96±1.91 | 53.28±3.13 | 44.16±2.43 | 50.56±2.68 | 24.16±1.91 | 29.12±3.56 | 38.72±3.18 | 39.52±2.86 | 13.44±2.07 | 17.76±1.91 | 17.28±4.33 | 20.64±3.81 | 31.47±1.09 |
| MedGemma-4B | 48.00±1.26 | 66.56±1.82 | 55.68±1.66 | 60.64±2.22 | 45.12±3.08 | 41.44±2.07 | 54.56±1.64 | 54.24±4.06 | 41.60±2.88 | 17.12±1.21 | 21.76±2.49 | 30.56±1.82 | 44.77±0.50 |
| MedGemma-27B | 75.04±0.67 | 82.56±0.67 | 80.00±1.39 | 80.80±2.33 | 74.72±0.72 | 74.08±3.51 | 80.16±1.91 | 80.16±2.68 | 76.00±3.10 | 17.60±4.08 | 45.60±1.60 | 61.60±1.96 | 69.03±0.60 |
| MedReason-8B | 46.56±2.49 | 54.56±3.98 | 16.96±3.32 | 16.64±2.62 | 39.84±2.91 | 32.00±3.25 | 18.40±3.05 | 21.76±0.67 | 9.76±1.54 | 9.60±0.98 | 9.60±2.77 | 6.72±1.56 | 23.53±0.29 |
| HuatuoGPT-o1-7B | 59.20±1.50 | 58.08±3.74 | 62.24±1.82 | 63.52±2.30 | 56.48±2.63 | 54.72±0.72 | 57.92±1.56 | 57.76±2.91 | 10.72±0.72 | 11.68±2.75 | 9.76±2.07 | 7.68±1.34 | 42.48±0.95 |
| HuatuoGPT-o1-8B | 55.52±2.63 | 63.84±3.60 | 60.32±2.57 | 57.28±1.75 | 47.52±3.74 | 42.56±3.64 | 57.76±3.32 | 58.88±5.76 | 40.64±1.73 | 11.20±1.39 | 11.68±2.30 | 8.32±2.57 | 42.96±0.90 |
| HuatuoGPT-o1-70B | 62.08±2.30 | 80.16±1.43 | 78.88±1.84 | 79.84±1.73 | 64.00±1.50 | 72.48±2.81 | 79.36±1.04 | 80.80±2.26 | 70.88±2.57 | 33.92±1.66 | 43.84±2.43 | 54.24±3.12 | 66.71±0.61 |
| HuatuoGPT-o1-72B | 79.52±1.84 | 80.32±1.66 | 84.16±1.43 | 83.04±1.04 | 80.80±1.70 | 76.48±2.16 | 85.60±1.13 | 84.00±1.70 | 61.92±3.51 | 39.68±1.66 | 36.00±1.60 | 19.04±3.12 | 67.55±0.28 |
| OpenBioLLM-8B | 17.44±4.32 | 30.24±2.22 | 14.56±2.85 | 24.80±2.99 | 9.28±3.23 | 10.40±3.25 | 24.80±3.75 | 21.60±2.53 | 12.00±2.71 | 6.56±2.85 | 9.12±3.33 | 13.76±2.43 | 16.21±1.34 |
| OpenBioLLM-70B | 25.76±1.31 | 70.72±1.56 | 63.68±2.30 | 40.80±3.96 | 26.72±4.55 | 21.44±3.81 | 47.20±3.49 | 58.56±1.91 | 29.12±4.48 | 10.72±3.13 | 14.24±2.22 | 22.72±3.33 | 35.97±0.50 |

**Supplementary Tab. 109:** Performance evaluation of 56 LLMs on MedExpQA.

| LLMs | Chinese | English | French | German | Japanese | Korean | Portuguese | Spanish | Swahili | Wolof | Yoruba | Zulu |
|---|---|---|---|---|---|---|---|---|---|---|---|---|
| **Proprietary LLMs** | | | | | | | | | | | | |
| **Claude-3.5-Haiku** | 72.00 | 78.40 | 80.00 | 75.20 | 74.40 | 72.00 | 80.80 | 80.00 | 49.60 | 34.40 | 43.20 | 35.20 |
| **Claude-4.0-Sonnet** | 84.00 | 88.80 | 87.20 | 86.40 | 84.80 | 87.20 | 86.40 | 89.60 | 80.80 | 56.00 | 68.80 | 72.00 |
| **Gemini-2.5-Flash** | 85.60 | 88.80 | 88.80 | 89.60 | 88.80 | 86.40 | 88.00 | 86.40 | 89.60 | 77.60 | 84.00 | 81.60 |
| **GPT-4o-mini** | 68.80 | 74.40 | 82.40 | 75.20 | 69.60 | 63.20 | 76.00 | 75.20 | 63.20 | 28.80 | 37.60 | 44.00 |
| **GPT-4o** | 83.20 | 86.40 | 86.40 | 87.20 | 82.40 | 84.00 | 84.80 | 85.60 | 83.20 | 36.80 | 52.00 | 75.20 |
| **GPT-4.1-nano** | 63.20 | 81.60 | 73.60 | 69.60 | 61.60 | 52.80 | 74.40 | 71.20 | 51.20 | 31.20 | 32.80 | 45.60 |
| **GPT-4.1-mini** | 78.40 | 86.40 | 86.40 | 84.00 | 81.60 | 80.80 | 85.60 | 84.00 | 76.00 | 29.60 | 56.00 | 65.60 |
| **GPT-4.1** | 86.40 | 84.80 | 88.00 | 84.80 | 85.60 | 82.40 | 88.80 | 84.80 | 84.80 | 51.20 | 68.00 | 73.60 |
| **GPT-5-nano** | 54.40 | 62.40 | 60.80 | 62.40 | 68.00 | 49.60 | 65.60 | 61.60 | 59.20 | 19.20 | 39.20 | 40.00 |
| **GPT-5-mini** | 79.20 | 84.00 | 83.20 | 83.20 | 84.00 | 82.40 | 85.60 | 83.20 | 78.40 | 28.80 | 59.20 | 75.20 |
| **GPT-5** | 84.80 | 88.00 | 87.20 | 89.60 | 85.60 | 86.40 | 87.20 | 84.00 | 84.00 | 58.40 | 72.80 | 81.60 |
| **o4-mini** | 85.60 | 89.60 | 90.40 | 88.00 | 87.20 | 84.80 | 88.80 | 88.80 | 87.20 | 52.80 | 79.20 | 83.20 |
| **Open-Weight LLMs** | | | | | | | | | | | | |
| **DeepSeek-V3** | 84.00 | 84.00 | 85.60 | 82.40 | 84.80 | 80.80 | 84.80 | 85.60 | 76.80 | 49.60 | 54.40 | 58.40 |
| **DeepSeek-R1** | 84.00 | 88.00 | 88.00 | 84.80 | 88.00 | 80.00 | 88.00 | 90.40 | 85.60 | 70.40 | 72.00 | 71.20 |
| **DeepSeek-R1-Qwen3-8B** | 70.40 | 76.00 | 72.80 | 60.80 | 60.80 | 56.80 | 67.20 | 66.40 | 31.20 | 13.60 | 14.40 | 20.80 |
| **Gemma-3-4B** | 40.00 | 49.60 | 46.40 | 48.80 | 40.00 | 41.60 | 43.20 | 51.20 | 32.80 | 8.80 | 17.60 | 12.80 |
| **Gemma-3-12B** | 59.20 | 68.80 | 61.60 | 64.00 | 56.80 | 51.20 | 64.00 | 66.40 | 52.00 | 8.80 | 28.00 | 40.00 |
| **Gemma-3-27B** | 69.60 | 69.60 | 71.20 | 72.00 | 67.20 | 63.20 | 73.60 | 75.20 | 62.40 | 16.80 | 47.20 | 54.40 |
| **gpt-oss-20B** | 82.40 | 76.00 | 84.00 | 85.60 | 84.80 | 72.00 | 80.80 | 83.20 | 64.80 | 26.40 | 55.20 | 63.20 |
| **gpt-oss-120B** | 85.60 | 86.40 | 87.20 | 90.40 | 87.20 | 86.40 | 87.20 | 88.00 | 76.80 | 58.40 | 74.40 | 73.60 |
| **LLaMA-3.1-8B** | 40.00 | 65.60 | 47.20 | 49.60 | 39.20 | 22.40 | 56.80 | 48.00 | 35.20 | 23.20 | 11.20 | 12.00 |
| **LLaMA-3.1-70B** | 72.00 | 81.60 | 79.20 | 76.00 | 72.00 | 67.20 | 76.00 | 78.40 | 67.20 | 42.40 | 34.40 | 46.40 |
| **LLaMA-3.2-3B** | 41.60 | 57.60 | 47.20 | 45.60 | 38.40 | 25.60 | 48.00 | 34.40 | 24.00 | 12.00 | 19.20 | 12.00 |
| **LLaMA-3.3-70B** | 68.00 | 80.00 | 80.00 | 80.80 | 58.40 | 65.60 | 79.20 | 84.00 | 68.00 | 42.40 | 40.80 | 48.80 |
| **LLaMA-4-Scout** | 79.20 | 84.00 | 78.40 | 80.00 | 74.40 | 73.60 | 78.40 | 79.20 | 68.00 | 44.00 | 37.60 | 58.40 |
| **LLaMA-4-Maverick** | 81.60 | 86.40 | 88.80 | 85.60 | 84.80 | 81.60 | 89.60 | 87.20 | 81.60 | 55.20 | 68.00 | 72.80 |
| **Mistral-7B-v0.3** | 24.80 | 34.40 | 15.20 | 36.00 | 17.60 | 20.00 | 17.60 | 32.00 | 12.80 | 21.60 | 14.40 | 11.20 |
| **Mistral-Small-3.1-24B** | 68.80 | 80.80 | 70.40 | 76.00 | 66.40 | 51.20 | 77.60 | 71.20 | 31.20 | 14.40 | 12.80 | 10.40 |
| **Phi-4-mini** | 34.40 | 60.00 | 35.20 | 41.60 | 38.40 | 32.00 | 36.80 | 32.80 | 24.00 | 11.20 | 15.20 | 16.00 |
| **Phi-4-mini-Reasoning** | 24.80 | 60.80 | 40.80 | 49.60 | 16.00 | 16.80 | 36.00 | 40.80 | 22.40 | 24.80 | 15.20 | 21.60 |
| **Phi-4** | 59.20 | 81.60 | 70.40 | 75.20 | 57.60 | 55.20 | 75.20 | 69.60 | 39.20 | 16.00 | 27.20 | 24.80 |
| **Phi-4-Reasoning** | 83.20 | 84.80 | 84.80 | 84.80 | 85.60 | 76.00 | 76.00 | 83.20 | 74.40 | 19.20 | 49.60 | 34.40 |
| **Qwen2.5-3B** | 48.00 | 51.20 | 45.60 | 44.80 | 42.40 | 36.00 | 43.20 | 43.20 | 12.00 | 12.80 | 11.20 | 9.60 |
| **Qwen2.5-7B** | 54.40 | 64.00 | 60.00 | 47.20 | 53.60 | 48.80 | 55.20 | 53.60 | 31.20 | 15.20 | 24.00 | 15.20 |
| **Qwen2.5-14B** | 70.40 | 72.80 | 60.00 | 72.00 | 65.60 | 51.20 | 64.80 | 66.40 | 33.60 | 20.80 | 27.20 | 28.00 |
| **Qwen2.5-72B** | 78.40 | 80.00 | 80.80 | 77.60 | 77.60 | 72.80 | 79.20 | 80.00 | 51.20 | 40.80 | 31.20 | 37.60 |
| **QwQ-32B** | 82.40 | 82.40 | 84.80 | 83.20 | 80.80 | 68.80 | 84.80 | 80.80 | 66.40 | 25.60 | 12.80 | 16.80 |
| **Qwen3-1.7B** | 52.00 | 52.80 | 41.60 | 43.20 | 36.00 | 33.60 | 43.20 | 43.20 | 13.60 | 23.20 | 28.00 | 21.60 |
| **Qwen3-4B** | 60.80 | 67.20 | 53.60 | 60.80 | 52.00 | 44.80 | 60.80 | 58.40 | 14.40 | 12.80 | 18.40 | 12.00 |
| **Qwen3-4B-thinking** | 63.20 | 72.80 | 63.20 | 69.60 | 60.80 | 56.00 | 66.40 | 66.40 | 28.80 | 18.40 | 17.60 | 12.00 |
| **Qwen3-8B** | 61.60 | 71.20 | 60.80 | 56.80 | 56.80 | 56.80 | 61.60 | 59.20 | 16.80 | 8.00 | 20.00 | 9.60 |
| **Qwen3-8B-thinking** | 73.60 | 76.00 | 73.60 | 75.20 | 74.40 | 68.80 | 76.00 | 70.40 | 32.00 | 16.00 | 12.80 | 9.60 |
| **Qwen3-14B** | 75.20 | 74.40 | 73.60 | 72.80 | 65.60 | 60.80 | 77.60 | 71.20 | 36.80 | 14.40 | 12.00 | 10.40 |
| **Qwen3-14B-thinking** | 78.40 | 81.60 | 80.00 | 80.00 | 80.80 | 72.80 | 80.00 | 76.00 | 59.20 | 17.60 | 10.40 | 9.60 |
| **Baichuan-M2-32B** | 77.60 | 82.40 | 80.80 | 78.40 | 77.60 | 67.20 | 80.80 | 72.80 | 23.20 | 22.40 | 20.80 | 24.80 |
| **Bio-Medical-LLaMA-3-8B** | 42.40 | 63.20 | 48.80 | 49.60 | 40.80 | 30.40 | 40.80 | 43.20 | 28.80 | 21.60 | 23.20 | 21.60 |
| **MediPhi** | 28.00 | 53.60 | 45.60 | 52.00 | 21.60 | 23.20 | 39.20 | 35.20 | 13.60 | 19.20 | 18.40 | 19.20 |
| **MedGemma-4B** | 47.20 | 65.60 | 56.00 | 60.00 | 48.00 | 41.60 | 52.80 | 55.20 | 39.20 | 18.40 | 24.00 | 31.20 |
| **MedGemma-27B** | 75.20 | 81.60 | 77.60 | 84.80 | 75.20 | 75.20 | 79.20 | 82.40 | 73.60 | 20.00 | 46.40 | 63.20 |
| **MedReason-8B** | 44.80 | 54.40 | 17.60 | 12.80 | 40.80 | 33.60 | 21.60 | 21.60 | 12.00 | 9.60 | 11.20 | 5.60 |
| **HuatuoGPT-o1-7B** | 58.40 | 57.60 | 61.60 | 61.60 | 52.80 | 55.20 | 58.40 | 56.80 | 11.20 | 10.40 | 7.20 | 8.80 |
| **HuatuoGPT-o1-8B** | 56.00 | 64.00 | 57.60 | 57.60 | 44.80 | 46.40 | 54.40 | 61.60 | 43.20 | 12.00 | 12.80 | 8.00 |
| **HuatuoGPT-o1-70B** | 62.40 | 79.20 | 81.60 | 80.00 | 62.40 | 69.60 | 79.20 | 81.60 | 75.20 | 35.20 | 45.60 | 51.20 |
| **HuatuoGPT-o1-72B** | 81.60 | 78.40 | 85.60 | 84.00 | 78.40 | 78.40 | 87.20 | 84.80 | 64.80 | 38.40 | 37.60 | 13.60 |
| **OpenBioLLM-8B** | 12.00 | 32.00 | 10.40 | 20.80 | 4.80 | 8.80 | 28.80 | 17.60 | 8.00 | 4.00 | 8.80 | 14.40 |
| **OpenBioLLM-70B** | 26.40 | 72.00 | 67.20 | 40.00 | 32.00 | 17.60 | 43.20 | 59.20 | 28.00 | 13.60 | 17.60 | 18.40 |

**Supplementary Tab. 110:** Zero-Shot performance evaluation of 56 LLMs on MedExpQA (Run 1).

| LLMs | Chinese | English | French | German | Japanese | Korean | Portuguese | Spanish | Swahili | Wolof | Yoruba | Zulu |
|---|---|---|---|---|---|---|---|---|---|---|---|---|
| **Proprietary LLMs** | | | | | | | | | | | | |
| **Claude-3.5-Haiku** | 72.80 | 78.40 | 80.00 | 75.20 | 74.40 | 72.00 | 80.80 | 80.00 | 50.40 | 33.60 | 43.20 | 35.20 |
| **Claude-4.0-Sonnet** | 84.00 | 87.20 | 88.00 | 86.40 | 84.00 | 87.20 | 88.80 | 87.20 | 80.80 | 58.40 | 69.60 | 70.40 |
| **Gemini-2.5-Flash** | 84.80 | 88.00 | 88.00 | 87.20 | 88.80 | 87.20 | 89.60 | 87.20 | 88.00 | 77.60 | 85.60 | 83.20 |
| **GPT-4o-mini** | 65.60 | 74.40 | 77.60 | 74.40 | 70.40 | 63.20 | 80.80 | 72.80 | 62.40 | 26.40 | 40.80 | 42.40 |
| **GPT-4o** | 83.20 | 86.40 | 88.00 | 88.80 | 82.40 | 80.80 | 85.60 | 85.60 | 80.00 | 30.40 | 52.80 | 69.60 |
| **GPT-4.1-nano** | 64.00 | 78.40 | 73.60 | 72.80 | 59.20 | 59.20 | 76.80 | 73.60 | 51.20 | 29.60 | 36.00 | 40.80 |
| **GPT-4.1-mini** | 80.00 | 87.20 | 86.40 | 88.80 | 80.80 | 80.00 | 85.60 | 84.80 | 76.80 | 26.40 | 60.80 | 68.00 |
| **GPT-4.1** | 84.80 | 84.80 | 85.60 | 84.80 | 87.20 | 82.40 | 88.80 | 84.80 | 85.60 | 54.40 | 64.00 | 75.20 |
| **GPT-5-nano** | 54.40 | 65.60 | 62.40 | 67.20 | 67.20 | 51.20 | 63.20 | 68.00 | 59.20 | 13.60 | 40.80 | 40.00 |
| **GPT-5-mini** | 78.40 | 82.40 | 85.60 | 85.60 | 81.60 | 80.00 | 85.60 | 85.60 | 78.40 | 26.40 | 57.60 | 74.40 |
| **GPT-5** | 86.40 | 88.00 | 86.40 | 84.80 | 87.20 | 85.60 | 88.80 | 88.80 | 88.80 | 55.20 | 73.60 | 82.40 |
| **o4-mini** | 84.00 | 87.20 | 89.60 | 88.80 | 88.00 | 84.80 | 88.00 | 88.80 | 88.00 | 42.40 | 78.40 | 82.40 |
| **Open-Weight LLMs** | | | | | | | | | | | | |
| **DeepSeek-V3** | 82.40 | 84.00 | 88.00 | 84.00 | 85.60 | 79.20 | 81.60 | 87.20 | 76.80 | 43.20 | 54.40 | 57.60 |
| **DeepSeek-R1** | 83.20 | 86.40 | 87.20 | 89.60 | 86.40 | 74.40 | 88.80 | 87.20 | 84.00 | 64.80 | 71.20 | 76.00 |
| **DeepSeek-R1-Qwen3-8B** | 70.40 | 68.00 | 64.00 | 63.20 | 65.60 | 66.40 | 65.60 | 69.60 | 24.80 | 11.20 | 18.40 | 12.00 |
| **Gemma-3-4B** | 44.80 | 50.40 | 47.20 | 48.80 | 44.00 | 34.40 | 47.20 | 41.60 | 37.60 | 14.40 | 11.20 | 11.20 |
| **Gemma-3-12B** | 63.20 | 64.80 | 62.40 | 66.40 | 55.20 | 52.80 | 66.40 | 70.40 | 55.20 | 7.20 | 29.60 | 44.00 |
| **Gemma-3-27B** | 65.60 | 73.60 | 71.20 | 76.00 | 71.20 | 64.00 | 71.20 | 72.80 | 61.60 | 16.00 | 42.40 | 58.40 |
| **gpt-oss-20B** | 80.80 | 72.80 | 85.60 | 84.00 | 80.80 | 76.80 | 84.00 | 80.80 | 68.80 | 25.60 | 59.20 | 68.80 |
| **gpt-oss-120B** | 87.20 | 88.00 | 88.80 | 89.60 | 87.20 | 83.20 | 89.60 | 88.00 | 77.60 | 54.40 | 74.40 | 76.00 |
| **LLaMA-3.1-8B** | 50.40 | 73.60 | 53.60 | 52.00 | 35.20 | 28.00 | 55.20 | 44.80 | 25.60 | 26.40 | 12.80 | 8.00 |
| **LLaMA-3.1-70B** | 72.80 | 77.60 | 78.40 | 76.80 | 72.00 | 65.60 | 76.00 | 77.60 | 69.60 | 45.60 | 38.40 | 42.40 |
| **LLaMA-3.2-3B** | 38.40 | 60.80 | 46.40 | 36.80 | 37.60 | 24.00 | 41.60 | 28.80 | 29.60 | 9.60 | 23.20 | 11.20 |
| **LLaMA-3.3-70B** | 62.40 | 83.20 | 80.80 | 77.60 | 55.20 | 67.20 | 79.20 | 80.00 | 65.60 | 43.20 | 44.00 | 50.40 |
| **LLaMA-4-Scout** | 71.20 | 82.40 | 76.00 | 81.60 | 72.80 | 74.40 | 79.20 | 79.20 | 68.80 | 44.00 | 38.40 | 57.60 |
| **LLaMA-4-Maverick** | 85.60 | 86.40 | 88.80 | 86.40 | 83.20 | 82.40 | 89.60 | 85.60 | 79.20 | 51.20 | 71.20 | 73.60 |
| **Mistral-7B-v0.3** | 29.60 | 32.80 | 17.60 | 34.40 | 24.00 | 16.00 | 22.40 | 32.80 | 22.40 | 16.00 | 9.60 | 9.60 |
| **Mistral-Small-3.1-24B** | 67.20 | 77.60 | 74.40 | 74.40 | 68.80 | 58.40 | 72.80 | 73.60 | 28.00 | 10.40 | 9.60 | 13.60 |
| **Phi-4-mini** | 38.40 | 56.80 | 35.20 | 41.60 | 37.60 | 28.00 | 36.80 | 31.20 | 23.20 | 24.80 | 17.60 | 12.00 |
| **Phi-4-mini-Reasoning** | 28.80 | 62.40 | 52.00 | 50.40 | 25.60 | 18.40 | 36.80 | 37.60 | 26.40 | 20.00 | 19.20 | 17.60 |
| **Phi-4** | 62.40 | 76.80 | 65.60 | 76.00 | 64.00 | 45.60 | 74.40 | 71.20 | 43.20 | 22.40 | 39.20 | 28.80 |
| **Phi-4-Reasoning** | 78.40 | 86.40 | 85.60 | 84.00 | 84.00 | 75.20 | 76.80 | 85.60 | 72.80 | 19.20 | 41.60 | 27.20 |
| **Qwen2.5-3B** | 52.00 | 50.40 | 47.20 | 46.40 | 44.00 | 32.80 | 47.20 | 44.00 | 19.20 | 12.00 | 18.40 | 11.20 |
| **Qwen2.5-7B** | 56.80 | 65.60 | 57.60 | 50.40 | 53.60 | 48.00 | 53.60 | 54.40 | 26.40 | 14.40 | 18.40 | 13.60 |
| **Qwen2.5-14B** | 68.00 | 72.00 | 59.20 | 68.80 | 64.00 | 52.00 | 71.20 | 68.80 | 32.80 | 22.40 | 28.00 | 35.20 |
| **Qwen2.5-72B** | 76.00 | 80.00 | 80.80 | 76.80 | 72.80 | 69.60 | 80.80 | 78.40 | 48.00 | 42.40 | 27.20 | 35.20 |
| **QwQ-32B** | 80.80 | 84.80 | 83.20 | 83.20 | 80.00 | 70.40 | 85.60 | 84.80 | 63.20 | 17.60 | 20.80 | 24.00 |
| **Qwen3-1.7B** | 51.20 | 49.60 | 42.40 | 43.20 | 39.20 | 31.20 | 44.00 | 35.20 | 16.80 | 18.40 | 26.40 | 17.60 |
| **Qwen3-4B** | 64.00 | 67.20 | 56.80 | 60.80 | 55.20 | 50.40 | 53.60 | 56.00 | 19.20 | 15.20 | 21.60 | 8.00 |
| **Qwen3-4B-thinking** | 68.00 | 71.20 | 66.40 | 65.60 | 64.80 | 53.60 | 68.00 | 67.20 | 20.80 | 17.60 | 12.80 | 10.40 |
| **Qwen3-8B** | 61.60 | 70.40 | 51.20 | 64.00 | 59.20 | 52.80 | 57.60 | 57.60 | 20.80 | 10.40 | 21.60 | 8.00 |
| **Qwen3-8B-thinking** | 72.00 | 76.00 | 73.60 | 73.60 | 72.00 | 65.60 | 74.40 | 78.40 | 21.60 | 14.40 | 10.40 | 6.40 |
| **Qwen3-14B** | 72.00 | 76.00 | 76.00 | 71.20 | 70.40 | 61.60 | 74.40 | 72.00 | 40.00 | 11.20 | 14.40 | 14.40 |
| **Qwen3-14B-thinking** | 78.40 | 80.00 | 80.00 | 82.40 | 81.60 | 71.20 | 79.20 | 78.40 | 57.60 | 21.60 | 15.20 | 18.40 |
| **Baichuan-M2-32B** | 77.60 | 82.40 | 76.00 | 77.60 | 75.20 | 64.80 | 77.60 | 77.60 | 23.20 | 16.00 | 24.00 | 20.80 |
| **Bio-Medical-LLaMA-3-8B** | 44.80 | 60.80 | 44.80 | 47.20 | 43.20 | 32.80 | 41.60 | 44.80 | 32.80 | 24.80 | 28.80 | 24.80 |
| **MediPhi** | 32.00 | 51.20 | 43.20 | 50.40 | 24.00 | 32.80 | 42.40 | 39.20 | 14.40 | 20.00 | 23.20 | 25.60 |
| **MedGemma-4B** | 48.80 | 64.00 | 58.40 | 58.40 | 44.00 | 42.40 | 56.00 | 57.60 | 42.40 | 16.80 | 24.00 | 32.80 |
| **MedGemma-27B** | 75.20 | 82.40 | 80.80 | 79.20 | 75.20 | 74.40 | 78.40 | 80.00 | 75.20 | 19.20 | 46.40 | 58.40 |
| **MedReason-8B** | 47.20 | 55.20 | 14.40 | 19.20 | 42.40 | 33.60 | 14.40 | 22.40 | 10.40 | 10.40 | 7.20 | 7.20 |
| **HuatuoGPT-o1-7B** | 57.60 | 57.60 | 63.20 | 65.60 | 60.00 | 54.40 | 58.40 | 60.80 | 11.20 | 11.20 | 9.60 | 7.20 |
| **HuatuoGPT-o1-8B** | 56.00 | 60.00 | 61.60 | 54.40 | 43.20 | 41.60 | 57.60 | 67.20 | 39.20 | 12.00 | 10.40 | 4.80 |
| **HuatuoGPT-o1-70B** | 64.00 | 79.20 | 79.20 | 80.00 | 64.00 | 73.60 | 78.40 | 81.60 | 69.60 | 34.40 | 45.60 | 51.20 |
| **HuatuoGPT-o1-72B** | 79.20 | 79.20 | 83.20 | 84.00 | 80.80 | 72.80 | 85.60 | 86.40 | 60.80 | 40.00 | 34.40 | 21.60 |
| **OpenBioLLM-8B** | 17.60 | 31.20 | 17.60 | 28.80 | 12.80 | 16.00 | 27.20 | 23.20 | 13.60 | 4.80 | 5.60 | 9.60 |
| **OpenBioLLM-70B** | 26.40 | 69.60 | 61.60 | 45.60 | 28.00 | 20.00 | 50.40 | 55.20 | 28.80 | 13.60 | 12.00 | 26.40 |

**Supplementary Tab. 111:** Zero-Shot performance evaluation of 56 LLMs on MedExpQA (Run 2).

| LLMs | Chinese | English | French | German | Japanese | Korean | Portuguese | Spanish | Swahili | Wolof | Yoruba | Zulu |
|---|---|---|---|---|---|---|---|---|---|---|---|---|
| **Proprietary LLMs** | | | | | | | | | | | | |
| **Claude-3.5-Haiku** | 72.80 | 78.40 | 80.00 | 75.20 | 74.40 | 72.00 | 80.80 | 80.00 | 49.60 | 32.80 | 43.20 | 35.20 |
| **Claude-4.0-Sonnet** | 81.60 | 87.20 | 87.20 | 87.20 | 84.00 | 86.40 | 88.80 | 88.00 | 80.00 | 52.80 | 68.80 | 69.60 |
| **Gemini-2.5-Flash** | 87.20 | 87.20 | 88.00 | 89.60 | 86.40 | 87.20 | 88.00 | 89.60 | 87.20 | 78.40 | 81.60 | 82.40 |
| **GPT-4o-mini** | 68.00 | 77.60 | 77.60 | 73.60 | 68.80 | 63.20 | 75.20 | 76.00 | 60.80 | 26.40 | 37.60 | 48.00 |
| **GPT-4o** | 81.60 | 84.00 | 84.80 | 88.00 | 84.00 | 84.00 | 87.20 | 85.60 | 82.40 | 39.20 | 52.80 | 73.60 |
| **GPT-4.1-nano** | 71.20 | 76.00 | 73.60 | 72.00 | 61.60 | 62.40 | 74.40 | 72.80 | 54.40 | 29.60 | 38.40 | 44.00 |
| **GPT-4.1-mini** | 83.20 | 84.80 | 85.60 | 86.40 | 84.00 | 80.80 | 86.40 | 84.80 | 75.20 | 27.20 | 56.80 | 71.20 |
| **GPT-4.1** | 83.20 | 82.40 | 89.60 | 86.40 | 86.40 | 80.80 | 88.00 | 84.80 | 84.80 | 52.00 | 66.40 | 73.60 |
| **GPT-5-nano** | 48.00 | 62.40 | 59.20 | 64.00 | 70.40 | 53.60 | 64.80 | 66.40 | 56.80 | 22.40 | 35.20 | 40.80 |
| **GPT-5-mini** | 81.60 | 85.60 | 87.20 | 84.80 | 82.40 | 84.00 | 86.40 | 84.80 | 80.80 | 33.60 | 63.20 | 70.40 |
| **GPT-5** | 83.20 | 87.20 | 88.00 | 88.80 | 87.20 | 88.80 | 86.40 | 85.60 | 84.00 | 59.20 | 74.40 | 80.00 |
| **o4-mini** | 84.80 | 89.60 | 88.00 | 88.00 | 86.40 | 84.80 | 89.60 | 89.60 | 88.00 | 43.20 | 79.20 | 79.20 |
| **Open-Weight LLMs** | | | | | | | | | | | | |
| **DeepSeek-V3** | 81.60 | 83.20 | 82.40 | 80.80 | 84.00 | 80.80 | 82.40 | 88.80 | 75.20 | 48.00 | 57.60 | 60.80 |
| **DeepSeek-R1** | 85.60 | 88.80 | 89.60 | 86.40 | 86.40 | 80.80 | 86.40 | 86.40 | 84.80 | 61.60 | 71.20 | 75.20 |
| **DeepSeek-R1-Qwen3-8B** | 69.60 | 74.40 | 63.20 | 62.40 | 64.80 | 67.20 | 65.60 | 69.60 | 28.80 | 11.20 | 11.20 | 12.80 |
| **Gemma-3-4B** | 42.40 | 55.20 | 47.20 | 43.20 | 41.60 | 29.60 | 41.60 | 43.20 | 37.60 | 8.00 | 11.20 | 9.60 |
| **Gemma-3-12B** | 65.60 | 72.00 | 68.00 | 68.00 | 54.40 | 52.00 | 65.60 | 61.60 | 55.20 | 14.40 | 24.00 | 40.00 |
| **Gemma-3-27B** | 68.80 | 76.00 | 75.20 | 71.20 | 67.20 | 66.40 | 72.80 | 72.80 | 61.60 | 13.60 | 36.80 | 56.80 |
| **gpt-oss-20B** | 81.60 | 76.80 | 80.80 | 85.60 | 80.80 | 76.80 | 84.00 | 81.60 | 67.20 | 30.40 | 60.80 | 68.00 |
| **gpt-oss-120B** | 86.40 | 88.80 | 88.00 | 89.60 | 88.00 | 86.40 | 88.00 | 88.80 | 80.80 | 53.60 | 73.60 | 76.00 |
| **LLaMA-3.1-8B** | 45.60 | 71.20 | 49.60 | 51.20 | 43.20 | 32.80 | 52.00 | 51.20 | 30.40 | 20.00 | 11.20 | 8.00 |
| **LLaMA-3.1-70B** | 76.00 | 78.40 | 79.20 | 76.00 | 72.80 | 64.80 | 77.60 | 75.20 | 64.80 | 40.80 | 42.40 | 41.60 |
| **LLaMA-3.2-3B** | 43.20 | 59.20 | 48.00 | 40.00 | 36.80 | 24.00 | 48.00 | 25.60 | 27.20 | 12.00 | 13.60 | 17.60 |
| **LLaMA-3.3-70B** | 65.60 | 84.80 | 81.60 | 82.40 | 59.20 | 68.00 | 77.60 | 81.60 | 67.20 | 44.80 | 42.40 | 49.60 |
| **LLaMA-4-Scout** | 73.60 | 84.80 | 77.60 | 81.60 | 71.20 | 73.60 | 80.00 | 75.20 | 66.40 | 40.00 | 38.40 | 61.60 |
| **LLaMA-4-Maverick** | 82.40 | 85.60 | 87.20 | 88.00 | 83.20 | 83.20 | 85.60 | 88.00 | 79.20 | 55.20 | 68.80 | 75.20 |
| **Mistral-7B-v0.3** | 28.00 | 32.80 | 14.40 | 42.40 | 16.80 | 24.80 | 14.40 | 28.80 | 14.40 | 18.40 | 8.80 | 8.80 |
| **Mistral-Small-3.1-24B** | 72.00 | 76.80 | 70.40 | 74.40 | 64.80 | 50.40 | 70.40 | 76.00 | 27.20 | 8.80 | 12.00 | 13.60 |
| **Phi-4-mini** | 32.00 | 57.60 | 36.80 | 39.20 | 32.00 | 28.00 | 30.40 | 34.40 | 20.00 | 10.40 | 17.60 | 21.60 |
| **Phi-4-mini-Reasoning** | 31.20 | 62.40 | 45.60 | 48.00 | 28.80 | 16.00 | 36.00 | 43.20 | 25.60 | 17.60 | 17.60 | 12.80 |
| **Phi-4** | 62.40 | 82.40 | 70.40 | 76.80 | 63.20 | 53.60 | 70.40 | 71.20 | 41.60 | 20.00 | 28.00 | 25.60 |
| **Phi-4-Reasoning** | 76.80 | 84.80 | 85.60 | 87.20 | 81.60 | 77.60 | 76.00 | 81.60 | 68.80 | 15.20 | 44.00 | 31.20 |
| **Qwen2.5-3B** | 53.60 | 44.80 | 48.00 | 44.80 | 41.60 | 36.80 | 38.40 | 45.60 | 13.60 | 13.60 | 13.60 | 14.40 |
| **Qwen2.5-7B** | 57.60 | 66.40 | 59.20 | 60.80 | 55.20 | 44.80 | 56.80 | 51.20 | 32.00 | 13.60 | 19.20 | 15.20 |
| **Qwen2.5-14B** | 63.20 | 74.40 | 61.60 | 68.80 | 56.80 | 57.60 | 68.00 | 68.80 | 31.20 | 20.00 | 28.00 | 28.80 |
| **Qwen2.5-72B** | 74.40 | 79.20 | 76.80 | 78.40 | 75.20 | 68.80 | 81.60 | 77.60 | 49.60 | 41.60 | 31.20 | 36.80 |
| **QwQ-32B** | 80.80 | 82.40 | 87.20 | 81.60 | 77.60 | 69.60 | 83.20 | 83.20 | 68.00 | 17.60 | 16.00 | 20.00 |
| **Qwen3-1.7B** | 43.20 | 52.80 | 48.80 | 37.60 | 32.80 | 36.00 | 40.00 | 43.20 | 16.00 | 25.60 | 28.80 | 20.80 |
| **Qwen3-4B** | 60.80 | 68.00 | 52.80 | 63.20 | 54.40 | 45.60 | 56.00 | 58.40 | 18.40 | 16.00 | 14.40 | 9.60 |
| **Qwen3-4B-thinking** | 65.60 | 71.20 | 68.80 | 65.60 | 61.60 | 55.20 | 62.40 | 61.60 | 18.40 | 18.40 | 13.60 | 7.20 |
| **Qwen3-8B** | 67.20 | 72.00 | 57.60 | 64.80 | 60.00 | 48.80 | 64.00 | 57.60 | 16.00 | 14.40 | 22.40 | 7.20 |
| **Qwen3-8B-thinking** | 72.80 | 76.00 | 72.00 | 73.60 | 69.60 | 68.80 | 76.00 | 73.60 | 23.20 | 11.20 | 15.20 | 8.80 |
| **Qwen3-14B** | 72.00 | 72.00 | 73.60 | 76.00 | 68.80 | 57.60 | 75.20 | 70.40 | 33.60 | 12.00 | 15.20 | 17.60 |
| **Qwen3-14B-thinking** | 80.00 | 79.20 | 80.80 | 83.20 | 80.80 | 72.80 | 80.00 | 78.40 | 58.40 | 15.20 | 9.60 | 11.20 |
| **Baichuan-M2-32B** | 80.00 | 84.00 | 76.80 | 81.60 | 75.20 | 69.60 | 80.80 | 72.00 | 30.40 | 19.20 | 23.20 | 32.80 |
| **Bio-Medical-LLaMA-3-8B** | 41.60 | 62.40 | 43.20 | 48.80 | 39.20 | 28.80 | 38.40 | 41.60 | 31.20 | 22.40 | 26.40 | 24.80 |
| **MediPhi** | 27.20 | 50.40 | 47.20 | 50.40 | 23.20 | 29.60 | 39.20 | 40.00 | 10.40 | 17.60 | 16.00 | 15.20 |
| **MedGemma-4B** | 46.40 | 67.20 | 54.40 | 59.20 | 48.80 | 40.80 | 56.00 | 56.00 | 38.40 | 17.60 | 20.00 | 28.00 |
| **MedGemma-27B** | 74.40 | 83.20 | 80.80 | 79.20 | 75.20 | 76.80 | 83.20 | 76.80 | 80.00 | 20.00 | 44.80 | 63.20 |
| **MedReason-8B** | 43.20 | 60.80 | 22.40 | 15.20 | 37.60 | 32.00 | 20.00 | 20.80 | 8.00 | 8.00 | 8.80 | 4.80 |
| **HuatuoGPT-o1-7B** | 59.20 | 53.60 | 60.00 | 62.40 | 56.00 | 53.60 | 59.20 | 54.40 | 10.40 | 8.00 | 10.40 | 8.00 |
| **HuatuoGPT-o1-8B** | 58.40 | 68.80 | 58.40 | 59.20 | 48.80 | 44.00 | 63.20 | 53.60 | 39.20 | 8.80 | 15.20 | 7.20 |
| **HuatuoGPT-o1-70B** | 58.40 | 82.40 | 79.20 | 82.40 | 64.00 | 76.80 | 80.00 | 81.60 | 71.20 | 35.20 | 45.60 | 54.40 |
| **HuatuoGPT-o1-72B** | 76.80 | 82.40 | 85.60 | 82.40 | 80.00 | 77.60 | 85.60 | 82.40 | 66.40 | 40.80 | 34.40 | 20.00 |
| **OpenBioLLM-8B** | 17.60 | 26.40 | 14.40 | 23.20 | 7.20 | 8.00 | 26.40 | 20.80 | 10.40 | 7.20 | 14.40 | 14.40 |
| **OpenBioLLM-70B** | 27.20 | 69.60 | 64.00 | 38.40 | 24.80 | 23.20 | 51.20 | 59.20 | 26.40 | 6.40 | 12.80 | 22.40 |

**Supplementary Tab. 112:** Zero-Shot performance evaluation of 56 LLMs on MedExpQA (Run 3).

| LLMs | Chinese | English | French | German | Japanese | Korean | Portuguese | Spanish | Swahili | Wolof | Yoruba | Zulu |
|---|---|---|---|---|---|---|---|---|---|---|---|---|
| **Proprietary LLMs** | | | | | | | | | | | | |
| **Claude-3.5-Haiku** | 72.80 | 78.40 | 80.00 | 75.20 | 74.40 | 72.00 | 80.80 | 80.00 | 49.60 | 33.60 | 43.20 | 35.20 |
| **Claude-4.0-Sonnet** | 82.40 | 85.60 | 89.60 | 88.00 | 84.00 | 88.00 | 88.80 | 88.00 | 80.00 | 58.40 | 68.80 | 69.60 |
| **Gemini-2.5-Flash** | 86.40 | 87.20 | 89.60 | 88.80 | 88.00 | 85.60 | 88.00 | 90.40 | 89.60 | 78.40 | 83.20 | 81.60 |
| **GPT-4o-mini** | 68.00 | 75.20 | 76.80 | 72.00 | 72.00 | 64.80 | 76.80 | 73.60 | 63.20 | 30.40 | 33.60 | 43.20 |
| **GPT-4o** | 84.00 | 85.60 | 87.20 | 87.20 | 81.60 | 82.40 | 85.60 | 84.80 | 80.00 | 36.00 | 50.40 | 69.60 |
| **GPT-4.1-nano** | 68.80 | 78.40 | 76.00 | 72.00 | 62.40 | 58.40 | 73.60 | 71.20 | 52.00 | 28.00 | 35.20 | 43.20 |
| **GPT-4.1-mini** | 81.60 | 85.60 | 84.80 | 86.40 | 81.60 | 78.40 | 83.20 | 84.00 | 72.80 | 20.80 | 55.20 | 70.40 |
| **GPT-4.1** | 84.80 | 85.60 | 85.60 | 84.00 | 88.80 | 80.00 | 87.20 | 86.40 | 86.40 | 52.80 | 63.20 | 75.20 |
| **GPT-5-nano** | 47.20 | 66.40 | 59.20 | 72.00 | 66.40 | 52.00 | 64.80 | 57.60 | 53.60 | 15.20 | 39.20 | 38.40 |
| **GPT-5-mini** | 80.00 | 84.00 | 84.80 | 88.00 | 83.20 | 79.20 | 83.20 | 84.80 | 83.20 | 33.60 | 64.00 | 71.20 |
| **GPT-5** | 86.40 | 85.60 | 87.20 | 88.00 | 84.80 | 86.40 | 83.20 | 86.40 | 89.60 | 57.60 | 72.00 | 84.00 |
| **o4-mini** | 83.20 | 87.20 | 91.20 | 88.80 | 86.40 | 88.00 | 89.60 | 88.00 | 88.00 | 41.60 | 76.80 | 85.60 |
| **Open-Weight LLMs** | | | | | | | | | | | | |
| **DeepSeek-V3** | 82.40 | 85.60 | 87.20 | 82.40 | 85.60 | 81.60 | 88.00 | 89.60 | 80.00 | 48.80 | 52.80 | 59.20 |
| **DeepSeek-R1** | 85.60 | 88.00 | 90.40 | 88.00 | 89.60 | 76.00 | 87.20 | 88.80 | 81.60 | 64.00 | 71.20 | 76.00 |
| **DeepSeek-R1-Qwen3-8B** | 68.00 | 73.60 | 65.60 | 66.40 | 63.20 | 64.80 | 66.40 | 64.00 | 28.00 | 12.80 | 5.60 | 9.60 |
| **Gemma-3-4B** | 43.20 | 57.60 | 51.20 | 44.80 | 44.00 | 32.00 | 47.20 | 44.00 | 36.00 | 10.40 | 19.20 | 12.00 |
| **Gemma-3-12B** | 59.20 | 66.40 | 66.40 | 65.60 | 53.60 | 61.60 | 68.80 | 63.20 | 52.00 | 12.00 | 27.20 | 38.40 |
| **Gemma-3-27B** | 68.00 | 78.40 | 67.20 | 76.00 | 64.80 | 62.40 | 71.20 | 69.60 | 68.00 | 16.80 | 40.00 | 56.00 |
| **gpt-oss-20B** | 80.00 | 78.40 | 79.20 | 86.40 | 79.20 | 79.20 | 82.40 | 80.00 | 69.60 | 30.40 | 64.80 | 66.40 |
| **gpt-oss-120B** | 82.40 | 87.20 | 89.60 | 91.20 | 86.40 | 85.60 | 88.00 | 86.40 | 77.60 | 52.00 | 75.20 | 70.40 |
| **LLaMA-3.1-8B** | 44.00 | 66.40 | 56.80 | 48.80 | 45.60 | 34.40 | 50.40 | 48.00 | 28.80 | 16.80 | 16.00 | 11.20 |
| **LLaMA-3.1-70B** | 73.60 | 84.00 | 76.80 | 82.40 | 73.60 | 65.60 | 75.20 | 79.20 | 68.00 | 44.80 | 40.80 | 43.20 |
| **LLaMA-3.2-3B** | 35.20 | 56.80 | 39.20 | 40.80 | 37.60 | 20.80 | 44.00 | 21.60 | 20.00 | 13.60 | 12.80 | 9.60 |
| **LLaMA-3.3-70B** | 65.60 | 80.80 | 82.40 | 80.80 | 60.80 | 64.00 | 78.40 | 80.00 | 68.00 | 40.80 | 41.60 | 49.60 |
| **LLaMA-4-Scout** | 77.60 | 81.60 | 82.40 | 80.80 | 70.40 | 71.20 | 77.60 | 76.00 | 68.00 | 46.40 | 39.20 | 56.80 |
| **LLaMA-4-Maverick** | 84.00 | 86.40 | 87.20 | 86.40 | 82.40 | 83.20 | 86.40 | 88.00 | 80.80 | 56.00 | 67.20 | 72.00 |
| **Mistral-7B-v0.3** | 28.00 | 32.80 | 17.60 | 41.60 | 12.80 | 18.40 | 22.40 | 32.00 | 20.80 | 16.80 | 11.20 | 12.80 |
| **Mistral-Small-3.1-24B** | 68.80 | 76.80 | 69.60 | 75.20 | 63.20 | 54.40 | 73.60 | 72.80 | 28.00 | 6.40 | 13.60 | 14.40 |
| **Phi-4-mini** | 32.00 | 57.60 | 36.00 | 40.00 | 36.80 | 25.60 | 36.00 | 39.20 | 27.20 | 11.20 | 16.00 | 16.80 |
| **Phi-4-mini-Reasoning** | 26.40 | 58.40 | 44.80 | 51.20 | 21.60 | 13.60 | 32.80 | 34.40 | 22.40 | 11.20 | 16.00 | 13.60 |
| **Phi-4** | 54.40 | 78.40 | 73.60 | 72.00 | 63.20 | 54.40 | 69.60 | 69.60 | 44.00 | 23.20 | 25.60 | 27.20 |
| **Phi-4-Reasoning** | 80.00 | 86.40 | 85.60 | 83.20 | 85.60 | 72.80 | 68.00 | 85.60 | 73.60 | 15.20 | 41.60 | 27.20 |
| **Qwen2.5-3B** | 50.40 | 54.40 | 48.00 | 39.20 | 43.20 | 40.00 | 44.00 | 47.20 | 15.20 | 12.80 | 16.00 | 12.80 |
| **Qwen2.5-7B** | 59.20 | 64.00 | 61.60 | 53.60 | 52.80 | 49.60 | 55.20 | 53.60 | 32.80 | 15.20 | 18.40 | 16.00 |
| **Qwen2.5-14B** | 68.80 | 74.40 | 58.40 | 69.60 | 60.00 | 60.00 | 66.40 | 66.40 | 32.80 | 17.60 | 28.00 | 33.60 |
| **Qwen2.5-72B** | 75.20 | 77.60 | 79.20 | 76.00 | 76.80 | 70.40 | 81.60 | 80.00 | 49.60 | 43.20 | 26.40 | 36.00 |
| **QwQ-32B** | 82.40 | 83.20 | 82.40 | 81.60 | 76.80 | 68.00 | 84.80 | 86.40 | 62.40 | 13.60 | 13.60 | 19.20 |
| **Qwen3-1.7B** | 51.20 | 51.20 | 42.40 | 40.80 | 36.00 | 32.80 | 39.20 | 45.60 | 16.80 | 24.00 | 21.60 | 22.40 |
| **Qwen3-4B** | 64.00 | 64.80 | 63.20 | 64.00 | 52.80 | 47.20 | 52.00 | 58.40 | 18.40 | 18.40 | 14.40 | 11.20 |
| **Qwen3-4B-thinking** | 69.60 | 74.40 | 67.20 | 64.80 | 60.80 | 54.40 | 67.20 | 64.00 | 20.80 | 16.80 | 14.40 | 6.40 |
| **Qwen3-8B** | 63.20 | 66.40 | 60.80 | 64.00 | 58.40 | 56.80 | 64.00 | 62.40 | 19.20 | 11.20 | 23.20 | 10.40 |
| **Qwen3-8B-thinking** | 72.80 | 78.40 | 75.20 | 75.20 | 66.40 | 67.20 | 71.20 | 75.20 | 25.60 | 16.80 | 18.40 | 15.20 |
| **Qwen3-14B** | 72.80 | 73.60 | 69.60 | 72.80 | 68.80 | 56.00 | 75.20 | 78.40 | 40.00 | 16.80 | 11.20 | 18.40 |
| **Qwen3-14B-thinking** | 78.40 | 80.00 | 77.60 | 81.60 | 81.60 | 73.60 | 80.00 | 79.20 | 55.20 | 14.40 | 8.00 | 14.40 |
| **Baichuan-M2-32B** | 80.80 | 82.40 | 74.40 | 80.80 | 77.60 | 70.40 | 79.20 | 76.00 | 31.20 | 21.60 | 25.60 | 20.80 |
| **Bio-Medical-LLaMA-3-8B** | 44.80 | 64.00 | 48.00 | 48.80 | 43.20 | 31.20 | 39.20 | 43.20 | 30.40 | 26.40 | 28.00 | 24.80 |
| **MediPhi** | 28.00 | 58.40 | 40.80 | 46.40 | 25.60 | 30.40 | 39.20 | 43.20 | 16.00 | 15.20 | 17.60 | 21.60 |
| **MedGemma-4B** | 49.60 | 68.80 | 54.40 | 64.00 | 42.40 | 38.40 | 52.80 | 47.20 | 45.60 | 15.20 | 22.40 | 29.60 |
| **MedGemma-27B** | 76.00 | 83.20 | 80.80 | 80.80 | 73.60 | 68.00 | 80.80 | 83.20 | 72.80 | 18.40 | 43.20 | 61.60 |
| **MedReason-8B** | 48.80 | 50.40 | 16.00 | 17.60 | 42.40 | 26.40 | 20.00 | 22.40 | 8.80 | 9.60 | 7.20 | 7.20 |
| **HuatuoGPT-o1-7B** | 59.20 | 64.00 | 61.60 | 61.60 | 57.60 | 55.20 | 58.40 | 56.00 | 9.60 | 14.40 | 12.80 | 5.60 |
| **HuatuoGPT-o1-8B** | 51.20 | 60.80 | 60.00 | 57.60 | 52.80 | 36.80 | 56.00 | 53.60 | 40.00 | 11.20 | 10.40 | 10.40 |
| **HuatuoGPT-o1-70B** | 64.00 | 79.20 | 76.80 | 77.60 | 66.40 | 71.20 | 78.40 | 82.40 | 68.80 | 31.20 | 41.60 | 56.00 |
| **HuatuoGPT-o1-72B** | 79.20 | 80.00 | 84.00 | 83.20 | 82.40 | 76.80 | 85.60 | 82.40 | 59.20 | 37.60 | 36.00 | 20.00 |
| **OpenBioLLM-8B** | 24.00 | 31.20 | 16.80 | 25.60 | 10.40 | 10.40 | 20.80 | 22.40 | 14.40 | 11.20 | 7.20 | 16.00 |
| **OpenBioLLM-70B** | 24.80 | 69.60 | 61.60 | 44.00 | 20.00 | 27.20 | 46.40 | 60.00 | 25.60 | 11.20 | 13.60 | 20.80 |

**Supplementary Tab. 113:** Zero-Shot performance evaluation of 56 LLMs on MedExpQA (Run 4).

| LLMs | Chinese | English | French | German | Japanese | Korean | Portuguese | Spanish | Swahili | Wolof | Yoruba | Zulu |
|---|---|---|---|---|---|---|---|---|---|---|---|---|
| **Proprietary LLMs** | | | | | | | | | | | | |
| **Claude-3.5-Haiku** | 72.80 | 78.40 | 80.00 | 76.00 | 74.40 | 72.00 | 80.80 | 80.00 | 49.60 | 34.40 | 43.20 | 35.20 |
| **Claude-4.0-Sonnet** | 83.20 | 86.40 | 88.80 | 86.40 | 81.60 | 86.40 | 88.00 | 88.80 | 78.40 | 56.00 | 68.80 | 69.60 |
| **Gemini-2.5-Flash** | 85.60 | 88.80 | 90.40 | 88.00 | 86.40 | 84.00 | 88.80 | 88.80 | 90.40 | 76.80 | 80.80 | 82.40 |
| **GPT-4o-mini** | 70.40 | 76.00 | 80.00 | 71.20 | 70.40 | 60.00 | 72.80 | 76.00 | 59.20 | 27.20 | 40.80 | 48.80 |
| **GPT-4o** | 84.00 | 84.80 | 88.80 | 86.40 | 84.00 | 80.80 | 85.60 | 85.60 | 84.80 | 38.40 | 54.40 | 71.20 |
| **GPT-4.1-nano** | 63.20 | 76.00 | 70.40 | 78.40 | 64.80 | 52.80 | 73.60 | 69.60 | 51.20 | 28.80 | 33.60 | 40.00 |
| **GPT-4.1-mini** | 78.40 | 86.40 | 86.40 | 86.40 | 82.40 | 82.40 | 84.80 | 85.60 | 78.40 | 23.20 | 54.40 | 67.20 |
| **GPT-4.1** | 82.40 | 82.40 | 88.00 | 84.00 | 85.60 | 80.00 | 88.00 | 85.60 | 83.20 | 52.80 | 62.40 | 72.00 |
| **GPT-5-nano** | 55.20 | 63.20 | 63.20 | 64.00 | 68.00 | 56.80 | 64.00 | 64.80 | 56.80 | 18.40 | 41.60 | 40.80 |
| **GPT-5-mini** | 80.80 | 84.00 | 87.20 | 87.20 | 83.20 | 76.00 | 86.40 | 84.00 | 79.20 | 27.20 | 64.00 | 75.20 |
| **GPT-5** | 84.00 | 87.20 | 88.80 | 87.20 | 84.80 | 84.80 | 88.00 | 86.40 | 87.20 | 54.40 | 73.60 | 81.60 |
| **o4-mini** | 83.20 | 88.80 | 91.20 | 89.60 | 87.20 | 84.80 | 88.00 | 88.00 | 88.00 | 44.80 | 81.60 | 83.20 |
| **Open-Weight LLMs** | | | | | | | | | | | | |
| **DeepSeek-V3** | 82.40 | 85.60 | 87.20 | 82.40 | 84.80 | 82.40 | 85.60 | 85.60 | 76.00 | 47.20 | 53.60 | 60.80 |
| **DeepSeek-R1** | 86.40 | 87.20 | 86.40 | 90.40 | 89.60 | 82.40 | 87.20 | 90.40 | 84.80 | 63.20 | 72.00 | 76.00 |
| **DeepSeek-R1-Qwen3-8B** | 72.80 | 76.80 | 67.20 | 67.20 | 64.80 | 63.20 | 68.00 | 61.60 | 28.80 | 9.60 | 15.20 | 10.40 |
| **Gemma-3-4B** | 44.00 | 52.00 | 46.40 | 43.20 | 40.80 | 30.40 | 45.60 | 44.80 | 33.60 | 12.00 | 17.60 | 13.60 |
| **Gemma-3-12B** | 62.40 | 68.00 | 65.60 | 63.20 | 56.80 | 56.00 | 66.40 | 67.20 | 53.60 | 8.00 | 26.40 | 43.20 |
| **Gemma-3-27B** | 67.20 | 76.00 | 72.00 | 76.00 | 64.00 | 64.80 | 72.80 | 74.40 | 58.40 | 17.60 | 42.40 | 57.60 |
| **gpt-oss-20B** | 81.60 | 80.80 | 80.80 | 80.00 | 78.40 | 76.00 | 80.80 | 80.00 | 72.80 | 29.60 | 52.80 | 68.00 |
| **gpt-oss-120B** | 90.40 | 90.40 | 89.60 | 90.40 | 86.40 | 83.20 | 87.20 | 87.20 | 77.60 | 57.60 | 75.20 | 73.60 |
| **LLaMA-3.1-8B** | 44.80 | 66.40 | 54.40 | 52.00 | 44.80 | 33.60 | 60.80 | 47.20 | 32.80 | 23.20 | 14.40 | 13.60 |
| **LLaMA-3.1-70B** | 73.60 | 81.60 | 80.80 | 79.20 | 72.00 | 69.60 | 77.60 | 76.00 | 69.60 | 44.80 | 36.80 | 44.80 |
| **LLaMA-3.2-3B** | 40.80 | 60.00 | 47.20 | 34.40 | 36.80 | 20.00 | 37.60 | 28.80 | 28.00 | 16.00 | 12.80 | 12.80 |
| **LLaMA-3.3-70B** | 68.80 | 83.20 | 82.40 | 80.80 | 60.00 | 69.60 | 78.40 | 81.60 | 68.80 | 44.00 | 44.80 | 45.60 |
| **LLaMA-4-Scout** | 74.40 | 83.20 | 81.60 | 82.40 | 72.00 | 71.20 | 77.60 | 76.00 | 68.80 | 47.20 | 37.60 | 57.60 |
| **LLaMA-4-Maverick** | 83.20 | 86.40 | 88.80 | 88.00 | 83.20 | 84.00 | 89.60 | 89.60 | 81.60 | 56.00 | 63.20 | 68.80 |
| **Mistral-7B-v0.3** | 32.00 | 30.40 | 12.80 | 40.80 | 25.60 | 18.40 | 19.20 | 26.40 | 14.40 | 20.00 | 12.00 | 11.20 |
| **Mistral-Small-3.1-24B** | 68.00 | 79.20 | 72.00 | 80.80 | 71.20 | 56.80 | 70.40 | 68.80 | 39.20 | 14.40 | 14.40 | 8.00 |
| **Phi-4-mini** | 36.00 | 58.40 | 40.80 | 41.60 | 34.40 | 26.40 | 38.40 | 32.80 | 20.00 | 15.20 | 16.80 | 16.80 |
| **Phi-4-mini-Reasoning** | 33.60 | 61.60 | 48.00 | 48.80 | 32.00 | 13.60 | 34.40 | 39.20 | 23.20 | 18.40 | 13.60 | 16.80 |
| **Phi-4** | 56.80 | 76.80 | 68.00 | 72.80 | 64.00 | 51.20 | 67.20 | 65.60 | 39.20 | 13.60 | 32.80 | 24.80 |
| **Phi-4-Reasoning** | 79.20 | 85.60 | 85.60 | 81.60 | 80.80 | 76.00 | 73.60 | 84.00 | 70.40 | 20.00 | 44.00 | 33.60 |
| **Qwen2.5-3B** | 48.00 | 45.60 | 45.60 | 43.20 | 48.80 | 35.20 | 46.40 | 44.00 | 14.40 | 14.40 | 15.20 | 14.40 |
| **Qwen2.5-7B** | 60.80 | 64.00 | 59.20 | 49.60 | 49.60 | 49.60 | 55.20 | 57.60 | 33.60 | 12.80 | 20.80 | 12.80 |
| **Qwen2.5-14B** | 67.20 | 72.80 | 60.80 | 67.20 | 64.00 | 56.00 | 69.60 | 66.40 | 28.80 | 20.80 | 29.60 | 33.60 |
| **Qwen2.5-72B** | 78.40 | 77.60 | 80.80 | 79.20 | 75.20 | 68.80 | 79.20 | 80.00 | 52.80 | 43.20 | 28.80 | 32.80 |
| **QwQ-32B** | 83.20 | 85.60 | 84.80 | 83.20 | 80.00 | 68.00 | 82.40 | 83.20 | 60.00 | 15.20 | 16.00 | 19.20 |
| **Qwen3-1.7B** | 48.80 | 50.40 | 44.80 | 41.60 | 33.60 | 33.60 | 40.80 | 42.40 | 16.00 | 20.80 | 24.00 | 25.60 |
| **Qwen3-4B** | 60.00 | 67.20 | 54.40 | 57.60 | 51.20 | 48.80 | 57.60 | 53.60 | 16.00 | 15.20 | 12.00 | 9.60 |
| **Qwen3-4B-thinking** | 64.00 | 70.40 | 65.60 | 64.80 | 64.00 | 60.80 | 68.80 | 58.40 | 16.80 | 19.20 | 9.60 | 8.00 |
| **Qwen3-8B** | 64.80 | 71.20 | 64.00 | 67.20 | 58.40 | 52.00 | 60.00 | 64.80 | 12.00 | 13.60 | 24.00 | 10.40 |
| **Qwen3-8B-thinking** | 75.20 | 76.80 | 72.00 | 76.00 | 72.00 | 66.40 | 73.60 | 76.00 | 28.00 | 15.20 | 9.60 | 10.40 |
| **Qwen3-14B** | 73.60 | 74.40 | 73.60 | 71.20 | 65.60 | 58.40 | 75.20 | 75.20 | 40.80 | 11.20 | 16.80 | 18.40 |
| **Qwen3-14B-thinking** | 78.40 | 77.60 | 79.20 | 81.60 | 78.40 | 77.60 | 81.60 | 77.60 | 57.60 | 21.60 | 15.20 | 14.40 |
| **Baichuan-M2-32B** | 80.80 | 84.00 | 76.00 | 79.20 | 80.00 | 65.60 | 81.60 | 74.40 | 24.80 | 24.00 | 21.60 | 20.00 |
| **Bio-Medical-LLaMA-3-8B** | 45.60 | 64.00 | 47.20 | 46.40 | 42.40 | 34.40 | 40.00 | 41.60 | 31.20 | 28.80 | 30.40 | 26.40 |
| **MediPhi** | 29.60 | 52.80 | 44.00 | 53.60 | 26.40 | 29.60 | 33.60 | 40.00 | 12.80 | 16.80 | 11.20 | 21.60 |
| **MedGemma-4B** | 48.00 | 67.20 | 55.20 | 61.60 | 42.40 | 44.00 | 55.20 | 55.20 | 42.40 | 17.60 | 18.40 | 31.20 |
| **MedGemma-27B** | 74.40 | 82.40 | 80.00 | 80.00 | 74.40 | 76.00 | 79.20 | 78.40 | 78.40 | 10.40 | 47.20 | 61.60 |
| **MedReason-8B** | 48.80 | 52.00 | 14.40 | 18.40 | 36.00 | 34.40 | 16.00 | 21.60 | 9.60 | 10.40 | 13.60 | 8.80 |
| **HuatuoGPT-o1-7B** | 61.60 | 57.60 | 64.80 | 66.40 | 56.00 | 55.20 | 55.20 | 60.80 | 11.20 | 14.40 | 8.80 | 8.80 |
| **HuatuoGPT-o1-8B** | 56.00 | 65.60 | 64.00 | 57.60 | 48.00 | 44.00 | 57.60 | 58.40 | 41.60 | 12.00 | 9.60 | 11.20 |
| **HuatuoGPT-o1-70B** | 61.60 | 80.80 | 77.60 | 79.20 | 63.20 | 71.20 | 80.80 | 76.80 | 69.60 | 33.60 | 40.80 | 58.40 |
| **HuatuoGPT-o1-72B** | 80.80 | 81.60 | 82.40 | 81.60 | 82.40 | 76.80 | 84.00 | 84.00 | 58.40 | 41.60 | 37.60 | 20.00 |
| **OpenBioLLM-8B** | 16.00 | 30.40 | 13.60 | 25.60 | 11.20 | 8.80 | 20.80 | 24.00 | 13.60 | 5.60 | 9.60 | 14.40 |
| **OpenBioLLM-70B** | 24.00 | 72.80 | 64.00 | 36.00 | 28.80 | 19.20 | 44.80 | 59.20 | 36.80 | 8.80 | 15.20 | 25.60 |

**Supplementary Tab. 114:** Zero-Shot performance evaluation of 56 LLMs on MedExpQA (Run 5).

| LLMs | Chinese | English | French | German | Japanese | Korean | Portuguese | Spanish | Swahili | Wolof | Yoruba | Zulu | Overall |
|---|---|---|---|---|---|---|---|---|---|---|---|---|---|
| **Proprietary LLMs** | | | | | | | | | | | | | |
| **Claude-3.5-Haiku** | 66.87±0.04 | 77.08±0.04 | 73.04±0.04 | 73.29±0.00 | 68.03±0.06 | 68.74±0.09 | 72.11±0.09 | 73.47±0.04 | 52.54±0.08 | 34.33±0.08 | 39.83±0.00 | 37.05±0.04 | 61.36±0.03 |
| **Claude-4.0-Sonnet** | 89.57±0.40 | 91.96±0.27 | 89.55±0.57 | 90.10±0.22 | 89.79±0.33 | 89.77±0.47 | 91.22±0.13 | 90.68±0.46 | 81.82±0.23 | 59.07±0.35 | 70.83±0.56 | 74.63±0.66 | 84.08±0.09 |
| **Gemini-2.5-Flash** | 90.16±0.42 | 91.26±0.35 | 90.89±0.26 | 91.39±0.12 | 90.04±0.30 | 90.06±0.22 | 90.92±0.49 | 91.14±0.38 | 89.68±0.36 | 79.31±0.84 | 82.81±0.66 | 84.77±0.40 | 88.54±0.03 |
| **GPT-4o-mini** | 67.60±0.47 | 77.88±0.50 | 71.77±0.61 | 73.39±0.41 | 68.47±0.92 | 65.26±0.35 | 72.90±0.48 | 73.56±0.49 | 60.74±0.53 | 32.49±0.51 | 43.93±0.76 | 50.50±0.86 | 63.21±0.22 |
| **GPT-4o** | 86.21±0.52 | 89.29±0.40 | 88.45±0.49 | 88.34±0.43 | 85.44±0.99 | 84.85±0.44 | 89.18±0.25 | 88.91±0.22 | 82.73±0.46 | 32.22±1.20 | 50.05±1.11 | 69.29±0.51 | 77.91±0.18 |
| **GPT-4.1-nano** | 66.90±0.86 | 81.41±0.80 | 72.33±1.11 | 71.25±0.85 | 63.31±1.25 | 60.77±1.00 | 72.54±1.01 | 73.87±0.45 | 53.17±0.70 | 27.84±0.85 | 38.32±1.07 | 45.37±1.02 | 60.59±0.23 |
| **GPT-4.1-mini** | 85.89±1.00 | 91.18±0.38 | 86.46±0.87 | 87.87±0.23 | 85.66±0.36 | 82.41±0.78 | 87.67±0.47 | 88.12±0.42 | 78.38±0.95 | 25.33±1.18 | 58.40±1.07 | 67.62±0.85 | 77.08±0.25 |
| **GPT-4.1** | 89.02±0.47 | 89.87±0.43 | 90.01±0.15 | 89.19±0.20 | 90.04±0.44 | 86.68±0.73 | 91.55±0.26 | 89.50±0.56 | 88.50±0.45 | 51.85±0.50 | 68.37±0.55 | 75.99±0.82 | 83.38±0.15 |
| **GPT-5-nano** | 51.31±1.24 | 69.43±1.36 | 65.53±0.79 | 69.50±0.42 | 65.25±0.63 | 54.88±1.27 | 62.56±0.93 | 71.58±1.04 | 56.60±1.45 | 20.02±1.32 | 38.67±0.70 | 43.82±0.93 | 55.76±0.26 |
| **GPT-5-mini** | 86.36±0.89 | 91.11±0.44 | 88.22±0.96 | 88.27±0.72 | 87.75±0.37 | 84.59±0.31 | 89.02±0.60 | 89.03±0.49 | 80.90±0.66 | 35.63±1.03 | 61.59±0.83 | 71.50±1.16 | 79.50±0.19 |
| **GPT-5** | 91.12±0.45 | 92.99±0.31 | 92.05±0.58 | 91.19±0.48 | 91.06±0.20 | 84.37±0.36 | 92.44±0.64 | 91.97±0.44 | 89.14±0.20 | 54.00±0.88 | 74.47±1.02 | 75.38±0.67 | 85.02±0.13 |
| **o4-mini** | 93.23±0.36 | 94.33±0.43 | 93.43±0.41 | 93.31±0.21 | 93.29±0.49 | 91.97±0.59 | 93.89±0.27 | 93.38±0.61 | 91.33±0.21 | 39.75±1.59 | 80.35±0.90 | 85.58±0.58 | 86.99±0.39 |
| **Open-Weight LLMs** | | | | | | | | | | | | | |
| **DeepSeek-V3** | 86.35±0.86 | 89.19±0.44 | 86.00±0.77 | 86.57±0.35 | 85.40±0.47 | 82.83±0.53 | 86.54±0.38 | 87.30±0.50 | 75.04±0.41 | 44.32±0.40 | 56.10±1.08 | 62.01±1.22 | 77.30±0.19 |
| **DeepSeek-R1** | 90.82±0.47 | 91.39±0.26 | 90.82±0.44 | 91.01±0.50 | 90.37±0.44 | 80.90±0.55 | 91.70±0.44 | 91.47±0.23 | 87.59±0.40 | 66.14±0.68 | 72.33±0.68 | 79.07±0.81 | 85.30±0.16 |
| **DeepSeek-R1-Qwen3-8B** | 73.07±0.57 | 79.76±0.39 | 69.02±0.95 | 70.54±0.30 | 66.66±0.67 | 68.42±1.22 | 71.70±0.77 | 69.43±0.56 | 33.56±1.47 | 10.73±0.66 | 11.36±0.69 | 12.17±0.89 | 53.04±0.25 |
| **Gemma-3-4B** | 43.80±0.54 | 51.89±0.62 | 45.53±0.74 | 44.93±0.91 | 42.06±0.62 | 40.06±1.25 | 46.30±0.42 | 47.07±0.57 | 37.63±0.60 | 9.84±0.95 | 16.67±0.76 | 14.66±0.33 | 36.70±0.13 |
| **Gemma-3-12B** | 61.48±0.90 | 69.79±0.65 | 64.93±0.70 | 64.91±0.75 | 60.91±0.60 | 59.85±0.94 | 64.29±0.75 | 65.89±0.83 | 56.21±1.02 | 9.79±0.79 | 36.28±0.62 | 43.65±1.44 | 54.83±0.22 |
| **Gemma-3-27B** | 69.35±0.83 | 76.06±0.72 | 72.17±0.64 | 72.66±0.52 | 69.74±0.75 | 68.50±0.75 | 72.11±0.74 | 72.37±0.32 | 65.31±0.80 | 19.12±0.90 | 46.17±0.85 | 53.04±0.61 | 63.05±0.29 |
| **gpt-oss-20B** | 82.70±0.41 | 80.36±0.66 | 84.12±0.83 | 83.52±0.86 | 83.10±1.29 | 80.82±0.92 | 84.09±0.78 | 84.09±0.21 | 68.09±0.97 | 30.93±1.37 | 60.82±0.88 | 66.69±0.86 | 74.11±0.13 |
| **gpt-oss-120B** | 89.32±0.78 | 91.45±0.54 | 90.41±0.45 | 90.37±0.59 | 89.49±0.64 | 87.68±0.57 | 90.78±0.91 | 90.81±0.44 | 81.15±0.50 | 54.74±0.61 | 71.88±0.61 | 77.64±0.39 | 83.81±0.18 |
| **LLaMA-3.1-8B** | 45.12±0.39 | 67.01±0.67 | 49.07±0.88 | 48.75±1.09 | 38.54±0.54 | 32.96±1.54 | 49.30±1.04 | 49.85±0.97 | 36.23±0.90 | 25.59±0.95 | 17.14±0.56 | 12.52±0.78 | 39.34±0.24 |
| **LLaMA-3.1-70B** | 71.58±0.74 | 83.24±0.92 | 76.56±0.78 | 76.91±0.27 | 69.08±0.83 | 62.69±0.80 | 76.54±0.92 | 78.30±0.43 | 61.89±0.40 | 40.06±1.68 | 43.14±0.90 | 47.04±0.64 | 65.59±0.25 |
| **LLaMA-3.2-3B** | 39.34±0.77 | 56.92±0.79 | 40.67±0.50 | 35.16±0.59 | 33.83±0.68 | 27.18±0.92 | 37.94±1.08 | 36.40±1.68 | 30.82±1.09 | 16.00±0.79 | 18.49±1.78 | 15.41±0.52 | 32.35±0.37 |
| **LLaMA-3.3-70B** | 61.26±0.55 | 84.89±0.15 | 80.16±0.86 | 80.68±0.73 | 58.84±0.38 | 70.59±0.39 | 80.69±0.53 | 81.16±0.71 | 67.90±0.39 | 41.40±0.85 | 42.98±0.35 | 47.45±0.53 | 66.50±0.22 |
| **LLaMA-4-Scout** | 79.64±0.70 | 85.29±0.67 | 81.64±0.86 | 82.75±0.42 | 79.18±1.02 | 78.24±0.63 | 81.62±0.60 | 82.19±0.54 | 74.45±0.53 | 46.44±0.96 | 46.54±0.10 | 63.72±0.35 | 73.48±0.16 |
| **LLaMA-4-Maverick** | 85.80±0.52 | 90.24±0.43 | 88.72±0.33 | 88.30±0.77 | 86.26±0.69 | 85.58±0.39 | 88.97±0.48 | 88.75±0.42 | 84.32±0.44 | 56.84±0.56 | 68.74±0.43 | 75.90±0.61 | 82.37±0.22 |
| **Mistral-7B-v0.3** | 22.32±0.83 | 28.95±1.72 | 14.50±0.90 | 29.28±0.74 | 20.82±1.18 | 17.97±0.76 | 22.29±1.30 | 27.71±1.21 | 19.68±1.05 | 16.76±0.94 | 10.07±0.66 | 10.20±0.65 | 20.05±0.13 |
| **Mistral-Small-3.1-24B** | 63.98±0.87 | 75.88±1.06 | 65.09±1.48 | 67.79±1.25 | 60.38±1.38 | 54.96±0.92 | 67.92±0.72 | 68.72±0.74 | 32.36±1.20 | 11.97±0.65 | 15.43±1.98 | 19.28±0.99 | 50.31±0.20 |
| **Phi-4-mini** | 32.82±1.32 | 53.83±0.76 | 36.02±1.20 | 38.68±1.48 | 31.80±0.89 | 24.07±0.98 | 36.01±0.99 | 37.72±0.91 | 27.82±1.63 | 17.53±0.42 | 19.57±1.18 | 17.11±0.98 | 31.08±0.29 |
| **Phi-4-mini-Reasoning** | 28.72±1.79 | 65.17±0.59 | 47.42±1.31 | 47.94±0.34 | 23.53±0.73 | 14.04±0.56 | 40.12±1.70 | 34.99±0.56 | 27.64±1.24 | 23.66±1.15 | 21.56±0.82 | 20.80±1.40 | 32.97±0.61 |
| **Phi-4** | 58.62±1.18 | 79.78±0.23 | 70.01±0.85 | 71.03±0.69 | 60.52±1.70 | 54.06±0.84 | 70.16±1.13 | 70.64±1.27 | 46.53±1.11 | 25.44±0.60 | 36.65±0.94 | 31.19±1.14 | 56.22±0.29 |
| **Phi-4-Reasoning** | 80.72±0.84 | 85.42±0.49 | 84.05±0.15 | 84.56±0.99 | 83.06±0.76 | 80.19±0.77 | 75.81±1.46 | 84.37±0.47 | 72.55±0.52 | 13.31±0.64 | 47.68±0.70 | 29.58±0.58 | 68.44±0.15 |
| **Qwen2.5-3B** | 46.18±1.29 | 49.71±1.30 | 40.46±0.59 | 40.68±0.87 | 39.00±0.97 | 33.37±0.49 | 38.92±0.63 | 39.31±0.29 | 14.56±0.99 | 9.76±0.53 | 11.09±0.56 | 8.96±0.41 | 31.00±0.06 |
| **Qwen2.5-7B** | 56.86±0.83 | 61.12±1.02 | 52.18±0.72 | 50.21±0.88 | 50.01±0.61 | 44.21±0.44 | 51.89±1.03 | 52.74±1.22 | 29.87±0.90 | 9.21±0.50 | 21.93±0.69 | 13.40±0.67 | 41.13±0.44 |
| **Qwen2.5-14B** | 65.39±0.63 | 70.30±0.80 | 62.11±1.11 | 62.36±0.87 | 60.21±0.80 | 55.08±0.72 | 61.56±0.36 | 62.42±0.64 | 36.86±0.26 | 21.16±1.27 | 36.07±0.24 | 33.20±0.48 | 52.23±0.26 |
| **Qwen2.5-72B** | 75.04±1.02 | 79.69±0.39 | 76.10±0.58 | 75.62±0.20 | 73.01±0.98 | 70.51±0.24 | 75.13±0.46 | 76.00±0.68 | 49.29±0.69 | 39.97±0.48 | 38.49±0.88 | 37.93±0.49 | 63.90±0.13 |
| **QwQ-32B** | 82.33±0.43 | 84.98±0.55 | 83.71±0.49 | 83.55±0.39 | 75.60±0.48 | 72.91±0.85 | 83.80±0.87 | 84.81±0.86 | 62.36±1.09 | 32.08±1.21 | 30.49±0.88 | 32.43±1.34 | 67.42±0.17 |
| **Qwen3-1.7B** | 44.10±0.69 | 49.14±0.88 | 40.02±0.82 | 38.93±0.78 | 35.88±0.52 | 34.28±1.01 | 37.82±0.66 | 39.43±0.93 | 26.61±0.54 | 26.27±1.13 | 25.64±0.52 | 25.25±0.86 | 35.28±0.17 |
| **Qwen3-4B** | 59.54±1.08 | 64.45±0.72 | 54.27±1.07 | 53.20±0.84 | 53.01±0.63 | 45.10±0.72 | 54.52±0.48 | 55.99±1.02 | 22.91±0.82 | 9.96±0.69 | 16.98±1.06 | 9.69±1.06 | 41.64±0.39 |
| **Qwen3-4B-thinking** | 65.94±0.56 | 72.38±0.40 | 68.78±0.62 | 68.45±0.74 | 65.72±0.83 | 60.06±0.56 | 67.83±1.01 | 68.17±0.47 | 26.35±0.31 | 15.27±1.10 | 12.08±0.67 | 8.91±0.54 | 49.99±0.23 |
| **Qwen3-8B** | 67.41±0.62 | 71.72±0.41 | 52.30±0.57 | 62.66±0.67 | 60.60±0.72 | 56.42±0.63 | 61.60±1.07 | 61.48±1.03 | 21.02±1.23 | 8.83±0.39 | 21.98±0.84 | 10.43±0.84 | 46.37±0.32 |
| **Qwen3-8B-thinking** | 74.15±0.78 | 80.17±0.94 | 76.95±0.71 | 77.39±0.30 | 75.10±0.42 | 73.39±0.44 | 76.65±0.62 | 76.23±0.51 | 28.82±1.11 | 10.54±1.56 | 14.72±0.70 | 10.12±1.16 | 56.19±0.24 |
| **Qwen3-14B** | 73.24±0.97 | 76.73±0.56 | 68.52±0.44 | 70.12±0.78 | 66.57±0.77 | 61.19±0.94 | 70.15±0.81 | 69.79±0.39 | 41.54±0.75 | 11.23±0.61 | 16.67±0.85 | 14.56±0.48 | 53.36±0.11 |
| **Qwen3-14B-thinking** | 79.43±0.32 | 83.02±0.53 | 80.66±0.72 | 80.71±0.55 | 80.52±0.63 | 77.75±1.00 | 80.72±0.59 | 81.26±0.64 | 62.00±0.96 | 13.26±0.71 | 18.22±1.09 | 14.27±0.84 | 62.65±0.21 |
| **Baichuan-M2-32B** | 82.89±1.02 | 85.47±0.29 | 73.39±0.83 | 78.98±1.28 | 77.17±0.96 | 70.17±1.06 | 77.56±0.51 | 75.60±0.81 | 30.93±0.72 | 24.26±0.77 | 28.92±1.81 | 27.24±1.72 | 61.05±0.40 |
| **Bio-Medical-LLaMA-3-8B** | 47.79±0.76 | 77.50±0.20 | 55.54±0.61 | 53.01±0.42 | 42.12±0.14 | 41.79±0.57 | 54.24±0.61 | 54.02±0.41 | 39.67±0.84 | 37.60±0.91 | 36.14±0.73 | 34.05±0.69 | 47.79±0.08 |
| **MediPhi** | 28.63±0.65 | 52.63±1.18 | 43.17±1.13 | 42.56±0.95 | 24.82±1.51 | 27.57±0.72 | 40.06±1.10 | 40.80±0.84 | 19.14±1.12 | 22.09±0.76 | 18.37±1.33 | 23.41±0.46 | 31.94±0.38 |
| **MedGemma-4B** | 47.67±0.91 | 62.88±0.84 | 52.94±0.97 | 54.60±0.45 | 47.89±1.13 | 43.57±0.43 | 53.01±0.64 | 53.56±1.20 | 42.81±0.68 | 16.47±0.66 | 19.37±1.29 | 28.80±1.16 | 43.63±0.18 |
| **MedGemma-27B** | 82.06±0.70 | 87.26±0.67 | 83.19±0.71 | 83.30±0.48 | 80.55±1.30 | 79.23±0.85 | 83.52±0.59 | 83.82±0.82 | 76.62±1.16 | 18.30±1.40 | 46.73±1.06 | 58.74±0.94 | 71.94±0.30 |
| **MedReason-8B** | 44.68±1.07 | 58.51±0.49 | 14.33±0.90 | 18.24±1.28 | 44.29±1.32 | 36.12±0.92 | 22.33±0.72 | 36.42±0.63 | 10.60±0.47 | 7.87±0.54 | 10.92±0.37 | 9.54±0.72 | 26.15±0.18 |
| **HuatuoGPT-o1-7B** | 63.28±0.97 | 65.36±0.65 | 64.57±1.11 | 65.14±0.83 | 54.67±0.89 | 54.26±0.94 | 65.10±0.43 | 65.89±0.71 | 7.29±0.62 | 7.37±0.65 | 6.96±0.49 | 6.66±0.52 | 43.88±0.16 |
| **HuatuoGPT-o1-8B** | 57.22±0.80 | 68.88±0.81 | 65.78±0.74 | 65.22±1.07 | 49.87±1.16 | 48.97±1.27 | 64.12±1.15 | 64.88±1.88 | 50.09±1.07 | 7.71±0.34 | 7.68±0.46 | 7.21±0.77 | 46.47±0.19 |
| **HuatuoGPT-o1-70B** | 67.51±0.56 | 87.51±0.50 | 85.42±0.32 | 85.17±0.31 | 66.44±0.64 | 74.22±0.67 | 83.77±0.84 | 85.47±0.50 | 73.81±0.18 | 29.91±0.59 | 46.74±0.86 | 57.25±1.49 | 70.27±0.28 |
| **HuatuoGPT-o1-72B** | 82.03±0.55 | 82.25±0.96 | 85.72±0.71 | 80.61±0.44 | 79.75±0.95 | 74.86±0.69 | 85.15±0.41 | 84.08±0.52 | 64.56±0.51 | 42.23±0.53 | 43.30±1.23 | 21.89±1.18 | 68.87±0.08 |
| **OpenBioLLM-8B** | 13.86±0.88 | 25.36±0.99 | 16.39±1.07 | 25.34±0.84 | 10.62±0.78 | 9.77±0.82 | 29.18±0.63 | 30.68±1.22 | 11.25±0.70 | 7.73±0.79 | 9.36±1.12 | 11.44±0.37 | 16.75±0.31 |
| **OpenBioLLM-70B** | 20.79±0.62 | 72.24±0.78 | 60.61±1.43 | 34.50±1.46 | 20.17±0.80 | 19.20±0.28 | 46.91±1.10 | 66.77±1.14 | 26.94±1.18 | 10.43±0.63 | 12.07±0.82 | 25.75±1.37 | 34.70±0.18 |

**Supplementary Tab. 115:** Performance evaluation of 56 LLMs on MedQA.

| LLMs | Chinese | English | French | German | Japanese | Korean | Portuguese | Spanish | Swahili | Wolof | Yoruba | Zulu |
|---|---|---|---|---|---|---|---|---|---|---|---|---|
| **Proprietary LLMs** | | | | | | | | | | | | |
| **Claude-3.5-Haiku** | 66.85 | 77.06 | 73.06 | 73.29 | 67.95 | 68.58 | 71.96 | 73.45 | 52.40 | 34.25 | 39.83 | 37.08 |
| **Claude-4.0-Sonnet** | 89.71 | 92.07 | 88.69 | 89.87 | 89.95 | 90.02 | 91.12 | 91.28 | 82.01 | 59.23 | 70.07 | 75.33 |
| **Gemini-2.5-Flash** | 90.10 | 91.67 | 91.28 | 91.36 | 90.49 | 89.95 | 90.89 | 91.52 | 89.71 | 78.55 | 82.88 | 84.52 |
| **GPT-4o-mini** | 67.71 | 78.32 | 71.56 | 73.06 | 69.21 | 65.12 | 72.58 | 74.16 | 60.80 | 33.23 | 42.89 | 51.30 |
| **GPT-4o** | 86.49 | 89.32 | 88.53 | 88.85 | 85.07 | 85.00 | 89.24 | 89.16 | 82.56 | 34.25 | 49.80 | 69.29 |
| **GPT-4.1-nano** | 66.61 | 82.48 | 72.98 | 70.15 | 63.94 | 61.51 | 72.35 | 74.39 | 52.71 | 29.22 | 38.88 | 46.43 |
| **GPT-4.1-mini** | 85.31 | 91.36 | 87.04 | 88.14 | 85.39 | 83.35 | 88.06 | 88.85 | 76.98 | 25.14 | 60.02 | 68.34 |
| **GPT-4.1** | 88.61 | 90.26 | 90.26 | 89.08 | 90.57 | 86.10 | 91.20 | 89.00 | 87.90 | 51.30 | 68.26 | 75.18 |
| **GPT-5-nano** | 50.27 | 69.84 | 65.59 | 70.23 | 65.83 | 55.77 | 63.39 | 72.43 | 57.42 | 21.13 | 38.10 | 43.21 |
| **GPT-5-mini** | 85.78 | 91.67 | 87.43 | 87.51 | 87.59 | 84.13 | 88.37 | 89.63 | 80.36 | 34.80 | 61.67 | 73.21 |
| **GPT-5** | 90.81 | 92.69 | 92.22 | 90.81 | 91.12 | 83.82 | 92.30 | 92.46 | 88.92 | 53.89 | 73.84 | 75.73 |
| **o4-mini** | 93.09 | 94.03 | 93.01 | 93.40 | 93.24 | 91.67 | 94.03 | 92.77 | 91.36 | 37.63 | 78.87 | 85.55 |
| **Open-Weight LLMs** | | | | | | | | | | | | |
| **DeepSeek-V3** | 85.15 | 88.85 | 85.39 | 86.25 | 85.15 | 82.64 | 86.41 | 87.59 | 75.57 | 44.07 | 56.64 | 61.04 |
| **DeepSeek-R1** | 91.12 | 91.20 | 90.57 | 90.81 | 90.34 | 80.68 | 92.14 | 91.36 | 87.90 | 65.36 | 72.19 | 78.71 |
| **DeepSeek-R1-Qwen3-8B** | 72.11 | 80.13 | 68.03 | 70.93 | 67.71 | 67.71 | 72.11 | 68.66 | 33.07 | 10.05 | 12.02 | 12.25 |
| **Gemma-3-4B** | 43.60 | 52.79 | 46.66 | 45.64 | 41.16 | 40.14 | 45.56 | 46.58 | 37.47 | 8.72 | 16.73 | 14.53 |
| **Gemma-3-12B** | 62.29 | 70.86 | 65.83 | 65.51 | 61.74 | 58.29 | 63.39 | 65.99 | 55.54 | 9.11 | 35.74 | 41.95 |
| **Gemma-3-27B** | 70.23 | 75.26 | 71.48 | 73.13 | 68.66 | 68.89 | 72.03 | 72.51 | 65.91 | 18.38 | 46.03 | 53.10 |
| **gpt-oss-20B** | 83.03 | 79.42 | 85.39 | 84.60 | 83.58 | 80.28 | 84.13 | 84.29 | 68.97 | 29.54 | 60.64 | 65.67 |
| **gpt-oss-120B** | 88.61 | 90.89 | 91.04 | 90.10 | 89.24 | 86.80 | 90.42 | 91.36 | 80.68 | 55.22 | 71.56 | 77.38 |
| **LLaMA-3.1-8B** | 45.33 | 67.95 | 48.47 | 49.10 | 39.28 | 32.05 | 47.92 | 49.80 | 36.92 | 27.02 | 16.97 | 11.63 |
| **LLaMA-3.1-70B** | 70.93 | 84.76 | 76.67 | 77.38 | 68.58 | 62.77 | 77.45 | 78.71 | 61.35 | 42.89 | 41.95 | 46.98 |
| **LLaMA-3.2-3B** | 39.12 | 57.66 | 39.91 | 35.59 | 33.39 | 26.47 | 38.26 | 34.56 | 29.22 | 15.48 | 16.97 | 14.85 |
| **LLaMA-3.3-70B** | 61.04 | 85.00 | 81.23 | 81.62 | 59.39 | 70.31 | 80.75 | 81.30 | 67.56 | 40.61 | 42.50 | 47.84 |
| **LLaMA-4-Scout** | 79.03 | 85.62 | 81.15 | 82.64 | 79.50 | 77.61 | 82.33 | 82.33 | 73.84 | 46.82 | 46.43 | 64.02 |
| **LLaMA-4-Maverick** | 85.55 | 89.95 | 88.14 | 87.04 | 87.20 | 85.23 | 88.85 | 88.14 | 83.97 | 56.48 | 69.21 | 75.41 |
| **Mistral-7B-v0.3** | 22.23 | 26.79 | 13.59 | 30.16 | 21.13 | 18.15 | 22.39 | 27.97 | 20.66 | 16.42 | 9.66 | 10.21 |
| **Mistral-Small-3.1-24B** | 64.10 | 75.65 | 64.41 | 67.16 | 58.92 | 54.36 | 68.34 | 69.99 | 32.76 | 12.41 | 14.93 | 18.54 |
| **Phi-4-mini** | 33.39 | 54.44 | 37.47 | 37.16 | 31.81 | 24.27 | 35.51 | 38.49 | 26.63 | 17.75 | 19.56 | 16.65 |
| **Phi-4-mini-Reasoning** | 29.77 | 65.59 | 48.55 | 48.08 | 24.35 | 14.77 | 41.63 | 34.88 | 29.46 | 24.90 | 22.94 | 22.62 |
| **Phi-4** | 57.34 | 79.97 | 69.44 | 71.72 | 58.60 | 53.81 | 70.70 | 69.68 | 47.60 | 25.92 | 37.23 | 30.01 |
| **Phi-4-Reasoning** | 80.99 | 85.07 | 84.05 | 85.70 | 82.48 | 80.28 | 76.28 | 83.66 | 72.03 | 12.96 | 46.98 | 29.14 |
| **Qwen2.5-3B** | 47.45 | 48.39 | 40.69 | 39.51 | 40.30 | 33.31 | 38.49 | 39.43 | 15.87 | 9.11 | 11.15 | 8.80 |
| **Qwen2.5-7B** | 57.82 | 60.80 | 53.02 | 50.90 | 50.51 | 44.93 | 52.00 | 53.89 | 30.24 | 8.80 | 22.55 | 13.75 |
| **Qwen2.5-14B** | 65.12 | 70.93 | 62.45 | 61.59 | 60.72 | 56.17 | 61.74 | 62.69 | 37.08 | 21.05 | 35.90 | 33.86 |
| **Qwen2.5-72B** | 73.84 | 79.58 | 75.96 | 75.81 | 74.00 | 70.46 | 75.88 | 76.83 | 48.39 | 39.83 | 39.75 | 37.47 |
| **QwQ-32B** | 81.70 | 84.92 | 83.82 | 83.35 | 75.26 | 74.08 | 83.58 | 85.23 | 62.14 | 32.13 | 30.24 | 34.41 |
| **Qwen3-1.7B** | 43.28 | 49.41 | 40.77 | 38.41 | 35.59 | 35.43 | 38.49 | 38.02 | 26.32 | 26.63 | 26.08 | 25.45 |
| **Qwen3-4B** | 59.54 | 64.73 | 54.75 | 52.24 | 52.24 | 44.07 | 54.12 | 56.95 | 22.86 | 10.45 | 16.03 | 8.64 |
| **Qwen3-4B-thinking** | 66.14 | 72.19 | 69.05 | 68.26 | 65.91 | 59.78 | 67.32 | 68.03 | 26.79 | 14.61 | 12.57 | 8.25 |
| **Qwen3-8B** | 67.16 | 72.03 | 52.71 | 62.22 | 61.35 | 57.11 | 61.27 | 62.92 | 22.15 | 9.27 | 21.45 | 11.55 |
| **Qwen3-8B-thinking** | 74.47 | 79.03 | 77.38 | 77.77 | 74.94 | 74.00 | 76.98 | 75.88 | 28.52 | 9.90 | 13.90 | 11.08 |
| **Qwen3-14B** | 73.29 | 77.22 | 68.74 | 69.91 | 66.14 | 60.17 | 71.01 | 69.76 | 41.16 | 12.10 | 17.99 | 13.75 |
| **Qwen3-14B-thinking** | 79.58 | 83.03 | 80.75 | 81.30 | 79.65 | 79.03 | 80.75 | 81.70 | 61.43 | 14.06 | 17.67 | 13.67 |
| **Baichuan-M2-32B** | 82.40 | 85.15 | 74.23 | 80.36 | 76.98 | 71.41 | 77.61 | 76.43 | 30.95 | 25.06 | 29.54 | 25.45 |
| **Bio-Medical-LLaMA-3-8B** | 48.08 | 77.53 | 54.91 | 52.95 | 42.11 | 41.87 | 53.65 | 53.89 | 40.85 | 36.92 | 37.08 | 34.25 |
| **MediPhi** | 27.89 | 52.71 | 43.83 | 43.28 | 26.39 | 26.71 | 41.71 | 40.14 | 19.95 | 21.45 | 17.75 | 23.17 |
| **MedGemma-4B** | 46.27 | 62.53 | 52.47 | 54.60 | 48.47 | 43.68 | 52.79 | 54.67 | 41.79 | 17.44 | 20.35 | 27.97 |
| **MedGemma-27B** | 82.33 | 88.22 | 82.17 | 82.88 | 79.42 | 79.89 | 83.27 | 84.76 | 77.61 | 19.32 | 47.76 | 59.47 |
| **MedReason-8B** | 44.62 | 58.37 | 15.48 | 19.80 | 43.21 | 35.98 | 22.47 | 37.31 | 10.29 | 7.78 | 11.39 | 9.82 |
| **HuatuoGPT-o1-7B** | 64.49 | 65.12 | 66.22 | 65.75 | 54.52 | 53.02 | 64.49 | 65.99 | 6.60 | 7.93 | 6.99 | 7.46 |
| **HuatuoGPT-o1-8B** | 56.87 | 67.95 | 65.51 | 66.22 | 49.96 | 46.90 | 63.71 | 65.59 | 50.35 | 7.31 | 7.86 | 7.86 |
| **HuatuoGPT-o1-70B** | 67.24 | 87.75 | 85.00 | 85.31 | 66.93 | 75.18 | 83.82 | 85.23 | 73.84 | 29.69 | 47.45 | 58.76 |
| **HuatuoGPT-o1-72B** | 82.09 | 82.80 | 86.88 | 80.75 | 80.13 | 75.41 | 85.31 | 83.66 | 64.81 | 42.81 | 42.97 | 20.42 |
| **OpenBioLLM-8B** | 13.98 | 24.12 | 16.03 | 24.67 | 11.94 | 9.74 | 29.22 | 29.14 | 11.31 | 8.01 | 10.29 | 11.86 |
| **OpenBioLLM-70B** | 20.27 | 72.82 | 60.57 | 36.76 | 20.82 | 19.17 | 47.68 | 66.85 | 26.55 | 11.00 | 11.08 | 24.19 |

**Supplementary Tab. 116:** Zero-Shot performance evaluation of 56 LLMs on MedQA (Run 1).

| LLMs | Chinese | English | French | German | Japanese | Korean | Portuguese | Spanish | Swahili | Wolof | Yoruba | Zulu |
|---|---|---|---|---|---|---|---|---|---|---|---|---|
| **Proprietary LLMs** | | | | | | | | | | | | |
| **Claude-3.5-Haiku** | 66.85 | 77.06 | 73.06 | 73.29 | 68.11 | 68.74 | 72.11 | 73.53 | 52.63 | 34.41 | 39.83 | 37.08 |
| **Claude-4.0-Sonnet** | 89.95 | 92.38 | 89.95 | 89.87 | 89.71 | 89.32 | 91.12 | 90.10 | 81.46 | 58.92 | 70.78 | 75.33 |
| **Gemini-2.5-Flash** | 89.55 | 90.81 | 90.65 | 91.44 | 90.18 | 89.95 | 91.12 | 91.44 | 89.63 | 80.75 | 82.64 | 84.21 |
| **GPT-4o-mini** | 68.34 | 78.08 | 72.27 | 74.00 | 69.05 | 65.83 | 73.21 | 73.37 | 60.02 | 32.76 | 43.60 | 51.53 |
| **GPT-4o** | 86.10 | 89.16 | 88.61 | 87.82 | 84.84 | 85.39 | 88.85 | 89.08 | 82.17 | 31.66 | 49.80 | 69.84 |
| **GPT-4.1-nano** | 66.61 | 82.01 | 73.84 | 72.51 | 63.63 | 60.09 | 72.58 | 73.45 | 52.63 | 27.73 | 37.31 | 44.85 |
| **GPT-4.1-mini** | 86.80 | 91.20 | 87.20 | 87.59 | 85.70 | 81.23 | 87.98 | 87.82 | 77.93 | 27.10 | 58.60 | 67.32 |
| **GPT-4.1** | 89.24 | 90.26 | 89.87 | 89.08 | 90.42 | 86.02 | 91.52 | 90.42 | 88.85 | 52.55 | 68.26 | 75.41 |
| **GPT-5-nano** | 52.95 | 67.64 | 64.65 | 69.44 | 64.73 | 52.95 | 61.59 | 70.93 | 57.34 | 20.35 | 37.78 | 45.40 |
| **GPT-5-mini** | 86.33 | 91.04 | 87.82 | 88.77 | 88.14 | 84.92 | 88.69 | 88.69 | 80.91 | 37.31 | 62.53 | 71.56 |
| **GPT-5** | 91.67 | 92.62 | 92.93 | 91.75 | 90.73 | 84.21 | 92.69 | 91.52 | 89.08 | 53.26 | 75.18 | 75.18 |
| **o4-mini** | 93.79 | 94.58 | 93.32 | 93.09 | 93.56 | 92.38 | 93.87 | 93.95 | 91.12 | 41.48 | 80.68 | 86.49 |
| **Open-Weight LLMs** | | | | | | | | | | | | |
| **DeepSeek-V3** | 85.70 | 89.40 | 86.49 | 86.25 | 85.31 | 82.72 | 86.33 | 86.96 | 74.47 | 45.01 | 56.25 | 61.12 |
| **DeepSeek-R1** | 91.20 | 91.59 | 90.26 | 90.49 | 89.95 | 81.07 | 92.07 | 91.83 | 88.06 | 66.38 | 72.19 | 78.48 |
| **DeepSeek-R1-Qwen3-8B** | 73.37 | 79.73 | 68.42 | 70.78 | 66.54 | 67.24 | 71.48 | 69.05 | 31.97 | 11.23 | 10.45 | 13.04 |
| **Gemma-3-4B** | 44.70 | 51.92 | 45.40 | 43.75 | 42.34 | 40.61 | 46.50 | 46.82 | 37.23 | 10.13 | 15.48 | 14.45 |
| **Gemma-3-12B** | 61.27 | 69.76 | 65.20 | 64.41 | 60.33 | 60.02 | 64.34 | 65.36 | 55.38 | 10.53 | 36.92 | 45.33 |
| **Gemma-3-27B** | 69.99 | 76.36 | 71.48 | 71.88 | 70.31 | 67.87 | 71.48 | 72.35 | 64.41 | 18.22 | 44.93 | 52.87 |
| **gpt-oss-20B** | 82.72 | 80.13 | 84.45 | 82.95 | 81.23 | 80.99 | 82.88 | 84.21 | 68.58 | 33.23 | 59.78 | 66.46 |
| **gpt-oss-120B** | 89.24 | 92.14 | 90.49 | 89.87 | 89.87 | 87.75 | 89.79 | 90.18 | 81.15 | 54.28 | 72.58 | 77.22 |
| **LLaMA-3.1-8B** | 44.93 | 67.40 | 49.10 | 49.49 | 38.73 | 33.23 | 50.51 | 48.70 | 37.39 | 25.92 | 17.67 | 13.28 |
| **LLaMA-3.1-70B** | 71.72 | 82.95 | 77.69 | 76.83 | 69.84 | 63.94 | 76.20 | 77.77 | 61.67 | 39.91 | 44.15 | 46.27 |
| **LLaMA-3.2-3B** | 38.49 | 57.66 | 41.08 | 35.27 | 34.41 | 28.52 | 38.18 | 36.84 | 30.24 | 14.93 | 18.54 | 15.95 |
| **LLaMA-3.3-70B** | 62.14 | 84.84 | 79.81 | 80.20 | 58.68 | 70.62 | 79.81 | 81.93 | 67.48 | 40.46 | 42.97 | 46.58 |
| **LLaMA-4-Scout** | 80.60 | 86.10 | 82.64 | 82.95 | 78.95 | 77.77 | 81.46 | 81.62 | 74.78 | 46.35 | 46.58 | 63.47 |
| **LLaMA-4-Maverick** | 85.55 | 90.57 | 88.92 | 88.22 | 85.39 | 85.23 | 89.08 | 88.53 | 84.68 | 57.74 | 69.05 | 76.12 |
| **Mistral-7B-v0.3** | 23.72 | 31.42 | 14.85 | 28.44 | 18.93 | 18.07 | 20.82 | 29.30 | 18.85 | 15.40 | 10.68 | 10.92 |
| **Mistral-Small-3.1-24B** | 63.39 | 74.23 | 67.48 | 68.03 | 59.94 | 54.83 | 67.09 | 68.42 | 30.87 | 11.70 | 18.62 | 18.54 |
| **Phi-4-mini** | 33.39 | 53.50 | 35.66 | 37.31 | 31.50 | 25.29 | 35.35 | 36.53 | 27.57 | 17.36 | 18.62 | 16.03 |
| **Phi-4-mini-Reasoning** | 29.07 | 65.59 | 48.47 | 48.39 | 22.78 | 13.98 | 40.93 | 35.19 | 27.26 | 22.39 | 21.13 | 21.13 |
| **Phi-4** | 59.47 | 79.89 | 69.05 | 71.72 | 58.76 | 53.81 | 70.93 | 71.72 | 45.25 | 24.90 | 37.78 | 32.91 |
| **Phi-4-Reasoning** | 81.15 | 85.39 | 83.97 | 85.55 | 84.37 | 81.07 | 74.94 | 84.13 | 72.11 | 12.96 | 48.86 | 29.77 |
| **Qwen2.5-3B** | 46.98 | 49.80 | 40.30 | 40.69 | 38.88 | 34.09 | 38.02 | 39.51 | 13.90 | 9.90 | 11.47 | 9.43 |
| **Qwen2.5-7B** | 56.79 | 60.49 | 51.45 | 50.67 | 49.73 | 43.75 | 50.90 | 53.02 | 28.44 | 8.88 | 21.60 | 13.59 |
| **Qwen2.5-14B** | 64.81 | 70.70 | 60.96 | 61.82 | 60.57 | 54.28 | 61.35 | 62.14 | 36.76 | 19.64 | 36.29 | 32.84 |
| **Qwen2.5-72B** | 76.59 | 79.97 | 75.65 | 75.41 | 73.53 | 70.23 | 74.63 | 76.28 | 49.80 | 40.14 | 38.73 | 37.63 |
| **QwQ-32B** | 82.17 | 84.76 | 83.90 | 83.58 | 75.26 | 72.11 | 84.60 | 84.76 | 61.98 | 30.64 | 30.16 | 32.13 |
| **Qwen3-1.7B** | 43.52 | 47.68 | 40.77 | 38.49 | 36.29 | 34.88 | 38.57 | 38.96 | 26.71 | 25.61 | 25.77 | 25.45 |
| **Qwen3-4B** | 60.72 | 64.49 | 54.20 | 53.34 | 53.73 | 45.09 | 53.97 | 56.72 | 21.60 | 9.66 | 16.10 | 8.96 |
| **Qwen3-4B-thinking** | 66.69 | 72.03 | 69.36 | 68.97 | 64.26 | 60.72 | 69.60 | 68.81 | 25.92 | 15.32 | 12.88 | 9.58 |
| **Qwen3-8B** | 68.42 | 71.96 | 52.95 | 63.79 | 60.72 | 55.70 | 60.57 | 60.80 | 20.90 | 8.25 | 22.78 | 10.60 |
| **Qwen3-8B-thinking** | 75.33 | 81.15 | 76.98 | 77.53 | 74.63 | 73.21 | 77.45 | 75.88 | 29.30 | 9.82 | 14.38 | 9.35 |
| **Qwen3-14B** | 72.90 | 76.90 | 68.97 | 69.21 | 67.71 | 61.12 | 68.89 | 69.13 | 41.40 | 10.76 | 16.03 | 15.00 |
| **Qwen3-14B-thinking** | 79.10 | 82.64 | 79.97 | 80.83 | 80.13 | 77.22 | 81.54 | 80.44 | 60.72 | 13.67 | 17.05 | 14.53 |
| **Baichuan-M2-32B** | 83.58 | 85.47 | 72.90 | 78.95 | 78.48 | 68.74 | 78.32 | 74.86 | 32.13 | 23.25 | 30.71 | 26.71 |
| **Bio-Medical-LLaMA-3-8B** | 48.63 | 77.45 | 55.38 | 53.57 | 42.34 | 41.63 | 53.81 | 53.50 | 39.59 | 38.10 | 36.37 | 33.94 |
| **MediPhi** | 29.22 | 52.40 | 43.44 | 41.32 | 24.74 | 28.36 | 39.36 | 40.22 | 17.60 | 22.94 | 20.58 | 23.72 |
| **MedGemma-4B** | 48.15 | 61.82 | 53.97 | 54.28 | 46.11 | 42.89 | 54.05 | 53.42 | 43.21 | 15.63 | 18.54 | 28.52 |
| **MedGemma-27B** | 81.78 | 86.72 | 83.74 | 83.27 | 82.01 | 78.40 | 83.97 | 84.13 | 74.71 | 18.15 | 47.53 | 59.94 |
| **MedReason-8B** | 43.28 | 58.37 | 14.14 | 18.93 | 45.80 | 37.47 | 21.84 | 35.90 | 10.37 | 7.78 | 11.23 | 10.37 |
| **HuatuoGPT-o1-7B** | 62.22 | 66.46 | 64.18 | 65.99 | 54.75 | 54.36 | 65.44 | 66.46 | 6.83 | 8.01 | 6.44 | 6.68 |
| **HuatuoGPT-o1-8B** | 56.48 | 69.76 | 65.36 | 66.22 | 49.96 | 49.96 | 65.59 | 64.10 | 51.30 | 7.46 | 7.07 | 7.86 |
| **HuatuoGPT-o1-70B** | 66.93 | 88.06 | 85.31 | 84.84 | 65.75 | 74.16 | 84.13 | 84.76 | 74.00 | 29.77 | 45.88 | 57.34 |
| **HuatuoGPT-o1-72B** | 82.01 | 81.38 | 85.15 | 80.20 | 80.52 | 75.73 | 85.78 | 84.52 | 65.20 | 42.42 | 42.50 | 21.13 |
| **OpenBioLLM-8B** | 13.83 | 26.71 | 17.60 | 24.59 | 10.45 | 11.15 | 29.30 | 31.42 | 12.33 | 7.93 | 9.90 | 11.31 |
| **OpenBioLLM-70B** | 20.03 | 72.03 | 60.72 | 33.23 | 20.74 | 18.77 | 48.08 | 66.22 | 28.04 | 10.45 | 11.78 | 27.57 |

**Supplementary Tab. 117:** Zero-Shot performance evaluation of 56 LLMs on MedQA (Run 2).

| LLMs | Chinese | English | French | German | Japanese | Korean | Portuguese | Spanish | Swahili | Wolof | Yoruba | Zulu |
|---|---|---|---|---|---|---|---|---|---|---|---|---|
| **Proprietary LLMs** | | | | | | | | | | | | |
| **Claude-3.5-Haiku** | 66.85 | 77.06 | 73.06 | 73.29 | 68.03 | 68.81 | 72.11 | 73.45 | 52.55 | 34.33 | 39.83 | 37.00 |
| **Claude-4.0-Sonnet** | 88.92 | 91.75 | 89.40 | 90.34 | 90.26 | 90.42 | 91.12 | 90.97 | 82.01 | 59.54 | 71.41 | 74.39 |
| **Gemini-2.5-Flash** | 90.18 | 91.04 | 90.89 | 91.52 | 89.95 | 89.87 | 91.52 | 90.57 | 89.32 | 79.18 | 83.66 | 84.92 |
| **GPT-4o-mini** | 67.56 | 77.38 | 72.11 | 73.53 | 67.24 | 65.36 | 73.45 | 72.90 | 60.49 | 32.05 | 44.30 | 49.96 |
| **GPT-4o** | 85.47 | 89.32 | 89.16 | 88.61 | 87.20 | 85.07 | 89.00 | 88.92 | 83.03 | 32.36 | 48.70 | 69.21 |
| **GPT-4.1-nano** | 67.87 | 80.75 | 72.27 | 71.17 | 64.41 | 61.12 | 71.25 | 74.16 | 52.63 | 27.10 | 37.23 | 46.43 |
| **GPT-4.1-mini** | 84.52 | 91.04 | 85.55 | 88.06 | 85.39 | 82.25 | 87.20 | 87.90 | 79.18 | 25.77 | 57.89 | 68.50 |
| **GPT-4.1** | 89.24 | 89.71 | 89.95 | 89.55 | 89.63 | 87.43 | 91.91 | 89.55 | 88.77 | 51.85 | 67.79 | 76.67 |
| **GPT-5-nano** | 49.88 | 70.93 | 66.61 | 69.13 | 65.67 | 55.38 | 62.77 | 70.31 | 57.42 | 17.99 | 39.36 | 43.21 |
| **GPT-5-mini** | 85.23 | 90.73 | 88.45 | 87.75 | 87.98 | 84.68 | 89.40 | 88.45 | 82.01 | 35.27 | 61.35 | 69.99 |
| **GPT-5** | 91.12 | 93.24 | 91.67 | 90.81 | 91.04 | 84.45 | 92.07 | 92.07 | 89.16 | 54.05 | 75.33 | 74.39 |
| **o4-mini** | 93.32 | 94.58 | 93.32 | 93.56 | 93.87 | 92.62 | 94.27 | 93.87 | 91.28 | 41.16 | 81.30 | 85.70 |
| **Open-Weight LLMs** | | | | | | | | | | | | |
| **DeepSeek-V3** | 86.96 | 89.63 | 86.80 | 86.80 | 85.47 | 82.33 | 87.04 | 87.98 | 75.02 | 44.30 | 54.44 | 62.22 |
| **DeepSeek-R1** | 90.81 | 91.44 | 90.81 | 90.97 | 90.18 | 80.99 | 91.59 | 91.44 | 87.04 | 65.75 | 73.37 | 80.20 |
| **DeepSeek-R1-Qwen3-8B** | 73.37 | 79.18 | 68.81 | 70.23 | 66.85 | 69.99 | 72.35 | 69.60 | 35.19 | 11.31 | 11.39 | 12.49 |
| **Gemma-3-4B** | 43.91 | 51.06 | 44.78 | 44.23 | 42.73 | 41.79 | 46.35 | 47.37 | 38.49 | 10.53 | 16.89 | 14.45 |
| **Gemma-3-12B** | 60.57 | 69.13 | 64.26 | 63.94 | 61.12 | 60.57 | 63.71 | 66.93 | 55.62 | 9.82 | 36.68 | 44.46 |
| **Gemma-3-27B** | 69.29 | 75.81 | 72.58 | 73.06 | 69.29 | 67.56 | 73.21 | 72.74 | 64.57 | 20.03 | 46.50 | 52.71 |
| **gpt-oss-20B** | 82.80 | 80.28 | 83.82 | 84.13 | 82.72 | 79.97 | 83.90 | 83.82 | 66.46 | 30.71 | 61.27 | 68.03 |
| **gpt-oss-120B** | 90.65 | 91.44 | 90.26 | 90.65 | 90.42 | 88.37 | 90.97 | 90.65 | 81.38 | 53.89 | 72.51 | 77.53 |
| **LLaMA-3.1-8B** | 45.09 | 66.30 | 50.27 | 46.82 | 38.41 | 33.86 | 49.73 | 50.43 | 35.19 | 24.90 | 17.75 | 12.41 |
| **LLaMA-3.1-70B** | 72.11 | 82.72 | 76.51 | 76.83 | 69.60 | 62.22 | 75.10 | 77.93 | 62.37 | 39.75 | 43.44 | 47.21 |
| **LLaMA-3.2-3B** | 40.06 | 55.77 | 40.61 | 34.41 | 34.17 | 26.79 | 38.81 | 34.80 | 31.34 | 16.89 | 17.75 | 15.32 |
| **LLaMA-3.3-70B** | 60.80 | 85.07 | 80.75 | 81.30 | 58.76 | 71.25 | 80.83 | 80.83 | 68.11 | 42.18 | 43.28 | 47.92 |
| **LLaMA-4-Scout** | 78.95 | 85.55 | 82.17 | 82.25 | 77.53 | 78.79 | 82.09 | 82.95 | 74.94 | 47.84 | 46.66 | 64.18 |
| **LLaMA-4-Maverick** | 86.72 | 90.57 | 88.77 | 89.08 | 86.02 | 86.10 | 89.63 | 88.85 | 84.52 | 56.95 | 68.74 | 76.43 |
| **Mistral-7B-v0.3** | 22.15 | 29.14 | 15.87 | 28.67 | 20.58 | 19.09 | 21.13 | 27.26 | 20.50 | 16.73 | 10.53 | 9.90 |
| **Mistral-Small-3.1-24B** | 63.63 | 76.98 | 63.79 | 69.68 | 59.70 | 54.99 | 68.89 | 68.74 | 32.36 | 11.47 | 14.93 | 20.58 |
| **Phi-4-mini** | 31.26 | 54.83 | 36.68 | 38.57 | 32.99 | 24.12 | 37.55 | 37.63 | 26.00 | 17.67 | 18.22 | 16.81 |
| **Phi-4-mini-Reasoning** | 30.40 | 64.41 | 47.13 | 47.53 | 23.02 | 13.51 | 38.88 | 34.72 | 26.39 | 22.55 | 21.37 | 20.66 |
| **Phi-4** | 60.02 | 79.58 | 71.25 | 70.38 | 61.90 | 53.42 | 71.17 | 71.33 | 47.60 | 25.06 | 36.53 | 30.48 |
| **Phi-4-Reasoning** | 80.83 | 86.25 | 84.29 | 83.58 | 83.03 | 80.68 | 73.92 | 84.52 | 73.06 | 13.59 | 47.60 | 30.24 |
| **Qwen2.5-3B** | 44.23 | 48.39 | 41.32 | 41.95 | 39.36 | 33.15 | 39.36 | 39.04 | 15.08 | 9.35 | 10.53 | 9.35 |
| **Qwen2.5-7B** | 56.40 | 62.06 | 52.87 | 50.59 | 50.75 | 44.07 | 50.90 | 52.55 | 29.69 | 9.19 | 21.21 | 12.25 |
| **Qwen2.5-14B** | 66.14 | 71.01 | 61.67 | 63.79 | 59.47 | 55.15 | 61.51 | 61.43 | 37.00 | 23.17 | 35.74 | 32.99 |
| **Qwen2.5-72B** | 75.26 | 79.42 | 76.67 | 75.81 | 73.06 | 70.38 | 75.02 | 75.26 | 48.94 | 39.43 | 37.31 | 38.73 |
| **QwQ-32B** | 82.40 | 85.94 | 83.11 | 84.05 | 75.49 | 72.27 | 84.68 | 86.02 | 61.51 | 33.62 | 29.46 | 31.19 |
| **Qwen3-1.7B** | 44.54 | 50.04 | 39.28 | 38.33 | 35.74 | 34.56 | 37.16 | 39.98 | 25.84 | 25.92 | 25.14 | 24.27 |
| **Qwen3-4B** | 60.25 | 65.44 | 55.62 | 54.52 | 52.55 | 45.95 | 55.15 | 56.40 | 23.33 | 10.92 | 17.52 | 9.74 |
| **Qwen3-4B-thinking** | 65.28 | 72.98 | 68.42 | 69.05 | 65.99 | 59.39 | 67.09 | 67.71 | 26.32 | 16.89 | 11.86 | 9.11 |
| **Qwen3-8B** | 67.48 | 71.01 | 51.61 | 62.45 | 61.12 | 55.85 | 63.24 | 61.90 | 20.74 | 9.11 | 23.02 | 10.29 |
| **Qwen3-8B-thinking** | 73.92 | 81.15 | 77.69 | 76.98 | 75.73 | 73.61 | 76.20 | 75.81 | 30.40 | 11.39 | 14.69 | 9.03 |
| **Qwen3-14B** | 72.03 | 75.96 | 68.26 | 70.93 | 66.85 | 62.37 | 70.15 | 70.07 | 41.48 | 11.15 | 16.97 | 14.53 |
| **Qwen3-14B-thinking** | 79.10 | 83.27 | 80.28 | 80.91 | 81.23 | 77.06 | 80.60 | 81.30 | 61.98 | 13.28 | 18.46 | 13.83 |
| **Baichuan-M2-32B** | 81.78 | 85.23 | 72.98 | 76.90 | 77.14 | 69.44 | 76.98 | 74.71 | 30.79 | 23.72 | 29.62 | 26.16 |
| **Bio-Medical-LLaMA-3-8B** | 48.23 | 77.53 | 56.32 | 52.95 | 41.95 | 41.24 | 54.05 | 53.81 | 38.57 | 38.96 | 35.43 | 35.04 |
| **MediPhi** | 29.38 | 51.85 | 43.75 | 43.60 | 26.24 | 27.34 | 40.22 | 41.87 | 20.11 | 22.70 | 18.62 | 24.04 |
| **MedGemma-4B** | 48.55 | 63.16 | 51.53 | 54.36 | 49.02 | 43.52 | 52.32 | 53.26 | 43.60 | 16.58 | 17.52 | 30.40 |
| **MedGemma-27B** | 81.07 | 86.57 | 82.72 | 82.88 | 81.85 | 79.58 | 82.64 | 82.56 | 76.98 | 17.91 | 46.43 | 58.52 |
| **MedReason-8B** | 44.46 | 58.68 | 13.28 | 18.38 | 43.13 | 35.11 | 23.33 | 36.53 | 10.68 | 8.33 | 10.76 | 8.48 |
| **HuatuoGPT-o1-7B** | 64.10 | 64.81 | 63.47 | 63.86 | 54.75 | 55.46 | 64.96 | 66.69 | 8.17 | 7.07 | 7.62 | 6.28 |
| **HuatuoGPT-o1-8B** | 57.42 | 69.68 | 66.06 | 64.26 | 48.86 | 49.80 | 62.92 | 62.92 | 50.82 | 8.09 | 8.33 | 5.97 |
| **HuatuoGPT-o1-70B** | 68.11 | 87.59 | 85.70 | 85.62 | 65.83 | 74.16 | 84.52 | 85.86 | 73.76 | 29.46 | 46.98 | 58.44 |
| **HuatuoGPT-o1-72B** | 82.88 | 82.33 | 85.15 | 80.52 | 79.42 | 74.71 | 84.76 | 83.66 | 64.26 | 42.42 | 41.95 | 23.49 |
| **OpenBioLLM-8B** | 12.96 | 25.06 | 17.44 | 24.98 | 10.13 | 9.19 | 28.91 | 32.05 | 10.45 | 6.60 | 10.29 | 11.70 |
| **OpenBioLLM-70B** | 21.45 | 71.01 | 59.39 | 33.46 | 20.58 | 19.40 | 46.19 | 68.58 | 27.57 | 10.37 | 12.10 | 26.63 |

**Supplementary Tab. 118:** Zero-Shot performance evaluation of 56 LLMs on MedQA (Run 3).

| LLMs | Chinese | English | French | German | Japanese | Korean | Portuguese | Spanish | Swahili | Wolof | Yoruba | Zulu |
|---|---|---|---|---|---|---|---|---|---|---|---|---|
| **Proprietary LLMs** | | | | | | | | | | | | |
| **Claude-3.5-Haiku** | 66.93 | 77.06 | 72.98 | 73.29 | 68.03 | 68.81 | 72.19 | 73.45 | 52.55 | 34.41 | 39.83 | 37.08 |
| **Claude-4.0-Sonnet** | 89.79 | 91.75 | 90.18 | 90.18 | 89.63 | 89.79 | 91.36 | 90.57 | 81.78 | 59.07 | 70.54 | 74.16 |
| **Gemini-2.5-Flash** | 90.26 | 91.28 | 90.65 | 91.20 | 89.87 | 90.42 | 90.89 | 91.04 | 89.47 | 78.95 | 83.03 | 85.07 |
| **GPT-4o-mini** | 67.32 | 77.30 | 70.78 | 72.98 | 69.13 | 64.96 | 72.27 | 73.92 | 61.43 | 32.05 | 43.91 | 49.65 |
| **GPT-4o** | 86.10 | 88.77 | 87.98 | 88.45 | 85.07 | 84.52 | 89.40 | 88.69 | 83.35 | 31.26 | 50.20 | 68.50 |
| **GPT-4.1-nano** | 65.75 | 80.68 | 71.41 | 71.41 | 61.19 | 59.39 | 74.08 | 74.00 | 53.81 | 27.97 | 39.75 | 45.01 |
| **GPT-4.1-mini** | 85.94 | 90.65 | 85.47 | 87.82 | 85.55 | 82.72 | 87.98 | 87.90 | 78.55 | 24.59 | 57.11 | 66.38 |
| **GPT-4.1** | 89.55 | 89.87 | 89.95 | 89.16 | 89.63 | 87.51 | 91.44 | 89.08 | 88.85 | 51.45 | 68.26 | 75.65 |
| **GPT-5-nano** | 51.85 | 68.42 | 65.91 | 69.36 | 64.41 | 56.01 | 63.47 | 72.82 | 56.79 | 19.48 | 38.81 | 43.91 |
| **GPT-5-mini** | 87.12 | 91.44 | 89.79 | 88.06 | 87.20 | 84.45 | 89.87 | 89.00 | 80.52 | 35.90 | 60.33 | 71.64 |
| **GPT-5** | 91.44 | 93.17 | 91.44 | 90.89 | 91.12 | 84.68 | 93.40 | 91.52 | 89.08 | 55.46 | 75.02 | 75.41 |
| **o4-mini** | 92.85 | 94.74 | 94.11 | 93.40 | 93.24 | 92.07 | 93.64 | 92.69 | 91.20 | 39.51 | 80.60 | 85.15 |
| **Open-Weight LLMs** | | | | | | | | | | | | |
| **DeepSeek-V3** | 86.88 | 88.61 | 85.00 | 87.04 | 86.17 | 83.74 | 86.10 | 86.72 | 74.86 | 44.15 | 57.34 | 61.67 |
| **DeepSeek-R1** | 90.97 | 91.67 | 91.36 | 91.83 | 91.12 | 81.62 | 91.04 | 91.20 | 87.51 | 67.16 | 72.43 | 78.32 |
| **DeepSeek-R1-Qwen3-8B** | 72.98 | 79.65 | 70.46 | 70.46 | 66.22 | 69.44 | 72.11 | 69.84 | 32.52 | 11.08 | 10.92 | 12.41 |
| **Gemma-3-4B** | 43.44 | 52.00 | 45.80 | 45.88 | 41.71 | 39.20 | 46.50 | 46.66 | 37.00 | 8.96 | 16.65 | 15.24 |
| **Gemma-3-12B** | 62.53 | 69.44 | 65.20 | 65.75 | 60.33 | 60.57 | 64.81 | 66.38 | 57.74 | 10.60 | 36.53 | 42.34 |
| **Gemma-3-27B** | 69.13 | 77.14 | 72.82 | 72.82 | 69.99 | 68.81 | 72.43 | 72.35 | 66.22 | 18.85 | 47.29 | 54.05 |
| **gpt-oss-20B** | 82.95 | 81.07 | 83.27 | 83.42 | 83.19 | 80.52 | 84.60 | 83.90 | 68.42 | 30.48 | 62.06 | 66.85 |
| **gpt-oss-120B** | 89.08 | 91.83 | 89.79 | 89.95 | 89.08 | 87.59 | 90.49 | 90.89 | 81.85 | 55.07 | 71.33 | 78.16 |
| **LLaMA-3.1-8B** | 44.62 | 66.85 | 48.00 | 49.25 | 38.49 | 34.80 | 49.80 | 49.18 | 35.74 | 25.53 | 16.42 | 13.35 |
| **LLaMA-3.1-70B** | 70.70 | 82.40 | 75.49 | 76.75 | 67.87 | 61.82 | 76.90 | 78.63 | 61.90 | 38.41 | 43.68 | 48.00 |
| **LLaMA-3.2-3B** | 40.22 | 56.64 | 41.16 | 35.82 | 34.33 | 26.39 | 36.06 | 37.31 | 31.34 | 16.18 | 17.67 | 15.95 |
| **LLaMA-3.3-70B** | 61.43 | 84.68 | 79.03 | 80.13 | 58.37 | 70.38 | 80.83 | 80.13 | 68.42 | 41.48 | 42.81 | 47.37 |
| **LLaMA-4-Scout** | 80.05 | 84.52 | 80.44 | 82.56 | 80.13 | 79.03 | 81.38 | 81.70 | 73.92 | 45.33 | 46.58 | 63.55 |
| **LLaMA-4-Maverick** | 85.55 | 89.63 | 88.92 | 88.61 | 86.65 | 85.47 | 89.00 | 89.16 | 84.68 | 56.32 | 68.58 | 75.10 |
| **Mistral-7B-v0.3** | 22.00 | 28.04 | 14.30 | 29.30 | 21.37 | 17.12 | 23.49 | 26.00 | 18.30 | 17.52 | 10.37 | 10.68 |
| **Mistral-Small-3.1-24B** | 65.44 | 75.96 | 64.26 | 66.30 | 60.80 | 54.12 | 67.87 | 68.11 | 34.09 | 11.39 | 13.20 | 20.11 |
| **Phi-4-mini** | 31.66 | 53.10 | 36.06 | 40.38 | 32.13 | 22.55 | 35.19 | 38.73 | 28.99 | 16.89 | 20.50 | 17.44 |
| **Phi-4-mini-Reasoning** | 28.59 | 65.59 | 47.60 | 47.68 | 24.27 | 14.45 | 37.78 | 35.82 | 28.28 | 24.59 | 21.52 | 18.70 |
| **Phi-4** | 57.50 | 79.97 | 69.99 | 70.31 | 61.27 | 53.73 | 68.50 | 68.89 | 46.66 | 26.24 | 35.27 | 31.66 |
| **Phi-4-Reasoning** | 79.26 | 85.31 | 83.90 | 83.90 | 82.56 | 79.81 | 77.77 | 84.68 | 73.13 | 12.73 | 47.45 | 29.93 |
| **Qwen2.5-3B** | 45.56 | 50.82 | 40.22 | 40.46 | 38.81 | 32.76 | 39.43 | 39.59 | 14.61 | 10.45 | 10.53 | 8.56 |
| **Qwen2.5-7B** | 57.50 | 62.29 | 51.69 | 50.20 | 49.25 | 44.07 | 53.34 | 53.50 | 30.87 | 10.05 | 22.78 | 13.43 |
| **Qwen2.5-14B** | 64.89 | 69.44 | 63.86 | 62.53 | 59.23 | 55.22 | 62.06 | 62.92 | 37.00 | 21.13 | 36.29 | 33.54 |
| **Qwen2.5-72B** | 74.94 | 79.26 | 76.75 | 75.41 | 71.41 | 70.86 | 75.18 | 76.28 | 50.12 | 40.69 | 38.26 | 37.78 |
| **QwQ-32B** | 82.48 | 84.60 | 83.35 | 83.74 | 75.57 | 73.53 | 82.56 | 84.05 | 64.26 | 32.84 | 30.79 | 33.07 |
| **Qwen3-1.7B** | 44.93 | 49.18 | 40.22 | 40.14 | 36.53 | 32.91 | 37.47 | 40.22 | 27.02 | 28.04 | 25.06 | 24.59 |
| **Qwen3-4B** | 57.89 | 63.55 | 54.05 | 52.95 | 53.02 | 45.56 | 54.60 | 55.38 | 23.80 | 9.51 | 18.54 | 11.39 |
| **Qwen3-4B-thinking** | 65.51 | 72.58 | 69.21 | 67.24 | 66.22 | 60.57 | 67.56 | 67.79 | 26.39 | 13.98 | 11.15 | 8.48 |
| **Qwen3-8B** | 66.77 | 71.88 | 52.40 | 62.14 | 60.25 | 56.48 | 60.88 | 61.51 | 19.17 | 8.72 | 21.29 | 9.19 |
| **Qwen3-8B-thinking** | 73.76 | 79.89 | 75.81 | 77.45 | 74.94 | 73.29 | 76.75 | 76.75 | 28.44 | 12.80 | 14.85 | 11.63 |
| **Qwen3-14B** | 73.29 | 76.36 | 68.74 | 69.60 | 65.67 | 61.90 | 70.70 | 69.91 | 42.81 | 10.60 | 15.87 | 14.77 |
| **Qwen3-14B-thinking** | 79.81 | 82.40 | 81.85 | 79.81 | 80.68 | 78.63 | 80.83 | 82.01 | 63.08 | 12.18 | 19.95 | 13.67 |
| **Baichuan-M2-32B** | 82.40 | 85.62 | 72.51 | 79.18 | 77.45 | 70.54 | 77.22 | 76.36 | 30.56 | 24.35 | 25.92 | 28.12 |
| **Bio-Medical-LLaMA-3-8B** | 47.29 | 77.77 | 55.07 | 53.18 | 42.11 | 41.48 | 54.52 | 54.44 | 39.36 | 37.16 | 36.45 | 33.86 |
| **MediPhi** | 28.20 | 54.60 | 41.16 | 42.73 | 22.94 | 28.28 | 40.22 | 40.22 | 18.30 | 21.21 | 17.44 | 22.94 |
| **MedGemma-4B** | 48.08 | 62.77 | 53.10 | 54.36 | 48.31 | 43.68 | 52.87 | 54.67 | 42.73 | 16.50 | 19.95 | 29.54 |
| **MedGemma-27B** | 82.17 | 87.20 | 83.66 | 84.05 | 80.13 | 80.05 | 84.13 | 83.58 | 76.43 | 19.87 | 46.82 | 58.05 |
| **MedReason-8B** | 46.27 | 57.89 | 15.00 | 17.67 | 43.68 | 35.51 | 22.55 | 35.74 | 11.39 | 7.07 | 10.68 | 9.19 |
| **HuatuoGPT-o1-7B** | 63.00 | 65.36 | 65.12 | 65.04 | 53.42 | 53.73 | 65.59 | 65.12 | 7.31 | 7.38 | 6.52 | 6.13 |
| **HuatuoGPT-o1-8B** | 56.79 | 68.66 | 66.93 | 63.94 | 48.86 | 48.63 | 65.04 | 67.79 | 49.10 | 7.70 | 7.54 | 7.15 |
| **HuatuoGPT-o1-70B** | 67.16 | 86.72 | 85.31 | 84.92 | 66.54 | 74.31 | 82.33 | 86.02 | 73.53 | 30.95 | 45.80 | 55.07 |
| **HuatuoGPT-o1-72B** | 81.78 | 81.23 | 85.55 | 80.28 | 80.44 | 74.08 | 84.84 | 84.76 | 64.65 | 41.40 | 45.01 | 22.00 |
| **OpenBioLLM-8B** | 13.28 | 24.98 | 15.63 | 26.08 | 10.60 | 9.11 | 28.36 | 29.69 | 11.31 | 7.38 | 8.48 | 11.39 |
| **OpenBioLLM-70B** | 21.29 | 72.98 | 62.92 | 35.11 | 18.93 | 19.48 | 47.21 | 66.69 | 25.06 | 9.43 | 13.35 | 24.74 |

**Supplementary Tab. 119:** Zero-Shot performance evaluation of 56 LLMs on MedQA (Run 4).

| LLMs | Chinese | English | French | German | Japanese | Korean | Portuguese | Spanish | Swahili | Wolof | Yoruba | Zulu |
|---|---|---|---|---|---|---|---|---|---|---|---|---|
| **Proprietary LLMs** | | | | | | | | | | | | |
| **Claude-3.5-Haiku** | 66.85 | 77.14 | 73.06 | 73.29 | 68.03 | 68.74 | 72.19 | 73.45 | 52.55 | 34.25 | 39.83 | 37.00 |
| **Claude-4.0-Sonnet** | 89.47 | 91.83 | 89.55 | 90.26 | 89.40 | 89.32 | 91.36 | 90.49 | 81.85 | 58.60 | 71.33 | 73.92 |
| **Gemini-2.5-Flash** | 90.73 | 91.52 | 90.97 | 91.44 | 89.71 | 90.10 | 90.18 | 91.12 | 90.26 | 79.10 | 81.85 | 85.15 |
| **GPT-4o-mini** | 67.09 | 78.32 | 72.11 | 73.37 | 67.71 | 65.04 | 72.98 | 73.45 | 60.96 | 32.36 | 44.93 | 50.04 |
| **GPT-4o** | 86.88 | 89.87 | 87.98 | 87.98 | 85.00 | 84.29 | 89.40 | 88.69 | 82.56 | 31.58 | 51.77 | 69.60 |
| **GPT-4.1-nano** | 67.64 | 81.15 | 71.17 | 71.01 | 63.39 | 61.74 | 72.43 | 73.37 | 54.05 | 27.18 | 38.41 | 44.15 |
| **GPT-4.1-mini** | 86.88 | 91.67 | 87.04 | 87.75 | 86.25 | 82.48 | 87.12 | 88.14 | 79.26 | 24.04 | 58.37 | 67.56 |
| **GPT-4.1** | 88.45 | 89.24 | 90.02 | 89.08 | 89.95 | 86.33 | 91.67 | 89.47 | 88.14 | 52.08 | 69.29 | 77.06 |
| **GPT-5-nano** | 51.61 | 70.31 | 64.89 | 69.36 | 65.59 | 54.28 | 61.59 | 71.41 | 54.05 | 21.13 | 39.28 | 43.36 |
| **GPT-5-mini** | 87.35 | 90.65 | 87.59 | 89.24 | 87.82 | 84.76 | 88.77 | 89.40 | 80.68 | 34.88 | 62.06 | 71.09 |
| **GPT-5** | 90.57 | 93.24 | 91.99 | 91.67 | 91.28 | 84.68 | 91.75 | 92.30 | 89.47 | 53.34 | 72.98 | 76.20 |
| **o4-mini** | 93.09 | 93.72 | 93.40 | 93.09 | 92.54 | 91.12 | 93.64 | 93.64 | 91.67 | 38.96 | 80.28 | 85.00 |
| **Open-Weight LLMs** | | | | | | | | | | | | |
| **DeepSeek-V3** | 87.04 | 89.47 | 86.33 | 86.49 | 84.92 | 82.72 | 86.80 | 87.27 | 75.26 | 44.07 | 55.85 | 64.02 |
| **DeepSeek-R1** | 90.02 | 91.04 | 91.12 | 90.97 | 90.26 | 80.13 | 91.67 | 91.52 | 87.43 | 66.06 | 71.48 | 79.65 |
| **DeepSeek-R1-Qwen3-8B** | 73.53 | 80.13 | 69.36 | 70.31 | 65.99 | 67.71 | 70.46 | 69.99 | 35.04 | 9.98 | 12.02 | 10.68 |
| **Gemma-3-4B** | 43.36 | 51.69 | 45.01 | 45.17 | 42.34 | 38.57 | 46.58 | 47.92 | 37.94 | 10.84 | 17.60 | 14.61 |
| **Gemma-3-12B** | 60.72 | 69.76 | 64.18 | 64.96 | 61.04 | 59.78 | 65.20 | 64.81 | 56.79 | 8.88 | 35.51 | 44.15 |
| **Gemma-3-27B** | 68.11 | 75.73 | 72.51 | 72.43 | 70.46 | 69.36 | 71.41 | 71.88 | 65.44 | 20.11 | 46.11 | 52.47 |
| **gpt-oss-20B** | 82.01 | 80.91 | 83.66 | 82.48 | 84.76 | 82.33 | 84.92 | 84.21 | 68.03 | 30.71 | 60.33 | 66.46 |
| **gpt-oss-120B** | 89.00 | 90.97 | 90.49 | 91.28 | 88.85 | 87.90 | 92.22 | 90.97 | 80.68 | 55.22 | 71.41 | 77.93 |
| **LLaMA-3.1-8B** | 45.64 | 66.54 | 49.49 | 49.10 | 37.78 | 30.87 | 48.55 | 51.14 | 35.90 | 24.59 | 16.89 | 11.94 |
| **LLaMA-3.1-70B** | 72.43 | 83.35 | 76.43 | 76.75 | 69.52 | 62.69 | 77.06 | 78.48 | 62.14 | 39.36 | 42.50 | 46.74 |
| **LLaMA-3.2-3B** | 38.81 | 56.87 | 40.61 | 34.72 | 32.84 | 27.73 | 38.41 | 38.49 | 31.97 | 16.50 | 21.52 | 15.00 |
| **LLaMA-3.3-70B** | 60.88 | 84.84 | 79.97 | 80.13 | 58.99 | 70.38 | 81.23 | 81.62 | 67.95 | 42.26 | 43.36 | 47.53 |
| **LLaMA-4-Scout** | 79.58 | 84.68 | 81.78 | 83.35 | 79.81 | 78.00 | 80.83 | 82.33 | 74.78 | 45.88 | 46.43 | 63.39 |
| **LLaMA-4-Maverick** | 85.62 | 90.49 | 88.85 | 88.53 | 86.02 | 85.86 | 88.30 | 89.08 | 83.74 | 56.72 | 68.11 | 76.43 |
| **Mistral-7B-v0.3** | 21.52 | 29.38 | 13.90 | 29.85 | 22.07 | 17.44 | 23.64 | 28.04 | 20.11 | 17.75 | 9.11 | 9.27 |
| **Mistral-Small-3.1-24B** | 63.32 | 76.59 | 65.51 | 67.79 | 62.53 | 56.48 | 67.40 | 68.34 | 31.74 | 12.88 | 15.48 | 18.62 |
| **Phi-4-mini** | 34.41 | 53.26 | 34.25 | 39.98 | 30.56 | 24.12 | 36.45 | 37.23 | 29.93 | 17.99 | 20.97 | 18.62 |
| **Phi-4-mini-Reasoning** | 25.77 | 64.65 | 45.33 | 48.00 | 23.25 | 13.51 | 41.40 | 34.33 | 26.79 | 23.88 | 20.82 | 20.90 |
| **Phi-4** | 58.76 | 79.50 | 70.31 | 71.01 | 62.06 | 55.54 | 69.52 | 71.56 | 45.56 | 25.06 | 36.45 | 30.87 |
| **Phi-4-Reasoning** | 81.38 | 85.07 | 84.05 | 84.05 | 82.88 | 79.10 | 76.12 | 84.84 | 72.43 | 14.30 | 47.53 | 28.83 |
| **Qwen2.5-3B** | 46.66 | 51.14 | 39.75 | 40.77 | 37.63 | 33.54 | 39.28 | 38.96 | 13.35 | 9.98 | 11.78 | 8.64 |
| **Qwen2.5-7B** | 55.77 | 59.94 | 51.85 | 48.70 | 49.80 | 44.23 | 52.32 | 50.75 | 30.09 | 9.11 | 21.52 | 13.98 |
| **Qwen2.5-14B** | 65.99 | 69.44 | 61.59 | 62.06 | 61.04 | 54.60 | 61.12 | 62.92 | 36.45 | 20.82 | 36.14 | 32.76 |
| **Qwen2.5-72B** | 74.55 | 80.20 | 75.49 | 75.65 | 73.06 | 70.62 | 74.94 | 75.33 | 49.18 | 39.75 | 38.41 | 38.02 |
| **QwQ-32B** | 82.88 | 84.68 | 84.37 | 83.03 | 76.43 | 72.58 | 83.58 | 83.97 | 61.90 | 31.19 | 31.81 | 31.34 |
| **Qwen3-1.7B** | 44.23 | 49.41 | 39.04 | 39.28 | 35.27 | 33.62 | 37.39 | 39.98 | 27.18 | 25.14 | 26.16 | 26.47 |
| **Qwen3-4B** | 59.31 | 64.02 | 52.71 | 52.95 | 53.50 | 44.85 | 54.75 | 54.52 | 22.94 | 9.27 | 16.73 | 9.74 |
| **Qwen3-4B-thinking** | 66.06 | 72.11 | 67.87 | 68.74 | 66.22 | 59.86 | 67.56 | 68.50 | 26.32 | 15.55 | 11.94 | 9.11 |
| **Qwen3-8B** | 67.24 | 71.72 | 51.85 | 62.69 | 59.54 | 56.95 | 62.06 | 60.25 | 22.15 | 8.80 | 21.37 | 10.53 |
| **Qwen3-8B-thinking** | 73.29 | 79.65 | 76.90 | 77.22 | 75.26 | 72.82 | 75.88 | 76.83 | 27.42 | 8.80 | 15.79 | 9.51 |
| **Qwen3-14B** | 74.71 | 77.22 | 67.87 | 70.93 | 66.46 | 60.41 | 69.99 | 70.07 | 40.85 | 11.55 | 16.50 | 14.77 |
| **Qwen3-14B-thinking** | 79.58 | 83.74 | 80.44 | 80.68 | 80.91 | 76.83 | 79.89 | 80.83 | 62.77 | 13.12 | 17.99 | 15.63 |
| **Baichuan-M2-32B** | 84.29 | 85.86 | 74.31 | 79.50 | 75.81 | 70.70 | 77.69 | 75.65 | 30.24 | 24.90 | 28.83 | 29.77 |
| **Bio-Medical-LLaMA-3-8B** | 46.74 | 77.22 | 56.01 | 52.40 | 42.11 | 42.73 | 55.15 | 54.44 | 39.98 | 36.84 | 35.35 | 33.15 |
| **MediPhi** | 28.44 | 51.61 | 43.68 | 41.87 | 23.80 | 27.18 | 38.81 | 41.56 | 19.72 | 22.15 | 17.44 | 23.17 |
| **MedGemma-4B** | 47.29 | 64.10 | 53.65 | 55.38 | 47.53 | 44.07 | 53.02 | 51.77 | 42.73 | 16.18 | 20.50 | 27.57 |
| **MedGemma-27B** | 82.95 | 87.59 | 83.66 | 83.42 | 79.34 | 78.24 | 83.58 | 84.05 | 77.38 | 16.26 | 45.09 | 57.74 |
| **MedReason-8B** | 44.78 | 59.23 | 13.75 | 16.42 | 45.64 | 36.53 | 21.45 | 36.61 | 10.29 | 8.41 | 10.53 | 9.82 |
| **HuatuoGPT-o1-7B** | 62.61 | 65.04 | 63.86 | 65.04 | 55.93 | 54.75 | 65.04 | 65.20 | 7.54 | 6.44 | 7.23 | 6.76 |
| **HuatuoGPT-o1-8B** | 58.52 | 68.34 | 65.04 | 65.44 | 51.69 | 49.57 | 63.32 | 64.02 | 48.86 | 8.01 | 7.62 | 7.23 |
| **HuatuoGPT-o1-70B** | 68.11 | 87.43 | 85.78 | 85.15 | 67.16 | 73.29 | 84.05 | 85.47 | 73.92 | 29.69 | 47.60 | 56.64 |
| **HuatuoGPT-o1-72B** | 81.38 | 83.50 | 85.86 | 81.30 | 78.24 | 74.39 | 85.07 | 83.82 | 63.86 | 42.11 | 44.07 | 22.39 |
| **OpenBioLLM-8B** | 15.24 | 25.92 | 15.24 | 26.39 | 9.98 | 9.66 | 30.09 | 31.11 | 10.84 | 8.72 | 7.86 | 10.92 |
| **OpenBioLLM-70B** | 20.90 | 72.35 | 59.47 | 33.94 | 19.80 | 19.17 | 45.40 | 65.51 | 27.49 | 10.92 | 12.02 | 25.61 |

**Supplementary Tab. 120:** Zero-Shot performance evaluation of 56 LLMs on MedQA (Run 5).

| LLMs | Chinese | English | French | German | Japanese | Korean | Portuguese | Spanish | Swahili | Wolof | Yoruba | Zulu | Overall |
|---|---|---|---|---|---|---|---|---|---|---|---|---|---|
| **Proprietary LLMs** | | | | | | | | | | | | | |
| Claude-3.5-Haiku | 59.36±0.00 | 68.45±0.00 | 71.66±0.00 | 63.10±0.00 | 59.89±0.00 | 55.61±0.00 | 68.45±0.00 | 72.62±0.24 | 36.36±0.00 | 15.51±1.46 | 20.96±0.45 | 29.84±0.24 | 51.82 ± 0.12 |
| Claude-4.0-Sonnet | 77.65±1.57 | 82.78±0.59 | 80.75±1.07 | 77.86±0.97 | 78.61±0.76 | 76.47±0.54 | 80.00±0.81 | 81.39±1.22 | 70.59±1.85 | 35.08±1.54 | 55.83±1.54 | 63.31±2.35 | 71.69 ± 0.39 |
| Gemini-2.5-Flash | 76.26±1.11 | 75.73±0.90 | 76.90±0.88 | 77.33±1.63 | 75.72±0.48 | 76.79±1.63 | 78.72±1.22 | 78.29±0.89 | 75.72±0.61 | 63.42±1.54 | 68.66±1.17 | 68.12±0.90 | 74.31 ± 0.42 |
| GPT-4o-mini | 57.11±1.91 | 71.98±1.11 | 68.13±1.59 | 65.88±1.44 | 60.43±2.48 | 56.68±3.10 | 66.63±1.23 | 68.34±0.88 | 48.02±1.83 | 16.47±0.88 | 27.81±2.14 | 34.54±1.04 | 53.50 ± 0.61 |
| GPT-4o | 72.73±1.07 | 76.26±1.95 | 75.72±2.09 | 73.58±1.88 | 73.26±1.25 | 69.52±1.25 | 76.37±1.75 | 76.37±1.83 | 66.41±2.34 | 25.99±2.26 | 42.68±1.28 | 55.51±1.83 | 65.37 ± 0.40 |
| GPT-4.1-nano | 56.58±3.24 | 72.73±2.03 | 65.56±1.17 | 65.99±1.23 | 54.87±1.95 | 47.49±2.57 | 66.52±0.97 | 68.34±2.25 | 37.22±2.90 | 14.65±2.95 | 21.28±1.71 | 30.59±1.98 | 50.15 ± 0.68 |
| GPT-4.1-mini | 70.05±1.69 | 78.07±1.07 | 72.41±1.29 | 74.97±1.16 | 70.91±1.34 | 66.10±0.48 | 75.51±1.49 | 73.05±2.32 | 62.03±2.07 | 15.51±1.47 | 44.06±0.90 | 53.16±2.52 | 62.99 ± 0.63 |
| GPT-4.1 | 74.65±0.97 | 77.65±1.33 | 77.54±0.84 | 75.72±1.54 | 74.87±1.41 | 71.34±1.54 | 79.57±1.28 | 79.25±0.96 | 74.12±1.34 | 31.44±3.24 | 46.10±1.83 | 63.32±1.59 | 68.80 ± 0.48 |
| GPT-5-nano | 47.59±2.17 | 65.78±1.96 | 58.29±1.82 | 60.32±2.09 | 52.83±1.62 | 40.43±3.48 | 56.47±2.22 | 59.68±1.17 | 39.14±1.38 | 12.51±1.54 | 20.75±2.37 | 26.84±1.62 | 45.05 ± 0.47 |
| GPT-5-mini | 73.16±1.67 | 76.69±0.61 | 74.22±2.34 | 73.58±2.64 | 71.34±1.72 | 64.81±2.46 | 74.76±1.03 | 76.26±1.29 | 61.50±2.39 | 19.89±1.91 | 44.17±1.23 | 56.47±2.66 | 63.90 ± 0.65 |
| GPT-5 | 75.30±1.67 | 76.90±0.95 | 77.33±1.17 | 74.12±2.06 | 74.22±1.43 | 68.55±1.67 | 79.78±1.03 | 79.36±0.61 | 70.70±0.79 | 36.47±2.25 | 55.61±2.80 | 63.42±2.74 | 69.31 ± 0.33 |
| o4-mini | 74.97±1.48 | 79.79±0.24 | 80.10±0.79 | 74.87±1.85 | 75.72±0.90 | 71.23±1.91 | 80.43±0.72 | 78.39±1.91 | 72.51±1.45 | 30.38±1.83 | 60.11±1.40 | 67.27±1.27 | 70.48 ± 0.37 |
| **Open-Weight LLMs** | | | | | | | | | | | | | |
| DeepSeek-V3 | 74.98±1.33 | 79.68±0.85 | 75.72±0.61 | 75.72±1.34 | 75.40±1.07 | 69.20±1.17 | 78.82±1.49 | 75.30±1.48 | 66.20±1.62 | 30.16±1.91 | 36.90±1.89 | 46.74±2.09 | 65.40 ± 0.11 |
| DeepSeek-R1 | 70.05±1.00 | 78.39±0.81 | 78.18±1.22 | 76.79±0.72 | 76.36±1.22 | 63.85±2.09 | 78.29±0.90 | 77.33±0.97 | 74.44±1.58 | 46.10±1.79 | 57.01±2.64 | 61.18±0.97 | 69.83 ± 0.34 |
| DeepSeek-R1-Qwen3-8B | 61.39±1.33 | 68.23±0.81 | 58.40±3.87 | 61.93±2.76 | 54.87±2.49 | 56.90±1.54 | 63.74±2.46 | 59.68±1.59 | 19.57±2.66 | 10.91±2.41 | 10.91±2.84 | 10.59±2.15 | 44.76 ± 0.65 |
| Gemma-3-4B | 31.02±1.20 | 40.86±2.69 | 31.98±1.58 | 35.29±2.75 | 27.81±1.14 | 22.67±2.22 | 29.95±2.04 | 27.70±2.94 | 22.35±1.03 | 11.44±2.02 | 11.02±1.58 | 10.05±3.15 | 25.18 ± 0.60 |
| Gemma-3-12B | 51.12±2.53 | 61.50±0.66 | 58.71±1.44 | 58.39±2.25 | 53.05±2.63 | 51.87±2.67 | 59.04±2.32 | 58.61±1.63 | 45.99±2.90 | 9.41±2.02 | 23.63±1.28 | 32.94±1.80 | 47.02 ± 0.61 |
| Gemma-3-27B | 62.46±3.52 | 67.38±1.65 | 68.98±2.39 | 64.49±1.17 | 58.50±0.72 | 57.97±1.72 | 70.27±1.80 | 67.70±2.94 | 54.54±2.00 | 13.15±1.34 | 27.59±0.81 | 44.28±1.09 | 54.78 ± 0.34 |
| gpt-oss-20B | 63.32±2.09 | 64.92±3.13 | 69.09±1.67 | 67.49±1.58 | 64.07±0.45 | 63.74±1.91 | 68.34±1.48 | 69.09±2.22 | 49.09±1.28 | 22.78±0.81 | 42.67±3.01 | 47.38±2.02 | 57.66 ± 0.85 |
| gpt-oss-120B | 70.27±0.61 | 77.22±1.34 | 75.93±2.07 | 74.12±1.29 | 72.94±1.39 | 69.84±2.19 | 75.83±1.03 | 75.83±2.12 | 62.03±1.69 | 35.40±3.06 | 54.44±2.55 | 58.72±2.66 | 66.88 ± 0.38 |
| LLaMA-3.1-8B | 34.87±1.71 | 54.44±1.38 | 38.61±3.24 | 38.93±1.48 | 24.60±1.73 | 20.00±3.48 | 38.29±1.44 | 40.21±2.05 | 22.25±0.81 | 15.83±2.23 | 10.70±3.07 | 10.48±3.87 | 29.10 ± 0.57 |
| LLaMA-3.1-70B | 59.89±2.73 | 72.30±1.22 | 66.63±1.04 | 67.81±2.21 | 59.89±2.62 | 52.30±1.33 | 66.63±1.63 | 68.66±2.55 | 50.80±2.17 | 20.75±1.39 | 22.99±2.33 | 28.77±3.37 | 53.12 ± 0.75 |
| LLaMA-3.2-3B | 29.84±1.16 | 40.64±2.33 | 27.59±1.95 | 30.91±1.91 | 19.68±2.76 | 12.73±1.67 | 22.99±1.20 | 25.13±3.05 | 16.04±1.46 | 11.01±2.66 | 8.13±2.96 | 10.80±1.48 | 21.29 ± 0.87 |
| LLaMA-3.3-70B | 53.15±2.02 | 75.94±0.65 | 73.58±1.54 | 71.98±0.97 | 51.87±2.30 | 60.32±1.16 | 71.66±0.00 | 72.83±0.70 | 57.65±1.48 | 24.06±1.77 | 27.17±1.58 | 34.23±1.56 | 56.20 ± 0.43 |
| LLaMA-4-Scout | 69.41±1.16 | 74.44±1.43 | 73.69±1.58 | 70.37±0.48 | 65.99±1.29 | 67.59±0.97 | 74.12±1.17 | 73.05±1.23 | 62.14±2.25 | 29.95±2.30 | 29.41±0.38 | 54.23±0.97 | 62.03 ± 0.47 |
| LLaMA-4-Maverick | 73.90±1.03 | 80.85±0.70 | 77.65±1.16 | 75.83±1.16 | 74.33±2.14 | 71.98±1.04 | 77.33±1.11 | 77.22±1.11 | 72.09±1.16 | 37.32±0.96 | 53.48±2.10 | 59.68±1.34 | 69.30 ± 0.29 |
| Mistral-7B-v0.3 | 13.90±1.31 | 20.75±3.90 | 13.90±1.26 | 17.65±2.68 | 12.94±1.09 | 11.76±2.33 | 18.08±1.75 | 16.79±1.23 | 9.41±1.72 | 10.05±0.79 | 10.38±1.39 | 10.27±1.66 | 13.82 ± 0.36 |
| Mistral-Small-3.1-24B | 56.79±2.05 | 70.05±2.17 | 65.24±3.49 | 64.92±2.35 | 58.93±1.28 | 48.98±3.68 | 65.13±2.43 | 65.78±1.81 | 23.64±3.99 | 11.34±2.25 | 10.48±1.49 | 11.23±2.78 | 46.04 ± 1.12 |
| Phi-4-mini | 25.45±2.74 | 48.56±3.17 | 31.77±3.64 | 32.09±2.42 | 20.53±4.01 | 16.26±2.69 | 26.52±3.52 | 28.13±2.97 | 16.15±0.45 | 10.37±1.59 | 11.34±1.16 | 10.91±1.23 | 23.17 ± 0.66 |
| Phi-4-mini-Reasoning | 20.86±2.10 | 60.00±2.31 | 43.32±3.36 | 44.17±2.35 | 18.39±3.31 | 10.91±3.15 | 36.47±5.91 | 31.87±2.35 | 16.69±2.99 | 13.91±3.47 | 11.12±1.79 | 14.01±1.03 | 26.81 ± 0.75 |
| Phi-4 | 51.66±2.44 | 72.51±0.90 | 65.56±2.02 | 68.02±1.66 | 56.68±1.73 | 48.88±3.13 | 68.02±0.70 | 66.31±0.84 | 33.58±1.94 | 13.58±1.68 | 18.61±1.38 | 18.08±2.25 | 48.46 ± 0.82 |
| Phi-4-Reasoning | 67.48±0.88 | 74.76±2.12 | 74.44±1.33 | 72.19±1.65 | 71.66±1.36 | 64.81±2.01 | 68.77±1.54 | 74.33±1.00 | 56.04±1.75 | 16.58±1.69 | 32.19±2.05 | 25.35±1.54 | 58.22 ± 0.29 |
| Qwen2.5-3B | 37.75±1.80 | 39.79±1.88 | 32.19±3.10 | 28.23±2.94 | 24.92±1.72 | 20.32±3.65 | 31.77±2.12 | 32.84±1.91 | 11.34±0.79 | 10.05±1.90 | 11.87±2.57 | 9.63±4.25 | 24.22 ± 0.93 |
| Qwen2.5-7B | 47.70±2.84 | 58.82±3.36 | 49.41±2.06 | 42.57±2.19 | 40.86±1.95 | 31.77±2.29 | 45.56±2.28 | 46.52±1.36 | 17.22±1.16 | 11.02±1.34 | 12.41±1.75 | 10.16±1.25 | 34.50 ± 0.76 |
| Qwen2.5-14B | 59.57±2.26 | 68.13±1.54 | 57.97±2.63 | 57.75±2.24 | 52.51±0.45 | 49.20±2.83 | 60.86±2.31 | 59.89±3.36 | 20.43±1.53 | 16.90±2.79 | 14.12±1.23 | 19.47±1.50 | 44.73 ± 0.38 |
| Qwen2.5-72B | 69.95±0.45 | 73.80±1.20 | 69.73±0.97 | 68.24±1.40 | 66.84±1.00 | 61.60±1.58 | 71.55±1.98 | 69.30±0.97 | 35.51±2.32 | 18.18±1.07 | 15.40±0.45 | 21.50±1.27 | 53.47 ± 0.15 |
| QwQ-32B | 69.41±1.83 | 74.76±1.58 | 73.90±1.03 | 72.94±1.29 | 68.45±1.00 | 64.38±2.22 | 76.15±1.23 | 74.97±1.67 | 46.63±2.12 | 18.61±2.38 | 19.14±2.22 | 20.86±1.65 | 56.68 ± 0.51 |
| Qwen3-1.7B | 29.95±0.54 | 38.40±2.49 | 28.77±2.22 | 23.32±2.92 | 20.54±1.04 | 21.07±2.79 | 24.28±1.76 | 25.78±2.09 | 14.76±2.44 | 11.44±2.35 | 11.98±0.30 | 13.58±1.40 | 21.99 ± 0.65 |
| Qwen3-4B | 53.16±1.95 | 59.15±1.76 | 50.80±2.48 | 50.59±1.72 | 43.64±2.22 | 37.75±2.23 | 48.34±1.45 | 50.27±3.49 | 12.51±1.80 | 11.23±0.65 | 11.02±2.29 | 10.70±1.46 | 36.60 ± 0.43 |
| Qwen3-4B-thinking | 58.18±0.88 | 63.75±1.44 | 64.39±1.72 | 60.75±1.63 | 57.43±1.04 | 53.58±2.84 | 61.50±2.30 | 61.60±0.70 | 15.30±1.91 | 12.08±0.97 | 12.84±3.03 | 9.84±2.74 | 44.27 ± 0.32 |
| Qwen3-8B | 58.50±2.16 | 66.95±1.16 | 54.65±0.79 | 54.76±1.71 | 52.30±2.12 | 45.67±2.52 | 56.47±1.84 | 57.86±3.01 | 19.68±2.86 | 11.66±1.38 | 13.05±1.88 | 12.09±3.28 | 41.97 ± 0.78 |
| Qwen3-8B-thinking | 62.46±2.44 | 72.30±1.53 | 68.87±2.02 | 66.52±2.29 | 64.92±1.59 | 62.57±2.42 | 68.13±2.47 | 68.13±0.81 | 24.06±3.23 | 12.94±2.66 | 13.58±2.16 | 13.05±1.44 | 49.79 ± 0.36 |
| Qwen3-14B | 65.13±1.38 | 71.34±2.41 | 65.24±1.41 | 65.99±1.34 | 58.40±2.08 | 57.22±2.48 | 68.77±2.03 | 65.67±1.75 | 24.38±0.72 | 11.34±2.52 | 12.83±1.20 | 16.15±3.15 | 48.54 ± 0.61 |
| Qwen3-14B-thinking | 66.31±1.31 | 75.29±1.75 | 72.84±0.79 | 71.55±1.09 | 70.69±1.75 | 66.31±1.69 | 71.02±2.05 | 71.34±1.63 | 44.92±1.97 | 12.73±1.62 | 17.43±2.06 | 16.04±2.21 | 54.71 ± 0.42 |
| Baichuan-M2-32B | 69.52±1.93 | 75.40±1.81 | 69.41±1.38 | 67.91±1.56 | 63.85±2.63 | 56.04±1.53 | 67.91±1.19 | 67.59±4.29 | 18.61±2.46 | 16.36±2.97 | 18.29±2.96 | 15.83±1.87 | 50.56 ± 0.77 |
| Bio-Medical-LLaMA-3-8B | 29.84±1.44 | 55.08±1.13 | 36.79±1.62 | 39.57±1.00 | 28.34±1.36 | 22.89±1.43 | 35.51±0.61 | 38.93±0.70 | 25.99±1.68 | 19.04±1.54 | 13.37±0.76 | 15.83±0.90 | 30.10 ± 0.42 |
| MediPhi | 19.15±2.68 | 42.03±2.32 | 32.94±3.64 | 28.23±2.22 | 12.51±1.17 | 10.27±2.08 | 31.45±1.48 | 30.70±0.90 | 9.41±3.35 | 10.59±1.86 | 11.12±1.27 | 10.59±1.22 | 20.75 ± 0.43 |
| MedGemma-4B | 33.48±1.04 | 46.31±1.95 | 42.25±3.12 | 37.22±1.72 | 30.91±1.75 | 29.63±2.23 | 36.04±1.49 | 36.68±1.29 | 23.64±2.18 | 13.05±2.44 | 12.41±2.40 | 16.26±2.99 | 29.82 ± 0.55 |
| MedGemma-27B | 68.88±2.08 | 75.29±1.87 | 73.80±1.89 | 74.01±1.23 | 69.20±1.54 | 67.27±1.38 | 73.15±1.71 | 73.69±2.31 | 61.28±2.06 | 14.86±2.63 | 29.73±0.61 | 46.31±3.44 | 60.62 ± 0.47 |
| MedReason-8B | 38.07±2.46 | 48.13±4.02 | 20.75±0.79 | 21.82±1.16 | 37.22±1.39 | 26.52±1.23 | 21.39±3.76 | 28.88±3.32 | 14.54±1.49 | 11.66±0.70 | 9.31±0.30 | 10.59±1.48 | 24.07 ± 0.44 |
| HuatuoGPT-o1-7B | 53.80±1.67 | 58.29±3.38 | 57.43±2.35 | 53.16±3.13 | 48.77±1.48 | 41.39±1.34 | 59.36±0.54 | 58.61±2.99 | 10.37±1.11 | 11.34±0.88 | 9.52±2.08 | 10.48±0.89 | 39.38 ± 0.99 |
| HuatuoGPT-o1-8B | 47.06±1.31 | 54.33±1.11 | 51.66±4.32 | 49.52±3.25 | 41.18±1.93 | 35.61±1.44 | 48.66±1.07 | 45.24±3.22 | 28.88±1.13 | 11.66±1.83 | 11.76±2.30 | 10.27±1.94 | 36.32 ± 0.40 |
| HuatuoGPT-o1-70B | 60.53±2.19 | 72.94±1.23 | 69.62±2.05 | 69.20±1.34 | 61.39±2.22 | 65.45±1.83 | 70.59±2.17 | 70.69±1.03 | 57.11±1.10 | 24.17±1.66 | 30.48±1.65 | 38.82±2.19 | 57.58 ± 0.34 |
| HuatuoGPT-o1-72B | 69.52±1.81 | 75.30±1.79 | 73.37±1.16 | 68.77±2.55 | 70.69±1.67 | 65.56±0.90 | 74.55±1.17 | 73.90±1.44 | 44.92±2.33 | 20.64±2.74 | 24.49±1.48 | 17.11±3.10 | 56.57 ± 0.45 |
| OpenBioLLM-8B | 12.41±1.27 | 22.25±3.64 | 16.15±2.37 | 16.79±1.68 | 11.44±1.63 | 10.59±3.30 | 17.33±1.50 | 15.08±1.49 | 10.59±1.48 | 10.37±2.02 | 9.52±2.76 | 9.84±1.39 | 13.53 ± 0.53 |
| OpenBioLLM-70B | 26.42±3.37 | 62.35±1.23 | 56.90±2.26 | 38.50±2.00 | 24.49±3.19 | 21.93±2.45 | 42.68±3.77 | 59.89±1.36 | 26.31±1.91 | 9.20±1.49 | 11.12±1.90 | 15.29±2.41 | 32.92 ± 0.83 |

**Supplementary Tab. 121:** Performance evaluation of 56 LLMs on MMLU-Pro.

| LLMs | Chinese | English | French | German | Japanese | Korean | Portuguese | Spanish | Swahili | Wolof | Yoruba | Zulu |
|---|---|---|---|---|---|---|---|---|---|---|---|---|
| **Proprietary LLMs** | | | | | | | | | | | | |
| **Claude-3.5-Haiku** | 59.36 | 68.45 | 71.66 | 63.10 | 59.89 | 55.61 | 68.45 | 72.73 | 36.36 | 16.58 | 20.32 | 29.95 |
| **Claude-4.0-Sonnet** | 75.94 | 82.35 | 80.21 | 77.54 | 78.61 | 77.01 | 81.28 | 82.89 | 73.26 | 33.69 | 54.01 | 66.84 |
| **Gemini-2.5-Flash** | 77.01 | 74.87 | 76.47 | 76.47 | 76.47 | 75.94 | 78.61 | 78.07 | 75.94 | 63.10 | 68.45 | 67.91 |
| **GPT-4o-mini** | 54.01 | 73.26 | 66.31 | 65.24 | 58.29 | 51.87 | 67.91 | 66.84 | 49.73 | 14.97 | 26.74 | 35.29 |
| **GPT-4o** | 71.66 | 77.54 | 75.40 | 75.94 | 74.33 | 70.59 | 74.33 | 77.01 | 62.57 | 25.67 | 40.64 | 55.08 |
| **GPT-4.1-nano** | 54.01 | 71.66 | 65.24 | 64.71 | 55.08 | 48.13 | 67.38 | 66.31 | 35.29 | 12.30 | 23.53 | 29.41 |
| **GPT-4.1-mini** | 72.19 | 79.14 | 70.59 | 74.33 | 72.19 | 66.31 | 76.47 | 70.05 | 62.57 | 17.65 | 43.85 | 54.01 |
| **GPT-4.1** | 74.87 | 78.61 | 76.47 | 76.47 | 76.47 | 71.66 | 78.07 | 80.21 | 74.87 | 31.02 | 45.99 | 65.78 |
| **GPT-5-nano** | 48.66 | 66.31 | 59.36 | 59.36 | 51.87 | 43.85 | 55.61 | 59.89 | 40.64 | 13.90 | 19.25 | 27.27 |
| **GPT-5-mini** | 75.94 | 77.54 | 75.40 | 75.40 | 71.12 | 67.38 | 75.94 | 77.01 | 62.03 | 17.11 | 45.45 | 60.43 |
| **GPT-5** | 76.47 | 77.54 | 78.61 | 76.47 | 74.33 | 68.45 | 79.68 | 79.14 | 71.12 | 36.90 | 52.41 | 65.78 |
| **o4-mini** | 74.33 | 79.68 | 81.28 | 77.54 | 76.47 | 73.26 | 79.68 | 76.47 | 71.66 | 29.41 | 59.89 | 65.78 |
| **Open-Weight LLMs** | | | | | | | | | | | | |
| **DeepSeek-V3** | 77.01 | 78.61 | 74.87 | 75.94 | 74.87 | 70.59 | 81.28 | 77.54 | 65.24 | 28.88 | 35.29 | 44.39 |
| **DeepSeek-R1** | 68.98 | 79.68 | 79.14 | 77.54 | 74.33 | 62.57 | 78.61 | 77.01 | 71.66 | 44.92 | 56.68 | 60.96 |
| **DeepSeek-R1-Qwen3-8B** | 61.50 | 68.98 | 57.22 | 64.17 | 57.75 | 55.61 | 64.71 | 58.29 | 23.53 | 9.09 | 7.49 | 8.02 |
| **Gemma-3-4B** | 30.48 | 41.71 | 34.22 | 39.57 | 28.88 | 19.79 | 28.88 | 26.74 | 24.06 | 10.16 | 11.76 | 7.49 |
| **Gemma-3-12B** | 52.41 | 62.57 | 56.68 | 57.22 | 55.08 | 48.13 | 59.36 | 56.15 | 43.32 | 8.02 | 22.99 | 35.83 |
| **Gemma-3-27B** | 65.24 | 68.45 | 69.52 | 64.17 | 57.75 | 55.08 | 69.52 | 64.17 | 55.61 | 11.23 | 27.81 | 43.85 |
| **gpt-oss-20B** | 63.10 | 59.89 | 70.59 | 65.78 | 64.71 | 64.17 | 66.31 | 66.84 | 48.66 | 22.99 | 43.85 | 49.20 |
| **gpt-oss-120B** | 71.12 | 77.54 | 73.26 | 72.73 | 72.73 | 71.66 | 74.33 | 75.94 | 63.10 | 34.76 | 55.61 | 63.10 |
| **LLaMA-3.1-8B** | 35.83 | 54.55 | 33.69 | 40.64 | 26.20 | 19.79 | 40.64 | 39.04 | 21.39 | 13.90 | 13.37 | 16.04 |
| **LLaMA-3.1-70B** | 63.64 | 73.26 | 67.38 | 68.98 | 62.57 | 54.01 | 67.38 | 67.38 | 51.34 | 22.46 | 25.13 | 26.20 |
| **LLaMA-3.2-3B** | 30.48 | 40.11 | 27.27 | 29.41 | 18.72 | 11.23 | 22.99 | 23.53 | 16.04 | 9.09 | 6.95 | 10.16 |
| **LLaMA-3.3-70B** | 55.61 | 76.47 | 71.66 | 70.59 | 52.94 | 59.36 | 71.66 | 72.19 | 59.89 | 24.06 | 29.41 | 34.76 |
| **LLaMA-4-Scout** | 69.52 | 73.80 | 75.94 | 70.59 | 66.31 | 68.45 | 75.40 | 74.33 | 60.96 | 33.16 | 29.41 | 55.08 |
| **LLaMA-4-Maverick** | 72.19 | 81.28 | 78.61 | 76.47 | 74.87 | 72.73 | 77.54 | 76.47 | 71.12 | 38.50 | 50.80 | 60.96 |
| **Mistral-7B-v0.3** | 12.30 | 18.18 | 15.51 | 18.18 | 13.90 | 14.44 | 20.86 | 16.58 | 11.23 | 10.16 | 10.70 | 10.70 |
| **Mistral-Small-3.1-24B** | 57.22 | 68.98 | 64.17 | 67.91 | 58.29 | 51.87 | 67.91 | 64.71 | 21.93 | 12.83 | 10.16 | 12.83 |
| **Phi-4-mini** | 22.99 | 49.20 | 26.74 | 32.09 | 18.18 | 15.51 | 24.06 | 32.62 | 16.58 | 10.16 | 12.83 | 12.30 |
| **Phi-4-mini-Reasoning** | 22.46 | 61.50 | 43.32 | 46.52 | 14.44 | 7.49 | 41.18 | 33.16 | 14.44 | 12.30 | 12.30 | 14.44 |
| **Phi-4** | 54.55 | 73.26 | 65.78 | 66.31 | 56.68 | 48.13 | 67.91 | 66.84 | 33.69 | 14.97 | 20.32 | 18.72 |
| **Phi-4-Reasoning** | 67.91 | 76.47 | 74.87 | 74.33 | 72.73 | 62.57 | 70.05 | 75.40 | 57.22 | 16.04 | 29.41 | 25.67 |
| **Qwen2.5-3B** | 40.11 | 39.57 | 35.83 | 25.13 | 25.67 | 16.04 | 34.22 | 34.76 | 10.16 | 10.70 | 13.37 | 10.16 |
| **Qwen2.5-7B** | 47.59 | 52.94 | 48.66 | 40.64 | 40.64 | 32.62 | 42.25 | 46.52 | 16.58 | 11.23 | 11.23 | 9.09 |
| **Qwen2.5-14B** | 60.43 | 68.45 | 59.89 | 58.82 | 52.41 | 45.99 | 58.29 | 63.10 | 20.32 | 20.32 | 14.44 | 18.72 |
| **Qwen2.5-72B** | 70.05 | 74.33 | 70.05 | 68.45 | 65.78 | 61.50 | 74.33 | 68.45 | 34.76 | 17.11 | 15.51 | 22.99 |
| **QwQ-32B** | 66.84 | 74.33 | 74.33 | 72.19 | 67.38 | 62.57 | 74.87 | 73.26 | 47.06 | 19.79 | 15.51 | 21.93 |
| **Qwen3-1.7B** | 29.95 | 35.83 | 30.48 | 22.46 | 20.86 | 24.60 | 24.60 | 25.67 | 18.72 | 14.97 | 11.76 | 14.44 |
| **Qwen3-4B** | 54.55 | 59.36 | 54.01 | 49.73 | 42.78 | 37.97 | 49.20 | 51.34 | 11.23 | 11.76 | 13.37 | 9.63 |
| **Qwen3-4B-thinking** | 57.22 | 64.71 | 63.64 | 60.43 | 56.15 | 49.73 | 64.17 | 62.03 | 15.51 | 11.23 | 13.37 | 10.70 |
| **Qwen3-8B** | 59.36 | 68.45 | 55.08 | 52.41 | 49.73 | 45.45 | 58.29 | 55.61 | 19.79 | 9.63 | 12.83 | 9.63 |
| **Qwen3-8B-thinking** | 59.89 | 74.33 | 67.91 | 63.64 | 66.84 | 66.31 | 72.19 | 68.98 | 25.13 | 14.44 | 11.23 | 12.83 |
| **Qwen3-14B** | 65.78 | 70.05 | 65.78 | 66.31 | 59.89 | 59.36 | 66.84 | 65.78 | 24.06 | 12.30 | 11.23 | 15.51 |
| **Qwen3-14B-thinking** | 64.71 | 74.33 | 72.73 | 71.66 | 70.05 | 65.78 | 72.73 | 72.19 | 42.78 | 12.30 | 16.04 | 19.79 |
| **Baichuan-M2-32B** | 68.98 | 73.26 | 68.98 | 66.84 | 64.17 | 56.15 | 67.38 | 73.80 | 17.11 | 14.44 | 18.18 | 18.18 |
| **Bio-Medical-LLaMA-3-8B** | 30.48 | 54.01 | 37.43 | 39.57 | 27.81 | 21.93 | 35.29 | 37.97 | 25.13 | 18.18 | 12.30 | 17.11 |
| **MediPhi** | 16.58 | 42.25 | 28.34 | 29.41 | 12.30 | 10.16 | 31.55 | 32.09 | 12.30 | 8.56 | 9.63 | 11.76 |
| **MedGemma-4B** | 33.16 | 48.13 | 37.43 | 37.43 | 31.55 | 28.34 | 35.29 | 37.43 | 21.93 | 14.44 | 10.16 | 13.90 |
| **MedGemma-27B** | 67.91 | 74.33 | 74.87 | 74.87 | 70.59 | 68.45 | 72.19 | 76.47 | 61.50 | 11.76 | 30.48 | 43.32 |
| **MedReason-8B** | 40.64 | 50.80 | 21.93 | 22.46 | 36.90 | 27.81 | 19.25 | 26.20 | 13.90 | 10.70 | 9.09 | 8.02 |
| **HuatuoGPT-o1-7B** | 51.34 | 54.55 | 55.08 | 49.73 | 49.20 | 42.25 | 58.82 | 54.01 | 10.16 | 12.83 | 7.49 | 9.63 |
| **HuatuoGPT-o1-8B** | 47.59 | 55.61 | 45.45 | 48.13 | 43.85 | 36.36 | 48.66 | 45.99 | 27.81 | 11.76 | 14.97 | 11.23 |
| **HuatuoGPT-o1-70B** | 59.36 | 72.73 | 67.91 | 69.52 | 65.24 | 64.17 | 70.05 | 71.12 | 58.29 | 25.13 | 31.55 | 36.90 |
| **HuatuoGPT-o1-72B** | 72.19 | 72.73 | 71.66 | 65.24 | 71.12 | 65.78 | 74.33 | 75.40 | 41.18 | 22.46 | 26.20 | 14.44 |
| **OpenBioLLM-8B** | 11.76 | 27.81 | 17.11 | 17.65 | 12.83 | 10.70 | 18.72 | 12.83 | 9.09 | 10.16 | 11.76 | 9.63 |
| **OpenBioLLM-70B** | 28.88 | 61.50 | 58.29 | 40.11 | 27.27 | 23.53 | 47.06 | 58.29 | 24.60 | 10.70 | 11.23 | 17.65 |

**Supplementary Tab. 122:** Zero-Shot performance evaluation of 56 LLMs on MMLU-Pro (Run 1).

| LLMs | Chinese | English | French | German | Japanese | Korean | Portuguese | Spanish | Swahili | Wolof | Yoruba | Zulu |
|---|---|---|---|---|---|---|---|---|---|---|---|---|
| **Proprietary LLMs** | | | | | | | | | | | | |
| **Claude-3.5-Haiku** | 59.36 | 68.45 | 71.66 | 63.10 | 59.89 | 55.61 | 68.45 | 72.73 | 36.36 | 17.11 | 20.86 | 29.95 |
| **Claude-4.0-Sonnet** | 77.01 | 83.42 | 82.35 | 78.07 | 77.54 | 75.94 | 79.68 | 80.75 | 70.05 | 33.16 | 56.68 | 60.96 |
| **Gemini-2.5-Flash** | 74.87 | 75.94 | 78.07 | 77.01 | 75.40 | 79.14 | 77.54 | 79.14 | 76.47 | 65.24 | 68.45 | 68.98 |
| **GPT-4o-mini** | 57.75 | 72.19 | 67.38 | 67.38 | 59.36 | 58.29 | 67.38 | 68.45 | 47.06 | 16.58 | 28.88 | 34.22 |
| **GPT-4o** | 72.19 | 75.40 | 78.61 | 73.26 | 73.80 | 68.45 | 78.61 | 77.01 | 68.98 | 23.53 | 43.85 | 55.08 |
| **GPT-4.1-nano** | 58.29 | 74.33 | 65.78 | 67.91 | 54.01 | 50.80 | 65.78 | 68.98 | 34.76 | 11.76 | 19.25 | 32.09 |
| **GPT-4.1-mini** | 70.05 | 78.07 | 72.73 | 77.01 | 72.19 | 65.24 | 75.94 | 75.40 | 60.43 | 15.51 | 43.85 | 57.22 |
| **GPT-4.1** | 74.33 | 76.47 | 77.01 | 74.33 | 75.40 | 72.19 | 78.61 | 79.14 | 73.26 | 30.48 | 45.45 | 61.50 |
| **GPT-5-nano** | 45.99 | 65.78 | 60.43 | 59.36 | 55.08 | 35.83 | 55.61 | 57.75 | 37.43 | 11.76 | 22.46 | 26.20 |
| **GPT-5-mini** | 72.19 | 75.94 | 73.26 | 70.59 | 72.73 | 66.31 | 74.87 | 75.40 | 60.96 | 19.25 | 44.92 | 57.22 |
| **GPT-5** | 75.40 | 77.01 | 77.54 | 74.33 | 76.47 | 67.38 | 79.14 | 78.61 | 71.66 | 35.29 | 57.22 | 63.64 |
| **o4-mini** | 77.54 | 79.68 | 80.21 | 73.26 | 76.47 | 71.66 | 80.75 | 77.01 | 74.33 | 28.34 | 58.29 | 66.31 |
| **Open-Weight LLMs** | | | | | | | | | | | | |
| **DeepSeek-V3** | 73.80 | 79.14 | 75.40 | 73.80 | 75.94 | 69.52 | 78.07 | 74.33 | 68.45 | 28.88 | 40.11 | 49.20 |
| **DeepSeek-R1** | 71.12 | 78.61 | 78.07 | 76.47 | 77.54 | 61.50 | 79.68 | 78.61 | 74.87 | 45.99 | 56.15 | 60.96 |
| **DeepSeek-R1-Qwen3-8B** | 61.50 | 67.38 | 58.82 | 63.64 | 56.68 | 57.22 | 60.96 | 58.29 | 19.25 | 9.09 | 12.30 | 11.76 |
| **Gemma-3-4B** | 31.55 | 40.64 | 30.48 | 35.83 | 28.88 | 21.39 | 31.02 | 23.53 | 21.93 | 13.37 | 9.63 | 12.30 |
| **Gemma-3-12B** | 48.66 | 61.50 | 59.89 | 57.75 | 49.73 | 51.87 | 56.15 | 57.75 | 50.27 | 8.56 | 24.06 | 33.16 |
| **Gemma-3-27B** | 65.24 | 69.52 | 68.45 | 64.17 | 59.36 | 57.75 | 70.05 | 67.91 | 54.01 | 12.30 | 26.20 | 43.32 |
| **gpt-oss-20B** | 66.31 | 67.91 | 69.52 | 67.38 | 64.17 | 64.17 | 67.91 | 70.59 | 49.73 | 24.06 | 46.52 | 47.06 |
| **gpt-oss-120B** | 69.52 | 77.54 | 78.07 | 73.80 | 74.33 | 67.91 | 77.01 | 76.47 | 60.96 | 33.69 | 53.48 | 57.22 |
| **LLaMA-3.1-8B** | 35.29 | 54.01 | 39.04 | 39.57 | 22.99 | 16.58 | 37.43 | 40.64 | 21.93 | 14.97 | 10.16 | 7.49 |
| **LLaMA-3.1-70B** | 60.96 | 72.19 | 66.31 | 68.98 | 61.50 | 52.41 | 64.71 | 67.38 | 52.41 | 20.32 | 24.60 | 25.67 |
| **LLaMA-3.2-3B** | 29.41 | 40.64 | 28.34 | 31.55 | 19.25 | 15.51 | 23.53 | 26.20 | 18.18 | 11.76 | 7.49 | 10.70 |
| **LLaMA-3.3-70B** | 55.08 | 76.47 | 72.19 | 71.66 | 52.94 | 62.03 | 71.66 | 72.19 | 56.15 | 25.67 | 25.67 | 32.09 |
| **LLaMA-4-Scout** | 71.12 | 75.40 | 73.26 | 70.05 | 64.17 | 68.45 | 73.26 | 72.73 | 64.71 | 28.34 | 29.41 | 55.08 |
| **LLaMA-4-Maverick** | 74.33 | 79.68 | 76.47 | 74.87 | 74.87 | 73.26 | 79.14 | 75.94 | 72.73 | 37.43 | 53.48 | 58.29 |
| **Mistral-7B-v0.3** | 13.37 | 25.67 | 12.83 | 21.39 | 11.23 | 13.90 | 17.65 | 15.51 | 6.95 | 9.09 | 10.70 | 9.09 |
| **Mistral-Small-3.1-24B** | 59.36 | 71.66 | 62.03 | 62.03 | 58.29 | 43.85 | 67.38 | 64.17 | 25.13 | 11.23 | 8.02 | 7.49 |
| **Phi-4-mini** | 25.67 | 51.34 | 33.16 | 35.83 | 17.11 | 12.83 | 31.55 | 25.13 | 16.04 | 8.02 | 11.76 | 11.76 |
| **Phi-4-mini-Reasoning** | 19.79 | 63.10 | 44.92 | 42.78 | 22.99 | 7.49 | 35.29 | 33.69 | 15.51 | 13.37 | 12.30 | 14.97 |
| **Phi-4** | 49.20 | 71.12 | 64.17 | 68.98 | 54.01 | 51.87 | 67.38 | 66.31 | 30.48 | 11.76 | 16.58 | 19.79 |
| **Phi-4-Reasoning** | 66.84 | 72.73 | 76.47 | 70.59 | 69.52 | 63.10 | 66.31 | 74.33 | 57.75 | 17.11 | 34.76 | 24.06 |
| **Qwen2.5-3B** | 37.43 | 36.90 | 32.09 | 27.27 | 27.27 | 17.11 | 32.09 | 33.16 | 12.30 | 8.02 | 12.83 | 5.35 |
| **Qwen2.5-7B** | 44.39 | 59.89 | 51.87 | 43.32 | 40.11 | 32.09 | 45.99 | 45.45 | 16.04 | 9.09 | 13.37 | 8.56 |
| **Qwen2.5-14B** | 57.75 | 66.84 | 58.29 | 55.08 | 52.94 | 53.48 | 62.03 | 56.68 | 22.46 | 17.11 | 12.83 | 18.18 |
| **Qwen2.5-72B** | 69.52 | 75.40 | 68.98 | 69.52 | 66.31 | 61.50 | 70.05 | 70.59 | 33.16 | 18.18 | 15.51 | 22.46 |
| **QwQ-32B** | 70.05 | 75.40 | 75.40 | 71.66 | 67.91 | 67.91 | 75.40 | 73.26 | 49.73 | 14.97 | 21.39 | 20.32 |
| **Qwen3-1.7B** | 30.48 | 36.36 | 28.34 | 21.93 | 20.86 | 18.72 | 21.39 | 25.67 | 12.83 | 9.63 | 11.76 | 14.44 |
| **Qwen3-4B** | 55.08 | 60.43 | 52.41 | 51.87 | 40.64 | 39.04 | 49.73 | 44.39 | 11.23 | 11.23 | 8.56 | 12.83 |
| **Qwen3-4B-thinking** | 58.82 | 62.57 | 64.17 | 58.82 | 57.75 | 55.61 | 63.64 | 61.50 | 15.51 | 11.76 | 11.23 | 12.30 |
| **Qwen3-8B** | 59.89 | 67.91 | 53.48 | 54.55 | 52.41 | 43.32 | 56.15 | 60.43 | 20.32 | 11.23 | 14.97 | 16.04 |
| **Qwen3-8B-thinking** | 63.10 | 71.12 | 66.84 | 66.31 | 64.17 | 63.64 | 66.84 | 67.91 | 28.88 | 13.90 | 16.58 | 10.70 |
| **Qwen3-14B** | 65.24 | 68.45 | 67.38 | 64.17 | 58.29 | 56.15 | 71.66 | 66.31 | 25.13 | 14.44 | 12.30 | 21.39 |
| **Qwen3-14B-thinking** | 67.91 | 73.80 | 73.80 | 71.66 | 68.45 | 66.84 | 68.45 | 73.26 | 44.39 | 14.44 | 14.97 | 14.44 |
| **Baichuan-M2-32B** | 68.98 | 77.54 | 70.05 | 65.78 | 61.50 | 54.01 | 67.38 | 65.78 | 16.04 | 16.04 | 13.37 | 14.97 |
| **Bio-Medical-LLaMA-3-8B** | 30.48 | 55.61 | 36.36 | 39.57 | 29.41 | 25.13 | 35.29 | 39.57 | 26.74 | 19.79 | 13.37 | 14.97 |
| **MediPhi** | 20.86 | 42.78 | 33.16 | 31.55 | 12.30 | 13.90 | 32.09 | 29.95 | 4.81 | 9.63 | 12.83 | 10.70 |
| **MedGemma-4B** | 34.76 | 45.45 | 42.78 | 37.97 | 31.02 | 31.55 | 34.76 | 35.29 | 22.46 | 11.23 | 15.51 | 14.44 |
| **MedGemma-27B** | 71.12 | 77.54 | 74.33 | 72.73 | 66.84 | 68.98 | 75.40 | 74.33 | 57.75 | 14.97 | 28.88 | 48.13 |
| **MedReason-8B** | 38.50 | 47.06 | 20.32 | 22.99 | 36.90 | 26.20 | 20.86 | 27.27 | 16.58 | 12.30 | 9.63 | 11.76 |
| **HuatuoGPT-o1-7B** | 54.55 | 55.61 | 60.96 | 57.22 | 47.06 | 41.71 | 59.89 | 59.89 | 11.23 | 11.23 | 9.63 | 10.70 |
| **HuatuoGPT-o1-8B** | 45.45 | 52.94 | 49.20 | 53.48 | 41.18 | 34.76 | 47.59 | 44.39 | 27.81 | 13.37 | 11.23 | 11.76 |
| **HuatuoGPT-o1-70B** | 63.64 | 71.66 | 67.38 | 69.52 | 60.96 | 64.71 | 70.59 | 72.19 | 56.68 | 24.60 | 29.95 | 40.64 |
| **HuatuoGPT-o1-72B** | 68.45 | 74.87 | 73.80 | 67.38 | 70.59 | 66.31 | 75.94 | 75.40 | 46.52 | 17.65 | 22.46 | 16.58 |
| **OpenBioLLM-8B** | 13.37 | 19.25 | 17.65 | 17.11 | 9.63 | 12.30 | 16.04 | 15.51 | 12.30 | 9.09 | 8.02 | 8.56 |
| **OpenBioLLM-70B** | 21.39 | 63.64 | 57.22 | 40.11 | 20.32 | 24.60 | 41.18 | 59.89 | 26.74 | 6.95 | 8.56 | 13.37 |

**Supplementary Tab. 123:** Zero-Shot performance evaluation of 56 LLMs on MMLU-Pro (Run 2).

| LLMs | Chinese | English | French | German | Japanese | Korean | Portuguese | Spanish | Swahili | Wolof | Yoruba | Zulu |
|---|---|---|---|---|---|---|---|---|---|---|---|---|
| **Proprietary LLMs** | | | | | | | | | | | | |
| **Claude-3.5-Haiku** | 59.36 | 68.45 | 71.66 | 63.10 | 59.89 | 55.61 | 68.45 | 72.73 | 36.36 | 15.51 | 21.39 | 29.95 |
| **Claude-4.0-Sonnet** | 77.54 | 82.35 | 81.28 | 78.07 | 79.68 | 75.94 | 79.68 | 81.82 | 68.98 | 36.36 | 56.15 | 63.10 |
| **Gemini-2.5-Flash** | 75.40 | 74.87 | 76.47 | 76.47 | 75.40 | 76.47 | 79.68 | 78.07 | 75.40 | 61.50 | 67.38 | 66.84 |
| **GPT-4o-mini** | 56.68 | 70.59 | 67.91 | 67.38 | 63.10 | 59.89 | 66.84 | 68.98 | 46.52 | 17.11 | 25.67 | 33.16 |
| **GPT-4o** | 72.19 | 77.01 | 77.01 | 72.73 | 72.19 | 68.45 | 77.01 | 78.61 | 66.84 | 24.60 | 42.25 | 56.68 |
| **GPT-4.1-nano** | 60.96 | 70.59 | 67.38 | 66.31 | 57.75 | 45.45 | 66.84 | 69.52 | 35.29 | 16.58 | 20.86 | 28.34 |
| **GPT-4.1-mini** | 67.91 | 77.01 | 73.26 | 74.33 | 70.59 | 66.31 | 74.87 | 75.40 | 63.64 | 13.90 | 42.78 | 51.34 |
| **GPT-4.1** | 75.94 | 75.94 | 78.61 | 74.87 | 74.33 | 70.05 | 81.28 | 78.07 | 72.19 | 34.22 | 49.20 | 63.10 |
| **GPT-5-nano** | 50.80 | 62.57 | 55.61 | 63.64 | 54.01 | 40.11 | 57.22 | 59.89 | 39.57 | 13.37 | 20.86 | 29.41 |
| **GPT-5-mini** | 71.66 | 76.47 | 76.47 | 71.66 | 70.59 | 65.24 | 73.26 | 78.07 | 60.43 | 21.39 | 42.25 | 56.68 |
| **GPT-5** | 74.87 | 78.07 | 75.40 | 73.26 | 72.73 | 71.12 | 78.61 | 80.21 | 70.59 | 34.22 | 56.68 | 66.31 |
| **o4-mini** | 73.80 | 79.68 | 80.21 | 73.26 | 75.94 | 72.73 | 79.68 | 78.07 | 71.12 | 29.95 | 61.50 | 67.91 |
| **Open-Weight LLMs** | | | | | | | | | | | | |
| **DeepSeek-V3** | 75.40 | 80.21 | 76.47 | 77.54 | 73.80 | 67.38 | 79.14 | 74.87 | 66.84 | 28.88 | 36.90 | 45.99 |
| **DeepSeek-R1** | 71.12 | 78.07 | 77.01 | 76.47 | 77.01 | 65.78 | 77.54 | 77.54 | 75.40 | 45.45 | 56.15 | 62.57 |
| **DeepSeek-R1-Qwen3-8B** | 61.50 | 67.38 | 52.94 | 62.03 | 54.01 | 56.68 | 61.50 | 59.36 | 19.25 | 10.70 | 11.76 | 8.56 |
| **Gemma-3-4B** | 29.41 | 36.90 | 32.62 | 35.29 | 26.20 | 23.53 | 29.95 | 27.81 | 22.46 | 10.16 | 10.70 | 13.90 |
| **Gemma-3-12B** | 48.66 | 61.50 | 59.36 | 60.43 | 52.94 | 51.34 | 57.22 | 59.36 | 47.59 | 9.63 | 25.67 | 31.02 |
| **Gemma-3-27B** | 62.03 | 65.78 | 71.66 | 64.71 | 58.82 | 59.36 | 73.26 | 66.31 | 51.87 | 14.44 | 27.81 | 43.32 |
| **gpt-oss-20B** | 63.64 | 65.78 | 66.31 | 66.84 | 63.64 | 60.43 | 69.52 | 67.91 | 50.27 | 21.93 | 38.50 | 45.45 |
| **gpt-oss-120B** | 70.05 | 78.07 | 76.47 | 75.94 | 74.33 | 68.45 | 76.47 | 72.19 | 62.03 | 40.11 | 51.87 | 57.22 |
| **LLaMA-3.1-8B** | 36.90 | 55.08 | 37.43 | 36.90 | 23.53 | 21.39 | 37.97 | 37.43 | 23.53 | 13.90 | 10.70 | 9.09 |
| **LLaMA-3.1-70B** | 57.22 | 71.12 | 66.31 | 70.05 | 58.29 | 50.27 | 68.98 | 69.52 | 52.94 | 21.93 | 22.46 | 33.16 |
| **LLaMA-3.2-3B** | 28.88 | 37.97 | 29.95 | 31.02 | 16.04 | 12.30 | 24.60 | 20.86 | 14.44 | 13.90 | 4.81 | 10.16 |
| **LLaMA-3.3-70B** | 51.87 | 75.94 | 74.33 | 72.19 | 54.01 | 59.89 | 71.66 | 73.80 | 58.29 | 25.67 | 27.27 | 35.83 |
| **LLaMA-4-Scout** | 68.98 | 75.40 | 73.26 | 70.05 | 65.24 | 66.31 | 73.26 | 73.26 | 64.17 | 31.55 | 29.41 | 52.94 |
| **LLaMA-4-Maverick** | 74.33 | 81.28 | 79.14 | 77.01 | 73.80 | 71.66 | 76.47 | 78.07 | 71.12 | 37.97 | 52.94 | 58.29 |
| **Mistral-7B-v0.3** | 15.51 | 16.04 | 13.37 | 17.11 | 13.90 | 9.09 | 17.65 | 18.72 | 8.56 | 9.63 | 12.30 | 10.16 |
| **Mistral-Small-3.1-24B** | 55.61 | 70.59 | 71.12 | 63.10 | 59.36 | 50.27 | 64.71 | 68.45 | 29.95 | 13.90 | 11.23 | 12.83 |
| **Phi-4-mini** | 27.27 | 43.85 | 33.69 | 29.95 | 17.65 | 18.72 | 23.53 | 29.41 | 15.51 | 12.30 | 11.23 | 11.23 |
| **Phi-4-mini-Reasoning** | 22.46 | 57.22 | 47.59 | 41.18 | 17.11 | 12.83 | 35.29 | 32.09 | 18.72 | 19.79 | 8.02 | 12.30 |
| **Phi-4** | 49.20 | 72.19 | 64.17 | 66.31 | 58.82 | 43.85 | 68.98 | 65.24 | 34.22 | 14.44 | 18.18 | 16.58 |
| **Phi-4-Reasoning** | 68.45 | 72.19 | 74.33 | 70.59 | 71.12 | 64.71 | 70.05 | 74.33 | 55.08 | 19.25 | 32.62 | 27.81 |
| **Qwen2.5-3B** | 39.04 | 39.57 | 29.41 | 32.09 | 23.53 | 24.06 | 33.16 | 34.22 | 11.76 | 8.56 | 13.90 | 16.58 |
| **Qwen2.5-7B** | 45.99 | 60.96 | 50.80 | 45.45 | 38.50 | 32.62 | 45.45 | 48.13 | 18.18 | 10.70 | 10.70 | 11.23 |
| **Qwen2.5-14B** | 60.96 | 70.05 | 53.48 | 59.36 | 52.94 | 48.13 | 59.36 | 56.68 | 21.39 | 18.72 | 15.51 | 21.93 |
| **Qwen2.5-72B** | 69.52 | 73.26 | 68.45 | 67.38 | 67.91 | 60.96 | 71.12 | 68.45 | 37.97 | 19.25 | 14.97 | 19.79 |
| **QwQ-32B** | 71.66 | 72.73 | 73.80 | 74.33 | 67.91 | 63.10 | 78.07 | 77.01 | 47.06 | 21.39 | 20.32 | 18.72 |
| **Qwen3-1.7B** | 29.41 | 41.71 | 31.55 | 19.79 | 21.39 | 19.79 | 24.06 | 28.88 | 13.37 | 11.23 | 11.76 | 11.23 |
| **Qwen3-4B** | 51.34 | 56.15 | 50.27 | 48.13 | 43.32 | 35.83 | 49.20 | 50.27 | 15.51 | 11.76 | 10.70 | 9.09 |
| **Qwen3-4B-thinking** | 58.82 | 65.78 | 66.31 | 59.89 | 58.82 | 51.34 | 58.82 | 60.96 | 12.30 | 12.83 | 12.30 | 7.49 |
| **Qwen3-8B** | 55.08 | 65.78 | 54.55 | 56.68 | 51.87 | 45.99 | 54.01 | 54.01 | 23.53 | 12.30 | 12.83 | 11.23 |
| **Qwen3-8B-thinking** | 59.89 | 72.19 | 72.19 | 65.24 | 63.10 | 61.50 | 67.38 | 67.38 | 21.39 | 16.04 | 14.97 | 14.44 |
| **Qwen3-14B** | 63.10 | 74.87 | 64.71 | 67.91 | 60.43 | 59.36 | 69.52 | 67.91 | 24.06 | 8.02 | 12.83 | 15.51 |
| **Qwen3-14B-thinking** | 65.78 | 75.94 | 73.26 | 70.59 | 72.73 | 68.98 | 73.26 | 68.98 | 48.13 | 10.16 | 17.11 | 14.44 |
| **Baichuan-M2-32B** | 71.12 | 73.80 | 67.38 | 68.45 | 64.17 | 57.22 | 67.38 | 62.03 | 18.18 | 21.39 | 20.32 | 15.51 |
| **Bio-Medical-LLaMA-3-8B** | 31.55 | 56.68 | 34.22 | 39.04 | 26.20 | 21.93 | 34.76 | 38.50 | 23.53 | 17.65 | 14.44 | 16.04 |
| **MediPhi** | 17.65 | 38.50 | 32.09 | 26.74 | 11.23 | 9.09 | 32.62 | 31.02 | 9.63 | 12.30 | 11.76 | 10.70 |
| **MedGemma-4B** | 34.22 | 47.59 | 44.92 | 34.22 | 33.16 | 32.09 | 35.29 | 36.36 | 26.74 | 16.58 | 11.23 | 13.90 |
| **MedGemma-27B** | 69.52 | 75.40 | 73.80 | 72.73 | 70.05 | 66.31 | 71.12 | 73.80 | 62.03 | 16.58 | 29.95 | 51.34 |
| **MedReason-8B** | 37.43 | 41.71 | 19.79 | 20.86 | 35.83 | 25.67 | 25.67 | 29.95 | 13.90 | 11.76 | 9.09 | 10.70 |
| **HuatuoGPT-o1-7B** | 55.61 | 59.36 | 57.22 | 54.01 | 49.20 | 41.71 | 59.36 | 60.96 | 9.63 | 11.23 | 7.49 | 10.70 |
| **HuatuoGPT-o1-8B** | 48.66 | 55.08 | 52.94 | 45.99 | 40.64 | 33.69 | 49.73 | 48.13 | 28.88 | 13.37 | 8.56 | 10.16 |
| **HuatuoGPT-o1-70B** | 57.75 | 72.19 | 69.52 | 67.91 | 59.89 | 66.84 | 73.26 | 70.59 | 56.15 | 26.20 | 27.81 | 37.43 |
| **HuatuoGPT-o1-72B** | 69.52 | 77.01 | 74.87 | 70.59 | 72.19 | 65.24 | 72.73 | 73.26 | 45.45 | 18.18 | 25.67 | 13.90 |
| **OpenBioLLM-8B** | 13.90 | 20.32 | 18.18 | 13.90 | 10.16 | 6.42 | 17.65 | 14.44 | 9.09 | 9.63 | 11.23 | 8.56 |
| **OpenBioLLM-70B** | 27.81 | 63.64 | 57.75 | 37.97 | 21.93 | 18.18 | 37.97 | 59.89 | 29.41 | 9.63 | 10.70 | 13.90 |

**Supplementary Tab. 124:** Zero-Shot performance evaluation of 56 LLMs on MMLU-Pro (Run 3).

| LLMs | Chinese | English | French | German | Japanese | Korean | Portuguese | Spanish | Swahili | Wolof | Yoruba | Zulu |
|---|---|---|---|---|---|---|---|---|---|---|---|---|
| **Proprietary LLMs** | | | | | | | | | | | | |
| **Claude-3.5-Haiku** | 59.36 | 68.45 | 71.66 | 63.10 | 59.89 | 55.61 | 68.45 | 72.73 | 36.36 | 13.37 | 20.86 | 29.95 |
| **Claude-4.0-Sonnet** | 77.54 | 83.42 | 79.68 | 76.47 | 78.61 | 76.47 | 79.14 | 79.68 | 68.98 | 36.36 | 57.75 | 61.50 |
| **Gemini-2.5-Flash** | 77.54 | 75.94 | 75.94 | 80.21 | 75.40 | 74.87 | 77.54 | 77.01 | 74.87 | 64.71 | 70.59 | 68.98 |
| **GPT-4o-mini** | 58.29 | 71.12 | 68.45 | 65.24 | 58.29 | 55.61 | 66.31 | 68.45 | 46.52 | 17.11 | 31.02 | 34.22 |
| **GPT-4o** | 73.26 | 73.26 | 73.80 | 74.87 | 71.66 | 68.98 | 77.01 | 75.40 | 66.84 | 26.74 | 43.32 | 52.94 |
| **GPT-4.1-nano** | 56.68 | 71.66 | 64.17 | 65.24 | 52.41 | 44.39 | 67.38 | 65.78 | 40.11 | 13.90 | 22.46 | 29.95 |
| **GPT-4.1-mini** | 68.98 | 77.01 | 71.66 | 74.33 | 70.59 | 66.31 | 73.26 | 72.19 | 59.36 | 14.44 | 44.92 | 51.87 |
| **GPT-4.1** | 74.87 | 78.61 | 77.54 | 78.07 | 72.73 | 73.26 | 80.21 | 78.61 | 74.87 | 34.76 | 44.39 | 62.57 |
| **GPT-5-nano** | 45.45 | 67.91 | 58.29 | 58.29 | 51.34 | 38.50 | 59.89 | 60.96 | 40.11 | 13.37 | 17.65 | 26.20 |
| **GPT-5-mini** | 72.73 | 76.47 | 75.40 | 77.01 | 68.98 | 60.96 | 75.40 | 75.94 | 58.82 | 21.93 | 44.39 | 54.01 |
| **GPT-5** | 72.73 | 75.94 | 77.54 | 75.40 | 74.33 | 66.84 | 80.21 | 79.68 | 70.59 | 40.11 | 52.94 | 59.89 |
| **o4-mini** | 74.33 | 79.68 | 79.14 | 74.33 | 74.33 | 69.52 | 80.75 | 79.14 | 71.66 | 31.02 | 59.36 | 68.98 |
| **Open-Weight LLMs** | | | | | | | | | | | | |
| **DeepSeek-V3** | 73.80 | 80.75 | 75.94 | 75.94 | 75.94 | 69.52 | 78.07 | 73.80 | 66.31 | 31.02 | 35.83 | 48.66 |
| **DeepSeek-R1** | 69.52 | 77.54 | 77.01 | 77.54 | 76.47 | 63.10 | 78.07 | 77.54 | 74.87 | 49.20 | 61.50 | 59.89 |
| **DeepSeek-R1-Qwen3-8B** | 63.10 | 68.98 | 59.36 | 57.22 | 54.55 | 59.36 | 66.84 | 62.03 | 19.79 | 14.97 | 8.56 | 12.83 |
| **Gemma-3-4B** | 32.62 | 40.64 | 30.48 | 33.16 | 27.27 | 25.67 | 32.62 | 31.55 | 21.93 | 13.90 | 13.37 | 10.16 |
| **Gemma-3-12B** | 54.55 | 60.96 | 59.89 | 55.61 | 51.34 | 52.41 | 61.50 | 59.89 | 43.85 | 8.02 | 22.99 | 32.09 |
| **Gemma-3-27B** | 56.68 | 65.78 | 70.05 | 66.31 | 57.75 | 58.82 | 70.05 | 72.19 | 57.22 | 13.90 | 28.34 | 45.45 |
| **gpt-oss-20B** | 60.43 | 66.84 | 68.98 | 67.38 | 63.64 | 64.71 | 70.05 | 72.19 | 49.73 | 22.46 | 43.32 | 49.73 |
| **gpt-oss-120B** | 70.59 | 78.07 | 77.54 | 73.26 | 71.12 | 68.45 | 75.94 | 77.01 | 64.17 | 36.36 | 58.29 | 56.68 |
| **LLaMA-3.1-8B** | 32.62 | 56.15 | 41.18 | 37.97 | 26.74 | 17.11 | 38.50 | 42.78 | 22.46 | 18.72 | 13.37 | 6.95 |
| **LLaMA-3.1-70B** | 57.22 | 71.12 | 67.91 | 66.31 | 60.96 | 52.41 | 66.31 | 72.73 | 47.59 | 19.25 | 19.25 | 27.27 |
| **LLaMA-3.2-3B** | 31.55 | 44.39 | 27.81 | 28.88 | 20.86 | 12.83 | 22.46 | 28.88 | 16.58 | 12.83 | 8.56 | 13.37 |
| **LLaMA-3.3-70B** | 51.34 | 74.87 | 74.87 | 73.26 | 48.13 | 59.36 | 71.66 | 73.26 | 57.22 | 23.53 | 25.67 | 35.29 |
| **LLaMA-4-Scout** | 67.91 | 75.40 | 71.66 | 71.12 | 66.84 | 66.84 | 73.26 | 73.80 | 59.36 | 28.88 | 28.88 | 54.55 |
| **LLaMA-4-Maverick** | 73.80 | 81.28 | 77.01 | 76.47 | 77.01 | 71.66 | 76.47 | 78.61 | 73.80 | 36.36 | 53.48 | 60.96 |
| **Mistral-7B-v0.3** | 13.37 | 23.53 | 12.83 | 17.65 | 12.83 | 11.23 | 18.18 | 17.11 | 9.63 | 11.23 | 8.56 | 8.56 |
| **Mistral-Small-3.1-24B** | 57.75 | 72.19 | 63.64 | 65.78 | 57.75 | 52.41 | 62.57 | 64.71 | 20.32 | 10.70 | 11.76 | 13.90 |
| **Phi-4-mini** | 22.46 | 47.06 | 29.41 | 32.62 | 24.06 | 14.97 | 28.88 | 26.20 | 16.58 | 10.16 | 9.63 | 9.63 |
| **Phi-4-mini-Reasoning** | 21.93 | 59.36 | 38.50 | 43.85 | 20.32 | 13.90 | 42.78 | 27.81 | 20.86 | 13.37 | 11.76 | 13.90 |
| **Phi-4** | 53.48 | 72.73 | 68.98 | 68.45 | 56.68 | 49.73 | 68.45 | 65.78 | 35.83 | 14.97 | 19.25 | 20.32 |
| **Phi-4-Reasoning** | 66.31 | 76.47 | 73.26 | 73.26 | 72.19 | 66.84 | 68.45 | 74.87 | 53.48 | 15.51 | 33.16 | 25.13 |
| **Qwen2.5-3B** | 36.36 | 41.71 | 28.88 | 26.20 | 22.99 | 23.53 | 28.88 | 29.95 | 11.23 | 12.83 | 7.49 | 8.02 |
| **Qwen2.5-7B** | 48.66 | 60.96 | 49.20 | 43.32 | 41.18 | 27.81 | 48.66 | 44.92 | 18.72 | 11.23 | 11.76 | 11.23 |
| **Qwen2.5-14B** | 62.03 | 68.98 | 58.29 | 55.61 | 52.41 | 50.27 | 64.17 | 63.64 | 18.72 | 13.37 | 12.83 | 19.79 |
| **Qwen2.5-72B** | 70.05 | 73.80 | 70.59 | 66.31 | 67.91 | 64.17 | 72.73 | 68.98 | 33.69 | 19.25 | 14.97 | 20.86 |
| **QwQ-32B** | 70.05 | 74.33 | 73.26 | 72.19 | 69.52 | 63.10 | 76.47 | 75.40 | 44.92 | 18.72 | 19.25 | 22.99 |
| **Qwen3-1.7B** | 30.48 | 37.97 | 26.20 | 27.27 | 18.72 | 23.53 | 25.67 | 22.99 | 15.51 | 9.09 | 12.30 | 13.37 |
| **Qwen3-4B** | 54.01 | 59.36 | 47.59 | 50.80 | 46.52 | 40.64 | 46.52 | 51.87 | 11.76 | 10.16 | 9.09 | 10.70 |
| **Qwen3-4B-thinking** | 57.22 | 62.57 | 65.78 | 61.50 | 56.68 | 55.61 | 60.43 | 62.57 | 15.51 | 11.23 | 17.65 | 6.42 |
| **Qwen3-8B** | 57.75 | 66.31 | 54.55 | 54.01 | 55.61 | 49.73 | 55.61 | 60.96 | 19.25 | 13.37 | 14.44 | 14.97 |
| **Qwen3-8B-thinking** | 64.71 | 70.59 | 68.98 | 69.52 | 66.31 | 60.96 | 65.78 | 68.98 | 24.06 | 10.16 | 12.30 | 13.37 |
| **Qwen3-14B** | 64.71 | 72.19 | 64.71 | 65.78 | 55.08 | 57.75 | 66.84 | 63.10 | 25.13 | 9.63 | 14.44 | 15.51 |
| **Qwen3-14B-thinking** | 67.38 | 78.07 | 72.73 | 73.26 | 72.19 | 65.24 | 69.52 | 71.66 | 44.39 | 13.37 | 19.79 | 16.04 |
| **Baichuan-M2-32B** | 71.66 | 76.47 | 71.12 | 69.52 | 67.91 | 55.08 | 70.05 | 67.91 | 19.25 | 13.90 | 18.72 | 13.37 |
| **Bio-Medical-LLaMA-3-8B** | 28.34 | 54.01 | 38.50 | 38.50 | 28.88 | 21.93 | 35.83 | 39.57 | 27.81 | 21.39 | 13.37 | 16.04 |
| **MediPhi** | 17.65 | 41.71 | 38.50 | 27.27 | 12.30 | 9.09 | 32.09 | 29.95 | 12.83 | 12.83 | 11.23 | 8.56 |
| **MedGemma-4B** | 32.09 | 47.06 | 44.92 | 38.50 | 30.48 | 26.74 | 38.50 | 35.83 | 25.13 | 12.30 | 10.70 | 19.79 |
| **MedGemma-27B** | 65.78 | 72.73 | 75.40 | 75.40 | 70.05 | 66.84 | 74.33 | 73.80 | 63.10 | 18.18 | 29.95 | 43.32 |
| **MedReason-8B** | 34.22 | 49.20 | 20.86 | 22.46 | 36.90 | 27.81 | 24.60 | 26.74 | 12.83 | 11.23 | 9.09 | 11.23 |
| **HuatuoGPT-o1-7B** | 54.55 | 58.82 | 55.61 | 50.27 | 50.80 | 39.04 | 59.89 | 57.22 | 9.09 | 10.70 | 10.70 | 9.63 |
| **HuatuoGPT-o1-8B** | 47.59 | 53.48 | 56.15 | 52.41 | 41.71 | 37.43 | 49.73 | 40.11 | 30.48 | 10.70 | 11.76 | 11.23 |
| **HuatuoGPT-o1-70B** | 60.96 | 73.26 | 71.12 | 71.12 | 59.89 | 63.64 | 67.38 | 69.52 | 56.15 | 22.46 | 31.55 | 41.71 |
| **HuatuoGPT-o1-72B** | 70.05 | 74.87 | 73.26 | 68.98 | 71.66 | 64.17 | 74.87 | 72.19 | 44.39 | 20.86 | 24.06 | 20.32 |
| **OpenBioLLM-8B** | 12.30 | 24.06 | 12.30 | 17.11 | 11.23 | 14.97 | 15.51 | 16.58 | 10.70 | 13.90 | 5.35 | 10.70 |
| **OpenBioLLM-70B** | 29.41 | 60.96 | 52.94 | 39.04 | 25.67 | 21.39 | 45.99 | 62.03 | 25.13 | 8.56 | 11.23 | 18.18 |

**Supplementary Tab. 125:** Zero-Shot performance evaluation of 56 LLMs on MMLU-Pro (Run 4).

| LLMs | Chinese | English | French | German | Japanese | Korean | Portuguese | Spanish | Swahili | Wolof | Yoruba | Zulu |
|---|---|---|---|---|---|---|---|---|---|---|---|---|
| **Proprietary LLMs** | | | | | | | | | | | | |
| **Claude-3.5-Haiku** | 59.36 | 68.45 | 71.66 | 63.10 | 59.89 | 55.61 | 68.45 | 72.19 | 36.36 | 14.97 | 21.39 | 29.41 |
| **Claude-4.0-Sonnet** | 80.21 | 82.35 | 80.21 | 79.14 | 78.61 | 77.01 | 80.21 | 81.82 | 71.66 | 35.83 | 54.55 | 64.17 |
| **Gemini-2.5-Flash** | 76.47 | 77.01 | 77.54 | 76.47 | 75.94 | 77.54 | 80.21 | 79.14 | 75.94 | 62.57 | 68.45 | 67.91 |
| **GPT-4o-mini** | 58.82 | 72.73 | 70.59 | 64.17 | 63.10 | 57.75 | 64.71 | 68.98 | 50.27 | 16.58 | 26.74 | 35.83 |
| **GPT-4o** | 74.33 | 78.07 | 73.80 | 71.12 | 74.33 | 71.12 | 74.87 | 73.80 | 66.84 | 29.41 | 43.32 | 57.75 |
| **GPT-4.1-nano** | 52.94 | 75.40 | 65.24 | 65.78 | 55.08 | 48.66 | 65.24 | 71.12 | 40.64 | 18.72 | 20.32 | 33.16 |
| **GPT-4.1-mini** | 71.12 | 79.14 | 73.80 | 74.87 | 68.98 | 66.31 | 77.01 | 72.19 | 64.17 | 16.04 | 44.92 | 51.34 |
| **GPT-4.1** | 73.26 | 78.61 | 78.07 | 74.87 | 75.40 | 69.52 | 79.68 | 80.21 | 75.40 | 26.74 | 45.45 | 63.64 |
| **GPT-5-nano** | 47.06 | 66.31 | 57.75 | 60.96 | 51.87 | 43.85 | 54.01 | 59.89 | 37.97 | 10.16 | 23.53 | 25.13 |
| **GPT-5-mini** | 73.26 | 77.01 | 70.59 | 73.26 | 73.26 | 64.17 | 74.33 | 74.87 | 65.24 | 19.79 | 43.85 | 54.01 |
| **GPT-5** | 77.01 | 75.94 | 77.54 | 71.12 | 73.26 | 68.98 | 81.28 | 79.14 | 69.52 | 35.83 | 58.82 | 61.50 |
| **o4-mini** | 74.87 | 80.21 | 79.68 | 75.94 | 75.40 | 68.98 | 81.28 | 81.28 | 73.80 | 33.16 | 61.50 | 67.38 |
| **Open-Weight LLMs** | | | | | | | | | | | | |
| **DeepSeek-V3** | 74.87 | 79.68 | 75.94 | 75.40 | 76.47 | 68.98 | 77.54 | 75.94 | 64.17 | 33.16 | 36.36 | 45.45 |
| **DeepSeek-R1** | 69.52 | 78.07 | 79.68 | 75.94 | 76.47 | 66.31 | 77.54 | 75.94 | 75.40 | 44.92 | 54.55 | 61.50 |
| **DeepSeek-R1-Qwen3-8B** | 59.36 | 68.45 | 63.64 | 62.57 | 51.34 | 55.61 | 64.71 | 60.43 | 16.04 | 10.70 | 14.44 | 11.76 |
| **Gemma-3-4B** | 31.02 | 44.39 | 32.09 | 32.62 | 27.81 | 22.99 | 27.27 | 28.88 | 21.39 | 9.63 | 9.63 | 6.42 |
| **Gemma-3-12B** | 51.34 | 60.96 | 57.75 | 60.96 | 56.15 | 55.61 | 60.96 | 59.89 | 44.92 | 12.83 | 22.46 | 32.62 |
| **Gemma-3-27B** | 63.10 | 67.38 | 65.24 | 63.10 | 58.82 | 58.82 | 68.45 | 67.91 | 54.01 | 13.90 | 27.81 | 45.45 |
| **gpt-oss-20B** | 63.10 | 64.17 | 70.05 | 70.05 | 64.17 | 65.24 | 67.91 | 67.91 | 47.06 | 22.46 | 41.18 | 45.45 |
| **gpt-oss-120B** | 70.05 | 74.87 | 74.33 | 74.87 | 72.19 | 72.73 | 75.40 | 77.54 | 59.89 | 32.09 | 52.94 | 59.36 |
| **LLaMA-3.1-8B** | 33.69 | 52.41 | 41.71 | 39.57 | 23.53 | 25.13 | 36.90 | 41.18 | 21.93 | 17.65 | 5.88 | 12.83 |
| **LLaMA-3.1-70B** | 60.43 | 73.80 | 65.24 | 64.71 | 56.15 | 52.41 | 65.78 | 66.31 | 49.73 | 19.79 | 23.53 | 31.55 |
| **LLaMA-3.2-3B** | 28.88 | 40.11 | 24.60 | 33.69 | 23.53 | 11.76 | 21.39 | 26.20 | 14.97 | 7.49 | 12.83 | 9.63 |
| **LLaMA-3.3-70B** | 51.87 | 75.94 | 74.87 | 72.19 | 51.34 | 60.96 | 71.66 | 72.73 | 56.68 | 21.39 | 27.81 | 33.16 |
| **LLaMA-4-Scout** | 69.52 | 72.19 | 74.33 | 70.05 | 67.38 | 67.91 | 75.40 | 71.12 | 61.50 | 27.81 | 29.95 | 53.48 |
| **LLaMA-4-Maverick** | 74.87 | 80.75 | 77.01 | 74.33 | 71.12 | 70.59 | 77.01 | 77.01 | 71.66 | 36.36 | 56.68 | 59.89 |
| **Mistral-7B-v0.3** | 14.97 | 20.32 | 14.97 | 13.90 | 12.83 | 10.16 | 16.04 | 16.04 | 10.70 | 10.16 | 9.63 | 12.83 |
| **Mistral-Small-3.1-24B** | 54.01 | 66.84 | 65.24 | 65.78 | 60.96 | 46.52 | 63.10 | 66.84 | 20.86 | 8.02 | 11.23 | 9.09 |
| **Phi-4-mini** | 28.88 | 51.34 | 35.83 | 29.95 | 25.67 | 19.25 | 24.60 | 27.27 | 16.04 | 11.23 | 11.23 | 9.63 |
| **Phi-4-mini-Reasoning** | 17.65 | 58.82 | 42.25 | 46.52 | 17.11 | 12.83 | 27.81 | 32.62 | 13.90 | 10.70 | 11.23 | 14.44 |
| **Phi-4** | 51.87 | 73.26 | 64.71 | 70.05 | 57.22 | 50.80 | 67.38 | 67.38 | 33.69 | 11.76 | 18.72 | 14.97 |
| **Phi-4-Reasoning** | 67.91 | 75.94 | 73.26 | 72.19 | 72.73 | 66.84 | 68.98 | 72.73 | 56.68 | 14.97 | 31.02 | 24.06 |
| **Qwen2.5-3B** | 35.83 | 41.18 | 34.76 | 30.48 | 25.13 | 20.86 | 30.48 | 32.09 | 11.23 | 10.16 | 11.76 | 8.02 |
| **Qwen2.5-7B** | 51.87 | 59.36 | 46.52 | 40.11 | 43.85 | 33.69 | 45.45 | 47.59 | 16.58 | 12.83 | 14.97 | 10.70 |
| **Qwen2.5-14B** | 56.68 | 66.31 | 59.89 | 59.89 | 51.87 | 48.13 | 60.43 | 59.36 | 19.25 | 14.97 | 14.97 | 18.72 |
| **Qwen2.5-72B** | 70.59 | 72.19 | 70.59 | 69.52 | 66.31 | 59.89 | 69.52 | 70.05 | 37.97 | 17.11 | 16.04 | 21.39 |
| **QwQ-32B** | 68.45 | 77.01 | 72.73 | 74.33 | 69.52 | 65.24 | 75.94 | 75.94 | 44.39 | 18.18 | 19.25 | 20.32 |
| **Qwen3-1.7B** | 29.41 | 40.11 | 27.27 | 25.13 | 20.86 | 18.72 | 25.67 | 25.67 | 13.37 | 12.30 | 12.30 | 14.44 |
| **Qwen3-4B** | 50.80 | 60.43 | 49.73 | 52.41 | 44.92 | 35.29 | 47.06 | 53.48 | 12.83 | 11.23 | 13.37 | 11.23 |
| **Qwen3-4B-thinking** | 58.82 | 63.10 | 62.03 | 63.10 | 57.75 | 55.61 | 60.43 | 60.96 | 17.65 | 13.37 | 9.63 | 12.30 |
| **Qwen3-8B** | 60.43 | 66.31 | 55.61 | 56.15 | 51.87 | 43.85 | 58.29 | 58.29 | 15.51 | 11.76 | 10.16 | 8.56 |
| **Qwen3-8B-thinking** | 64.71 | 73.26 | 68.45 | 67.91 | 64.17 | 60.43 | 68.45 | 67.38 | 20.86 | 10.16 | 12.83 | 13.90 |
| **Qwen3-14B** | 66.84 | 71.12 | 63.64 | 65.78 | 58.29 | 53.48 | 68.98 | 65.24 | 23.53 | 12.30 | 13.37 | 12.83 |
| **Qwen3-14B-thinking** | 65.78 | 74.33 | 71.66 | 70.59 | 70.05 | 64.71 | 71.12 | 70.59 | 44.92 | 13.37 | 19.25 | 15.51 |
| **Baichuan-M2-32B** | 66.84 | 75.94 | 69.52 | 68.98 | 61.50 | 57.75 | 67.38 | 68.45 | 22.46 | 16.04 | 20.86 | 17.11 |
| **Bio-Medical-LLaMA-3-8B** | 28.34 | 55.08 | 37.43 | 41.18 | 29.41 | 23.53 | 36.36 | 39.04 | 26.74 | 18.18 | 13.37 | 14.97 |
| **MediPhi** | 22.99 | 44.92 | 32.62 | 26.20 | 14.44 | 9.09 | 28.88 | 30.48 | 7.49 | 9.63 | 10.16 | 11.23 |
| **MedGemma-4B** | 33.16 | 43.32 | 41.18 | 37.97 | 28.34 | 29.41 | 36.36 | 38.50 | 21.93 | 10.70 | 14.44 | 19.25 |
| **MedGemma-27B** | 70.05 | 76.47 | 70.59 | 74.33 | 68.45 | 65.78 | 72.73 | 70.05 | 62.03 | 12.83 | 29.41 | 45.45 |
| **MedReason-8B** | 39.57 | 51.87 | 20.86 | 20.32 | 39.57 | 25.13 | 16.58 | 34.22 | 15.51 | 12.30 | 9.63 | 11.23 |
| **HuatuoGPT-o1-7B** | 52.94 | 63.10 | 58.29 | 54.55 | 47.59 | 42.25 | 58.82 | 60.96 | 11.76 | 10.70 | 12.30 | 11.76 |
| **HuatuoGPT-o1-8B** | 45.99 | 54.55 | 54.55 | 47.59 | 38.50 | 35.83 | 47.59 | 47.59 | 29.41 | 9.09 | 12.30 | 6.95 |
| **HuatuoGPT-o1-70B** | 60.96 | 74.87 | 72.19 | 67.91 | 60.96 | 67.91 | 71.66 | 70.05 | 58.29 | 22.46 | 31.55 | 37.43 |
| **HuatuoGPT-o1-72B** | 67.38 | 77.01 | 73.26 | 71.66 | 67.91 | 66.31 | 74.87 | 73.26 | 47.06 | 24.06 | 24.06 | 20.32 |
| **OpenBioLLM-8B** | 10.70 | 19.79 | 15.51 | 18.18 | 13.37 | 8.56 | 18.72 | 16.04 | 11.76 | 9.09 | 11.23 | 11.76 |
| **OpenBioLLM-70B** | 24.60 | 62.03 | 58.29 | 35.29 | 27.27 | 21.93 | 41.18 | 59.36 | 25.67 | 10.16 | 13.90 | 13.37 |

**Supplementary Tab. 126:** Zero-Shot performance evaluation of 56 LLMs on MMLU-Pro (Run 5).

## S2.4. Language Disparity Analysis

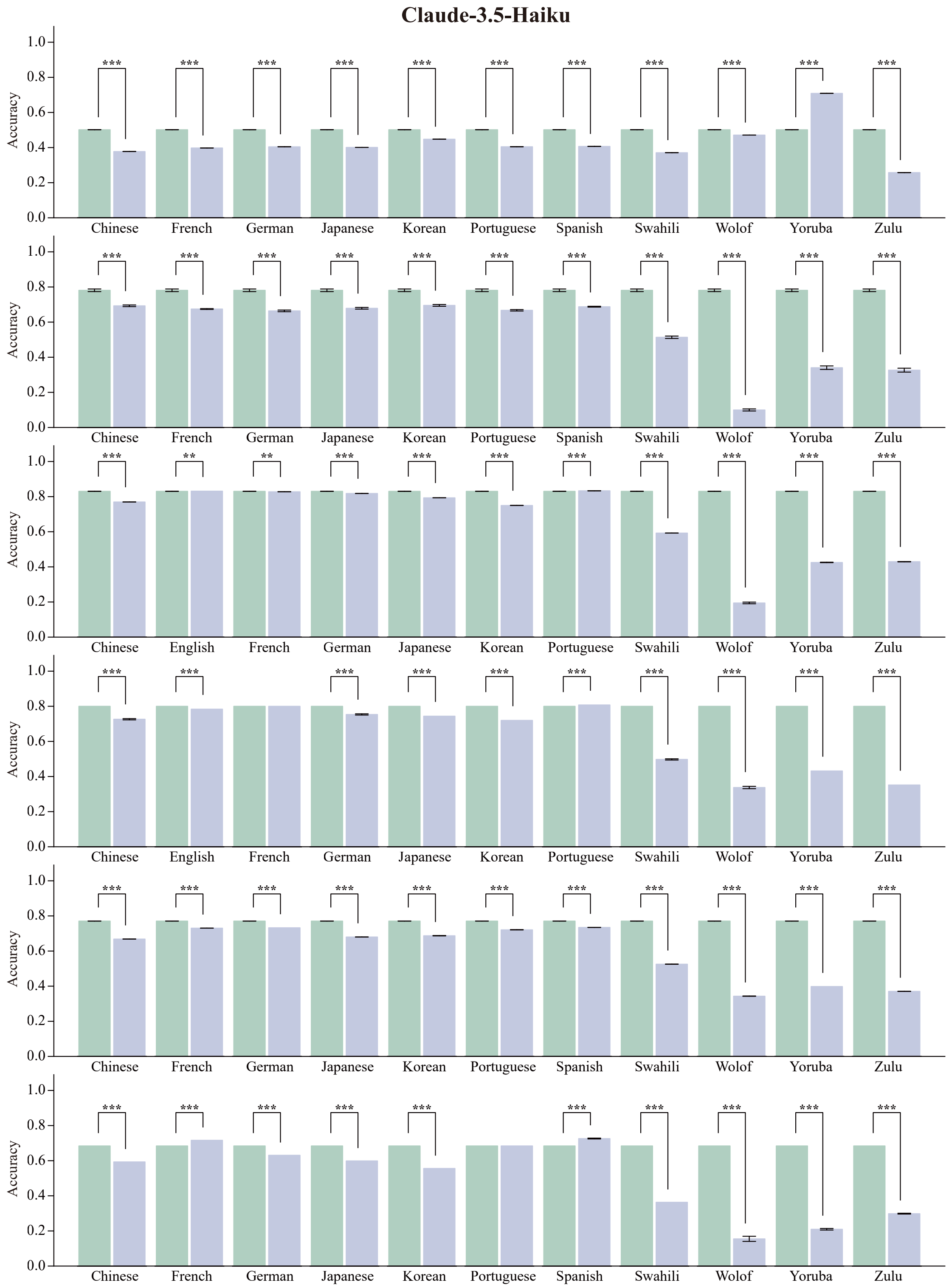

**Supplementary Fig. 2: Multilingual performance of Claude-3.5-Haiku on six medical benchmarks (top to bottom: BioNLI, MedNLI, HeadQA, MedExpQA, MedQA, MMLU-Pro).** The experiment compared the accuracy disparities between the original language (green) and target languages (purple), with each condition repeated five times. Statistical significance is indicated by asterisks (*p<0.05, **p<0.01, ***p<0.001).

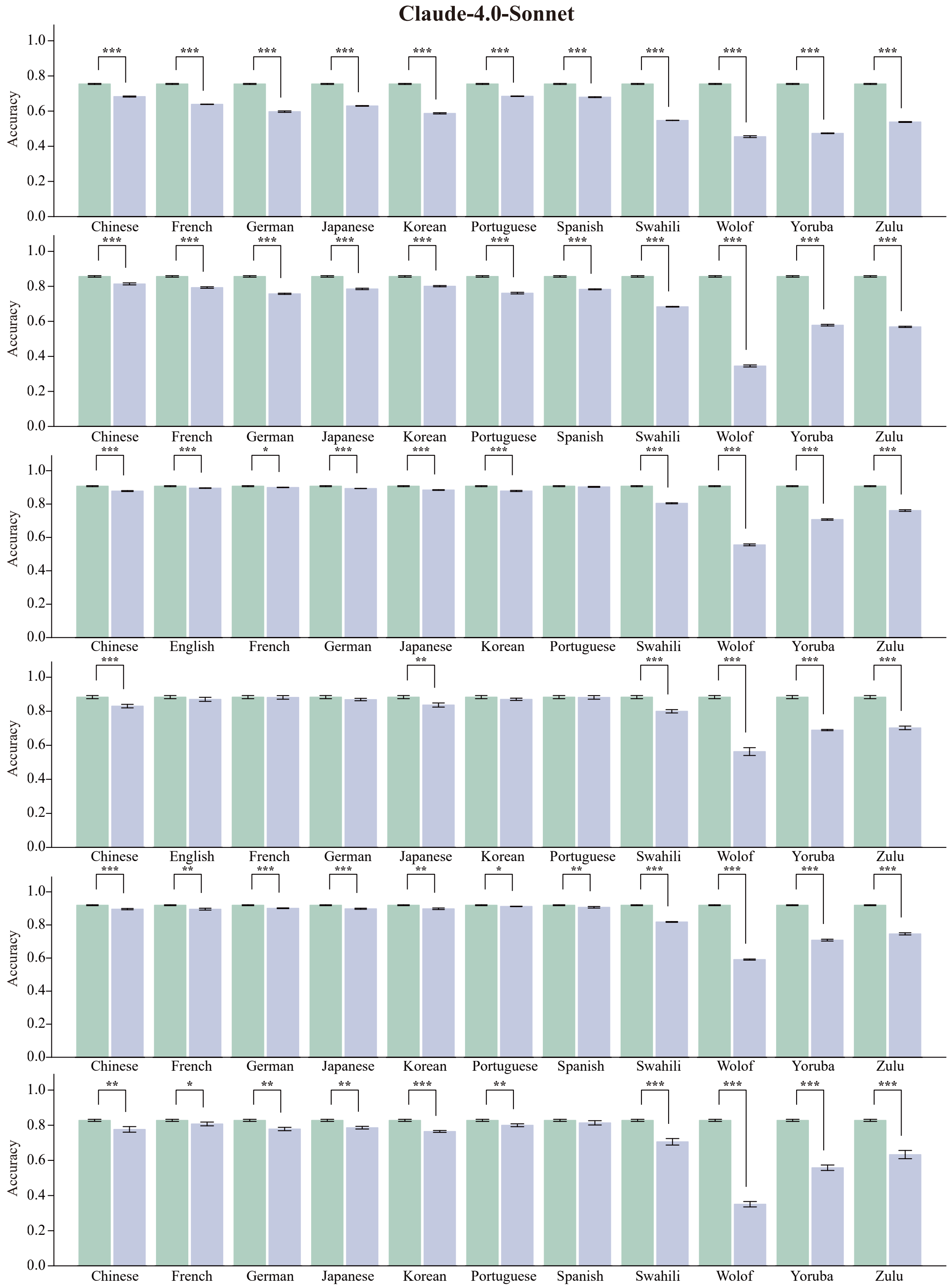

**Supplementary Fig. 3: Multilingual performance of Claude-4.0-Sonnet on six medical benchmarks (top to bottom: BioNLI, MedNLI, HeadQA, MedExpQA, MedQA, MMLU-Pro).** The experiment compared the accuracy disparities between the original language (green) and target languages (purple), with each condition repeated five times. Statistical significance is indicated by asterisks (*p<0.05, **p<0.01, ***p<0.001).

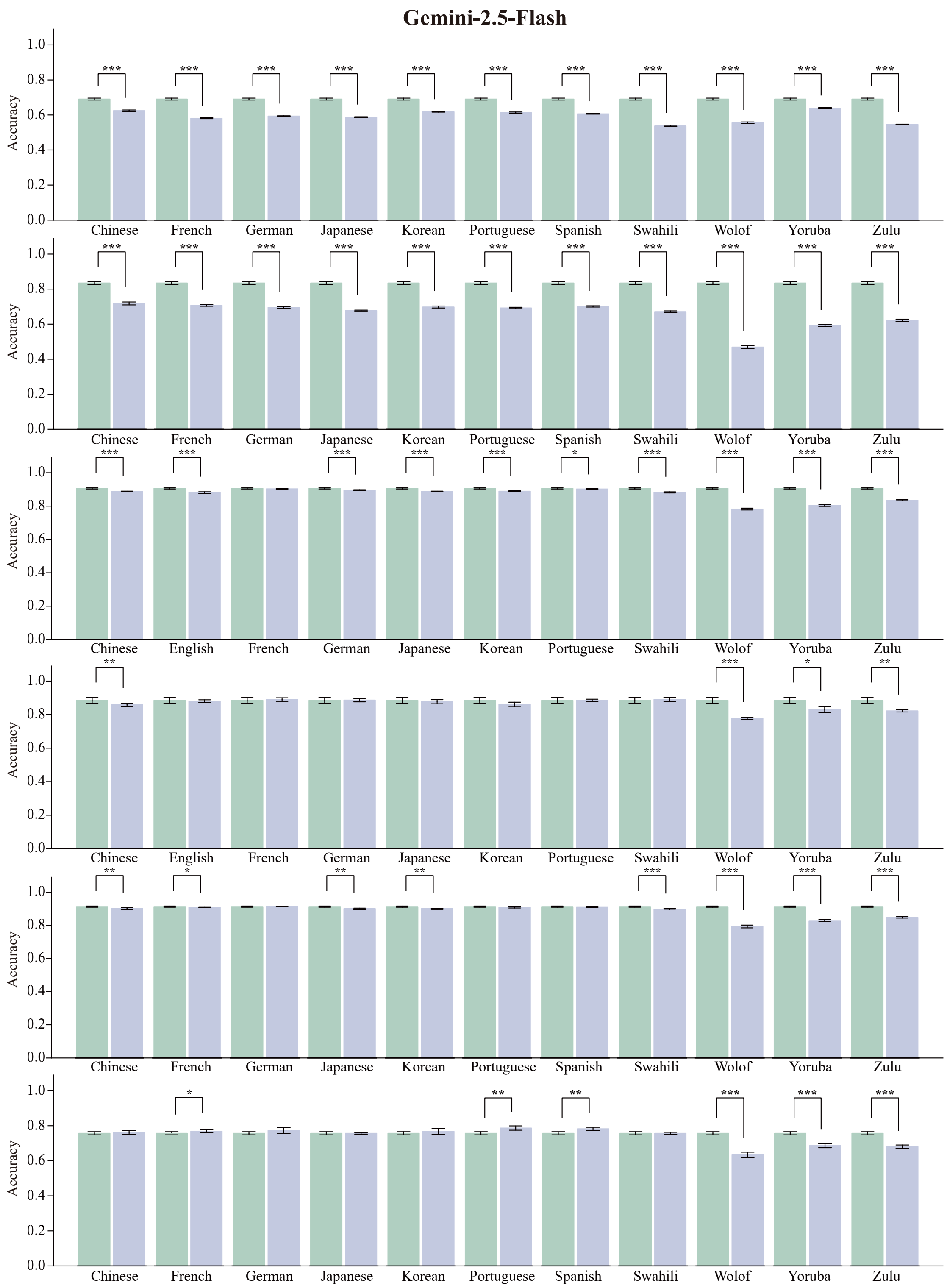

**Supplementary Fig. 4: Multilingual performance of Gemini-2.5-Flash on six medical benchmarks (top to bottom: BioNLI, MedNLI, HeadQA, MedExpQA, MedQA, MMLU-Pro).** The experiment compared the accuracy disparities between the original language (green) and target languages (purple), with each condition repeated five times. Statistical significance is indicated by asterisks (*p<0.05, **p<0.01, ***p<0.001).

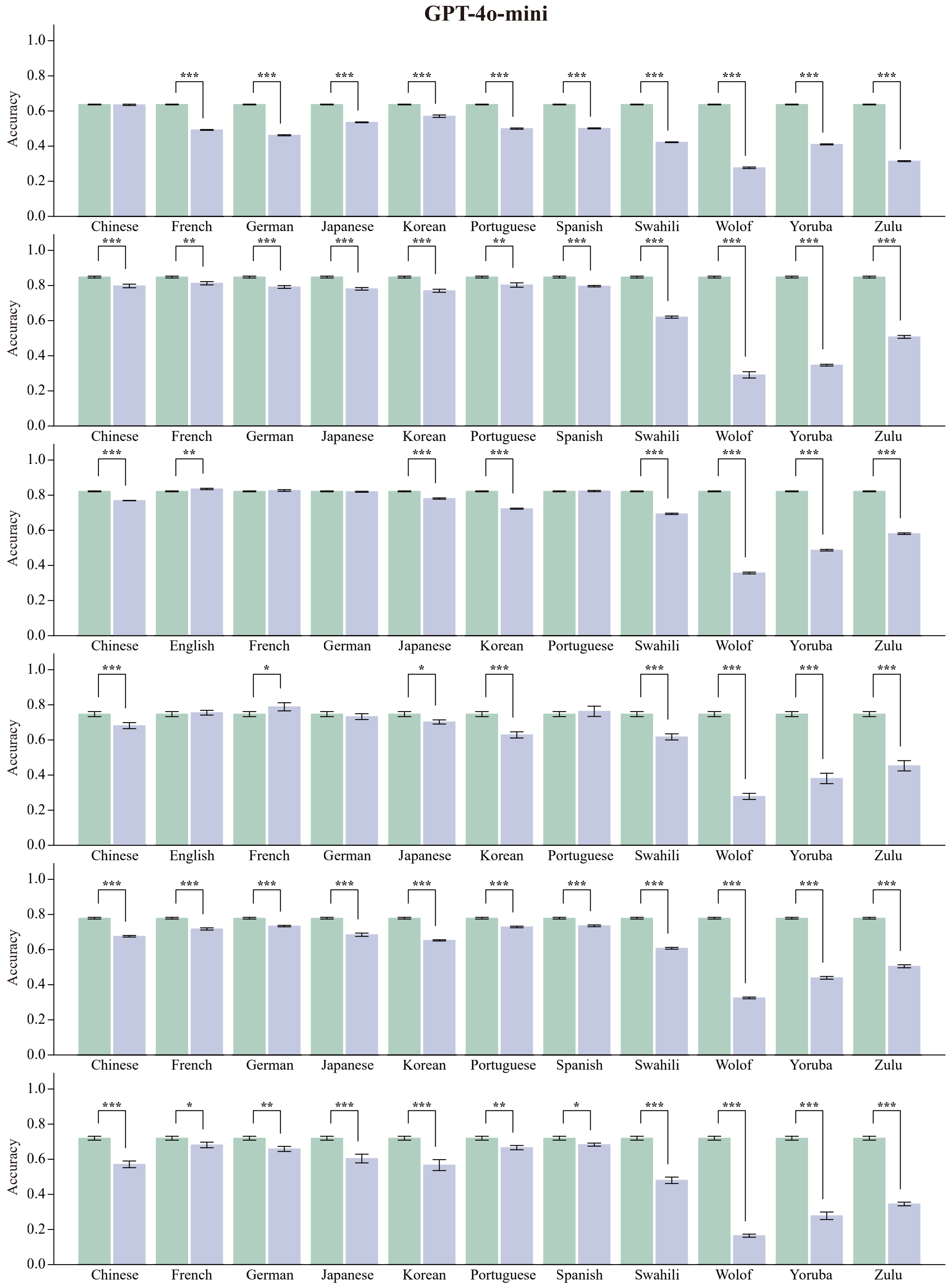

**Supplementary Fig. 5: Multilingual performance of GPT-4o-mini on six medical benchmarks (top to bottom: BioNLI, MedNLI, HeadQA, MedExpQA, MedQA, MMLU-Pro).** The experiment compared the accuracy disparities between the original language (green) and target languages (purple), with each condition repeated five times. Statistical significance is indicated by asterisks (*p<0.05, **p<0.01, ***p<0.001).

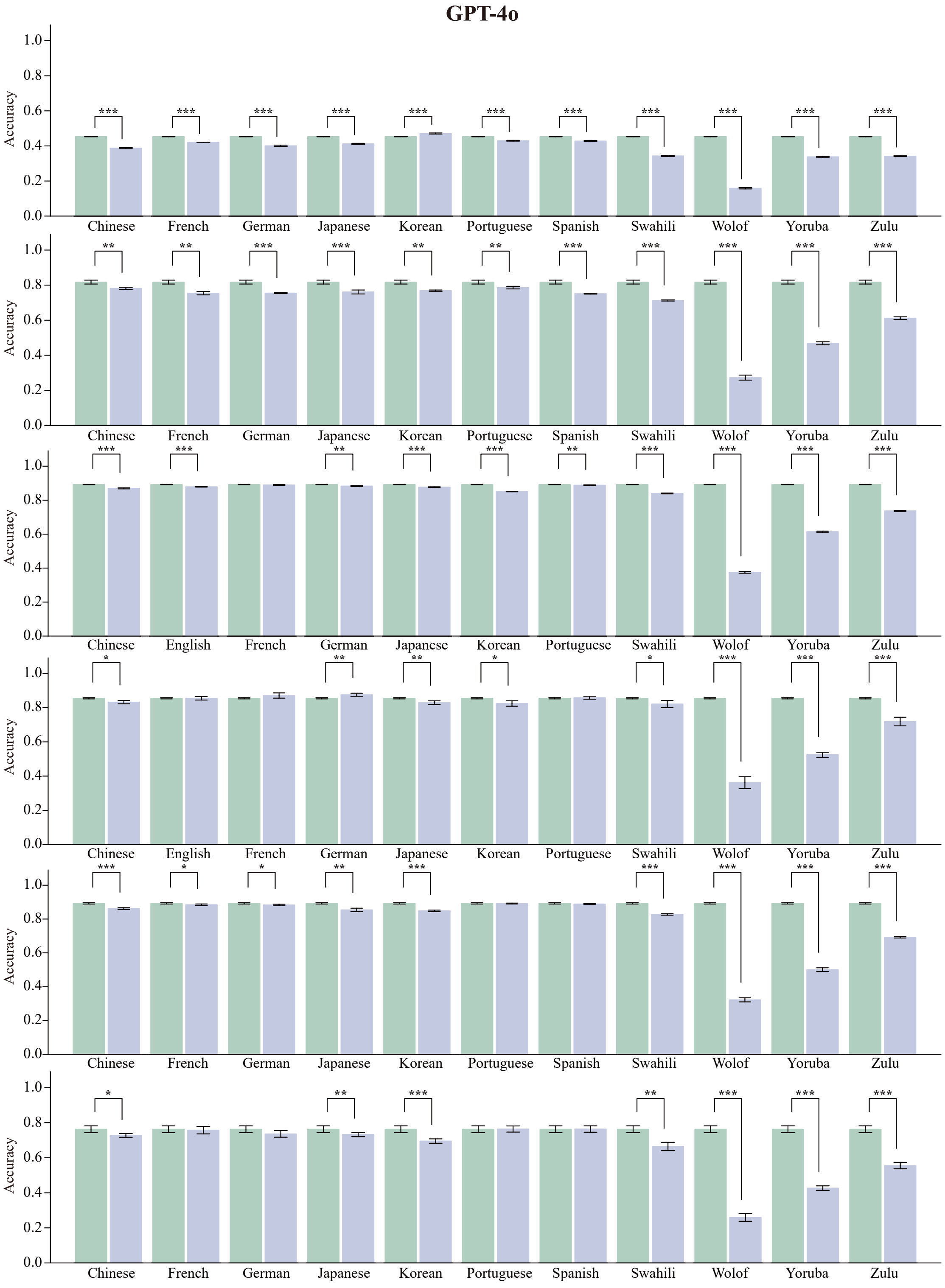

**Supplementary Fig. 6: Multilingual performance of GPT-4o on six medical benchmarks (top to bottom: BioNLI, MedNLI, HeadQA, MedExpQA, MedQA, MMLU-Pro).** The experiment compared the accuracy disparities between the original language (green) and target languages (purple), with each condition repeated five times. Statistical significance is indicated by asterisks (*$p<0.05$, **$p<0.01$, ***$p<0.001$).

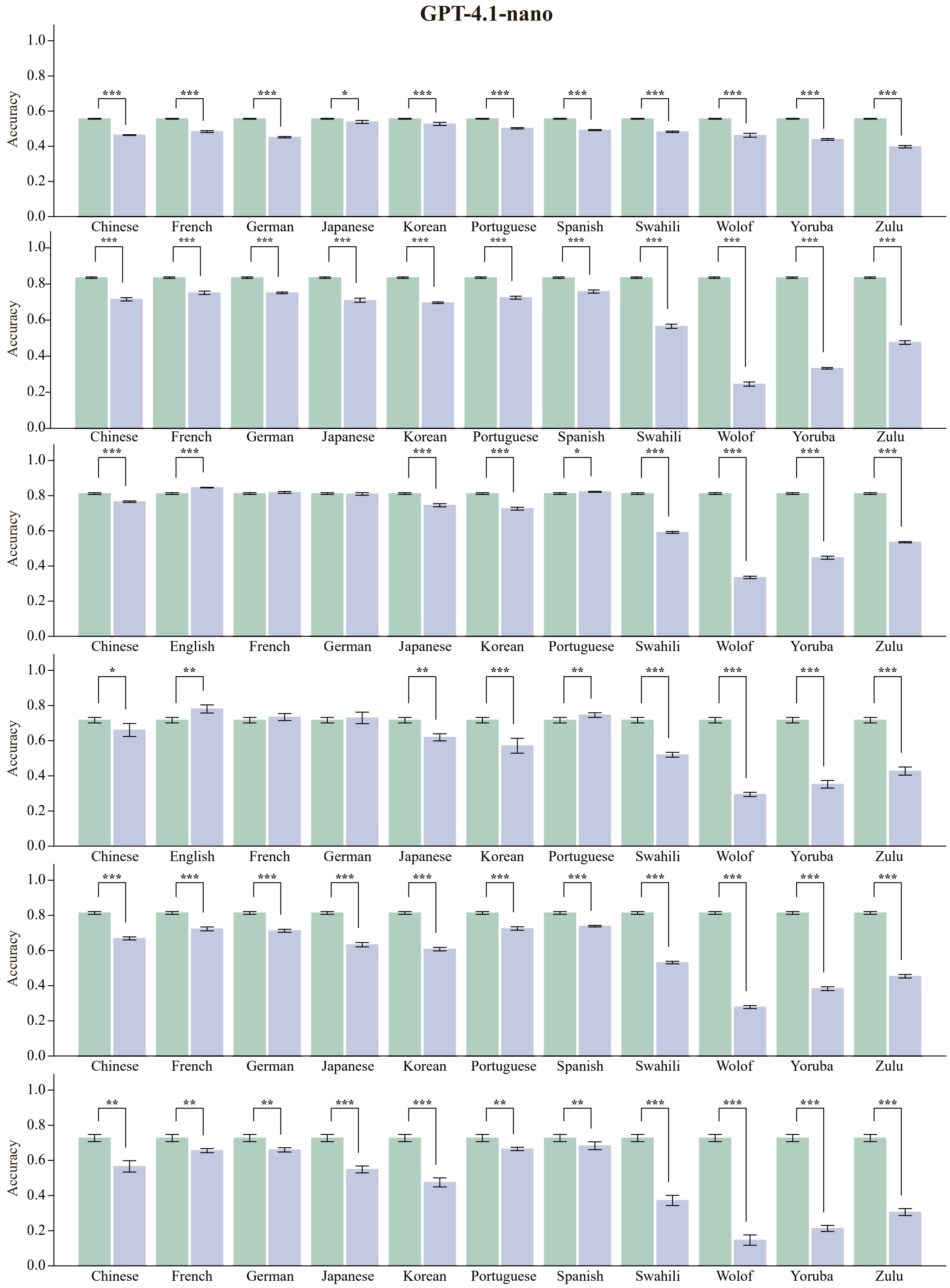

**Supplementary Fig. 7: Multilingual performance of GPT-4.1-nano on six medical benchmarks (top to bottom: BioNLI, MedNLI, HeadQA, MedExpQA, MedQA, MMLU-Pro).** The experiment compared the accuracy disparities between the original language (green) and target languages (purple), with each condition repeated five times. Statistical significance is indicated by asterisks (*p<0.05, **p<0.01, ***p<0.001).

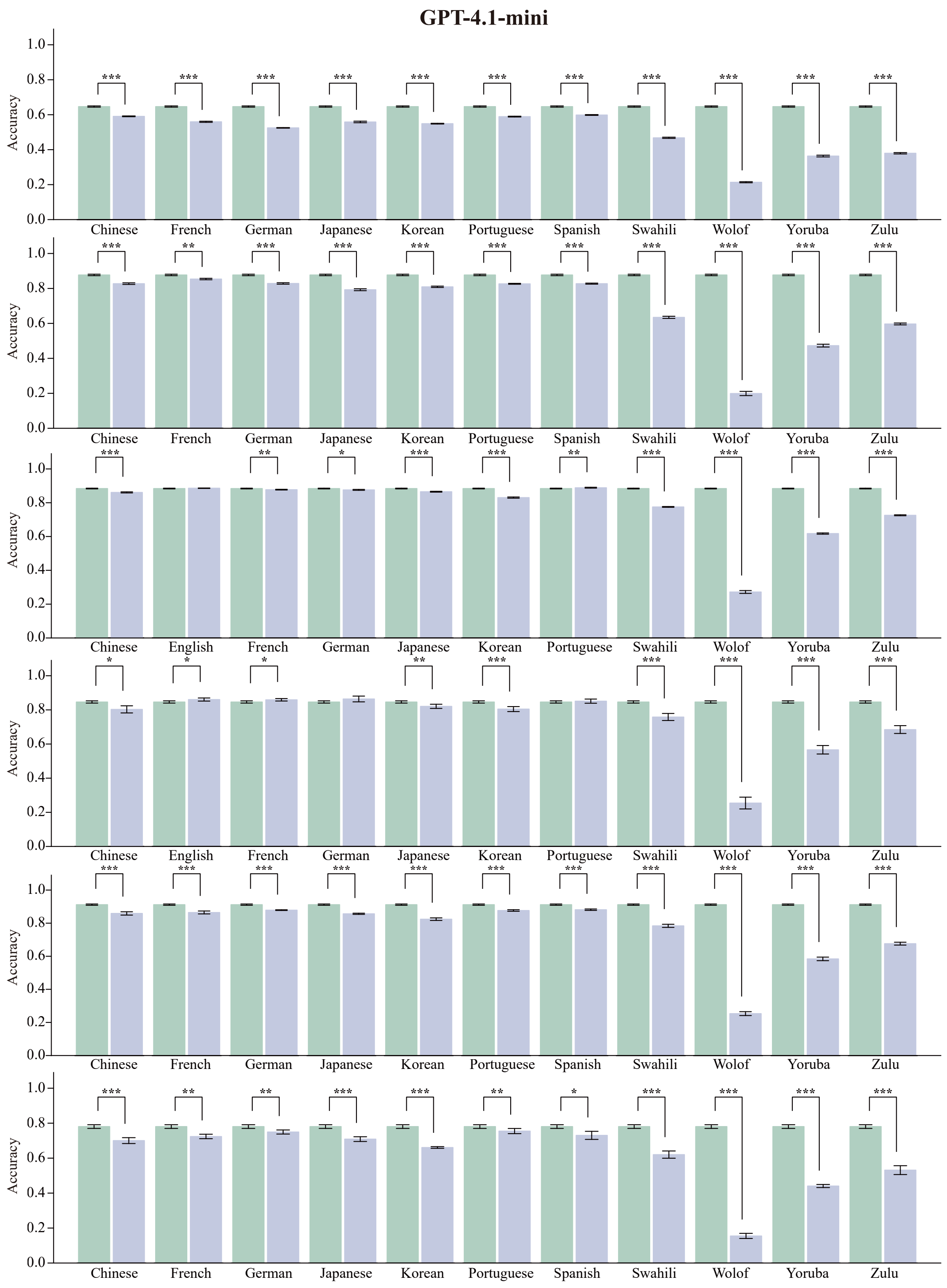

**Supplementary Fig. 8: Multilingual performance of GPT-4.1-mini on six medical benchmarks (top to bottom: BioNLI, MedNLI, HeadQA, MedExpQA, MedQA, MMLU-Pro).** The experiment compared the accuracy disparities between the original language (green) and target languages (purple), with each condition repeated five times. Statistical significance is indicated by asterisks (*p<0.05, **p<0.01, ***p<0.001).

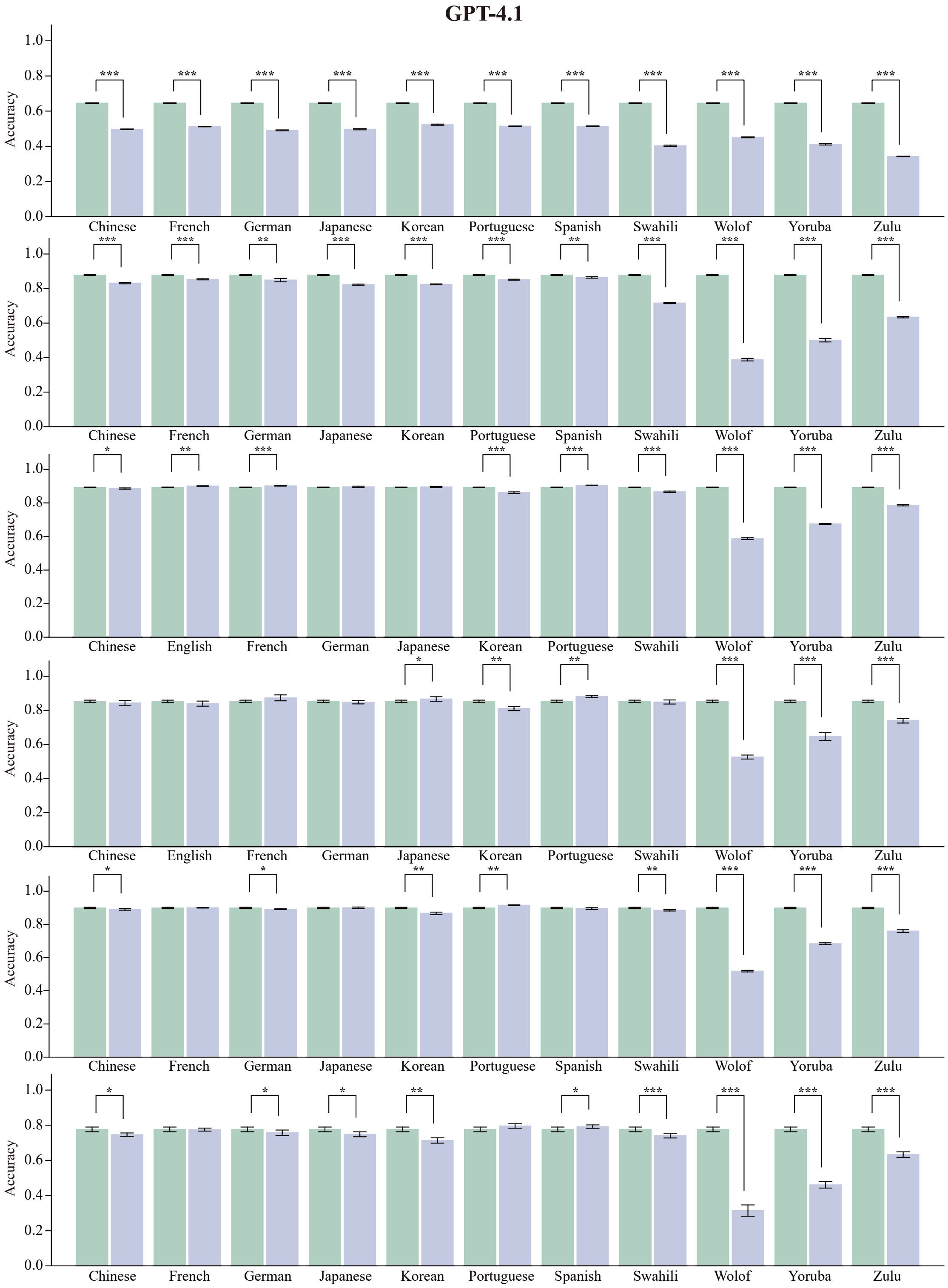

**Supplementary Fig. 9:** ***Multilingual performance of*** **GPT-4.1** ***on six medical benchmarks (top to bottom: BioNLI,*** **MedNLI, HeadQA, MedExpQA, MedQA, MMLU-Pro).** The experiment compared the accuracy disparities between the original language (green) and target languages (purple), with each condition repeated five times. Statistical significance is indicated by asterisks (*p<0.05, **p<0.01, ***p<0.001).

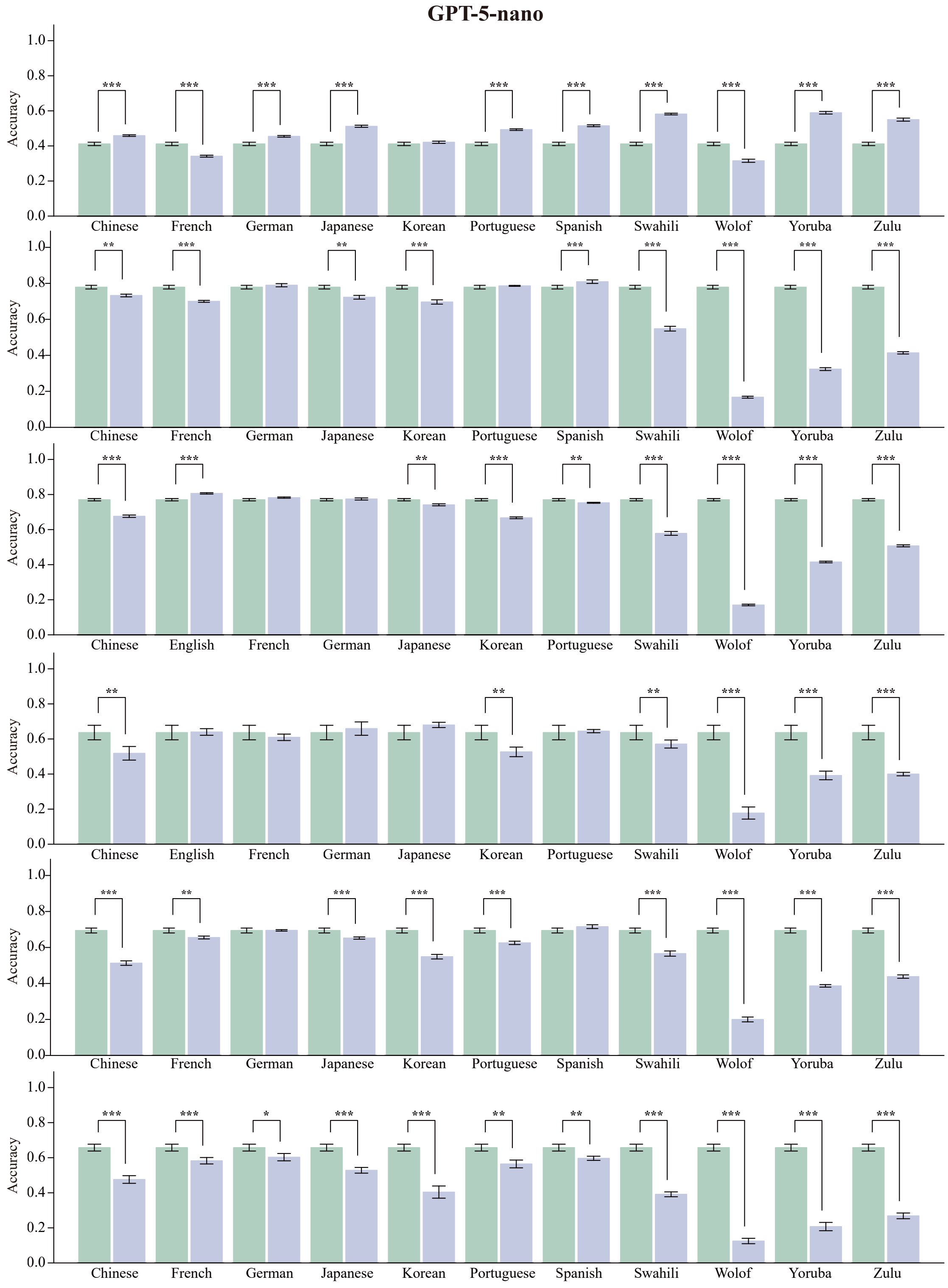

**Supplementary Fig. 10: Multilingual performance of GPT-5-nano on six medical benchmarks (top to bottom: BioNLI, MedNLI, HeadQA, MedExpQA, MedQA, MMLU-Pro).** The experiment compared the accuracy disparities between the original language (green) and target languages (purple), with each condition repeated five times. Statistical significance is indicated by asterisks (*p<0.05, **p<0.01, ***p<0.001).

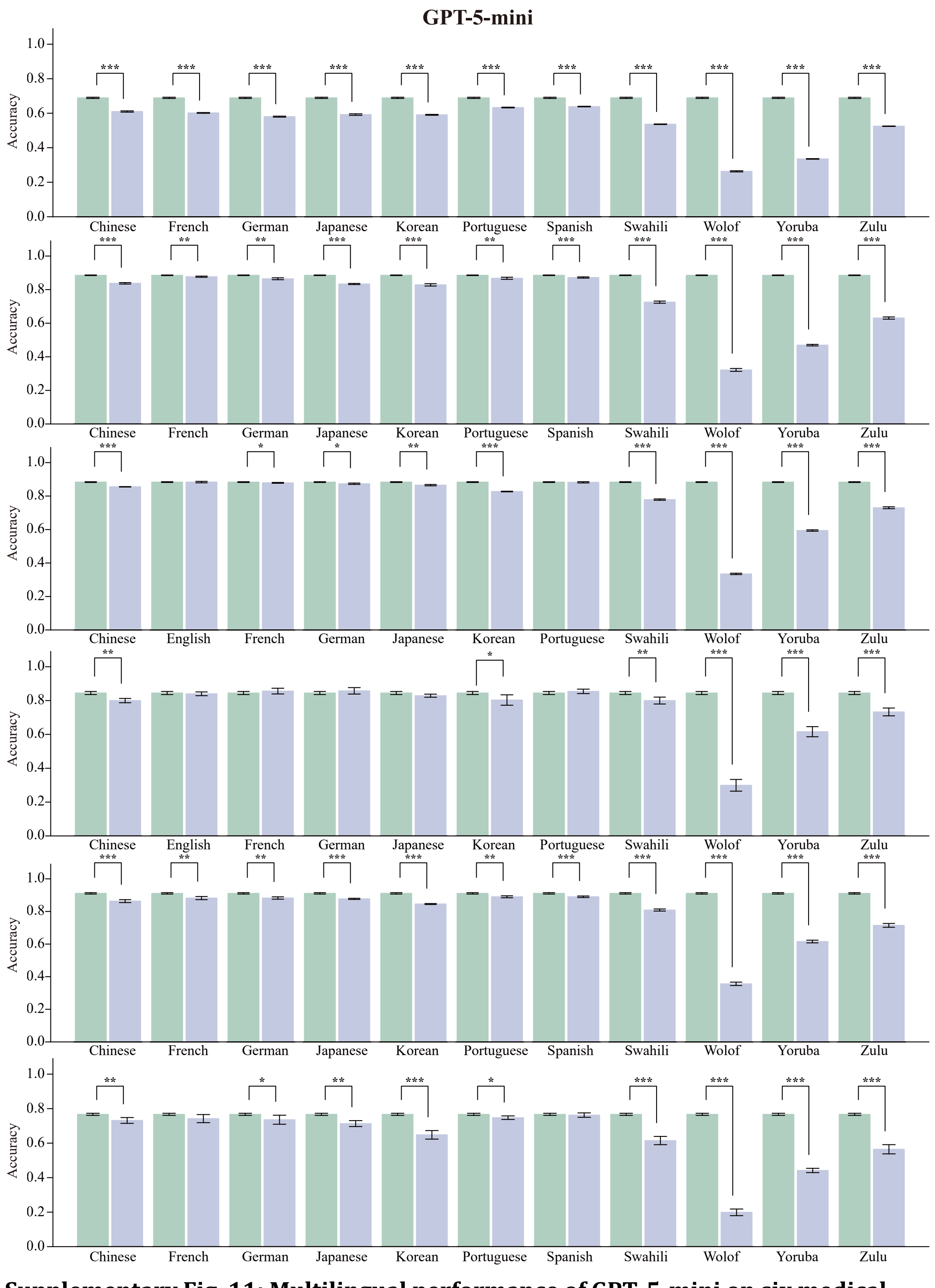

**Supplementary Fig. 11: Multilingual performance of GPT-5-mini on six medical benchmarks (top to bottom: BioNLI, MedNLI, HeadQA, MedExpQA, MedQA, MMLU-Pro).** The experiment compared the accuracy disparities between the original language (green) and target languages (purple), with each condition repeated five times. Statistical significance is indicated by asterisks (*p<0.05, **p<0.01, ***p<0.001).

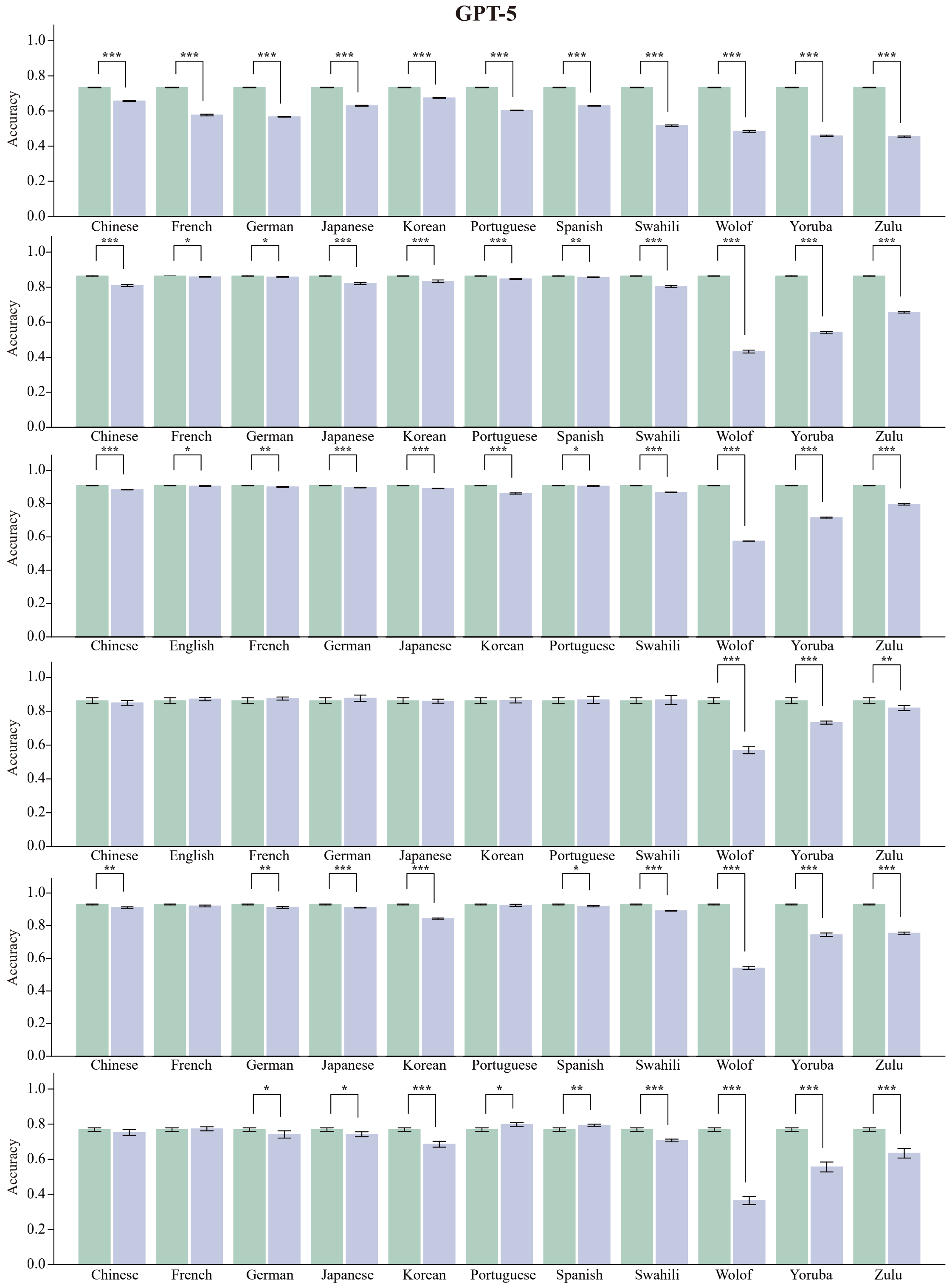

**Supplementary Fig. 12: Multilingual performance of GPT-5 on six medical benchmarks (top to bottom: BioNLI, MedNLI, HeadQA, MedExpQA, MedQA, MMLU-Pro).** The experiment compared the accuracy disparities between the original language (green) and target languages (purple), with each condition repeated five times. Statistical significance is indicated by asterisks (*p<0.05, **p<0.01, ***p<0.001).

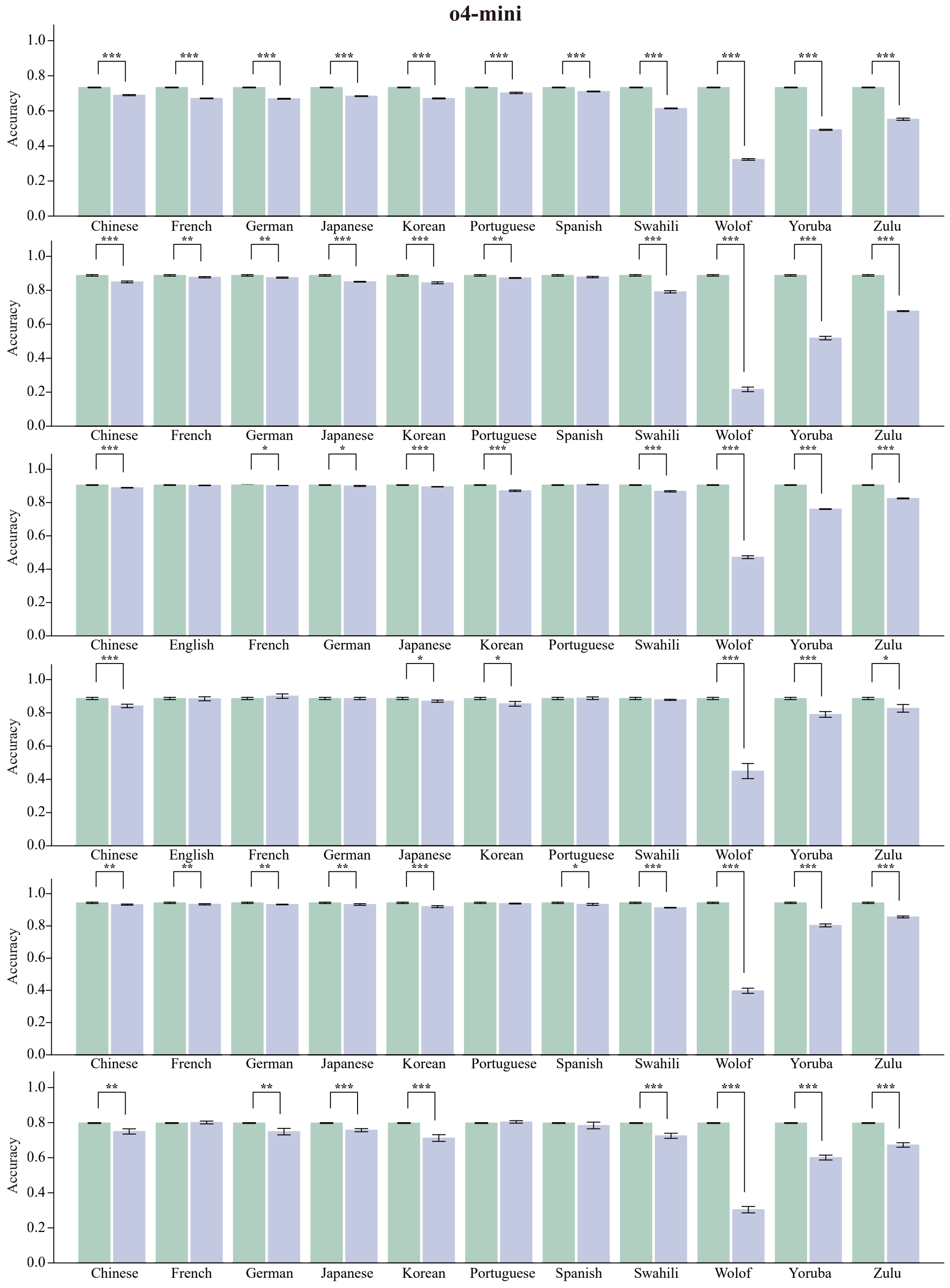

**Supplementary Fig. 13: Multilingual performance of o4-mini on six medical benchmarks (top to bottom: BioNLI, MedNLI, HeadQA, MedExpQA, MedQA, MMLU-Pro).** The experiment compared the accuracy disparities between the original language (green) and target languages (purple), with each condition repeated five times. Statistical significance is indicated by asterisks (*p<0.05, **p<0.01, ***p<0.001).

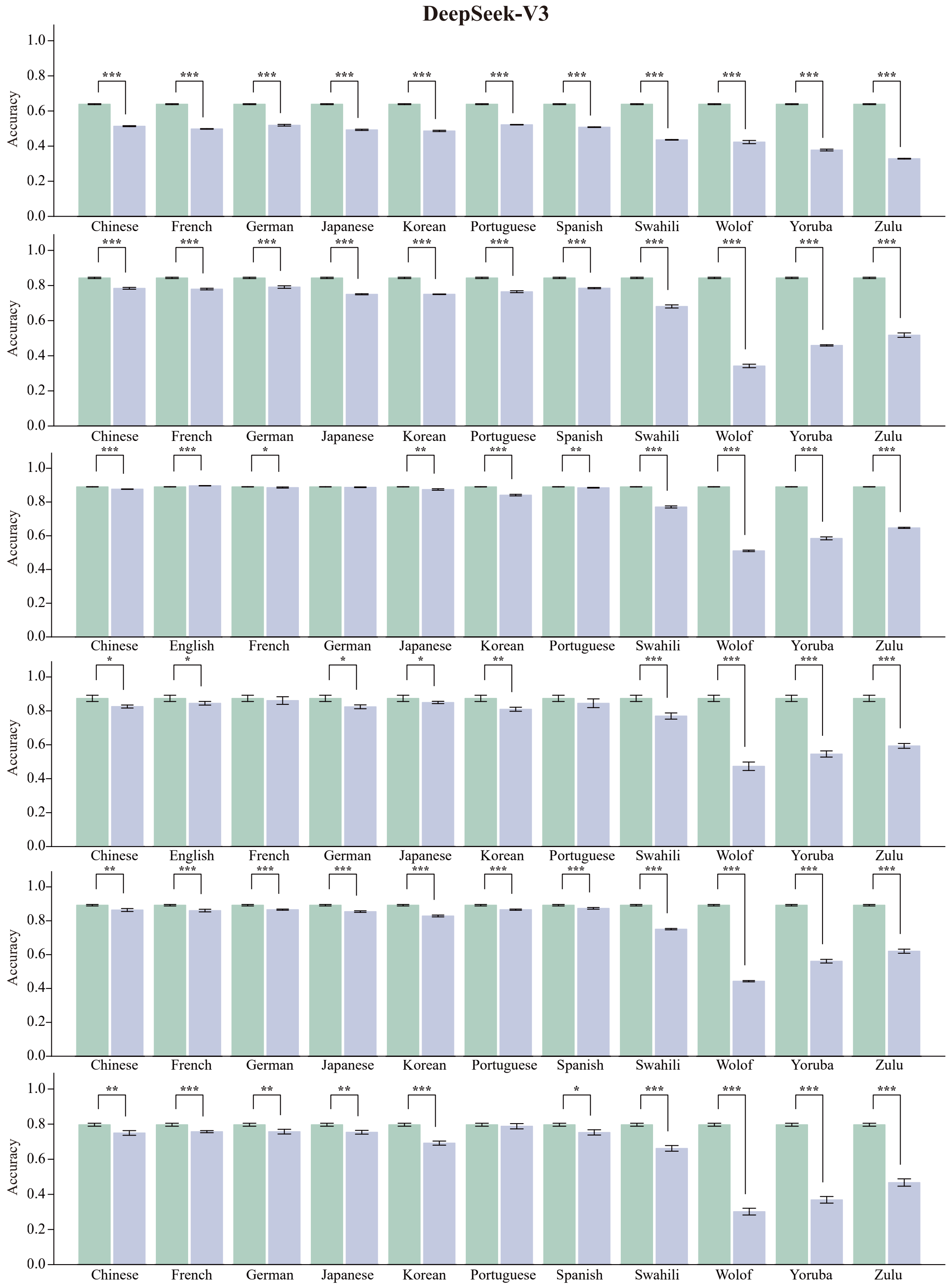

**Supplementary Fig. 14: Multilingual performance of DeepSeek-V3 on six medical benchmarks (top to bottom: BioNLI, MedNLI, HeadQA, MedExpQA, MedQA, MMLU-Pro).** The experiment compared the accuracy disparities between the original language (green) and target languages (purple), with each condition repeated five times. Statistical significance is indicated by asterisks (*p<0.05, **p<0.01, ***p<0.001).

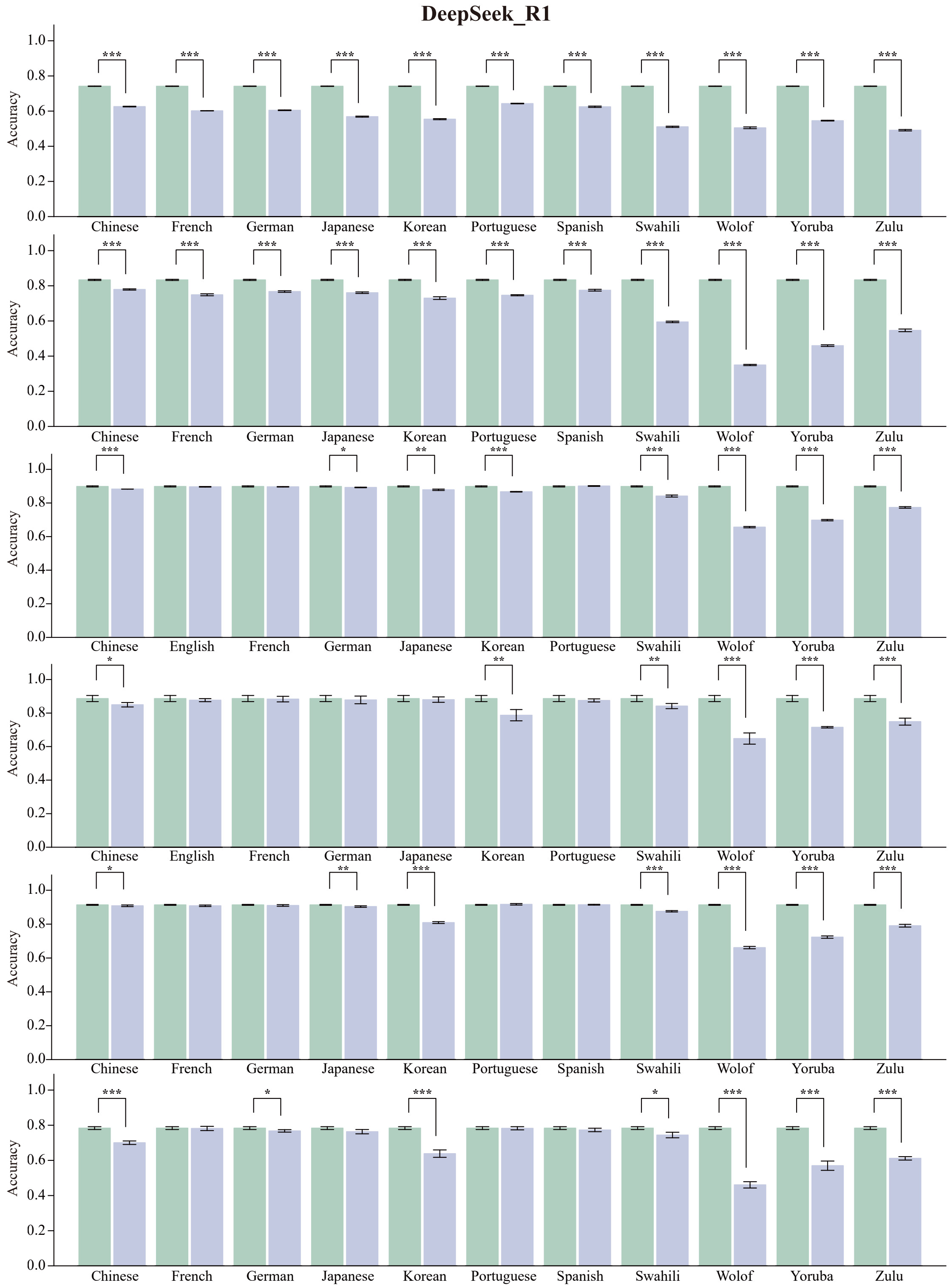

**Supplementary Fig. 15: Multilingual performance of DeepSeek-R1 on six medical benchmarks (top to bottom: BioNLI, MedNLI, HeadQA, MedExpQA, MedQA, MMLU-Pro).** The experiment compared the accuracy disparities between the original language (green) and target languages (purple), with each condition repeated five times. Statistical significance is indicated by asterisks (*p<0.05, **p<0.01, ***p<0.001).

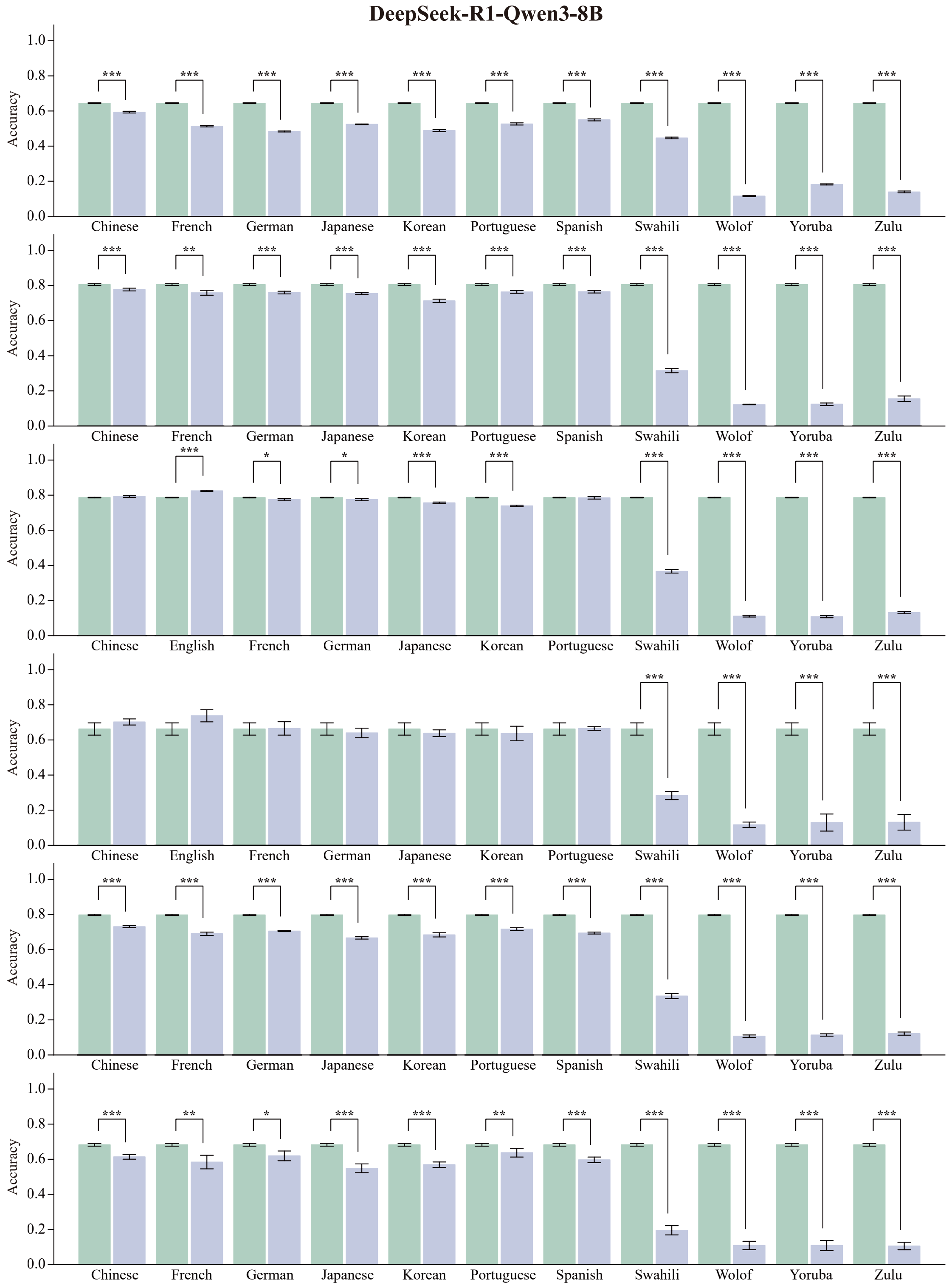

**Supplementary Fig. 16: Multilingual performance of DeepSeek-R1-Qwen3-8B on six medical benchmarks (top to bottom: BioNLI, MedNLI, HeadQA, MedExpQA, MedQA, MMLU-Pro).** The experiment compared the accuracy disparities between the original language (green) and target languages (purple), with each condition repeated five times. Statistical significance is indicated by asterisks (*p<0.05, **p<0.01, ***p<0.001).

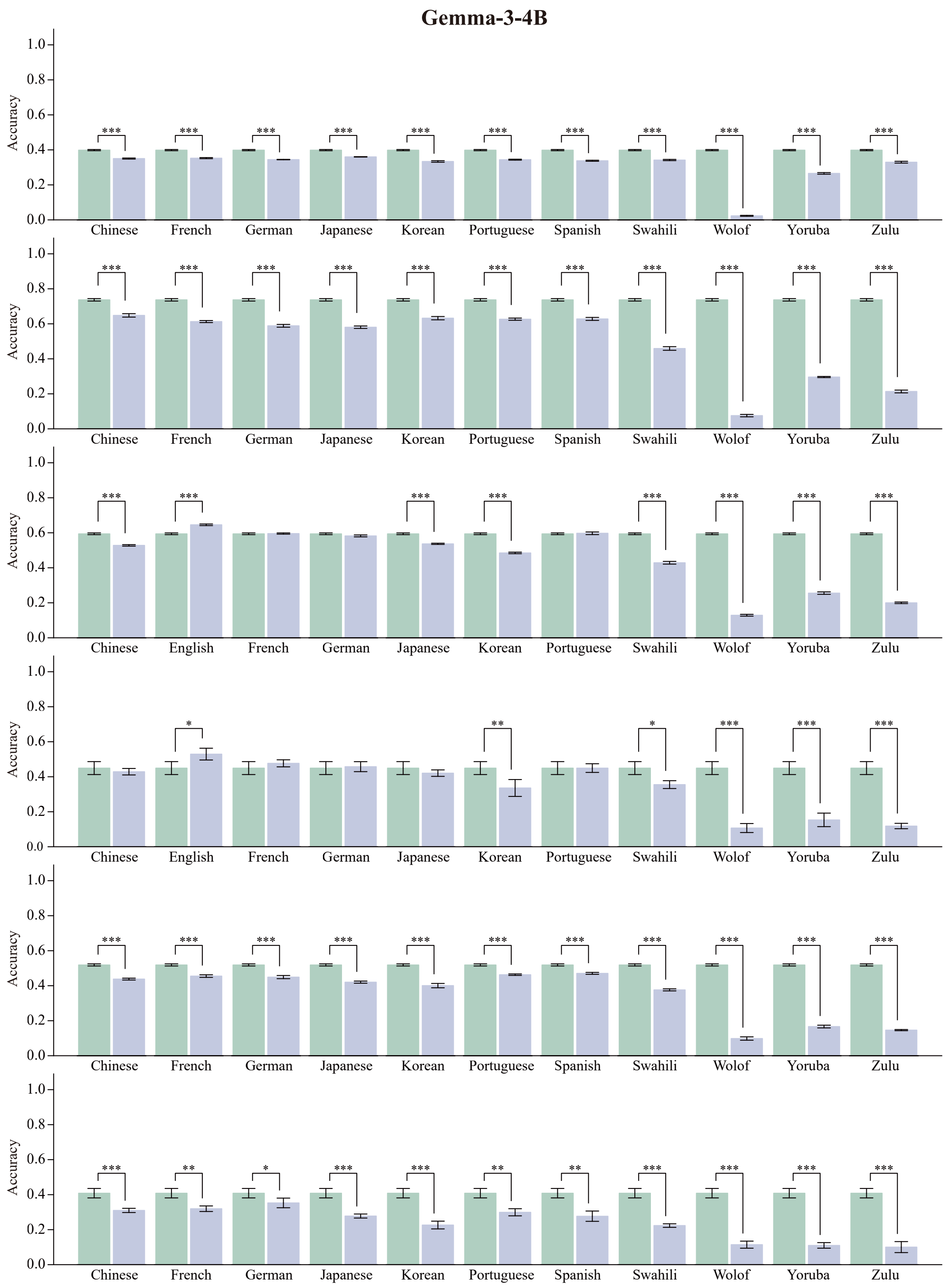

**Supplementary Fig. 17: Multilingual performance of Gemma-3-4B on six medical benchmarks (top to bottom: BioNLI, MedNLI, HeadQA, MedExpQA, MedQA, MMLU-Pro).** The experiment compared the accuracy disparities between the original language (green) and target languages (purple), with each condition repeated five times. Statistical significance is indicated by asterisks (*p<0.05, **p<0.01, ***p<0.001).

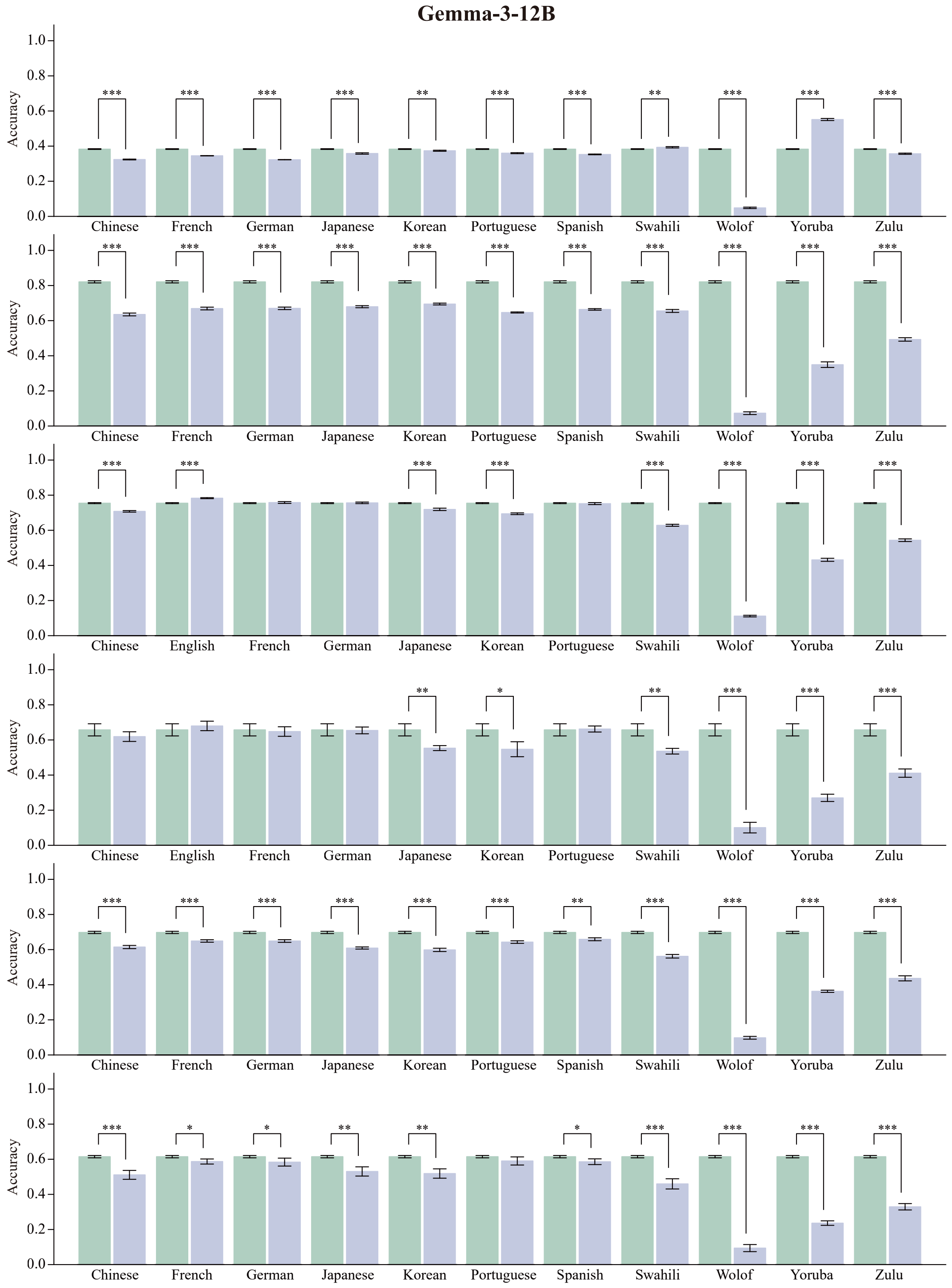

**Supplementary Fig. 18: Multilingual performance of Gemma-3-12B on six medical benchmarks (top to bottom: BioNLI, MedNLI, HeadQA, MedExpQA, MedQA, MMLU-Pro).** The experiment compared the accuracy disparities between the original language (green) and target languages (purple), with each condition repeated five times. Statistical significance is indicated by asterisks (*p<0.05, **p<0.01, ***p<0.001).

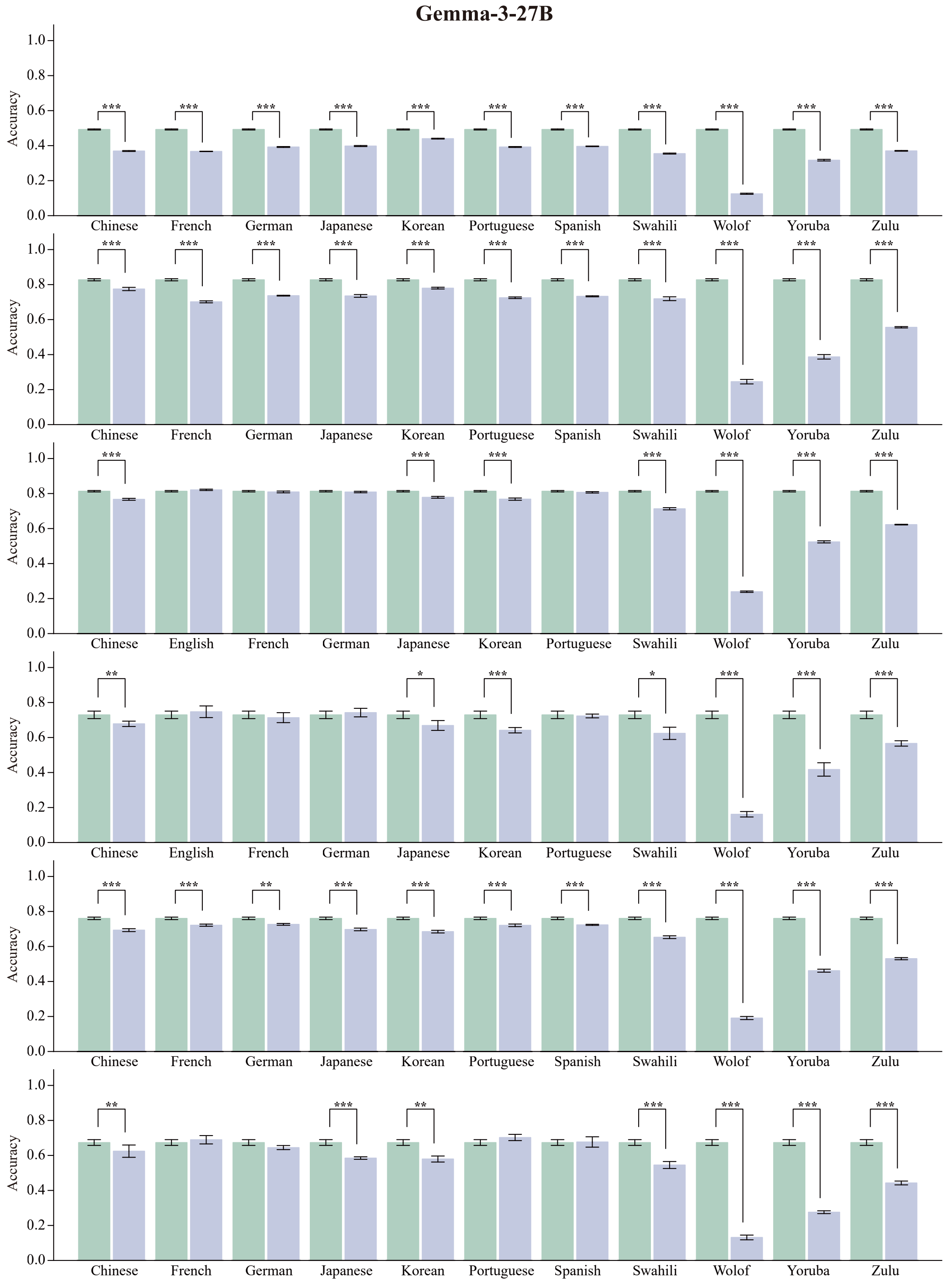

**Supplementary Fig. 19: Multilingual performance of Gemma-3-27B on six medical benchmarks (top to bottom: BioNLI, MedNLI, HeadQA, MedExpQA, MedQA, MMLU-Pro).** The experiment compared the accuracy disparities between the original language (green) and target languages (purple), with each condition repeated five times. Statistical significance is indicated by asterisks (*p<0.05, **p<0.01, ***p<0.001).

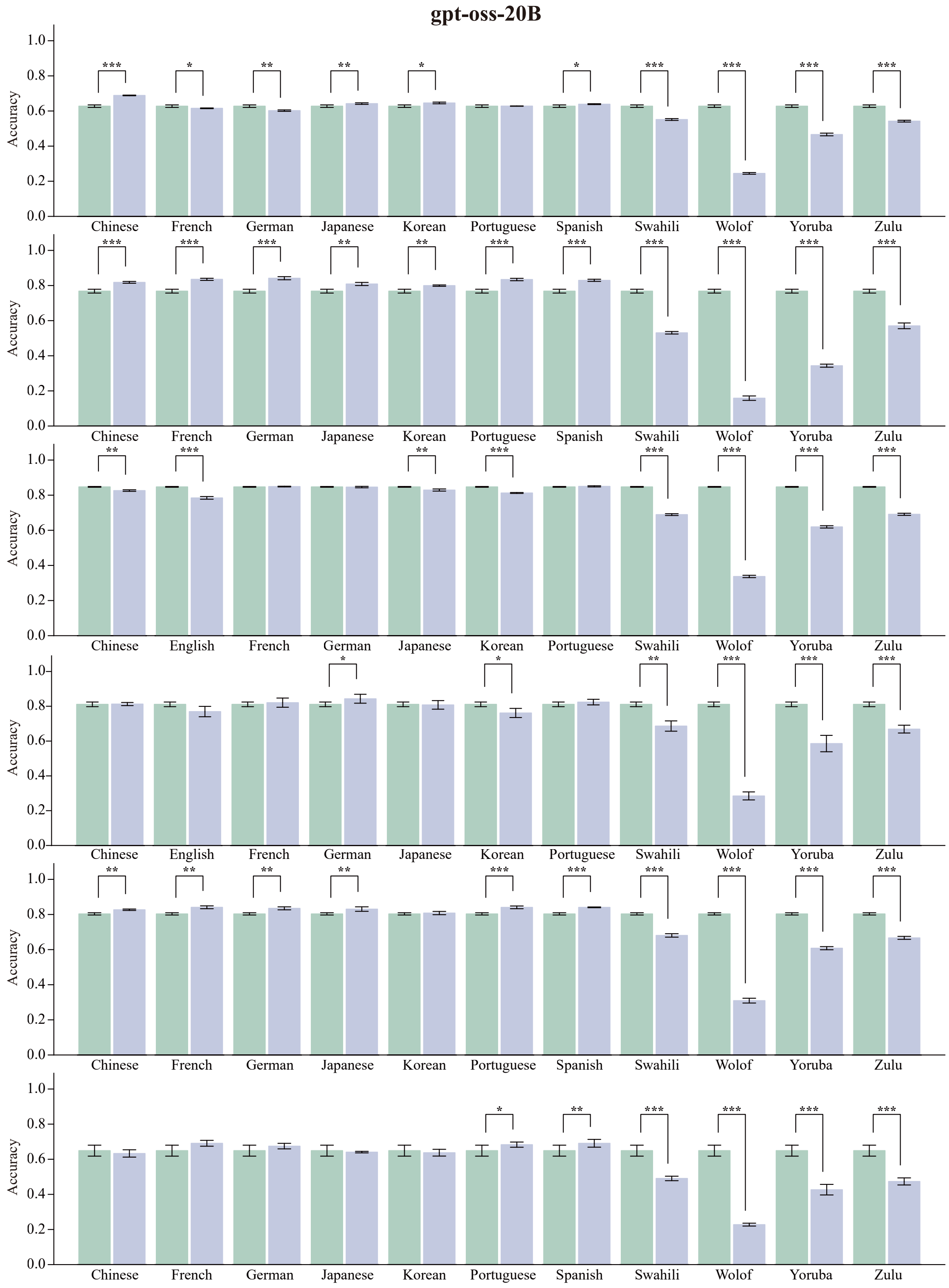

**Supplementary Fig. 20: Multilingual performance of gpt-oss-20B on six medical benchmarks (top to bottom: BioNLI, MedNLI, HeadQA, MedExpQA, MedQA, MMLU-Pro).** The experiment compared the accuracy disparities between the original language (green) and target languages (purple), with each condition repeated five times. Statistical significance is indicated by asterisks (*p<0.05, **p<0.01, ***p<0.001).

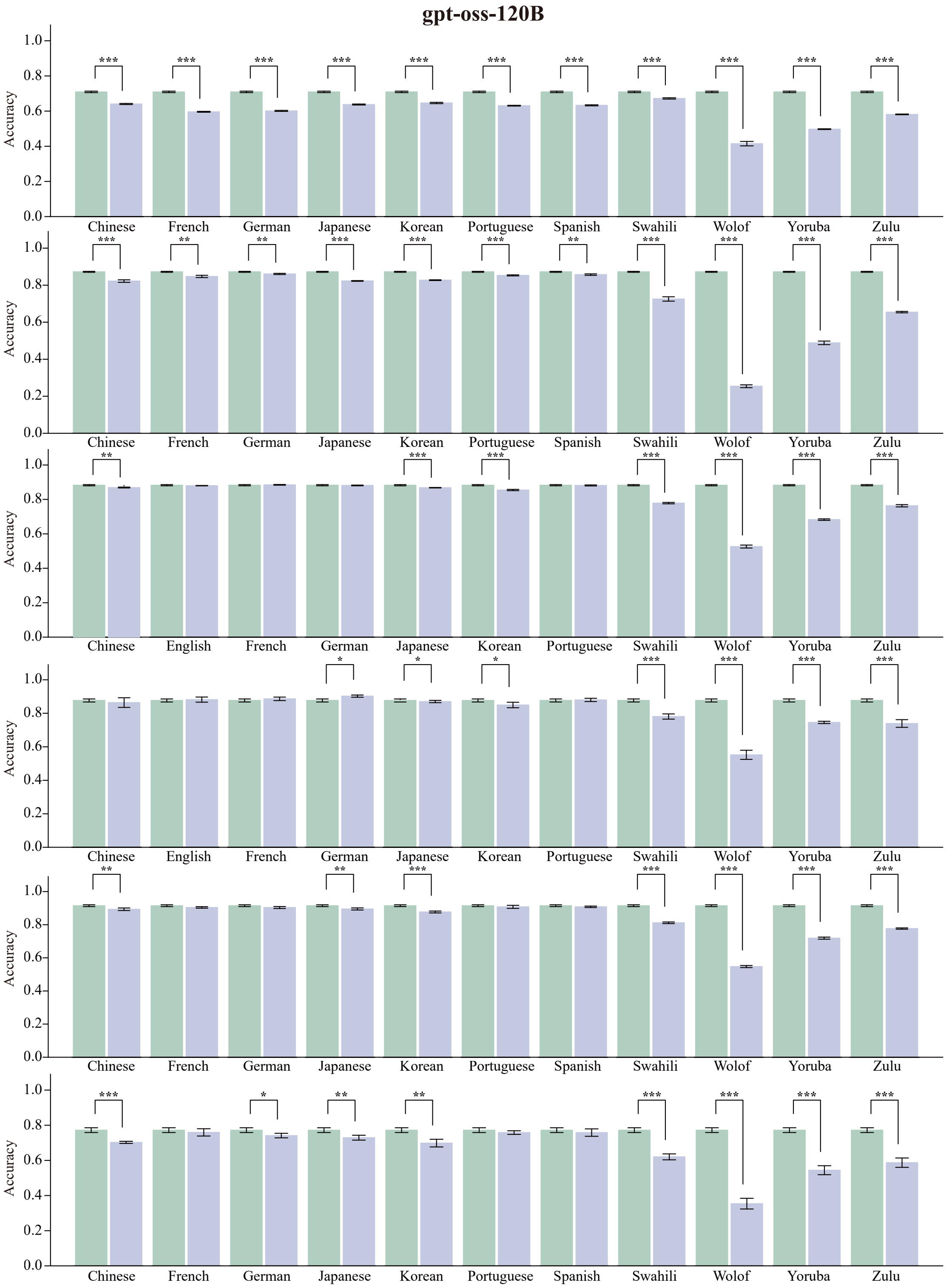

**Supplementary Fig. 21: Multilingual performance of gpt-oss-120B on six medical benchmarks (top to bottom: BioNLI, MedNLI, HeadQA, MedExpQA, MedQA, MMLU-Pro).** The experiment compared the accuracy disparities between the original language (green) and target languages (purple), with each condition repeated five times. Statistical significance is indicated by asterisks (*p<0.05, **p<0.01, ***p<0.001).

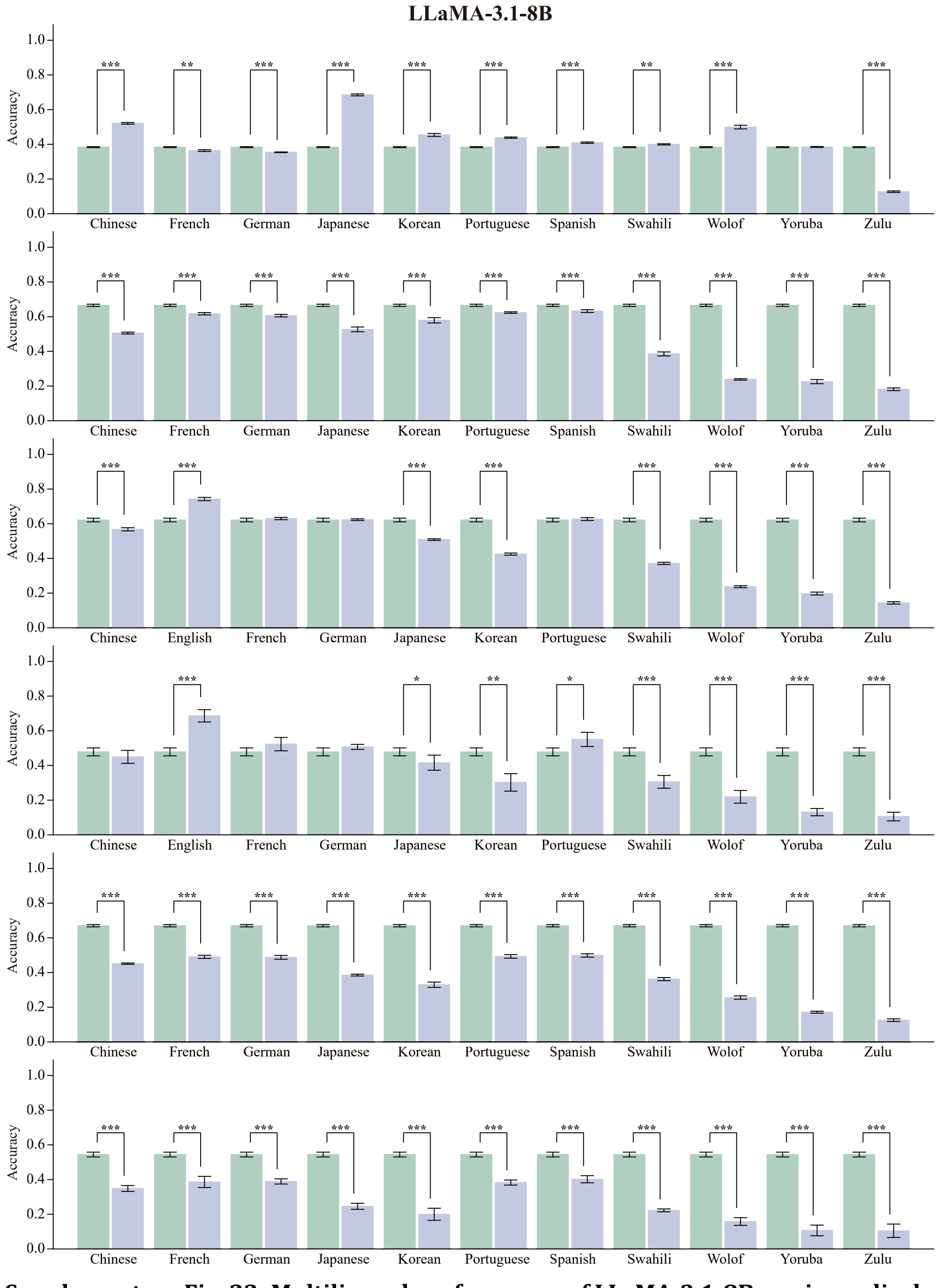

**Supplementary Fig. 22: Multilingual performance of LLaMA-3.1-8B on six medical benchmarks (top to bottom: BioNLI, MedNLI, HeadQA, MedExpQA, MedQA, MMLU-Pro).** The experiment compared the accuracy disparities between the original language (green) and target languages (purple), with each condition repeated five times. Statistical significance is indicated by asterisks (*p<0.05, **p<0.01, ***p<0.001).

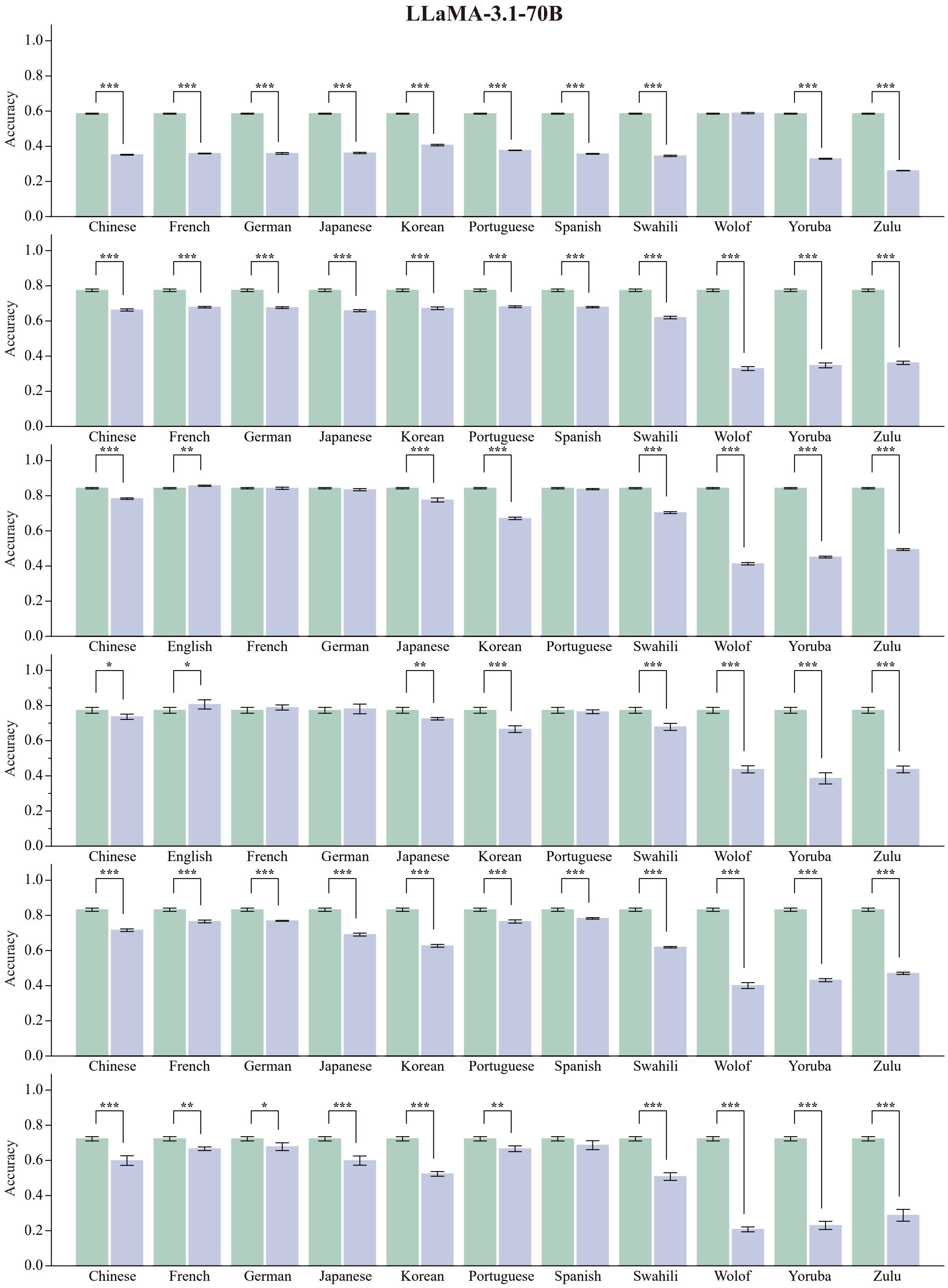

**Supplementary Fig. 23: Multilingual performance of LLaMA-3.1-70B on six medical benchmarks (top to bottom: BioNLI, MedNLI, HeadQA, MedExpQA, MedQA, MMLU-Pro).** The experiment compared the accuracy disparities between the original language (green) and target languages (purple), with each condition repeated five times. Statistical significance is indicated by asterisks (*p<0.05, **p<0.01, ***p<0.001).

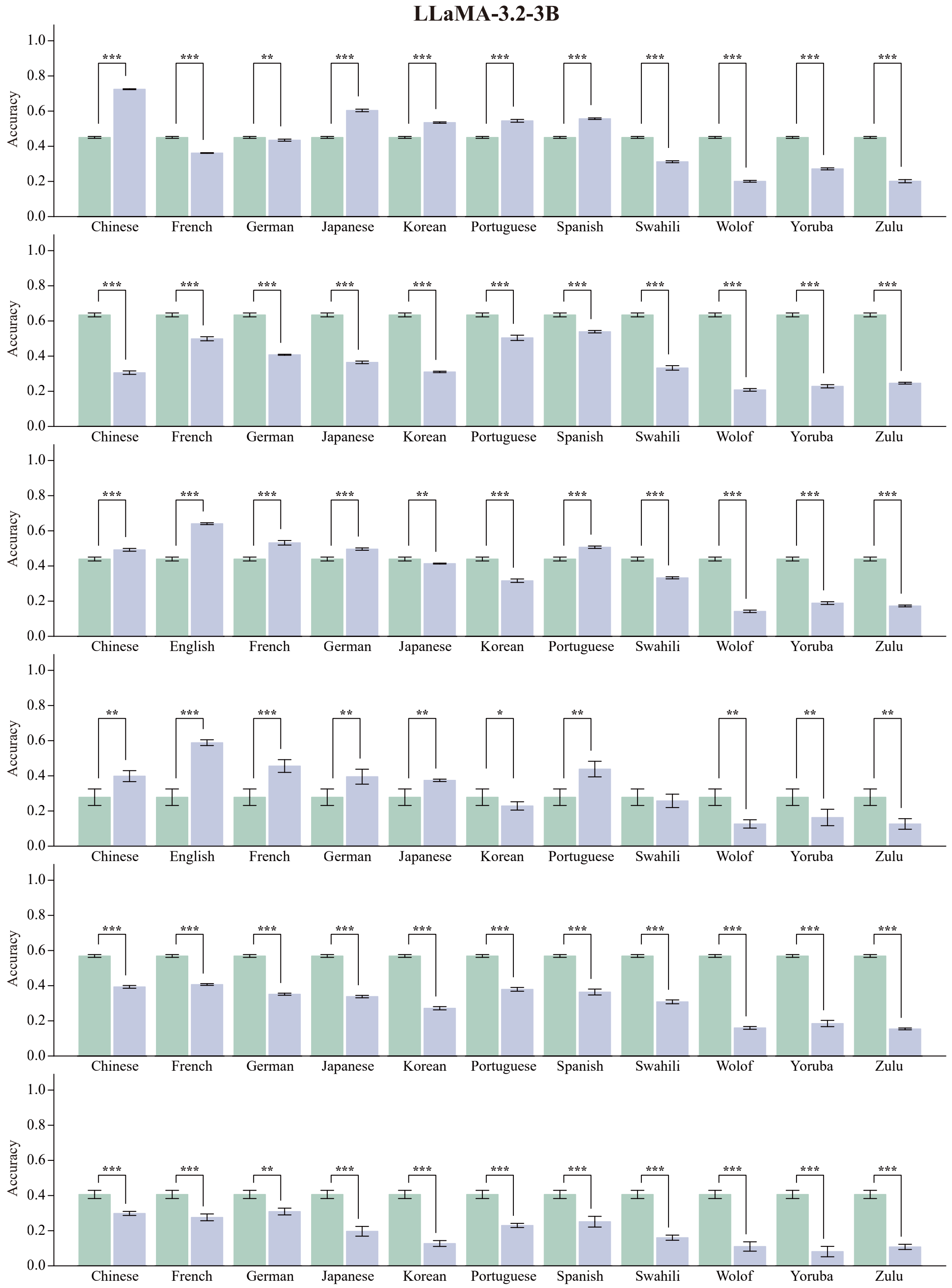

**Supplementary Fig. 24: Multilingual performance of LLaMA-3.2-3B on six medical benchmarks (top to bottom: BioNLI, MedNLI, HeadQA, MedExpQA, MedQA, MMLU-Pro).** The experiment compared the accuracy disparities between the original language (green) and target languages (purple), with each condition repeated five times. Statistical significance is indicated by asterisks (*p<0.05, **p<0.01, ***p<0.001).

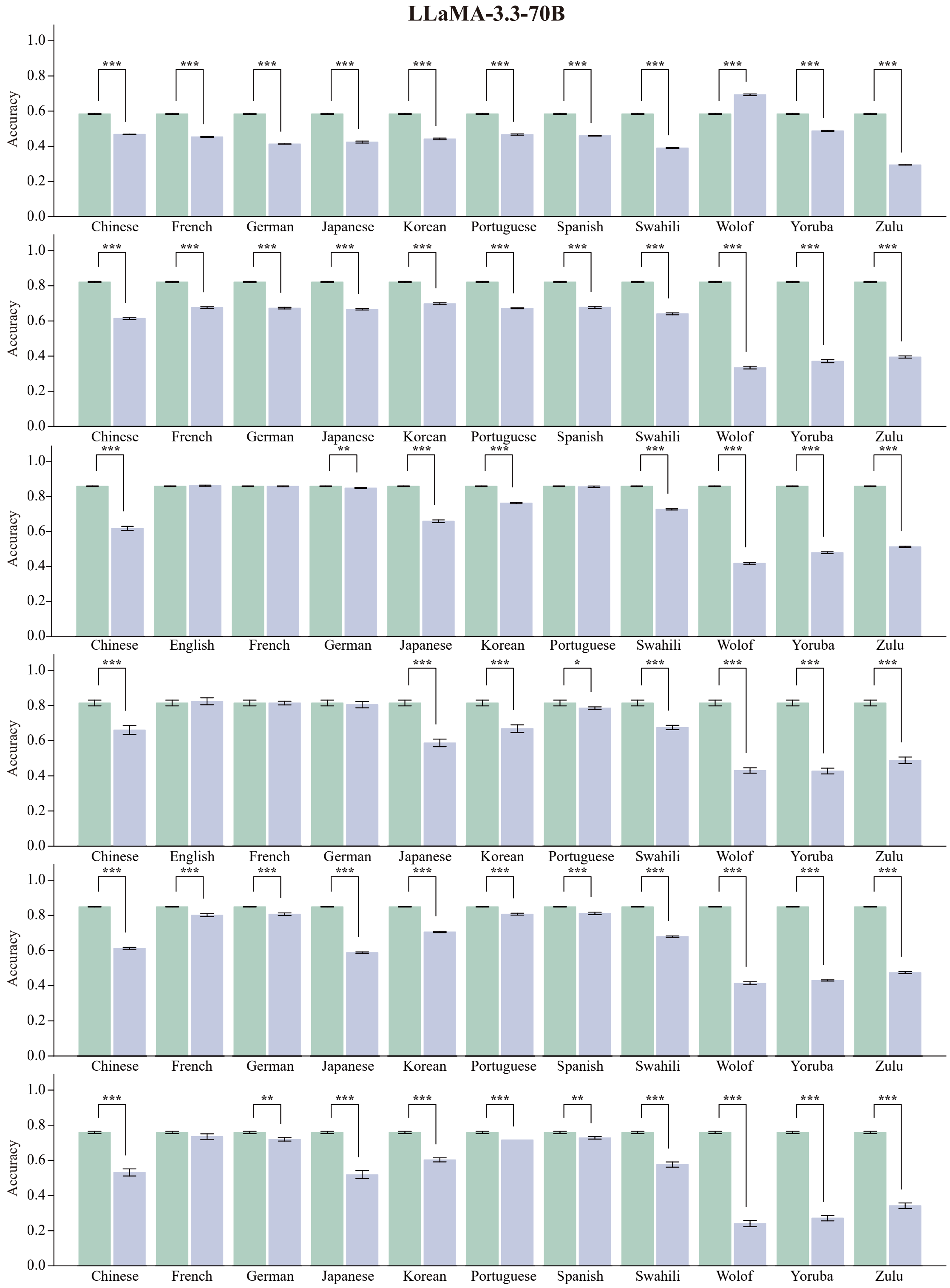

**Supplementary Fig. 25: Multilingual performance of LLaMA-3.3-70B on six medical benchmarks (top to bottom: BioNLI, MedNLI, HeadQA, MedExpQA, MedQA, MMLU-Pro).** The experiment compared the accuracy disparities between the original language (green) and target languages (purple), with each condition repeated five times. Statistical significance is indicated by asterisks (*p<0.05, **p<0.01, ***p<0.001).

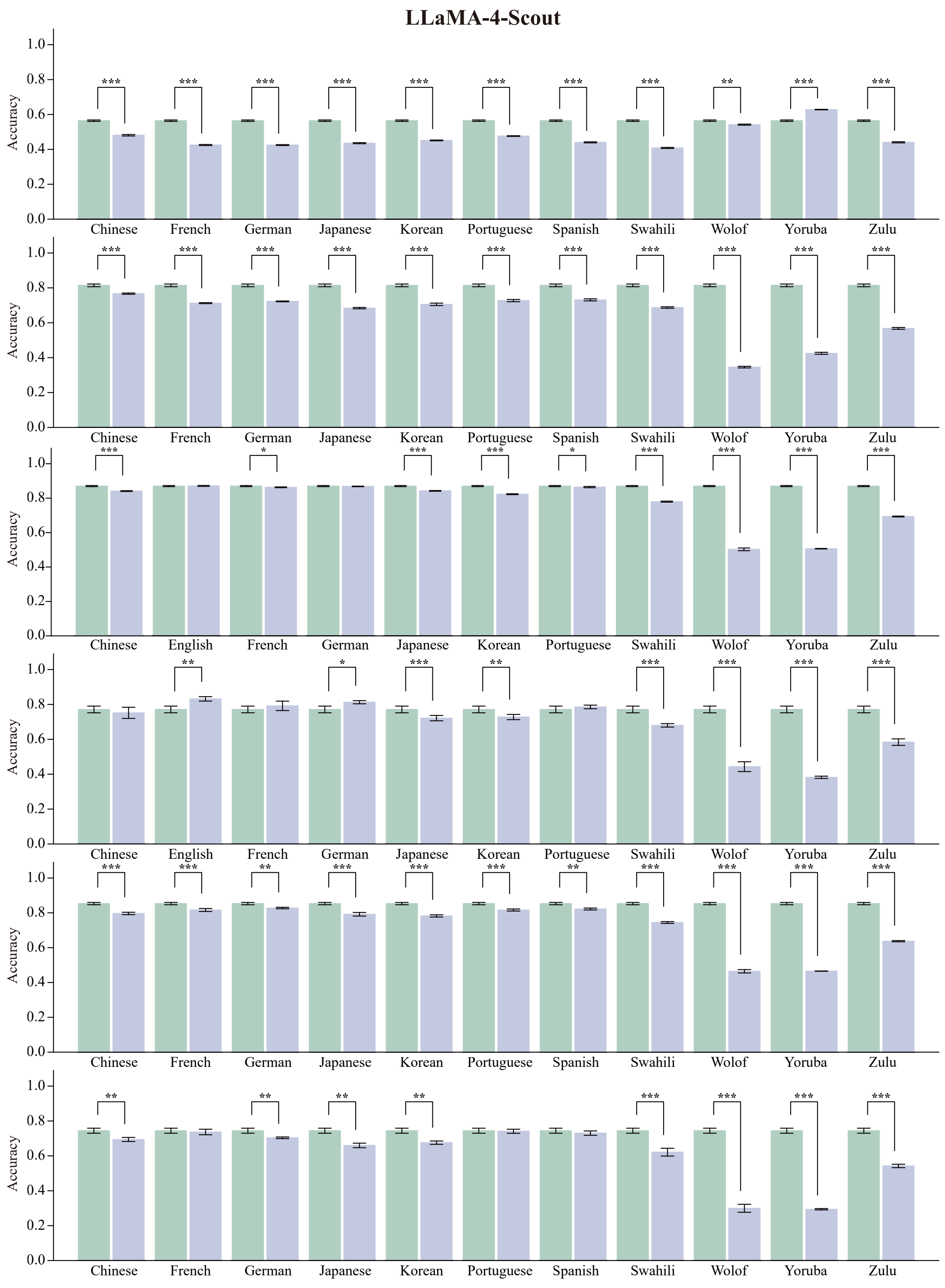

**Supplementary Fig. 26: Multilingual performance of LLaMA-4-Scout on six medical benchmarks (top to bottom: BioNLI, MedNLI, HeadQA, MedExpQA, MedQA, MMLU-Pro).** The experiment compared the accuracy disparities between the original language (green) and target languages (purple), with each condition repeated five times. Statistical significance is indicated by asterisks (*p<0.05, **p<0.01, ***p<0.001).

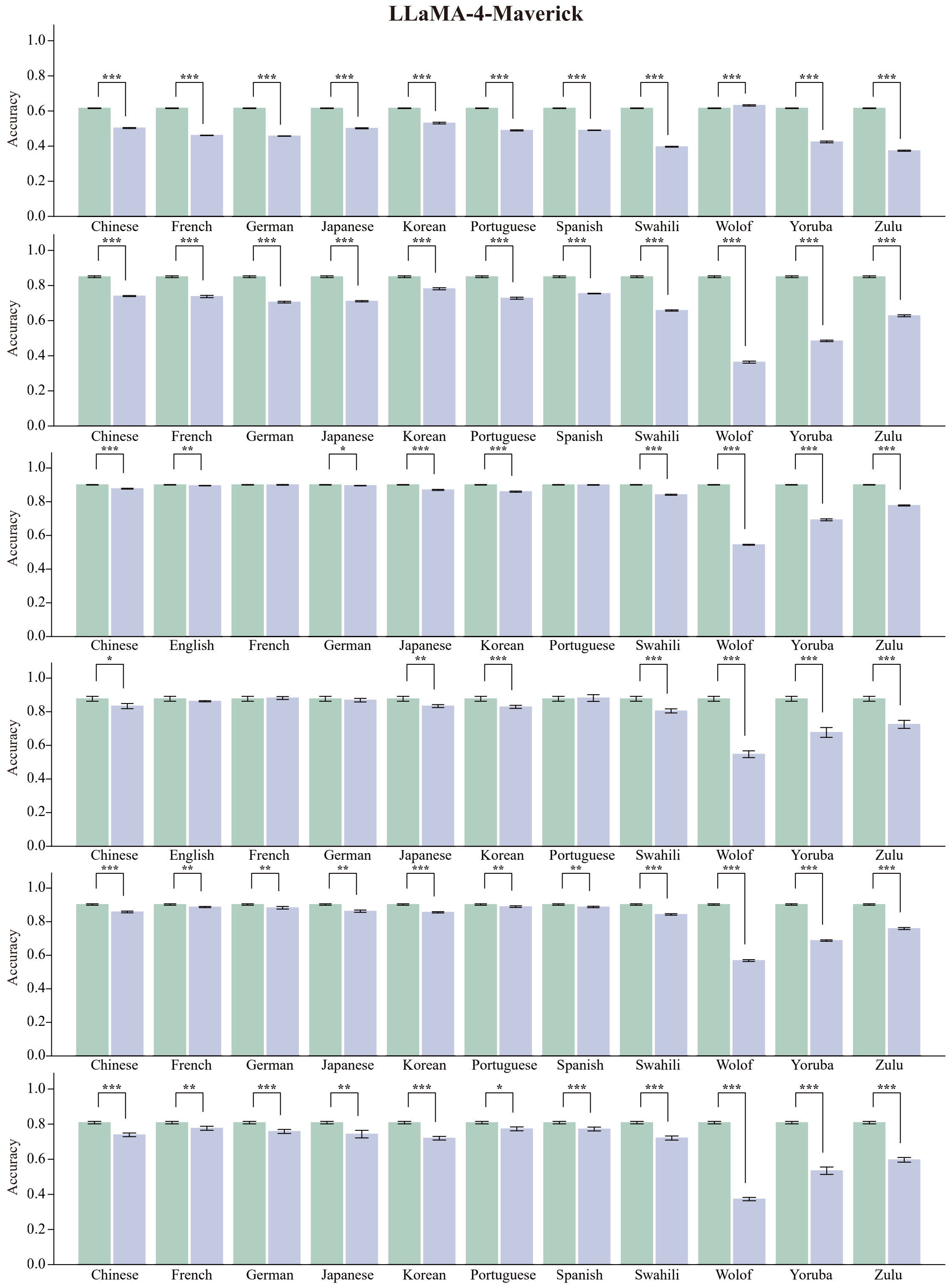

**Supplementary Fig. 27: Multilingual performance of LLaMA-4-Maverick on six medical benchmarks (top to bottom: BioNLI, MedNLI, HeadQA, MedExpQA, MedQA, MMLU-Pro).** The experiment compared the accuracy disparities between the original language (green) and target languages (purple), with each condition repeated five times. Statistical significance is indicated by asterisks (*p<0.05, **p<0.01, ***p<0.001).

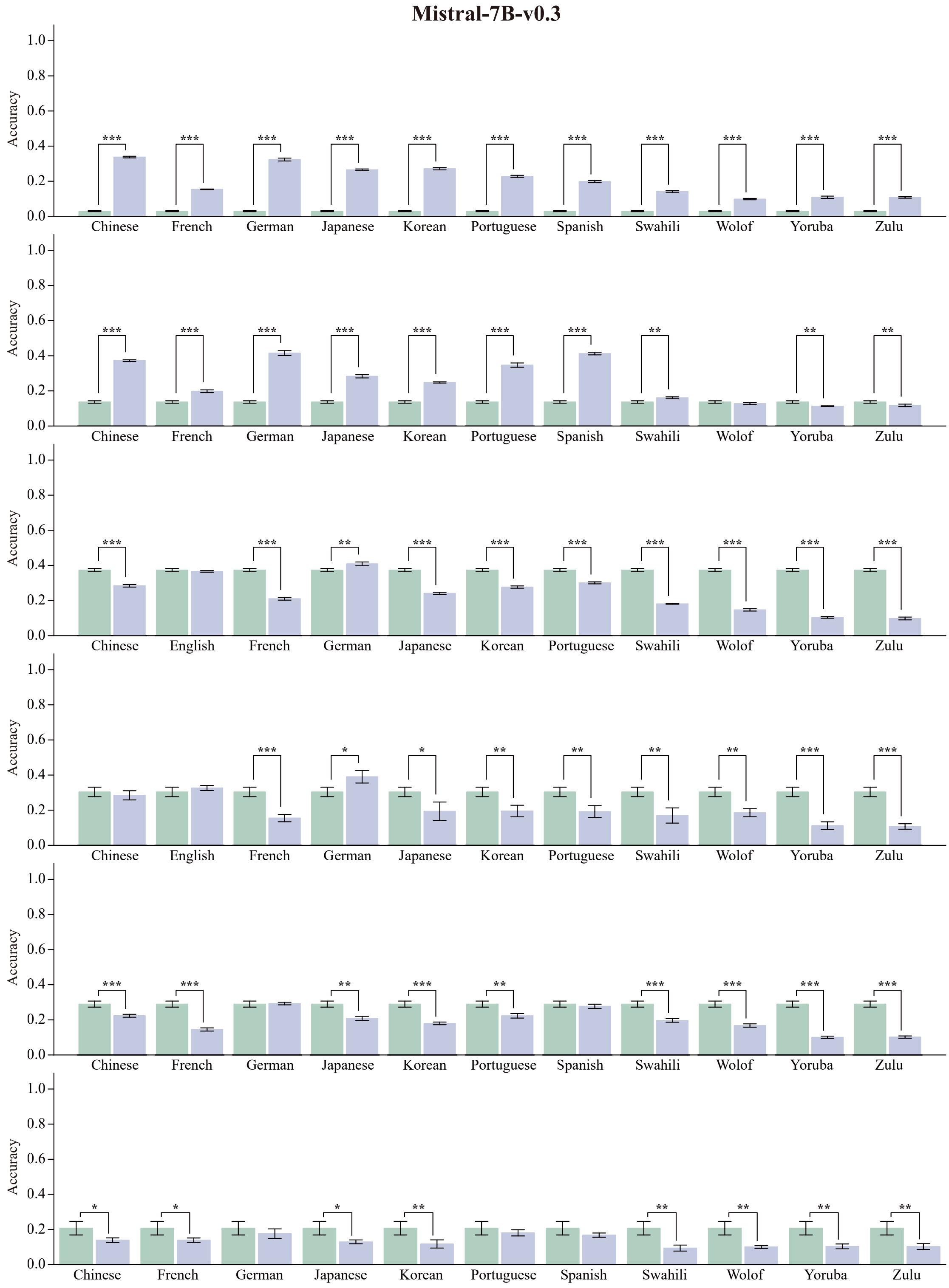

**Supplementary Fig. 28: Multilingual performance of Mistral-7B-v0.3 on six medical benchmarks (top to bottom: BioNLI, MedNLI, HeadQA, MedExpQA, MedQA, MMLU-Pro).** The experiment compared the accuracy disparities between the original language (green) and target languages (purple), with each condition repeated five times. Statistical significance is indicated by asterisks (*p<0.05, **p<0.01, ***p<0.001).

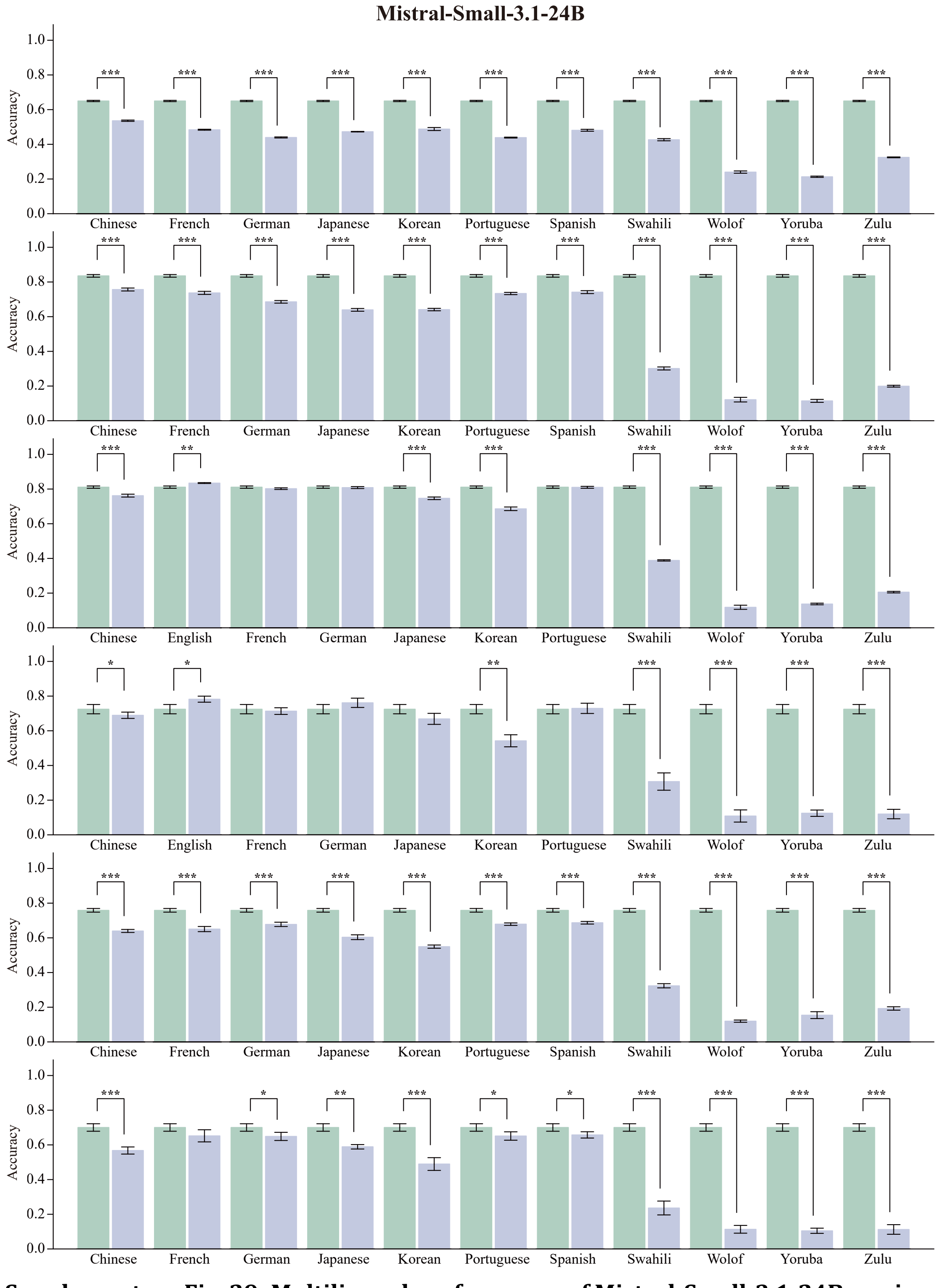

**Supplementary Fig. 29: Multilingual performance of Mistral-Small-3.1-24B on six medical benchmarks (top to bottom: BioNLI, MedNLI, HeadQA, MedExpQA, MedQA, MMLU-Pro).** The experiment compared the accuracy disparities between the original language (green) and target languages (purple), with each condition repeated five times. Statistical significance is indicated by asterisks (*p<0.05, **p<0.01, ***p<0.001).

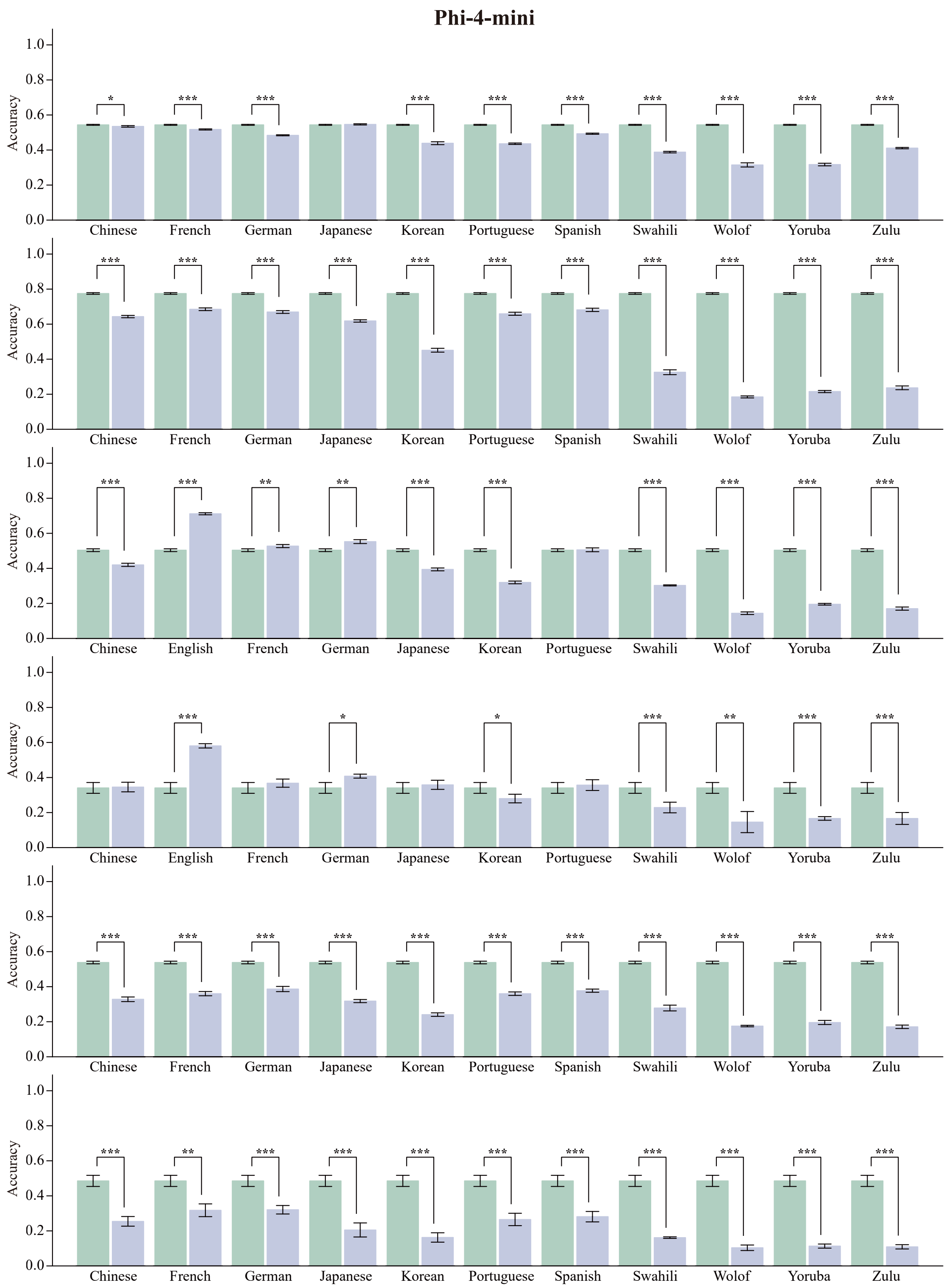

**Supplementary Fig. 30: Multilingual performance of Phi-4-mini on six medical benchmarks (top to bottom: BioNLI, MedNLI, HeadQA, MedExpQA, MedQA, MMLU-Pro).** The experiment compared the accuracy disparities between the original language (green) and target languages (purple), with each condition repeated five times. Statistical significance is indicated by asterisks (*p<0.05, **p<0.01, ***p<0.001).

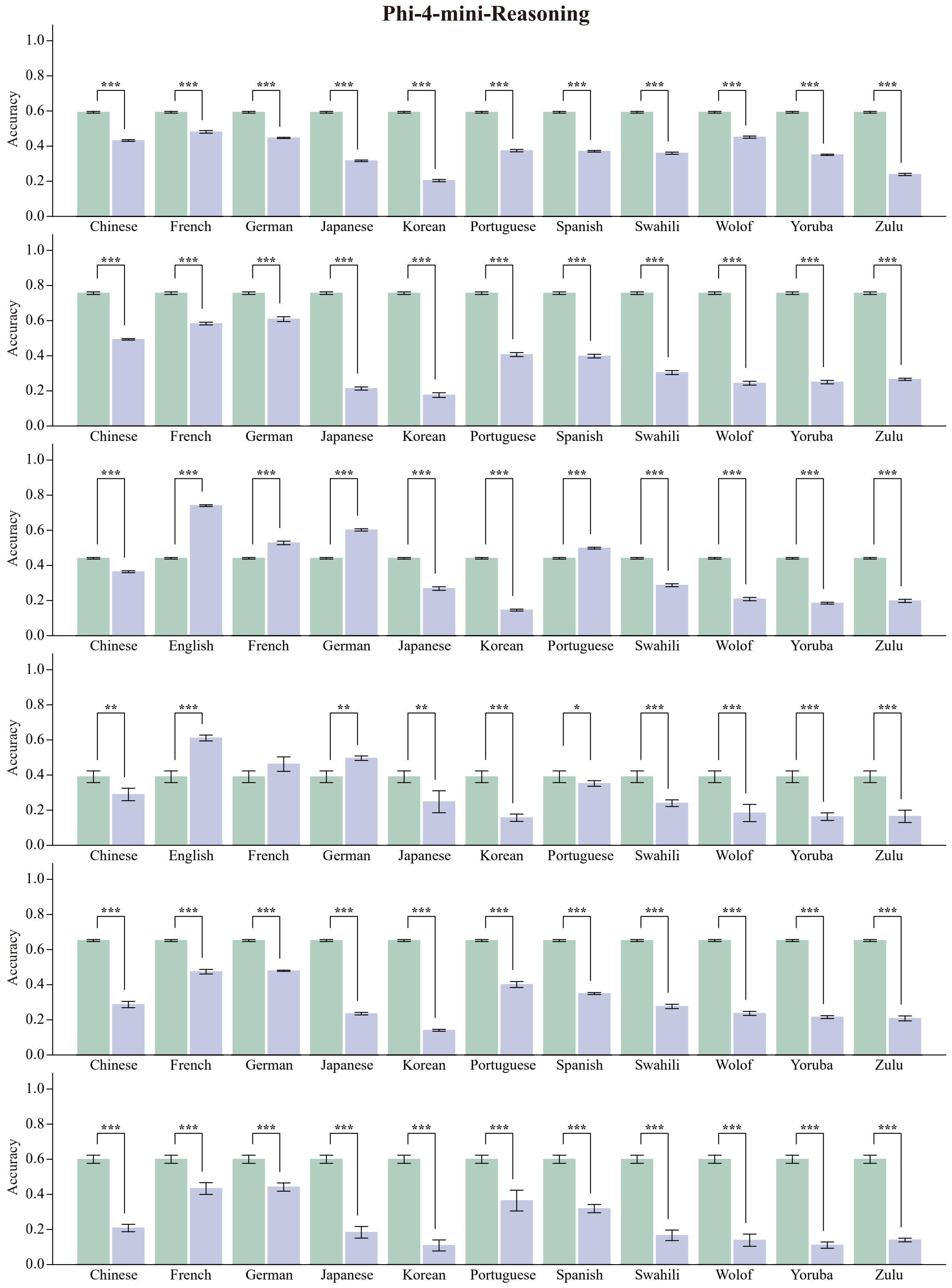

**Supplementary Fig. 31: Multilingual performance of Phi-4-mini-Reasoning on six medical benchmarks (top to bottom: BioNLI, MedNLI, HeadQA, MedExpQA, MedQA, MMLU-Pro).** The experiment compared the accuracy disparities between the original language (green) and target languages (purple), with each condition repeated five times. Statistical significance is indicated by asterisks (*p<0.05, **p<0.01, ***p<0.001).

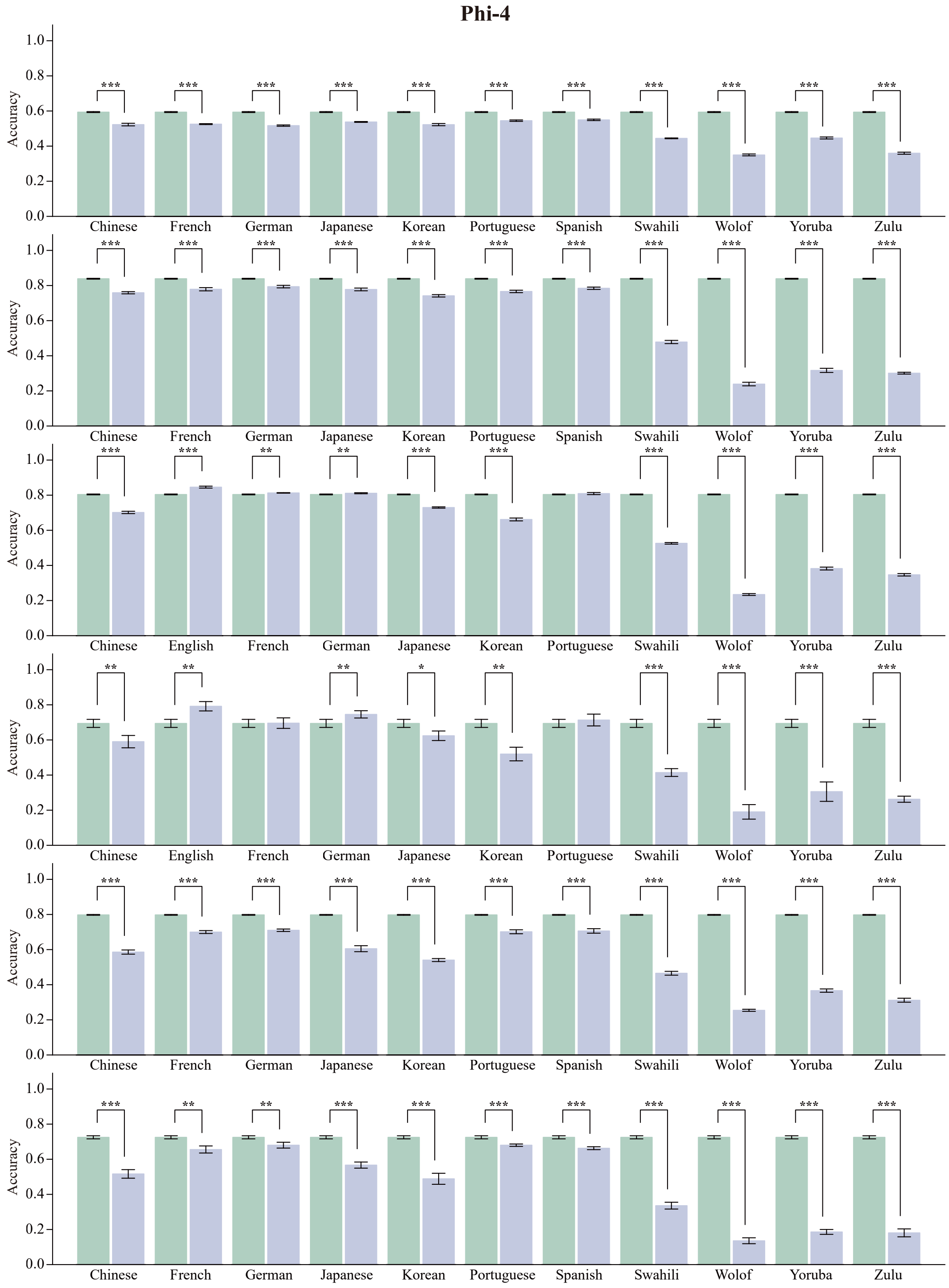

**Supplementary Fig. 32: Multilingual performance of Phi-4 on six medical benchmarks (top to bottom: BioNLI, MedNLI, HeadQA, MedExpQA, MedQA, MMLU-Pro).** The experiment compared the accuracy disparities between the original language (green) and target languages (purple), with each condition repeated five times. Statistical significance is indicated by asterisks (*p<0.05, **p<0.01, ***p<0.001).

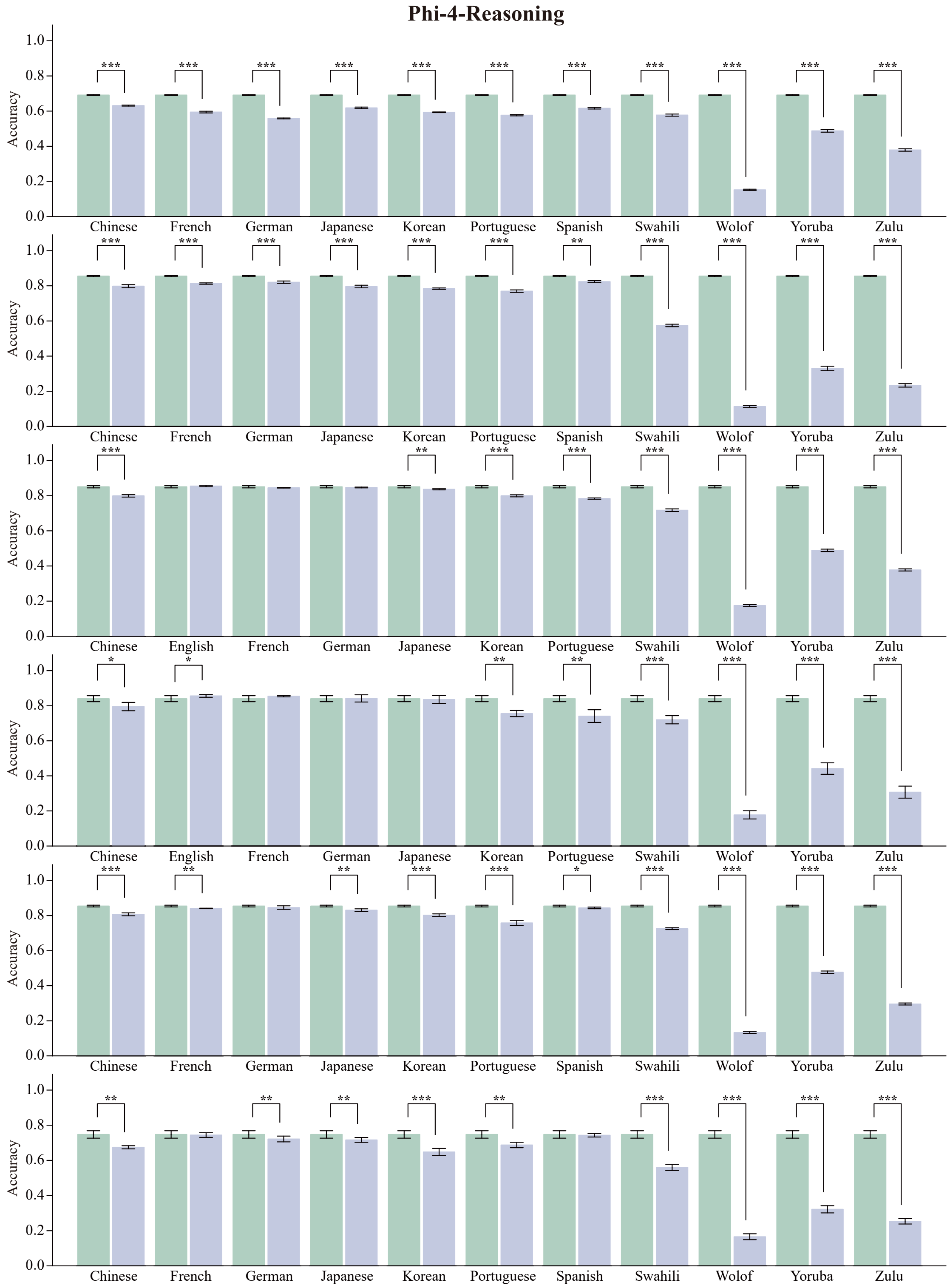

**Supplementary Fig. 33: Multilingual performance of Phi-4-Reasoning on six medical benchmarks (top to bottom: BioNLI, MedNLI, HeadQA, MedExpQA, MedQA, MMLU-Pro).** The experiment compared the accuracy disparities between the original language (green) and target languages (purple), with each condition repeated five times. Statistical significance is indicated by asterisks (*p<0.05, **p<0.01, ***p<0.001).

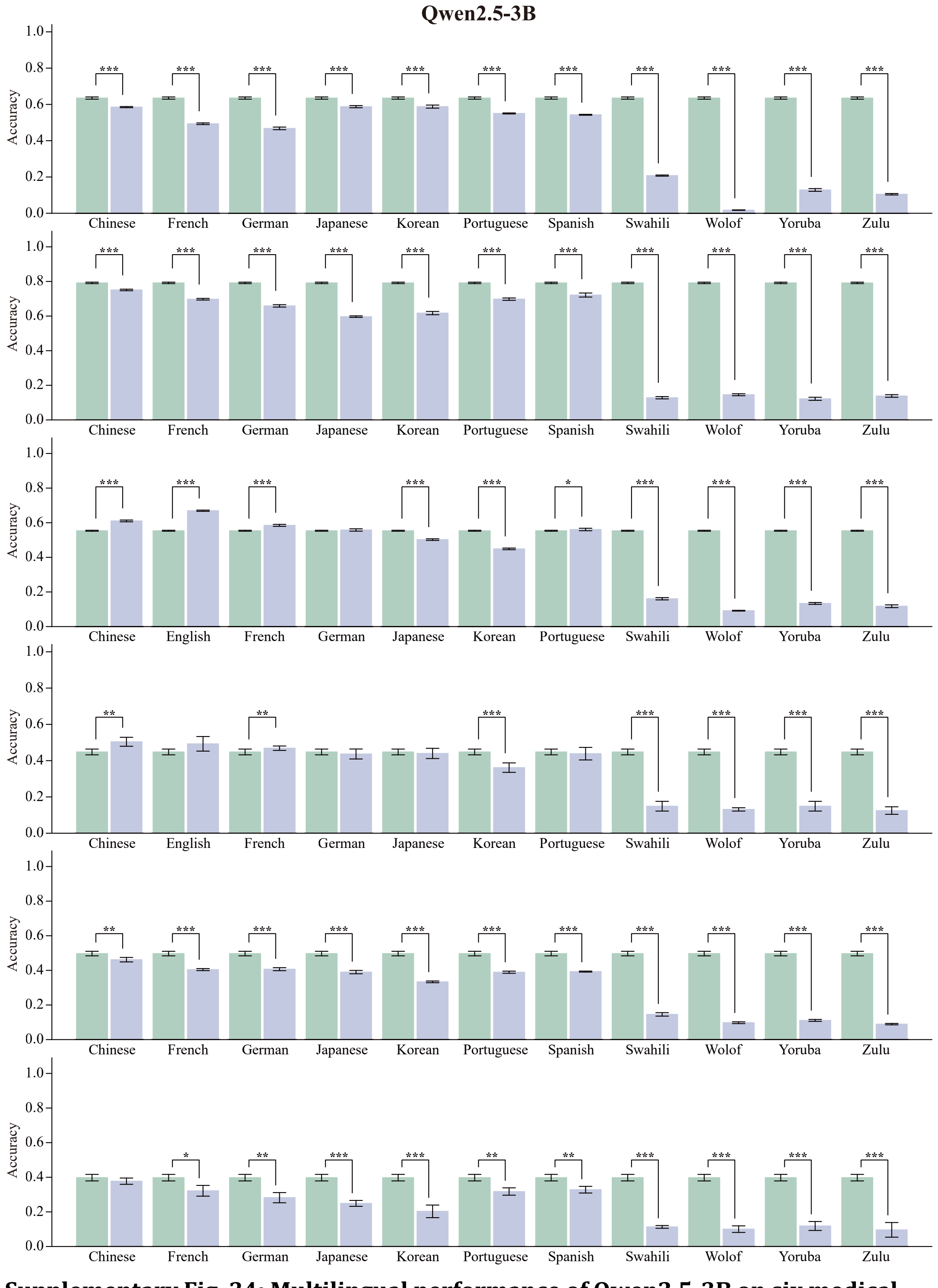

**Supplementary Fig. 34: Multilingual performance of Qwen2.5-3B on six medical benchmarks (top to bottom: BioNLI, MedNLI, HeadQA, MedExpQA, MedQA, MMLU-Pro).** The experiment compared the accuracy disparities between the original language (green) and target languages (purple), with each condition repeated five times. Statistical significance is indicated by asterisks (*p<0.05, **p<0.01, ***p<0.001).

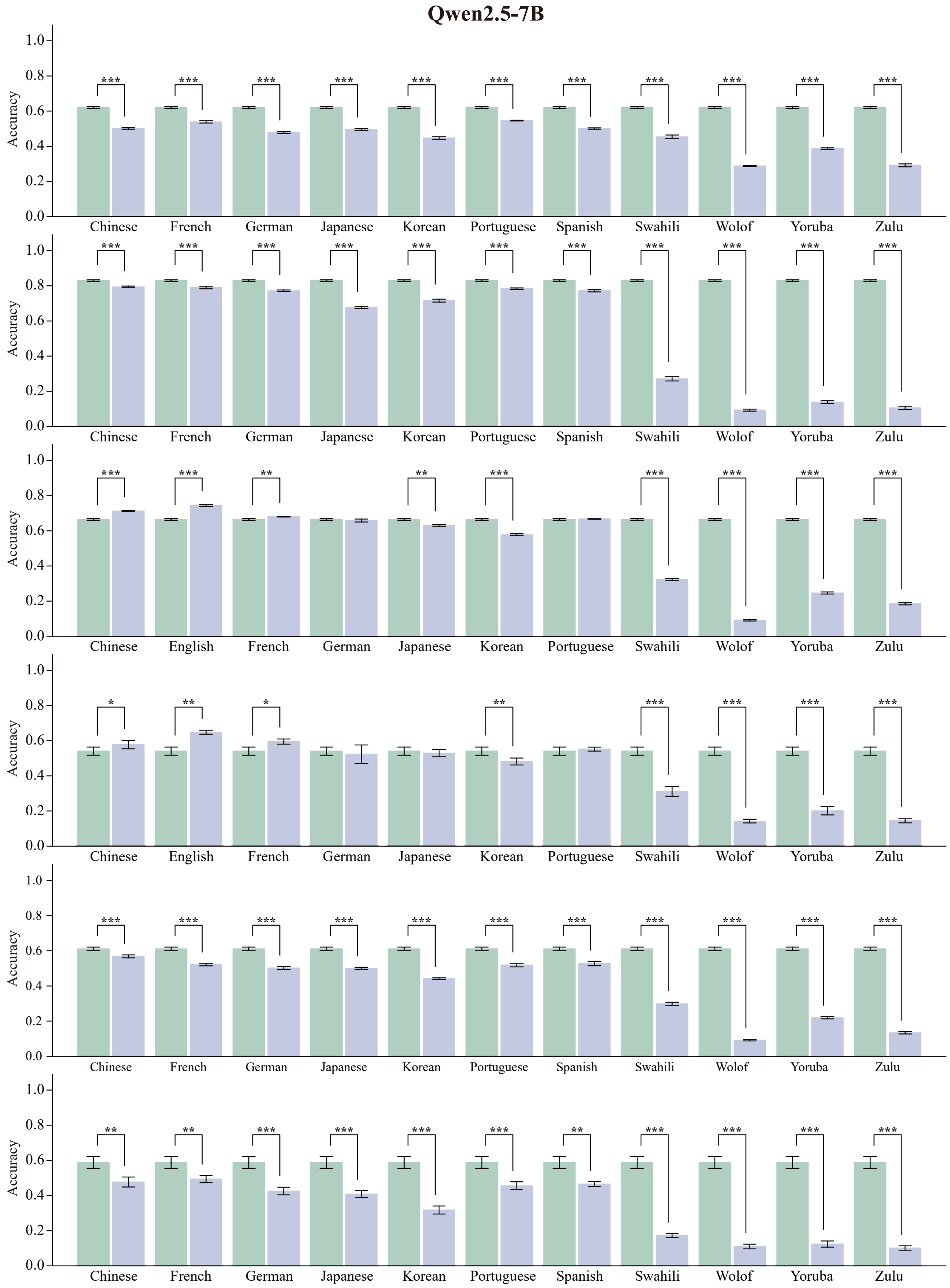

**Supplementary Fig. 35: Multilingual performance of Qwen2.5-7B on six medical benchmarks (top to bottom: BioNLI, MedNLI, HeadQA, MedExpQA, MedQA, MMLU-Pro).** The experiment compared the accuracy disparities between the original language (green) and target languages (purple), with each condition repeated five times. Statistical significance is indicated by asterisks (*p<0.05, **p<0.01, ***p<0.001).

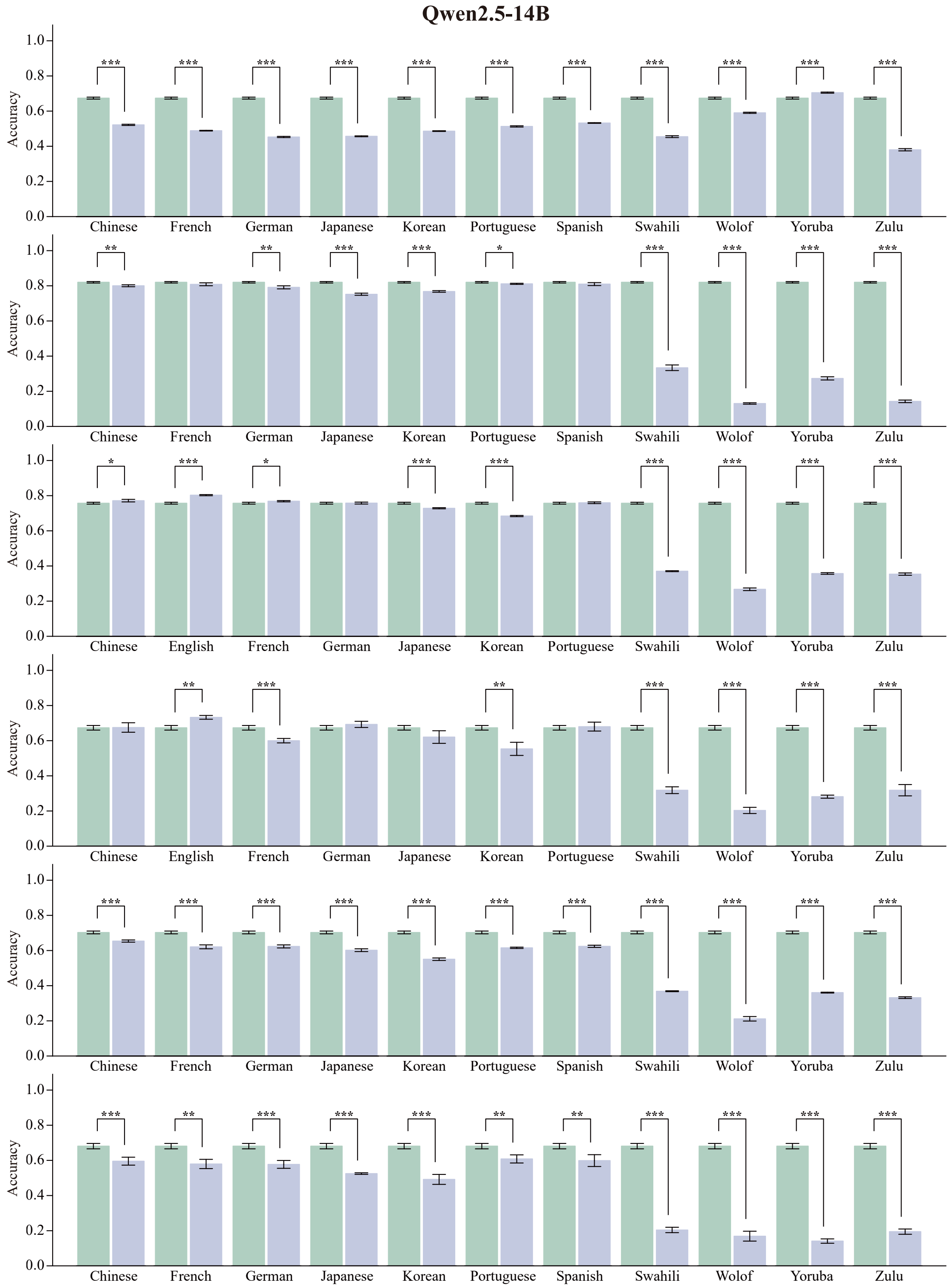

**Supplementary Fig. 36: Multilingual performance of Qwen2.5-14B on six medical benchmarks (top to bottom: BioNLI, MedNLI, HeadQA, MedExpQA, MedQA, MMLU-Pro).** The experiment compared the accuracy disparities between the original language (green) and target languages (purple), with each condition repeated five times. Statistical significance is indicated by asterisks (*p<0.05, **p<0.01, ***p<0.001).

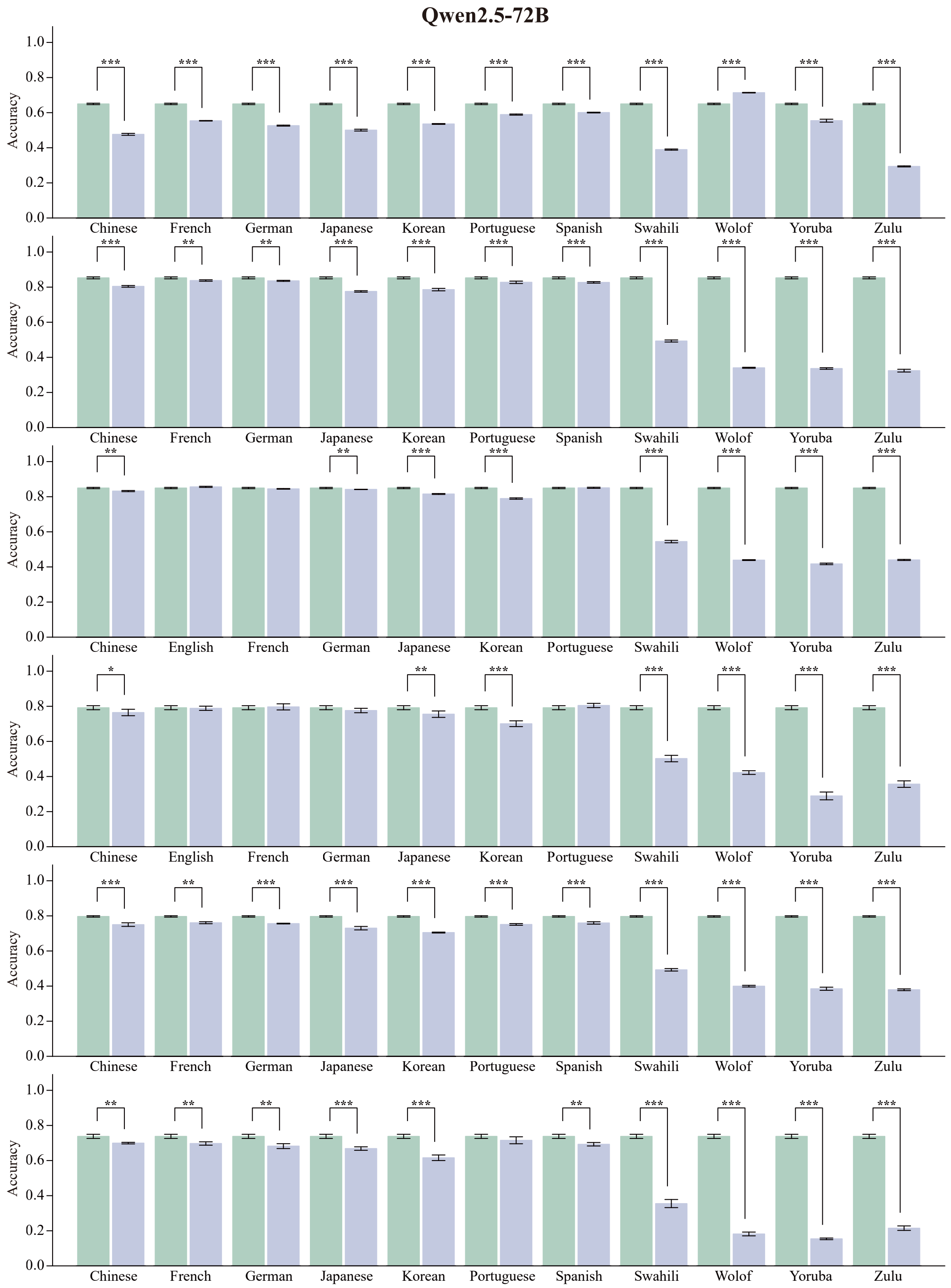

**Supplementary Fig. 37: Multilingual performance of Qwen2.5-72B on six medical benchmarks (top to bottom: BioNLI, MedNLI, HeadQA, MedExpQA, MedQA, MMLU-Pro).** The experiment compared the accuracy disparities between the original language (green) and target languages (purple), with each condition repeated five times. Statistical significance is indicated by asterisks (*p<0.05, **p<0.01, ***p<0.001).

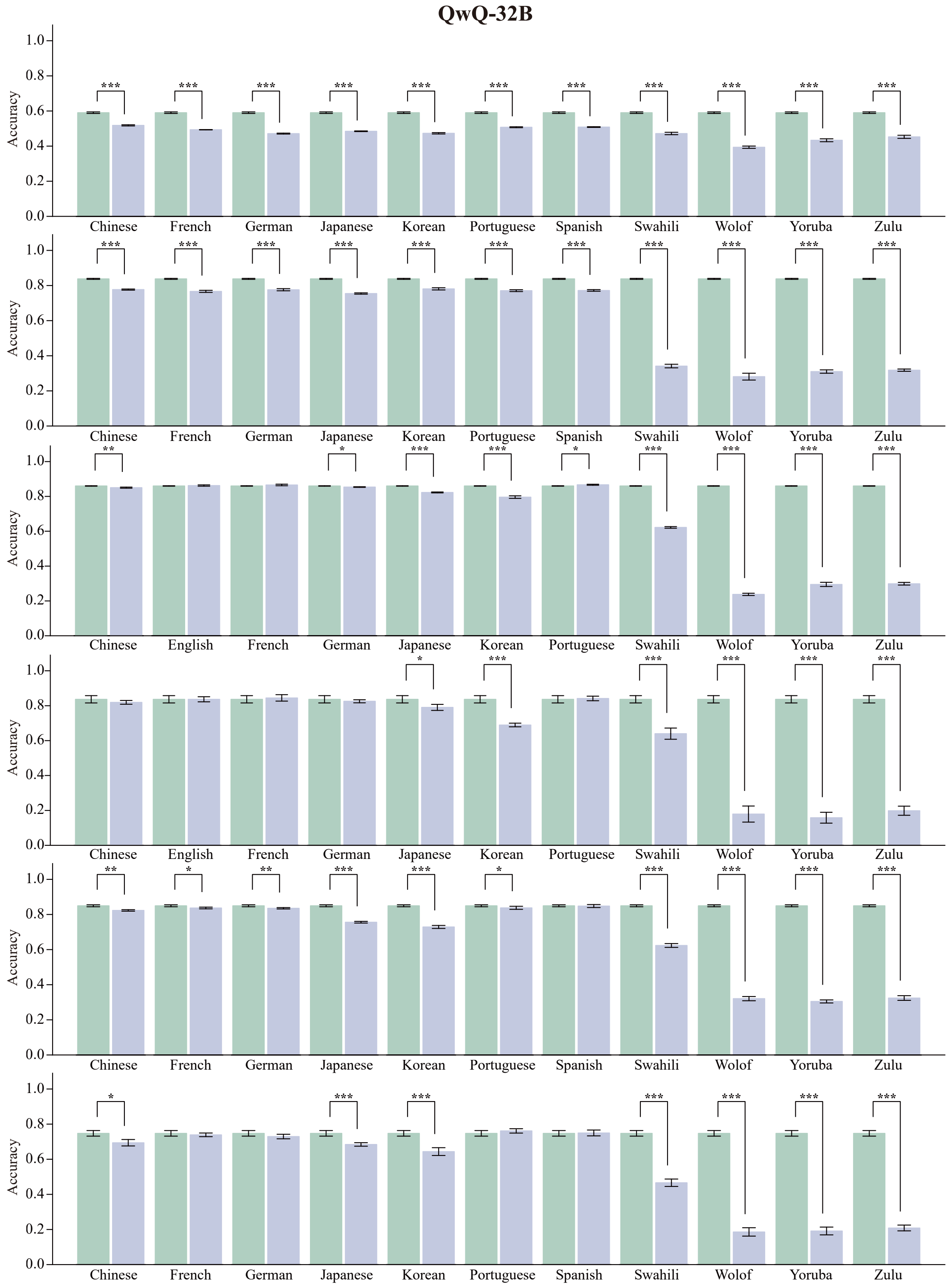

**Supplementary Fig. 38: Multilingual performance of QwQ-32B on six medical benchmarks (top to bottom: BioNLI, MedNLI, HeadQA, MedExpQA, MedQA, MMLU-Pro).** The experiment compared the accuracy disparities between the original language (green) and target languages (purple), with each condition repeated five times. Statistical significance is indicated by asterisks (*p<0.05, **p<0.01, ***p<0.001).

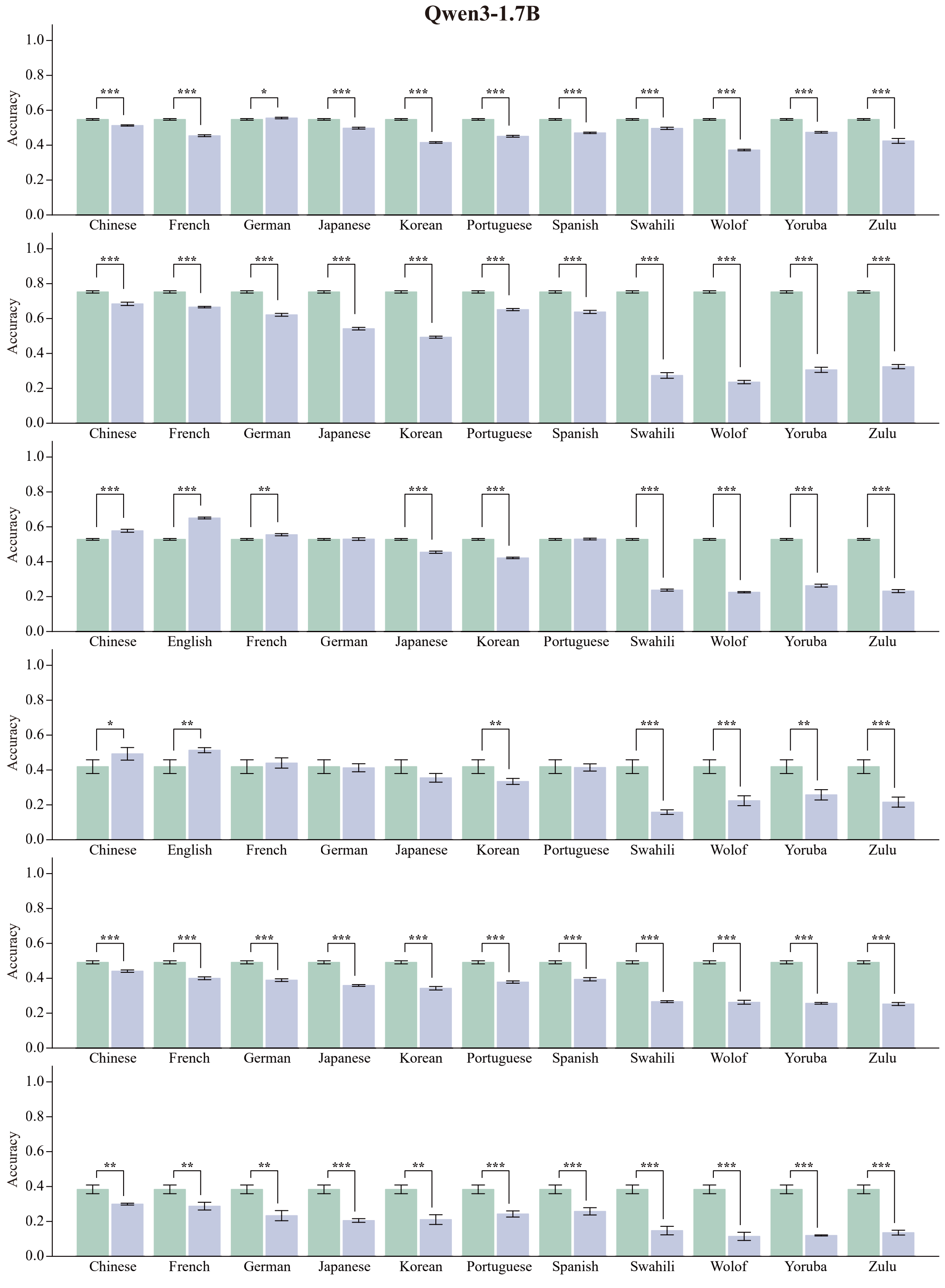

**Supplementary Fig. 39: Multilingual performance of Qwen3-1.7B on six medical benchmarks (top to bottom: BioNLI, MedNLI, HeadQA, MedExpQA, MedQA, MMLU-Pro).** The experiment compared the accuracy disparities between the original language (green) and target languages (purple), with each condition repeated five times. Statistical significance is indicated by asterisks (*p<0.05, **p<0.01, ***p<0.001).

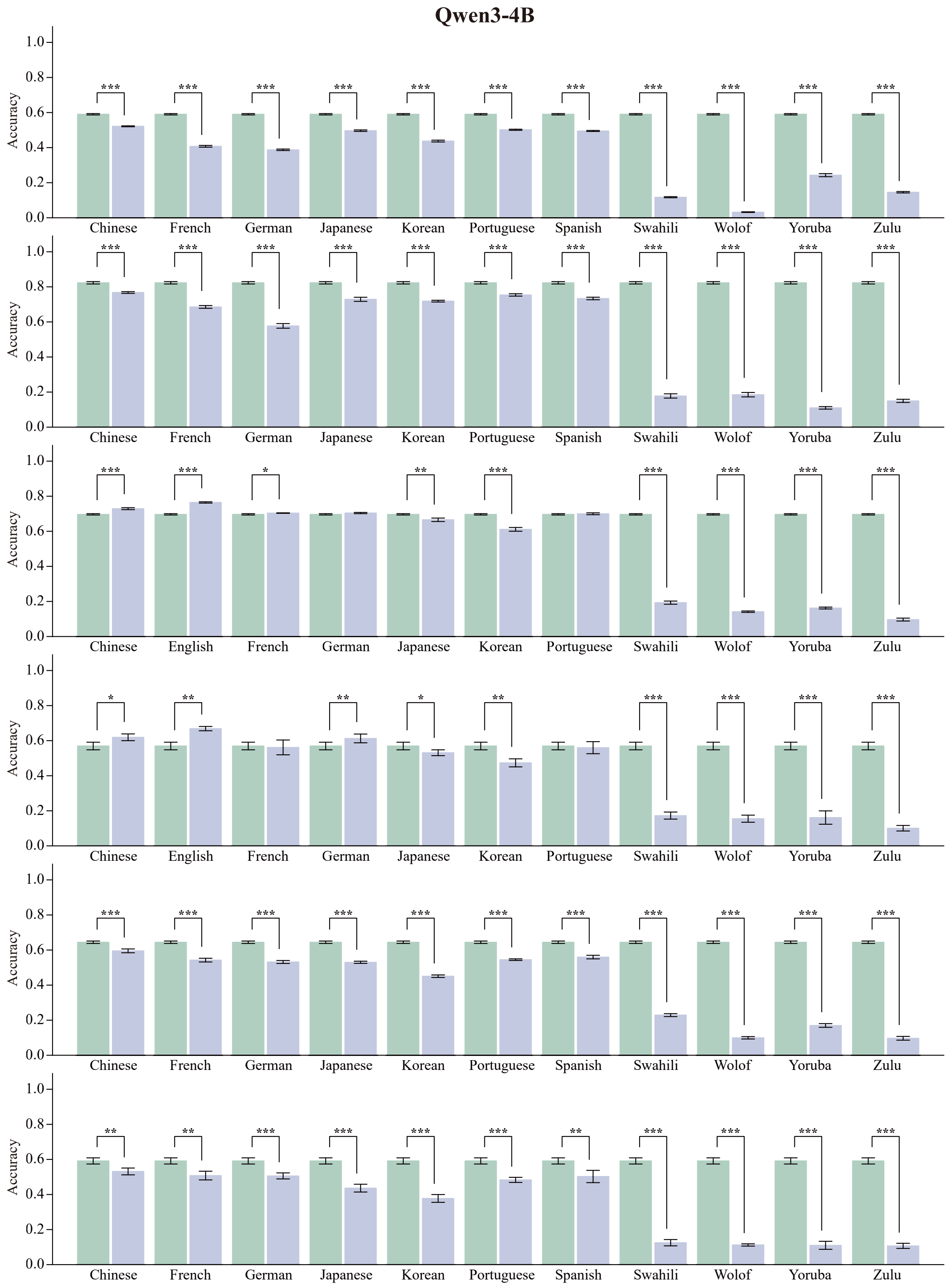

**Supplementary Fig. 40: Multilingual performance of Qwen3-4B on six medical benchmarks (top to bottom: BioNLI, MedNLI, HeadQA, MedExpQA, MedQA, MMLU-Pro).** The experiment compared the accuracy disparities between the original language (green) and target languages (purple), with each condition repeated five times. Statistical significance is indicated by asterisks (*p<0.05, **p<0.01, ***p<0.001).

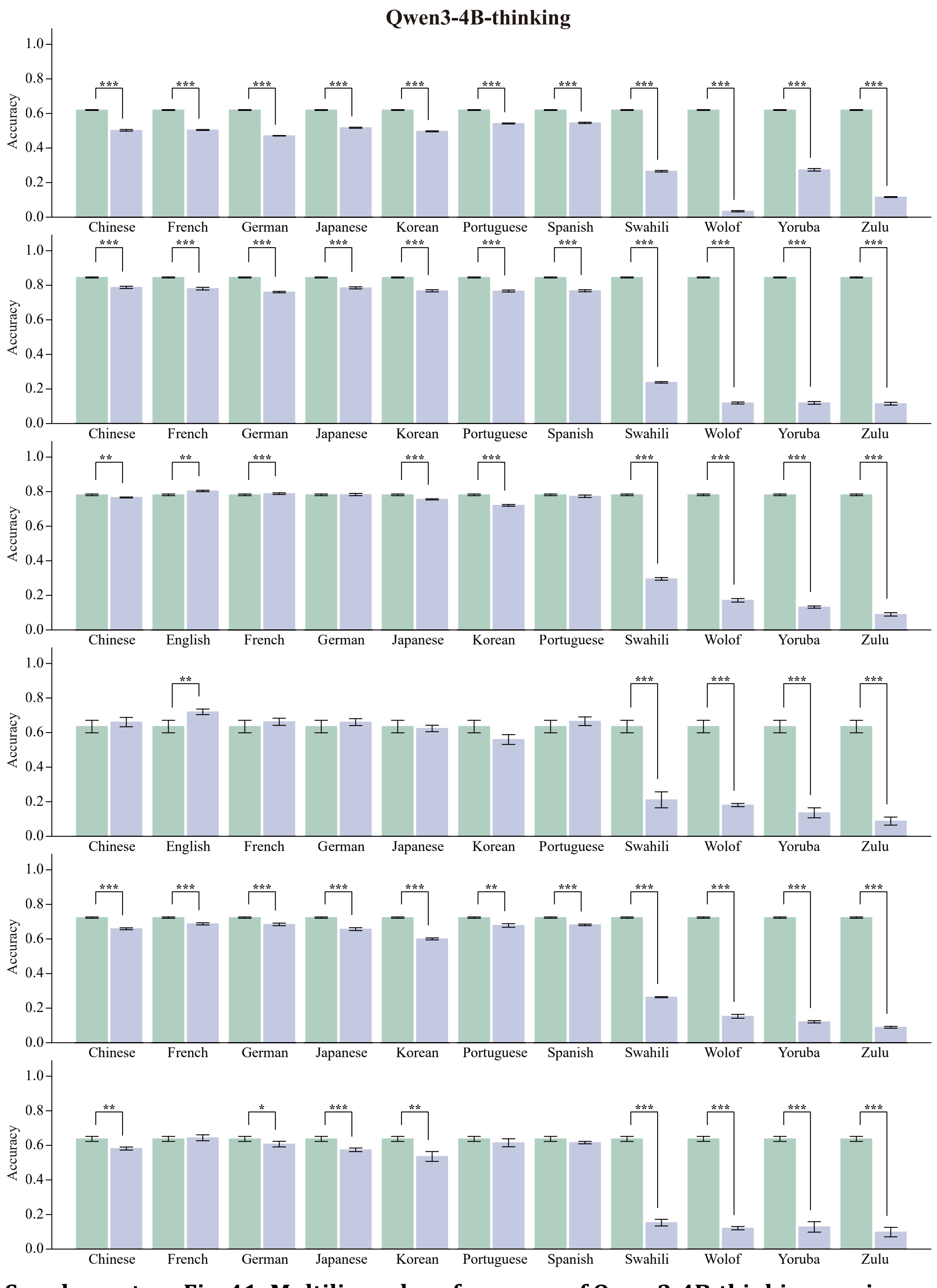

**Supplementary Fig. 41: Multilingual performance of Qwen3-4B-thinking on six medical benchmarks (top to bottom: BioNLI, MedNLI, HeadQA, MedExpQA, MedQA, MMLU-Pro).** The experiment compared the accuracy disparities between the original language (green) and target languages (purple), with each condition repeated five times. Statistical significance is indicated by asterisks (*p<0.05, **p<0.01, ***p<0.001).

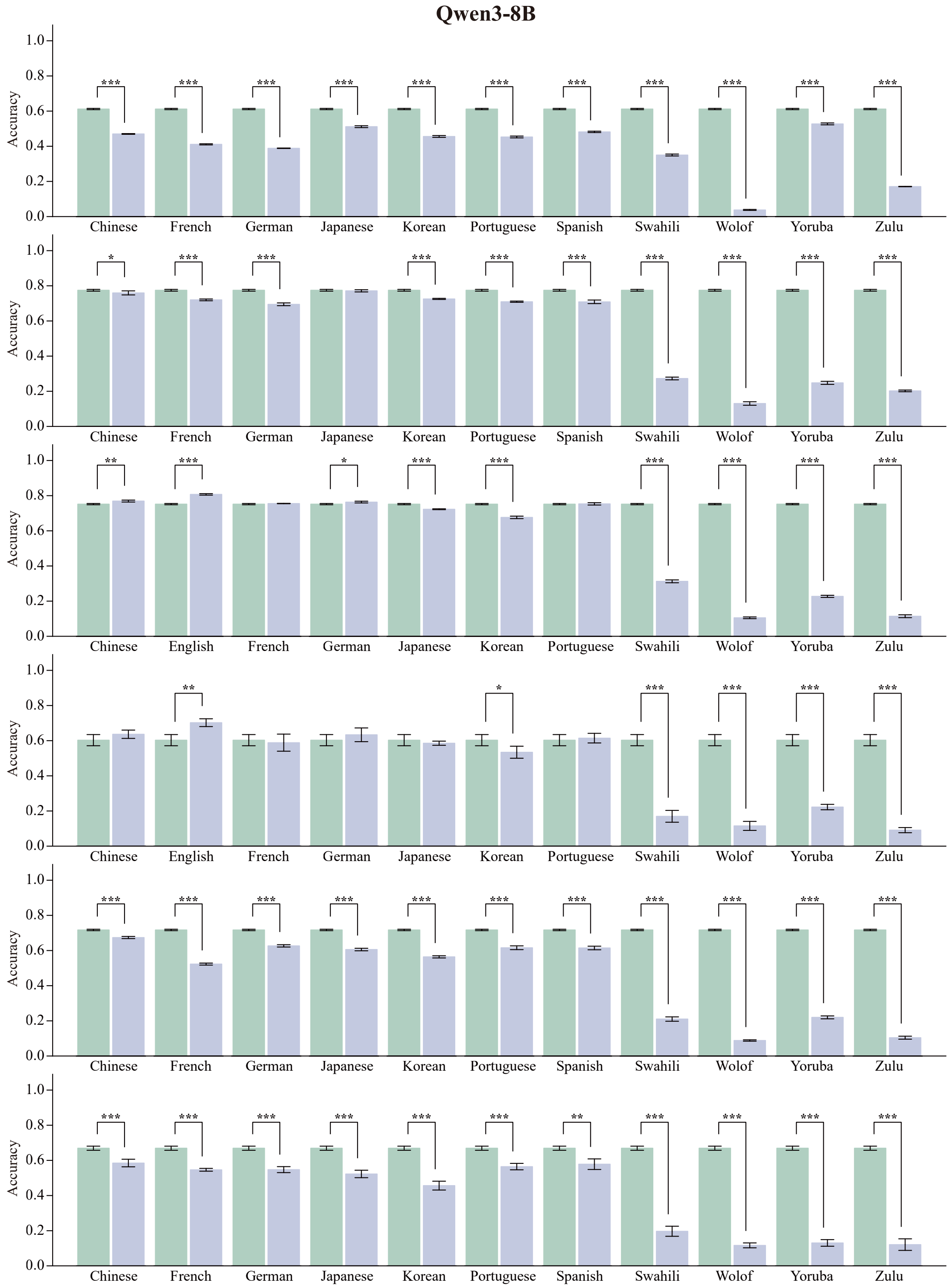

**Supplementary Fig. 42: Multilingual performance of Qwen3-8B on six medical benchmarks (top to bottom: BioNLI, MedNLI, HeadQA, MedExpQA, MedQA, MMLU-Pro).** The experiment compared the accuracy disparities between the original language (green) and target languages (purple), with each condition repeated five times. Statistical significance is indicated by asterisks (*p<0.05, **p<0.01, ***p<0.001).

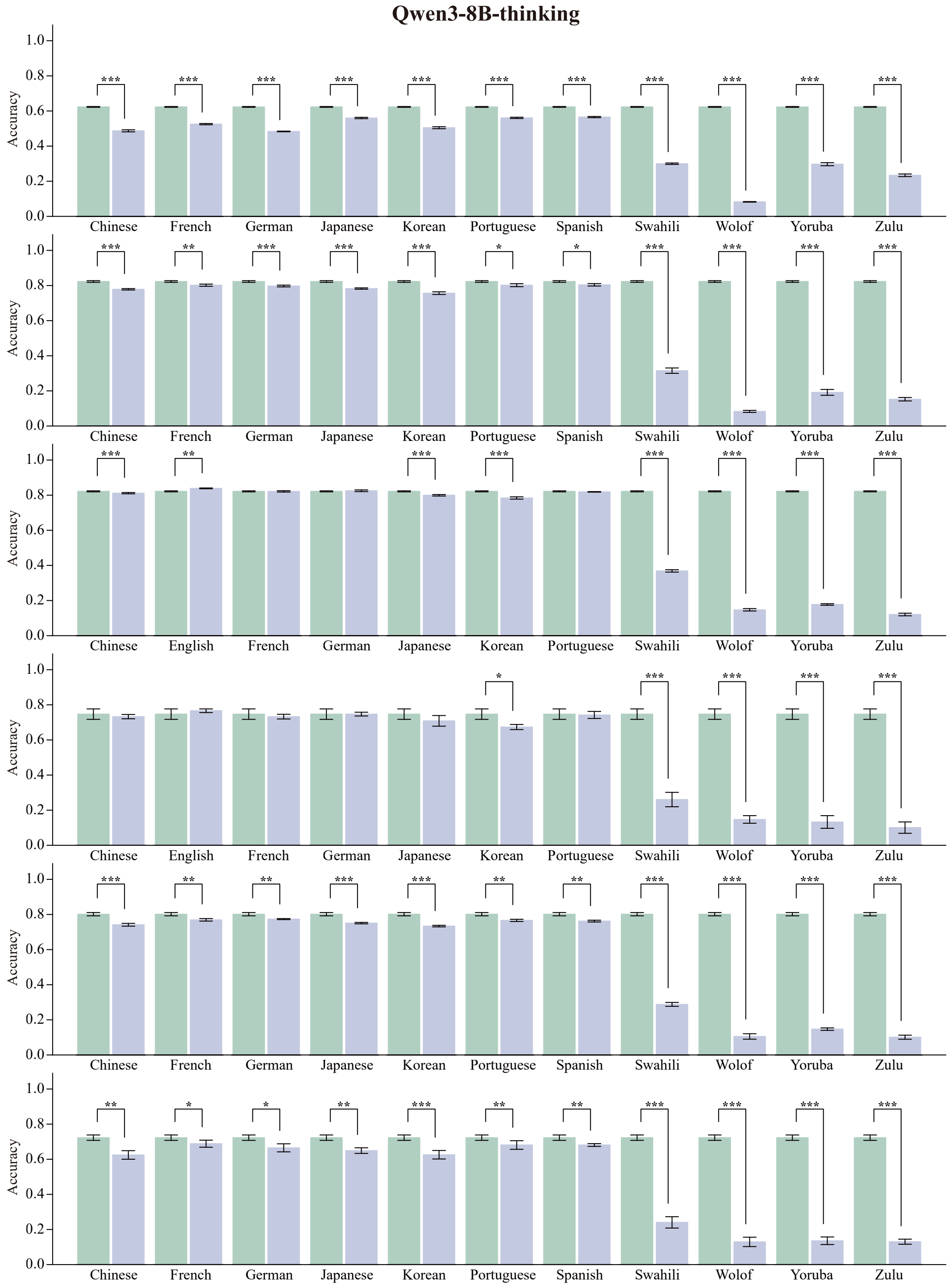

**Supplementary Fig. 43: Multilingual performance of Qwen3-8B-thinking on six medical benchmarks (top to bottom: BioNLI, MedNLI, HeadQA, MedExpQA, MedQA, MMLU-Pro).** The experiment compared the accuracy disparities between the original language (green) and target languages (purple), with each condition repeated five times. Statistical significance is indicated by asterisks (*p<0.05, **p<0.01, ***p<0.001).

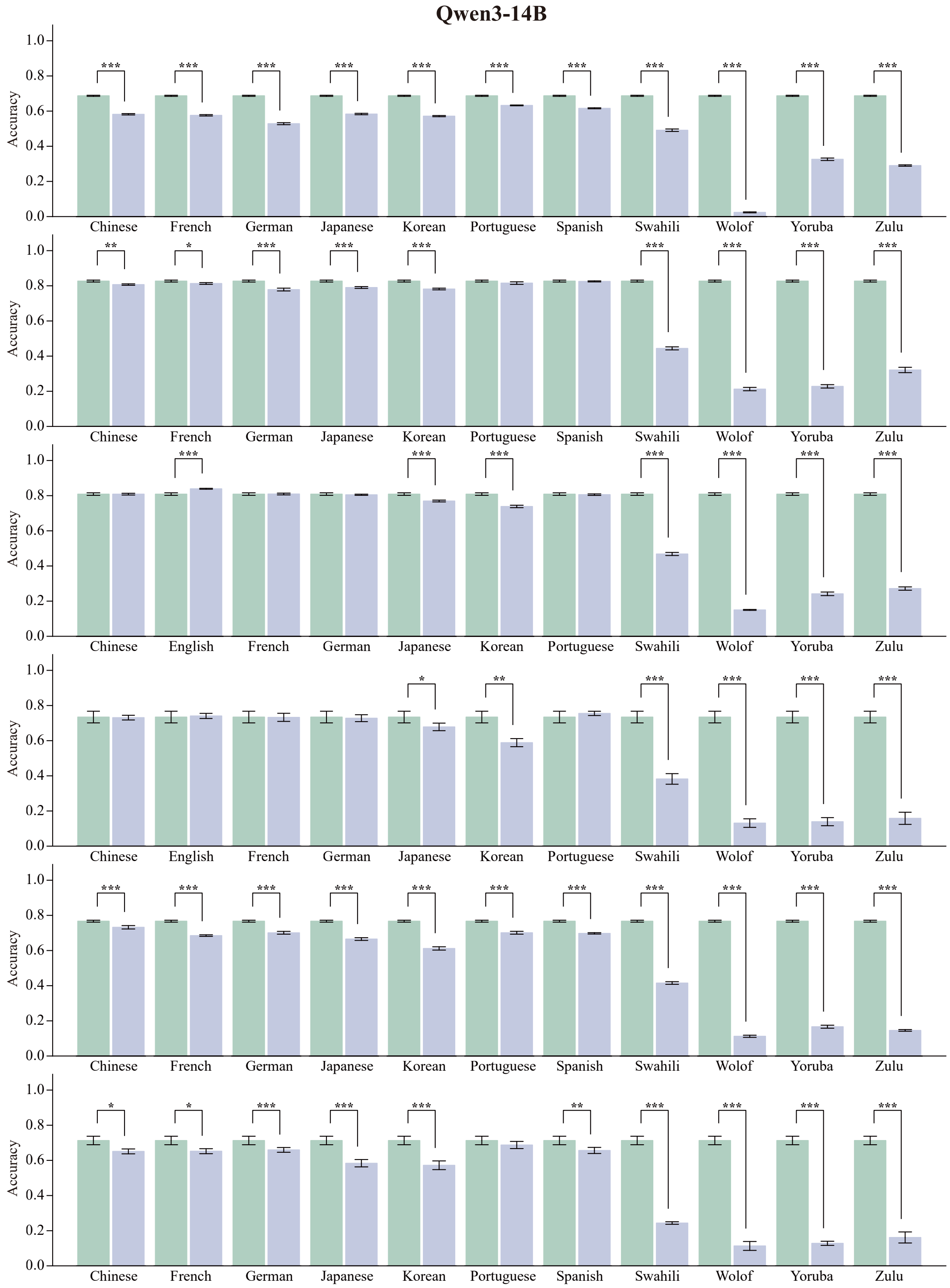

**Supplementary Fig. 44: Multilingual performance of Qwen3-14B on six medical benchmarks (top to bottom: BioNLI, MedNLI, HeadQA, MedExpQA, MedQA, MMLU-Pro).** The experiment compared the accuracy disparities between the original language (green) and target languages (purple), with each condition repeated five times. Statistical significance is indicated by asterisks (*p<0.05, **p<0.01, ***p<0.001).

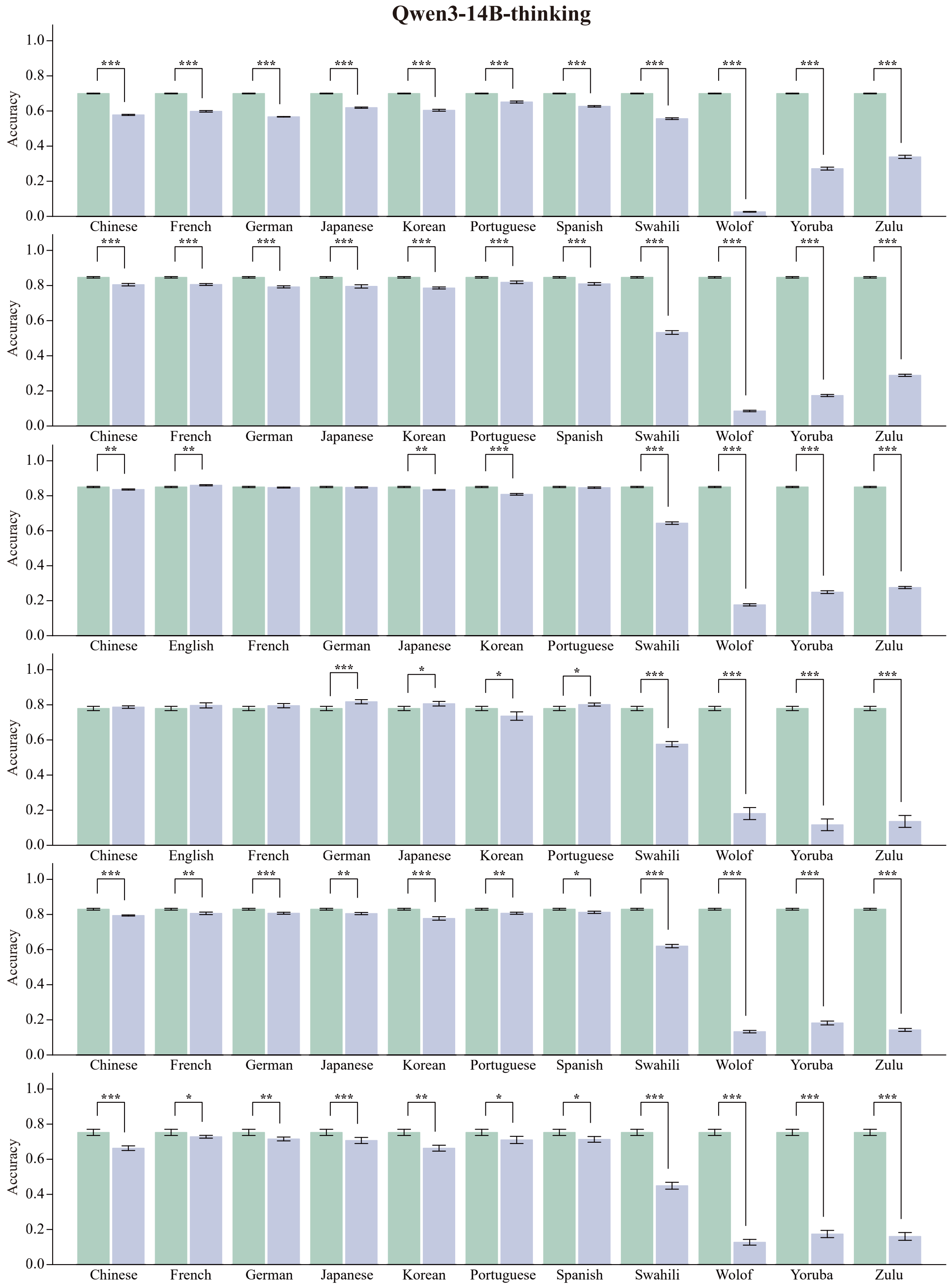

**Supplementary Fig. 45: Multilingual performance of Qwen3-14B-thinking on six medical benchmarks (top to bottom: BioNLI, MedNLI, HeadQA, MedExpQA, MedQA, MMLU-Pro).** The experiment compared the accuracy disparities between the original language (green) and target languages (purple), with each condition repeated five times. Statistical significance is indicated by asterisks (*p<0.05, **p<0.01, ***p<0.001).

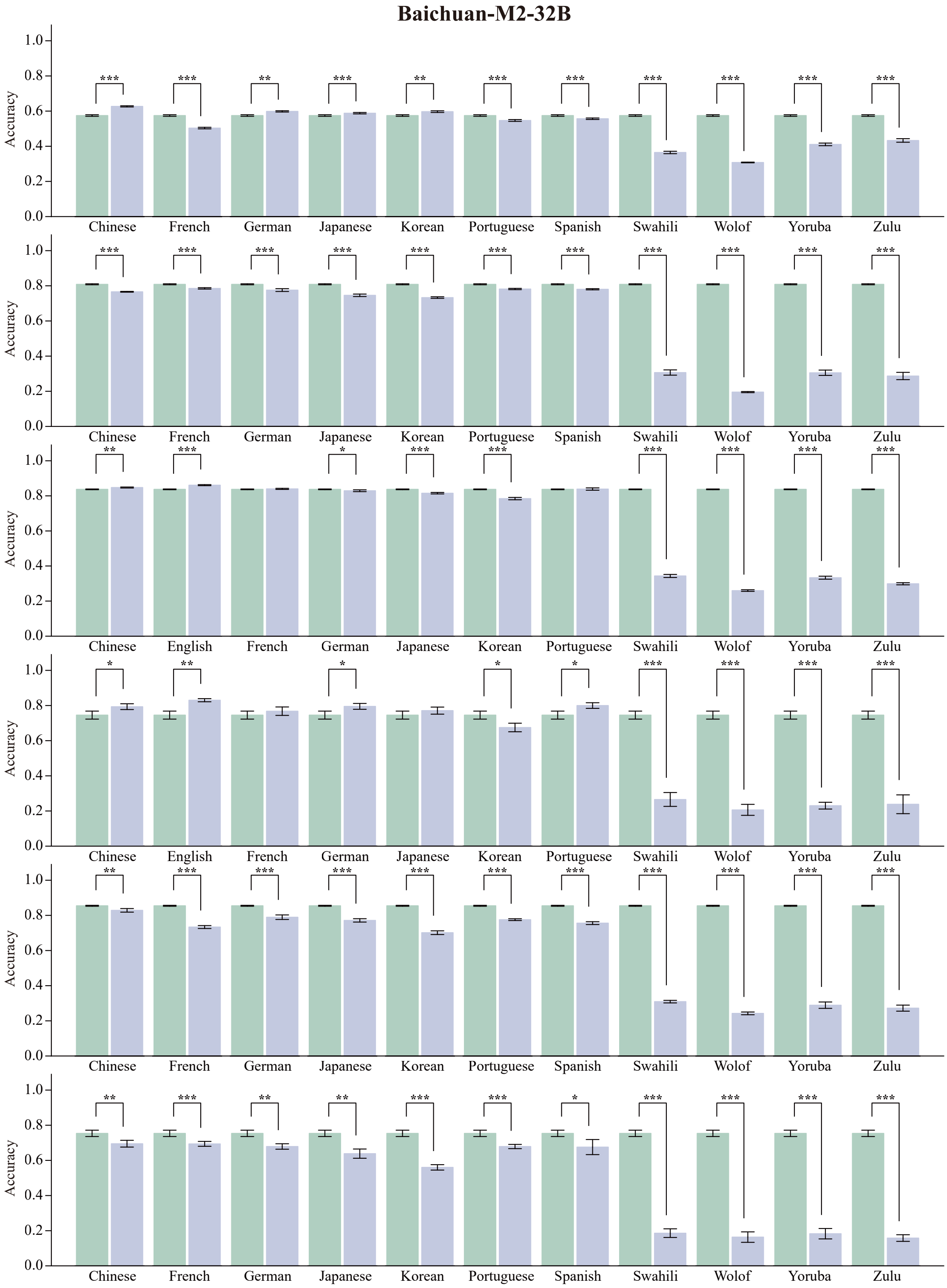

**Supplementary Fig. 46: Multilingual performance of Baichuan-M2-32B on six medical benchmarks (top to bottom: BioNLI, MedNLI, HeadQA, MedExpQA, MedQA, MMLU-Pro).** The experiment compared the accuracy disparities between the original language (green) and target languages (purple), with each condition repeated five times. Statistical significance is indicated by asterisks (*p<0.05, **p<0.01, ***p<0.001).

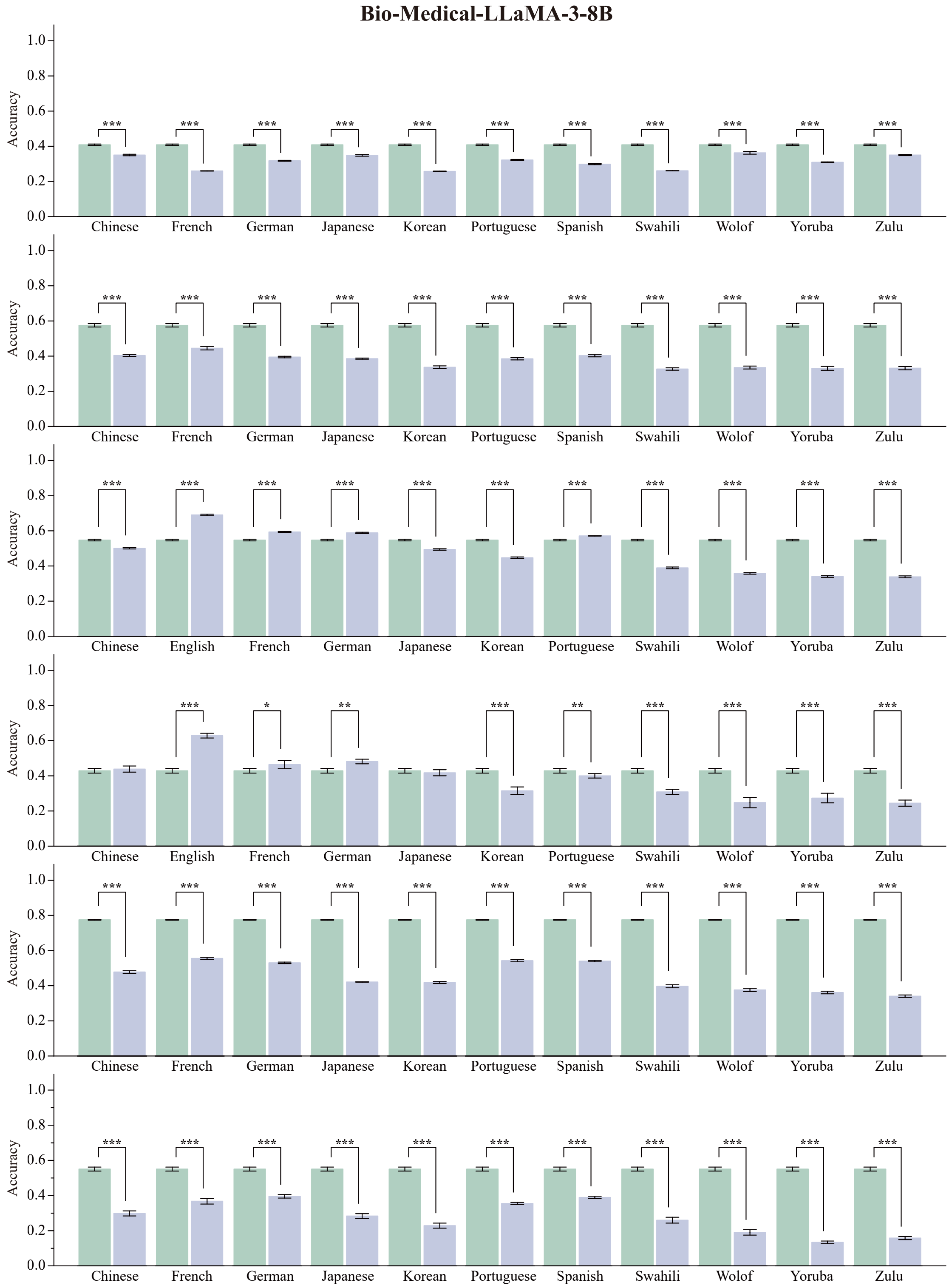

**Supplementary Fig. 47: Multilingual performance of Bio-Medical-LLaMA-3-8B on six medical benchmarks (top to bottom: BioNLI, MedNLI, HeadQA, MedExpQA, MedQA, MMLU-Pro).** The experiment compared the accuracy disparities between the original language (green) and target languages (purple), with each condition repeated five times. Statistical significance is indicated by asterisks (*p<0.05, **p<0.01, ***p<0.001).

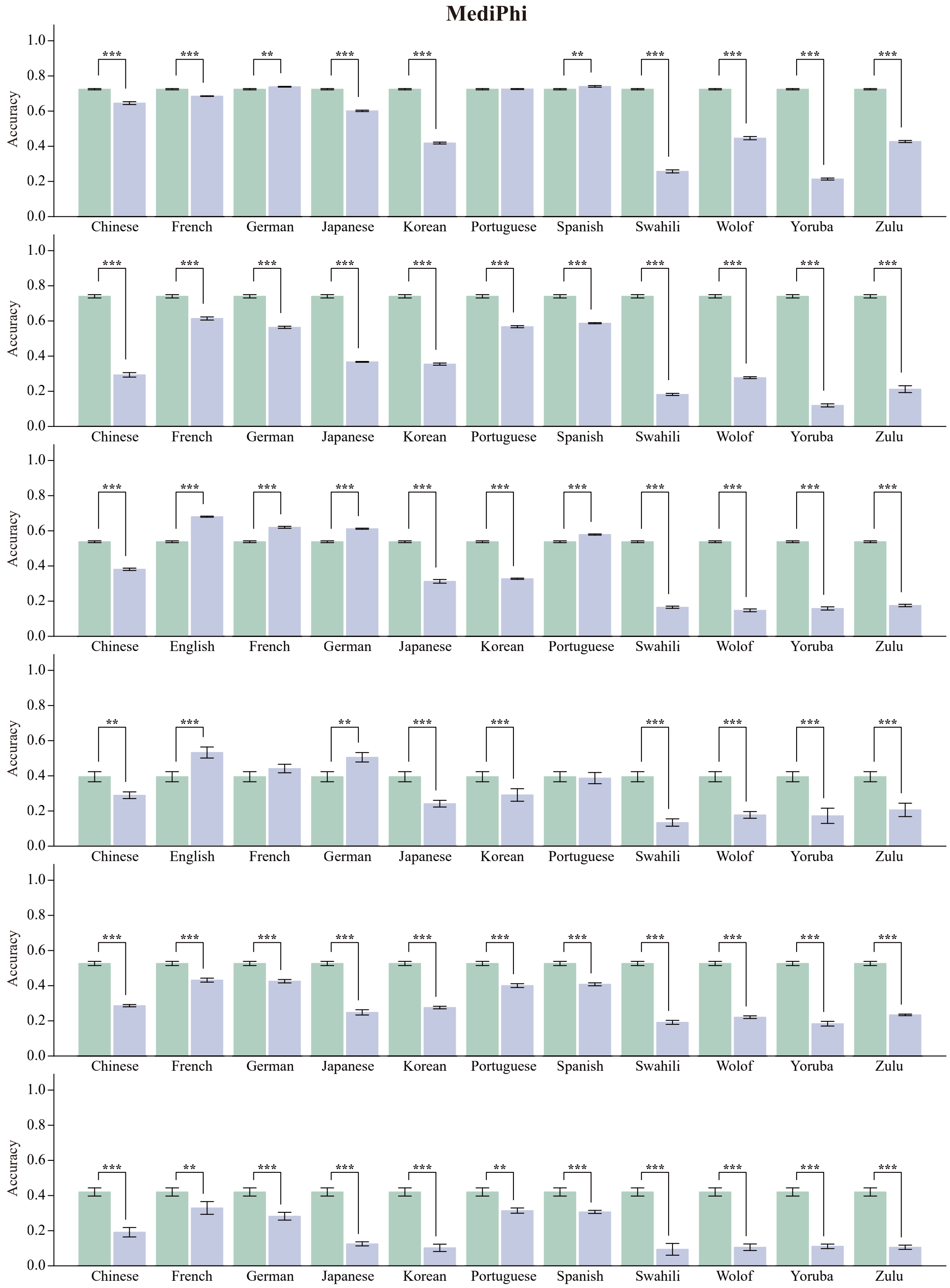

**Supplementary Fig. 48: Multilingual performance of MediPhi on six medical benchmarks (top to bottom: BioNLI, MedNLI, HeadQA, MedExpQA, MedQA, MMLU-Pro).** The experiment compared the accuracy disparities between the original language (green) and target languages (purple), with each condition repeated five times. Statistical significance is indicated by asterisks (*p<0.05, **p<0.01, ***p<0.001).

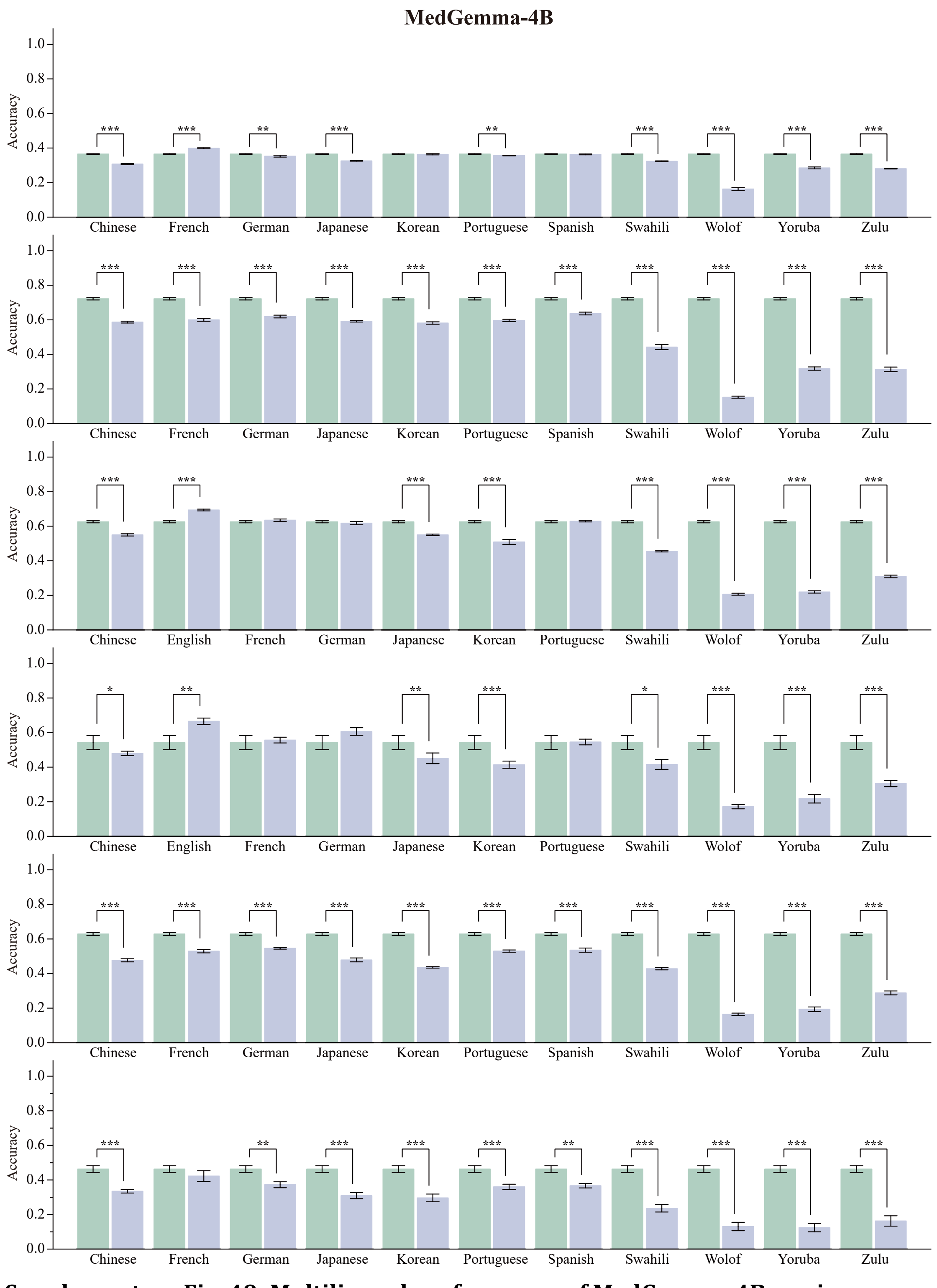

**Supplementary Fig. 49: Multilingual performance of MedGemma-4B on six medical benchmarks (top to bottom: BioNLI, MedNLI, HeadQA, MedExpQA, MedQA, MMLU-Pro).** The experiment compared the accuracy disparities between the original language (green) and target languages (purple), with each condition repeated five times. Statistical significance is indicated by asterisks (*p<0.05, **p<0.01, ***p<0.001).

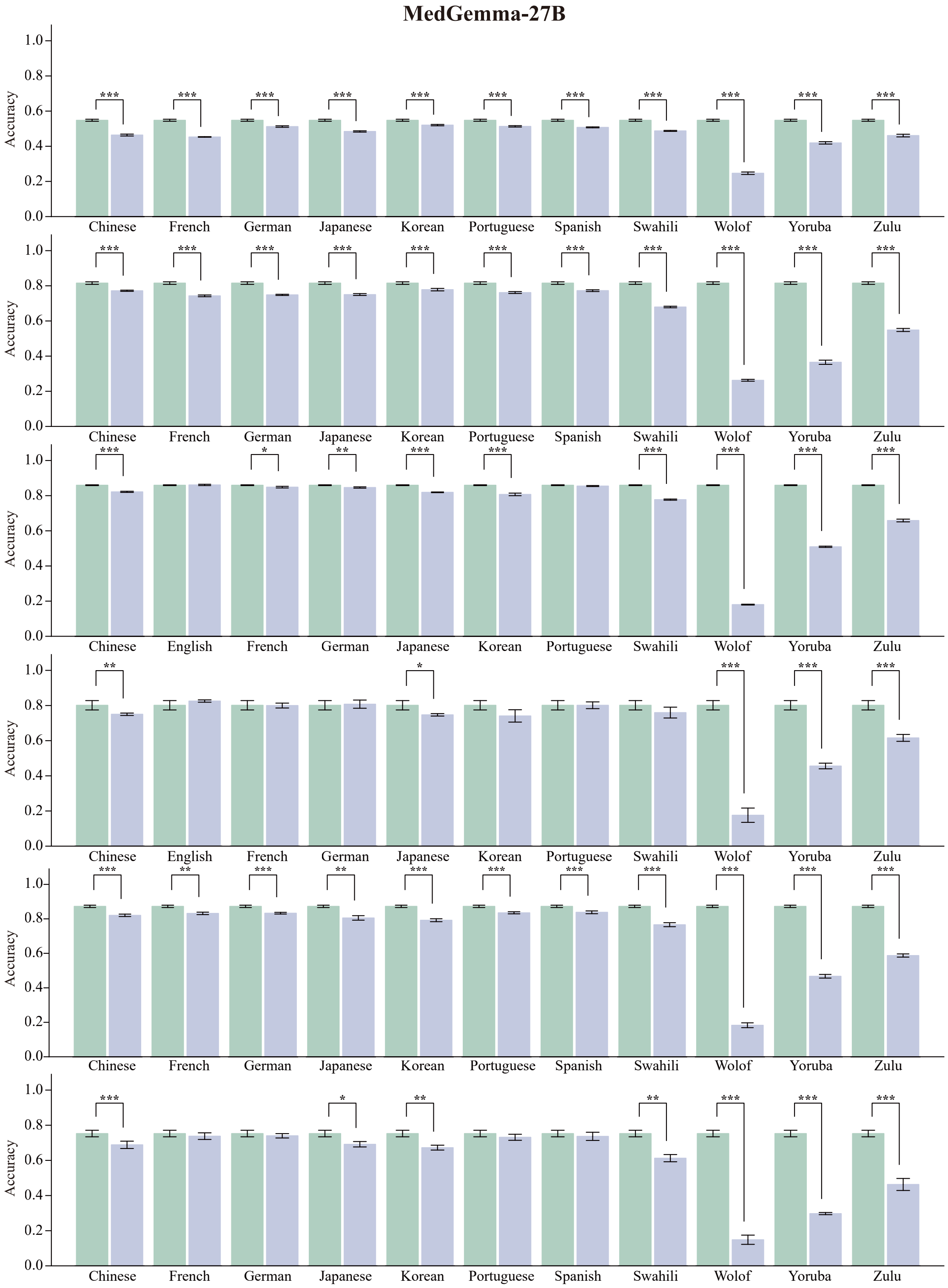

**Supplementary Fig. 50: Multilingual performance of MedGemma-27B on six medical benchmarks (top to bottom: BioNLI, MedNLI, HeadQA, MedExpQA, MedQA, MMLU-Pro).** The experiment compared the accuracy disparities between the original language (green) and target languages (purple), with each condition repeated five times. Statistical significance is indicated by asterisks (*p<0.05, **p<0.01, ***p<0.001).

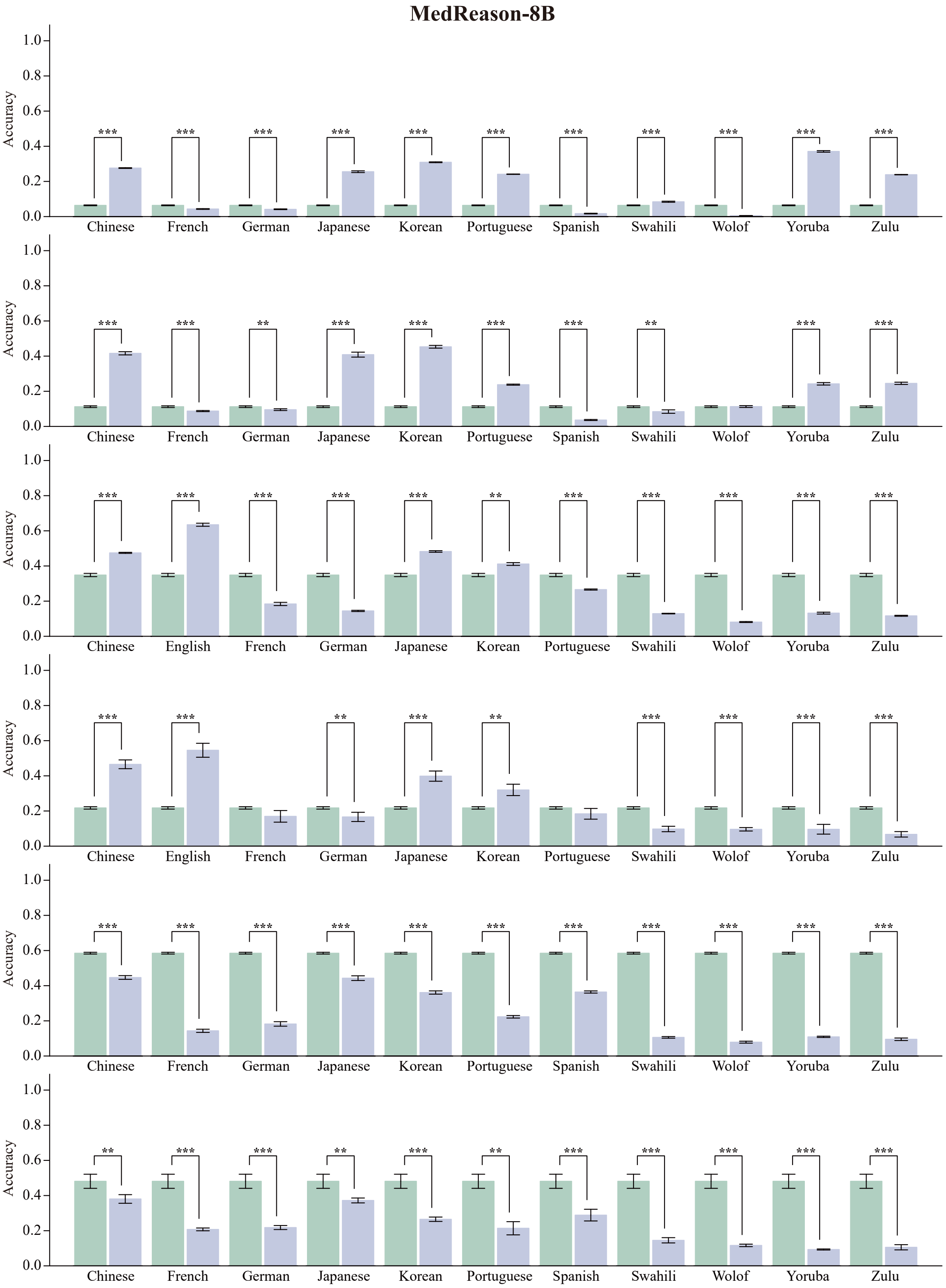

**Supplementary Fig. 51: Multilingual performance of MedReason-8B on six medical benchmarks (top to bottom: BioNLI, MedNLI, HeadQA, MedExpQA, MedQA, MMLU-Pro).** The experiment compared the accuracy disparities between the original language (green) and target languages (purple), with each condition repeated five times. Statistical significance is indicated by asterisks (*p<0.05, **p<0.01, ***p<0.001).

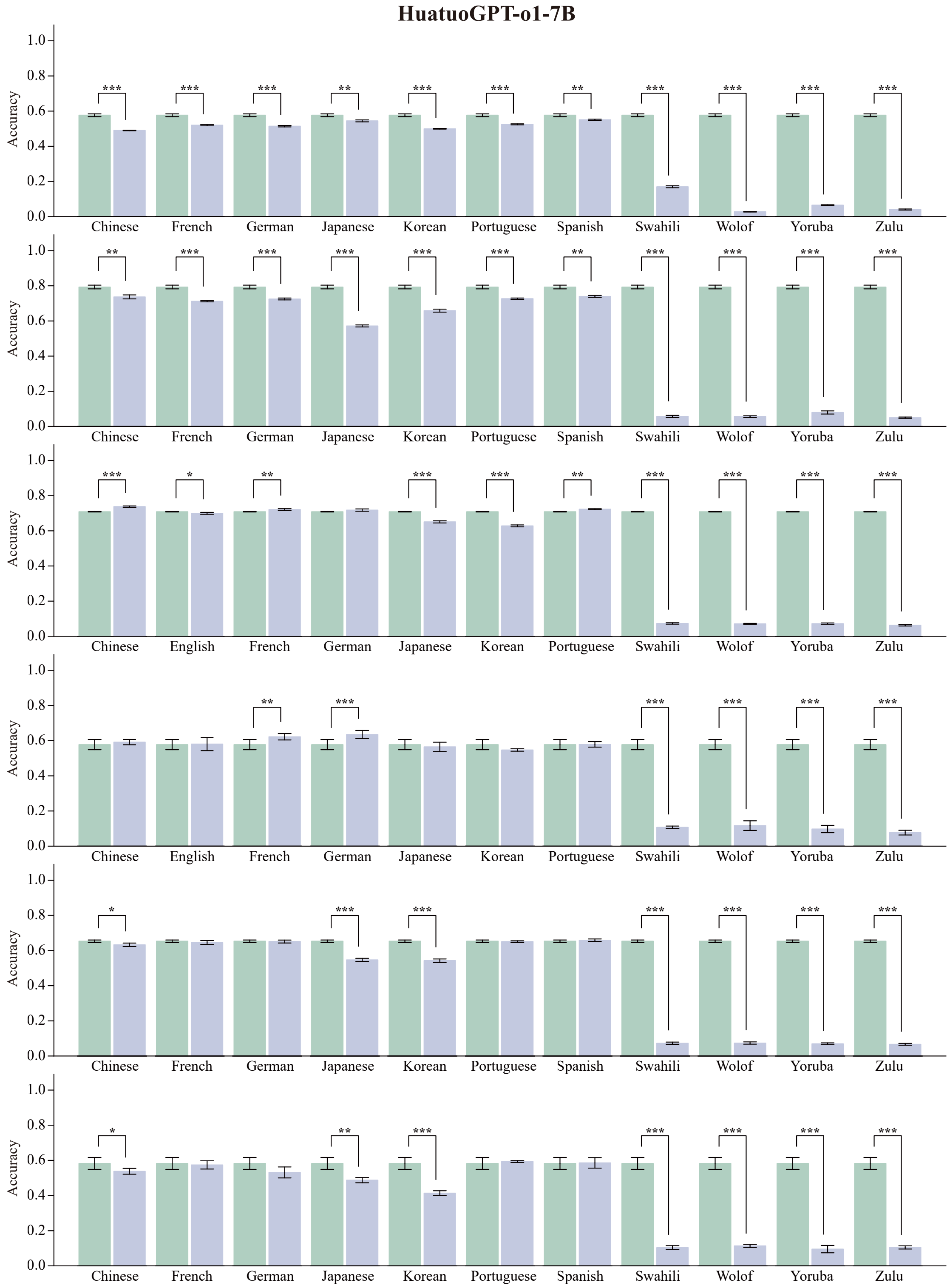

**Supplementary Fig. 52: Multilingual performance of HuatuoGPT-o1-7B on six medical benchmarks (top to bottom: BioNLI, MedNLI, HeadQA, MedExpQA, MedQA, MMLU-Pro).** The experiment compared the accuracy disparities between the original language (green) and target languages (purple), with each condition repeated five times. Statistical significance is indicated by asterisks (*p<0.05, **p<0.01, ***p<0.001).

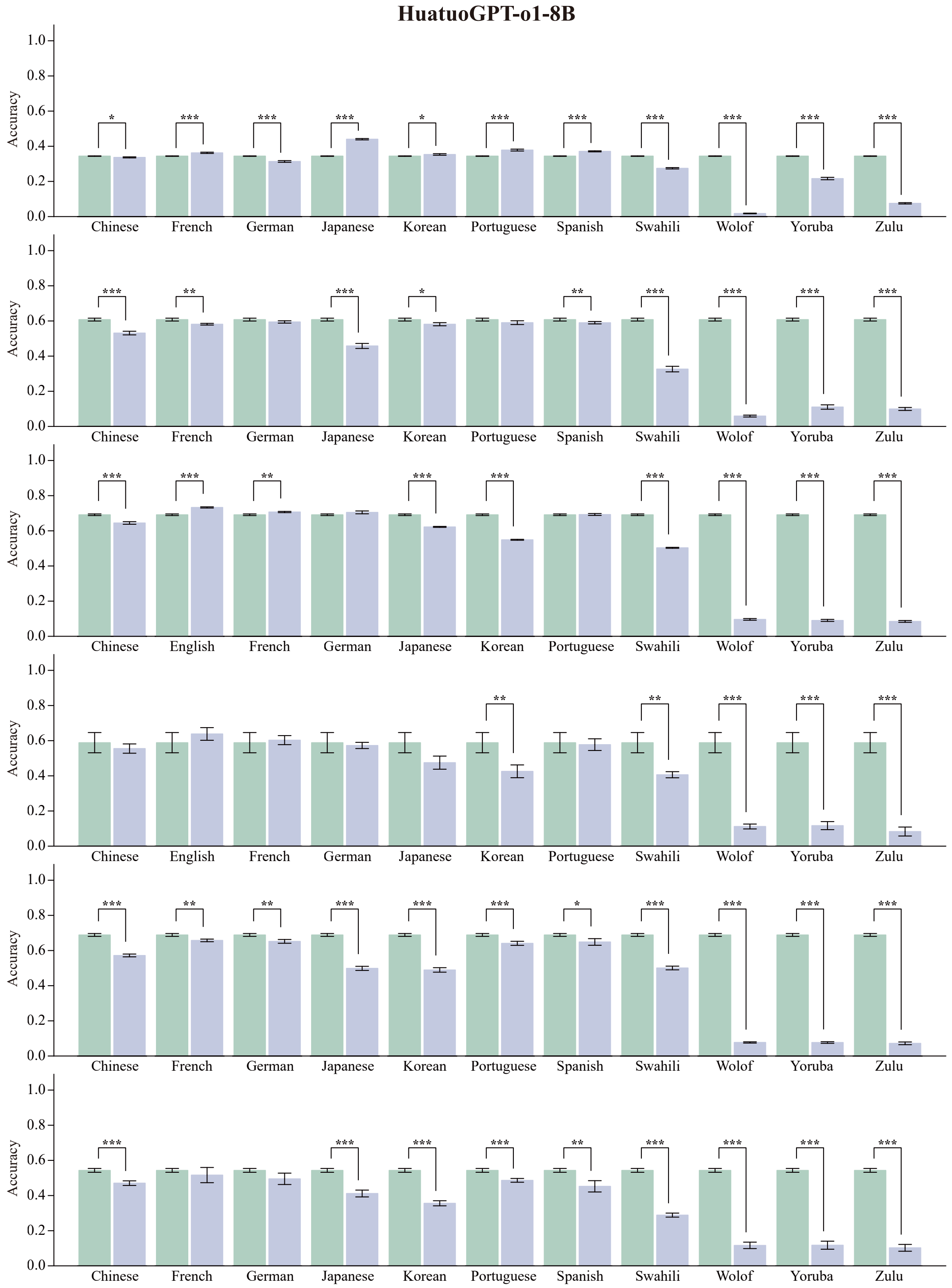

**Supplementary Fig. 53: Multilingual performance of HuatuoGPT-o1-8B on six medical benchmarks (top to bottom: BioNLI, MedNLI, HeadQA, MedExpQA, MedQA, MMLU-Pro).** The experiment compared the accuracy disparities between the original language (green) and target languages (purple), with each condition repeated five times. Statistical significance is indicated by asterisks (*p<0.05, **p<0.01, ***p<0.001).

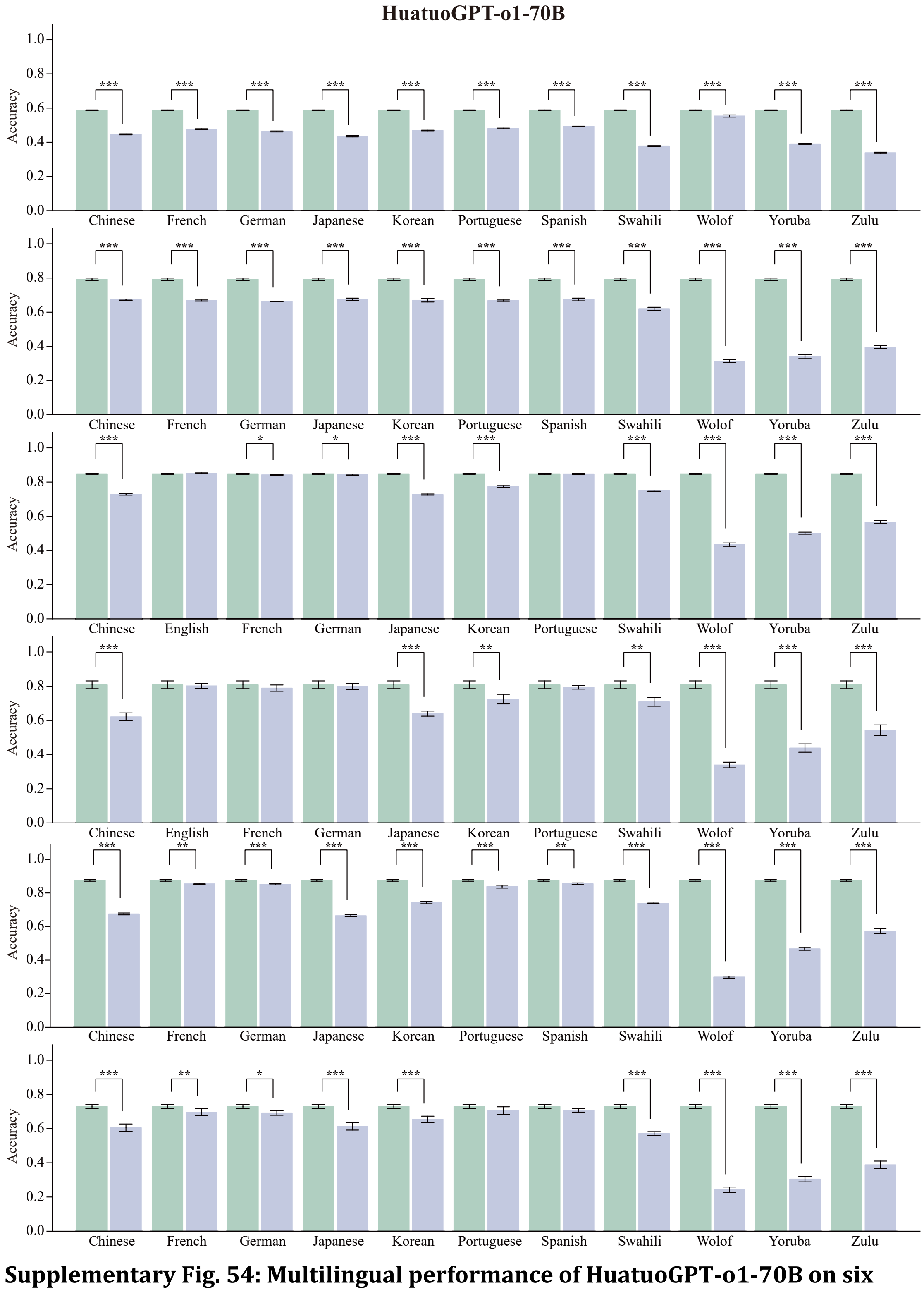

**Supplementary Fig. 54: Multilingual performance of HuatuoGPT-o1-70B on six medical benchmarks (top to bottom: BioNLI, MedNLI, HeadQA, MedExpQA, MedQA, MMLU-Pro).** The experiment compared the accuracy disparities between the original language (green) and target languages (purple), with each condition repeated five times. Statistical significance is indicated by asterisks (*p<0.05, **p<0.01, ***p<0.001).

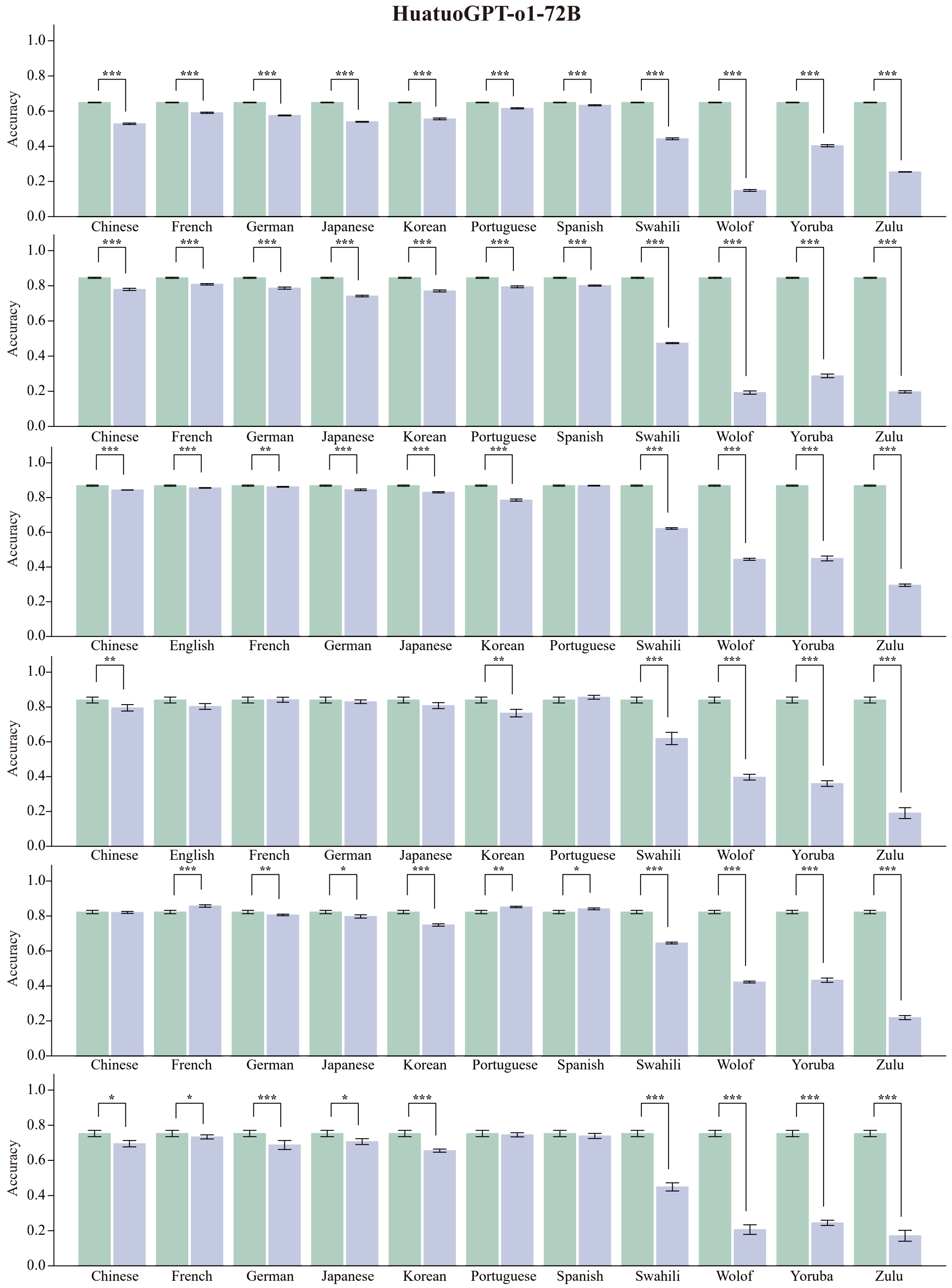

**Supplementary Fig. 55: Multilingual performance of HuatuoGPT-o1-72B on six medical benchmarks (top to bottom: BioNLI, MedNLI, HeadQA, MedExpQA, MedQA, MMLU-Pro).** The experiment compared the accuracy disparities between the original language (green) and target languages (purple), with each condition repeated five times. Statistical significance is indicated by asterisks (*p<0.05, **p<0.01, ***p<0.001).

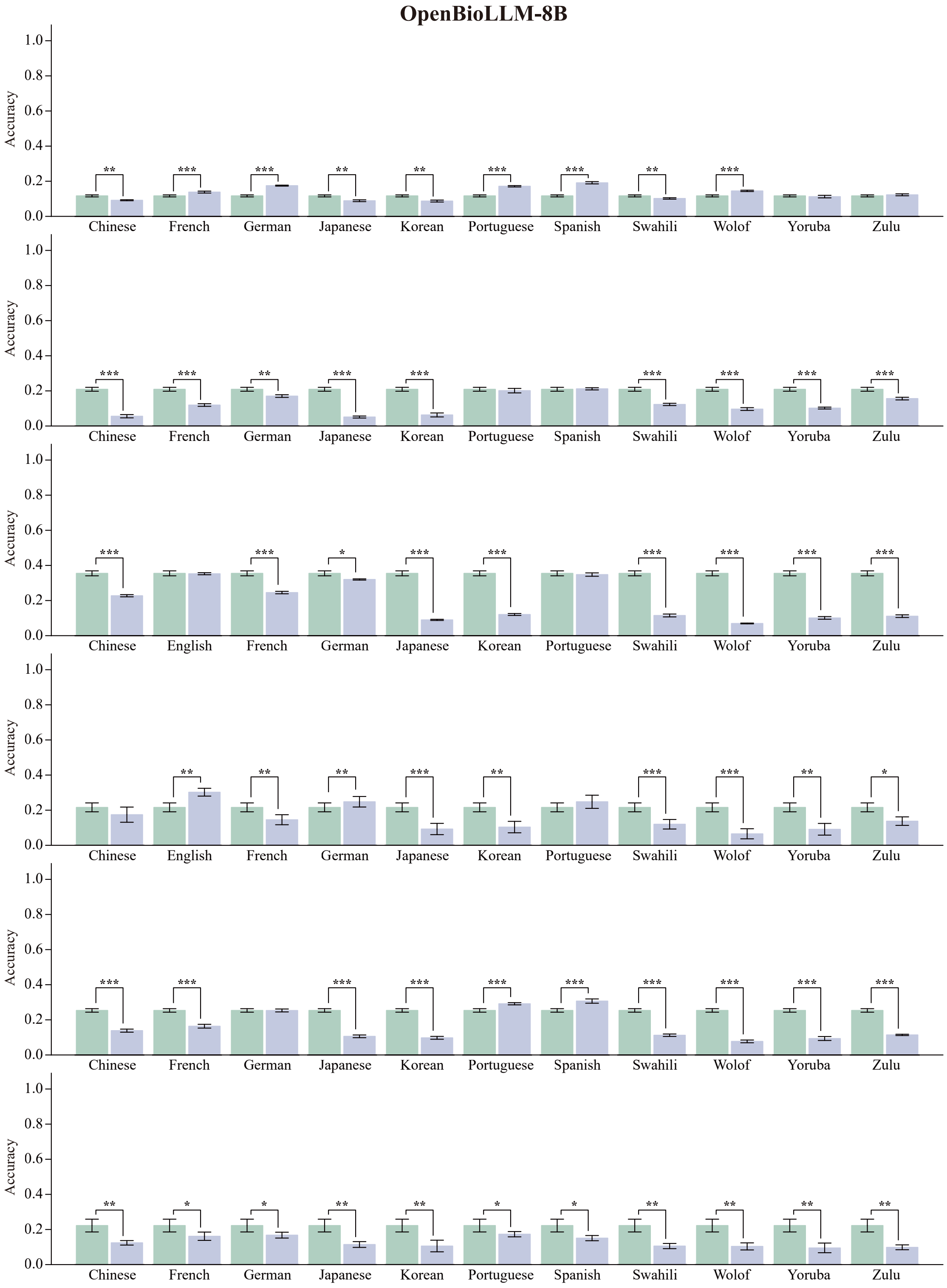

**Supplementary Fig. 56: Multilingual performance of OpenBioLLM-8B on six medical benchmarks (top to bottom: BioNLI, MedNLI, HeadQA, MedExpQA, MedQA, MMLU-Pro).** The experiment compared the accuracy disparities between the original language (green) and target languages (purple), with each condition repeated five times. Statistical significance is indicated by asterisks (*p<0.05, **p<0.01, ***p<0.001).

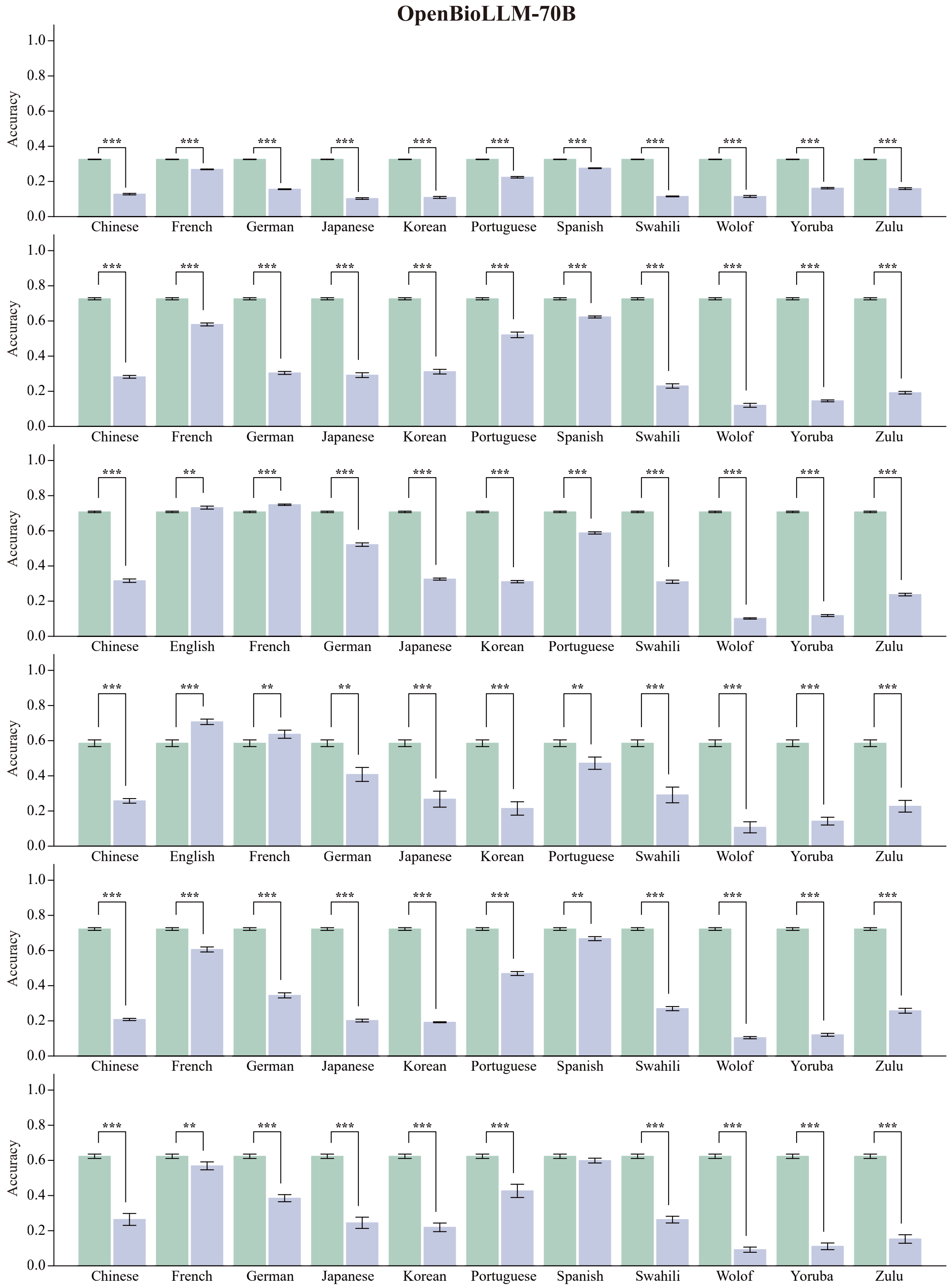

**Supplementary Fig. 57: Multilingual performance of OpenBioLLM-70B on six medical benchmarks (top to bottom: BioNLI, MedNLI, HeadQA, MedExpQA, MedQA, MMLU-Pro).** The experiment compared the accuracy disparities between the original language (green) and target languages (purple), with each condition repeated five times. Statistical significance is indicated by asterisks (*p<0.05, **p<0.01, ***p<0.001).

## S2.5. HealthBench Evaluation

We selected 30 samples from HealthBench and extended them to six languages: English, French, Japanese, Korean, Swahili, and Yoruba. We evaluated the high-performing proprietary LLM and open-weight LLM on GlobMed-Bench, Gemini-2.5-Flash and DeepSeek-R1, respectively, generating a total of 360 model responses. English responses were scored by GPT-5.5 following the original HealthBench evaluation framework. For the other five languages, each model response was independently scored under blinded conditions by two medical experts proficient in the corresponding target language, and the two expert scores were averaged. **Supplementary Tab. 127** reports the mean response-level HealthBench score of each model for each language, with Overall representing the arithmetic mean across the six languages. Both models exhibited substantial performance variation across languages, indicating that cross-lingual performance disparities persisted in more open-ended and clinically relevant scenarios.

| Language | Gemini-2.5-Flash | DeepSeek-R1 |
|---|---|---|
| **English** | 0.4355 | 0.5110 |
| **French** | 0.5241 | 0.6592 |
| **Korean** | 0.6940 | 0.7160 |
| **Japanese** | 0.6172 | 0.7190 |
| **Swahili** | 0.6805 | 0.6598 |
| **Yoruba** | 0.5399 | 0.6076 |
| **Overall** | 0.5819 | 0.6454 |

**Supplementary Tab. 127:** HealthBench evaluation across six languages.

# S3. GlobMed-LLMs

## S3.1. GlobMed-Qwen3-1.7B/8B through Direct Supervised Fine-Tuning

In addition to GlobMed-MedGemma-4B and GlobMed-Qwen3-4B discussed in the main text, both GlobMed-Qwen3-1.7B and GlobMed-Qwen3-8B demonstrated significant improvements on multilingual medical benchmarks as well (**Supplementary Fig. 58**). GlobMed-Qwen3-1.7B improved overall accuracy from 39.14% to 54.44%, while GlobMed-Qwen3-8B increased from 47.85% to 66.15%, surpassing their baseline models. Improvements in average accuracy were consistent across all benchmarks and languages, with particularly strong gains in low-resource languages such as Swahili, Wolof, Yoruba, and Zulu.

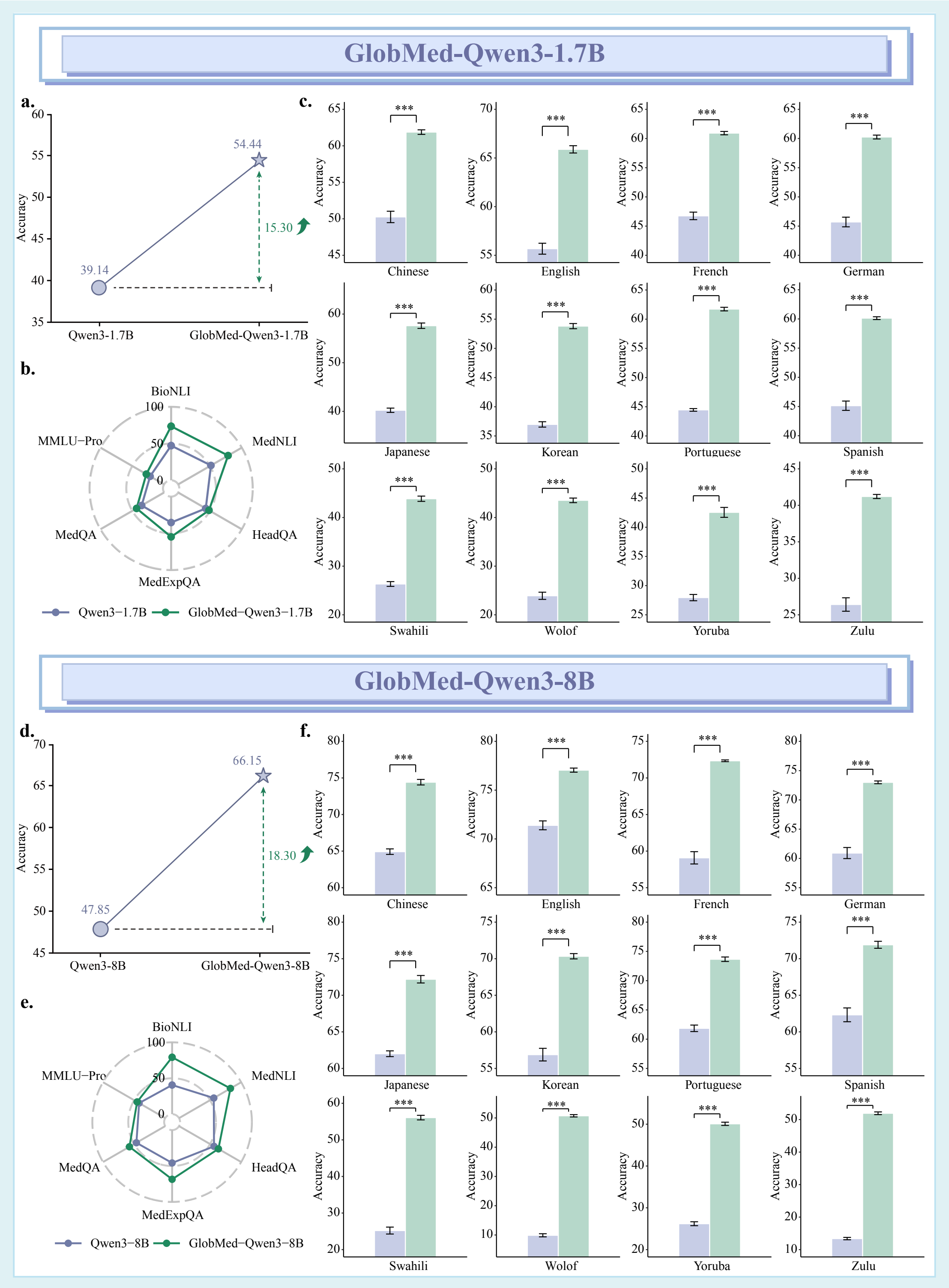

**Supplementary Fig. 58: Performance comparison of GlobMed-LLMs versus baseline LLMs. a, GlobMed-Qwen3-1.7B overall performance:** Average accuracy across all benchmarks and languages improved from 39.14% to 54.44% compared with Qwen3-1.7B. **b, Task-wise performance:** GlobMed-Qwen3-1.7B outperformed Qwen3-1.7B across all medical benchmarks. **c, Language-wise performance:** GlobMed-Qwen3-1.7B achieved higher average accuracy across all 12 languages compared with Qwen3-1.7B, with particularly notable improvements in low-resource languages. **d, GlobMed-**

**Qwen3-8B overall performance:** Average accuracy across all benchmarks and languages improved from 47.85% to 66.15% compared with Qwen3-8B. **e, Task-wise performance:** GlobMed-Qwen3-8B consistently outperformed Qwen3-8B across all medical benchmarks. **f, Language-wise performance:** GlobMed-Qwen3-8B achieved higher average accuracy across all 12 languages compared with Qwen3-8B, with particularly notable improvements in low-resource languages. Statistical significance is indicated by asterisks (*p<0.05, **p<0.01, ***p<0.001).

## S3.2. Performance Comparison between GlobMed-LLMs and Baseline LLMs

| LLMs | Chinese | English | French | German | Japanese | Korean | Portuguese | Spanish | Swahili | Wolof | Yoruba | Zulu | Overall |
|---|---|---|---|---|---|---|---|---|---|---|---|---|---|
| **MedGemma-4B** | $30.72_{\pm0.30}$ | $36.53_{\pm0.19}$ | $39.85_{\pm0.30}$ | $35.28_{\pm0.54}$ | $32.59_{\pm0.22}$ | $36.40_{\pm0.33}$ | $35.67_{\pm0.20}$ | $36.33_{\pm0.25}$ | $32.33_{\pm0.23}$ | $16.32_{\pm0.80}$ | $28.52_{\pm0.55}$ | $28.07_{\pm0.19}$ | $32.38_{\pm 0.11}$ |
| **GlobMed-MedGemma-4B** | $75.22_{\pm0.47}$ | $83.99_{\pm0.27}$ | $72.43_{\pm0.28}$ | $73.10_{\pm0.51}$ | $75.43_{\pm0.19}$ | $74.61_{\pm0.30}$ | $77.24_{\pm0.39}$ | $77.39_{\pm0.53}$ | $74.45_{\pm0.44}$ | $67.19_{\pm0.56}$ | $69.34_{\pm0.32}$ | $73.48_{\pm0.42}$ | $74.49_{\pm 0.07}$ |
| **Qwen3-1.7B** | $51.29_{\pm0.36}$ | $54.77_{\pm0.46}$ | $45.41_{\pm0.53}$ | $55.56_{\pm0.44}$ | $49.77_{\pm0.53}$ | $41.56_{\pm0.44}$ | $45.10_{\pm0.52}$ | $47.07_{\pm0.40}$ | $49.64_{\pm0.65}$ | $37.22_{\pm0.44}$ | $47.39_{\pm0.45}$ | $42.44_{\pm1.41}$ | $47.27_{\pm 0.18}$ |
| **GlobMed-Qwen3-1.7B** | $77.11_{\pm0.36}$ | $85.50_{\pm0.14}$ | $72.13_{\pm0.38}$ | $73.18_{\pm0.29}$ | $75.91_{\pm0.40}$ | $75.65_{\pm0.26}$ | $78.80_{\pm0.38}$ | $78.44_{\pm0.41}$ | $67.41_{\pm0.50}$ | $67.51_{\pm0.66}$ | $65.68_{\pm0.59}$ | $66.15_{\pm0.23}$ | $73.62_{\pm 0.09}$ |
| **Qwen3-4B** | $52.17_{\pm0.26}$ | $59.04_{\pm0.38}$ | $40.76_{\pm0.50}$ | $38.76_{\pm0.44}$ | $49.69_{\pm0.43}$ | $43.75_{\pm0.51}$ | $50.20_{\pm0.32}$ | $49.51_{\pm0.30}$ | $11.72_{\pm0.31}$ | $3.21_{\pm0.19}$ | $24.30_{\pm0.88}$ | $14.56_{\pm0.41}$ | $36.47_{\pm 0.21}$ |
| **GlobMed-Qwen3-4B** | $80.92_{\pm0.10}$ | $88.62_{\pm0.12}$ | $74.31_{\pm0.19}$ | $77.26_{\pm0.22}$ | $80.08_{\pm0.29}$ | $79.07_{\pm0.36}$ | $81.76_{\pm0.37}$ | $81.80_{\pm0.17}$ | $74.26_{\pm0.15}$ | $69.63_{\pm0.26}$ | $70.84_{\pm0.79}$ | $70.27_{\pm0.50}$ | $77.40_{\pm 0.06}$ |
| **Qwen3-8B** | $47.02_{\pm0.20}$ | $61.20_{\pm0.37}$ | $41.10_{\pm0.30}$ | $38.88_{\pm0.12}$ | $51.16_{\pm0.53}$ | $45.59_{\pm0.50}$ | $45.29_{\pm0.50}$ | $48.19_{\pm0.39}$ | $35.02_{\pm0.53}$ | $3.79_{\pm0.29}$ | $52.74_{\pm0.54}$ | $17.08_{\pm0.11}$ | $40.59_{\pm 0.06}$ |
| **GlobMed-Qwen3-8B** | $82.29_{\pm0.20}$ | $90.44_{\pm0.10}$ | $76.21_{\pm0.25}$ | $78.70_{\pm0.17}$ | $81.44_{\pm0.22}$ | $81.93_{\pm0.53}$ | $84.23_{\pm0.26}$ | $83.71_{\pm0.15}$ | $77.46_{\pm0.46}$ | $70.32_{\pm0.42}$ | $72.43_{\pm0.47}$ | $72.35_{\pm0.50}$ | $79.29_{\pm 0.16}$ |

**Supplementary Tab. 128: Performance comparison across 12 languages on BioNLI.**

| LLMs | Chinese | English | French | German | Japanese | Korean | Portuguese | Spanish | Swahili | Wolof | Yoruba | Zulu |
|---|---|---|---|---|---|---|---|---|---|---|---|---|
| **MedGemma-4B** | 30.79 | 36.36 | 39.55 | 34.38 | 32.88 | 36.31 | 35.89 | 36.63 | 32.36 | 14.92 | 28.11 | 28.20 |
| **GlobMed-MedGemma-4B** | 75.91 | 83.64 | 72.40 | 73.12 | 75.69 | 75.01 | 76.56 | 76.72 | 74.65 | 67.21 | 68.94 | 73.75 |
| **Qwen3-1.7B** | 51.24 | 54.67 | 46.07 | 55.30 | 50.02 | 41.46 | 45.24 | 46.47 | 50.40 | 37.51 | 47.10 | 42.74 |
| **GlobMed-Qwen3-1.7B** | 76.85 | 85.37 | 72.00 | 73.66 | 76.16 | 75.84 | 78.61 | 78.34 | 67.28 | 67.53 | 65.10 | 66.07 |
| **Qwen3-4B** | 51.91 | 58.92 | 40.47 | 38.20 | 49.55 | 43.06 | 50.29 | 49.24 | 11.35 | 3.17 | 23.15 | 15.19 |
| **GlobMed-Qwen3-4B** | 80.81 | 88.83 | 74.40 | 77.24 | 79.91 | 78.74 | 82.34 | 81.51 | 74.22 | 69.51 | 71.30 | 70.04 |
| **Qwen3-8B** | 47.17 | 61.26 | 41.21 | 39.01 | 51.12 | 45.78 | 45.73 | 48.74 | 34.83 | 3.55 | 52.52 | 17.12 |
| **GlobMed-Qwen3-8B** | 82.29 | 90.38 | 76.20 | 78.52 | 81.30 | 82.58 | 84.54 | 83.89 | 77.78 | 70.18 | 72.94 | 72.70 |

**Supplementary Tab. 129: Zero-Shot performance comparison across 12 languages on BioNLI (Run 1).**

| LLMs | Chinese | English | French | German | Japanese | Korean | Portuguese | Spanish | Swahili | Wolof | Yoruba | Zulu |
|---|---|---|---|---|---|---|---|---|---|---|---|---|
| **MedGemma-4B** | 30.65 | 36.31 | 39.84 | 35.78 | 32.40 | 36.90 | 35.75 | 36.18 | 32.20 | 16.70 | 28.11 | 27.75 |
| **GlobMed-MedGemma-4B** | 75.15 | 84.09 | 72.31 | 72.85 | 75.55 | 74.49 | 77.55 | 78.11 | 74.56 | 66.76 | 69.37 | 74.02 |
| **Qwen3-1.7B** | 51.51 | 54.52 | 44.76 | 55.01 | 50.22 | 42.00 | 44.25 | 47.30 | 49.84 | 36.81 | 48.13 | 40.25 |
| **GlobMed-Qwen3-1.7B** | 77.64 | 85.55 | 72.54 | 73.10 | 75.75 | 75.75 | 78.92 | 77.98 | 67.55 | 67.26 | 66.00 | 66.36 |
| **Qwen3-4B** | 52.13 | 59.17 | 40.18 | 39.08 | 49.44 | 44.09 | 50.20 | 49.15 | 11.60 | 3.35 | 23.64 | 14.72 |
| **GlobMed-Qwen3-4B** | 80.90 | 88.54 | 74.58 | 77.24 | 79.98 | 78.81 | 81.89 | 81.82 | 74.52 | 69.75 | 69.80 | 70.34 |
| **Qwen3-8B** | 46.74 | 60.65 | 41.55 | 38.99 | 51.66 | 44.97 | 44.94 | 48.22 | 35.21 | 3.91 | 53.21 | 17.03 |
| **GlobMed-Qwen3-8B** | 82.34 | 90.40 | 76.25 | 78.92 | 81.51 | 81.10 | 83.82 | 83.51 | 76.74 | 70.02 | 71.89 | 72.00 |

**Supplementary Tab. 130: Zero-Shot performance comparison across 12 languages on BioNLI (Run 2).**

| LLMs | Chinese | English | French | German | Japanese | Korean | Portuguese | Spanish | Swahili | Wolof | Yoruba | Zulu |
|---|---|---|---|---|---|---|---|---|---|---|---|---|
| **MedGemma-4B** | 31.03 | 36.76 | 39.62 | 35.39 | 32.56 | 36.36 | 35.78 | 36.45 | 32.20 | 16.81 | 28.52 | 28.22 |
| **GlobMed-MedGemma-4B** | 75.44 | 84.34 | 72.36 | 73.91 | 75.39 | 74.18 | 77.33 | 77.71 | 73.69 | 66.81 | 69.10 | 73.35 |
| **Qwen3-1.7B** | 51.78 | 54.18 | 44.99 | 56.02 | 49.71 | 40.94 | 45.55 | 47.42 | 49.44 | 37.35 | 47.03 | 42.45 |
| **GlobMed-Qwen3-1.7B** | 77.30 | 85.57 | 72.02 | 73.17 | 75.53 | 75.87 | 78.76 | 78.18 | 67.89 | 68.07 | 64.99 | 66.27 |
| **Qwen3-4B** | 52.45 | 59.42 | 41.51 | 39.28 | 49.19 | 44.36 | 50.67 | 49.60 | 12.16 | 3.21 | 24.90 | 14.49 |
| **GlobMed-Qwen3-4B** | 80.88 | 88.56 | 74.29 | 77.06 | 80.58 | 78.94 | 81.44 | 81.80 | 74.20 | 69.62 | 71.89 | 70.76 |
| **Qwen3-8B** | 47.01 | 61.21 | 41.03 | 38.83 | 50.31 | 46.25 | 45.82 | 47.69 | 35.71 | 4.13 | 52.38 | 17.08 |
| **GlobMed-Qwen3-8B** | 81.98 | 90.56 | 76.38 | 78.61 | 81.46 | 82.00 | 84.31 | 83.82 | 77.91 | 69.89 | 71.98 | 72.72 |

**Supplementary Tab. 131: Zero-Shot performance comparison across 12 languages on BioNLI (Run 3).**

| LLMs | Chinese | English | French | German | Japanese | Korean | Portuguese | Spanish | Swahili | Wolof | Yoruba | Zulu |
|---|---|---|---|---|---|---|---|---|---|---|---|---|
| MedGemma-4B | 30.25 | 36.56 | 39.96 | 35.26 | 32.38 | 36.43 | 35.42 | 36.40 | 32.72 | 16.38 | 29.44 | 28.16 |
| GlobMed-MedGemma-4B | 74.70 | 83.82 | 72.92 | 72.54 | 75.21 | 74.65 | 77.30 | 77.21 | 74.85 | 67.03 | 69.64 | 72.94 |
| Qwen3-1.7B | 50.90 | 55.19 | 45.69 | 55.51 | 48.88 | 41.46 | 45.44 | 46.88 | 48.65 | 37.73 | 47.17 | 42.58 |
| GlobMed-Qwen3-1.7B | 77.03 | 85.33 | 71.62 | 73.08 | 75.62 | 75.53 | 79.37 | 78.67 | 67.71 | 68.16 | 66.22 | 66.25 |
| Qwen3-4B | 52.43 | 58.43 | 40.94 | 38.79 | 50.00 | 43.51 | 49.89 | 49.75 | 11.91 | 2.92 | 24.58 | 14.13 |
| GlobMed-Qwen3-4B | 81.08 | 88.61 | 74.09 | 77.62 | 79.87 | 79.28 | 81.44 | 81.98 | 74.25 | 69.28 | 70.67 | 70.67 |
| Qwen3-8B | 46.94 | 61.21 | 40.74 | 38.83 | 51.53 | 45.73 | 44.65 | 48.34 | 34.27 | 3.42 | 53.42 | 16.94 |
| GlobMed-Qwen3-8B | 82.29 | 90.34 | 76.43 | 78.83 | 81.75 | 81.93 | 84.18 | 83.66 | 77.35 | 70.72 | 72.76 | 72.67 |

**Supplementary Tab. 132: Zero-Shot performance comparison across 12 languages on BioNLI (Run 4).**

| LLMs | Chinese | English | French | German | Japanese | Korean | Portuguese | Spanish | Swahili | Wolof | Yoruba | Zulu |
|---|---|---|---|---|---|---|---|---|---|---|---|---|
| MedGemma-4B | 30.88 | 36.65 | 40.29 | 35.60 | 32.74 | 35.98 | 35.51 | 35.98 | 32.18 | 16.81 | 28.40 | 28.04 |
| GlobMed-MedGemma-4B | 74.92 | 84.04 | 72.18 | 73.10 | 75.30 | 74.70 | 77.44 | 77.21 | 74.49 | 68.13 | 69.64 | 73.35 |
| Qwen3-1.7B | 51.03 | 55.28 | 45.55 | 55.98 | 50.02 | 41.96 | 45.03 | 47.30 | 49.87 | 36.72 | 47.51 | 44.18 |
| GlobMed-Qwen3-1.7B | 76.74 | 85.66 | 72.49 | 72.90 | 76.49 | 75.24 | 78.36 | 79.01 | 66.61 | 66.54 | 66.11 | 65.78 |
| Qwen3-4B | 51.93 | 59.24 | 40.70 | 38.47 | 50.25 | 43.71 | 49.93 | 49.80 | 11.60 | 3.42 | 25.24 | 14.29 |
| GlobMed-Qwen3-4B | 80.94 | 88.58 | 74.18 | 77.12 | 80.04 | 79.60 | 81.71 | 81.87 | 74.11 | 69.98 | 70.52 | 69.53 |
| Qwen3-8B | 47.24 | 61.69 | 40.97 | 38.72 | 51.17 | 45.21 | 45.33 | 47.98 | 35.10 | 3.93 | 52.18 | 17.24 |
| GlobMed-Qwen3-8B | 82.54 | 90.54 | 75.80 | 78.61 | 81.17 | 82.02 | 84.31 | 83.69 | 77.53 | 70.81 | 72.56 | 71.64 |

**Supplementary Tab. 133: Zero-Shot performance comparison across 12 languages on BioNLI (Run 5).**

| LLMs | Chinese | English | French | German | Japanese | Korean | Portuguese | Spanish | Swahili | Wolof | Yoruba | Zulu | Overall |
|---|---|---|---|---|---|---|---|---|---|---|---|---|---|
| MedGemma-4B | $58.70_{\pm 0.53}$ | $72.20_{\pm 0.68}$ | $60.00_{\pm 0.84}$ | $61.92_{\pm 0.79}$ | $59.16_{\pm 0.46}$ | $58.16_{\pm 0.72}$ | $59.66_{\pm 0.63}$ | $63.66_{\pm 0.78}$ | $44.26_{\pm 1.44}$ | $15.23_{\pm 0.58}$ | $31.80_{\pm 0.95}$ | $31.39_{\pm 1.30}$ | $51.34 \pm 0.21$ |
| GlobMed-MedGemma-4B | $86.89_{\pm 0.44}$ | $88.55_{\pm 0.37}$ | $87.69_{\pm 0.38}$ | $87.06_{\pm 0.84}$ | $86.46_{\pm 0.53}$ | $85.17_{\pm 0.55}$ | $86.96_{\pm 0.42}$ | $87.78_{\pm 0.50}$ | $82.96_{\pm 0.69}$ | $57.95_{\pm 0.43}$ | $64.42_{\pm 0.86}$ | $70.40_{\pm 0.65}$ | $81.02 \pm 0.14$ |
| Qwen3-1.7B | $68.44_{\pm 0.96}$ | $75.29_{\pm 0.66}$ | $66.58_{\pm 0.43}$ | $62.15_{\pm 0.82}$ | $54.20_{\pm 0.75}$ | $49.33_{\pm 0.60}$ | $65.14_{\pm 0.63}$ | $63.79_{\pm 0.91}$ | $27.38_{\pm 1.61}$ | $23.59_{\pm 0.96}$ | $30.61_{\pm 1.47}$ | $32.43_{\pm 1.20}$ | $51.58 \pm 0.32$ |
| GlobMed-Qwen3-1.7B | $87.31_{\pm 0.44}$ | $88.60_{\pm 0.45}$ | $86.80_{\pm 0.27}$ | $86.99_{\pm 0.50}$ | $86.08_{\pm 0.37}$ | $84.42_{\pm 0.66}$ | $86.38_{\pm 0.40}$ | $86.48_{\pm 0.34}$ | $69.98_{\pm 0.98}$ | $60.58_{\pm 0.48}$ | $61.67_{\pm 0.62}$ | $56.95_{\pm 0.56}$ | $78.52 \pm 0.19$ |
| Qwen3-4B | $76.78_{\pm 0.51}$ | $82.33_{\pm 0.69}$ | $68.57_{\pm 0.78}$ | $57.73_{\pm 1.32}$ | $72.90_{\pm 1.18}$ | $71.87_{\pm 0.50}$ | $75.40_{\pm 0.68}$ | $73.37_{\pm 0.71}$ | $17.78_{\pm 1.22}$ | $18.53_{\pm 1.25}$ | $11.01_{\pm 0.71}$ | $14.99_{\pm 0.92}$ | $53.44 \pm 0.11$ |
| GlobMed-Qwen3-4B | $89.13_{\pm 0.27}$ | $89.99_{\pm 0.23}$ | $88.86_{\pm 0.28}$ | $89.14_{\pm 0.21}$ | $88.68_{\pm 0.33}$ | $86.83_{\pm 0.17}$ | $88.57_{\pm 0.41}$ | $89.17_{\pm 0.47}$ | $76.90_{\pm 0.65}$ | $64.22_{\pm 1.29}$ | $65.04_{\pm 1.12}$ | $64.62_{\pm 1.07}$ | $81.76 \pm 0.18$ |
| Qwen3-8B | $75.98_{\pm 1.16}$ | $77.49_{\pm 0.50}$ | $72.01_{\pm 0.49}$ | $69.49_{\pm 0.81}$ | $77.15_{\pm 0.63}$ | $72.55_{\pm 0.31}$ | $70.97_{\pm 0.36}$ | $70.90_{\pm 1.01}$ | $27.30_{\pm 0.78}$ | $13.06_{\pm 1.01}$ | $24.80_{\pm 0.88}$ | $20.24_{\pm 0.48}$ | $55.99 \pm 0.18$ |
| GlobMed-Qwen3-8B | $89.22_{\pm 0.37}$ | $90.32_{\pm 0.17}$ | $89.68_{\pm 0.17}$ | $89.64_{\pm 0.33}$ | $88.70_{\pm 0.19}$ | $87.88_{\pm 0.37}$ | $88.89_{\pm 0.32}$ | $88.74_{\pm 0.42}$ | $78.93_{\pm 0.37}$ | $63.73_{\pm 0.65}$ | $69.74_{\pm 0.65}$ | $68.33_{\pm 0.67}$ | $82.81 \pm 0.11$ |

**Supplementary Tab. 134: Performance comparison across 12 languages on MedNLI.**

| LLMs | Chinese | English | French | German | Japanese | Korean | Portuguese | Spanish | Swahili | Wolof | Yoruba | Zulu |
|---|---|---|---|---|---|---|---|---|---|---|---|---|
| MedGemma-4B | 59.35 | 72.05 | 59.00 | 62.31 | 59.07 | 57.73 | 59.77 | 64.29 | 45.73 | 15.60 | 30.28 | 30.77 |
| GlobMed-MedGemma-4B | 86.45 | 88.07 | 88.14 | 86.45 | 87.01 | 84.90 | 86.31 | 88.00 | 82.85 | 57.87 | 63.87 | 69.65 |
| Qwen3-1.7B | 68.03 | 75.37 | 66.69 | 61.54 | 54.20 | 49.26 | 64.22 | 63.30 | 27.45 | 22.23 | 31.90 | 33.17 |
| GlobMed-Qwen3-1.7B | 86.66 | 89.13 | 86.59 | 86.31 | 86.03 | 85.25 | 86.17 | 86.24 | 68.88 | 60.13 | 60.90 | 56.60 |
| Qwen3-4B | 76.71 | 82.99 | 67.68 | 58.65 | 72.34 | 71.63 | 75.44 | 72.34 | 18.35 | 19.97 | 10.80 | 14.68 |
| GlobMed-Qwen3-4B | 89.48 | 90.33 | 88.92 | 88.99 | 88.85 | 86.59 | 87.93 | 89.13 | 76.08 | 63.59 | 66.48 | 66.48 |
| Qwen3-8B | 76.64 | 77.42 | 72.12 | 68.45 | 77.63 | 72.90 | 71.21 | 70.71 | 27.52 | 14.11 | 24.35 | 20.89 |
| GlobMed-Qwen3-8B | 89.56 | 90.19 | 89.84 | 89.34 | 88.43 | 87.65 | 88.57 | 88.57 | 78.69 | 63.59 | 70.85 | 67.61 |

**Supplementary Tab. 135: Zero-Shot performance comparison across 12 languages on MedNLI (Run 1).**

| LLMs | Chinese | English | French | German | Japanese | Korean | Portuguese | Spanish | Swahili | Wolof | Yoruba | Zulu |
|---|---|---|---|---|---|---|---|---|---|---|---|---|
| MedGemma-4B | 59.00 | 72.27 | 59.56 | 62.88 | 59.14 | 58.50 | 58.86 | 62.67 | 42.70 | 15.74 | 31.90 | 31.55 |
| GlobMed-MedGemma-4B | 87.01 | 88.85 | 87.37 | 87.23 | 87.01 | 84.47 | 86.87 | 88.36 | 82.99 | 57.23 | 64.64 | 70.85 |
| Qwen3-1.7B | 67.47 | 76.29 | 66.06 | 63.37 | 53.00 | 49.82 | 65.21 | 63.44 | 27.38 | 22.94 | 30.70 | 32.53 |
| GlobMed-Qwen3-1.7B | 87.16 | 88.29 | 87.23 | 87.65 | 86.66 | 84.47 | 85.96 | 86.80 | 70.64 | 60.41 | 61.82 | 56.32 |
| Qwen3-4B | 77.21 | 82.07 | 69.58 | 55.96 | 73.39 | 71.35 | 74.95 | 73.82 | 17.64 | 16.73 | 10.94 | 16.30 |
| GlobMed-Qwen3-4B | 88.99 | 89.84 | 88.92 | 88.99 | 89.13 | 86.87 | 88.57 | 88.92 | 77.21 | 64.86 | 64.36 | 64.50 |
| Qwen3-8B | 76.57 | 76.85 | 72.27 | 70.29 | 77.56 | 72.69 | 71.00 | 71.56 | 28.51 | 13.41 | 23.57 | 19.83 |
| GlobMed-Qwen3-8B | 88.71 | 90.33 | 89.77 | 89.91 | 88.85 | 88.14 | 88.71 | 89.20 | 78.55 | 63.02 | 69.65 | 67.75 |

**Supplementary Tab. 136: Zero-Shot performance comparison across 12 languages on MedNLI (Run 2).**

| LLMs | Chinese | English | French | German | Japanese | Korean | Portuguese | Spanish | Swahili | Wolof | Yoruba | Zulu |
|---|---|---|---|---|---|---|---|---|---|---|---|---|
| MedGemma-4B | 58.01 | 71.14 | 60.97 | 62.10 | 58.93 | 59.21 | 59.21 | 64.29 | 45.73 | 14.26 | 32.89 | 30.35 |
| GlobMed-MedGemma-4B | 86.80 | 88.92 | 87.37 | 88.43 | 86.24 | 85.32 | 86.94 | 87.65 | 82.50 | 58.15 | 63.37 | 71.21 |
| Qwen3-1.7B | 68.17 | 75.09 | 66.97 | 61.47 | 54.41 | 48.34 | 65.28 | 65.42 | 26.61 | 24.28 | 28.86 | 30.42 |
| GlobMed-Qwen3-1.7B | 87.86 | 88.99 | 86.59 | 87.16 | 86.17 | 83.91 | 86.87 | 86.38 | 71.14 | 60.20 | 61.61 | 57.80 |
| Qwen3-4B | 75.94 | 81.79 | 68.88 | 59.21 | 72.12 | 71.56 | 76.15 | 73.89 | 17.71 | 19.27 | 12.07 | 14.61 |
| GlobMed-Qwen3-4B | 88.99 | 90.12 | 89.27 | 89.06 | 88.36 | 86.73 | 88.64 | 89.48 | 77.21 | 62.39 | 64.64 | 64.15 |
| Qwen3-8B | 77.13 | 78.19 | 71.42 | 68.81 | 77.56 | 72.12 | 70.64 | 69.23 | 26.46 | 12.21 | 25.34 | 20.61 |
| GlobMed-Qwen3-8B | 89.27 | 90.19 | 89.41 | 89.56 | 88.64 | 87.79 | 88.92 | 89.13 | 79.25 | 63.73 | 69.16 | 69.23 |

**Supplementary Tab. 137: Zero-Shot performance comparison across 12 languages on MedNLI (Run 3).**

| LLMs | Chinese | English | French | German | Japanese | Korean | Portuguese | Spanish | Swahili | Wolof | Yoruba | Zulu |
|---|---|---|---|---|---|---|---|---|---|---|---|---|
| MedGemma-4B | 58.79 | 72.55 | 60.76 | 61.47 | 59.92 | 57.37 | 60.06 | 64.08 | 44.11 | 15.31 | 31.83 | 33.59 |
| GlobMed-MedGemma-4B | 87.58 | 88.29 | 87.51 | 86.80 | 86.24 | 85.96 | 87.37 | 87.01 | 84.12 | 58.29 | 64.57 | 70.43 |
| Qwen3-1.7B | 70.01 | 74.45 | 66.20 | 61.75 | 55.05 | 49.47 | 65.00 | 63.44 | 29.92 | 24.42 | 32.18 | 33.52 |
| GlobMed-Qwen3-1.7B | 87.37 | 88.50 | 86.87 | 87.09 | 85.89 | 83.63 | 86.73 | 86.10 | 69.09 | 61.26 | 61.40 | 56.95 |
| Qwen3-4B | 77.13 | 83.13 | 67.89 | 57.94 | 74.74 | 72.41 | 75.94 | 72.90 | 15.95 | 17.93 | 10.09 | 15.46 |
| GlobMed-Qwen3-4B | 88.85 | 89.91 | 88.57 | 89.20 | 88.36 | 87.01 | 88.64 | 89.77 | 77.63 | 65.77 | 65.91 | 64.08 |
| Qwen3-8B | 74.38 | 77.28 | 72.62 | 69.87 | 76.22 | 72.69 | 70.57 | 71.77 | 27.03 | 11.79 | 25.83 | 19.83 |
| GlobMed-Qwen3-8B | 88.99 | 90.26 | 89.77 | 90.05 | 88.92 | 88.36 | 88.85 | 88.57 | 79.39 | 64.78 | 69.51 | 68.38 |

**Supplementary Tab. 138: Zero-Shot performance comparison across 12 languages on MedNLI (Run 4).**

| LLMs | Chinese | English | French | German | Japanese | Korean | Portuguese | Spanish | Swahili | Wolof | Yoruba | Zulu |
|---|---|---|---|---|---|---|---|---|---|---|---|---|
| MedGemma-4B | 58.36 | 72.97 | 59.70 | 60.83 | 58.72 | 58.01 | 60.41 | 62.95 | 43.05 | 15.24 | 32.11 | 30.70 |
| GlobMed-MedGemma-4B | 86.59 | 88.64 | 88.07 | 86.38 | 85.82 | 85.18 | 87.30 | 87.86 | 82.36 | 58.22 | 65.63 | 69.87 |
| Qwen3-1.7B | 68.53 | 75.23 | 66.97 | 62.60 | 54.34 | 49.75 | 65.98 | 63.37 | 25.55 | 24.06 | 29.43 | 32.53 |
| GlobMed-Qwen3-1.7B | 87.51 | 88.07 | 86.73 | 86.73 | 85.67 | 84.83 | 86.17 | 86.87 | 70.15 | 60.90 | 62.60 | 57.09 |
| Qwen3-4B | 76.92 | 81.65 | 68.81 | 56.88 | 71.91 | 72.41 | 74.52 | 73.89 | 19.27 | 18.77 | 11.15 | 13.90 |
| GlobMed-Qwen3-4B | 89.34 | 89.77 | 88.64 | 89.48 | 88.71 | 86.94 | 89.06 | 88.57 | 76.36 | 64.50 | 63.80 | 63.87 |
| Qwen3-8B | 75.16 | 77.70 | 71.63 | 70.01 | 76.78 | 72.34 | 71.42 | 71.21 | 26.96 | 13.76 | 24.91 | 20.04 |
| GlobMed-Qwen3-8B | 89.56 | 90.61 | 89.63 | 89.34 | 88.64 | 87.44 | 89.41 | 88.21 | 78.76 | 63.51 | 69.51 | 68.67 |

**Supplementary Tab. 139: Zero-Shot performance comparison across 12 languages on MedNLI (Run 5).**

| LLMs | Chinese | English | French | German | Japanese | Korean | Portuguese | Spanish | Swahili | Wolof | Yoruba | Zulu | Overall |
|---|---|---|---|---|---|---|---|---|---|---|---|---|---|
| MedGemma-4B | $55.00_{\pm0.66}$ | $69.39_{\pm0.50}$ | $63.51_{\pm0.68}$ | $61.79_{\pm0.91}$ | $55.02_{\pm0.42}$ | $50.91_{\pm1.44}$ | $62.94_{\pm0.47}$ | $62.57_{\pm0.61}$ | $45.48_{\pm0.32}$ | $20.66_{\pm0.57}$ | $21.98_{\pm0.68}$ | $30.95_{\pm0.75}$ | $50.02_{\pm0.37}$ |
| GlobMed-MedGemma-4B | $49.97_{\pm0.31}$ | $61.40_{\pm0.42}$ | $56.44_{\pm0.42}$ | $55.65_{\pm0.56}$ | $50.88_{\pm0.59}$ | $49.03_{\pm0.77}$ | $55.73_{\pm0.49}$ | $43.57_{\pm0.97}$ | $44.36_{\pm1.04}$ | $36.63_{\pm0.60}$ | $36.69_{\pm0.65}$ | $40.82_{\pm0.45}$ | $48.43_{\pm0.17}$ |
| Qwen3-1.7B | $57.75_{\pm0.84}$ | $65.09_{\pm0.52}$ | $55.49_{\pm0.62}$ | $52.98_{\pm0.72}$ | $45.46_{\pm0.65}$ | $42.18_{\pm0.44}$ | $52.99_{\pm0.51}$ | $52.82_{\pm0.50}$ | $23.75_{\pm0.59}$ | $22.52_{\pm0.38}$ | $26.30_{\pm0.85}$ | $23.17_{\pm0.88}$ | $43.38_{\pm0.26}$ |
| GlobMed-Qwen3-1.7B | $58.76_{\pm0.29}$ | $63.54_{\pm0.66}$ | $57.04_{\pm0.31}$ | $56.68_{\pm0.51}$ | $50.95_{\pm0.96}$ | $47.69_{\pm0.28}$ | $56.63_{\pm0.31}$ | $49.56_{\pm0.34}$ | $34.93_{\pm0.58}$ | $36.37_{\pm0.31}$ | $34.45_{\pm0.58}$ | $34.96_{\pm0.95}$ | $48.46_{\pm0.22}$ |
| Qwen3-4B | $72.91_{\pm0.56}$ | $76.47_{\pm0.37}$ | $70.40_{\pm0.15}$ | $70.43_{\pm0.35}$ | $66.57_{\pm0.97}$ | $61.10_{\pm1.07}$ | $70.00_{\pm0.54}$ | $69.67_{\pm0.39}$ | $19.33_{\pm0.91}$ | $14.16_{\pm0.41}$ | $16.25_{\pm0.53}$ | $9.66_{\pm0.83}$ | $51.41_{\pm0.17}$ |
| GlobMed-Qwen3-4B | $68.88_{\pm0.26}$ | $71.71_{\pm0.20}$ | $67.76_{\pm0.34}$ | $68.38_{\pm0.38}$ | $65.15_{\pm0.19}$ | $60.71_{\pm0.43}$ | $65.88_{\pm0.32}$ | $56.43_{\pm0.17}$ | $43.52_{\pm0.74}$ | $41.70_{\pm0.83}$ | $39.76_{\pm0.17}$ | $39.38_{\pm0.54}$ | $57.44_{\pm0.11}$ |
| Qwen3-8B | $76.95_{\pm0.56}$ | $80.75_{\pm0.42}$ | $75.54_{\pm0.11}$ | $76.37_{\pm0.49}$ | $72.29_{\pm0.25}$ | $67.65_{\pm0.70}$ | $75.36_{\pm0.67}$ | $75.22_{\pm0.37}$ | $31.28_{\pm0.81}$ | $10.53_{\pm0.50}$ | $22.73_{\pm0.61}$ | $11.42_{\pm0.82}$ | $56.34_{\pm0.18}$ |
| GlobMed-Qwen3-8B | $74.65_{\pm0.26}$ | $77.29_{\pm0.34}$ | $73.67_{\pm0.46}$ | $73.90_{\pm0.37}$ | $71.27_{\pm0.25}$ | $69.19_{\pm0.40}$ | $74.02_{\pm0.21}$ | $64.18_{\pm0.39}$ | $49.60_{\pm0.86}$ | $45.79_{\pm0.22}$ | $43.21_{\pm0.65}$ | $44.53_{\pm0.60}$ | $63.44_{\pm0.14}$ |

**Supplementary Tab. 140: Performance comparison across 12 languages on HeadQA.**

| LLMs | Chinese | English | French | German | Japanese | Korean | Portuguese | Spanish | Swahili | Wolof | Yoruba | Zulu |
|---|---|---|---|---|---|---|---|---|---|---|---|---|
| MedGemma-4B | 55.23 | 69.57 | 63.71 | 61.00 | 54.69 | 50.36 | 63.17 | 62.67 | 45.18 | 20.20 | 21.46 | 29.94 |
| GlobMed-MedGemma-4B | 50.05 | 61.41 | 56.85 | 55.95 | 50.63 | 49.86 | 55.59 | 43.42 | 43.91 | 36.52 | 36.88 | 40.67 |
| Qwen3-1.7B | 57.39 | 65.15 | 54.73 | 52.21 | 44.72 | 42.06 | 53.11 | 52.98 | 23.62 | 22.45 | 26.38 | 22.09 |
| GlobMed-Qwen3-1.7B | 58.88 | 63.26 | 56.63 | 56.76 | 49.91 | 47.84 | 56.31 | 49.46 | 35.48 | 36.74 | 35.08 | 35.17 |
| Qwen3-4B | 72.72 | 76.56 | 70.24 | 70.65 | 67.45 | 62.53 | 69.48 | 69.16 | 18.94 | 14.52 | 16.14 | 10.37 |
| GlobMed-Qwen3-4B | 68.49 | 71.73 | 67.45 | 68.26 | 64.92 | 61.09 | 66.19 | 56.49 | 43.42 | 42.61 | 39.86 | 39.77 |
| Qwen3-8B | 76.92 | 80.39 | 75.52 | 77.19 | 72.27 | 67.40 | 75.11 | 74.80 | 32.42 | 11.00 | 22.99 | 11.18 |
| GlobMed-Qwen3-8B | 74.62 | 77.01 | 73.04 | 73.44 | 71.28 | 69.12 | 74.39 | 64.52 | 50.86 | 46.08 | 42.52 | 44.86 |

**Supplementary Tab. 141: Zero-Shot performance comparison across 12 languages on HeadQA (Run 1).**

| LLMs | Chinese | English | French | German | Japanese | Korean | Portuguese | Spanish | Swahili | Wolof | Yoruba | Zulu |
|---|---|---|---|---|---|---|---|---|---|---|---|---|
| MedGemma-4B | 54.55 | 68.53 | 63.35 | 61.27 | 54.46 | 49.28 | 62.85 | 62.35 | 45.49 | 20.20 | 21.06 | 31.51 |
| GlobMed-MedGemma-4B | 50.14 | 61.05 | 56.13 | 55.37 | 51.40 | 49.64 | 55.59 | 44.32 | 46.17 | 36.38 | 36.79 | 41.25 |
| Qwen3-1.7B | 57.57 | 64.74 | 55.91 | 53.74 | 46.39 | 42.92 | 53.34 | 53.61 | 22.77 | 23.13 | 26.96 | 23.99 |
| GlobMed-Qwen3-1.7B | 58.39 | 62.62 | 56.99 | 56.76 | 51.53 | 47.20 | 57.12 | 49.32 | 34.94 | 36.34 | 33.63 | 34.72 |
| Qwen3-4B | 73.31 | 76.83 | 70.65 | 70.47 | 66.46 | 61.77 | 70.65 | 69.57 | 19.88 | 13.53 | 15.55 | 10.14 |
| GlobMed-Qwen3-4B | 68.76 | 71.60 | 67.63 | 68.76 | 65.33 | 60.60 | 65.69 | 56.54 | 44.45 | 42.56 | 39.72 | 39.40 |
| Qwen3-8B | 77.41 | 81.02 | 75.70 | 76.38 | 72.05 | 68.39 | 75.02 | 74.84 | 31.83 | 10.37 | 23.62 | 12.04 |
| GlobMed-Qwen3-8B | 74.53 | 77.14 | 73.90 | 74.21 | 71.69 | 69.34 | 73.90 | 64.56 | 49.55 | 45.94 | 42.83 | 43.91 |

**Supplementary Tab. 142: Zero-Shot performance comparison across 12 languages on HeadQA (Run 2).**

| LLMs | Chinese | English | French | German | Japanese | Korean | Portuguese | Spanish | Swahili | Wolof | Yoruba | Zulu |
|---|---|---|---|---|---|---|---|---|---|---|---|---|
| MedGemma-4B | 55.68 | 69.66 | 64.07 | 63.03 | 55.32 | 51.98 | 62.22 | 63.07 | 45.94 | 21.19 | 22.32 | 31.83 |
| GlobMed-MedGemma-4B | 50.27 | 60.96 | 56.45 | 56.45 | 50.41 | 48.11 | 55.23 | 44.72 | 44.27 | 36.25 | 37.20 | 40.31 |
| Qwen3-1.7B | 56.76 | 64.70 | 55.00 | 53.02 | 45.81 | 42.06 | 53.43 | 52.66 | 24.12 | 22.09 | 26.96 | 22.81 |
| GlobMed-Qwen3-1.7B | 59.06 | 64.43 | 57.48 | 57.17 | 52.16 | 47.75 | 56.67 | 49.23 | 34.45 | 36.56 | 34.58 | 36.34 |
| Qwen3-4B | 72.45 | 76.78 | 70.42 | 69.84 | 66.41 | 60.78 | 69.66 | 69.88 | 20.56 | 14.47 | 16.01 | 9.24 |
| GlobMed-Qwen3-4B | 69.03 | 71.91 | 68.03 | 67.99 | 65.33 | 60.87 | 66.05 | 56.54 | 43.01 | 41.16 | 39.99 | 38.64 |
| Qwen3-8B | 77.50 | 80.34 | 75.38 | 75.92 | 72.05 | 68.39 | 74.57 | 75.47 | 30.88 | 10.28 | 22.72 | 12.08 |
| GlobMed-Qwen3-8B | 74.44 | 77.05 | 73.44 | 74.26 | 71.01 | 68.53 | 73.99 | 64.25 | 48.65 | 45.67 | 43.10 | 44.14 |

**Supplementary Tab. 143: Zero-Shot performance comparison across 12 languages on HeadQA (Run 3).**

| LLMs | Chinese | English | French | German | Japanese | Korean | Portuguese | Spanish | Swahili | Wolof | Yoruba | Zulu |
|---|---|---|---|---|---|---|---|---|---|---|---|---|
| **MedGemma-4B** | 55.46 | 69.79 | 64.02 | 61.18 | 55.37 | 52.80 | 63.48 | 61.63 | 45.18 | 20.33 | 22.45 | 30.88 |
| **GlobMed-MedGemma-4B** | 49.95 | 61.99 | 56.85 | 55.00 | 50.36 | 49.14 | 55.68 | 43.10 | 43.60 | 36.29 | 35.57 | 41.34 |
| **Qwen3-1.7B** | 59.02 | 65.96 | 55.59 | 53.65 | 45.13 | 41.75 | 52.93 | 52.30 | 24.17 | 22.41 | 24.89 | 24.17 |
| **GlobMed-Qwen3-1.7B** | 58.93 | 63.62 | 56.99 | 55.82 | 51.08 | 47.84 | 56.58 | 50.09 | 35.53 | 36.29 | 34.13 | 34.90 |
| **Qwen3-4B** | 72.41 | 76.01 | 70.38 | 70.74 | 65.10 | 60.69 | 69.70 | 69.52 | 18.21 | 13.98 | 16.86 | 10.14 |
| **GlobMed-Qwen3-4B** | 68.98 | 71.42 | 68.21 | 68.80 | 65.15 | 60.01 | 65.42 | 56.13 | 42.65 | 41.39 | 39.63 | 39.99 |
| **Qwen3-8B** | 76.83 | 80.66 | 75.56 | 76.19 | 72.54 | 66.86 | 76.10 | 75.47 | 30.48 | 9.92 | 22.14 | 10.10 |
| **GlobMed-Qwen3-8B** | 74.57 | 77.41 | 74.26 | 73.99 | 71.15 | 69.43 | 73.85 | 63.80 | 49.91 | 45.72 | 43.37 | 44.36 |

**Supplementary Tab. 144: Zero-Shot performance comparison across 12 languages on HeadQA (Run 4).**

| LLMs | Chinese | English | French | German | Japanese | Korean | Portuguese | Spanish | Swahili | Wolof | Yoruba | Zulu |
|---|---|---|---|---|---|---|---|---|---|---|---|---|
| **MedGemma-4B** | 54.10 | 69.39 | 62.40 | 62.49 | 55.28 | 50.14 | 62.98 | 63.12 | 45.63 | 21.37 | 22.59 | 30.57 |
| **GlobMed-MedGemma-4B** | 49.46 | 61.59 | 55.91 | 55.50 | 51.62 | 48.38 | 56.54 | 42.29 | 43.87 | 37.69 | 37.02 | 40.53 |
| **Qwen3-1.7B** | 58.03 | 64.88 | 56.22 | 52.30 | 45.27 | 42.11 | 52.16 | 52.57 | 24.08 | 22.54 | 26.33 | 22.77 |
| **GlobMed-Qwen3-1.7B** | 58.52 | 63.75 | 57.12 | 56.90 | 50.05 | 47.84 | 56.45 | 49.68 | 34.27 | 35.93 | 34.85 | 33.68 |
| **Qwen3-4B** | 73.67 | 76.15 | 70.33 | 70.47 | 67.45 | 59.74 | 70.51 | 70.20 | 19.07 | 14.29 | 16.68 | 8.39 |
| **GlobMed-Qwen3-4B** | 69.16 | 71.87 | 67.49 | 68.08 | 65.01 | 61.00 | 66.05 | 56.45 | 44.09 | 40.80 | 39.59 | 39.09 |
| **Qwen3-8B** | 76.10 | 81.33 | 75.56 | 76.15 | 72.54 | 67.22 | 76.01 | 75.52 | 30.79 | 11.09 | 22.18 | 11.68 |
| **GlobMed-Qwen3-8B** | 75.11 | 77.82 | 73.72 | 73.58 | 71.24 | 69.52 | 73.99 | 63.75 | 49.01 | 45.54 | 44.23 | 45.40 |

**Supplementary Tab. 145: Zero-Shot performance comparison across 12 languages on HeadQA (Run 5).**

| LLMs | Chinese | English | French | German | Japanese | Korean | Portuguese | Spanish | Swahili | Wolof | Yoruba | Zulu | Overall |
|---|---|---|---|---|---|---|---|---|---|---|---|---|---|
| **MedGemma-4B** | $48.00_{\pm 1.26}$ | $66.56_{\pm 1.82}$ | $55.68_{\pm 1.66}$ | $60.64_{\pm 2.22}$ | $45.12_{\pm 3.08}$ | $41.44_{\pm 2.07}$ | $54.56_{\pm 1.64}$ | $54.24_{\pm 4.06}$ | $41.60_{\pm 2.88}$ | $17.12_{\pm 1.21}$ | $21.76_{\pm 2.49}$ | $30.56_{\pm 1.82}$ | $44.77_{\pm 0.50}$ |
| **GlobMed-MedGemma-4B** | $67.52_{\pm 2.57}$ | $71.68_{\pm 2.37}$ | $66.24_{\pm 3.60}$ | $64.32_{\pm 0.91}$ | $65.12_{\pm 0.91}$ | $63.36_{\pm 2.68}$ | $64.64_{\pm 2.15}$ | $61.28_{\pm 2.69}$ | $56.64_{\pm 2.74}$ | $49.92_{\pm 2.37}$ | $49.28_{\pm 1.66}$ | $52.16_{\pm 1.43}$ | $61.01_{\pm 0.47}$ |
| **Qwen3-1.7B** | $49.28_{\pm 3.60}$ | $51.36_{\pm 1.43}$ | $44.00_{\pm 2.94}$ | $41.28_{\pm 2.30}$ | $35.52_{\pm 2.50}$ | $33.44_{\pm 1.73}$ | $41.44_{\pm 2.07}$ | $41.92_{\pm 3.94}$ | $15.84_{\pm 1.31}$ | $22.40_{\pm 2.83}$ | $25.76_{\pm 2.96}$ | $21.60_{\pm 2.88}$ | $35.32_{\pm 0.50}$ |
| **GlobMed-Qwen3-1.7B** | $64.48_{\pm 1.84}$ | $65.92_{\pm 0.91}$ | $64.32_{\pm 1.34}$ | $61.92_{\pm 1.84}$ | $61.76_{\pm 2.85}$ | $51.20_{\pm 2.04}$ | $64.16_{\pm 1.31}$ | $65.76_{\pm 0.36}$ | $36.00_{\pm 2.88}$ | $40.64_{\pm 2.49}$ | $42.40_{\pm 3.84}$ | $39.84_{\pm 2.07}$ | $54.87_{\pm 0.91}$ |
| **Qwen3-4B** | $61.92_{\pm 1.93}$ | $66.88_{\pm 1.21}$ | $56.16_{\pm 4.21}$ | $61.28_{\pm 2.50}$ | $53.12_{\pm 1.66}$ | $47.36_{\pm 2.29}$ | $56.00_{\pm 3.44}$ | $56.96_{\pm 2.15}$ | $17.28_{\pm 2.01}$ | $15.52_{\pm 2.01}$ | $16.16_{\pm 3.81}$ | $10.08_{\pm 1.56}$ | $43.23_{\pm 0.87}$ |
| **GlobMed-Qwen3-4B** | $69.12_{\pm 1.45}$ | $72.80_{\pm 2.33}$ | $69.60_{\pm 0.57}$ | $69.76_{\pm 2.29}$ | $66.24_{\pm 2.22}$ | $64.00_{\pm 1.88}$ | $69.60_{\pm 0.80}$ | $69.12_{\pm 1.66}$ | $50.88_{\pm 2.57}$ | $48.48_{\pm 0.72}$ | $44.16_{\pm 2.79}$ | $48.80_{\pm 3.96}$ | $61.88_{\pm 0.64}$ |
| **Qwen3-8B** | $63.68_{\pm 2.37}$ | $70.24_{\pm 2.22}$ | $58.88_{\pm 4.85}$ | $63.36_{\pm 3.89}$ | $58.56_{\pm 1.19}$ | $53.44_{\pm 3.41}$ | $61.44_{\pm 2.74}$ | $60.32_{\pm 3.18}$ | $16.96_{\pm 3.37}$ | $11.52_{\pm 2.57}$ | $22.24_{\pm 1.54}$ | $9.12_{\pm 1.45}$ | $45.81_{\pm 1.01}$ |
| **GlobMed-Qwen3-8B** | $79.52_{\pm 1.21}$ | $77.60_{\pm 0.98}$ | $73.92_{\pm 0.44}$ | $76.48_{\pm 0.72}$ | $76.16_{\pm 1.82}$ | $73.44_{\pm 1.31}$ | $74.40_{\pm 2.33}$ | $75.52_{\pm 1.66}$ | $57.44_{\pm 2.96}$ | $52.96_{\pm 2.22}$ | $47.04_{\pm 1.73}$ | $58.24_{\pm 4.02}$ | $68.56_{\pm 0.39}$ |

**Supplementary Tab. 146: Performance comparison across 12 languages on MedExpQA.**

| LLMs | Chinese | English | French | German | Japanese | Korean | Portuguese | Spanish | Swahili | Wolof | Yoruba | Zulu |
|---|---|---|---|---|---|---|---|---|---|---|---|---|
| **MedGemma-4B** | 47.20 | 65.60 | 56.00 | 60.00 | 48.00 | 41.60 | 52.80 | 55.20 | 39.20 | 18.40 | 24.00 | 31.20 |
| **GlobMed-MedGemma-4B** | 67.20 | 75.20 | 63.20 | 65.60 | 64.80 | 61.60 | 63.20 | 58.40 | 54.40 | 48.80 | 47.20 | 53.60 |
| **Qwen3-1.7B** | 52.00 | 52.80 | 41.60 | 43.20 | 36.00 | 33.60 | 43.20 | 43.20 | 13.60 | 23.20 | 28.00 | 21.60 |
| **GlobMed-Qwen3-1.7B** | 64.80 | 64.80 | 64.80 | 60.00 | 62.40 | 53.60 | 63.20 | 65.60 | 40.00 | 44.80 | 41.60 | 39.20 |
| **Qwen3-4B** | 60.80 | 67.20 | 53.60 | 60.80 | 52.00 | 44.80 | 60.80 | 58.40 | 14.40 | 12.80 | 18.40 | 12.00 |
| **GlobMed-Qwen3-4B** | 69.60 | 74.40 | 69.60 | 69.60 | 69.60 | 62.40 | 70.40 | 68.00 | 49.60 | 48.80 | 45.60 | 55.20 |
| **Qwen3-8B** | 61.60 | 71.20 | 60.80 | 56.80 | 56.80 | 56.80 | 61.60 | 59.20 | 16.80 | 8.00 | 20.00 | 9.60 |
| **GlobMed-Qwen3-8B** | 80.80 | 77.60 | 73.60 | 76.80 | 74.40 | 73.60 | 76.00 | 76.80 | 57.60 | 56.00 | 44.00 | 54.40 |

**Supplementary Tab. 147: Zero-Shot performance comparison across 12 languages on MedExpQA (Run 1).**

| LLMs | Chinese | English | French | German | Japanese | Korean | Portuguese | Spanish | Swahili | Wolof | Yoruba | Zulu |
|---|---|---|---|---|---|---|---|---|---|---|---|---|
| **MedGemma-4B** | 48.80 | 64.00 | 58.40 | 58.40 | 44.00 | 42.40 | 56.00 | 57.60 | 42.40 | 16.80 | 24.00 | 32.80 |
| **GlobMed-MedGemma-4B** | 71.20 | 69.60 | 68.80 | 64.80 | 66.40 | 63.20 | 66.40 | 58.40 | 60.00 | 46.40 | 48.00 | 51.20 |
| **Qwen3-1.7B** | 51.20 | 49.60 | 42.40 | 43.20 | 39.20 | 31.20 | 44.00 | 35.20 | 16.80 | 18.40 | 26.40 | 17.60 |
| **GlobMed-Qwen3-1.7B** | 62.40 | 65.60 | 63.20 | 60.00 | 56.80 | 51.20 | 62.40 | 65.60 | 36.00 | 40.80 | 38.40 | 40.00 |
| **Qwen3-4B** | 64.00 | 67.20 | 56.80 | 60.80 | 55.20 | 50.40 | 53.60 | 56.00 | 19.20 | 15.20 | 21.60 | 8.00 |
| **GlobMed-Qwen3-4B** | 71.20 | 73.60 | 70.40 | 72.80 | 64.80 | 63.20 | 70.40 | 70.40 | 51.20 | 48.80 | 44.00 | 46.40 |
| **Qwen3-8B** | 61.60 | 70.40 | 51.20 | 64.00 | 59.20 | 52.80 | 57.60 | 57.60 | 20.80 | 10.40 | 21.60 | 8.00 |
| **GlobMed-Qwen3-8B** | 78.40 | 77.60 | 73.60 | 76.80 | 77.60 | 74.40 | 76.00 | 75.20 | 53.60 | 52.00 | 47.20 | 62.40 |

**Supplementary Tab. 148: Zero-Shot performance comparison across 12 languages on MedExpQA (Run 2).**

| LLMs | Chinese | English | French | German | Japanese | Korean | Portuguese | Spanish | Swahili | Wolof | Yoruba | Zulu |
|---|---|---|---|---|---|---|---|---|---|---|---|---|
| **MedGemma-4B** | 46.40 | 67.20 | 54.40 | 59.20 | 48.80 | 40.80 | 56.00 | 56.00 | 38.40 | 17.60 | 20.00 | 28.00 |
| **GlobMed-MedGemma-4B** | 68.00 | 71.20 | 61.60 | 64.00 | 65.60 | 62.40 | 65.60 | 64.00 | 54.40 | 50.40 | 49.60 | 53.60 |
| **Qwen3-1.7B** | 43.20 | 52.80 | 48.80 | 37.60 | 32.80 | 36.00 | 40.00 | 43.20 | 16.00 | 25.60 | 28.80 | 20.80 |
| **GlobMed-Qwen3-1.7B** | 67.20 | 67.20 | 64.00 | 62.40 | 64.00 | 49.60 | 64.80 | 66.40 | 35.20 | 39.20 | 46.40 | 37.60 |
| **Qwen3-4B** | 60.80 | 68.00 | 52.80 | 63.20 | 54.40 | 45.60 | 56.00 | 58.40 | 18.40 | 16.00 | 14.40 | 9.60 |
| **GlobMed-Qwen3-4B** | 68.80 | 68.80 | 68.80 | 67.20 | 67.20 | 66.40 | 68.80 | 71.20 | 48.80 | 47.20 | 42.40 | 48.80 |
| **Qwen3-8B** | 67.20 | 72.00 | 57.60 | 64.80 | 60.00 | 48.80 | 64.00 | 57.60 | 16.00 | 14.40 | 22.40 | 7.20 |
| **GlobMed-Qwen3-8B** | 79.20 | 76.80 | 74.40 | 75.20 | 78.40 | 73.60 | 74.40 | 76.80 | 56.00 | 54.40 | 48.00 | 57.60 |

**Supplementary Tab. 149: Zero-Shot performance comparison across 12 languages on MedExpQA (Run 3).**

| LLMs | Chinese | English | French | German | Japanese | Korean | Portuguese | Spanish | Swahili | Wolof | Yoruba | Zulu |
|---|---|---|---|---|---|---|---|---|---|---|---|---|
| **MedGemma-4B** | 49.60 | 68.80 | 54.40 | 64.00 | 42.40 | 38.40 | 52.80 | 47.20 | 45.60 | 15.20 | 22.40 | 29.60 |
| **GlobMed-MedGemma-4B** | 64.00 | 69.60 | 69.60 | 63.20 | 64.00 | 68.00 | 66.40 | 63.20 | 55.20 | 52.00 | 51.20 | 50.40 |
| **Qwen3-1.7B** | 51.20 | 51.20 | 42.40 | 40.80 | 36.00 | 32.80 | 39.20 | 45.60 | 16.80 | 24.00 | 21.60 | 22.40 |
| **GlobMed-Qwen3-1.7B** | 63.20 | 66.40 | 66.40 | 64.00 | 62.40 | 48.80 | 65.60 | 65.60 | 36.80 | 40.00 | 46.40 | 43.20 |
| **Qwen3-4B** | 64.00 | 64.80 | 63.20 | 64.00 | 52.80 | 47.20 | 52.00 | 58.40 | 18.40 | 18.40 | 14.40 | 11.20 |
| **GlobMed-Qwen3-4B** | 68.80 | 72.80 | 69.60 | 71.20 | 65.60 | 62.40 | 69.60 | 67.20 | 49.60 | 48.80 | 48.00 | 48.80 |
| **Qwen3-8B** | 63.20 | 66.40 | 60.80 | 64.00 | 58.40 | 56.80 | 64.00 | 62.40 | 19.20 | 11.20 | 23.20 | 10.40 |
| **GlobMed-Qwen3-8B** | 80.80 | 79.20 | 74.40 | 76.80 | 76.00 | 74.40 | 70.40 | 76.00 | 58.40 | 50.40 | 48.00 | 62.40 |

**Supplementary Tab. 150: Zero-Shot performance comparison across 12 languages on MedExpQA (Run 4).**

| LLMs | Chinese | English | French | German | Japanese | Korean | Portuguese | Spanish | Swahili | Wolof | Yoruba | Zulu |
|---|---|---|---|---|---|---|---|---|---|---|---|---|
| **MedGemma-4B** | 48.00 | 67.20 | 55.20 | 61.60 | 42.40 | 44.00 | 55.20 | 55.20 | 42.40 | 17.60 | 18.40 | 31.20 |
| **GlobMed-MedGemma-4B** | 67.20 | 72.80 | 68.00 | 64.00 | 64.80 | 61.60 | 61.60 | 62.40 | 59.20 | 52.00 | 50.40 | 52.00 |
| **Qwen3-1.7B** | 48.80 | 50.40 | 44.80 | 41.60 | 33.60 | 33.60 | 40.80 | 42.40 | 16.00 | 20.80 | 24.00 | 25.60 |
| **GlobMed-Qwen3-1.7B** | 64.80 | 65.60 | 63.20 | 63.20 | 63.20 | 52.80 | 64.80 | 65.60 | 32.00 | 38.40 | 39.20 | 39.20 |
| **Qwen3-4B** | 60.00 | 67.20 | 54.40 | 57.60 | 51.20 | 48.80 | 57.60 | 53.60 | 16.00 | 15.20 | 12.00 | 9.60 |
| **GlobMed-Qwen3-4B** | 67.20 | 74.40 | 69.60 | 68.00 | 64.00 | 65.60 | 68.80 | 68.80 | 55.20 | 48.80 | 40.80 | 44.80 |
| **Qwen3-8B** | 64.80 | 71.20 | 64.00 | 67.20 | 58.40 | 52.00 | 60.00 | 64.80 | 12.00 | 13.60 | 24.00 | 10.40 |
| **GlobMed-Qwen3-8B** | 78.40 | 76.80 | 73.60 | 76.80 | 74.40 | 71.20 | 75.20 | 72.80 | 61.60 | 52.00 | 48.00 | 54.40 |

**Supplementary Tab. 151: Zero-Shot performance comparison across 12 languages on MedExpQA (Run 5).**

| LLMs | Chinese | English | French | German | Japanese | Korean | Portuguese | Spanish | Swahili | Wolof | Yoruba | Zulu | Overall |
|---|---|---|---|---|---|---|---|---|---|---|---|---|---|
| MedGemma-4B | 47.67$_{\pm0.91}$ | 62.88$_{\pm0.84}$ | 52.94$_{\pm0.97}$ | 54.60$_{\pm0.45}$ | 47.89$_{\pm1.13}$ | 43.57$_{\pm0.43}$ | 53.01$_{\pm0.64}$ | 53.56$_{\pm1.20}$ | 42.81$_{\pm0.68}$ | 16.47$_{\pm0.66}$ | 19.37$_{\pm1.29}$ | 28.80$_{\pm1.16}$ | 43.63 $_{\pm 0.18}$ |
| GlobMed-MedGemma-4B | 46.98$_{\pm0.84}$ | 54.75$_{\pm0.51}$ | 50.90$_{\pm0.50}$ | 50.81$_{\pm0.63}$ | 47.34$_{\pm1.20}$ | 47.46$_{\pm0.67}$ | 50.95$_{\pm0.85}$ | 50.38$_{\pm0.51}$ | 46.20$_{\pm0.63}$ | 39.50$_{\pm0.88}$ | 37.92$_{\pm0.68}$ | 41.51$_{\pm0.98}$ | 47.06 $_{\pm 0.33}$ |
| Qwen3-1.7B | 44.10$_{\pm0.69}$ | 49.14$_{\pm0.88}$ | 40.02$_{\pm0.82}$ | 38.93$_{\pm0.78}$ | 35.88$_{\pm0.52}$ | 34.28$_{\pm1.01}$ | 37.82$_{\pm0.66}$ | 39.43$_{\pm0.93}$ | 26.61$_{\pm0.54}$ | 26.27$_{\pm1.13}$ | 25.64$_{\pm0.52}$ | 25.25$_{\pm0.86}$ | 35.28 $_{\pm 0.17}$ |
| GlobMed-Qwen3-1.7B | 49.00$_{\pm0.39}$ | 52.51$_{\pm0.56}$ | 48.22$_{\pm0.82}$ | 47.78$_{\pm0.81}$ | 44.19$_{\pm0.70}$ | 40.71$_{\pm0.41}$ | 47.60$_{\pm0.33}$ | 47.75$_{\pm0.66}$ | 35.46$_{\pm0.65}$ | 37.86$_{\pm0.59}$ | 35.02$_{\pm0.64}$ | 33.10$_{\pm0.81}$ | 43.27 $_{\pm 0.10}$ |
| Qwen3-4B | 59.54$_{\pm1.08}$ | 64.45$_{\pm0.72}$ | 54.27$_{\pm1.07}$ | 53.20$_{\pm0.84}$ | 53.01$_{\pm0.63}$ | 45.10$_{\pm0.72}$ | 54.52$_{\pm0.48}$ | 55.99$_{\pm1.02}$ | 22.91$_{\pm0.82}$ | 9.96$_{\pm0.69}$ | 16.98$_{\pm1.06}$ | 9.69$_{\pm1.06}$ | 41.64 $_{\pm 0.39}$ |
| GlobMed-Qwen3-4B | 58.52$_{\pm0.92}$ | 62.26$_{\pm0.67}$ | 58.48$_{\pm0.44}$ | 57.04$_{\pm0.21}$ | 54.69$_{\pm0.42}$ | 50.34$_{\pm1.07}$ | 59.56$_{\pm0.26}$ | 59.10$_{\pm0.29}$ | 42.22$_{\pm0.32}$ | 43.76$_{\pm0.54}$ | 39.00$_{\pm0.79}$ | 42.23$_{\pm0.59}$ | 52.27 $_{\pm 0.16}$ |
| Qwen3-8B | 67.41$_{\pm0.62}$ | 71.72$_{\pm0.41}$ | 52.30$_{\pm0.57}$ | 62.66$_{\pm0.67}$ | 60.60$_{\pm0.72}$ | 56.42$_{\pm0.63}$ | 61.60$_{\pm1.07}$ | 61.48$_{\pm1.03}$ | 21.02$_{\pm1.23}$ | 8.83$_{\pm0.39}$ | 21.98$_{\pm0.84}$ | 10.43$_{\pm0.84}$ | 46.37 $_{\pm 0.32}$ |
| GlobMed-Qwen3-8B | 63.94$_{\pm0.41}$ | 66.77$_{\pm0.47}$ | 63.69$_{\pm0.54}$ | 63.88$_{\pm0.41}$ | 61.90$_{\pm0.39}$ | 58.79$_{\pm0.66}$ | 65.23$_{\pm0.52}$ | 63.98$_{\pm0.52}$ | 47.15$_{\pm0.52}$ | 47.02$_{\pm0.77}$ | 43.36$_{\pm1.00}$ | 45.40$_{\pm0.95}$ | 57.59 $_{\pm 0.11}$ |

**Supplementary Tab. 152: Performance comparison across 12 languages on MedQA.**

| LLMs | Chinese | English | French | German | Japanese | Korean | Portuguese | Spanish | Swahili | Wolof | Yoruba | Zulu |
|---|---|---|---|---|---|---|---|---|---|---|---|---|
| MedGemma-4B | 46.27 | 62.53 | 52.47 | 54.60 | 48.47 | 43.68 | 52.79 | 54.67 | 41.79 | 17.44 | 20.35 | 27.97 |
| GlobMed-MedGemma-4B | 46.03 | 54.20 | 50.35 | 50.04 | 46.90 | 46.90 | 50.59 | 51.06 | 46.11 | 38.96 | 37.78 | 40.06 |
| Qwen3-1.7B | 43.28 | 49.41 | 40.77 | 38.41 | 35.59 | 35.43 | 38.49 | 38.02 | 26.32 | 26.63 | 26.08 | 25.45 |
| GlobMed-Qwen3-1.7B | 48.78 | 52.71 | 48.70 | 48.63 | 45.17 | 40.46 | 47.29 | 47.84 | 35.98 | 38.18 | 34.49 | 32.52 |
| Qwen3-4B | 59.54 | 64.73 | 54.75 | 52.24 | 52.24 | 44.07 | 54.12 | 56.95 | 22.86 | 10.45 | 16.03 | 8.64 |
| GlobMed-Qwen3-4B | 58.52 | 63.32 | 58.29 | 57.34 | 54.52 | 49.10 | 59.94 | 59.47 | 41.79 | 43.68 | 38.57 | 42.26 |
| Qwen3-8B | 67.16 | 72.03 | 52.71 | 62.22 | 61.35 | 57.11 | 61.27 | 62.92 | 22.15 | 9.27 | 21.45 | 11.55 |
| GlobMed-Qwen3-8B | 63.32 | 66.77 | 63.32 | 64.10 | 61.51 | 58.60 | 64.96 | 64.57 | 47.29 | 47.21 | 43.13 | 45.48 |

**Supplementary Tab. 153: Zero-Shot performance comparison across 12 languages on MedQA (Run 1).**

| LLMs | Chinese | English | French | German | Japanese | Korean | Portuguese | Spanish | Swahili | Wolof | Yoruba | Zulu |
|---|---|---|---|---|---|---|---|---|---|---|---|---|
| MedGemma-4B | 48.15 | 61.82 | 53.97 | 54.28 | 46.11 | 42.89 | 54.05 | 53.42 | 43.21 | 15.63 | 18.54 | 28.52 |
| GlobMed-MedGemma-4B | 46.11 | 55.54 | 50.82 | 51.45 | 45.95 | 46.58 | 49.73 | 50.75 | 45.40 | 40.77 | 38.88 | 40.93 |
| Qwen3-1.7B | 43.52 | 47.68 | 40.77 | 38.49 | 36.29 | 34.88 | 38.57 | 38.96 | 26.71 | 25.61 | 25.77 | 25.45 |
| GlobMed-Qwen3-1.7B | 49.57 | 52.55 | 47.45 | 48.23 | 43.75 | 40.77 | 47.60 | 46.74 | 35.66 | 38.02 | 35.82 | 33.39 |
| Qwen3-4B | 60.72 | 64.49 | 54.20 | 53.34 | 53.73 | 45.09 | 53.97 | 56.72 | 21.60 | 9.66 | 16.10 | 8.96 |
| GlobMed-Qwen3-4B | 58.37 | 62.45 | 58.60 | 56.87 | 54.05 | 50.51 | 59.23 | 59.15 | 42.11 | 43.52 | 38.18 | 41.24 |
| Qwen3-8B | 68.42 | 71.96 | 52.95 | 63.79 | 60.72 | 55.70 | 60.57 | 60.80 | 20.90 | 8.25 | 22.78 | 10.60 |
| GlobMed-Qwen3-8B | 63.86 | 67.16 | 64.49 | 63.71 | 61.82 | 57.89 | 65.20 | 63.71 | 47.92 | 46.27 | 43.44 | 44.70 |

**Supplementary Tab. 154: Zero-Shot performance comparison across 12 languages on MedQA (Run 2).**

| LLMs | Chinese | English | French | German | Japanese | Korean | Portuguese | Spanish | Swahili | Wolof | Yoruba | Zulu |
|---|---|---|---|---|---|---|---|---|---|---|---|---|
| MedGemma-4B | 48.55 | 63.16 | 51.53 | 54.36 | 49.02 | 43.52 | 52.32 | 53.26 | 43.60 | 16.58 | 17.52 | 30.40 |
| GlobMed-MedGemma-4B | 47.53 | 54.83 | 51.45 | 50.43 | 46.98 | 48.00 | 51.61 | 49.96 | 46.03 | 38.73 | 37.78 | 42.11 |
| Qwen3-1.7B | 44.54 | 50.04 | 39.28 | 38.33 | 35.74 | 34.56 | 37.16 | 39.98 | 25.84 | 25.92 | 25.14 | 24.27 |
| GlobMed-Qwen3-1.7B | 48.78 | 52.79 | 48.78 | 47.37 | 43.36 | 40.53 | 47.45 | 48.00 | 35.74 | 38.33 | 34.25 | 34.09 |
| Qwen3-4B | 60.25 | 65.44 | 55.62 | 54.52 | 52.55 | 45.95 | 55.15 | 56.40 | 23.33 | 10.92 | 17.52 | 9.74 |
| GlobMed-Qwen3-4B | 58.21 | 61.90 | 57.82 | 56.87 | 54.83 | 51.92 | 59.54 | 58.68 | 42.66 | 44.07 | 38.73 | 42.42 |
| Qwen3-8B | 67.48 | 71.01 | 51.61 | 62.45 | 61.12 | 55.85 | 63.24 | 61.90 | 20.74 | 9.11 | 23.02 | 10.29 |
| GlobMed-Qwen3-8B | 64.41 | 67.09 | 63.08 | 63.32 | 62.45 | 59.07 | 65.28 | 64.10 | 47.21 | 46.50 | 45.01 | 44.46 |

**Supplementary Tab. 155: Zero-Shot performance comparison across 12 languages on MedQA (Run 3).**

| LLMs | Chinese | English | French | German | Japanese | Korean | Portuguese | Spanish | Swahili | Wolof | Yoruba | Zulu |
|---|---|---|---|---|---|---|---|---|---|---|---|---|
| **MedGemma-4B** | 48.08 | 62.77 | 53.10 | 54.36 | 48.31 | 43.68 | 52.87 | 54.67 | 42.73 | 16.50 | 19.95 | 29.54 |
| **GlobMed-MedGemma-4B** | 47.45 | 54.44 | 51.37 | 51.45 | 49.18 | 47.92 | 51.85 | 50.27 | 46.35 | 38.96 | 38.18 | 42.26 |
| **Qwen3-1.7B** | 44.93 | 49.18 | 40.22 | 40.14 | 36.53 | 32.91 | 37.47 | 40.22 | 27.02 | 28.04 | 25.06 | 24.59 |
| **GlobMed-Qwen3-1.7B** | 48.63 | 52.95 | 47.21 | 48.08 | 44.54 | 40.38 | 48.15 | 47.60 | 35.59 | 36.84 | 35.35 | 33.46 |
| **Qwen3-4B** | 57.89 | 63.55 | 54.05 | 52.95 | 53.02 | 45.56 | 54.60 | 55.38 | 23.80 | 9.51 | 18.54 | 11.39 |
| **GlobMed-Qwen3-4B** | 57.50 | 62.06 | 58.76 | 57.19 | 54.99 | 49.65 | 59.62 | 58.99 | 42.18 | 44.46 | 39.28 | 42.81 |
| **Qwen3-8B** | 66.77 | 71.88 | 52.40 | 62.14 | 60.25 | 56.48 | 60.88 | 61.51 | 19.17 | 8.72 | 21.29 | 9.19 |
| **GlobMed-Qwen3-8B** | 63.94 | 65.99 | 63.86 | 63.86 | 62.14 | 59.70 | 66.06 | 63.24 | 46.66 | 46.90 | 42.42 | 45.48 |

**Supplementary Tab. 156: Zero-Shot performance comparison across 12 languages on MedQA (Run 4).**

| LLMs | Chinese | English | French | German | Japanese | Korean | Portuguese | Spanish | Swahili | Wolof | Yoruba | Zulu |
|---|---|---|---|---|---|---|---|---|---|---|---|---|
| **MedGemma-4B** | 47.29 | 64.10 | 53.65 | 55.38 | 47.53 | 44.07 | 53.02 | 51.77 | 42.73 | 16.18 | 20.50 | 27.57 |
| **GlobMed-MedGemma-4B** | 47.76 | 54.75 | 50.51 | 50.67 | 47.68 | 47.92 | 50.98 | 49.88 | 47.13 | 40.06 | 37.00 | 42.18 |
| **Qwen3-1.7B** | 44.23 | 49.41 | 39.04 | 39.28 | 35.27 | 33.62 | 37.39 | 39.98 | 27.18 | 25.14 | 26.16 | 26.47 |
| **GlobMed-Qwen3-1.7B** | 49.25 | 51.53 | 48.94 | 46.58 | 44.15 | 41.40 | 47.53 | 48.55 | 34.33 | 37.94 | 35.19 | 32.05 |
| **Qwen3-4B** | 59.31 | 64.02 | 52.71 | 52.95 | 53.50 | 44.85 | 54.75 | 54.52 | 22.94 | 9.27 | 16.73 | 9.74 |
| **GlobMed-Qwen3-4B** | 60.02 | 61.59 | 58.92 | 56.95 | 55.07 | 50.51 | 59.47 | 59.23 | 42.34 | 43.05 | 40.22 | 42.42 |
| **Qwen3-8B** | 67.24 | 71.72 | 51.85 | 62.69 | 59.54 | 56.95 | 62.06 | 60.25 | 22.15 | 8.80 | 21.37 | 10.53 |
| **GlobMed-Qwen3-8B** | 64.18 | 66.85 | 63.71 | 64.41 | 61.59 | 58.68 | 64.65 | 64.26 | 46.66 | 48.23 | 42.81 | 46.90 |

**Supplementary Tab. 157: Zero-Shot performance comparison across 12 languages on MedQA (Run 5).**

| LLMs | Chinese | English | French | German | Japanese | Korean | Portuguese | Spanish | Swahili | Wolof | Yoruba | Zulu | Overall |
|---|---|---|---|---|---|---|---|---|---|---|---|---|---|
| **MedGemma-4B** | $33.48_{\pm1.04}$ | $46.31_{\pm1.95}$ | $42.25_{\pm3.12}$ | $37.22_{\pm1.72}$ | $30.91_{\pm1.75}$ | $29.63_{\pm2.23}$ | $36.04_{\pm1.49}$ | $36.68_{\pm1.29}$ | $23.64_{\pm2.18}$ | $13.05_{\pm2.44}$ | $12.41_{\pm2.40}$ | $16.26_{\pm2.99}$ | 29.82 ± 0.55 |
| **GlobMed-MedGemma-4B** | $32.94_{\pm1.80}$ | $41.07_{\pm1.91}$ | $37.97_{\pm1.25}$ | $39.89_{\pm2.97}$ | $32.73_{\pm1.48}$ | $30.16_{\pm1.68}$ | $36.36_{\pm0.54}$ | $34.22_{\pm1.25}$ | $30.27_{\pm1.63}$ | $20.54_{\pm3.24}$ | $20.96_{\pm3.88}$ | $24.49_{\pm1.33}$ | 31.80 ± 0.56 |
| **Qwen3-1.7B** | $29.95_{\pm0.54}$ | $38.40_{\pm2.49}$ | $28.77_{\pm2.22}$ | $23.32_{\pm2.92}$ | $20.54_{\pm1.04}$ | $21.07_{\pm2.79}$ | $24.28_{\pm1.76}$ | $25.78_{\pm2.09}$ | $14.76_{\pm2.44}$ | $11.44_{\pm2.35}$ | $11.98_{\pm0.30}$ | $13.58_{\pm1.40}$ | 21.99 ± 0.65 |
| **GlobMed-Qwen3-1.7B** | $34.65_{\pm1.16}$ | $39.25_{\pm0.90}$ | $36.79_{\pm1.90}$ | $34.97_{\pm1.39}$ | $26.74_{\pm2.07}$ | $23.21_{\pm2.52}$ | $36.79_{\pm1.16}$ | $32.83_{\pm1.80}$ | $19.36_{\pm1.79}$ | $18.18_{\pm2.81}$ | $16.04_{\pm2.60}$ | $16.26_{\pm2.02}$ | 27.92 ± 0.59 |
| **Qwen3-4B** | $53.16_{\pm1.95}$ | $59.15_{\pm1.76}$ | $50.80_{\pm2.48}$ | $50.59_{\pm1.72}$ | $43.64_{\pm2.22}$ | $37.75_{\pm2.23}$ | $48.34_{\pm1.45}$ | $50.27_{\pm3.49}$ | $12.51_{\pm1.80}$ | $11.23_{\pm0.65}$ | $11.02_{\pm2.29}$ | $10.70_{\pm1.46}$ | 36.60 ± 0.43 |
| **GlobMed-Qwen3-4B** | $50.59_{\pm2.16}$ | $56.04_{\pm0.58}$ | $53.26_{\pm1.80}$ | $54.12_{\pm1.66}$ | $42.03_{\pm2.13}$ | $40.75_{\pm0.70}$ | $54.01_{\pm0.85}$ | $55.93_{\pm0.81}$ | $25.99_{\pm2.29}$ | $26.63_{\pm2.02}$ | $24.71_{\pm1.44}$ | $23.42_{\pm2.43}$ | 42.29 ± 0.39 |
| **Qwen3-8B** | $58.50_{\pm2.16}$ | $66.95_{\pm1.16}$ | $54.65_{\pm0.79}$ | $54.76_{\pm1.71}$ | $52.30_{\pm2.12}$ | $45.67_{\pm2.52}$ | $56.47_{\pm1.84}$ | $57.86_{\pm3.01}$ | $19.68_{\pm2.86}$ | $11.66_{\pm1.38}$ | $13.05_{\pm1.88}$ | $12.09_{\pm3.28}$ | 41.97 ± 0.78 |
| **GlobMed-Qwen3-8B** | $56.90_{\pm2.09}$ | $59.89_{\pm1.26}$ | $56.90_{\pm1.29}$ | $55.40_{\pm1.11}$ | $53.69_{\pm1.23}$ | $50.80_{\pm0.85}$ | $55.19_{\pm1.16}$ | $55.29_{\pm0.81}$ | $25.99_{\pm1.23}$ | $24.60_{\pm2.17}$ | $24.81_{\pm1.95}$ | $22.78_{\pm0.97}$ | 45.19 ± 0.46 |

**Supplementary Tab. 158: Performance comparison across 12 languages on MMLU-Pro.**

| LLMs | Chinese | English | French | German | Japanese | Korean | Portuguese | Spanish | Swahili | Wolof | Yoruba | Zulu |
|---|---|---|---|---|---|---|---|---|---|---|---|---|
| **MedGemma-4B** | 33.16 | 48.13 | 37.43 | 37.43 | 31.55 | 28.34 | 35.29 | 37.43 | 21.93 | 14.44 | 10.16 | 13.90 |
| **GlobMed-MedGemma-4B** | 32.62 | 41.18 | 38.50 | 40.64 | 34.22 | 31.02 | 36.36 | 35.29 | 31.02 | 16.58 | 19.25 | 23.53 |
| **Qwen3-1.7B** | 29.95 | 35.83 | 30.48 | 22.46 | 20.86 | 24.60 | 24.60 | 25.67 | 18.72 | 14.97 | 11.76 | 14.44 |
| **GlobMed-Qwen3-1.7B** | 34.76 | 40.11 | 38.50 | 36.36 | 24.06 | 19.79 | 37.43 | 31.55 | 19.25 | 16.04 | 12.83 | 15.51 |
| **Qwen3-4B** | 54.55 | 59.36 | 54.01 | 49.73 | 42.78 | 37.97 | 49.20 | 51.34 | 11.23 | 11.76 | 13.37 | 9.63 |
| **GlobMed-Qwen3-4B** | 50.27 | 55.08 | 52.41 | 56.15 | 44.92 | 41.18 | 52.94 | 57.22 | 27.27 | 23.53 | 24.06 | 19.25 |
| **Qwen3-8B** | 59.36 | 68.45 | 55.08 | 52.41 | 49.73 | 45.45 | 58.29 | 55.61 | 19.79 | 9.63 | 12.83 | 9.63 |
| **GlobMed-Qwen3-8B** | 58.29 | 61.50 | 58.29 | 54.01 | 54.55 | 50.27 | 55.61 | 55.61 | 26.20 | 21.93 | 26.20 | 24.06 |

**Supplementary Tab. 159: Zero-Shot performance comparison across 12 languages on MMLU-Pro (Run 1).**

| LLMs | Chinese | English | French | German | Japanese | Korean | Portuguese | Spanish | Swahili | Wolof | Yoruba | Zulu |
|---|---|---|---|---|---|---|---|---|---|---|---|---|
| **MedGemma-4B** | 34.76 | 45.45 | 42.78 | 37.97 | 31.02 | 31.55 | 34.76 | 35.29 | 22.46 | 11.23 | 15.51 | 14.44 |
| **GlobMed-MedGemma-4B** | 35.83 | 43.85 | 39.04 | 38.50 | 31.02 | 30.48 | 35.83 | 35.29 | 32.62 | 21.93 | 22.99 | 24.06 |
| **Qwen3-1.7B** | 30.48 | 36.36 | 28.34 | 21.93 | 20.86 | 18.72 | 21.39 | 25.67 | 12.83 | 9.63 | 11.76 | 14.44 |
| **GlobMed-Qwen3-1.7B** | 34.22 | 39.04 | 38.50 | 33.16 | 27.27 | 26.20 | 35.83 | 34.22 | 19.25 | 21.93 | 17.11 | 15.51 |
| **Qwen3-4B** | 55.08 | 60.43 | 52.41 | 51.87 | 40.64 | 39.04 | 49.73 | 44.39 | 11.23 | 11.23 | 8.56 | 12.83 |
| **GlobMed-Qwen3-4B** | 52.94 | 56.15 | 54.01 | 52.41 | 39.57 | 41.18 | 54.55 | 55.08 | 28.34 | 27.81 | 25.67 | 25.13 |
| **Qwen3-8B** | 59.89 | 67.91 | 53.48 | 54.55 | 52.41 | 43.32 | 56.15 | 60.43 | 20.32 | 11.23 | 14.97 | 16.04 |
| **GlobMed-Qwen3-8B** | 58.82 | 60.96 | 55.08 | 56.15 | 54.01 | 49.73 | 55.08 | 56.15 | 24.06 | 25.13 | 22.99 | 22.99 |

**Supplementary Tab. 160: Zero-Shot performance comparison across 12 languages on MMLU-Pro (Run 2).**

| LLMs | Chinese | English | French | German | Japanese | Korean | Portuguese | Spanish | Swahili | Wolof | Yoruba | Zulu |
|---|---|---|---|---|---|---|---|---|---|---|---|---|
| **MedGemma-4B** | 34.22 | 47.59 | 44.92 | 34.22 | 33.16 | 32.09 | 35.29 | 36.36 | 26.74 | 16.58 | 11.23 | 13.90 |
| **GlobMed-MedGemma-4B** | 32.09 | 40.64 | 38.50 | 36.36 | 34.22 | 31.55 | 35.83 | 32.62 | 28.34 | 24.06 | 14.97 | 26.74 |
| **Qwen3-1.7B** | 29.41 | 41.71 | 31.55 | 19.79 | 21.39 | 19.79 | 24.06 | 28.88 | 13.37 | 11.23 | 11.76 | 11.23 |
| **GlobMed-Qwen3-1.7B** | 33.16 | 39.04 | 34.76 | 34.22 | 28.88 | 25.13 | 37.43 | 34.76 | 16.58 | 17.11 | 14.97 | 17.11 |
| **Qwen3-4B** | 51.34 | 56.15 | 50.27 | 48.13 | 43.32 | 35.83 | 49.20 | 50.27 | 15.51 | 11.76 | 10.70 | 9.09 |
| **GlobMed-Qwen3-4B** | 47.59 | 56.15 | 55.61 | 55.61 | 43.32 | 40.64 | 54.01 | 55.61 | 27.27 | 28.88 | 22.46 | 25.13 |
| **Qwen3-8B** | 55.08 | 65.78 | 54.55 | 56.68 | 51.87 | 45.99 | 54.01 | 54.01 | 23.53 | 12.30 | 12.83 | 11.23 |
| **GlobMed-Qwen3-8B** | 53.48 | 58.82 | 57.75 | 56.68 | 55.08 | 50.80 | 56.68 | 55.08 | 27.27 | 27.81 | 26.74 | 22.99 |

**Supplementary Tab. 161: Zero-Shot performance comparison across 12 languages on MMLU-Pro (Run 3).**

| LLMs | Chinese | English | French | German | Japanese | Korean | Portuguese | Spanish | Swahili | Wolof | Yoruba | Zulu |
|---|---|---|---|---|---|---|---|---|---|---|---|---|
| **MedGemma-4B** | 32.09 | 47.06 | 44.92 | 38.50 | 30.48 | 26.74 | 38.50 | 35.83 | 25.13 | 12.30 | 10.70 | 19.79 |
| **GlobMed-MedGemma-4B** | 33.16 | 41.18 | 35.83 | 39.57 | 31.55 | 30.48 | 36.90 | 34.76 | 29.95 | 22.46 | 24.60 | 24.60 |
| **Qwen3-1.7B** | 30.48 | 37.97 | 26.20 | 27.27 | 18.72 | 23.53 | 25.67 | 22.99 | 15.51 | 9.09 | 12.30 | 13.37 |
| **GlobMed-Qwen3-1.7B** | 34.76 | 40.11 | 34.76 | 36.36 | 25.13 | 22.46 | 35.29 | 33.16 | 20.32 | 15.51 | 19.79 | 13.90 |
| **Qwen3-4B** | 54.01 | 59.36 | 47.59 | 50.80 | 46.52 | 40.64 | 46.52 | 51.87 | 11.76 | 10.16 | 9.09 | 10.70 |
| **GlobMed-Qwen3-4B** | 49.73 | 56.68 | 53.48 | 52.94 | 41.71 | 39.57 | 53.48 | 55.61 | 23.53 | 26.20 | 25.67 | 24.06 |
| **Qwen3-8B** | 57.75 | 66.31 | 54.55 | 54.01 | 55.61 | 49.73 | 55.61 | 60.96 | 19.25 | 13.37 | 14.44 | 14.97 |
| **GlobMed-Qwen3-8B** | 56.68 | 58.82 | 56.15 | 55.61 | 52.41 | 51.87 | 55.08 | 55.61 | 25.67 | 23.53 | 25.67 | 21.39 |

**Supplementary Tab. 162: Zero-Shot performance comparison across 12 languages on MMLU-Pro (Run 4).**

| LLMs | Chinese | English | French | German | Japanese | Korean | Portuguese | Spanish | Swahili | Wolof | Yoruba | Zulu |
|---|---|---|---|---|---|---|---|---|---|---|---|---|
| **MedGemma-4B** | 33.16 | 43.32 | 41.18 | 37.97 | 28.34 | 29.41 | 36.36 | 38.50 | 21.93 | 10.70 | 14.44 | 19.25 |
| **GlobMed-MedGemma-4B** | 31.02 | 38.50 | 37.97 | 44.39 | 32.62 | 27.27 | 36.90 | 33.16 | 29.41 | 17.65 | 22.99 | 23.53 |
| **Qwen3-1.7B** | 29.41 | 40.11 | 27.27 | 25.13 | 20.86 | 18.72 | 25.67 | 25.67 | 13.37 | 12.30 | 12.30 | 14.44 |
| **GlobMed-Qwen3-1.7B** | 36.36 | 37.97 | 37.43 | 34.76 | 28.34 | 22.46 | 37.97 | 30.48 | 21.39 | 20.32 | 15.51 | 19.25 |
| **Qwen3-4B** | 50.80 | 60.43 | 49.73 | 52.41 | 44.92 | 35.29 | 47.06 | 53.48 | 12.83 | 11.23 | 13.37 | 11.23 |
| **GlobMed-Qwen3-4B** | 52.41 | 56.15 | 50.80 | 53.48 | 40.64 | 41.18 | 55.08 | 56.15 | 23.53 | 26.74 | 25.67 | 23.53 |
| **Qwen3-8B** | 60.43 | 66.31 | 55.61 | 56.15 | 51.87 | 43.85 | 58.29 | 58.29 | 15.51 | 11.76 | 10.16 | 8.56 |
| **GlobMed-Qwen3-8B** | 57.22 | 59.36 | 57.22 | 54.55 | 52.41 | 51.34 | 53.48 | 54.01 | 26.74 | 24.60 | 22.46 | 22.46 |

**Supplementary Tab. 163: Zero-Shot performance comparison across 12 languages on MMLU-Pro (Run 5).**

## S4. Multiple-Comparison Correction

We applied Benjamini–Hochberg false discovery rate (BH-FDR) correction separately to cross-lingual performance disparities, comparisons between general and medical LLMs, comparisons between non-reasoning and reasoning LLMs, and comparisons between GlobMed-LLMs and their corresponding baseline LLMs, covering 4,992 comparisons in total. Following correction, only 27 estimable comparisons changed from statistically significant to nonsignificant at the 0.05 threshold, while the overall conclusions remained unchanged.